\documentclass[journal]{IEEEtran}
\ifCLASSINFOpdf
\else
\fi

\usepackage{mathtools,booktabs,tabu,graphicx,hyperref,mystyle,epstopdf,url,rotating}

\usepackage{algorithm}
\usepackage{fancyref}
\usepackage[noend]{algpseudocode}

\newcommand{\vertiii}[1]{{\left\vert\kern-0.25ex\left\vert\kern-0.25ex\left\vert #1 
		\right\vert\kern-0.25ex\right\vert\kern-0.25ex\right\vert}}

\def\A{{\mathbf A}}
\def\b{{\mathbf {b}}}

\begin{document}
%
\title{Compressive MRI quantification using convex spatiotemporal priors and deep auto-encoders}
%
%
%
%

\author{Mohammad~Golbabaee,~\IEEEmembership{Member,~IEEE,}~Guido~Buonincontri,~Carolin M.~Pirkl, ~Marion~I.~Menzel,~Bjoern~H.~Menze,~Mike~Davies~and~Pedro A.~G\'omez
\thanks{MG is with the Computer Science department at the University of Bath, UK: (m.golbabaee@bath.ac.uk). 
GB is with Imago7 foundation. MD is with the University of Edinburgh.
MIM is with the GE Healthcare. CMP, BHM and PAG are with the Technical University of Munich. 
}}
\maketitle


\begin{abstract}
We propose a dictionary-matching-free pipeline for multi-parametric quantitative MRI image computing. Our approach has two stages based on compressed sensing reconstruction and deep learned quantitative inference. The reconstruction phase is convex and incorporates efficient spatiotemporal regularisations within an accelerated iterative shrinkage algorithm. This minimises the under-sampling (aliasing) artefacts from aggressively short scan times. The learned quantitative inference phase is purely trained on physical simulations (Bloch equations) that are flexible for producing rich training samples. We propose a deep and compact auto-encoder network with residual blocks in order to embed Bloch manifold projections through multi-scale piecewise affine approximations, and to replace the non-scalable dictionary-matching baseline. 
Tested on a number of datasets we demonstrate effectiveness of the proposed scheme for recovering accurate and consistent quantitative information from novel and aggressively subsampled 2D/3D quantitative MRI acquisition protocols. 
\end{abstract}


%
\IEEEpeerreviewmaketitle

\section{Introduction}
\label{sec:intro}
%
%
%
%
Quantification of the intrinsic NMR characteristics~\cite{toftsmri} has proven powerful for
tissue identification and tracking pathological changes. 
Despite many potentials, standard quantitative MRI (QMRI) approaches have very long acquisition times and for this reason, are not widely applicable in clinical setups. 
Magnetic Resonance Fingerprinting (MRF) has
emerged to overcome this challenge~\cite{MRF}. MRF uses short excitation sequences capable of simultaneously encoding multitudes of NMR properties and further adopts Compressed Sensing (CS) to subsample a tiny fraction of the spatiotemporal k-space information~\cite{FISP, EPIT1T2star,MRF-perfusion,MRF_T1T2diff,lustigCSMRI}. Estimating the underlying 
quantitative  maps therefore becomes a highly ill-posed inverse problem. 

Baseline approaches to solve the MRF inverse problem rely on 
\emph{dictionary matching} (DM), primarily for parameter inference i.e. estimating quantitative maps from back-projected images, or further for promoting temporal-domain priors within model-based MRF reconstructions 
\cite{BLIPsiam}.
 However DM's complexity (storage/runtime) does not scale well to the emerging multi-parametric QMRI applications. 
 \emph{Deep learning} MRF approaches recently emerged to address this issue~\cite{DRONE-MRF,betterreal,deepMRFgeom}. Back-projected images are fed into a compact neural network that temporally
processes voxel-wise MRF signal evolutions, so-called fingerprints, 
and replaces DM for quantitative inference. 
Trained with independently corrupted 
noisy fingerprints, such networks are unable to
correct for dominant spatially-correlated (aliasing) artefacts appearing in heavily
undersampled aquisitions. 
While larger convolutional models~\cite{hoppeDLMRF, balsigerMRFDL, fang2019deep} capture spatiotemporal information to resolve aliasing artefacts, labelled QMRI datasets (i.e. ground truth multi-parametric anatomical maps) that are necessary to train these models particularly in novel applications are scarce and hence place adaption of these models at the risk of overfitted predictions. 
Further, current
approaches along this line build customised de-noisers (de-aliasing) and require expensive re-training by changing
sampling parameters i.e. the forward model.

This work aims to address these shortcomings through a two-stage DM-free pipeline: First,  we take a CS approach to \emph{spatiotemporally process} the k-space data and minimise undersampling artefacts in the reconstructed image time-series, and second we feed the resulted  sequence to a deep and compact 
\emph{auto-encoder} network with residual blocks for per-voxel quantitative inference. 
We cast our spatiotemporally regularised reconstruction as a \emph{convex} optimisation problem which enjoys reproducible global solutions regardless of initialisation and can be implemented with a momentum-accelerated algorithm with fast convergence. 
We further provide geometrical insights to the mechanism behind the proposed deep inference approach.
We show that the network provides a \emph{multi-resolution piecewise affine approximation} to the Bloch response manifold projection.  Rather than memorising a large MRF dictionary, the network hierarchically clusters this manifold through deep layers and learns a compact set of \emph{deep regressing filters} for parameter inference. 
The proposed pipeline is validated on a number of experiments using a novel multi-parametric acquisition sequence for \emph{2D and 3D quantitative brain imaging}. Our approach can flexibly apply and report consistent predictions for different k-space readouts and further outperforms shallow learned inference models related to the Gaussian kernel fitting.


\emph{Paper organisation}: We review related works in section~\ref{sec:SOA}. Section~\ref{sec:model} presents the inverse imaging problem model. Section~\ref{sec:pipeline} presents our reconstruction and quantitive inference pipeline. Section~\ref{sec:partitioninig} presents our geometrical insight to the network's performance for deep quantitative inference. In Section~\ref{sec:expe} we present and discuss our experimental results, and finally we conclude in section~\ref{sec:conclusion}.

\emph{Notations: }
Throughout $\|.\|$ denotes the Euclidean norm of a vector or a matrix,  $\|.\|_{TV}$ denotes the Total Variation (TV) of a 2D or 3D spatial image defined by the sums of its gradient magnitudes~\cite{rof_tv}. Matrix rows and columns are denoted by $X_{(i,.)}$ and $X_i$ respectively.

\section{Related works}
\label{sec:SOA}
Multi-parametric quantification based on fingerprinting, DM and its SVD-compressed (low-rank) variant were proposed in~\cite{MRF, SVDMRF}. Reconstructing image time-series from k-space data was non-iterative and used zero-filling (ZF). 
Inspired by CS, later studies adopted model-based reconstructions to reduce subsampling (aliasing) artefacts 
and to pave the path for aggressively shorter scan times~\cite{BLIPsiam,asslander-ADMMMRF}. 
These methods are based on non-convex optimisation (iterative) without momentum-acceleration, and require DM per iteration in order to promote  temporal-domain Bloch dynamic priors.
While fast search schemes~\cite{MRF-GRM, AIRMRF, coverblip_iop} could partly improve the runtime, DM remains a computational/storage bottleneck for multi-dimensional imaging problems involving multi-parametric dictionaries. 
Further, for some k-space subsampling patterns, including those adopted in our experiments, using only a temporal-domain prior is insufficient to produce artefact-free reconstructions (see e.g.~\cite{AIRMRF, coverblip_iop}).  Spatial domain regularisations 
were integrated into DM~\cite{AIRMRF,simon_lrtv,Gomez2015a, Gomez2016}, however these methods require costly DM per iteration, are nonconvex and without momentum-acceleration.  
DM-free convex reconstructions based on low-rank priors were proposed, some~\cite{zhao-LR-MRF,Eldar-FLOR, song2019hydra} with no spatial regularisations hence prone to artefacts in highly ill-posed problems, some~\cite{Eldar-FLOR, song2019hydra} without temporal dimensionality reduction and long runtimes, and some using patch-based spatiotemporal low-rank regularisation~\cite{bustin2019high,jaubert2020free} but encountering long runtimes due to non-accelerated iterations and per-iteration costly SVD decompositions.
Our work mitigates these issues: we propose an alternative convex formulation for the MRF reconstruction problem (our preliminary results appeared in~\cite{LRTVismrm}). We add spatial TV regularisation for the dimension-reduced image time-series while enforcing temporal-domain priors through a (low-rank) subspace representation of the dictionary instead of DM. This optimisation can be solved with momentum-accelerated iterative algorithms with fast global convergence. 



%
%
%

On the other hand, deep learning MRF approaches recently emerged to address the non-scalability of DM. Many works use non-iterative  baselines~\cite{MRF, SVDMRF} for reconstruction,  
and for quantitative inference they replace DM with a neural network. These methods broadly divide in two camps: the first group learns temporal-domain dynamics from simulating Bloch equations; hence is rich with training data (see e.g.~\cite{DRONE-MRF,betterreal,deepMRFgeom,MRFRNN} 
and also a kernel machine approach for shallow learning~\cite{PERK}). The second group use convolutional layers to also learn spatial domain regularities, see e.g.~\cite{hoppeDLMRF, hoppe2020rinq,balsigerMRFDL, balsiger2019spatially, fang2019deep, fang2019rca}, but they require training on ground truth quantitative anatomical maps that may not be largely available as for the mainstream qualitative MRI. 
Our quantitative inference approach belongs to the first camp. We provide a geometrical interpretation for our deep inference approach. Importantly, we replace ZF by our DM-free spatiotemporally regularised (model-based) reconstruction to remove undersampling artefacts before being fed to the network. This enables aggressively short-time 2D/3D quantitative imaging protocols produce consistent results using a fast and non memory-intensive computational pipeline.

\section{Compressive QMRI acquisition model}
\label{sec:model}
The compressed sensing approach adopted by MRF for acquiring quantitative information follows a linear spatiotemporal model~\cite{MRF}:
\eql{\label{eq:forward}
	Y=\Aa (\overline X)+\xi,
	}
where $Y\in \CC^{T\times m}$ is the multi-coil k-space measurements collected at $t= 1,\hdots,T$ temporal frames and corrupted by some  noise $\xi$. The Time-Series of Magnetisation Images (TSMI) \textemdash to be reconstructed\textemdash is an image sequence represented by a complex-valued matrix $\overline X$ of spatiotemporal resolution $T\times n$  i.e. $n$ spatial voxels across $T$ temporal frames. 
The forward operator $\Aa:= F_\Omega  S$ models the multi-coil sensitivities operator $S$,  and the Fourier transform $F$  
subsampled according to a set of \emph{temporally-varying} k-space locations $\Omega$.

The tissues' quantitative properties in each voxel are encoded in a temporal signal at the corresponding column of the TMSI matrix. This signal records the magnetisation response of proton dipoles 
to dynamic \emph{excitations} in the form a sequence of flip angles (magnetic field rotations) applying with certain repetition (TR) and echo (TE) times. Tissues with different NMR characteristics respond distinctively to excitations. QMRI/MRF rely on this principle to estimate quantitative  characteristics from the (computed) TSMI. 
Per-voxel $v$ magnetisation responses of the TSMI scaled by the \emph{proton density} $\gamma_v$ are modelled as
\eql{ \label{eq:blochmodel}
\overline X_v \approx \gamma_v \Bb(\Theta_v) \quad \forall v\in 1,\hdots,n
}
where the \emph{Bloch response} $\Bb(\Theta_v): \RR^p \rightarrow \CC^T$ is a nonlinear mapping from per-voxel intrinsic NMR properties $\Theta_v$  to the corresponding (discrete-time) solution of the \emph{Bloch differential equations} which captures  the overall transient-state macroscopic dynamics of a voxel~\cite{jaynes1955matrix}. Our experiments use sequences that simultaneously encode $p=2$ characteristics in each voxel i.e. the T1 and T2 relaxation times. This could be further extended to include other properties e.g. off resonance frequencies, $T2^*$, diffusion and perfusion~\cite{EPIT1T2star,MRF-perfusion,MRF_T1T2diff}. 

\subsection{Low-dimensional manifold and subspace models}
Estimating  $\Theta$ (i.e. quantification) requires long enough sequences $T>p$ to create contrast between  different tissues' responses. As such the Bloch responses despite their high ambient dimension live on a \emph{low $p$-dimensional (nonlinear) sub-manifold} of $\CC^T$. 
Further it is observed that for certain excitation sequences, including those used in our experiments, this manifold is approximately embedded in a \emph{low-rank subspace} $\text{Range}(V)\subset \CC^T$ represented by an orthonormal matrix $V\in \CC^{T\times s}$ where $p< s\ll T$. 
Hence the following dimension-reduced alternatives for models \eqref{eq:forward} and \eqref{eq:blochmodel} can be deduced:
\begin{align}
&Y \approx \Aa (VX) \label{eq:lrmodel} \\
&X_v \approx \gamma_v V^H\Bb(\Theta_v) \label{eq:blochmodellr}
\end{align}
where $X\in \CC^{s\times n}$ is the dimension-reduced TSMI. This compact representation is the basis for the subspace compression methods~\cite{asslander-ADMMMRF, zhao-LR-MRF} and is proven beneficial to the runtime and accuracy (by noise trimming) of the reconstructions.


\subsection{Model-fitting for parameter inference}
Fitting computed TSMIs to the Bloch response model is central to QMRI. Per-voxel model-fitting according to~\eqref{eq:blochmodellr} for obtaining the NMR characteristics and proton density reads (see e.g.~\cite{SVDMRF,BLIPsiam}): 
\begin{align}
&\widehat \Theta_v = \Pp_\Bb(\widehat X_v):=\argmin_\Theta \|\widehat X_v-V^H\Bb(\Theta)\| \label{eq:proj} \\
&\widehat \gamma_v= \langle \widehat X_v, V^H\Bb(\widehat \Theta_v)\rangle \label{eq:PDestim}
\end{align}
We assumed without losing generality having normalised Bloch responses. 
We refer to $\Pp_\Bb(.)$ as the \emph{Bloch response manifold projection.}  This projection is nonconvex and oftentimes intractable for the generally complicated Bloch responses adopted by the MRF sequences. The MRF framework instead 
approximates~\eqref{eq:proj} by \emph{dictionary matching} (DM). A \emph{fingerprint} dictionary $D=\{D_j\}$ is constructed for sampling the manifold of Bloch responses through a fine-grid discretesation of the parameter space $[\Theta]=[T1]\times[T2]\times \ldots$ and exhaustively simulating the Bloch responses $D_j := \Bb([\Theta_j])$
for all combinations of the quantised parameters. The DM step identifies the most correlated fingerprint (and the underlying NMR parameters) for each voxel of the reconstructed TMSI:
\eql{\label{eq:NNS}
	\Pp_\Bb(\widehat X_v) \approx \argmin_j\|\widehat X_v-V^HD_j\|
}
through a nearest neighbour \emph{search} 
that is itself a projection onto the discrete set of fingerprints i.e. a \emph{point-wise approximation} to the (continuous) Bloch response manifold. 

Viewing fingerprints as training samples, the dictionary can be factorised through principal component analysis (PCA)~\cite{SVDMRF}: 
\eql{\label{eq:pca}
DD^H \approx V \Lambda V^H
}
for unsupervisedly learning the low-rank subspace representation of the Bloch responses. This representation helps to reduce temporal dimension and can be coupled with fast search schemes~\cite{MRF-GRM,coverblip_iop,AIRMRF} to accelerate DM runtime. However any form of DM (fast or exhaustive search) remains non-scalable and creates storage overhead in multi-parametric QMRI applications because the number of dictionary atoms \emph{exponentially} grows with $p$. 

\section{DM-free image reconstruction and parameter inference pipeline}
\label{sec:pipeline}
Our DM-free image computing pipeline consists of two stages: i) reconstructing TMSIs from undersampled k-space measurements and then  ii) approximate model-fitting according to~\eqref{eq:blochmodellr} for parameter inference. 
A set of simulated fingerprints (could be MRF dictionary) sample the Bloch response model and are used only for training (pre-processing) in order to learn three temporal-domain models: i) a dimension-reduced (low-rank) {subspace representation} for the Bloch responses, ii) an encoder network to map noisy fingerprints to the NMR parameters,  and iii) a decoder network to generate clean Bloch responses  from the NMR parameters.


\subsection{Convex TSMI reconstruction with LRTV algorithm}
The~\cite{SVDMRF}'s baseline
backprojects k-space measurements to form a dimension-reduced TSMI through the adjoint of~\eqref{eq:lrmodel}:  
 \eql{ \label{eq:zf}
 \widehat X =V^H\Aa^H(Y) \in \CC^{s\times n}
 } 
 prior to the DM inference. 
Modern QMRI/MRF acquisitions aggressively curtail the scan times by using short excitation sequences and severe spatial (k-space) subsampling. As such the inverse problem~\eqref{eq:forward} becomes highly ill-posed and ZF/backprojection (which is not an inversion) results in aliasing artefacts in the reconstructed TSMI. 
 Errors made at this stage can be indeed significant (see experiment results), they propagate to the parameter inference step and deteriorate the overall quantification accuracy. 

To address this issue, we propose the following convex and DM-free optimisation dubbed as LRTV to compute dimension-reduced TSMIs using joint spatiotemporal regularisations:
\eql{ \label{eq:optim}
\widehat X =\argmin_{X\in \CC^{s\times n}} \|Y-\Aa(VX)\|^2 + \sum_{i=1}^s \lambda_i \|X_{(i,.)}\|_{TV}
}
The first term minimises discrepancies between the k-space measurements and the solutions through the factorised forward model~\eqref{eq:lrmodel}. As such LRTV adopts a temporal-domain prior through the subspace model (i.e. the low-rank factorisation $\overline{X}\approx VX$) which provides a compact and convex (in fact linear)  \emph{relaxed representation} for the Bloch response model instead of using the MRF dictionary. 
LRTV additionally adopts Total Variation (TV) regularisation.
Each component of the TSMI corresponds to a spatial 2D or 3D volumetric image $X_{(i,.)}$ (matrix row), where penalising its TV norm promotes spatial-domain regularities via sparse image gradients. $\lambda_i>0$ control per (subspace) component regularisation levels.

%

Our LRTV formulation~\eqref{eq:optim} can be efficiently solved using Fast Iterative Shrinkage Algorithm with
Nesterov momentum acceleration and backtracking step-size~\cite{FISTA, nesterov-momentum}.
Each iteration $k=0,1,2\ldots$ computes:
\eql{\label{eq:fistalgo}
\begin{cases}
&\nabla= X^k-\mu_k V^H \Aa^H\left(\Aa(V X^k) -Y\right)\\
&Z_{(i,.)}^{k} =\textbf{Prox}_{\la_i \mu_k} (\nabla_{(i,.)}) \quad \forall i=1,\ldots s\\
&X^{k+1} = Z^k+\left(\frac{k-1}{k+2}\right)(Z^k-Z^{k-1})
\end{cases}
}
The first and third lines correspond to the gradient and momentum-acceleration updates, respectively. The second line computes a small number $s\ll T$ of shrinkage operations for the 2D/3D images in each subspace component $\textbf{Prox}_{\alpha} (x):=\argmin_u\frac{1}{2}\|x-u\|^2+\alpha\|u\|_{TV}$, which can be efficiently done on a GPU using the Primal-Dual algorithm~\cite{Chambolle2011}. Per iteration, the initial step size $\mu_k$ halves until the following criteria holds:\footnote{Optional warm-start could rescale the chosen $\mu_1$ by factor $\|Y\|/\|\Aa(X^1)\|$.}  
\begin{align*}
\|Y-\Aa(VZ^k)\|^2> &\|Y-\Aa(VX^k)\|^2+\\
&2\text{Re}\langle G,Z^k-X^k\rangle+\mu_k^{-1} \|Z^k-X^k\|^2
\end{align*}

With an all-zero initialisation, the first line of~\eqref{eq:fistalgo} recovers ZF~\cite{SVDMRF} in the first iteration. Setting $\la=0$ recovers the LR formulation in~\cite{zhao-LR-MRF}. LR is a convex relaxed alternative to DM-based models~\cite{BLIPsiam,asslander-ADMMMRF}, wherein temporal-only priors based on the MRF dictionary are replaced by the low-rank subspace. Note that the size of $V$ is independent of the number of fingerprints (used for training). Hence the solver does not face a memory bottleneck and the slow progress of computing DM per iteration.
While for certain (Cartesian) sampling schemes this temporal model can decently regularise the inversion~\cite{benjamin2019multi}, for other important sampling patterns e.g. non-cartesian spiral and radial readouts used in our experiments, it turns out to be inadequate and fails to output artefact-free TSMIs (see section~\ref{sec:expe}). Multi-prior CS solvers are proven
effective for highly undersampled systems by further restricting degrees of freedom of data~\cite{meLRJS, meLRTV}. The LRTV uses this fact by setting $\la>0$ and adding spatial priors to sufficiently regularise the problem. Besides being DM-free, the LRTV has other advantages over its non-convex spatiotemporal alternatives~\cite{AIRMRF,simon_lrtv}, including a tractable way to incorporate multiple priors\footnote{In non-convex (e.g. DM-based) approaches incorporating extra priors such as spatial regularity constraints is not always algorithmically tractable e.g. sequential projections on two sets where one is non-convex may not result in projecting onto the intersection.}, momentum-acceleration for fast convergence and reproducible global solutions regardless of initialisation.

\vspace{-.2cm}
\subsection{MRFResnet for parameter inference}

\begin{figure}[t!]
	\centering
	\begin{minipage}{\linewidth}
		\centering
		\includegraphics[trim=10 40 0 100,clip,width=\linewidth]{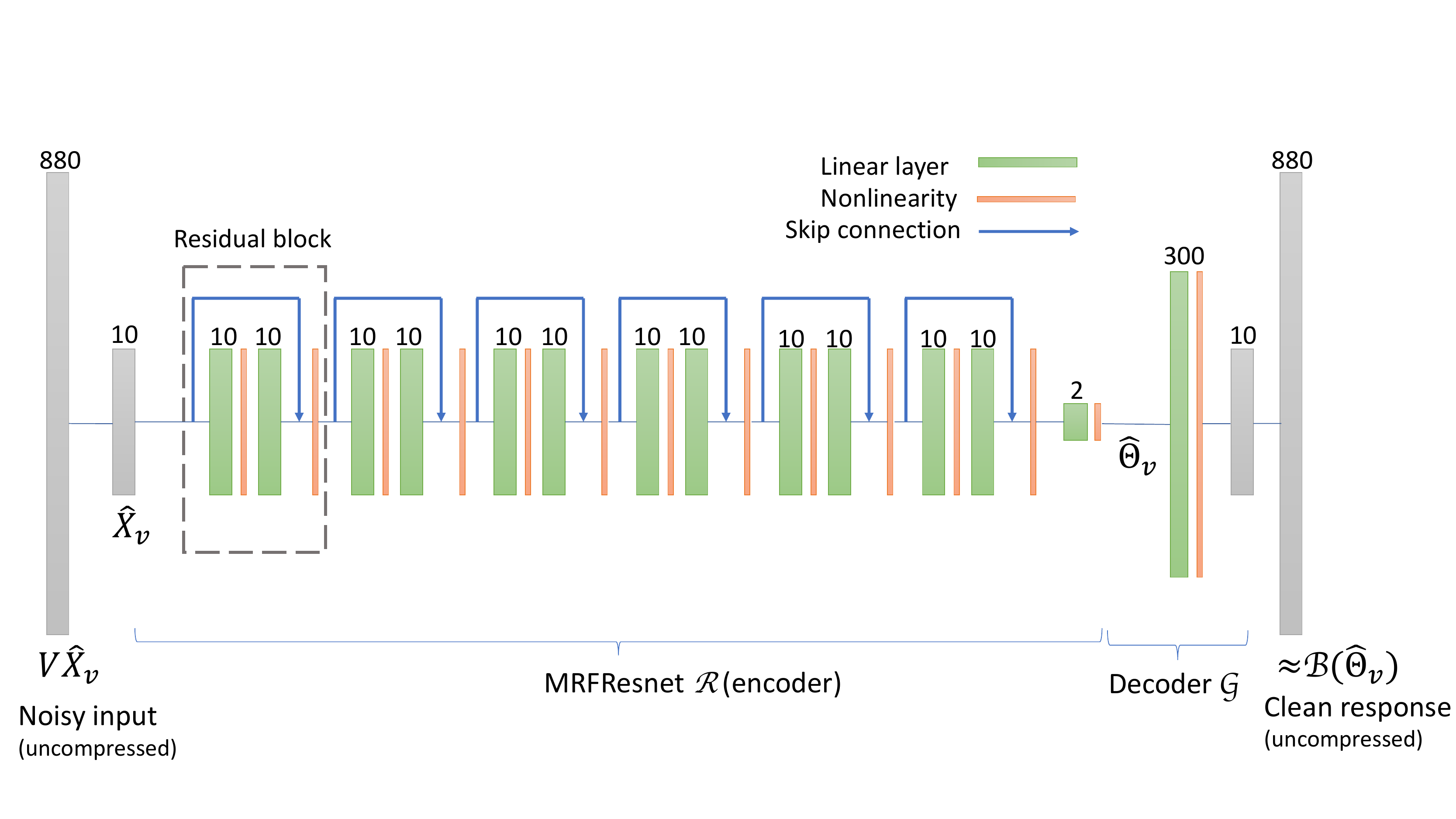} 	
		\caption{\footnotesize{MRFResnet (encoder) for T1/T2 inference, the Bloch response generative network $\Gg$ (decoder), and the \emph{implicit} linear dimensionality reduction/expansion (first/last) layers using the subspace model $V^H/V$. 
		} \label{fig:net}}
	\end{minipage}
\end{figure}

Instead of using a large-size dictionary for DM, we propose training and using a compact network coined as \emph{MRFResnet} in the form of an auto-encoder with deep residual blocks, shown in Figure~\ref{fig:net}.
Auto-encoders have proven powerful in denoising tasks through creating an \emph{information bottleneck} which corresponds to learning a low-dimensional manifold model for capturing (nonlinear) intrinsic signal structures~\cite{SDAE_bengio}. 
In our task computed TSMI voxels are processed by such a model to create clean magnetisation responses as well as estimating the intrinsic NMR parameters in a computably efficient manner. The $p=2$ neurons bottleneck (in Figure~\ref{fig:net}) has a physical interpretation: fitting noisy temporal trajectories to the nonlinear Bloch model with limited $p\ll T$ degrees of freedom determined by the T1 and T2 quantities.

\subsubsection{Encoder}

This network learns to approximate Bloch manifold projections through a continuous mapping  $\Rr: \widehat X_v \rightarrow\Theta_v$ parametrised by the network's weights and biases $\{W,\beta\}$:
\begin{align}
\Rr(x) \equiv h^{(N+1)}(x)=\phi\left(W^{(N+1)}h^{(N)}(x)+\beta^{(N+1)}\right) \label{eq:resnet1}
\end{align}
where $h^{(i)}$ the outputs of $i=1,\ldots,N$ residual blocks are
\begin{align*}
&h^{(i)}(x)= \phi\left( h^{(i-1)}(x) + g^{(i)}(x)  \right), \\
&\text{and} \quad g^{(i)} (x)= W^{(i,2)}\phi\left(W^{(i,1)}h^{(i-1)}(x)+\beta^{(i,1)}\right)+\beta^{(i,2)},
\end{align*}
$h^{(0)}(x)=x$ and the ReLU activations $\phi(x)=\max(x,0)$ are used throughout.
 The inputs are the normalised temporal voxels of the dimension-reduced TSMI. The network is trained on simulated noisy Bloch responses (see section~\ref{sec:algosetup}) so that the approximate projection holds
 \eql{ \Rr(x) \approx \Pp_\Bb(x)
 }
 in a neighbourhood of the (compressed) Bloch manifold.

 \subsubsection{Decoder}
The proton density (PD) is a scaling factor that amplifies the Bloch responses in each voxel. Hence after estimating other nonlinear NMR parameters (e.g. T1/T2) using the encoder part, PD can be explicitly resolved through~\eqref{eq:PDestim}. 
This would however require either storing a dense dictionary or evaluating Bloch responses for all voxels and their parameters $\Theta_v$, which can be computationally intensive. Instead we train a \emph{decoder} network $\Gg(.)$ which for given NMR parameters it approximately generates
\eql{\Gg(\Theta_v) \approx V^H\Bb(\Theta_v)
}
 the corresponding compressed Bloch responses (clean fingerprints) in short runtimes. This allows~\eqref{eq:PDestim} to be easily applied without significant computations.
For the sequence design used in our experiments, it turns out that a fully-connected shallow network with one hidden layer and ReLU activations can approximate well this step.\footnote{We also observed a similar network complexity for generating responses to the well-known FISP sequence~\cite{FISP}. However, we did not achieved accurate predictions using shallow architectures of comparable sizes for $\Rr$ (two layer were needed at least however with larger model than MRFResnet~\cite{deepMRFgeom, LRTVismrm}). This suggests that generating clean responses (decoding) was easier than projecting noisy fingerprints to their generative parameters (encoding), and the latter requires deep processing (see section~\ref{sec:partitioninig}).
} Unit dimensions are customised to a sequence used in our experiments encoding T1/T2 relaxation times, with reduced subspace dimension $s=10$.   
 Encoder has $N=6$ residual blocks 
 of 10 neurons width, and decoder has 300 neurons in its single hidden layer. 


The subspace compression helps reduce model sizes in both networks (hence reducing risk of overfitted predictions) and also reduce required training resources compared to uncompressed deep MRF approaches~\cite{DRONE-MRF,betterreal}. 
Further to avoid losing discrimination between fingerprints \textemdash e.g. by a magnitude-only data processing~\cite{DRONE-MRF} \textemdash we adopt 
a practical phase-correction step~\cite{coverblip_iop,AIRMRF} to align phases of the complex-valued TSMIs and training samples before being fed to the MRFResnet. Complex-valued phases estimated from the first principal components of the dimension-reduced TSMI pixels and the training samples are used for phase-alignment. 
This treatement allows the network without losing generality have real-valued parameters and approximate real-valued mappings.

\section{Hierarchical partitioning of the Bloch response manifold}
\label{sec:partitioninig}
In this part we show that the MRFResnet provides a multi-scale piecewise affine approximation to the Bloch response manifold projection \eqref{eq:proj}. 
Hierarchical partitioning and multi-scale approximations are also central to the fast search schemes proposed for the DM-based MRF (see illustrations in \cite{coverblip_iop,CoverBLIP-MLSP}).
However unlike any form of DM (fast or exhaustive) that creates point-wise approximations for 
\eqref{eq:NNS}, 
MRFResnet does not memorise a dictionary and rather uses it to learn and efficiently encode a compact set of partitions and \emph{deep matched-filters} for affine regression of the NMR quantities.

\begin{figure*}[t!]
	\centering
	\scalebox{1}{
	\begin{minipage}{\linewidth}
		\centering		
		\includegraphics[trim=115 25 110 75,clip, width=.165\linewidth]{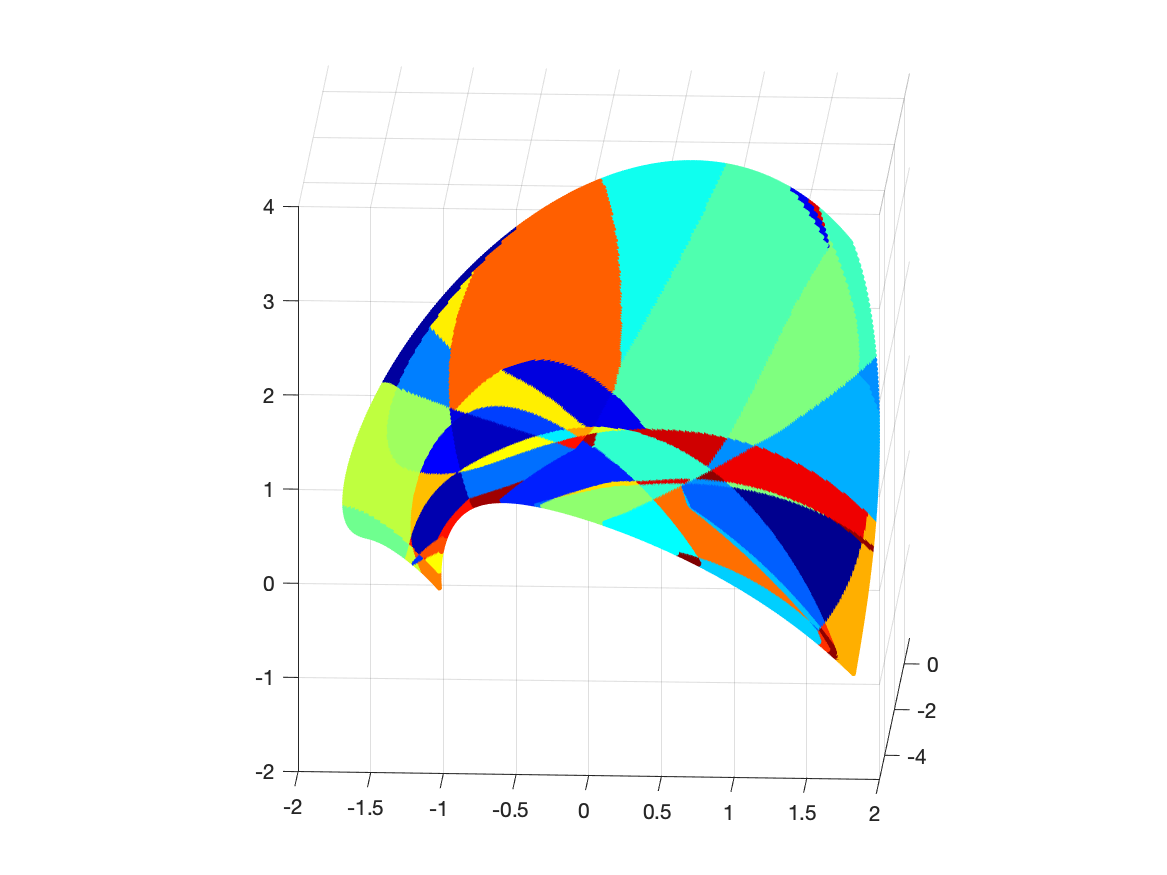}\hspace{-.1cm}
		\includegraphics[trim=115 25 110 75,clip,width=.165\linewidth]{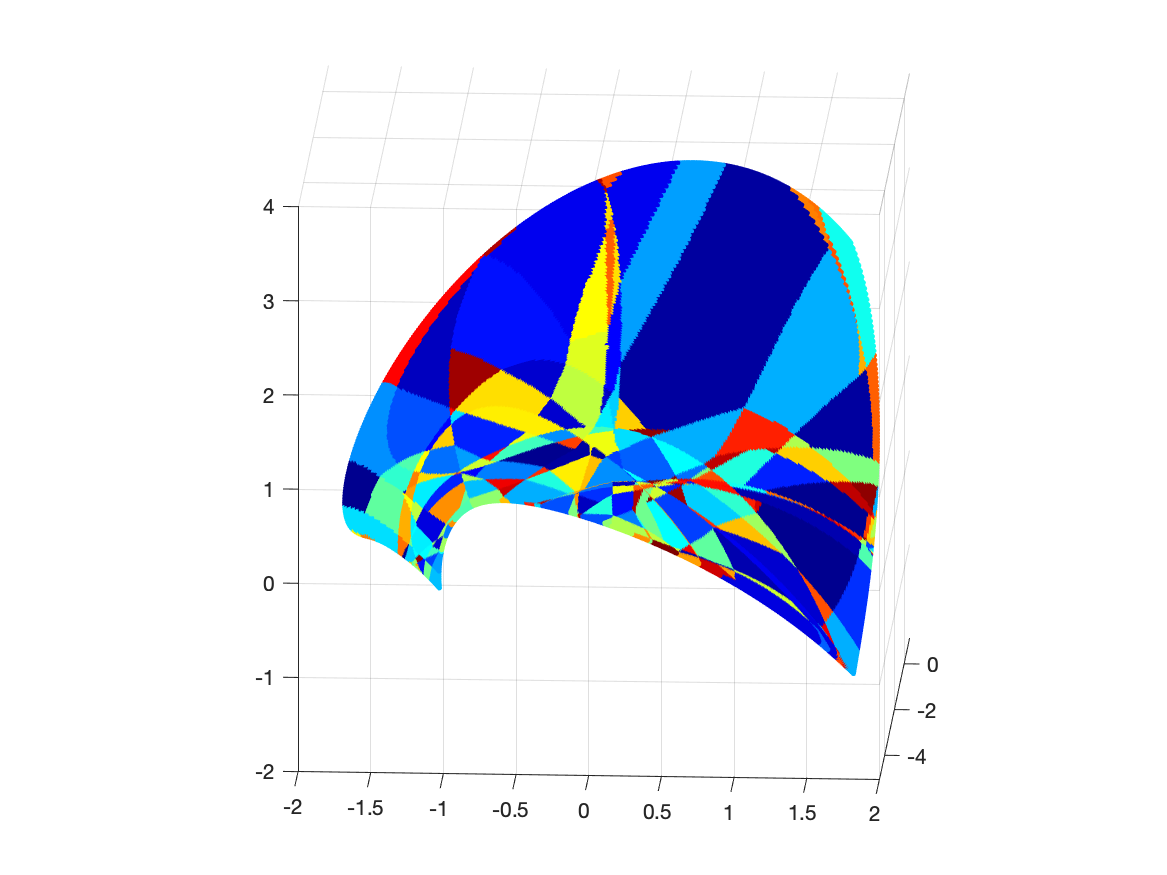}\hspace{-.1cm}
		\includegraphics[trim=115 25 110 75,clip,width=.165\linewidth]{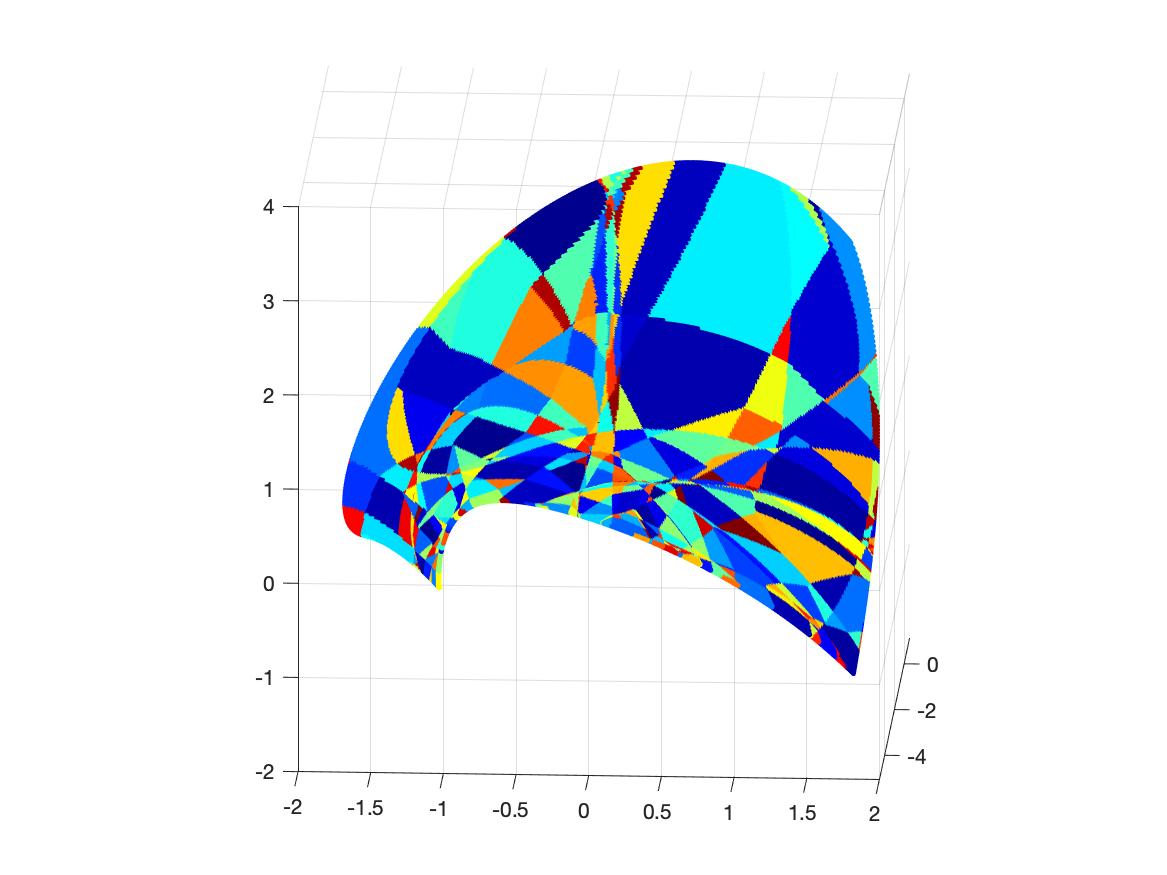}\hspace{-.1cm}
		\includegraphics[trim=115 25 110 75,clip,width=.165\linewidth]{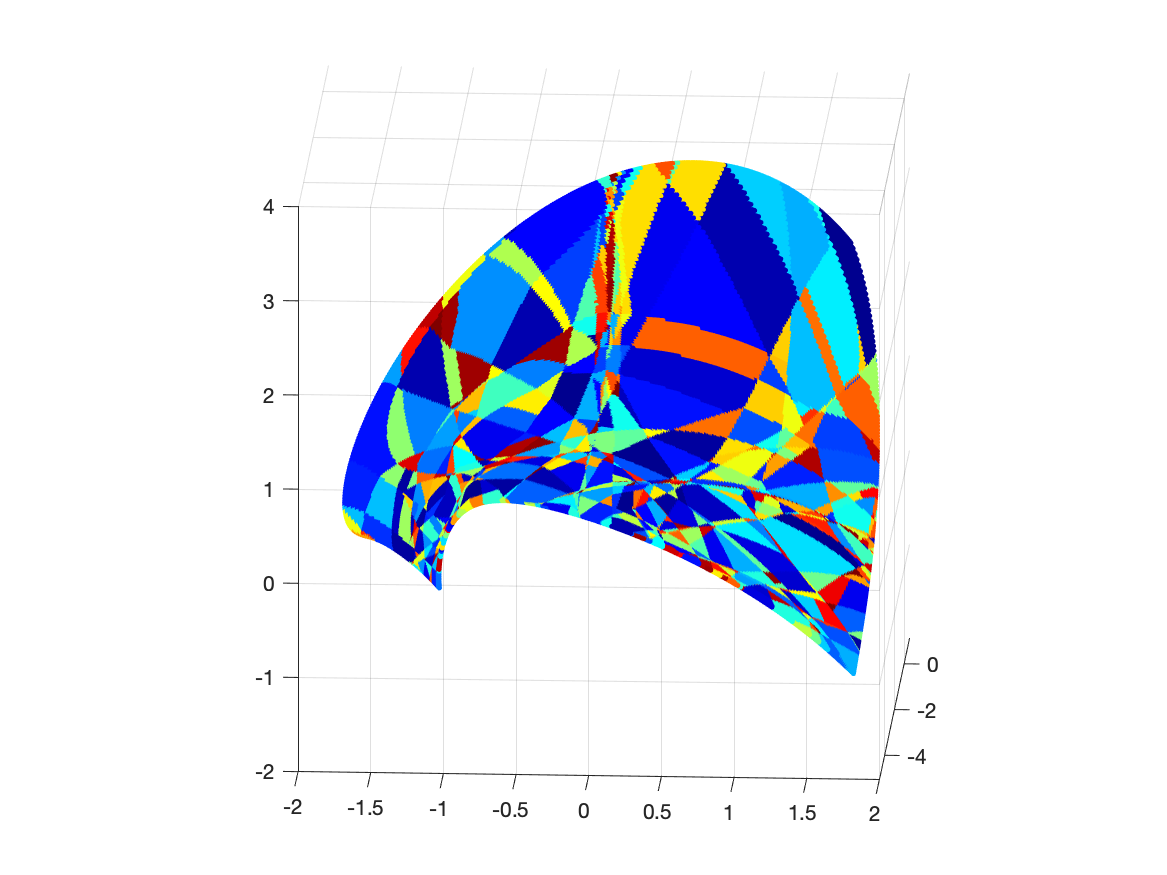}\hspace{-.1cm}
		\includegraphics[trim=115 25 110 75,clip,width=.165\linewidth]{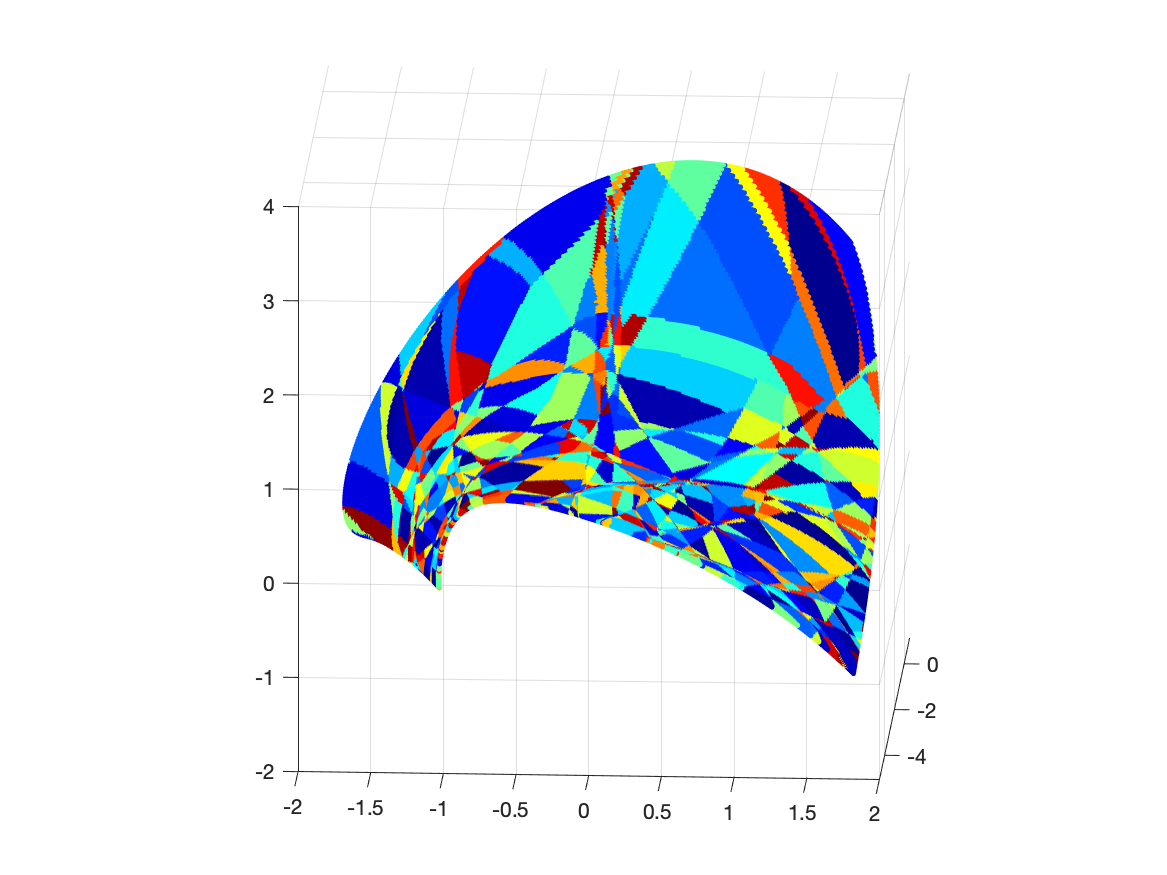}\hspace{-.1cm}
		\includegraphics[trim=115 25 110 75,clip,width=.165\linewidth]{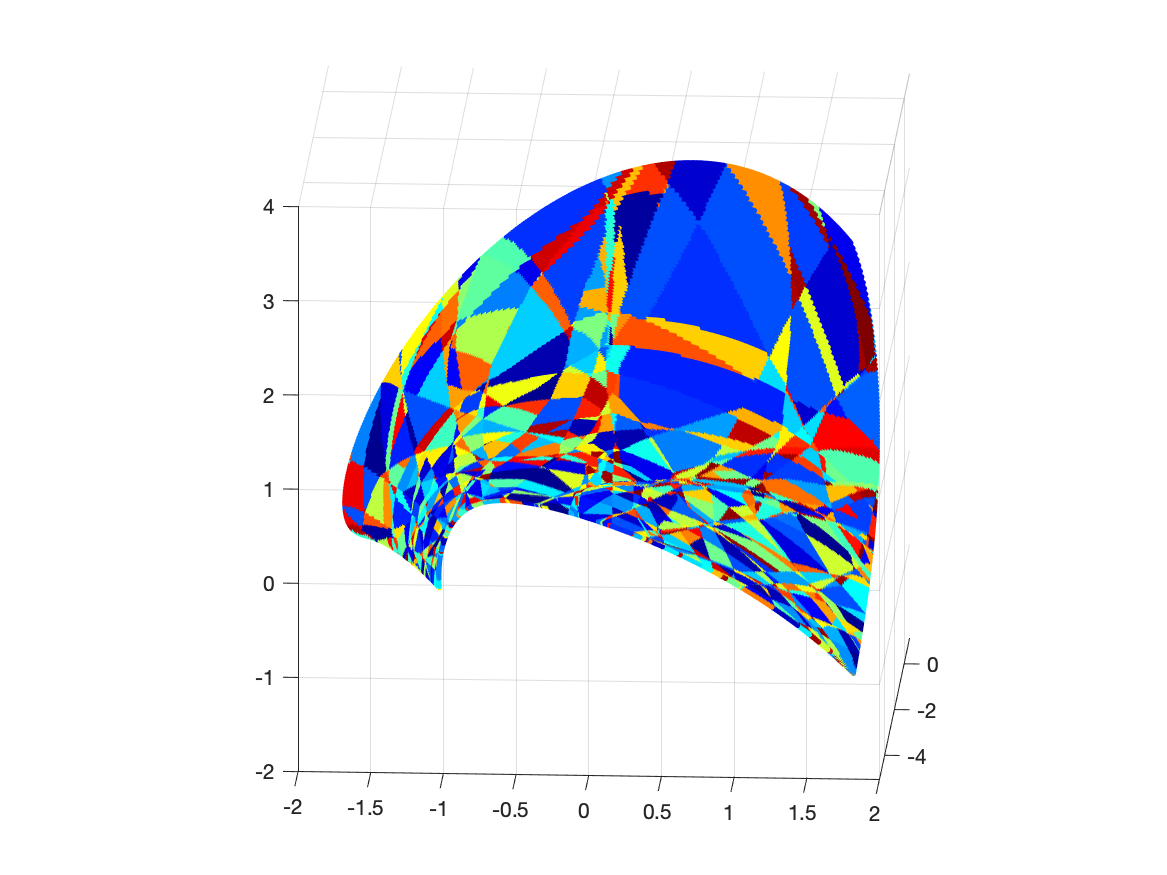}
		\\
		\includegraphics[width=.165\linewidth]{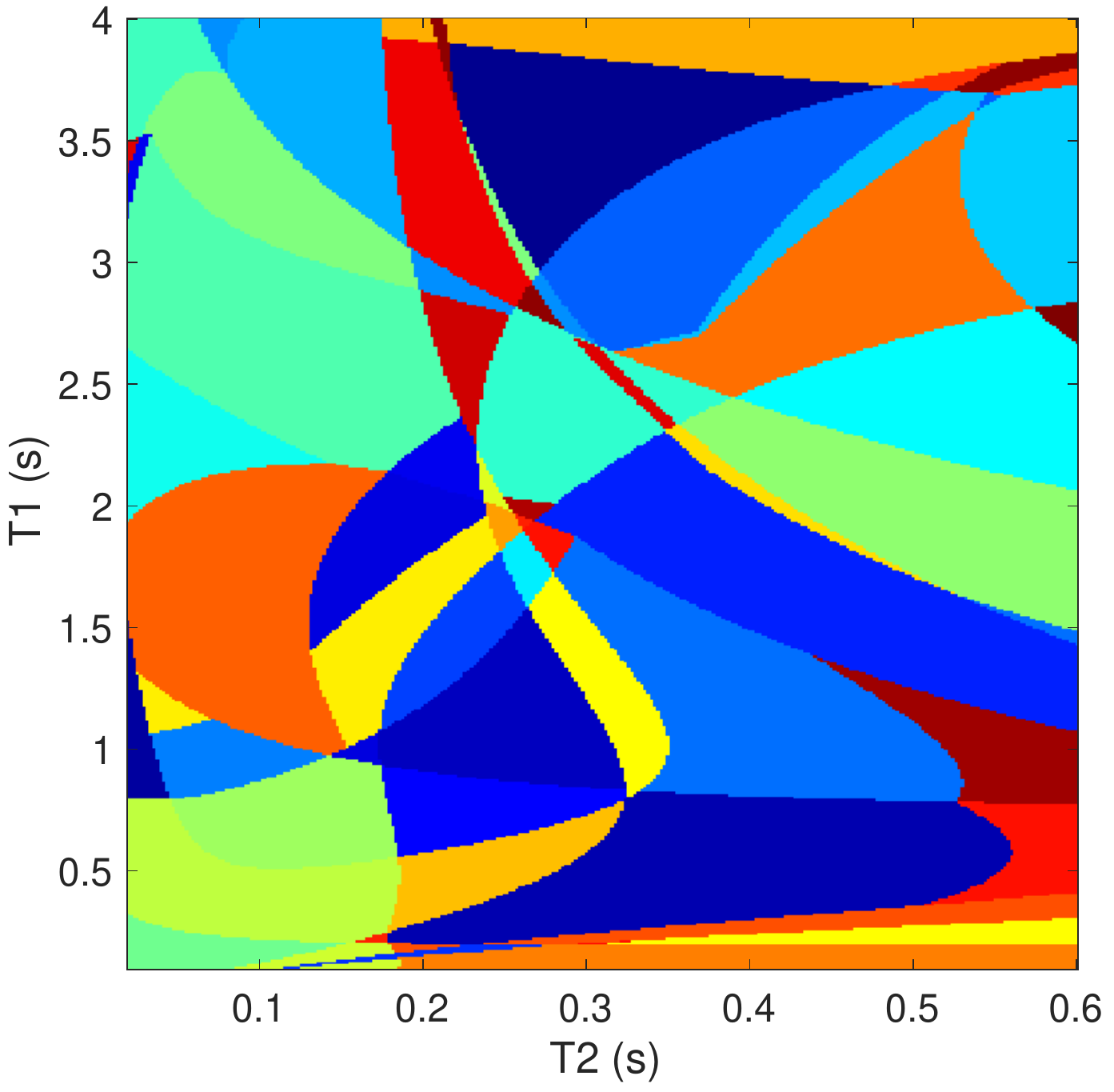}\hspace{-.1cm}
		\includegraphics[width=.165\linewidth]{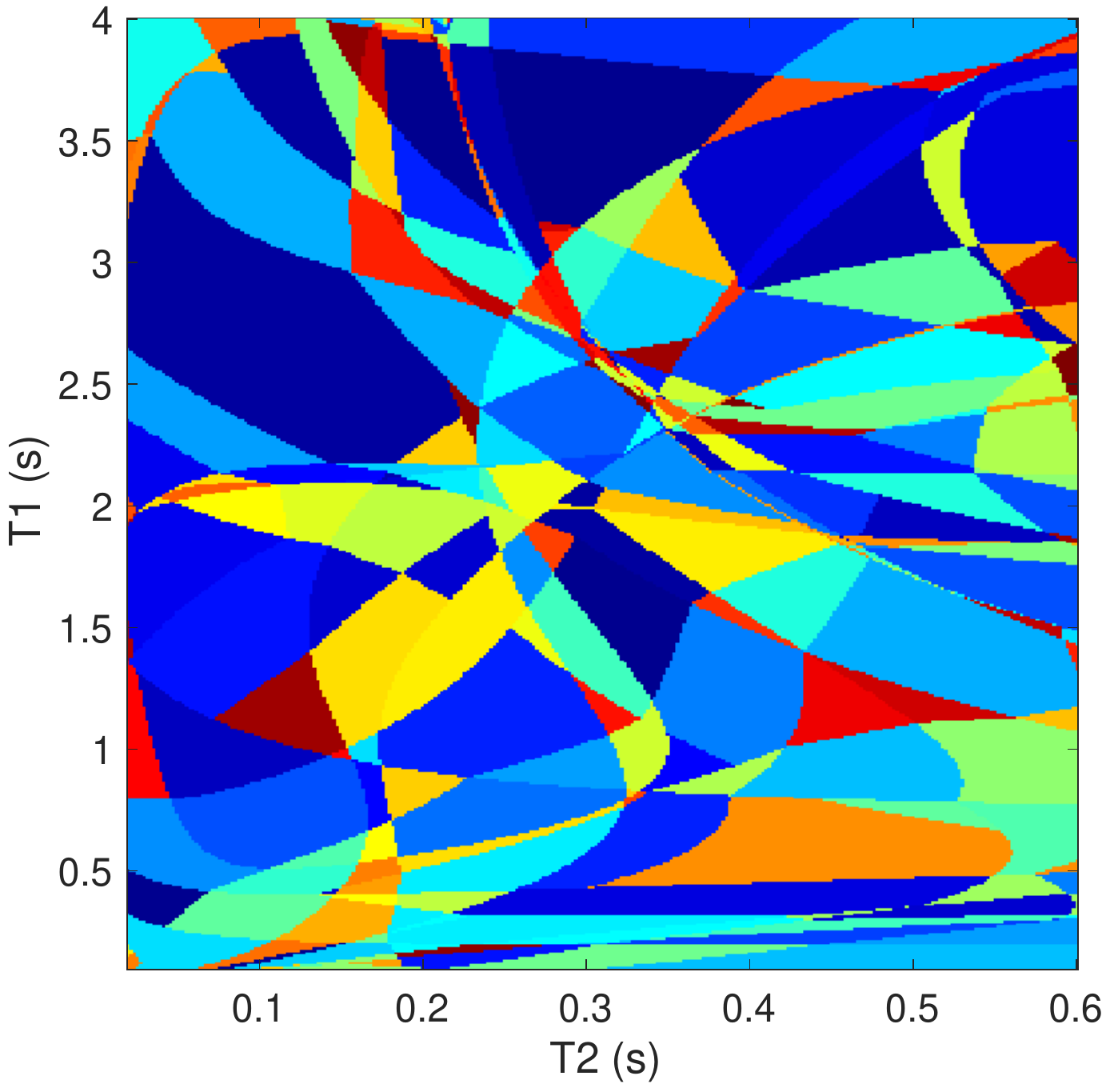}\hspace{-.1cm}
		\includegraphics[width=.165\linewidth]{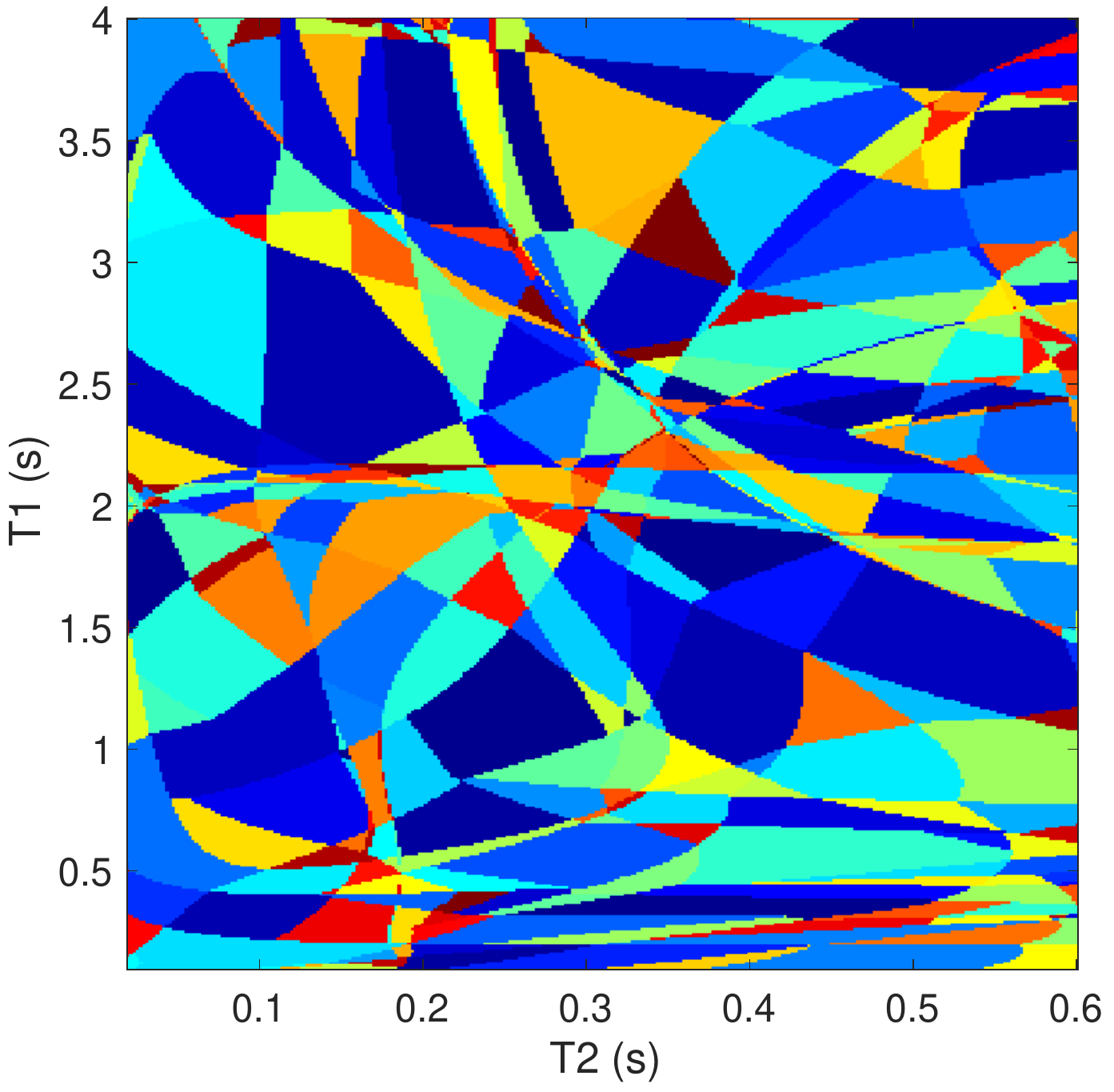}\hspace{-.1cm}
		\includegraphics[width=.165\linewidth]{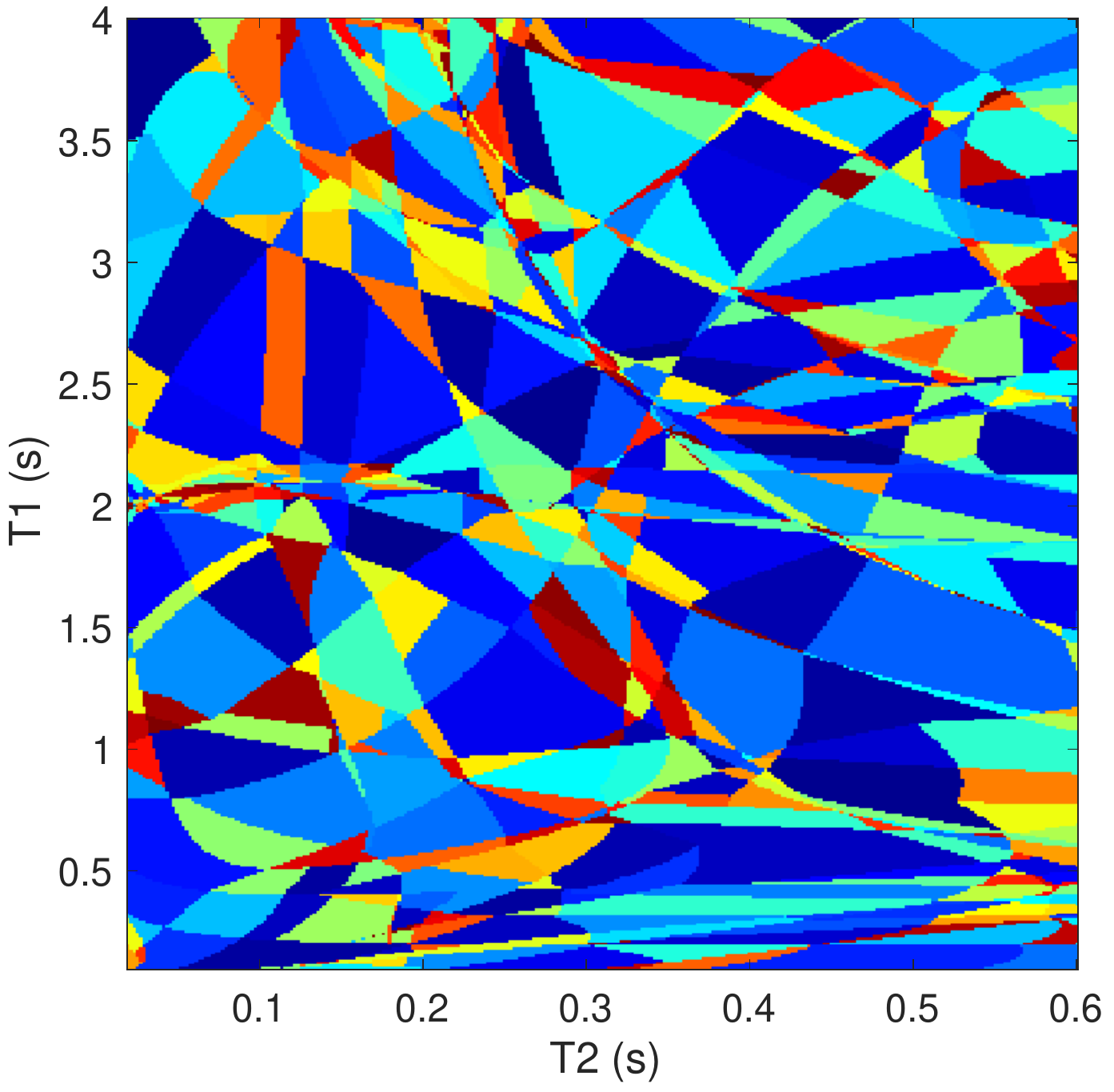}\hspace{-.1cm}
		\includegraphics[width=.165\linewidth]{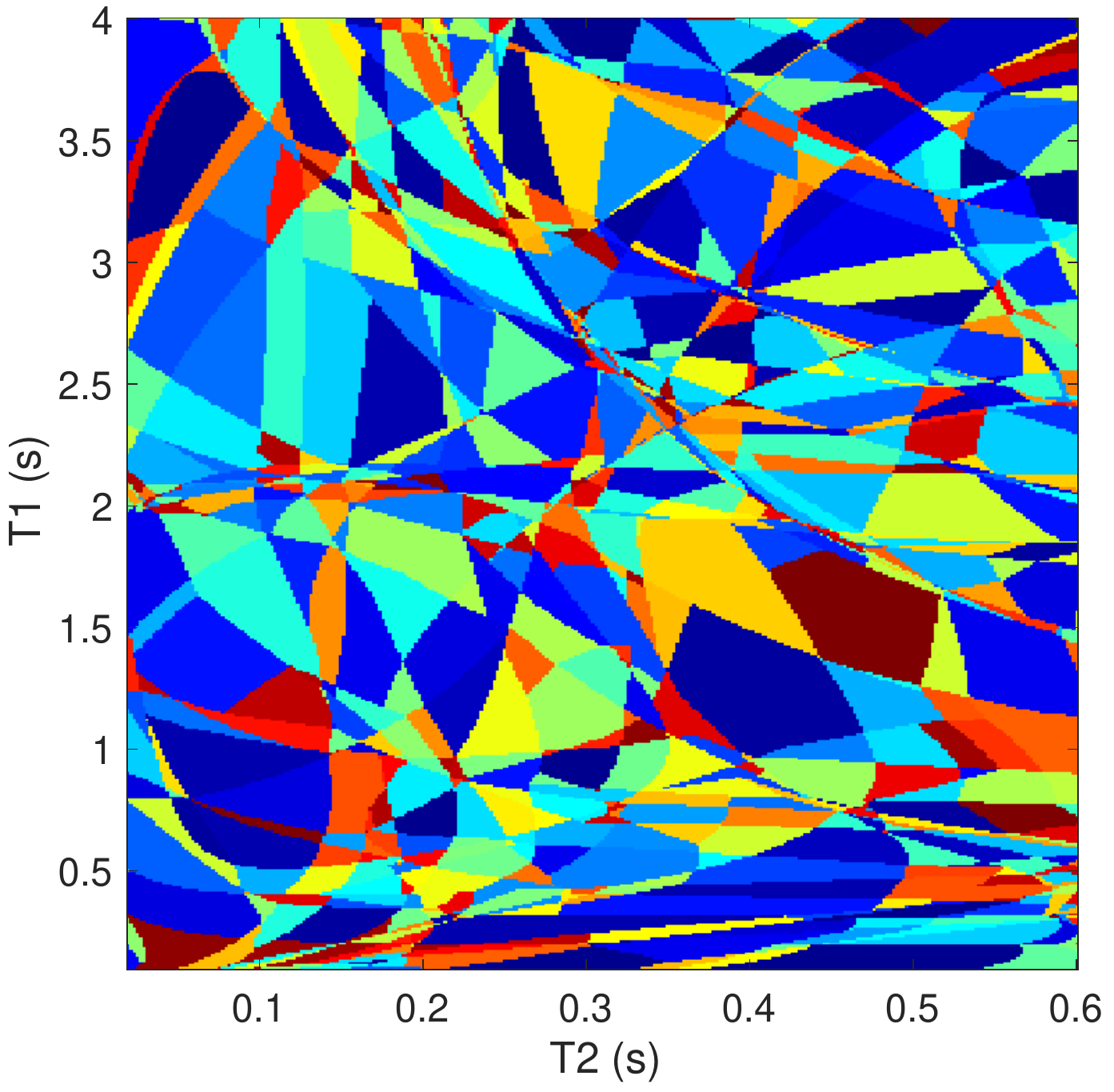}\hspace{-.1cm}
		\includegraphics[width=.165\linewidth]{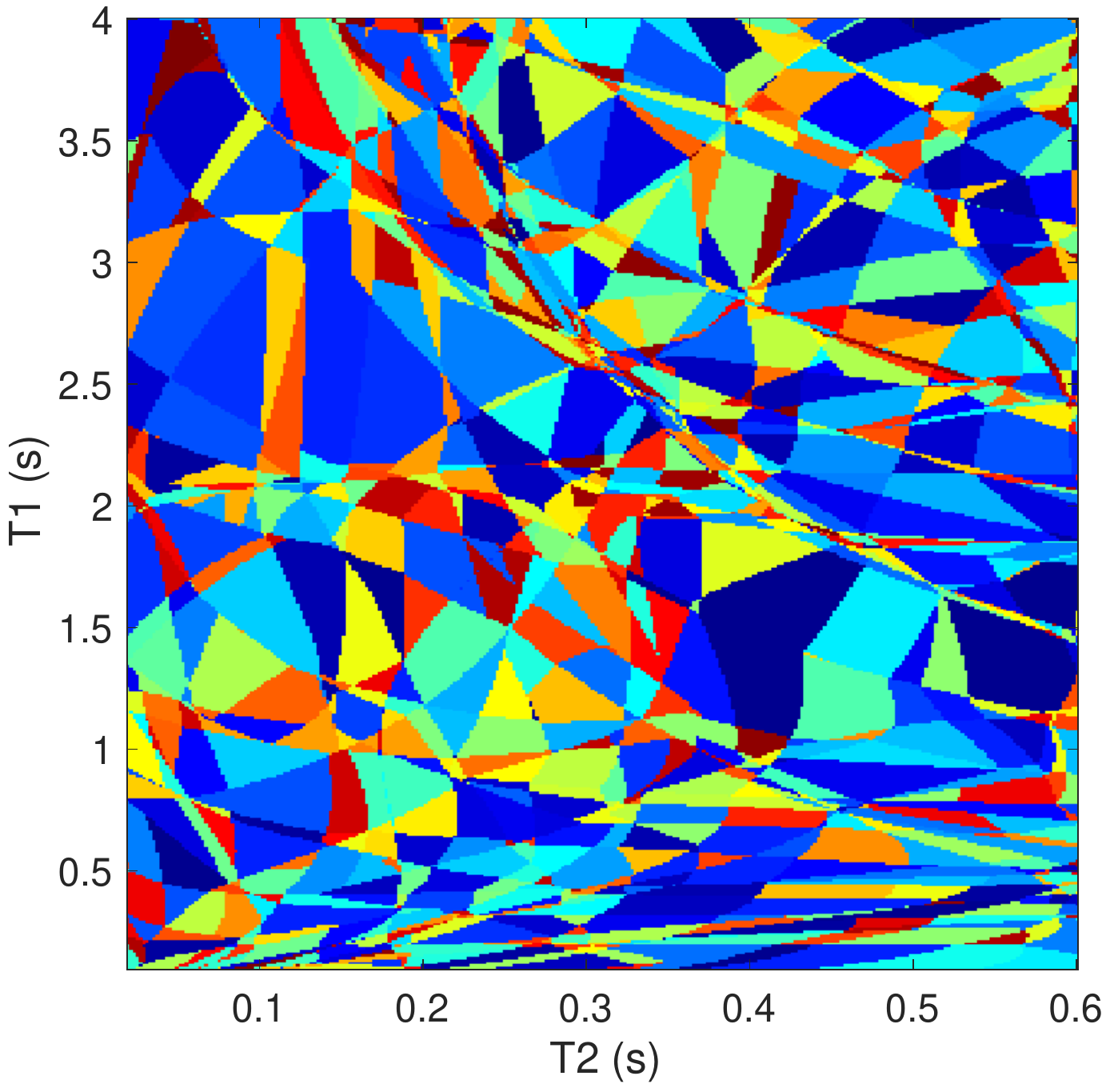}	
\caption{\footnotesize{Coarse-to-fine partitioning of the Bloch manifold (top row\textemdash manifold data is visualised using PCA) sampled by a dense fingerprinting dictionary, and their generative T1/T2 parameters (bottom row) using MRFResnet. From left to right figures illustrate learned colour-coded partitions after each residual block. \vspace{.3cm}}
\label{fig:seg}}
\end{minipage}}
\end{figure*}

\begin{figure}[t!]
	\centering
	\scalebox{1}{
	\begin{minipage}{\linewidth}
		\centering		
		\subfloat[The average magnetic response]{
		\includegraphics[width=.485\linewidth]{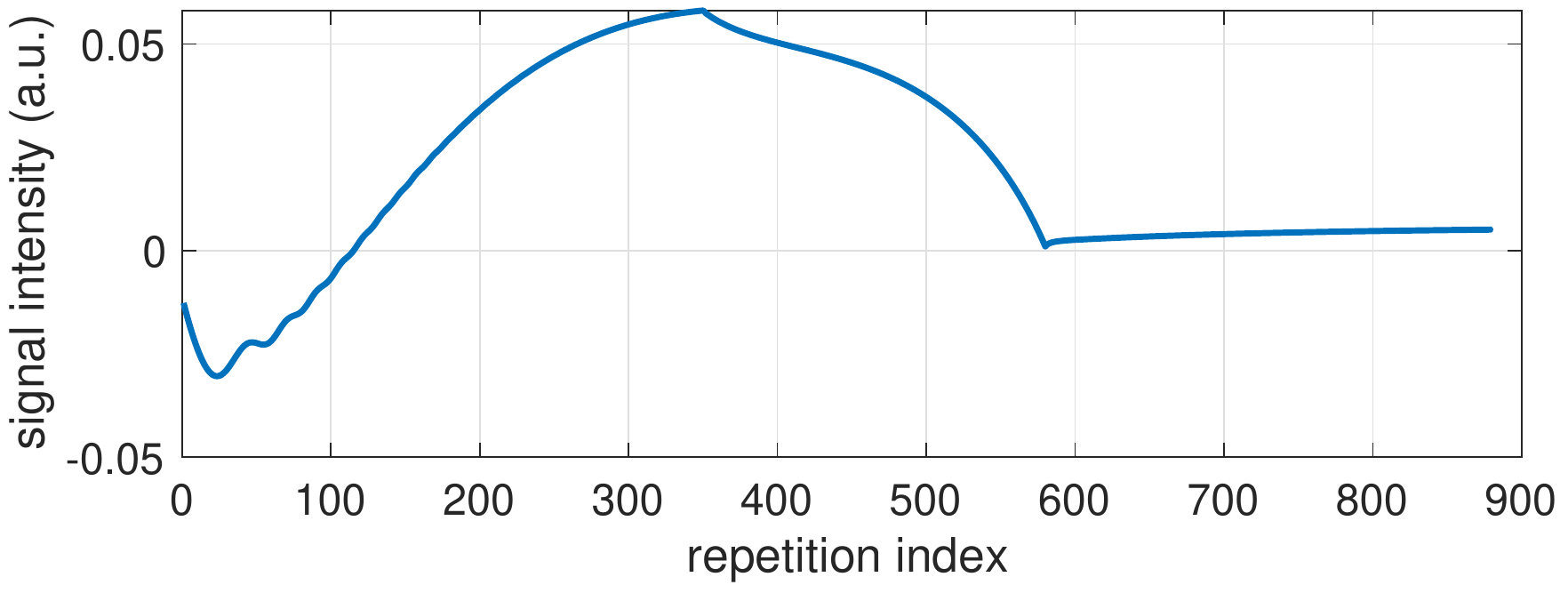}}\hspace{-.1cm}
		\subfloat[Centred magnetic responses]{
		\includegraphics[width=.485\linewidth]{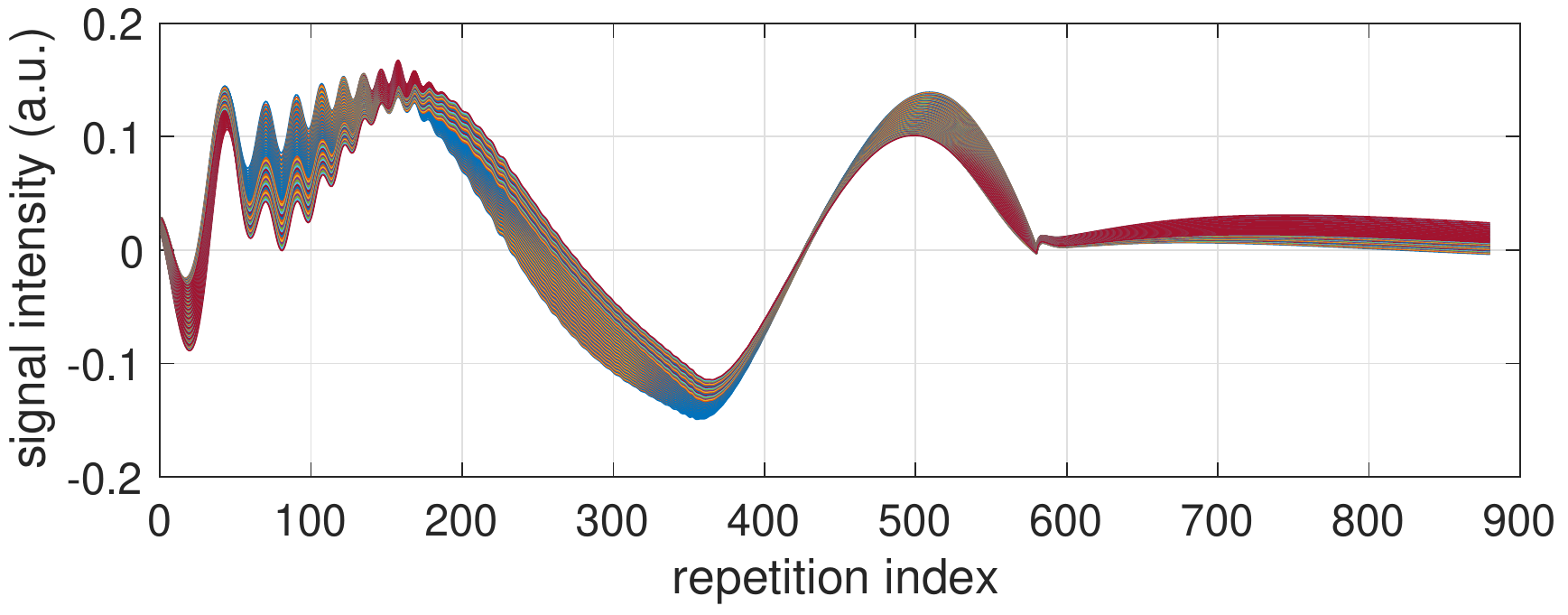}}\\
		\subfloat[T1 deep matched filter]{
		\includegraphics[width=.485\linewidth]{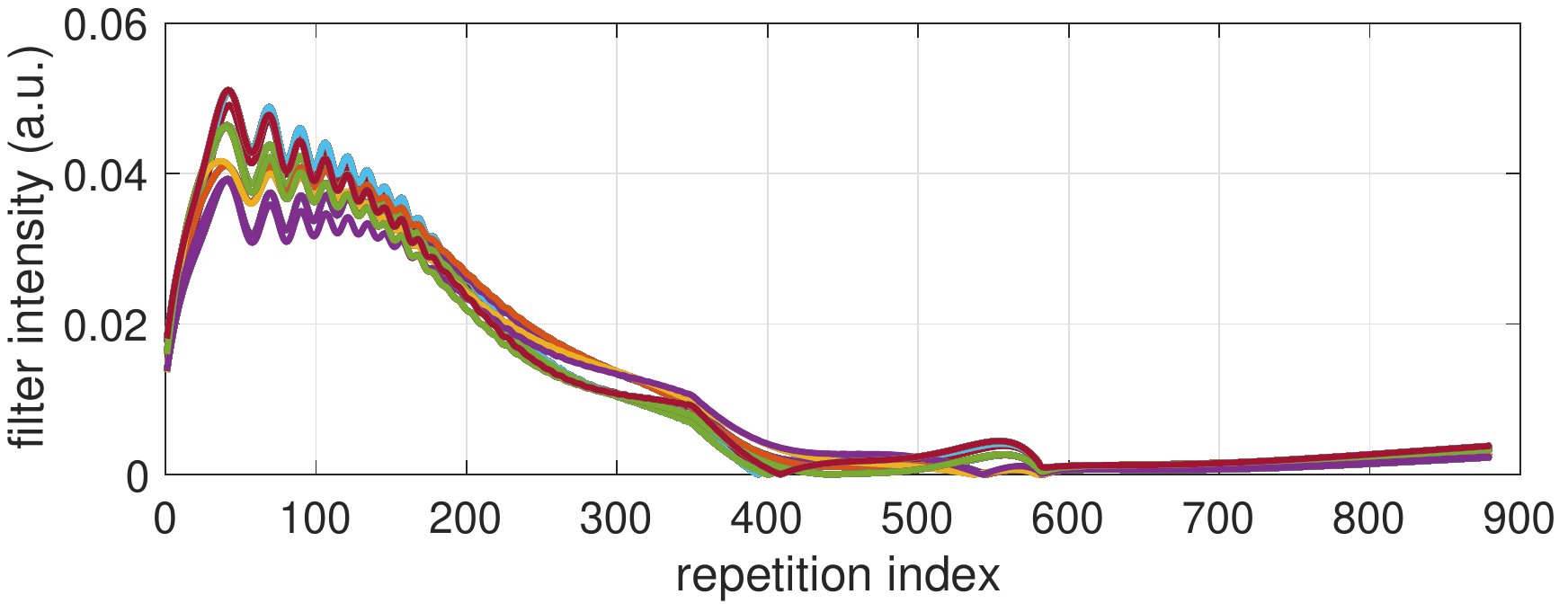}}\hspace{-.1cm}	
		\subfloat[T2 deep matched filter]{
		\includegraphics[width=.485\linewidth]{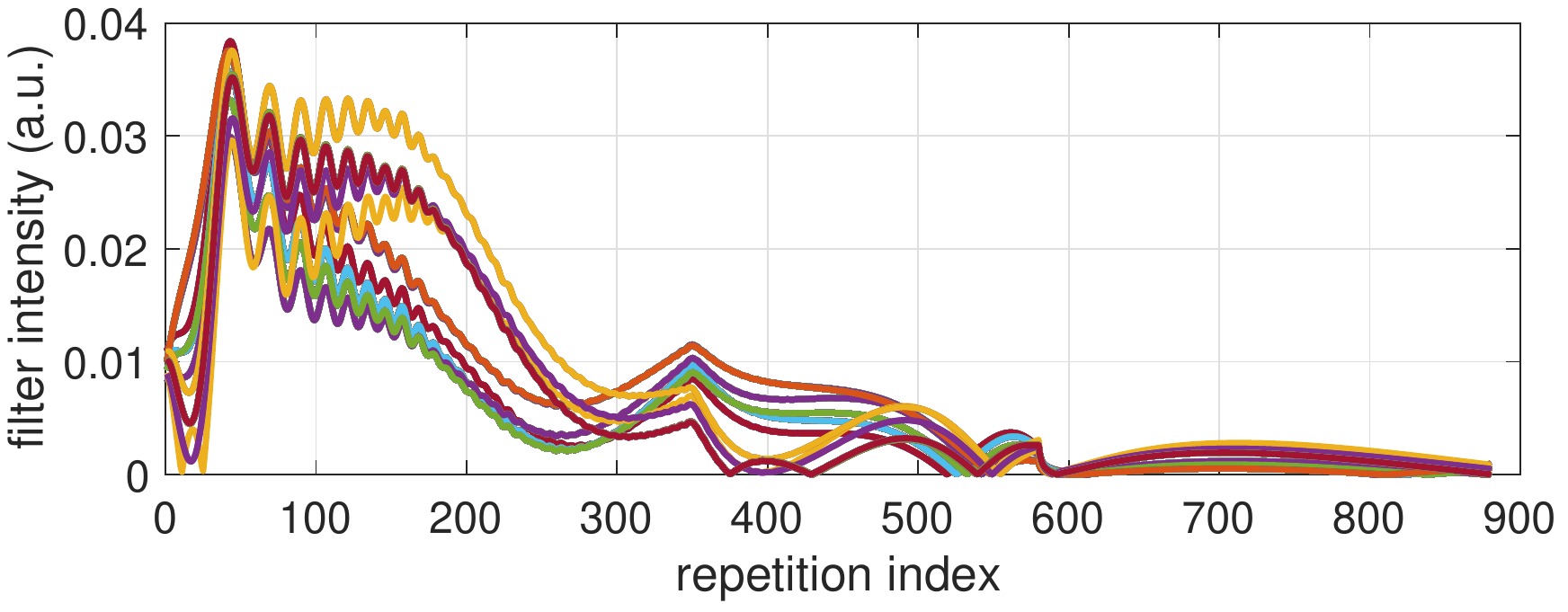}}	
\caption{ \footnotesize{(a) The mean and (b) centred Bloch responses within the range (T1, T2) $\in [1000-1200]\times[80-110]$ (ms). (c)-(d) The end-to-end match filters learned by the MRFResnet to regress T1/T2 values are shown across the original (non-compressed) temporal dimension.\vspace{.3cm}} \label{fig:matchfilt}}
\end{minipage}}
\end{figure}

\subsection{Affine spline function approximation}
The MRFResnet encoder 
(also its decoder network) 
is composed of linear connections and piecewise linear ReLU activations.
 This results  
in piecewise affine functions $h^{(i)}(x)$ after each residual block as well as the end-to-end mapping $\Rr(x)$~(see e.g.~\cite{splineDN, montufar2014number}). Further, $\Rr$ is \emph{Lipschitz continuous} for continuous activation functions as above and for bounded $\{W^{(i)}, \beta^{(i)}, i\}$. 


{\thm{
Denote by $z:\RR^s \rightarrow \RR^p$
\eql{ \label{eq:wo}
z(x):= W^{(N+1)} h^{(N) }(x)+\beta^{(N+1)} 
}
the \emph{weighted outputs} in \eqref{eq:resnet1} before the last non-linearity.\footnote{The last ReLU layer in $\Rr$ is for imposing the positivity of T1/T2 values, and therefore the prediction task is mainly done by the preceding layers. } The following \underline{affine spline representation} holds for MRFResnet:
\eql{ z(x) = \A[x]x+\b[x] :=\sum_r \left( A_r x+b_r \right)\iota_{\Omega_r}(x), \label{eq:spline}}
where $\iota_{\Omega_r}(x)$ is the indicator function with respect to a segment (set) $\Omega_r\in \RR^s$, returning $x$ if it belongs to the segment and 0 otherwise \textemdash segments form a disjoint partitioning of the input space with affine boundaries. Matrices $A_r\in\RR^{p\times s}$ and vectors $b_r\in\RR^p$ define the corresponding slopes and offsets for the input-output affine mapping in each segment. Shorthands $\A[x]: \RR^s\rightarrow \RR^p$ and $\b[x]:\RR^s\rightarrow \RR^p$ represent the input-dependent (piece-wise affine) mapping of $z(x)$. $\b[x]$ represents $p$ input-dependent offsets. Similarly, $\A[x]$ is an input-dependent $p\times s$ matrix where each row is a \underline{deep matched-filter} returning its correlation with $x$ for each output. }}

Proof can be found in~\cite{splineDN} for general feedforward networks with fully-connected, convolutional, pooling and/or residual layers and using any piecewise-linear activations. 
During training, MRFResnet encoder learns $\{W^{(i)}, \beta^{(i)}\}$ or equivalently $\{\A[x],\b[x]\}$ to provide a continuous and piece-wise affine \emph{approximation} for~\eqref{eq:proj}. 
The universal approximation theorem~\cite{cybenko1989} states that a shallow network with one but very wide hidden layer can do this.
 Deeper networks are however more practical to efficiently reduce the number of hidden units~\cite{shallowdeep-bengio}. Indeed, we experimentally observe this (section~\ref{sec:deepvskm}) by comparing MRFResnet to a shallow learning scheme related to~\cite{PERK} based on Kernel Machines (KM) and random features~\cite{fastfood}. 

\subsection{Visualising MRFResnet segments on Bloch manifold}

{\rem{\emph{Continuity of $z(x)$ implies that adjacent segments $\Omega_r,\Omega_{r'}$  correspond to distinct $A_r,A_{r'}$. 
Indeed, if $A_r=A_{r'}$ and the only difference is in the offsets $b_r\neq b_{r'}$, then $\Omega_r,\Omega_{r'}$ won't intersect on boundaries. Therefore they are not adjacent segments unless contradicting the continuity assumption.}}}

This remark gives an idea for visualizing the input space segments.  
For densely sampled input signals $x$, we compute derivatives of the weighted outputs~\eqref{eq:wo} with respect to inputs using \emph{back propagation}. These will determine the input-dependant slopes in the affine spline formulation \eqref{eq:spline} i.e. rows of $\A[x]$ at a point $x$ are populated as follows $\forall j=1,2,\ldots,p$:
\begin{align}
	\A[x]_{(j,.)}
	=\left[ \frac{\partial z_{j}(x)}{\partial x_1}, \frac{\partial z_{j}(x)}{\partial x_2},\ldots,\frac{\partial z_{j}(x)}{\partial x_s} \right]. \label{eq:grad}
\end{align} 
By vector quantisation (e.g. k-means clustering) we cluster regions of $x$ that output distinct slopes $A_r$  and identify the segments $\Omega_r$. Similar routine could apply to compute input space partitions by clustering back-propagated output derivates after each residual block (Theorem 1 and Remark 1 also hold for the intermediate blocks of $\Rr$). 

According to~\cite{splineDN} as we progress into deeper layers, partitions will be subdivided into smaller segments in a \emph{hierarchal} fashion. This can be observed in Figure~\ref{fig:seg} where we adopted the above routine for the T1/T2 encoding MRF sequence used in our experiments and visualised multi-scale (from coarse-to-fine) partitions obtained after each residual layer. 
The Bloch response manifold is sampled across fine-gridded T1/T2 values (i.e. MRF dictionary) to visualise the intersection of the input space segments with this manifold (results are visualised across the three dominant principal component axes). 
MRFResnet encoder learns about a thousand partitions for its end-to-end mapping $z(x)$. In the light of~\eqref{eq:spline} we know that for each partition $\Omega_r$ the network implicitly encodes $p=2$ deep matched-filters (the rows of $\A[x]$ or alternatively $A_r$) and an offset term to locally linearly regress the T1/T2 outputs in that segment. As such  instead of memorising $>$100K dictionary atoms used for training, the network learns a compact piece-wise affine approximation to the Bloch manifold projection~\eqref{eq:proj} as a rapid and memory-efficient alternative to DM's point-wise approximation~\eqref{eq:NNS}. 
The total number of parameters used by the MRFResnet (Table~\ref{tab:comp}) are two hundreds times less than the size of the dimension-reduced MRF dictionary. 
Figure~\ref{fig:matchfilt} shows the Bloch responses 
for a range of T1/T2 values, as well as deep matched-filters learned by MRFResnet to predict each of these quantities in this range from noisy inputs. 
Computed through \eqref{eq:grad}, match-filters are one-dimensional analogues of the \emph{saliency maps} a.k.a. \emph{deep dream images}~\cite{saliency}, measuring sensitivities of the T1/T2 output neurons with respect to the inputs. 

\section{Numerical experiments and discussions}
\label{sec:expe}
In the spirit of reproducible research, the source codes of
the proposed algorithms are available at \url{https://github.com/mgolbabaee/LRTV-MRFResnet-for-MRFingerprinting}.
\subsection{Datasets and 2D/3D acquisition parameters}
Methods are tested on the Brainweb \emph{in-silico} phantom (see supplementary materials), a EUROSPIN TO5 phantom (\emph{in-vitro})~\cite{eurospin}, and  
a healthy human brain (\emph{in-vivo}). 
\emph{In-vitro} and \emph{in-vivo} data were
acquired on a 1.5T GE HDxT scanner using 8-channel receive-only head RF coil. The novel adopted excitation sequence has $T=880$ repetitions and jointly encodes T1/T2 values using an inversion pulse followed by a flip angle schedule that linearly ramps up from $1^\circ$ to $70^\circ$ in repetitions 1-400, ramps down to $1^\circ$ in repetitions 400-600, and then stays constant to $1^\circ$ for repetitions 600-880 (see more details in~\cite{3DMRF_pedro}).
Three non-Cartesian readout trajectories were tested: 2D/3D variable density spiral and 2D radial k-space subsampling patterns.
Throughout we used inversion time=18 ms, fixed TR=12 ms, and TE = 0.46/2.08 ms for spiral/radial acquisitions, respectively. For the 2D/3D acquisitions we had $200^2/200^3$ (mm$^2$/mm$^3$) FOV and $200^2/200^3$ voxels image/tensor size, respectively. Further, the total number of interleaves for the 2D/3D spiral and 2D radial readouts were 377/48'400 and 967, respectively. Only one radial spoke (or spiral arm) was sampled at each of the 880 timeframes, resulting in aggressive acceleration (undersampling) factors  62$\times$, 252$\times$
and 262$\times$ with respect to a fully sampled 2D spiral, 2D radial and 3D spiral acquisitions, correspondingly.  
The total acquisition times for the 2D and 3D scans were 10:56 seconds and 9:51 minutes, respectively. For all cases coil sensitivities were computed from undersampled data using an adaptive coil combination scheme~\cite{adaptivesense}.


\subsection{Tested algorithms}
\label{sec:algosetup}
For TSMI reconstruction we compare model-based iterative methods: LRTV, LR~\cite{zhao-LR-MRF}  and FLOR~\cite{Eldar-FLOR} which are convex and DM-free, and the non-convex DM-based AIR-MRF~\cite{AIRMRF}.
LRTV and LR solve~\eqref{eq:optim} with  spatiotemporal ($\la>0$) and temporal-only ($\la=0$) regularisations, respectively. 
Note that the latter  
solves the same problem as~\cite{zhao-LR-MRF} but with Nesterov acceleration. Here, AIR-MRF uses Gaussian low-pass filtering (to avoid Gibbs artefacts) to regularise the spatial domain. It further uses MATLAB's in-build kd-tree searches for fast DM per-iteration. 
The maximum number of iterations for these algorithms were set to 30, but each stops earlier if the relative change in their objectives is below tolerance $10^{-4}$ (convergence).
Further we compare against reconstruction non-iterative baselines ZF~\eqref{eq:zf}~\cite{SVDMRF} and ViewSharing (VS)~\cite{VSguido}. VS aggregates spatial k-space data within neighbouring temporal frames to increase per-frame samples and enhance spatial resolutions in a non model-based fashion. 
For quantitative inference and besides the baseline DM, we compare learned models MRFResnet (deep learning) and a shallow learning method based on Gaussian Kernel Machines (KM) related to~\cite{PERK}. \footnote{Used hyperparameters: KM used optimised kernel scales by the MATLAB's $\mathrm{fitrkernel}$ function and 1000/500 random features~\cite{fastfood} per output index for the encoder/decoder models, respectively. VS used 880 shared views as in~\cite{3DMRF_pedro}. LRTV used the TV norm weighting $\forall i, \la_i=\la=0.2/0.04$ for the 2D/3D brain scans, respectively. AIRMRF used 2D/3D gaussian filters with spreads $\sigma=1/0.5$ for the 2D/3D brain scans. FLOR used the nuclear norm weighting $\la=10$ for 2D brain scans. Parameters were adjusted experimentally according to the visual impression of the results and where ground-truth was available (e.g. for the phantom or retrospective experiments) they were grid-searched to minimise the reconstruction errors.}.

%
%
%

\subsubsection{Learned models} Except FLOR which does not use dimensionality reduction, other methods above use a $s=10$ dimensional subspace model a-priori learned from Bloch response simulations using PCA. For this, a dictionary of $d = 113781$ atoms sampling the T1=[100:10:4000] (ms) and T2=[20:2:600] (ms) grid was simulated using the Extended Phase Graph formalism~\cite{EPGWeigel}. The subspace-compressed dictionary was directly used in DM and FGM, whereas for learning-based inference it was only used for training. Clean fingerprints were used for training MRFResnet decoder $\Gg$ i.e. Bloch response generation network. Noisy fingerprints each corrupted with i.i.d. noise $\sim \Nn(0,0.01)$ were used to train the MRFResnet encoder $\Rr$. 
We created fifty noisy realisation of each fingerprint (i.e. data augmentation), and for each we performed dictionary search to find correct T1/T2 (closest match) training labels 
and not those that originally generated the fingerprints. Noisy data augmentation creates a tube around the finely-sampled Bloch response manifold, and the label-search procedure enables learning a Euclidean \emph{projection} mapping (rather than a possible overfitted denoiser) onto this manifold. 
Trainings used Adam optimiser with MSE loss for 20 epochs, 0.01 initial learning rate with decay factors 0.8/0.95 and mini-batch sizes 500/20 for $\Rr$ and $\Gg$, respectively. The same datasets were used for training KM's encoder and decoder models using LBFGS optimiser.
 
 \begin{table}[t!]
	\centering
	\scalebox{.76}{
		\begin{tabular}
		{p{1.5cm}p{1cm}|p{1.2cm}p{1.1cm}|p{1.2cm}p{1.1cm}|p{1.1cm}}
			\toprule[0.2em]
			& {Total \# params.} & {T1 (ms) MAE} &  {T1 (\%) MAPE} & { T2 (ms) MAE} & { T2 (\%) MAPE} &$\Bb$ (\%) NRMSE \\
			\midrule[0.1em]
			\midrule[0.1em]
			MRFResnet   & 5.2K & 7.19 & 0.91 & 1.91 & 1.05 & 0.86 \\
			\midrule[0.05em]
			KM fitting &  44K & 29.93 & 3.97 & 15.84 & 9.68 & 12.88 \\
			\bottomrule[0.2em]
		\end{tabular}}
		\caption{\footnotesize{Prediction performances of MRFResnet and KM.\vspace{.2cm}}} \label{tab:comp}	
\end{table}

\subsection{Deep vs. shallow learning models' prediction results}
\label{sec:deepvskm}
To compare the prediction performances of the learned models MRFResnet and KM, 500K out-of-sample noisy fingerprints 
were randomly generated and fed to the corresponding encoder models to estimate the T1/T2 parameters. Predicted T1/T2s were then fed to the decoder models for generating the corresponding noise-free Bloch responses. The ground-truth (GT) T1/T2s from DM were used to measure encoders'  performances based on Mean Absolute (Percentage) Errors MAE $=\EE [ |\widehat T1 -  T1^{\text{\tiny GT}}| ]$ and  MAPE $=\EE [ \frac{| \widehat T1 -  T1^{\text{\tiny GT}}|}{T1^{\text{\tiny GT}}}]$ (similarly for T2). Corresponding clean fingerprints were used as GT to measure generative model (decoder) predictions based on Normalised-RMSE $=\EE\frac{\| \Gg(\widehat \Theta)-\Bb(\Theta^{\text{\tiny GT}})\|} {\|\Bb(\Theta^{\text{\tiny GT}})\|}$. 
Table~\ref{tab:comp} summarises our results. 

\vspace{-.2cm}	
\subsubsection{Discussion} 
MRFResnet outperforms KM and achieves reliable predictions for T1/T2 values and Bloch response generation, with about or less than $1\%$ average difference with the DM baseline. KM reports poor T2 and Bloch response estimations for the  number of random features used. By increasing this number KM's expressive capacity can improve (e.g. we observed by doubling random features KM's T2 error reduces to about $7\%$), however by comparing both model sizes at their current configurations, we can deduce the advantage of \emph{depth} in MRFResnet 
to embed DM more efficiently compared to its shallow alternative KM for the adopted acquisition sequence.


\vspace{-.3cm}
\subsection{\textit{In-vitro} phantom experiment}
\label{seq:vitroexpe}
\begin{figure}[t!]
	\centering
	\begin{minipage}{\linewidth}
		\centering
		{\includegraphics[width=.49\linewidth]{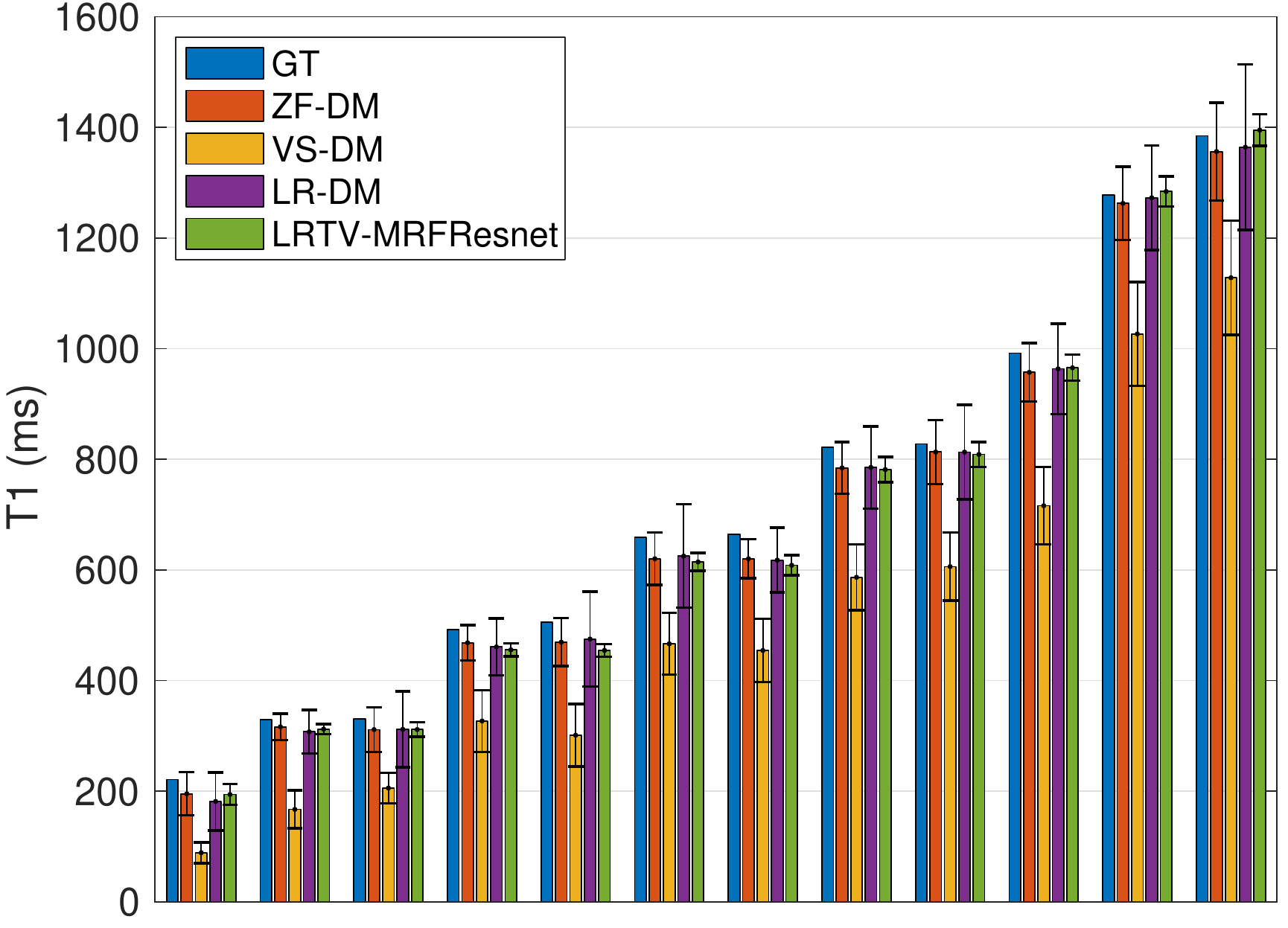} 	
		\includegraphics[width=.49\linewidth]{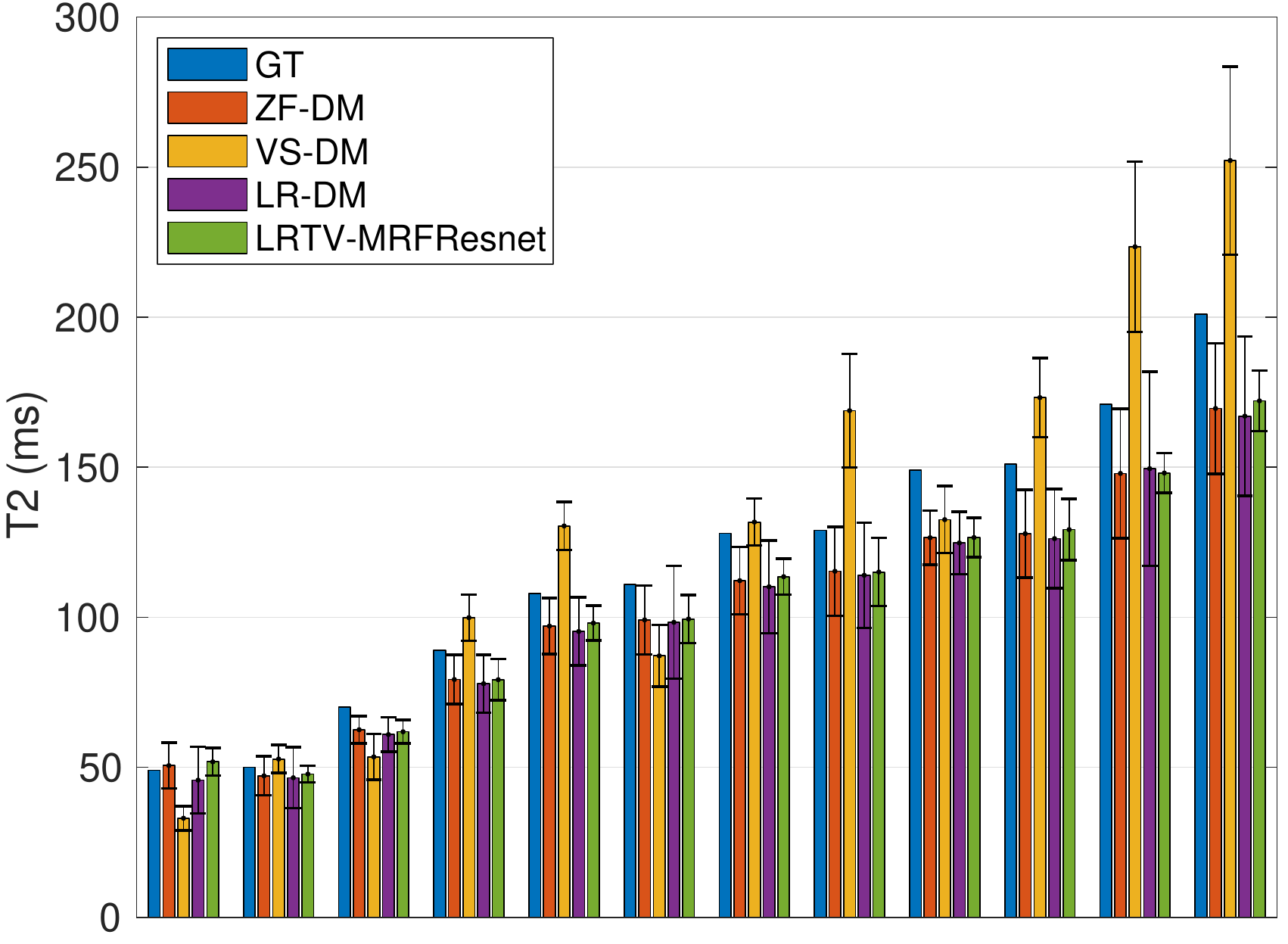} }
		\\
		{\includegraphics[width=.49\linewidth]{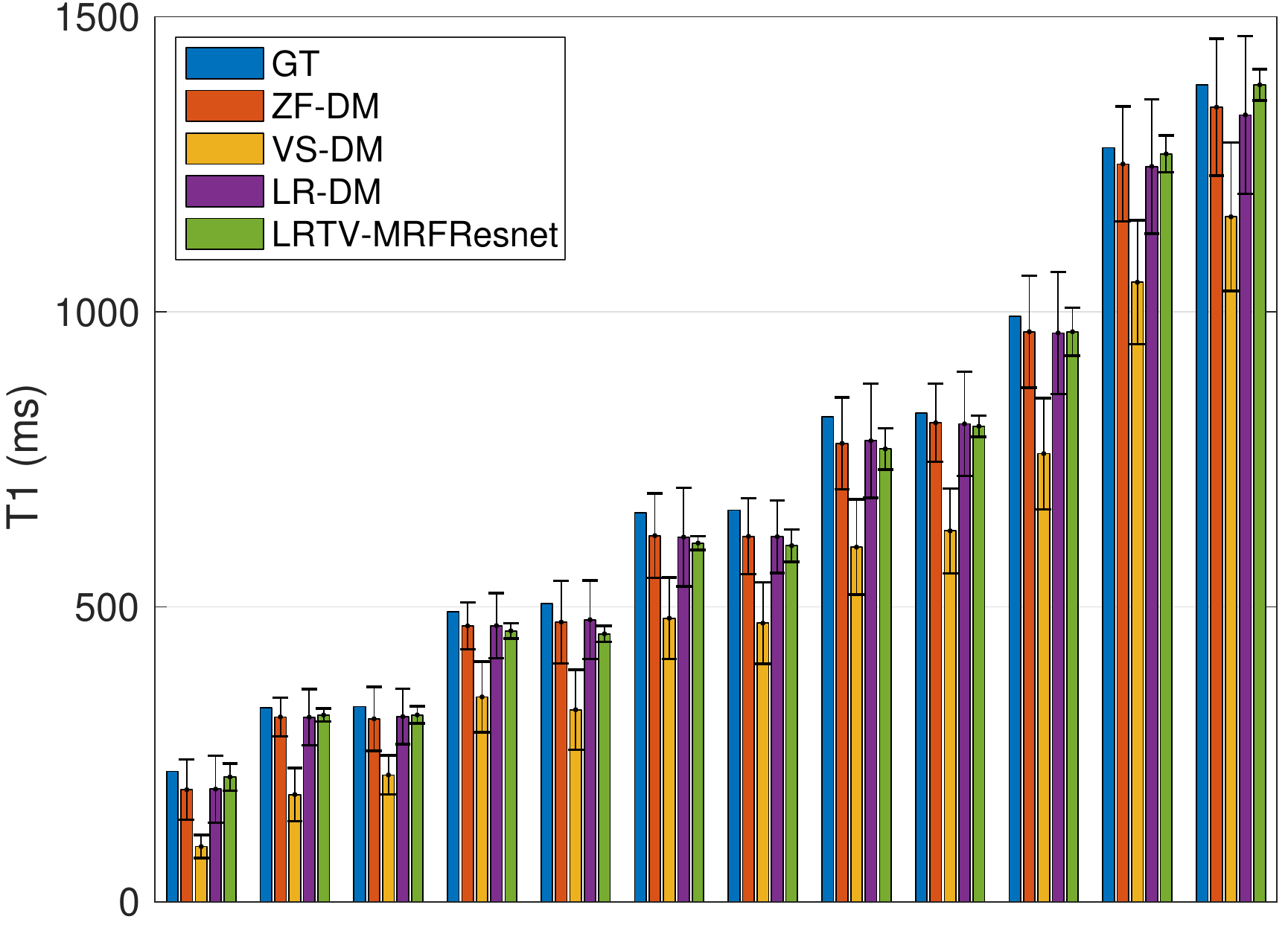} 	
		\includegraphics[width=.49\linewidth]{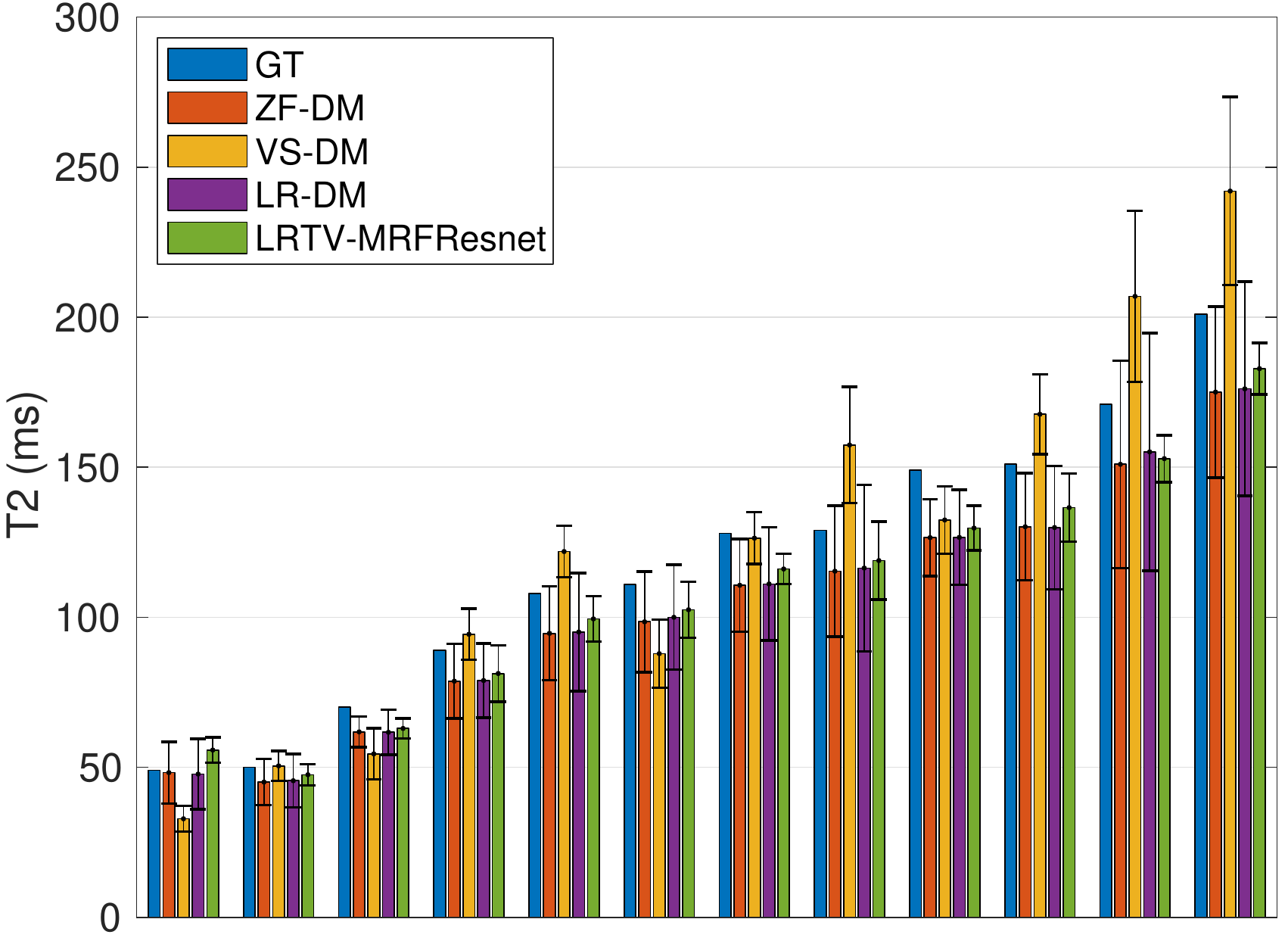} }
		\\
		{\includegraphics[width=.49\linewidth]{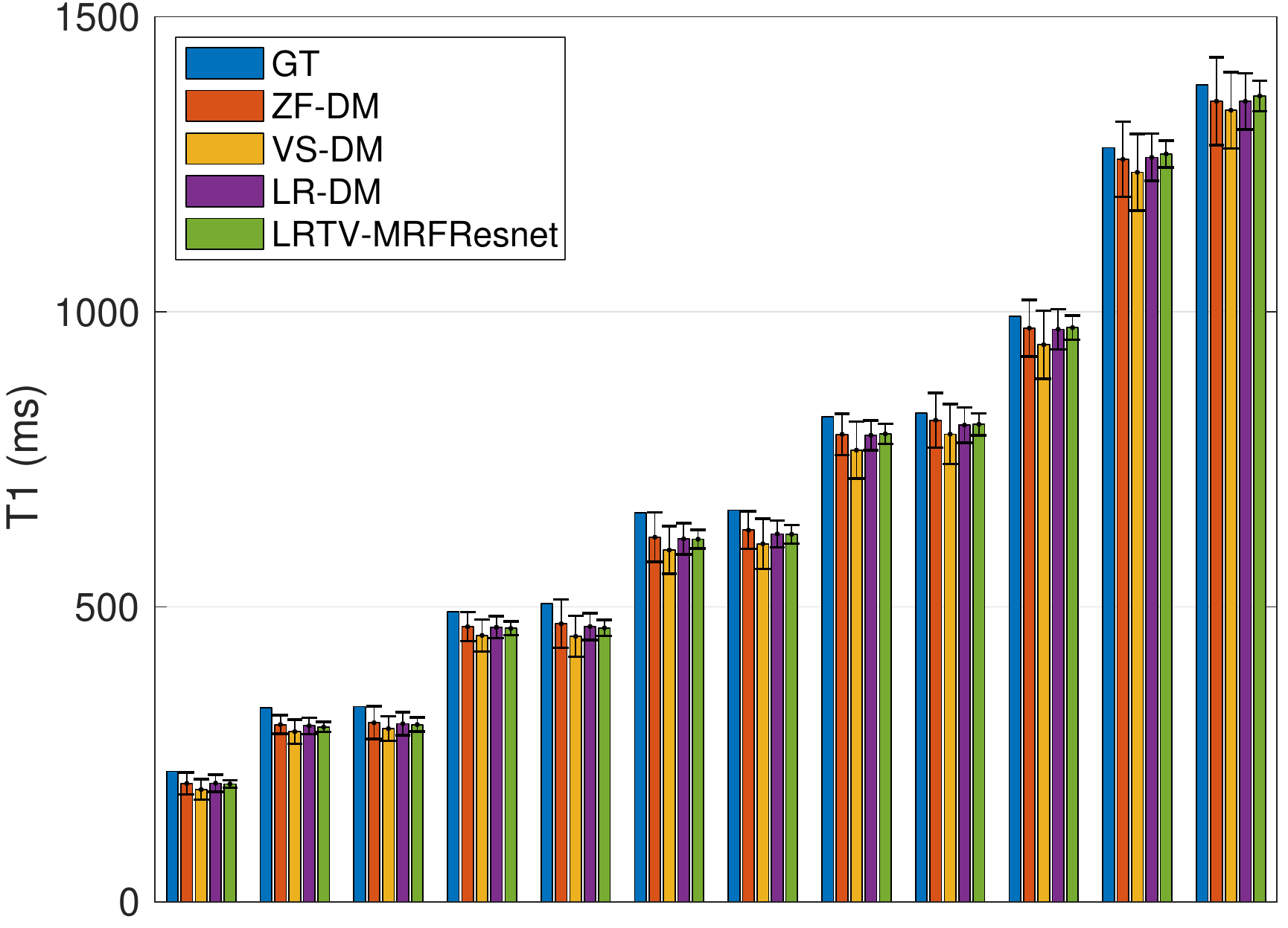} 
		\includegraphics[width=.49\linewidth]{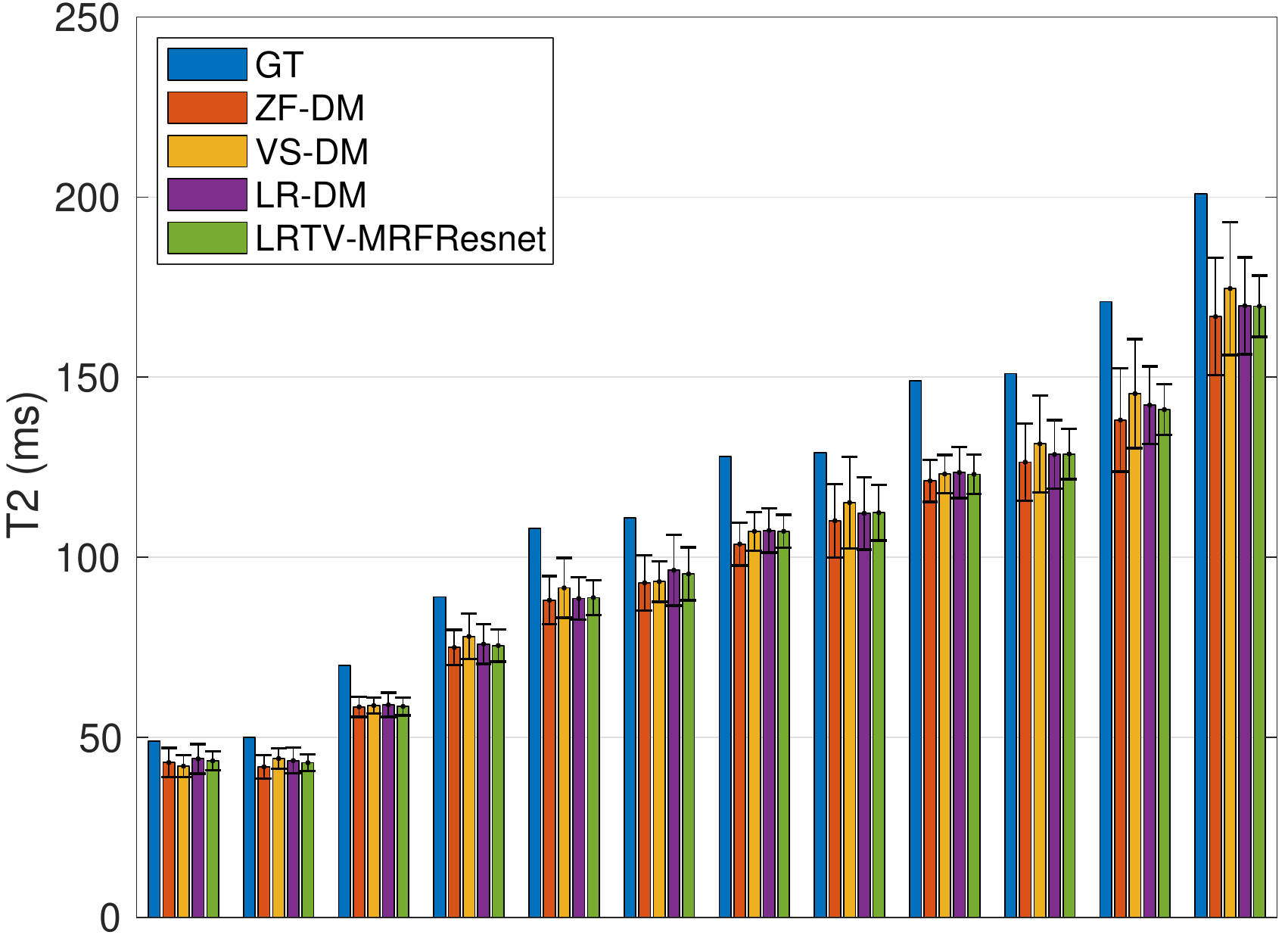} }	
		\caption{\footnotesize{The mean T1 (left column) and T2 (right column) values in milliseconds and their standard deviations (error bars) estimated via using four reconstruction methods compared to the reference values (GT)  in 12 phantom ROIs. Results are compared for 2D spiral (top row), 2D radial (middle row) and 3D spiral acquisitions (bottom row).} \label{fig:phanbars}}
	\end{minipage}
\end{figure}

The 2D (spiral/radial) and 3D (spiral) acquisition schemes were tested for measuring quantitative parameters in twelve tubes of the EUROSPIN TO5 phantom. 
Table~\ref{tab:comp2} shows the MAPE errors averaged over all ROIs (tubes) for predicting the T1/T2 values using different DM-free reconstructions algorithms ZF, LR, VS and the proposed LRTV, all cascaded to DM or the proposed MRFResent for quantitative inference. 
The gold standard T1/T2 values reported in the phantom’s manual were used as reference. 
DM and MRFResent score very similar quantitative inference accuracies regardless of the reconstruction scheme, which shows that the MRFResnet accurately embeds DM. In addition, Figure~\ref{fig:phanbars} displays the mean and standard deviation of the predicted T1/T2 values in each ROI.  For compactness we only display ZF-DM, LR-DM and VS-DM baselines, and the proposed LRTV-MRFResnet predictions. 
%
Computed parameter map images are also shown in the supplementary materials (Figure S4). 
Figure~\ref{fig:ci} displays the Bland-Altman plots of the percentage differences between T1/T2 values of the phantom ROIs in spiral and radial scans, estimated using the ZF-DM and LRTV-MRFResnet.

\begin{table}[t!]
	\centering
	\scalebox{.83}{
		\begin{tabular}
		{c|cc|cc|cc}
			\toprule[0.2em]
			\multicolumn{1}{c}{ }& \multicolumn{2}{c}{2D spiral} &  \multicolumn{2}{c}{2D radial} &  \multicolumn{2}{c}{3D spiral}  \\
			\midrule[0.05em]
			MAPE (\%) & T1&T2&  T1&T2 &  T1&T2
			\\
			\midrule[0.1em]
			\midrule[0.1em]
			ZF-DM &   7.36&12.80 & 9.93 & 15.46 & 6.69&16.29\\
			*-MRFResnet  &7.49 &13.42 & 9.89 & 15.79 & 6.54& 16.88 \\
			\midrule[0.15em]
			VS-DM &31.46& 19.53 & 28.54 & 16.04 & 8.47 & \textbf{14.43} \\
			*-MRFResnet   &  33.65&19.85 & 30.67 & 16.48 & 8.53& 14.90 \\
			\midrule[0.15em]
			LR-DM 		   &  	10.65 &15.99  & 10.75& 16.67 & 6.18 & 15.04\\
			*-MRFResnet   & 11.15& 16.27 & 10.94 & 17.05 & 5.99 & 15.25\\
			\midrule[0.15em]
			LRTV-DM  & \textbf{5.60} & \textbf{11.44} & \textbf{5.54} & \textbf{9.76} & \textbf{5.87} & \textbf{15.29}  \\
			*-MRFResnet   & \textbf{6.03} & \textbf{11.96} & \textbf{5.85} & \textbf{10.69} & \textbf{5.68} & 15.44 \\
			\bottomrule[0.2em]
		\end{tabular}}
		\caption{Average errors for predicting the T1/T2 values of the \emph{in-vitro} phantom ROIs using DM-free reconstruction algorithms ZF, VS, LR and LRTV (ours), followed by DM and MRFResnet (ours) for quantitative inference.\vspace{.3cm}} \label{tab:comp2}
	\end{table}

\begin{figure}[t!]
	\centering
	\begin{minipage}{\linewidth}
		\centering
		{\includegraphics[width=.47\linewidth]{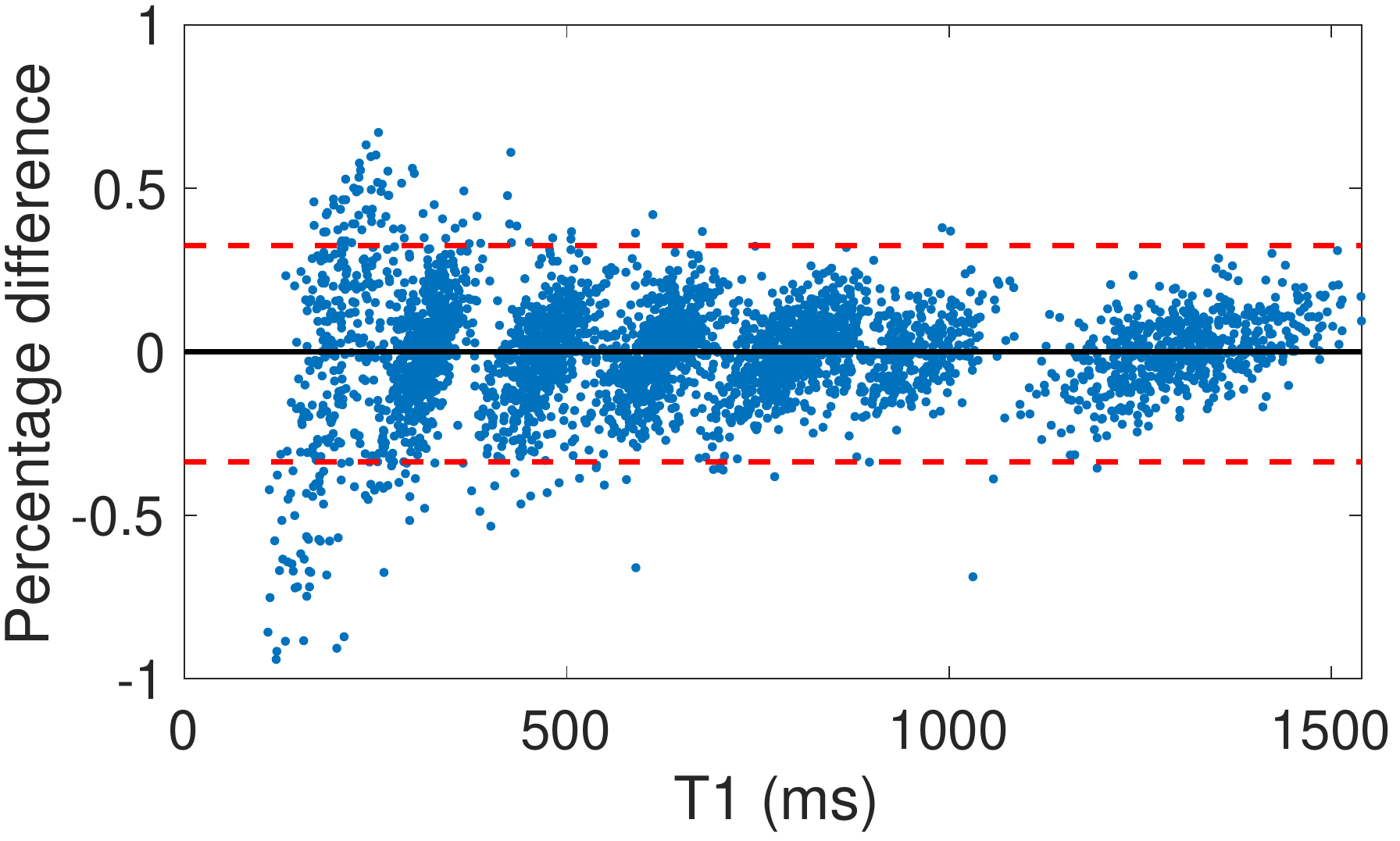} 
		\,
		\includegraphics[width=.47\linewidth]{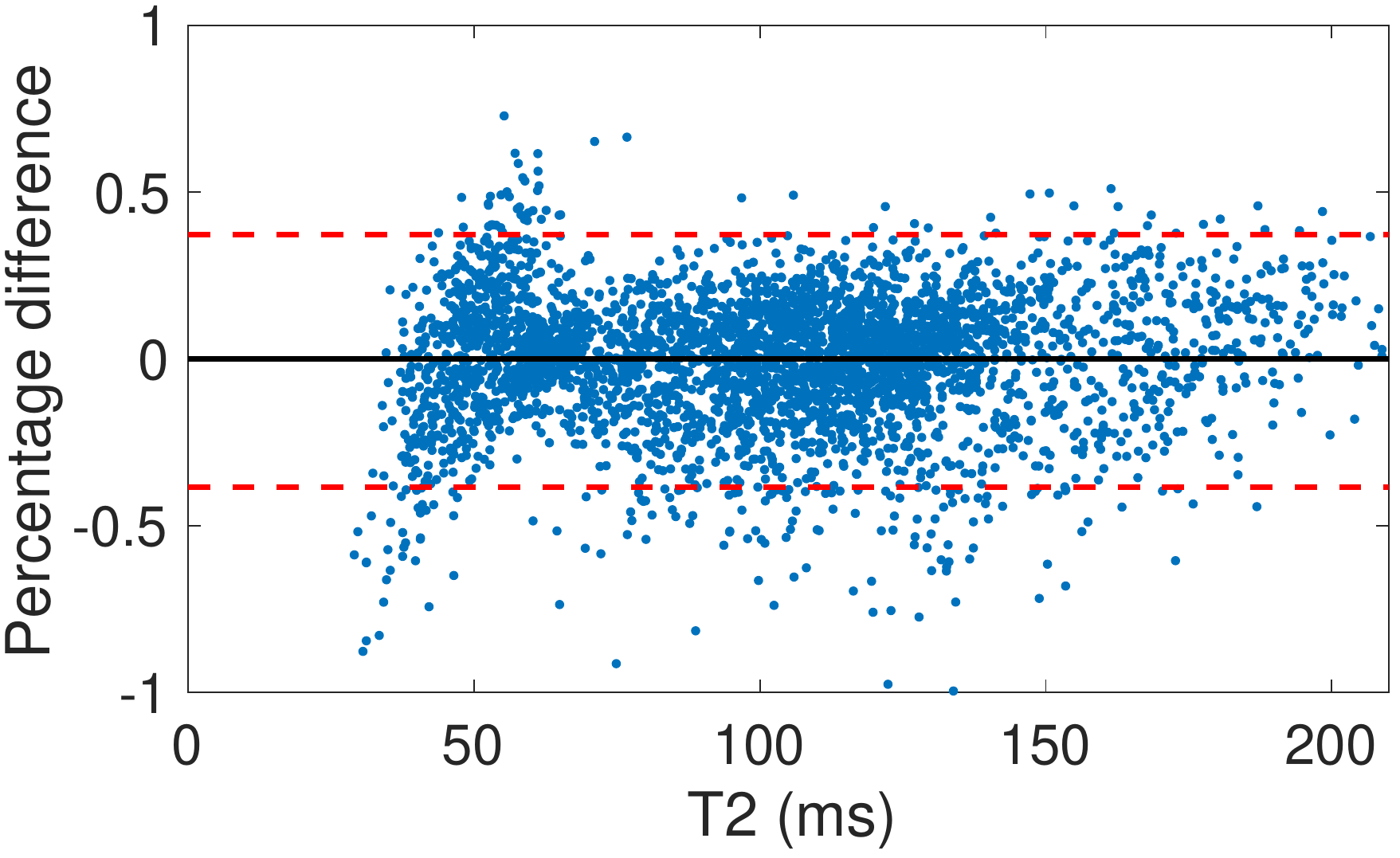} }
		\\
		{\includegraphics[width=.47\linewidth]{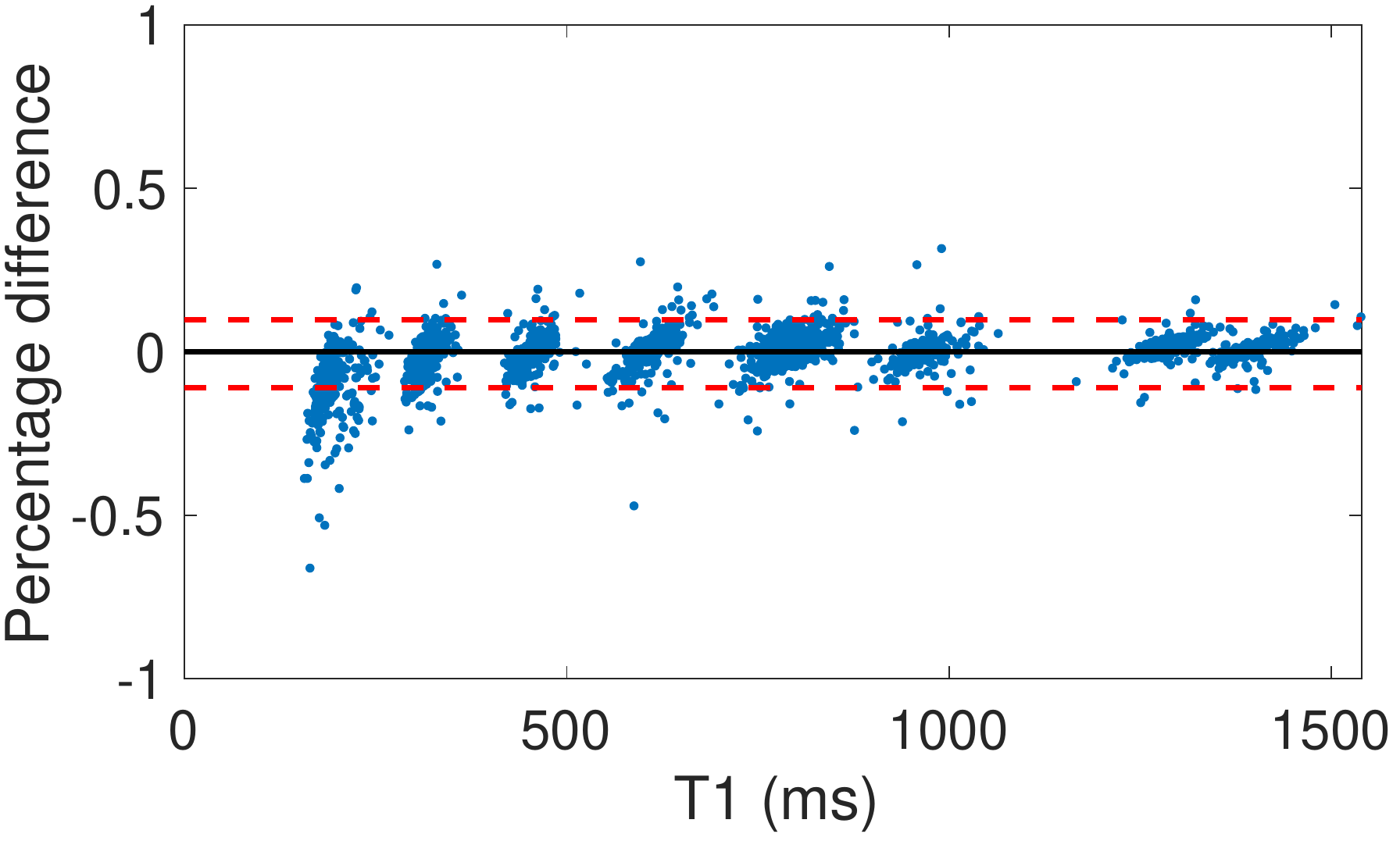} 
		\,
		\includegraphics[width=.47\linewidth]{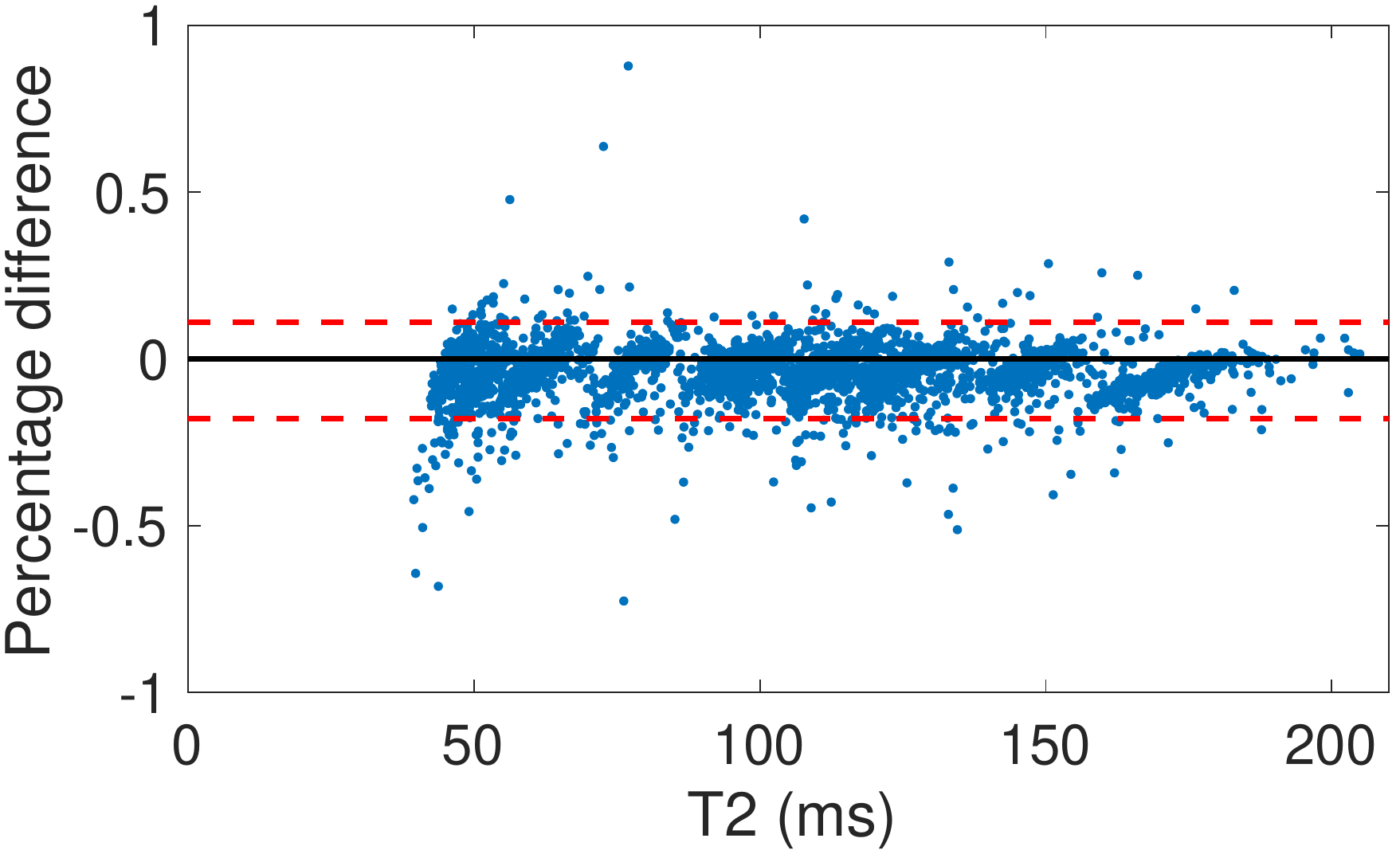} }	
		\caption{\footnotesize{Confidence intervals (CIs) for the percentage differences between predicted T1/T2 values of the phantom ROIs in 2D spiral and radial scans, using ZF-DM (top) and the LRTV-MRFResnet (bottom) reconstructions. The 0.95\%-CIs are -33.9\% to 32.3\% for T1 and -38.5\% to 37.4\% for T2 using ZF, whereas using LRTV CIs are tighter -10.2\% to 7.8\% for T1 and -17.0\% to 10.4\%  for T2.
		} \label{fig:ci}}
	\end{minipage}
\end{figure}

\subsubsection{Discussion} 

From Figure~\ref{fig:phanbars} we observe that 
tested methods (except VS for 2D) report comparable predictions for the mean T1/T2 values in each ROI.\footnote{We observe VS generally trades off image smoothness against overestimated T2s and underestimated T1s. This compromise is strongly unfavourable in 2D acquisitions. Larger k-space neighbourhood  information was available/shared in 3D (than 2D) acquisitions, which made 3D VS competitive.} T1 values are comparable to the GT (although ZF, LR and VS  slightly underestimate T1).
The predicted T2 values, especially in high T2 regimes, are under-estimated (negative bias). We hypothesise this is due to physical effects e.g. flip angle calibration errors, diffusion or magnetisation transfer that are currently un-modelled in the reconstruction schemes. 
Overall, the proposed LRTV predicts least biased T1/T2s. 
Notably, LRTV has the \emph{least variations} around the estimated values  (see the error bars in Figure~\ref{fig:phanbars}). For all experiments and averaged over all ROIs, the LRTV's standard deviation is $1.5/2.5$ times less than its closest competitor for predicting T1/T2, respectively.  This leads the LRTV to have the least (or close to least) predication errors in all tested acquisitions (see Table~\ref{tab:comp2}). 
From Figures~\ref{fig:ci} and images shown in Figure~S4 
we observe that LRTV \emph{enables highly consistent predictions} across radial and spiral sampling protocols i.e. per-pixel estimated T1/T2 values in all ROIs obtained from 2D radial and spiral measurements are  2 to 3 times more consistent with each other than those computed via ZF. 
We did not observed similar consistency level in other tested algorithms as the readout-dependent undersampling artefacts in images were not fully removed by them. 

\begin{figure}[t!]
	\centering
	\scalebox{.95}{
	\begin{minipage}{\linewidth}
		\centering	
		\begin{turn}{90} \quad\quad ZF-DM\end{turn}
		\includegraphics[width=.3\linewidth]{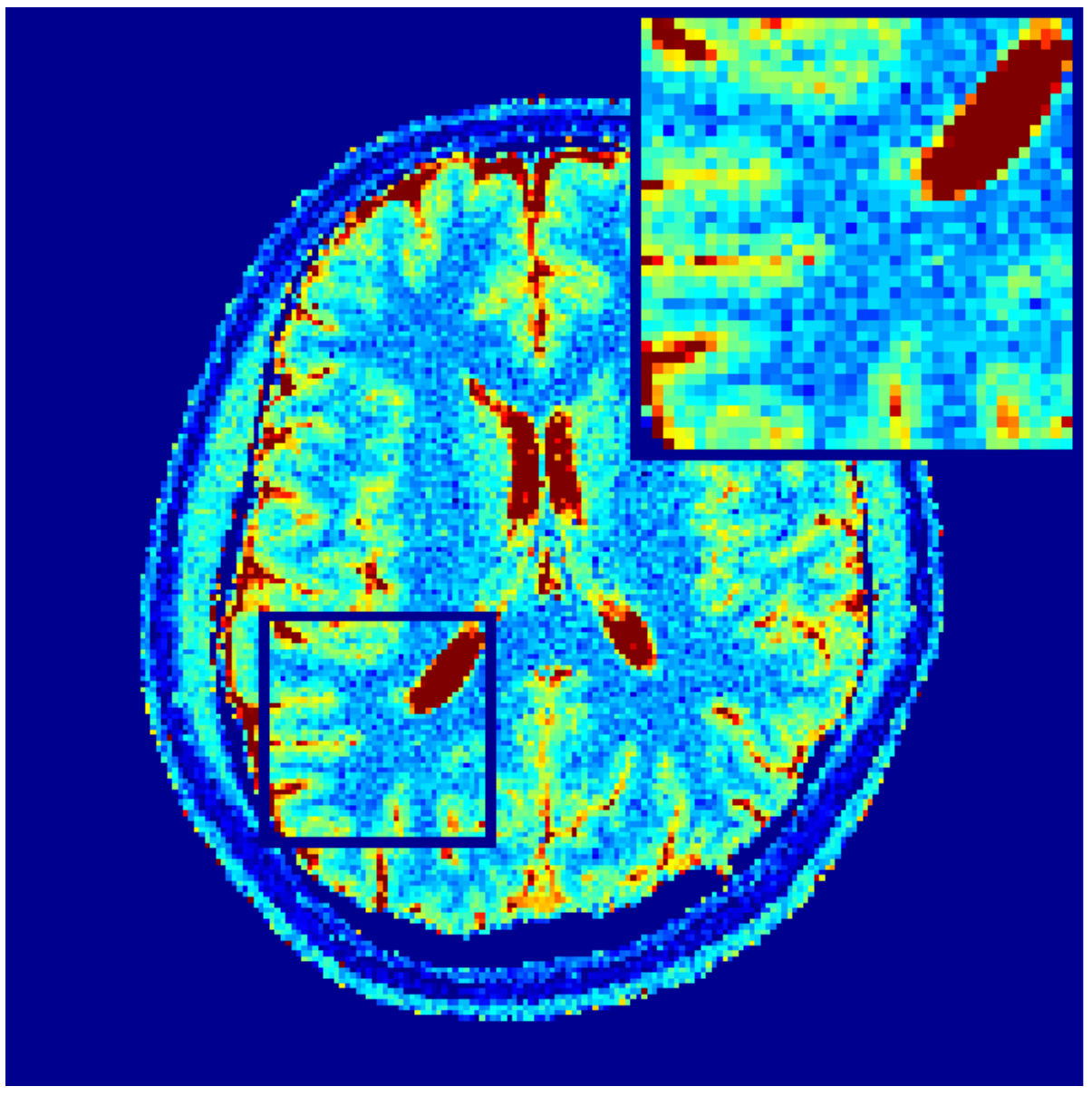}\hspace{-.1cm}
		\includegraphics[width=.3\linewidth]{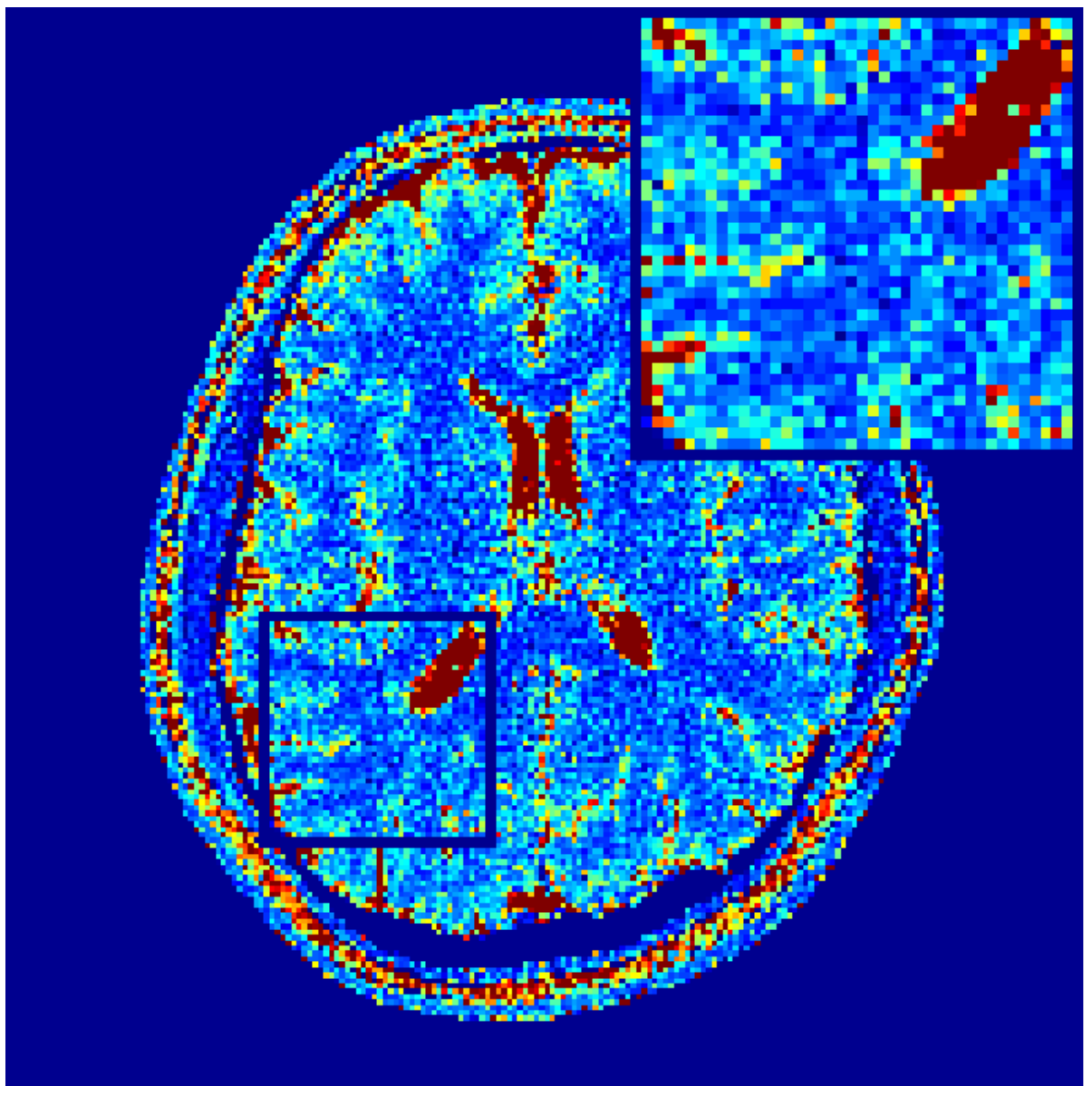}\hspace{-.1cm}
		\includegraphics[width=.3\linewidth]{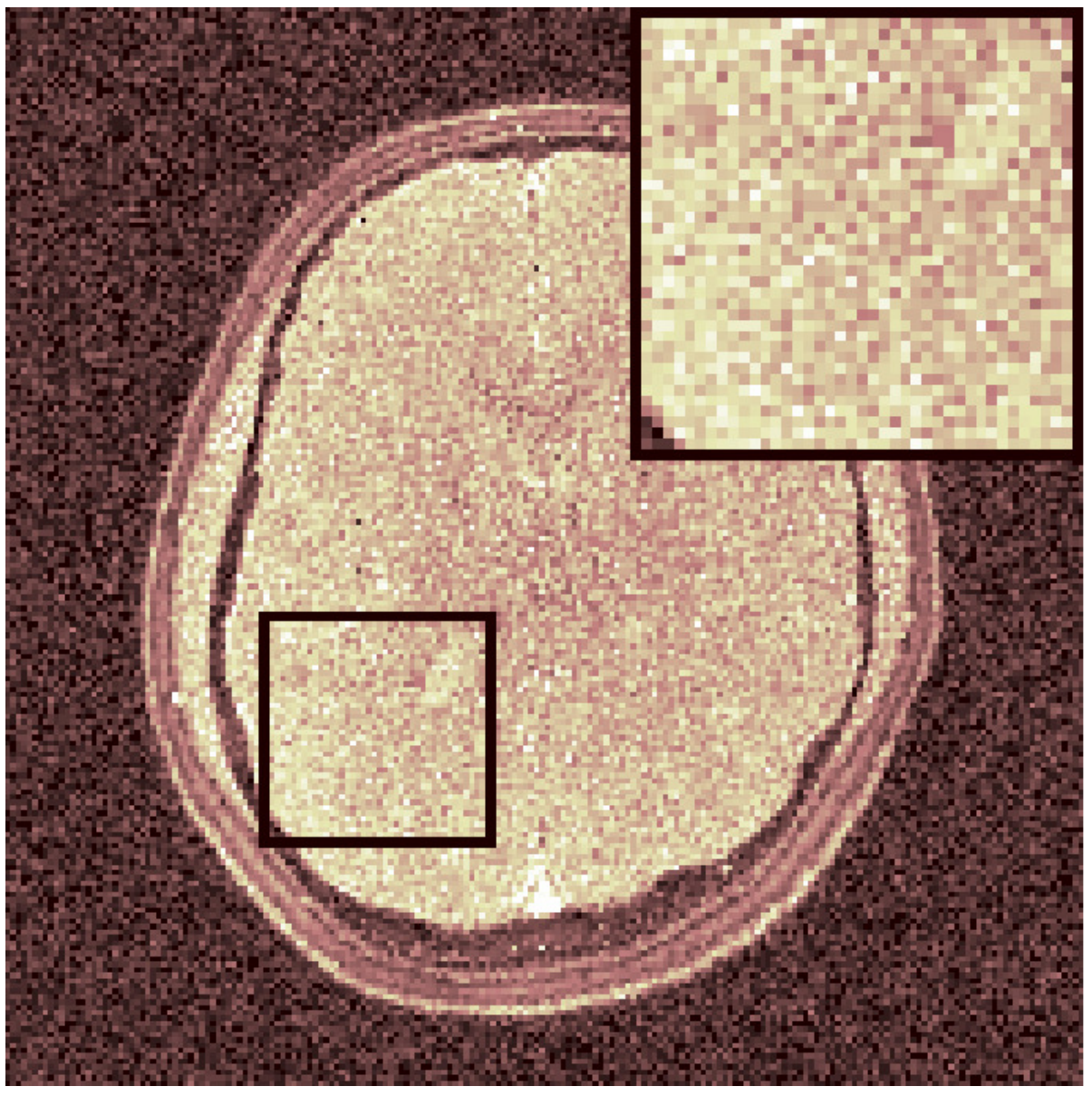}
		\\
		\begin{turn}{90} \quad\quad LR-DM\end{turn}
		\includegraphics[width=.3\linewidth]{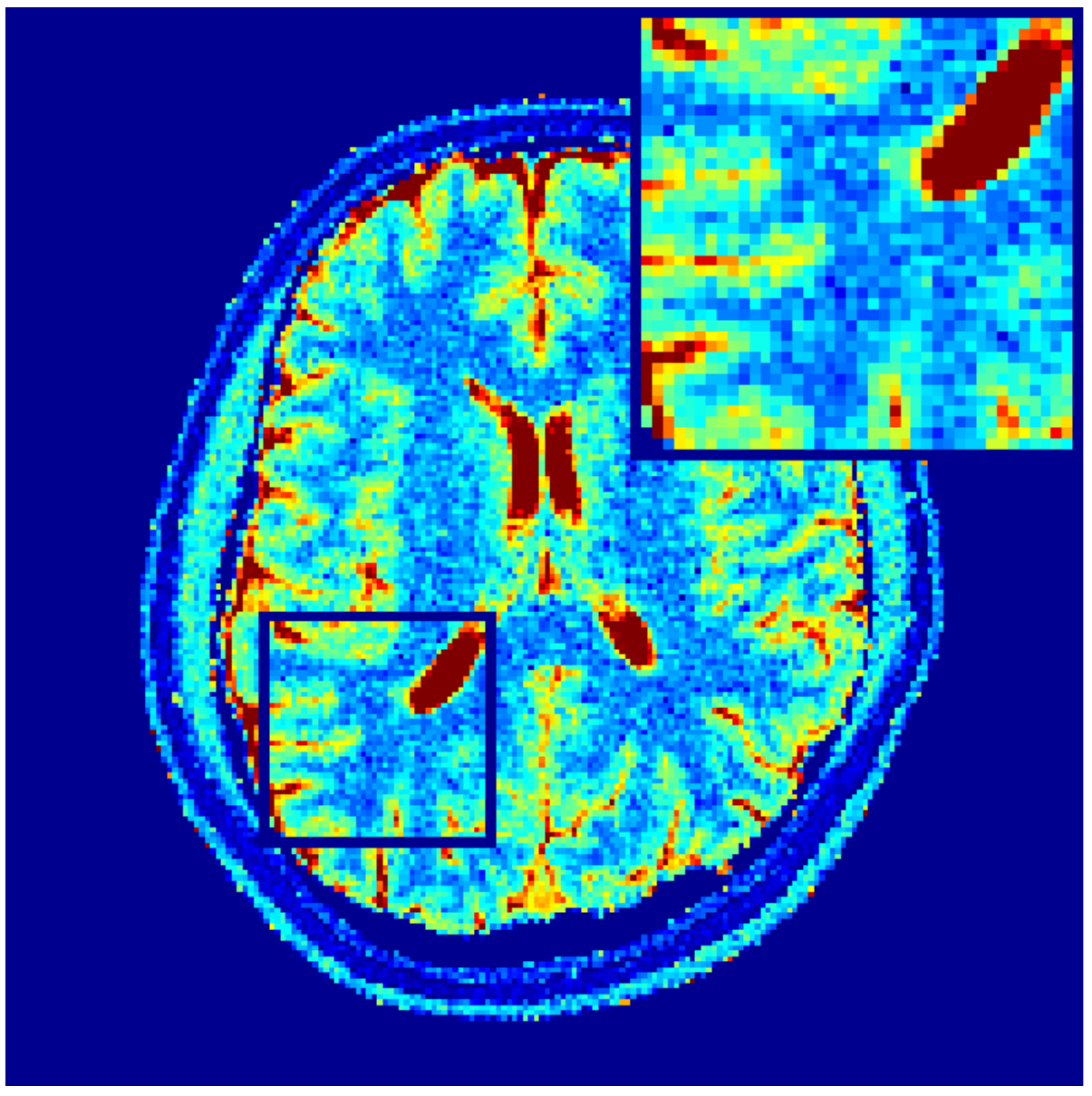}\hspace{-.1cm}
		\includegraphics[width=.3\linewidth]{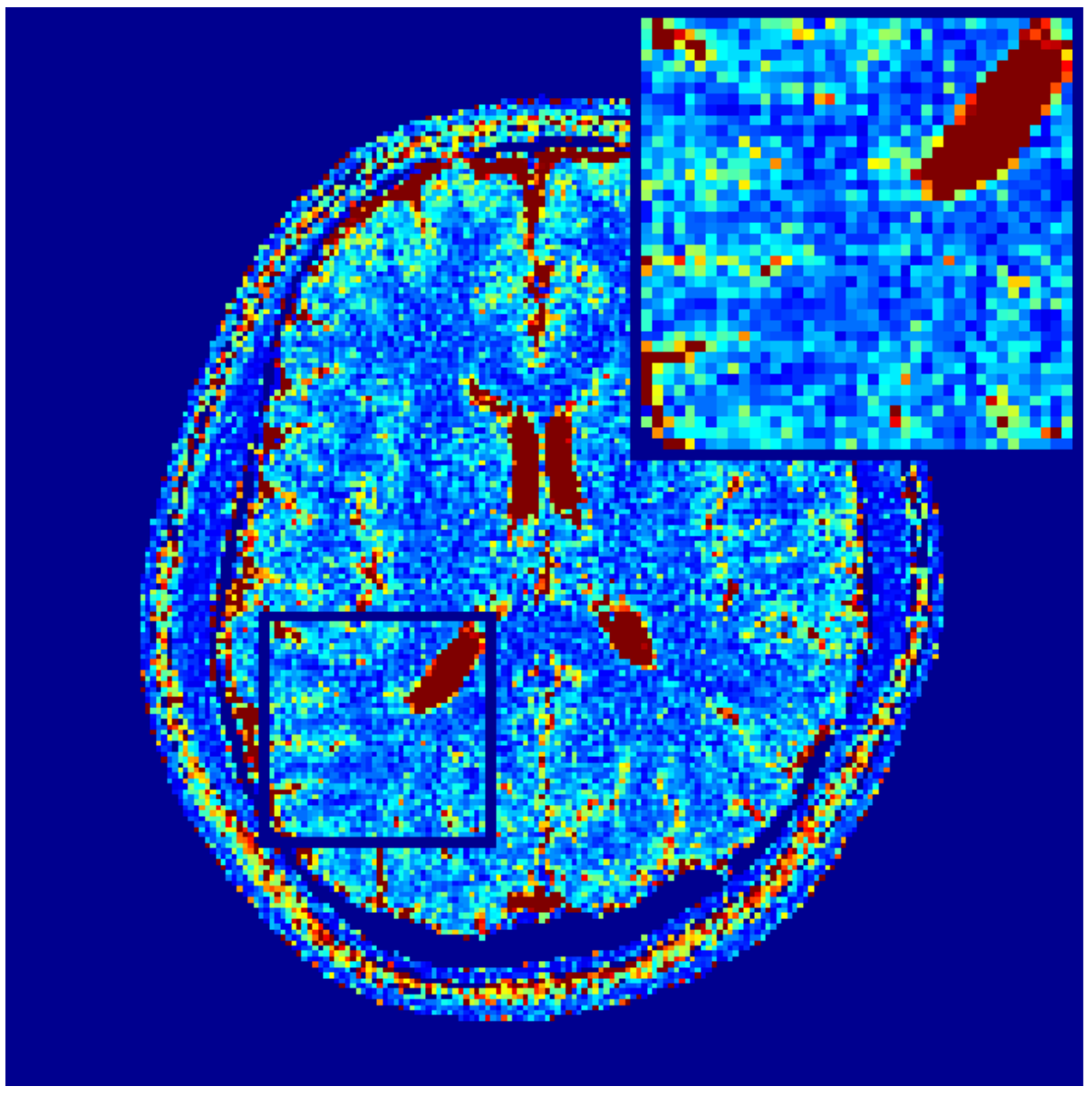}\hspace{-.1cm}
		\includegraphics[width=.3\linewidth]{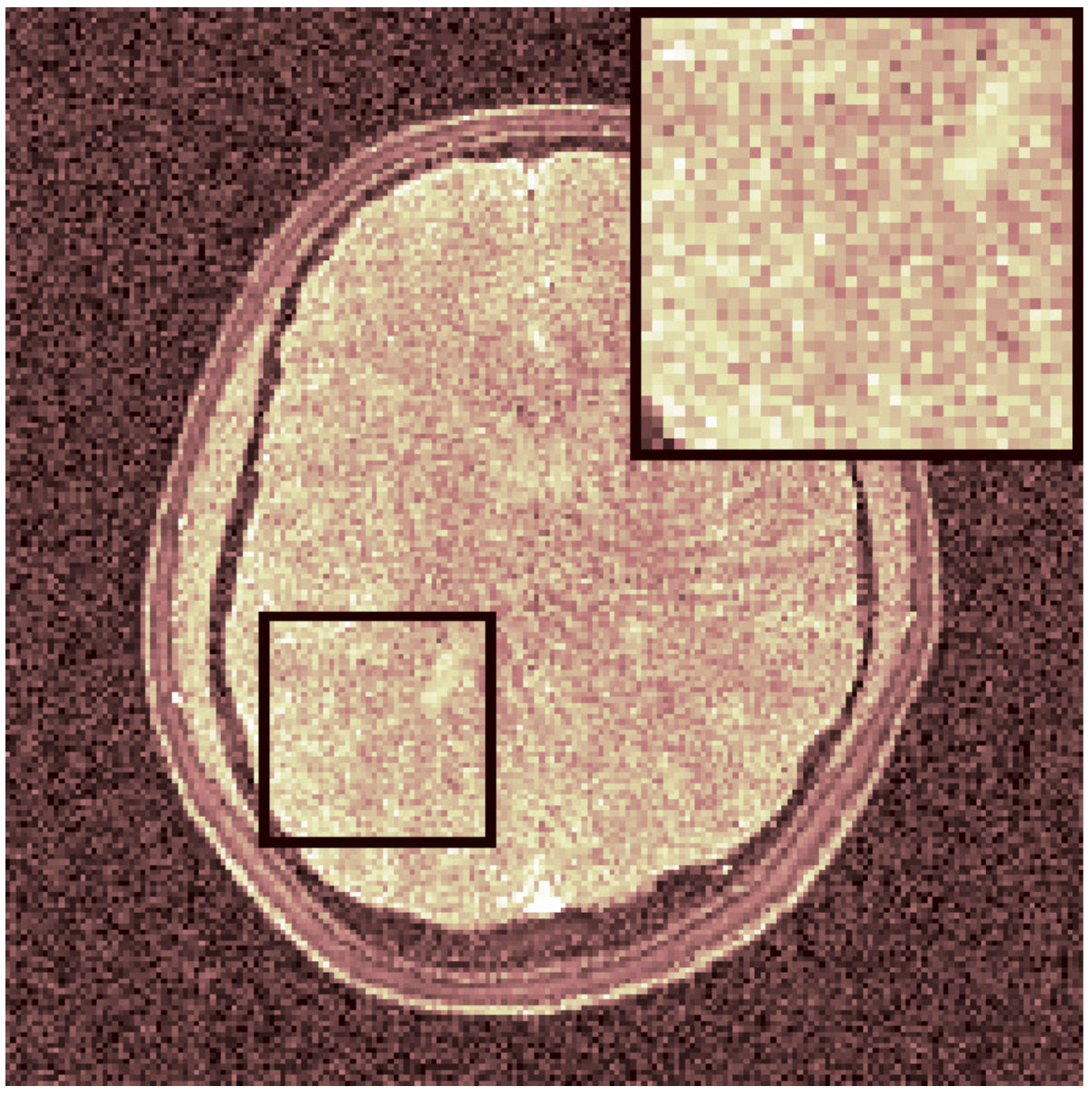}
		\\	
		\begin{turn}{90} \quad\quad VS-DM\end{turn}
		\includegraphics[width=.3\linewidth]{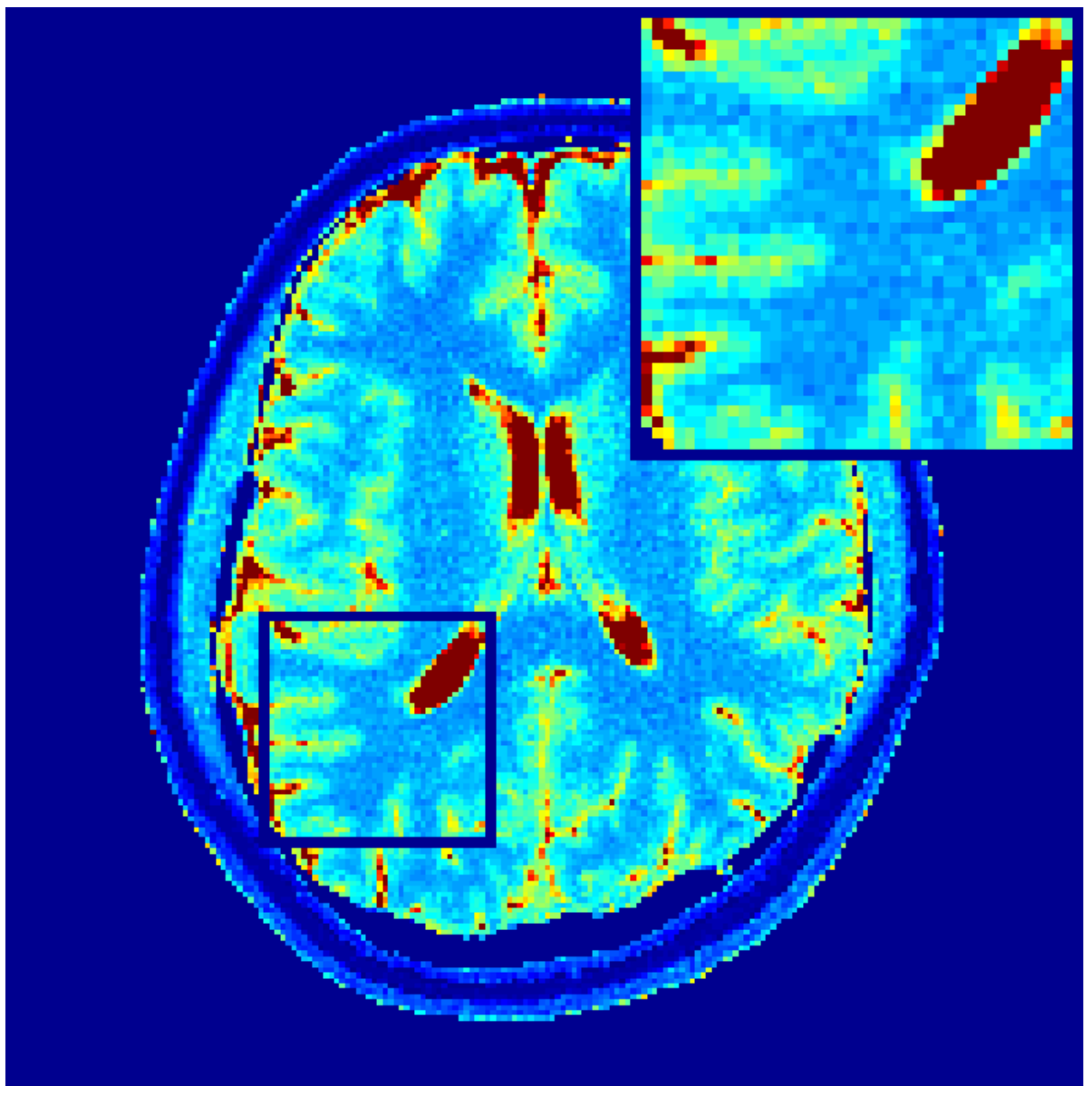}\hspace{-.1cm}
		\includegraphics[width=.3\linewidth]{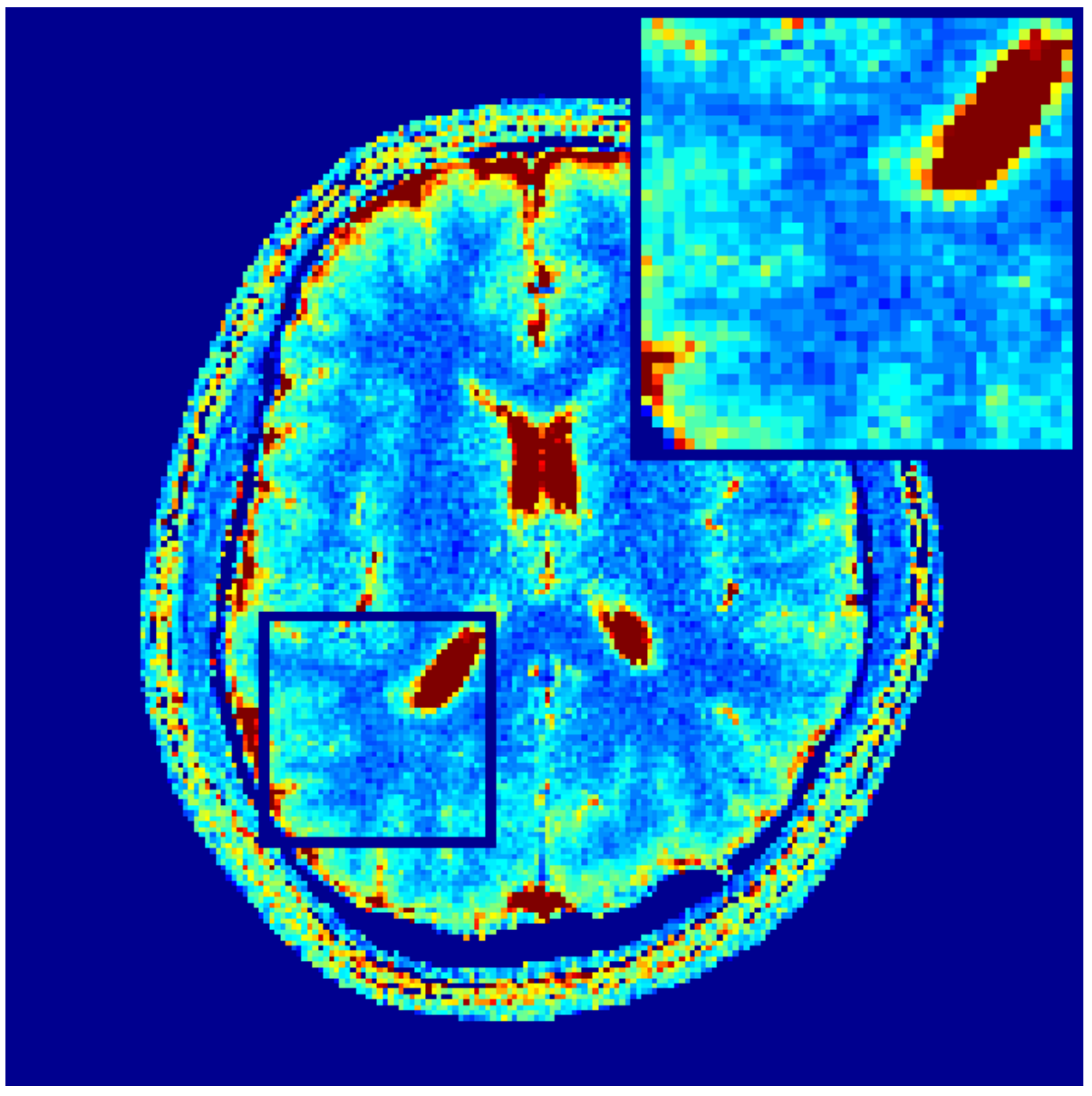}\hspace{-.1cm}
		\includegraphics[width=.3\linewidth]{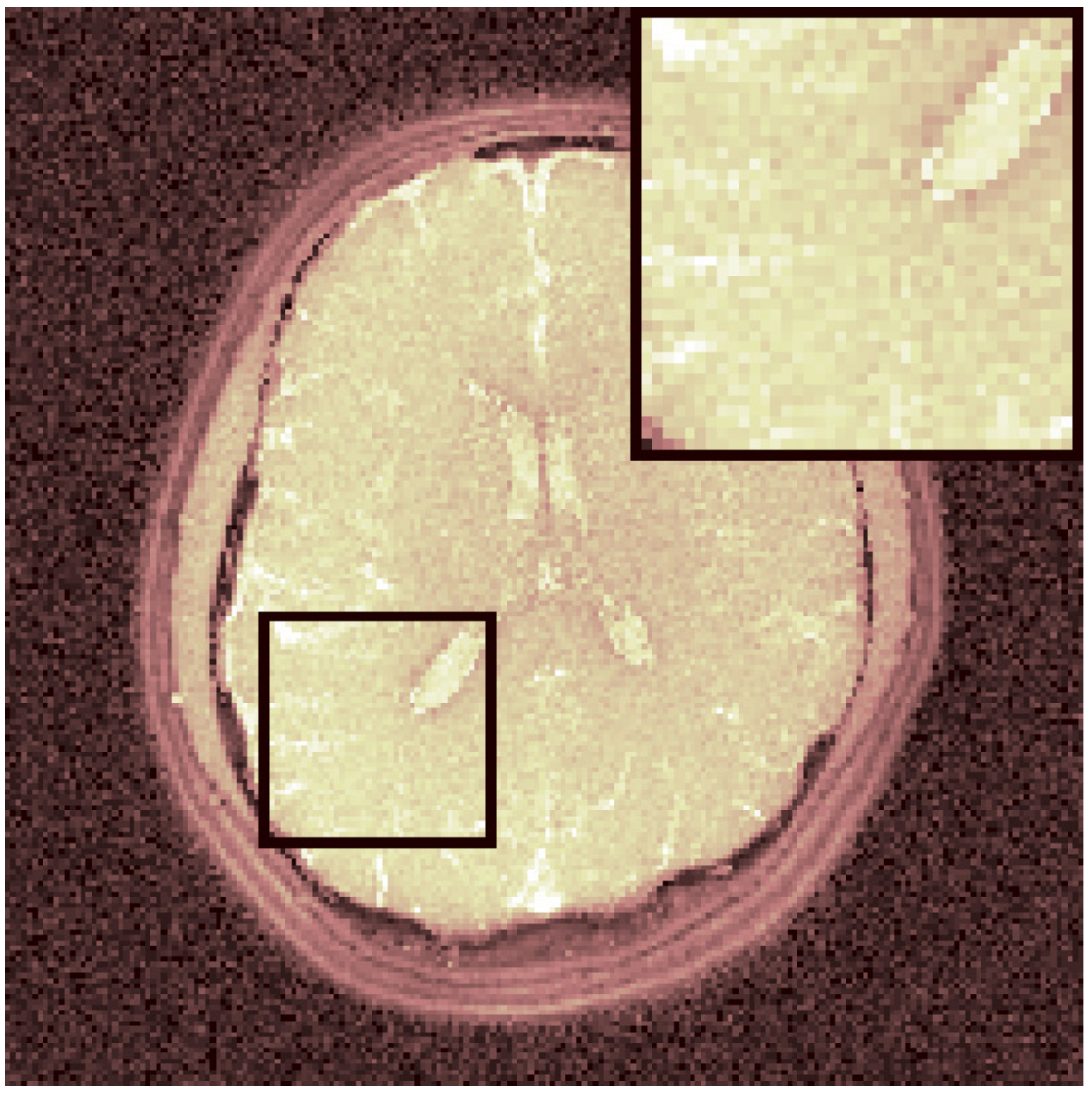}
		\\
		\begin{turn}{90} \quad\qquad FLOR\end{turn}
		\includegraphics[width=.3\linewidth]{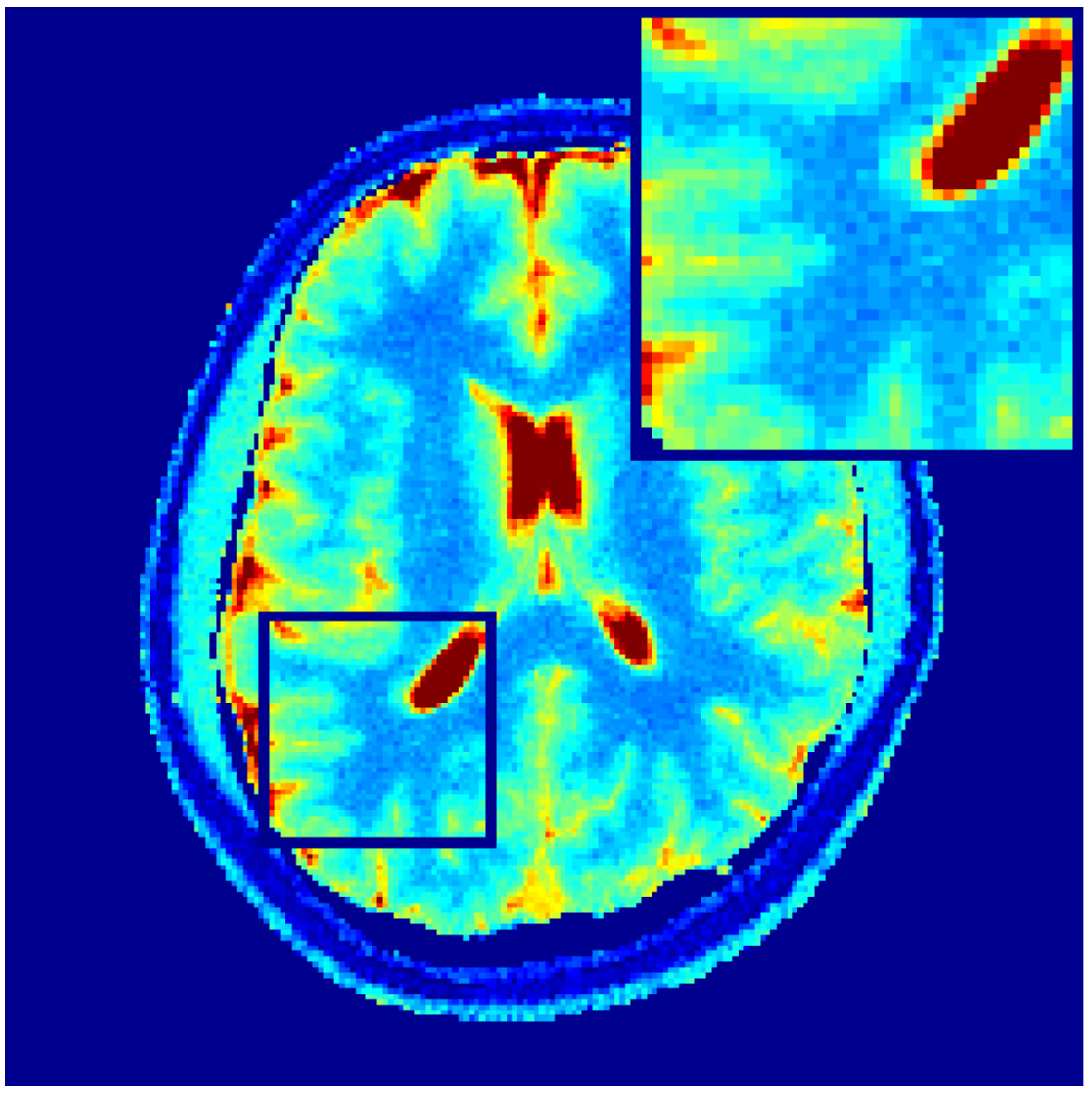}\hspace{-.1cm}
		\includegraphics[width=.3\linewidth]{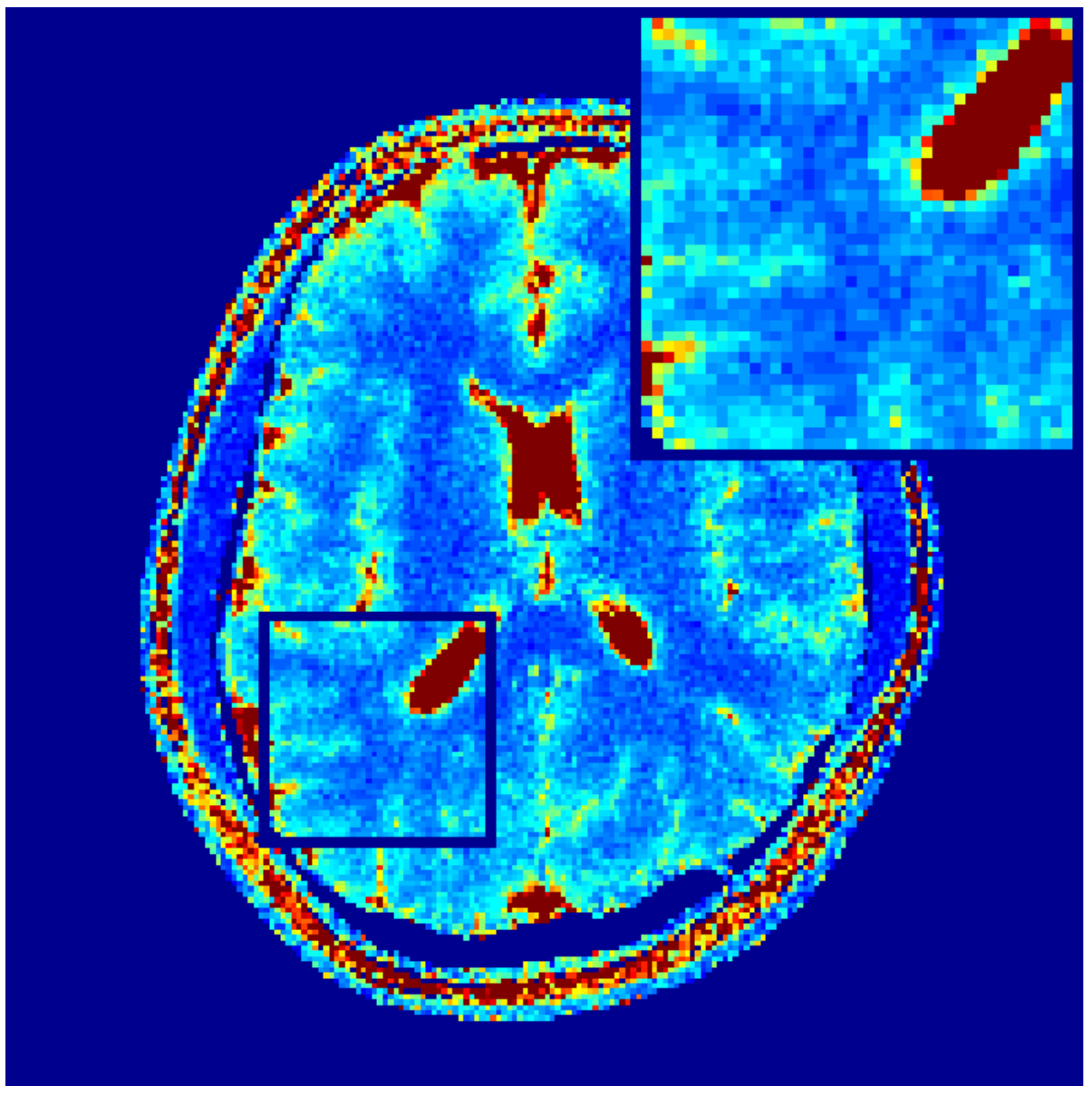}\hspace{-.1cm}
		\includegraphics[width=.3\linewidth]{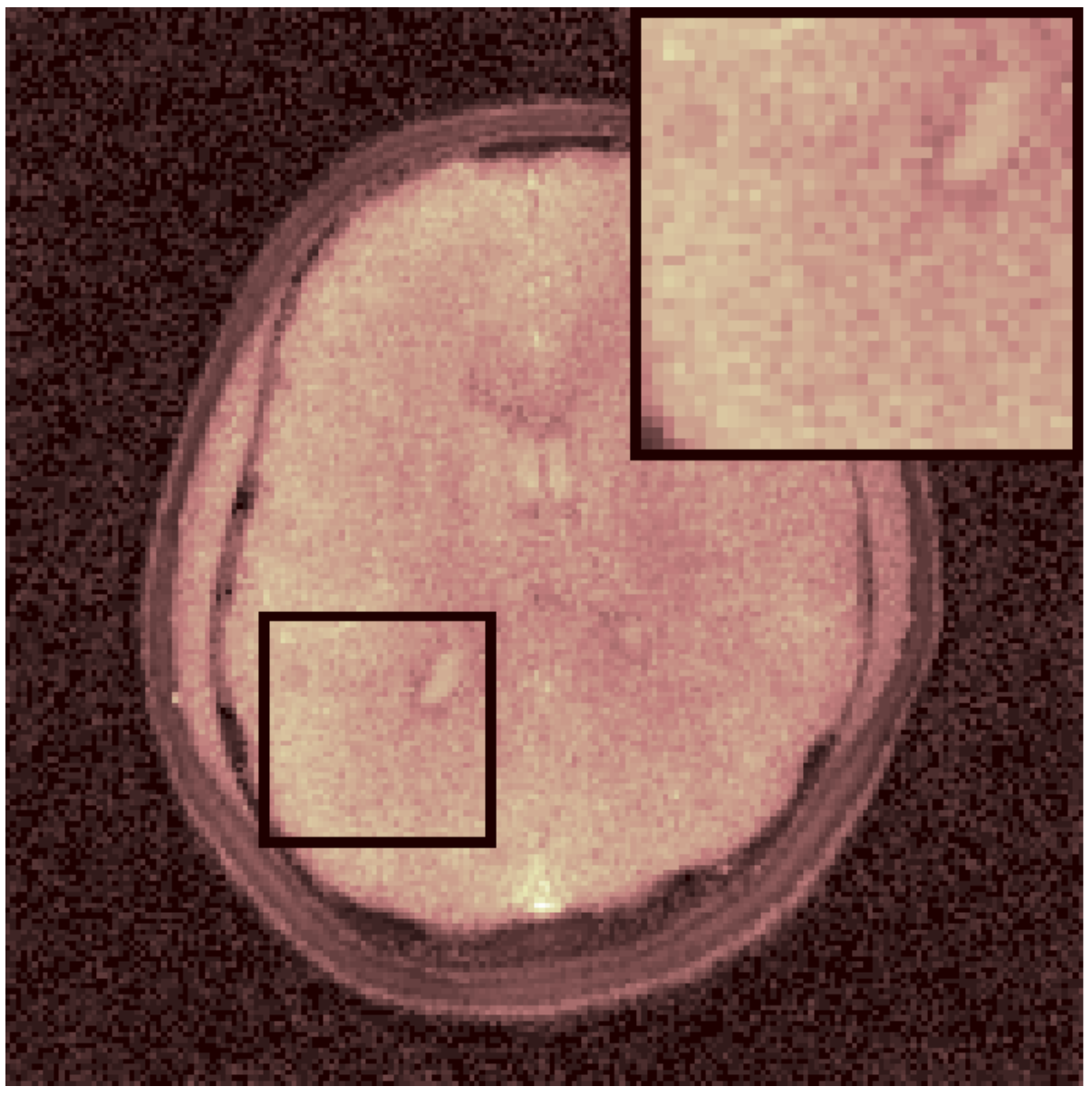}
		\\
		\begin{turn}{90} \quad\quad AIR-MRF\end{turn}
		\includegraphics[width=.3\linewidth]{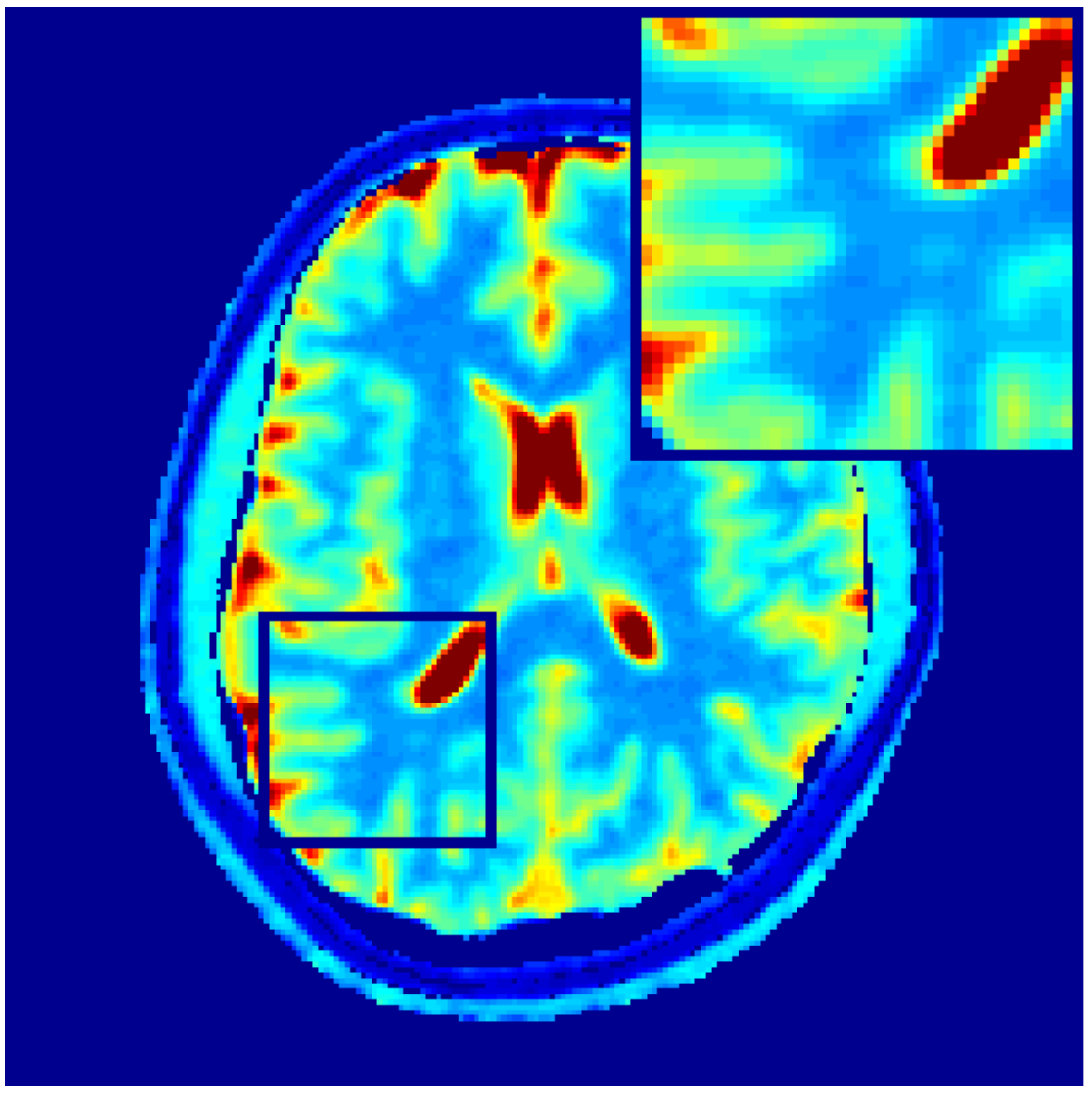}\hspace{-.1cm}
		\includegraphics[width=.3\linewidth]{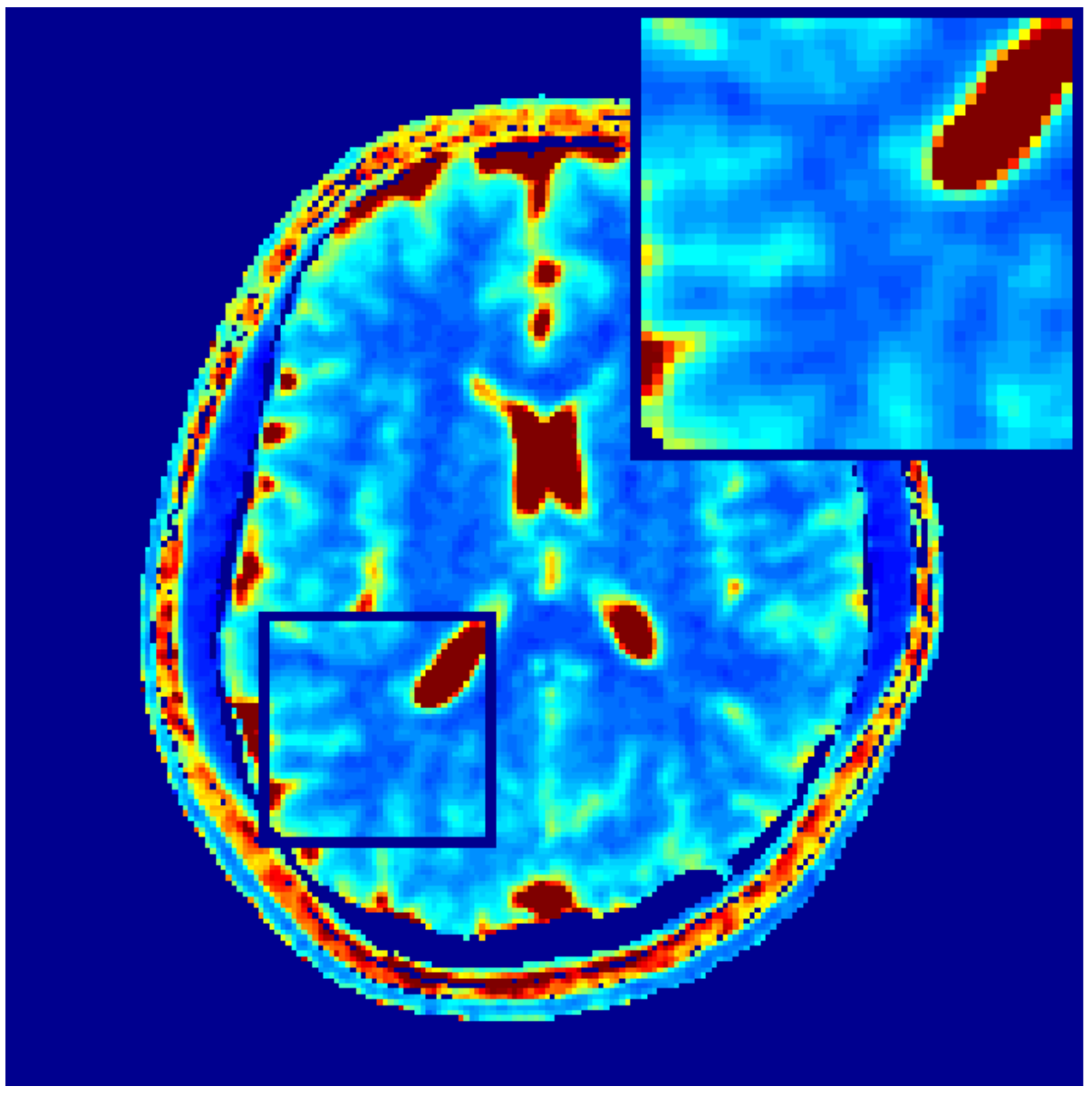}\hspace{-.1cm}
		\includegraphics[width=.3\linewidth]{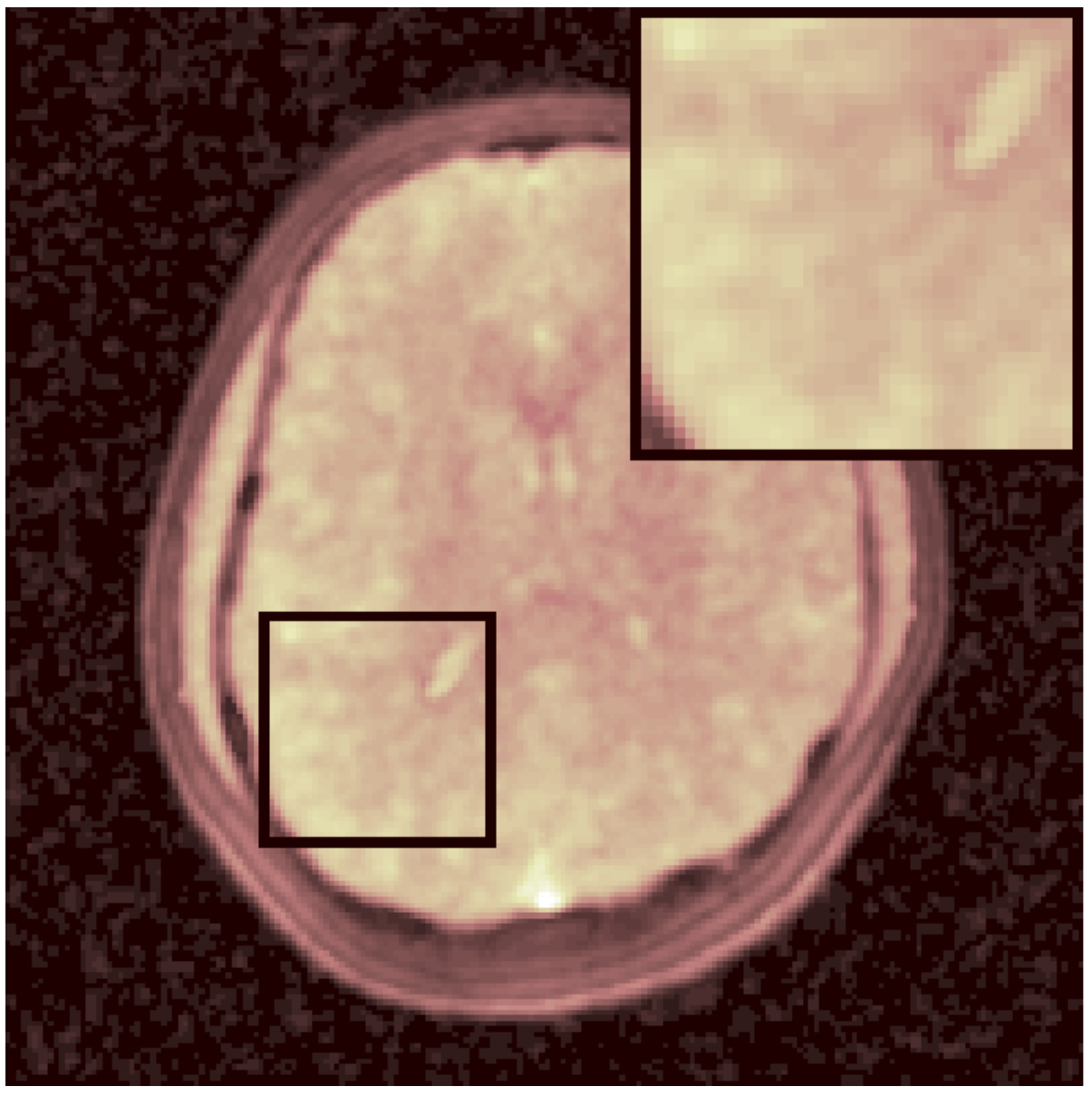}
		\\
		\hrule
		\hrule
		\hrule
		\begin{turn}{90} \quad\quad LRTV-DM\end{turn}
		\includegraphics[width=.3\linewidth]{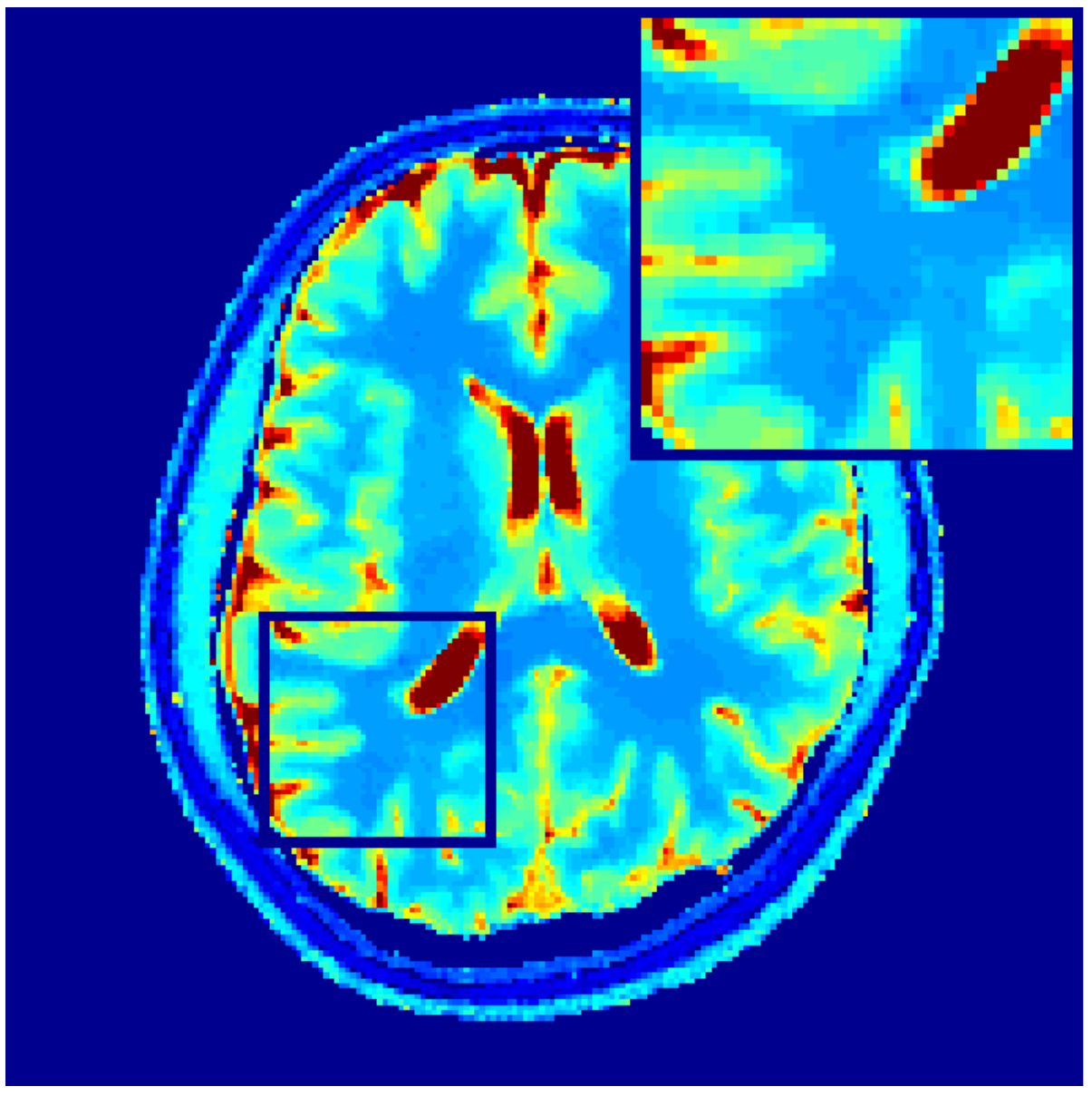}\hspace{-.1cm}
		\includegraphics[width=.3\linewidth]{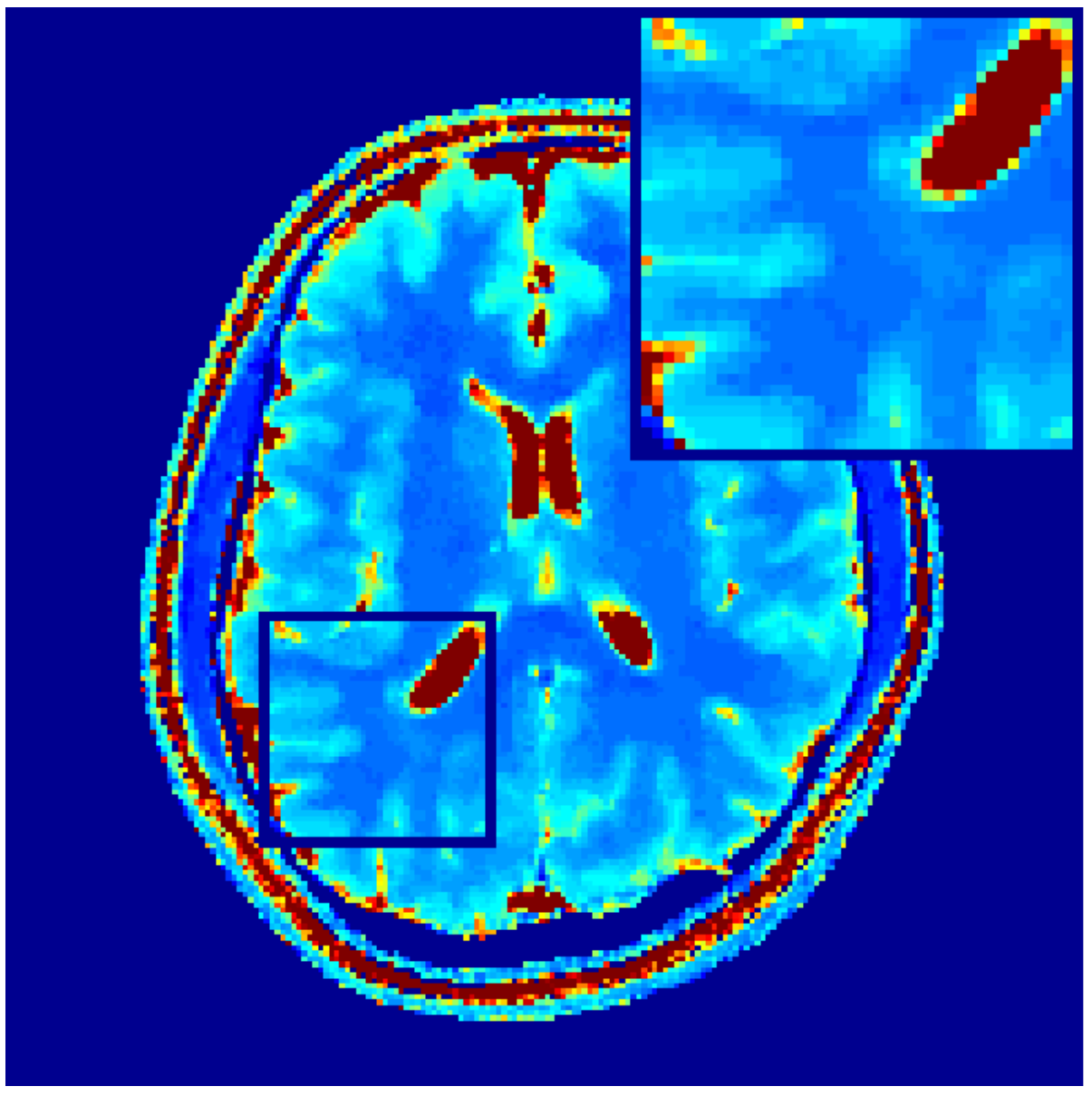}\hspace{-.1cm}
		\includegraphics[width=.3\linewidth]{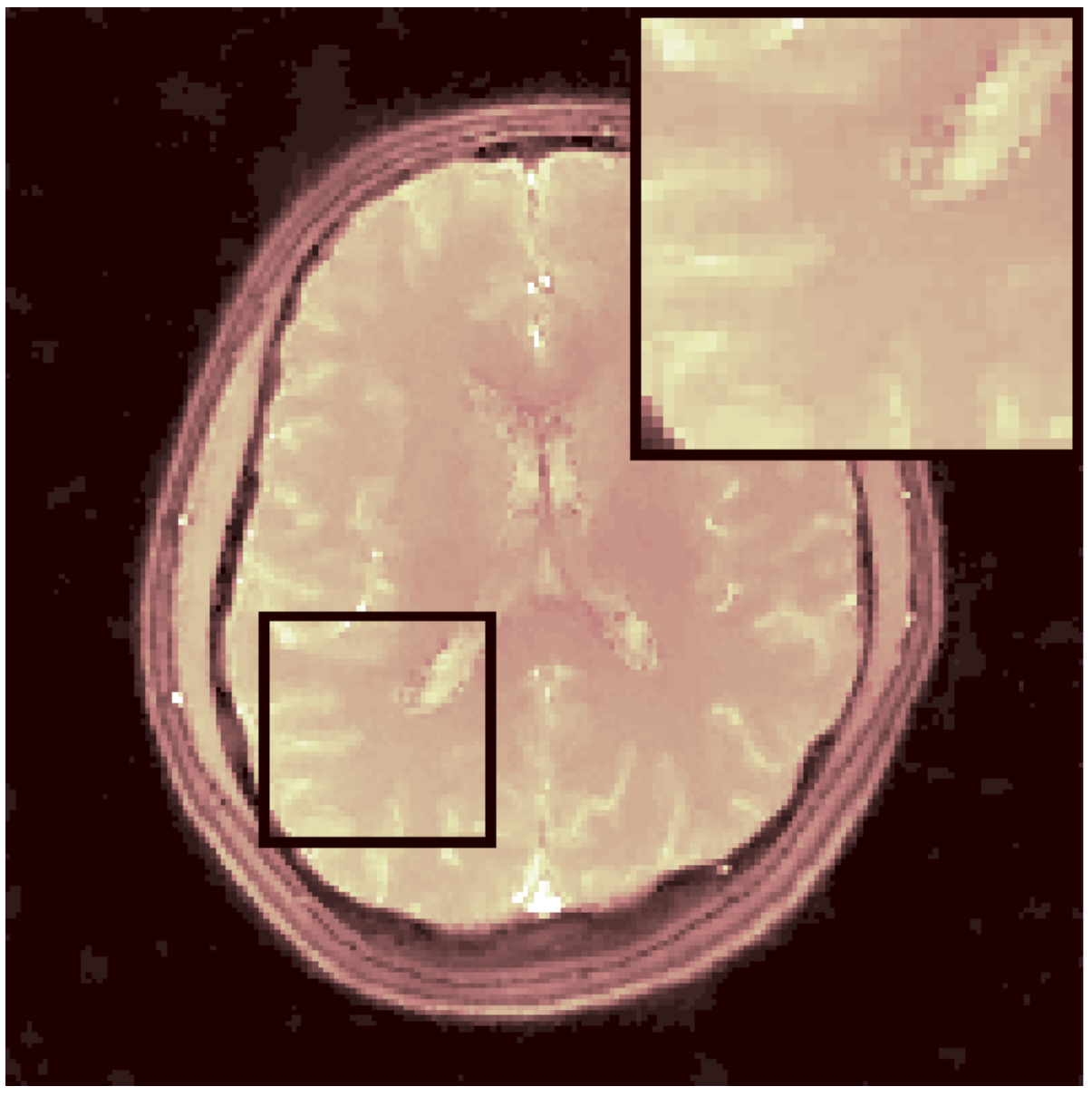}
		\\
		\begin{turn}{90} \quad\quad LRTV-KM\end{turn}
		\includegraphics[width=.3\linewidth]{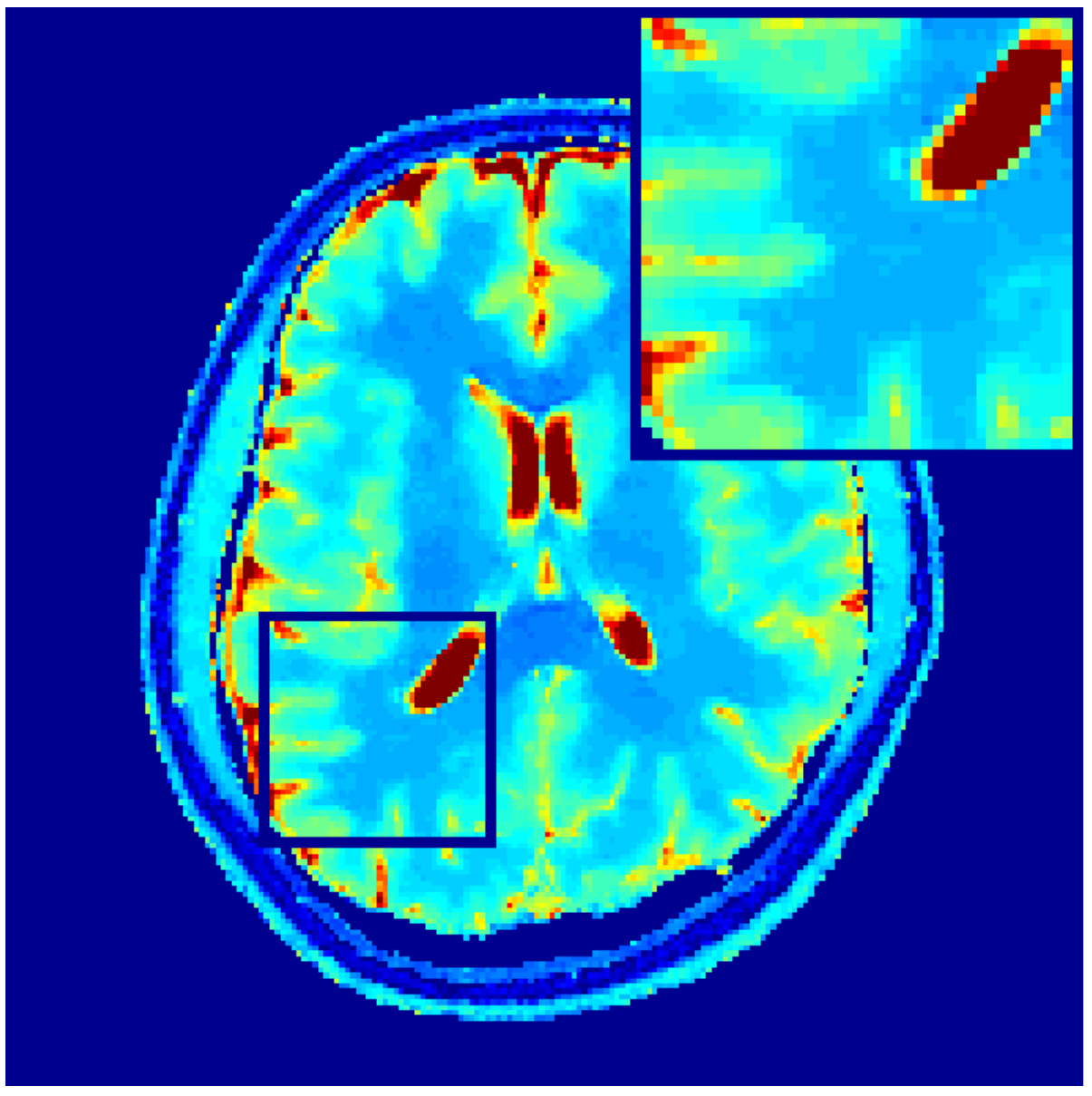}\hspace{-.1cm}
		\includegraphics[width=.3\linewidth]{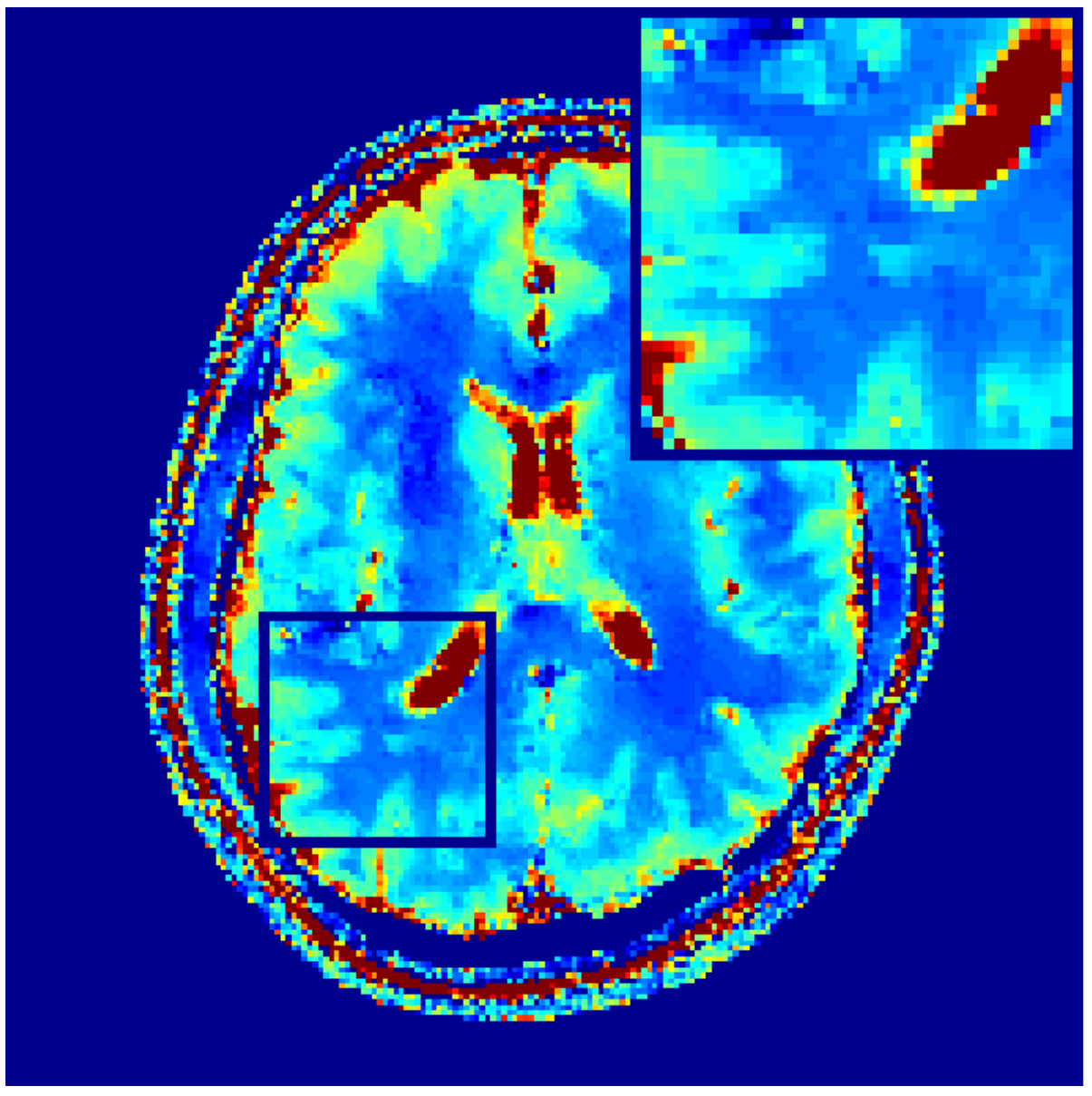}\hspace{-.1cm}
		\includegraphics[width=.3\linewidth]{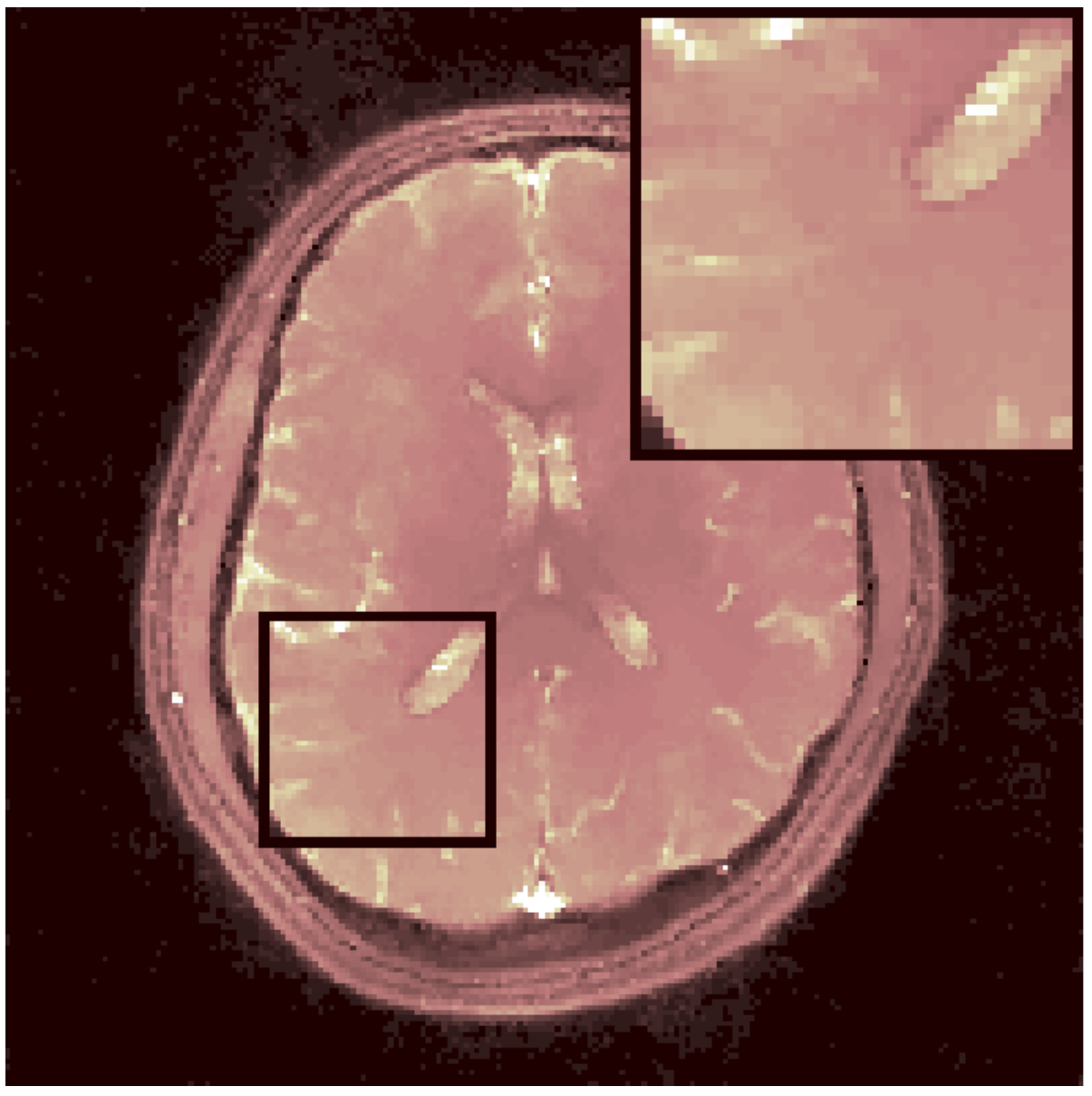}
		\\
		\begin{turn}{90} LRTV-MRFResnet\end{turn}
		\includegraphics[width=.3\linewidth]{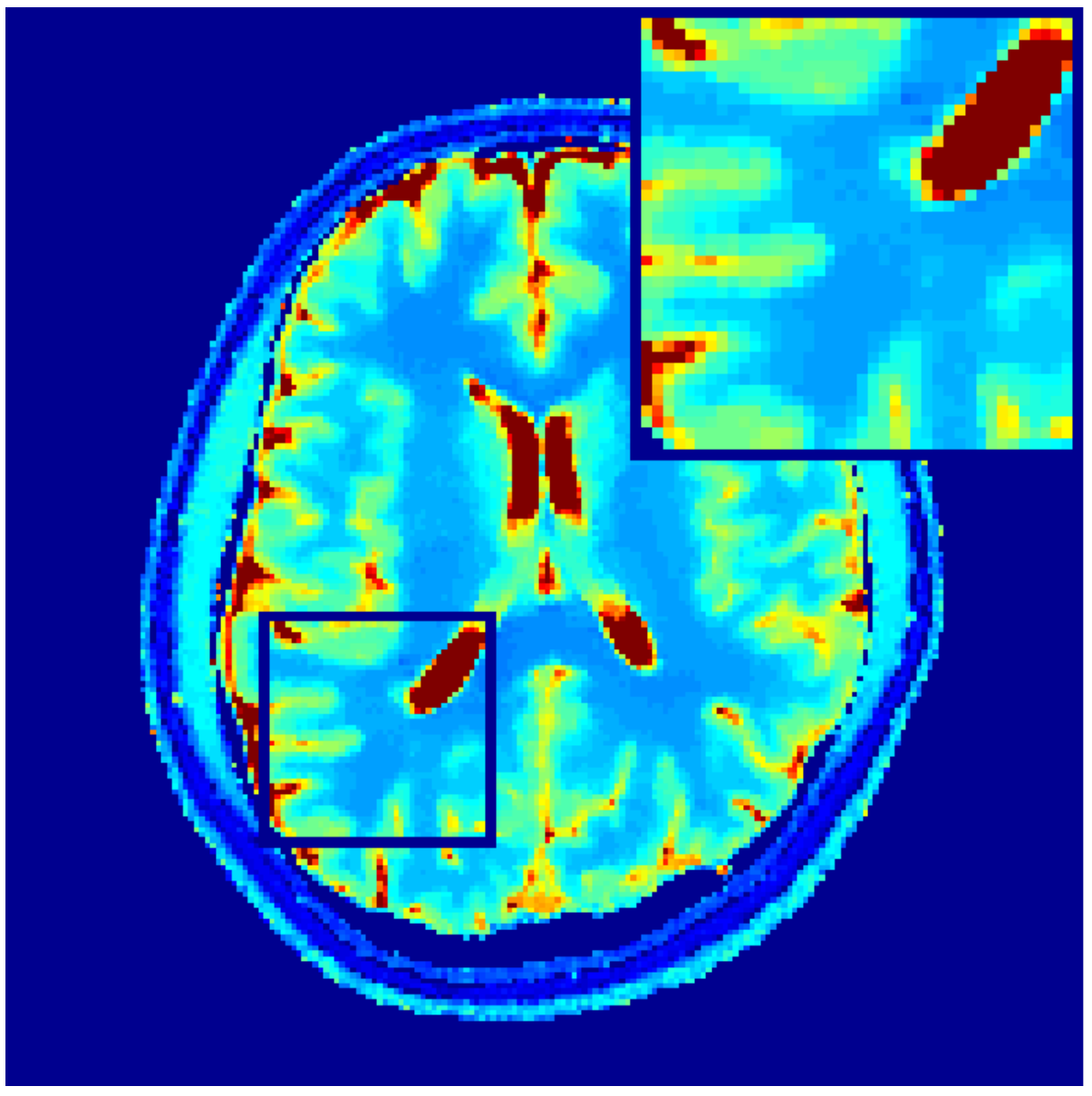}\hspace{-.1cm}
		\includegraphics[width=.3\linewidth]{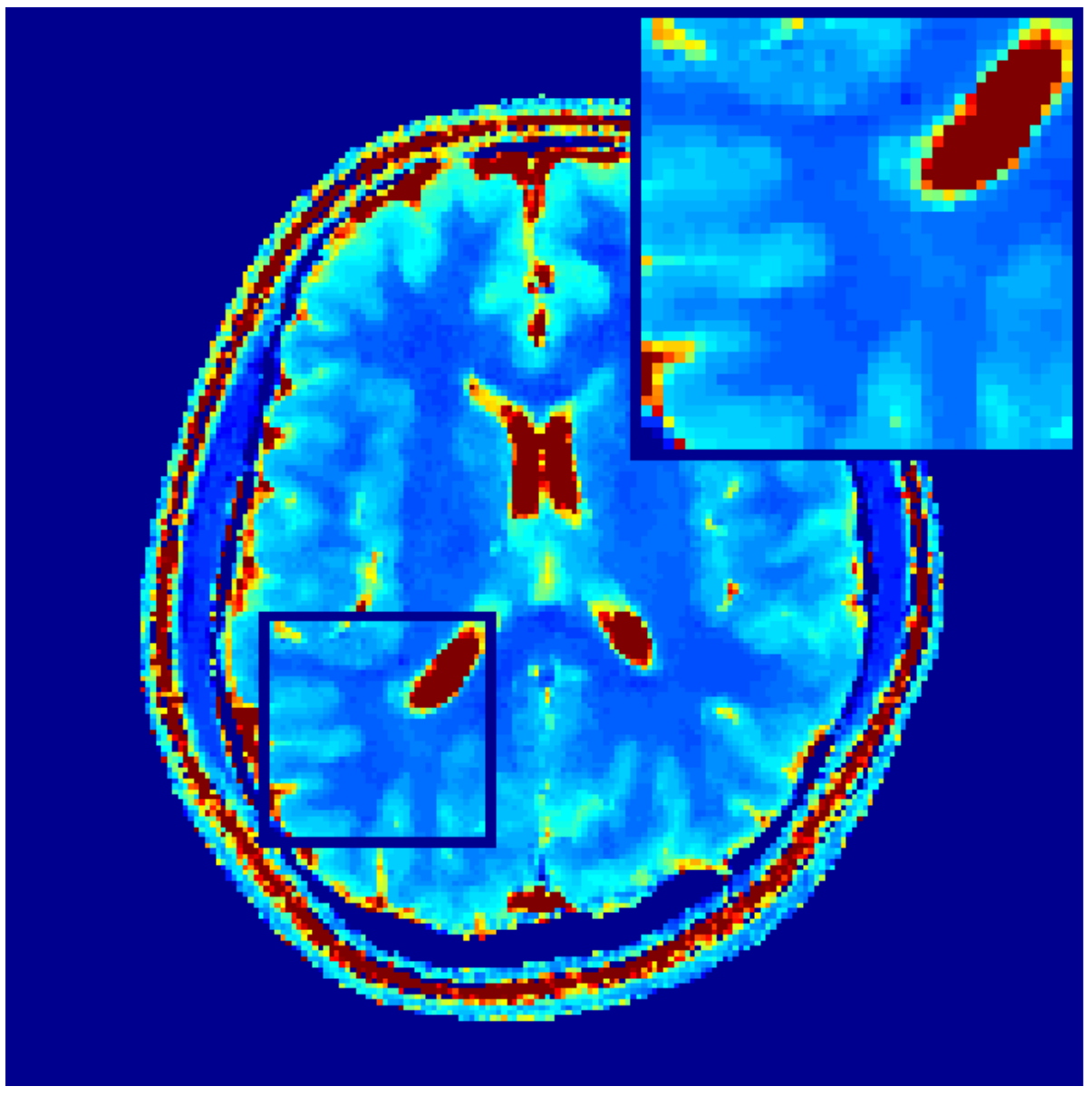}\hspace{-.1cm}
		\includegraphics[width=.3\linewidth]{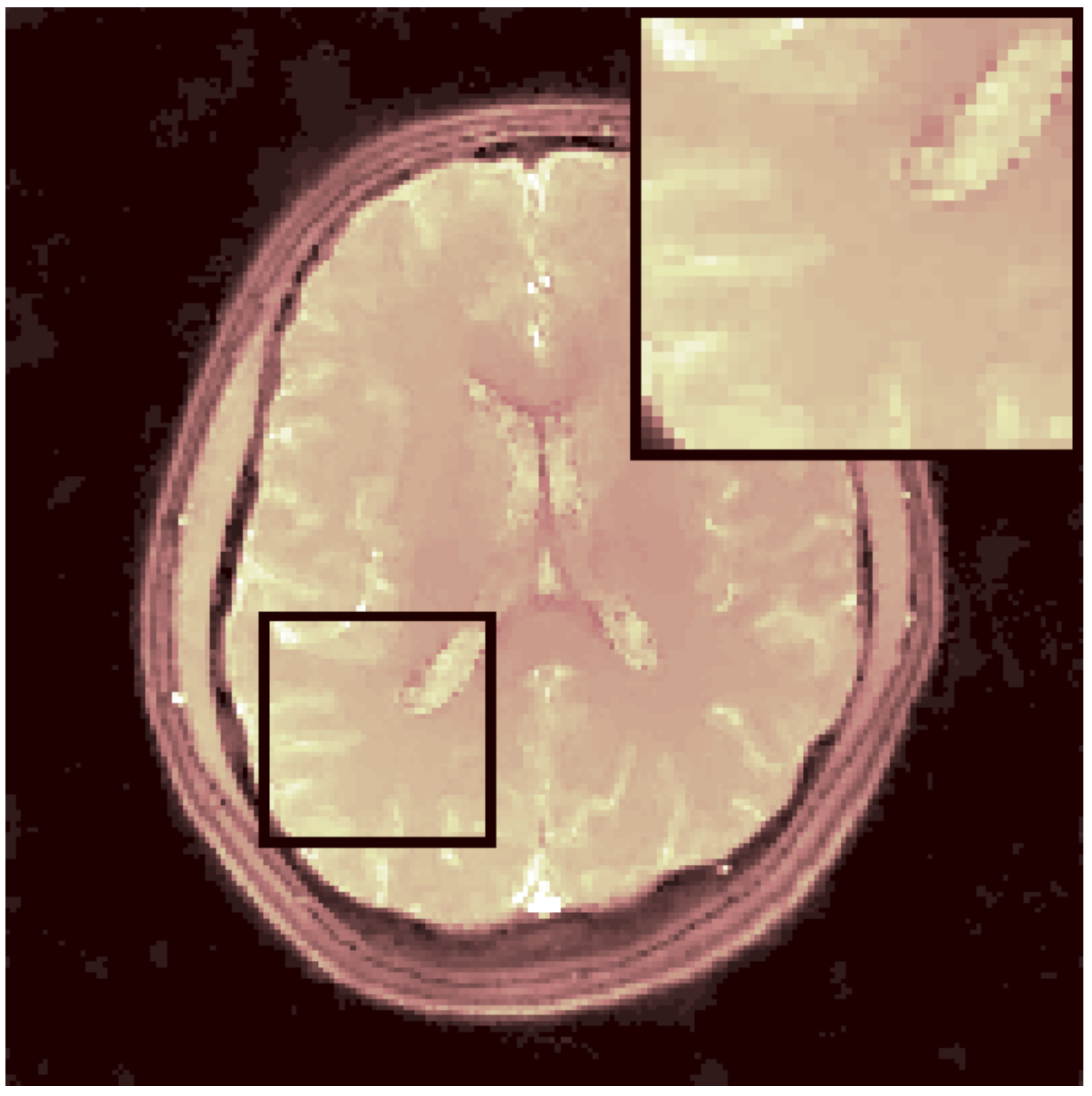}	
		\\
		\includegraphics[trim= -10 50 -20 720, clip,width=.3\linewidth]{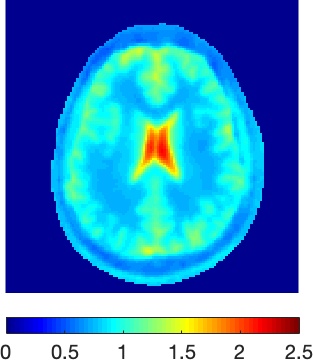}
		\includegraphics[trim= -10 50 -20 700, clip,width=.3\linewidth]{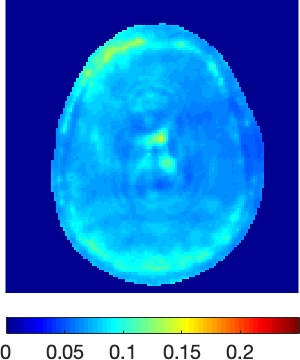}
		\includegraphics[trim= -10 50 -20 700, clip,width=.3\linewidth]{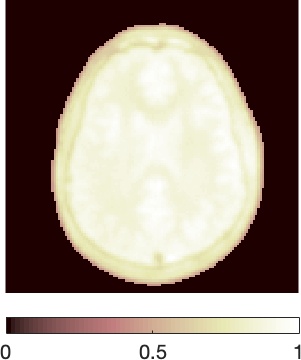}\vspace{.5cm}
				\\
		T1(s) \hspace{1.6cm} T2 (s) \hspace{1.6 cm}  PD (a.u.) \\
		\caption{\footnotesize{Reconstructed T1, T2 and PD maps from a 2D spiral acquisition (real-world scan) using different reconstruction and inference algorithms.}  \label{fig:2dvivo_spiral}}
\end{minipage}}
\end{figure}

\begin{figure}[t!]
	\centering
	\scalebox{.95}{
	\begin{minipage}{\linewidth}
		\centering		
		\begin{turn}{90} \quad\quad ZF-DM\end{turn}
		\includegraphics[width=.3\linewidth]{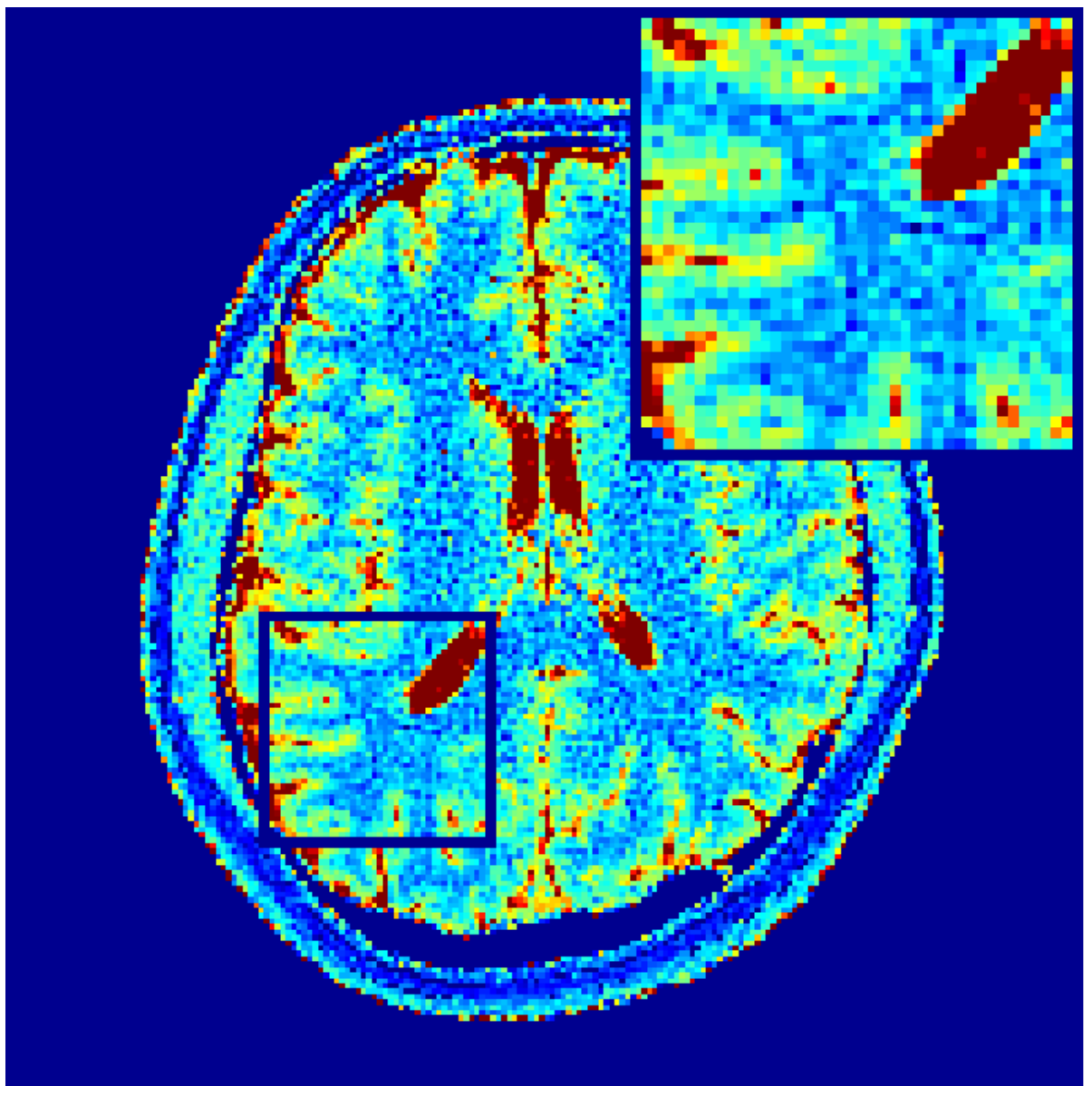}\hspace{-.1cm}
		\includegraphics[width=.3\linewidth]{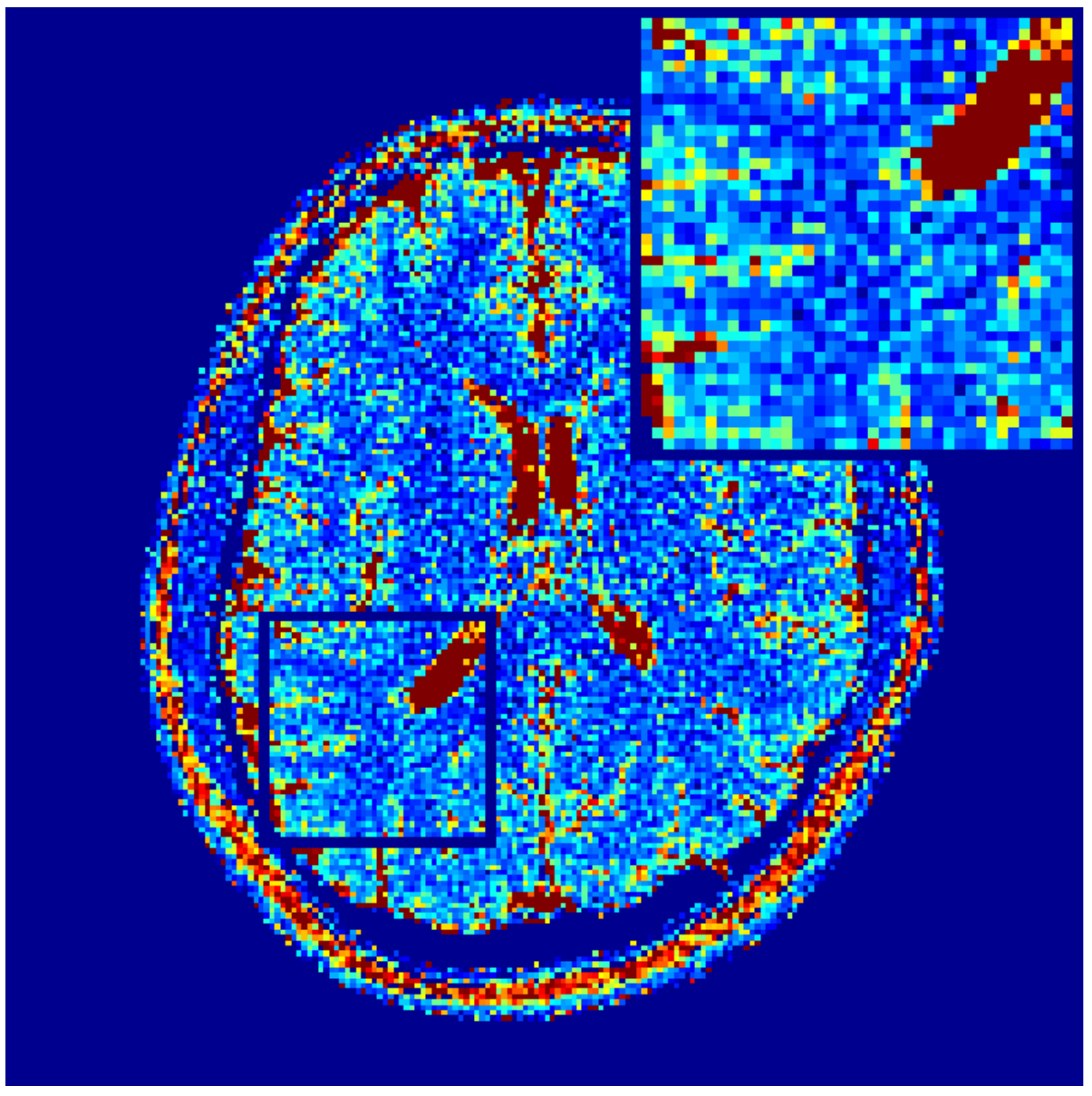}\hspace{-.1cm}
		\includegraphics[width=.3\linewidth]{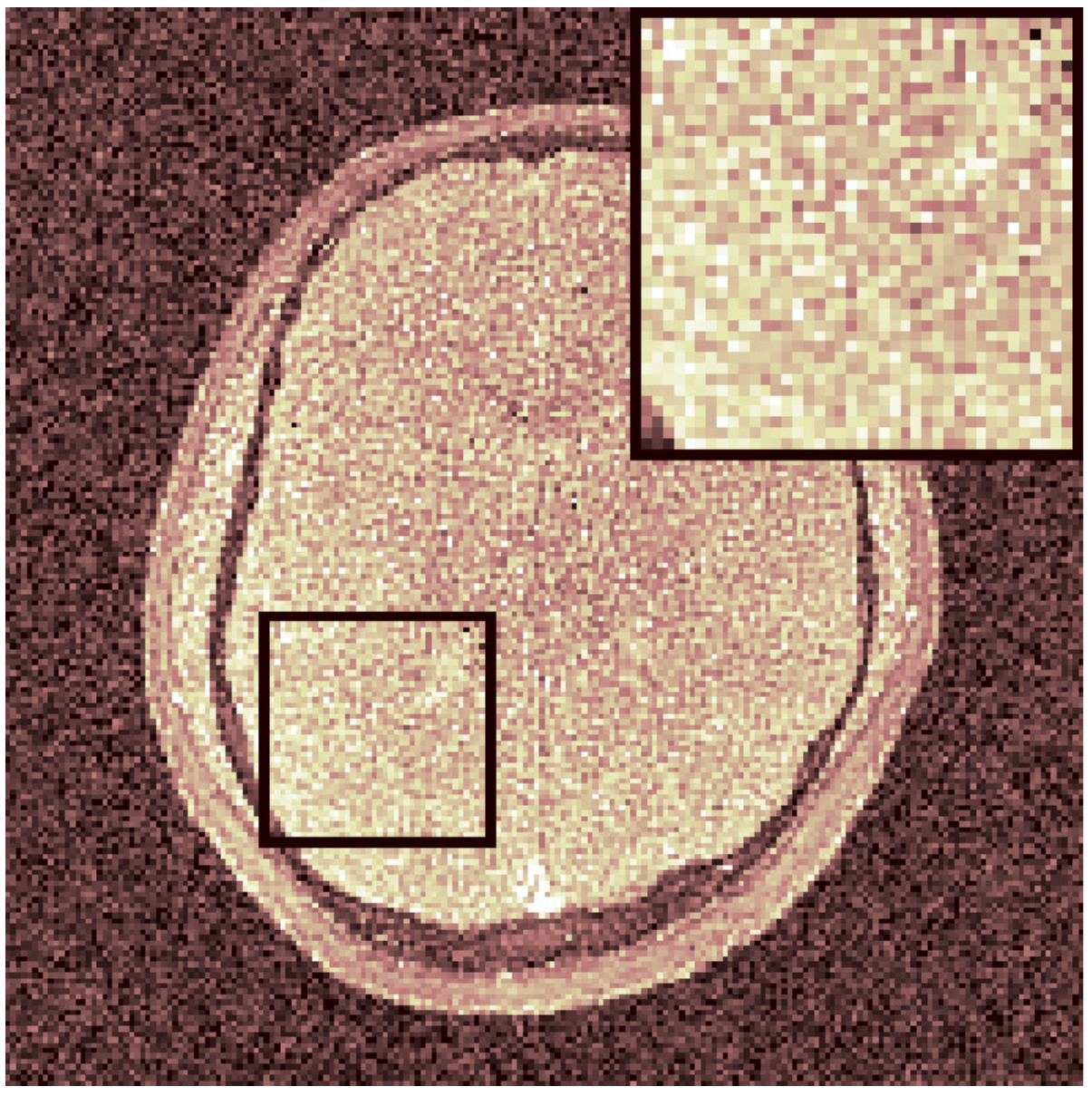}
		\\
		\begin{turn}{90} \quad\quad LR-DM\end{turn}
		\includegraphics[width=.3\linewidth]{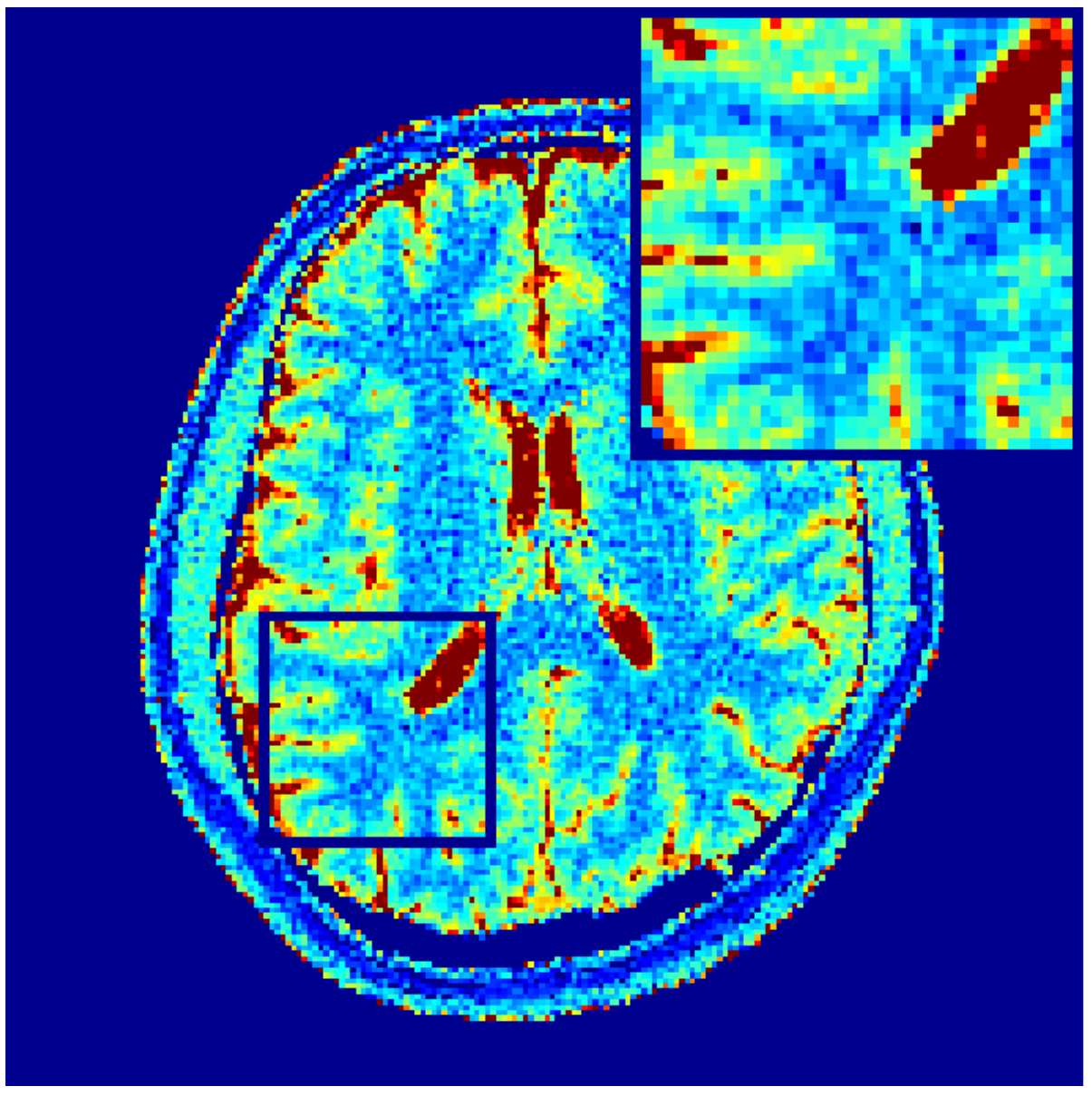}\hspace{-.1cm}
		\includegraphics[width=.3\linewidth]{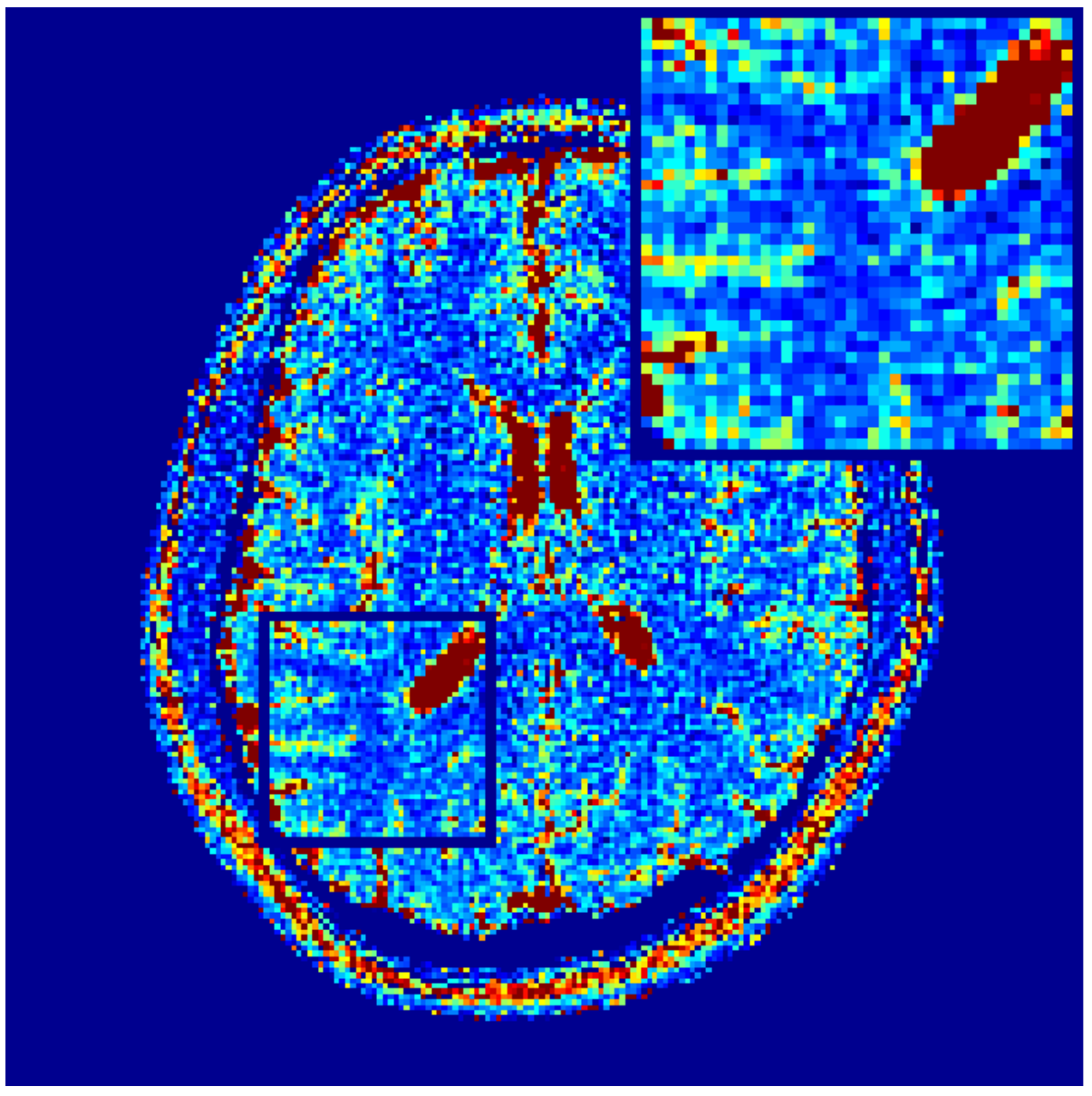}\hspace{-.1cm}
		\includegraphics[width=.3\linewidth]{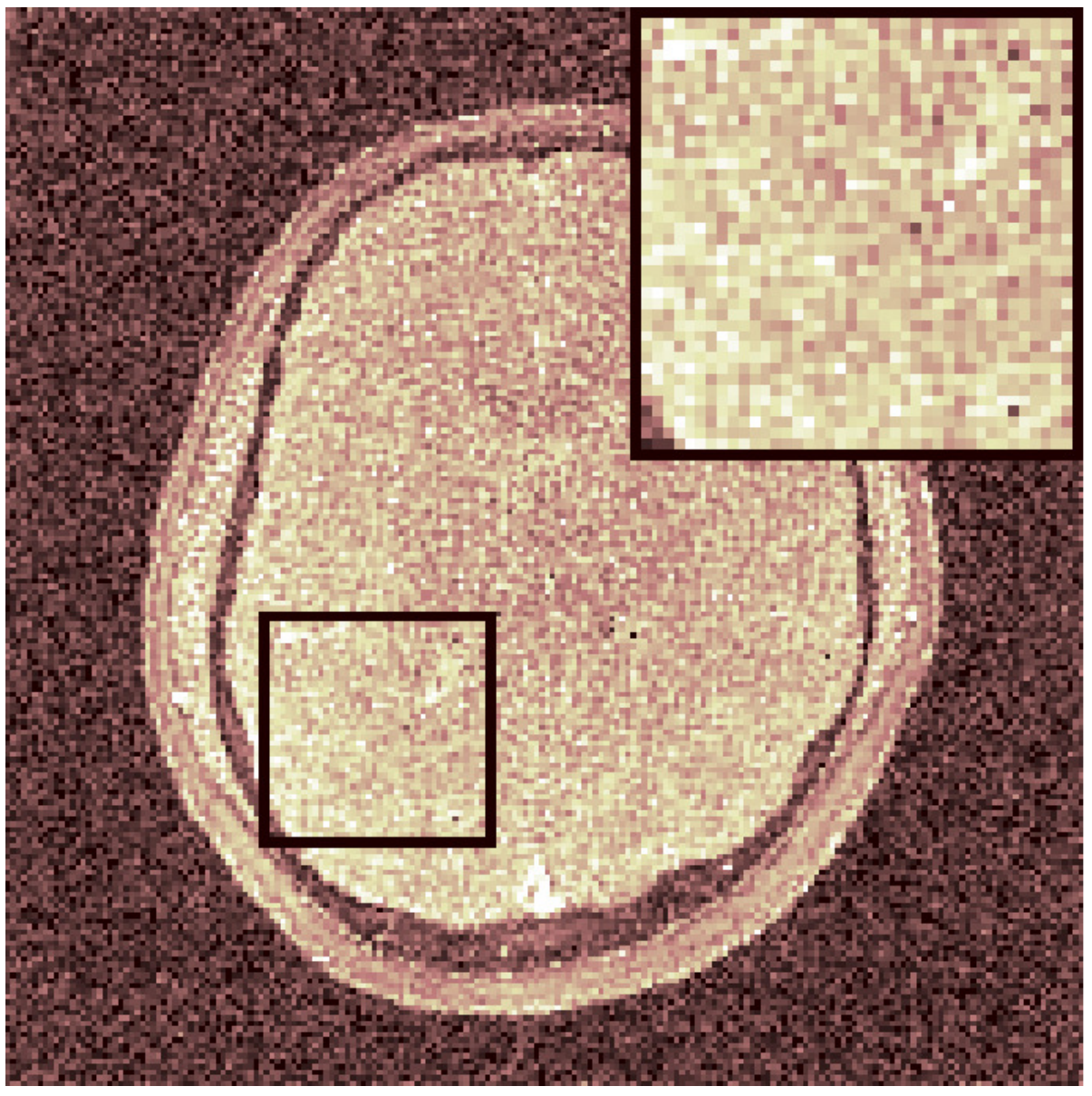}		
		\\
		\begin{turn}{90} \quad\quad VS-DM\end{turn}
		\includegraphics[width=.3\linewidth]{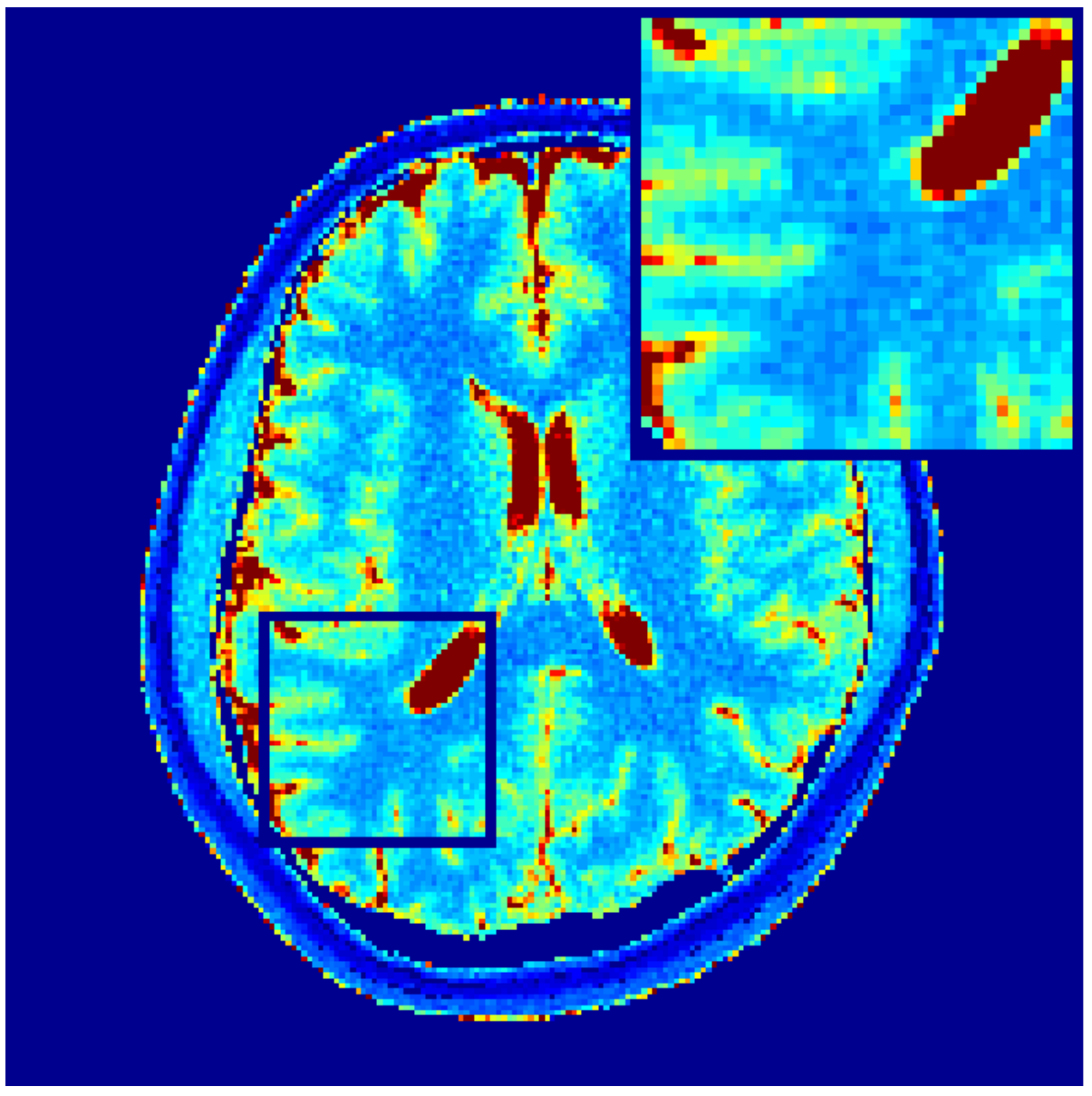}\hspace{-.1cm}
		\includegraphics[width=.3\linewidth]{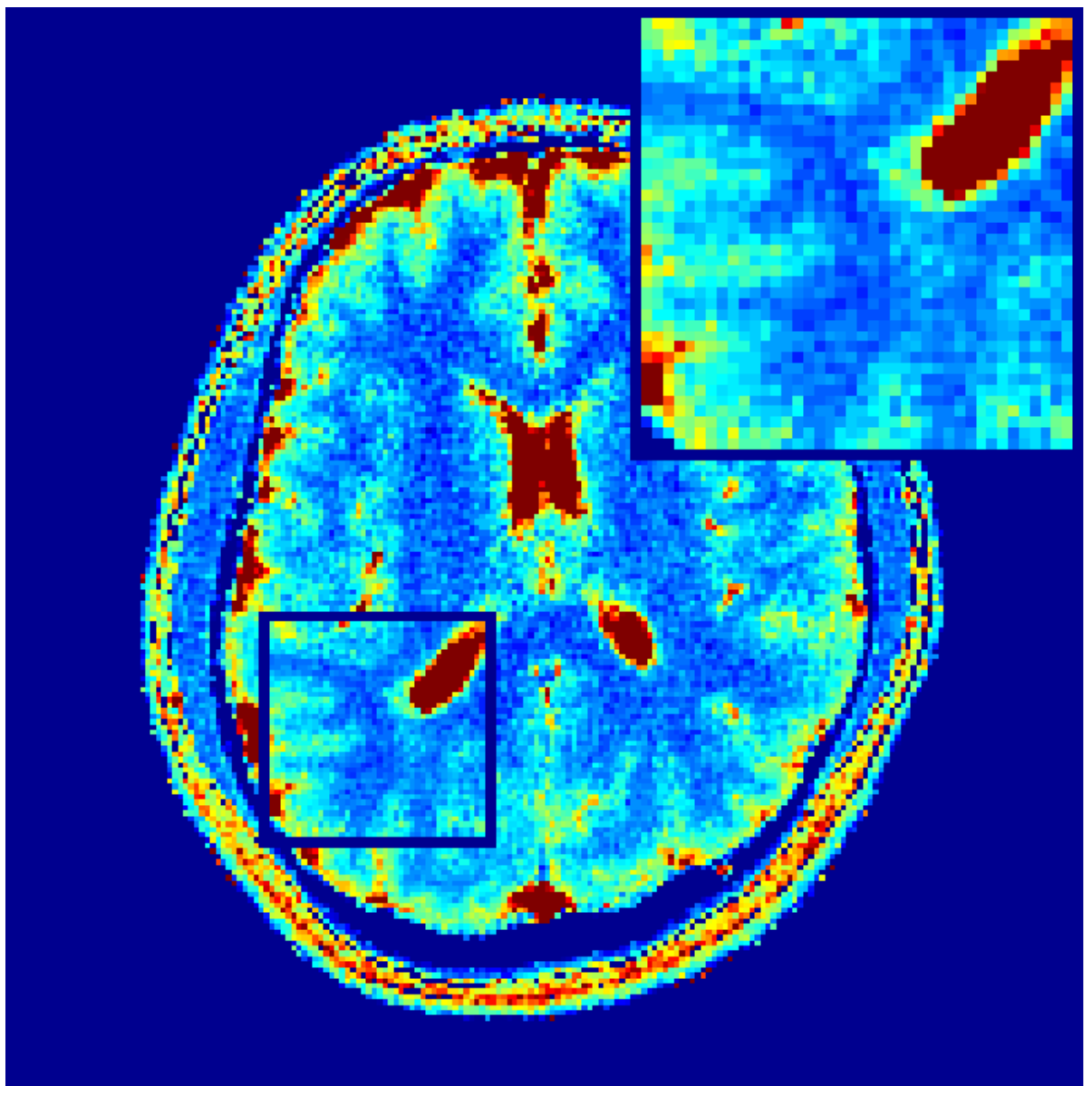}\hspace{-.1cm}
		\includegraphics[width=.3\linewidth]{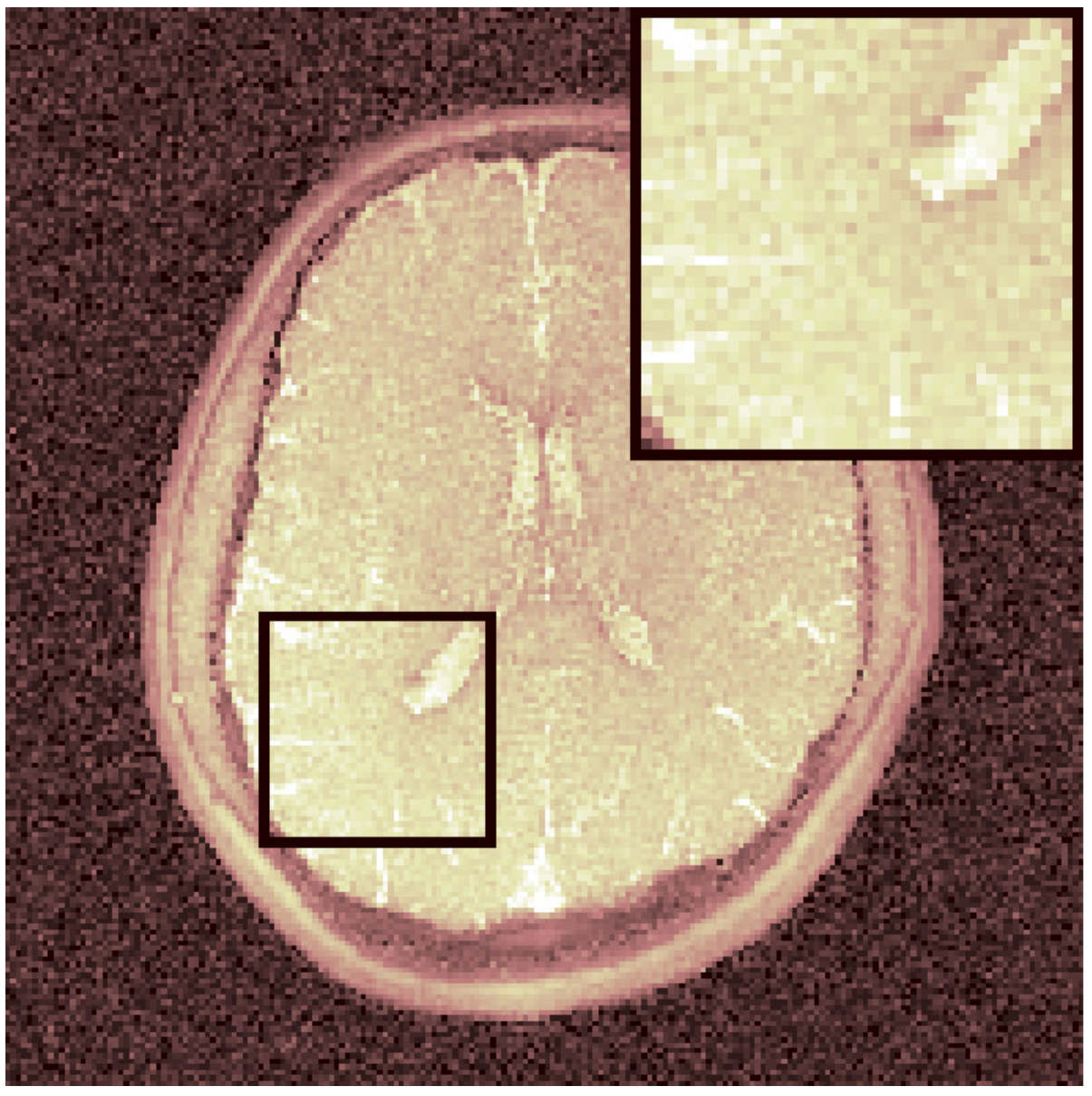}	
		\\
		\begin{turn}{90} \quad\qquad FLOR\end{turn}
		\includegraphics[width=.3\linewidth]{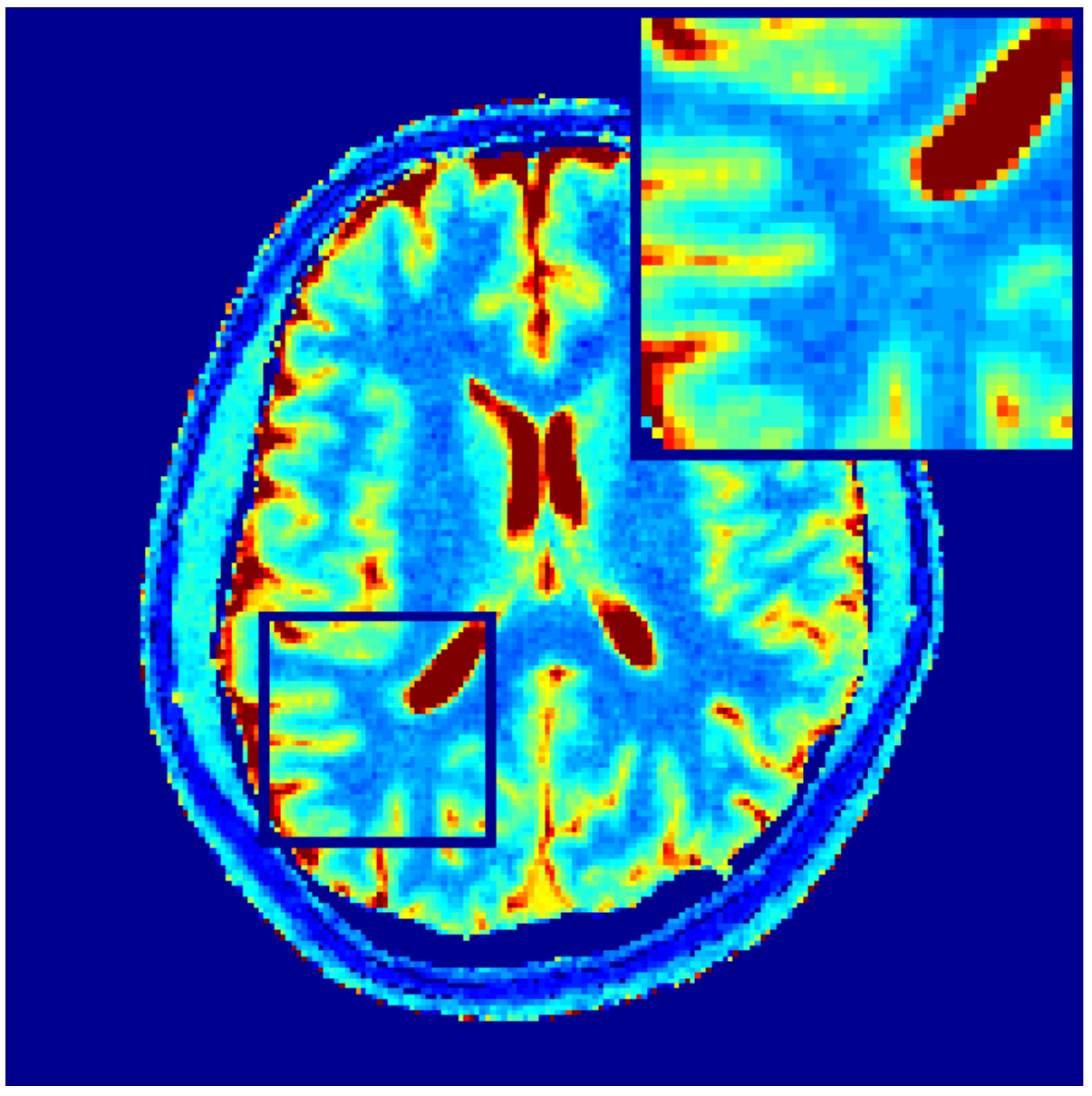}\hspace{-.1cm}
		\includegraphics[width=.3\linewidth]{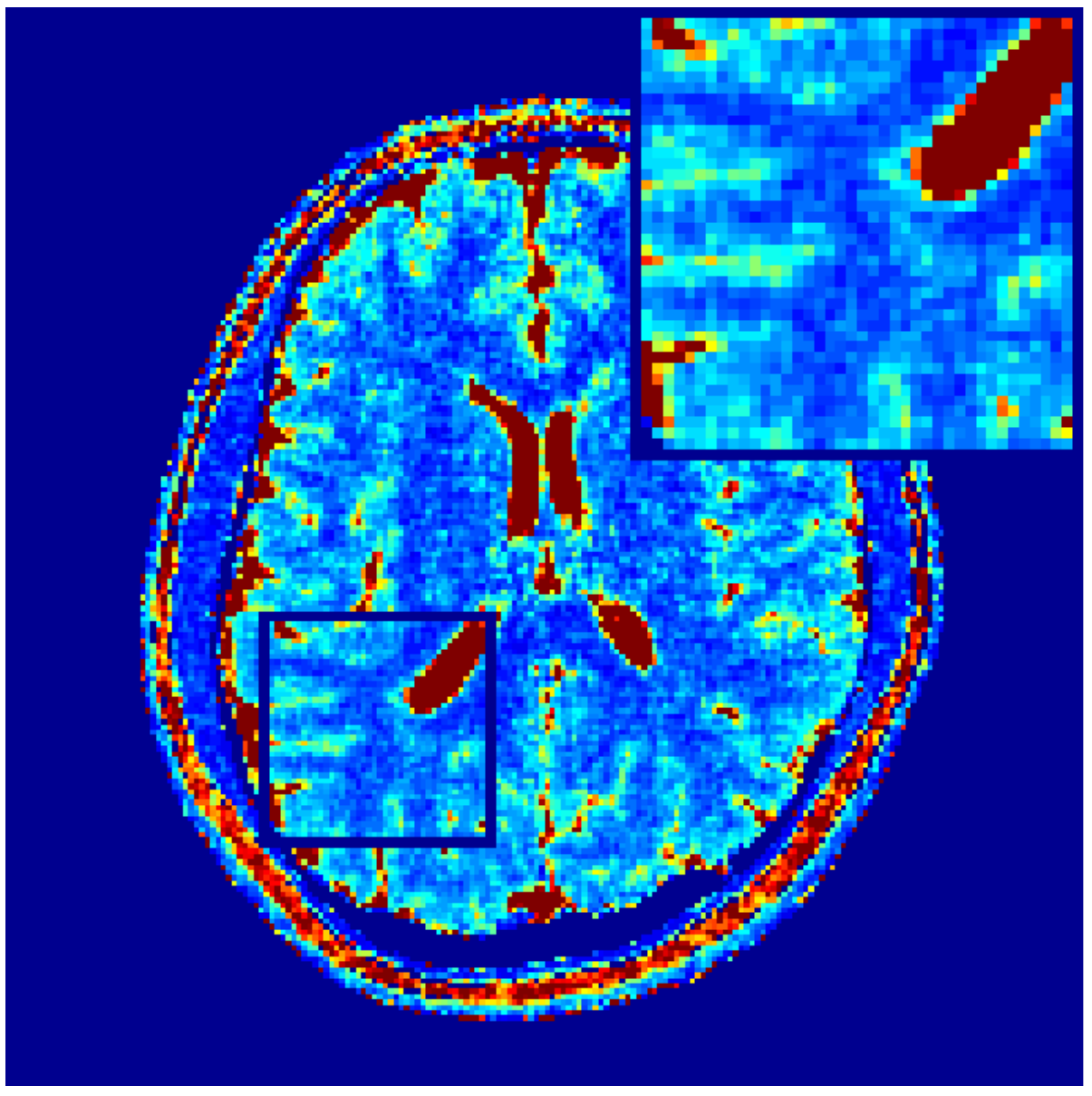}\hspace{-.1cm}
		\includegraphics[width=.3\linewidth]{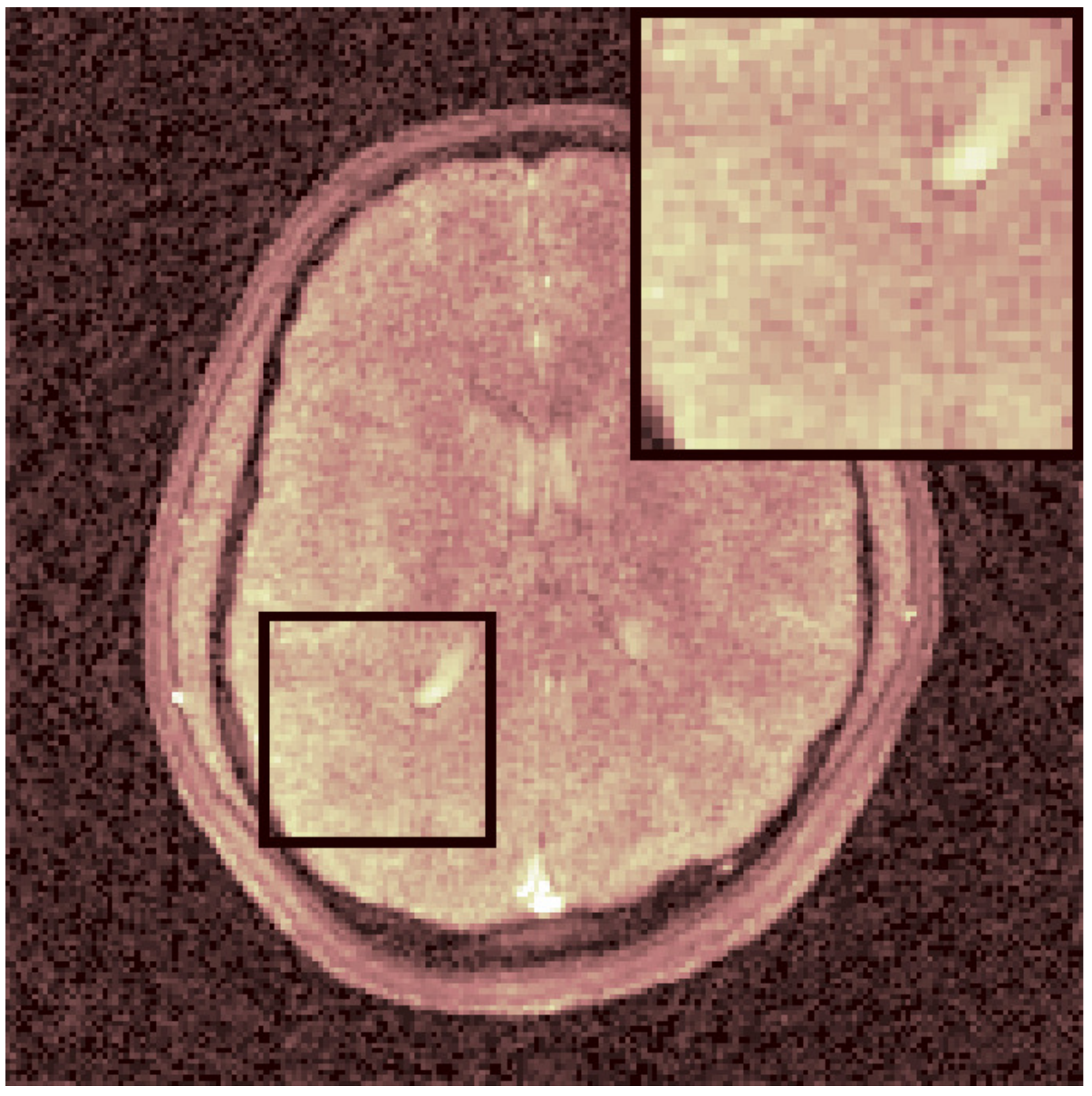}
		\\
		\begin{turn}{90} \quad\quad AIR-MRF\end{turn}
		\includegraphics[width=.3\linewidth]{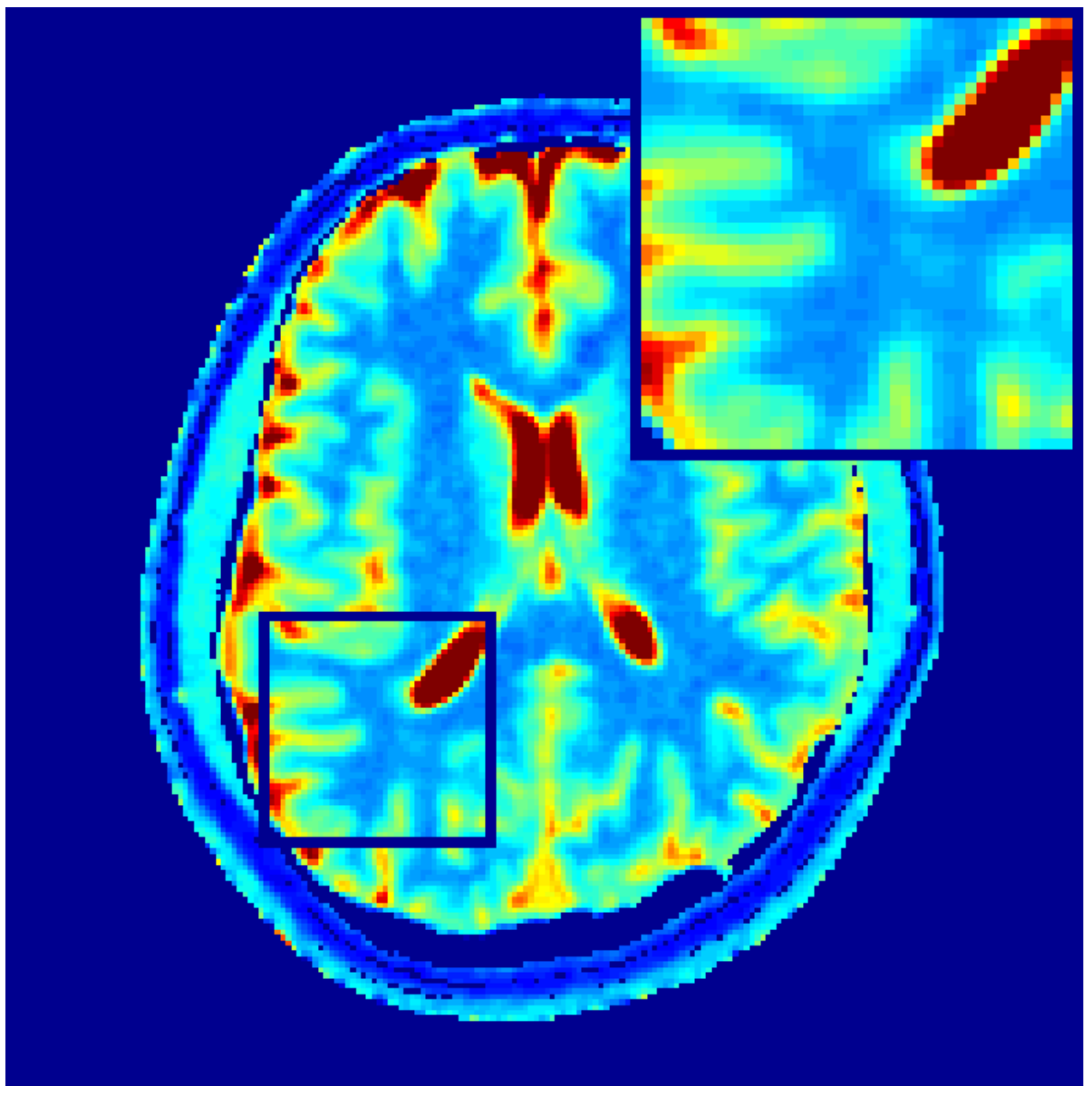}\hspace{-.1cm}
		\includegraphics[width=.3\linewidth]{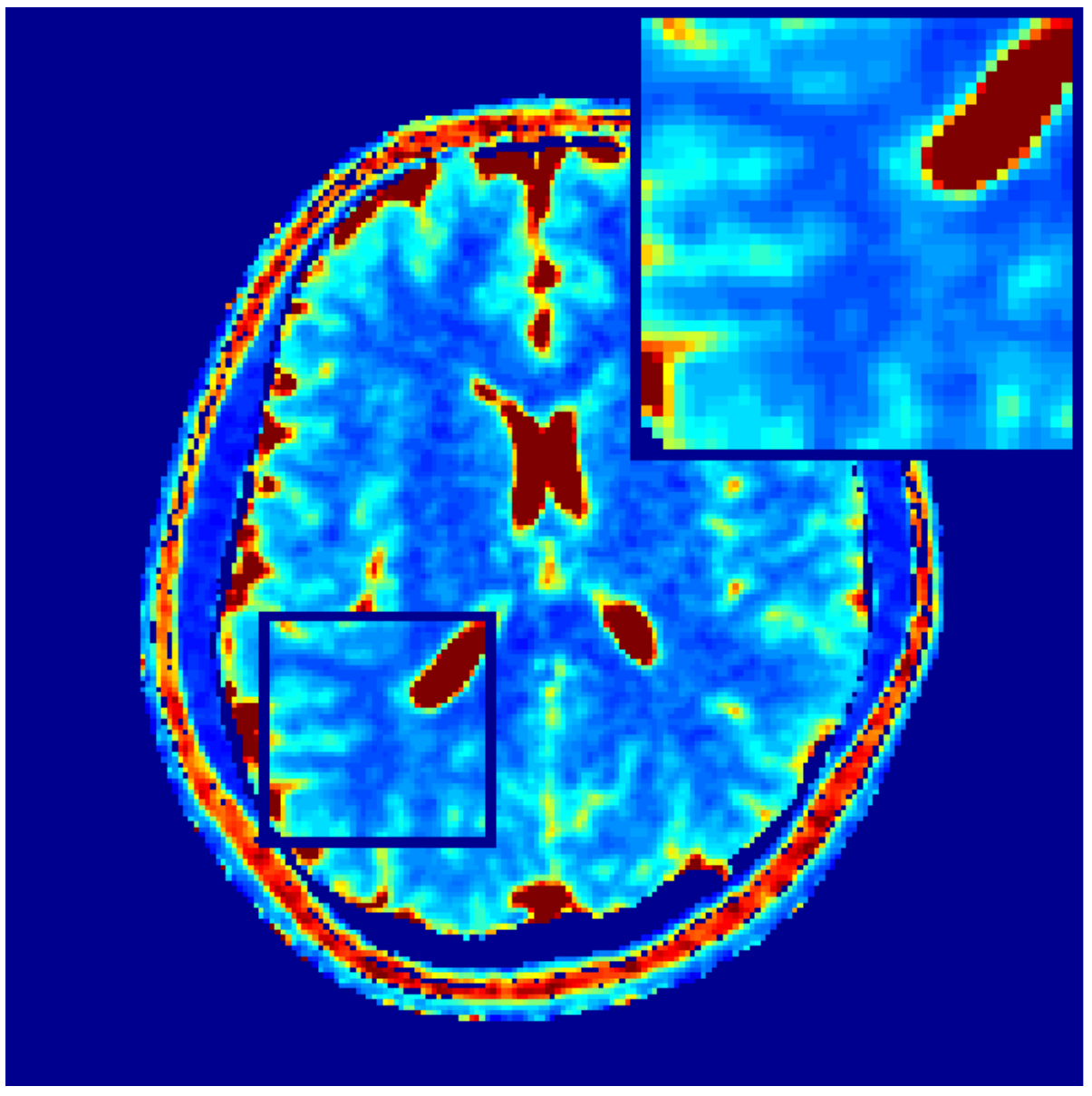}\hspace{-.1cm}
		\includegraphics[width=.3\linewidth]{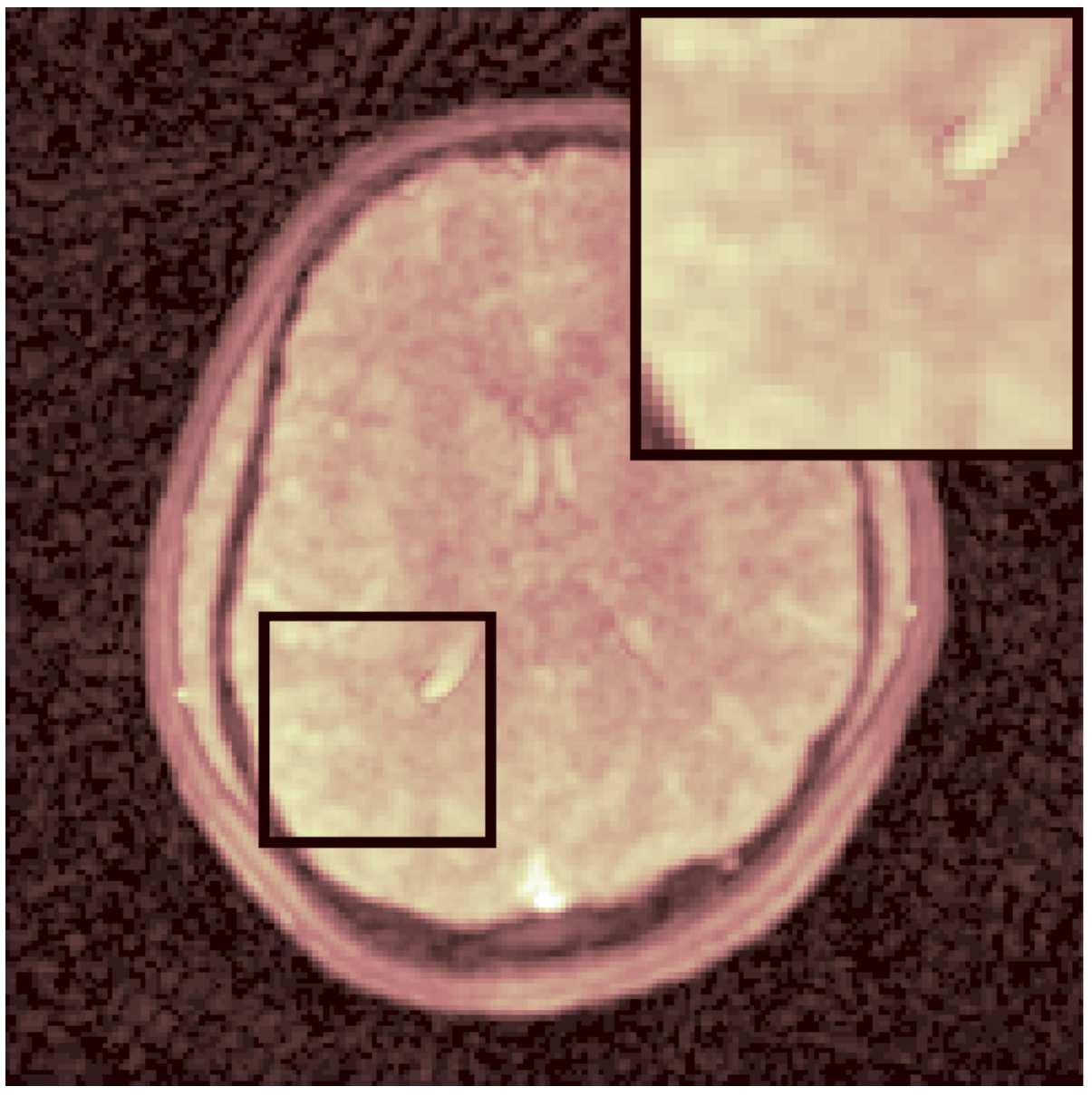}
		\\
		\hrule
		\hrule
		\hrule
		\begin{turn}{90} \quad\quad LRTV-DM\end{turn}
		\includegraphics[width=.3\linewidth]{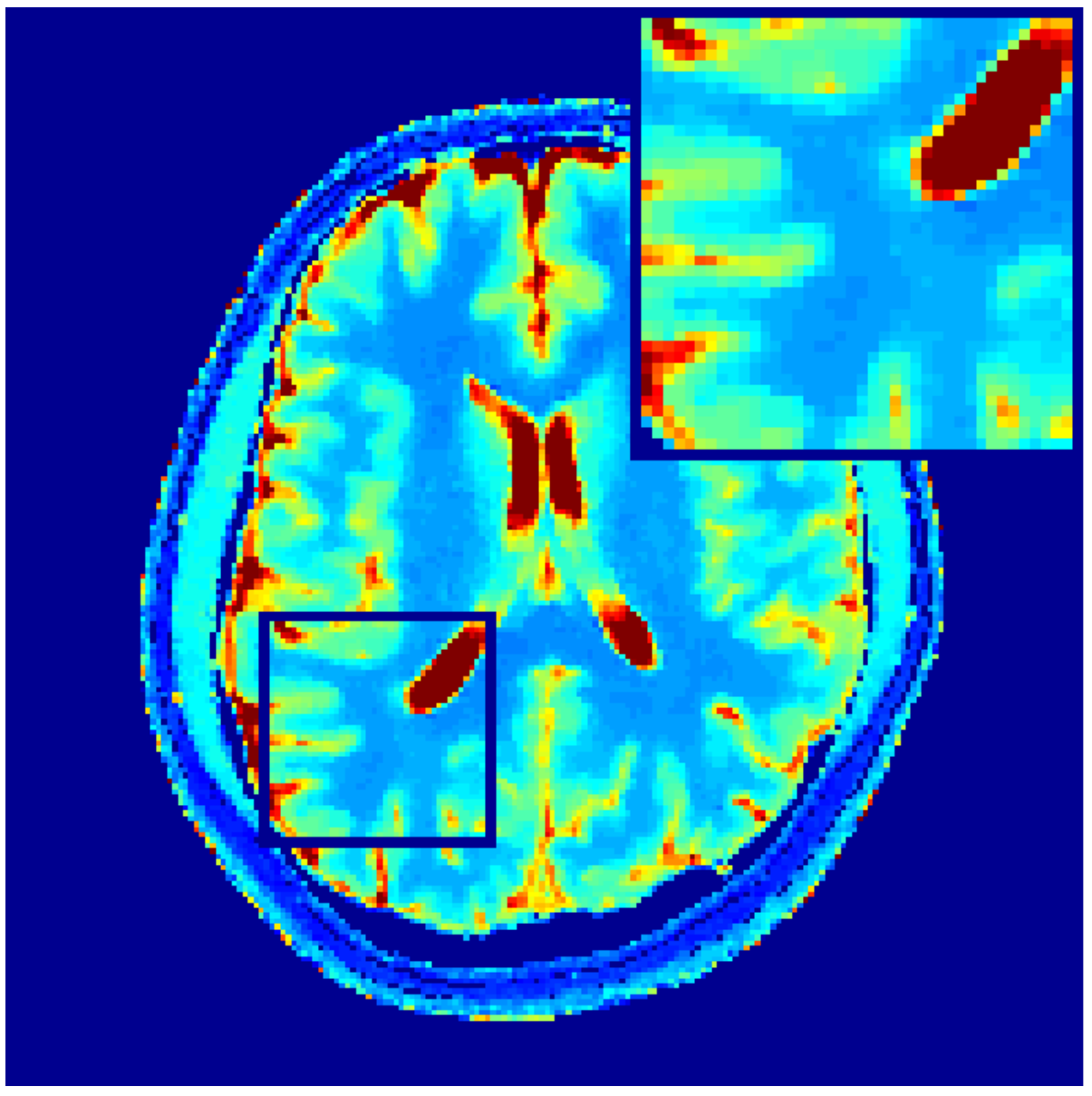}\hspace{-.1cm}
		\includegraphics[width=.3\linewidth]{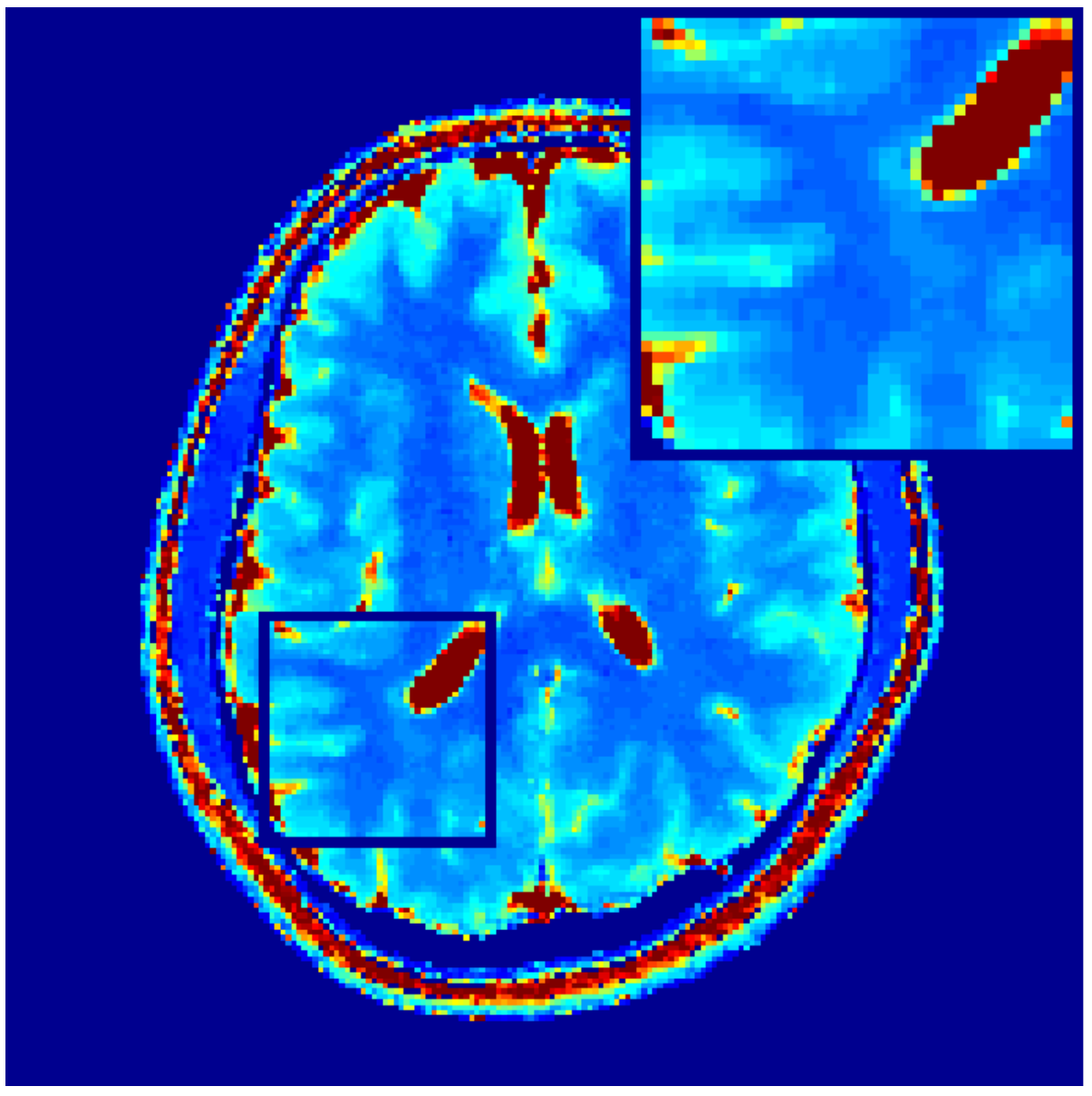}\hspace{-.1cm}
		\includegraphics[width=.3\linewidth]{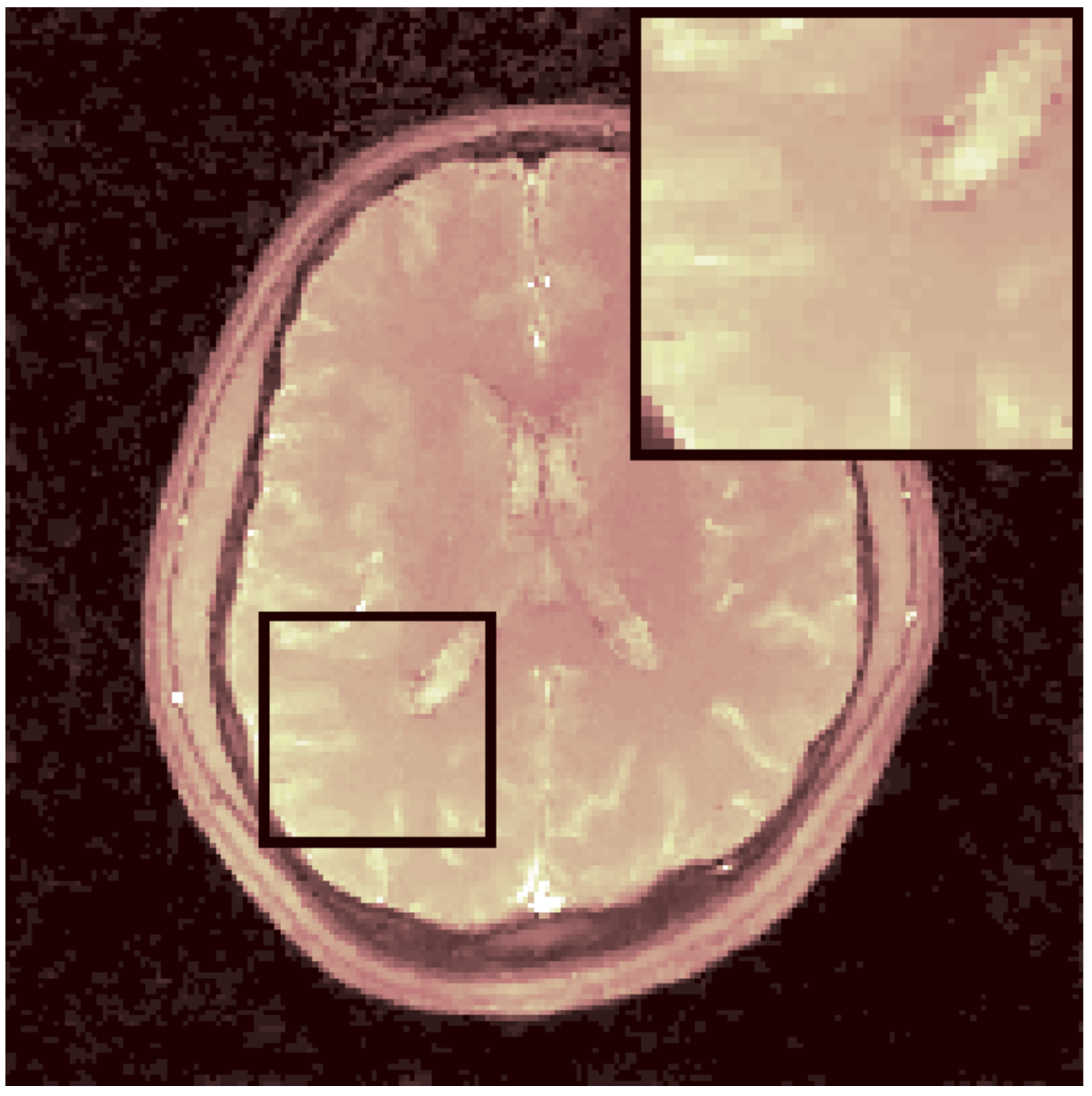}
		\\
		\begin{turn}{90} \quad\quad LRTV-KM\end{turn}
		\includegraphics[width=.3\linewidth]{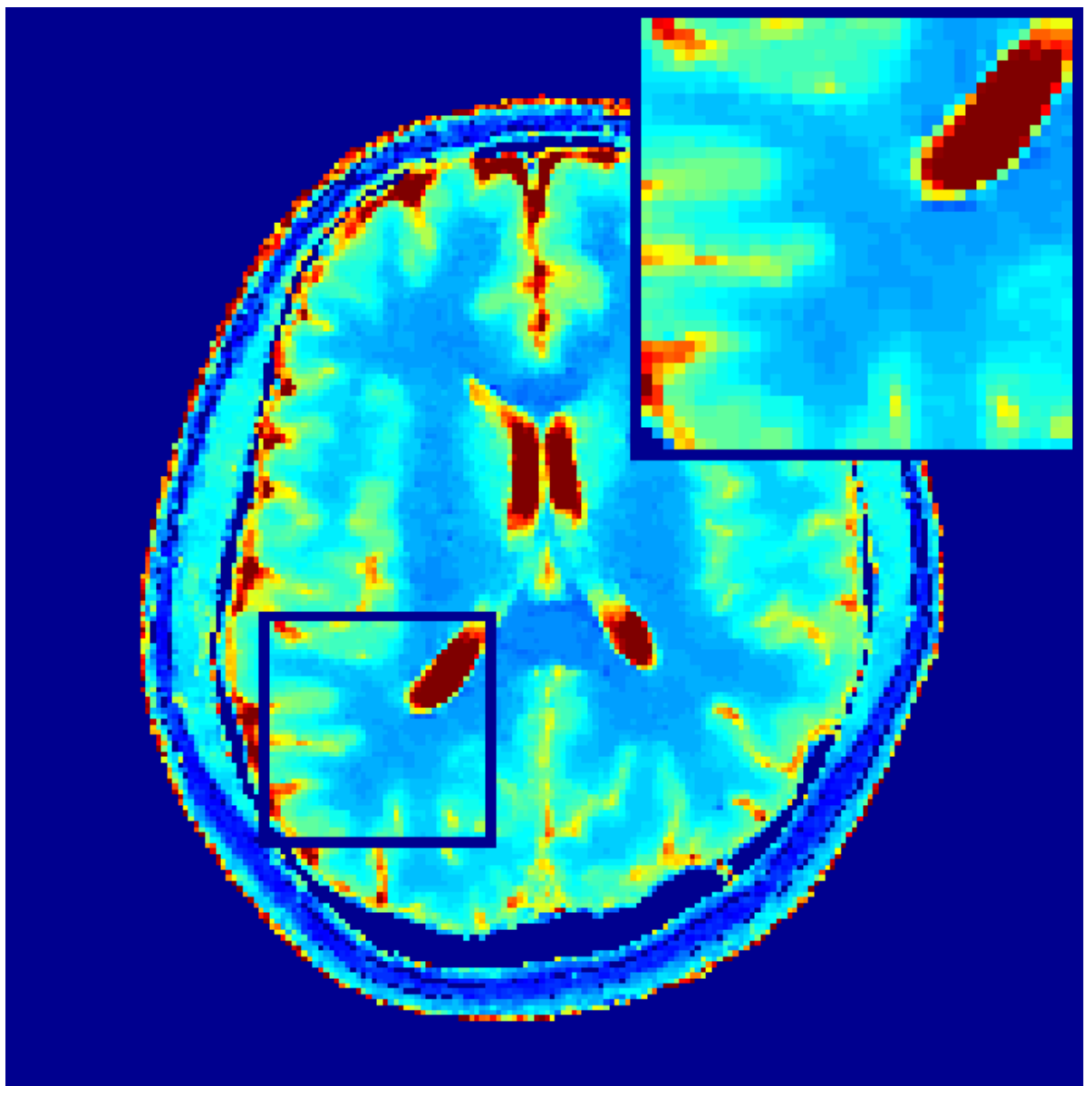}\hspace{-.1cm}
		\includegraphics[width=.3\linewidth]{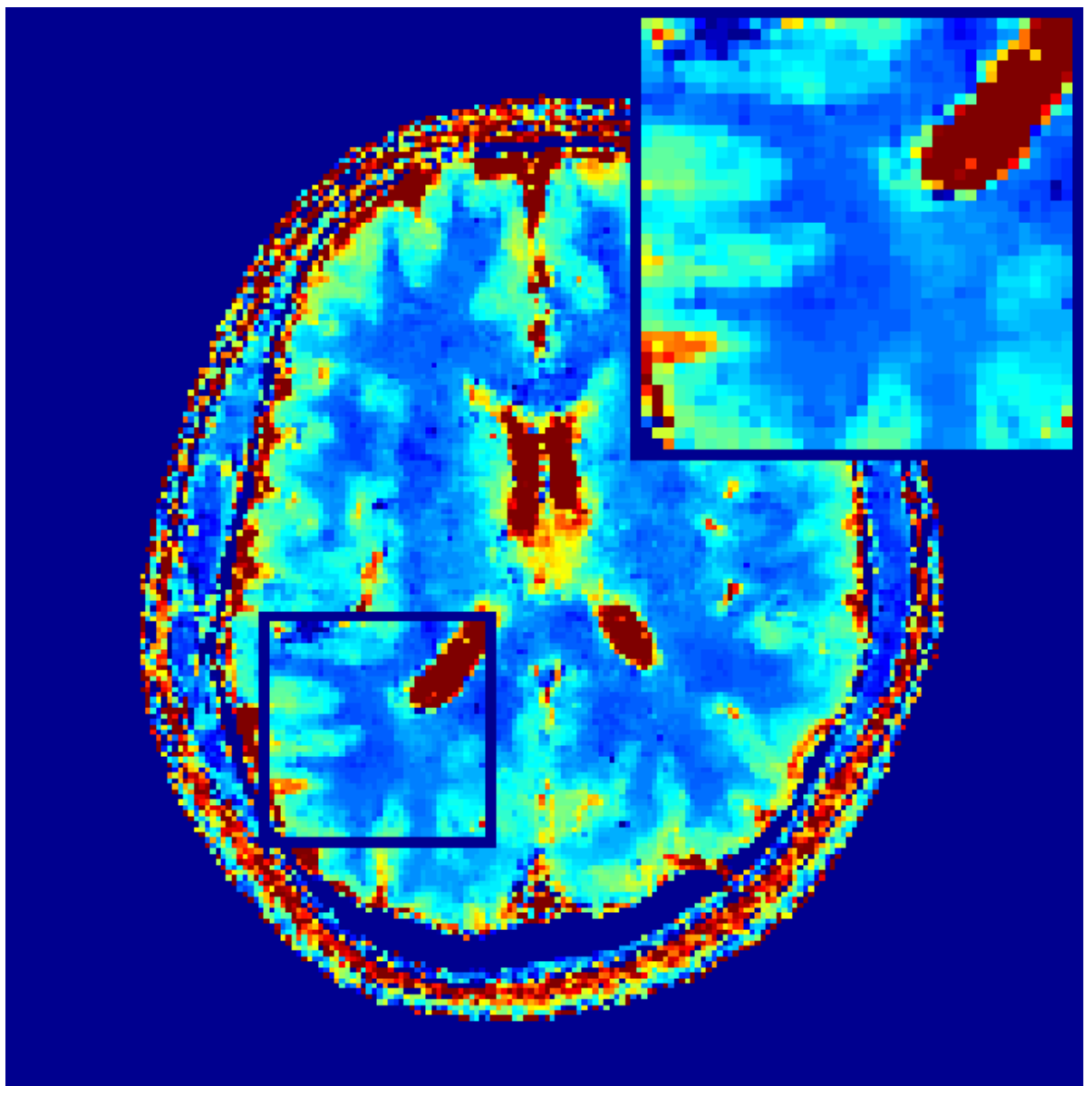}\hspace{-.1cm}
		\includegraphics[width=.3\linewidth]{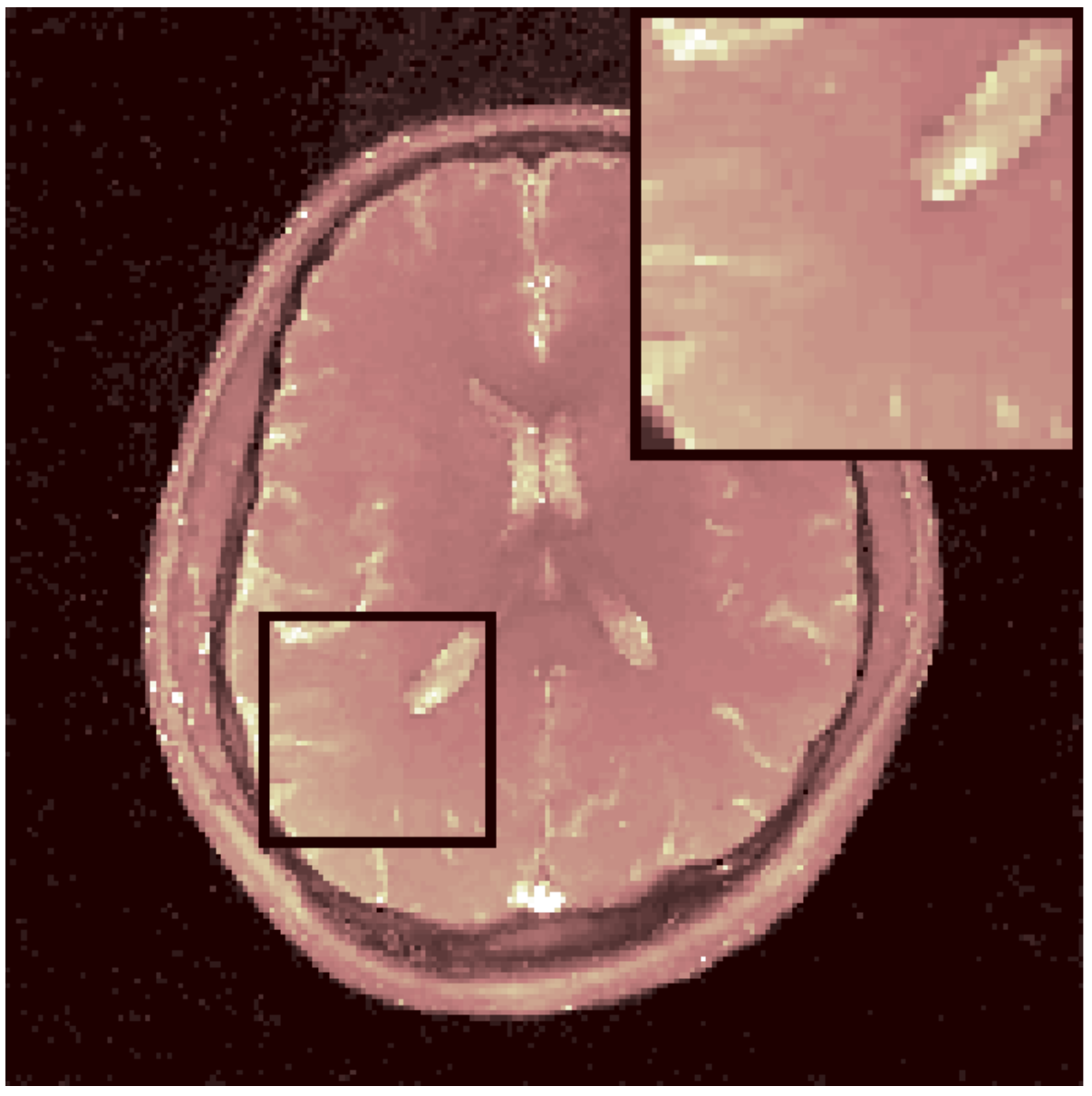}
		\\
		\begin{turn}{90} LRTV-MRFResnet\end{turn}
		\includegraphics[width=.3\linewidth]{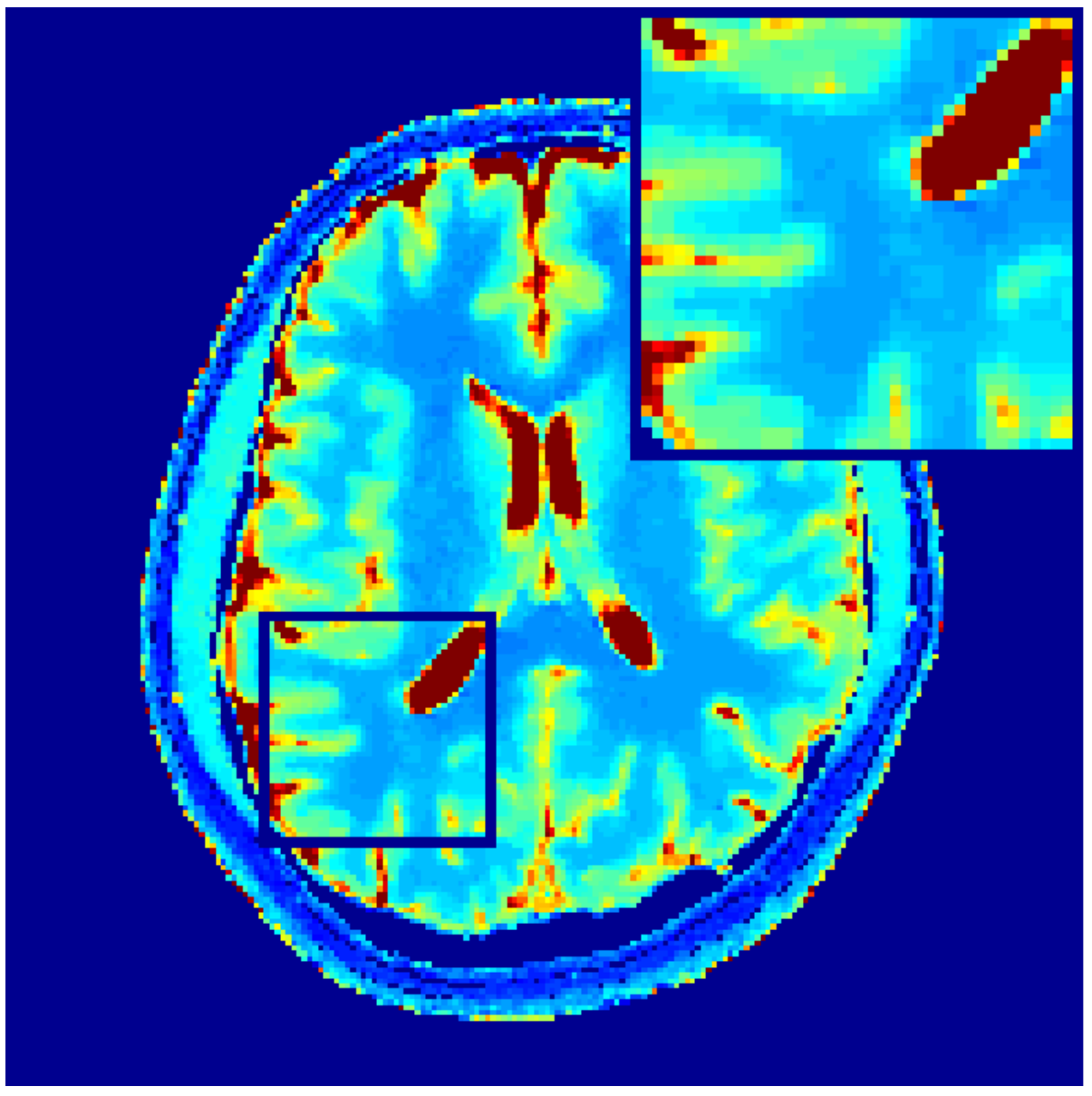}\hspace{-.1cm}
		\includegraphics[width=.3\linewidth]{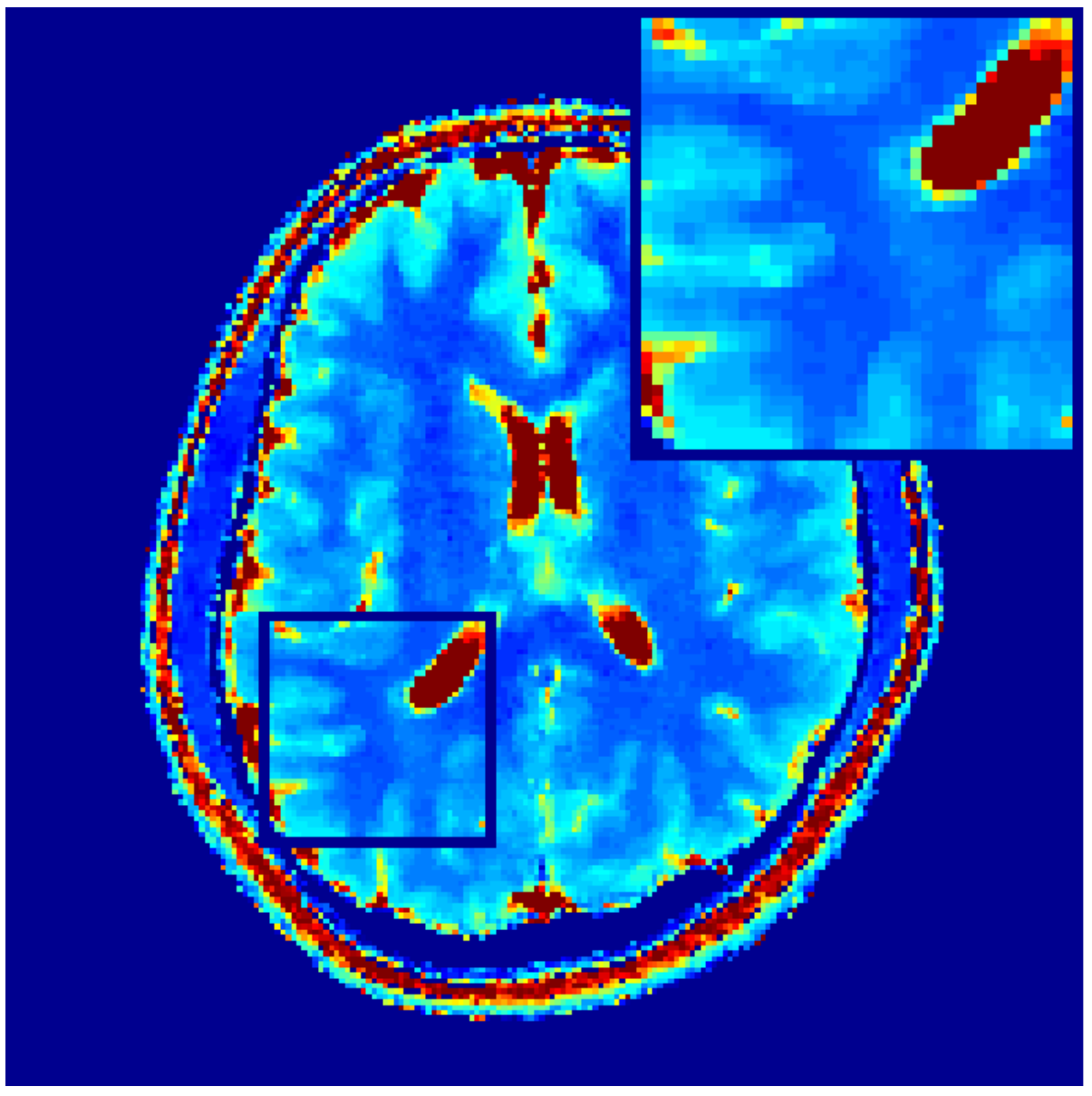}\hspace{-.1cm}
		\includegraphics[width=.3\linewidth]{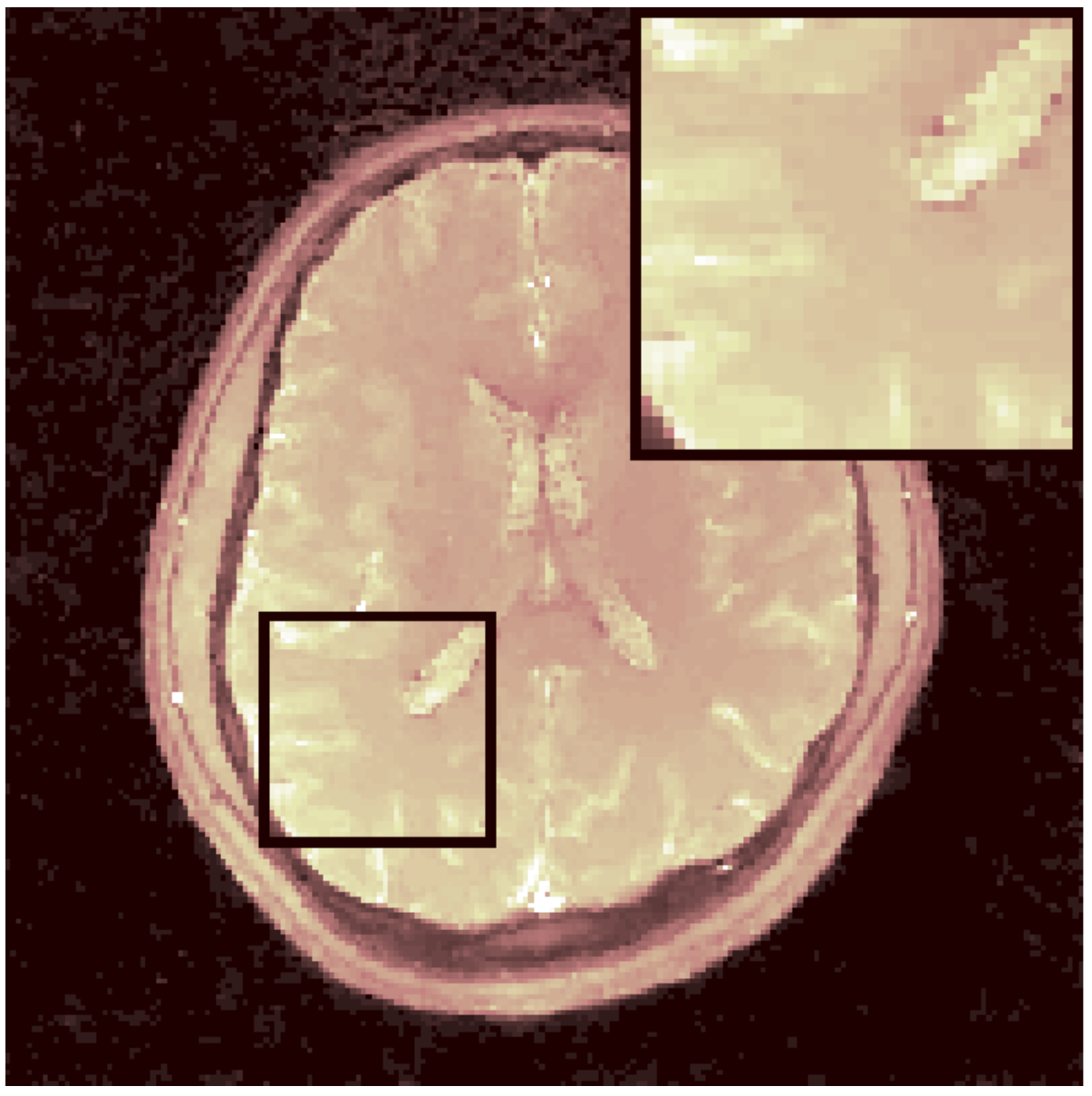}	
		\\
		\includegraphics[trim= -10 50 -20 720, clip,width=.3\linewidth]{./figs/T1barvivo.jpg}
		\includegraphics[trim= -10 50 -20 700, clip,width=.3\linewidth]{./figs/T2barvivo.jpg}
		\includegraphics[trim= -10 50 -20 700, clip,width=.3\linewidth]{./figs/PDbarvivo.jpg}\vspace{.5cm}
				\\
		T1(s) \hspace{1.6cm} T2 (s) \hspace{1.6cm}  PD (a.u.) \\
		\caption{\footnotesize{Reconstructed T1, T2 and PD maps from a 2D radial acquisition (real-world scan) using different reconstruction and inference algorithms.}  \label{fig:2dvivo_radial}}
\end{minipage}}
\end{figure}


\begin{figure*}[t!]
	\centering
	\scalebox{.85}{
	\begin{minipage}{\textwidth}
		\centering
		\includegraphics[trim= 25 15 25 25, clip, width=.16\linewidth]{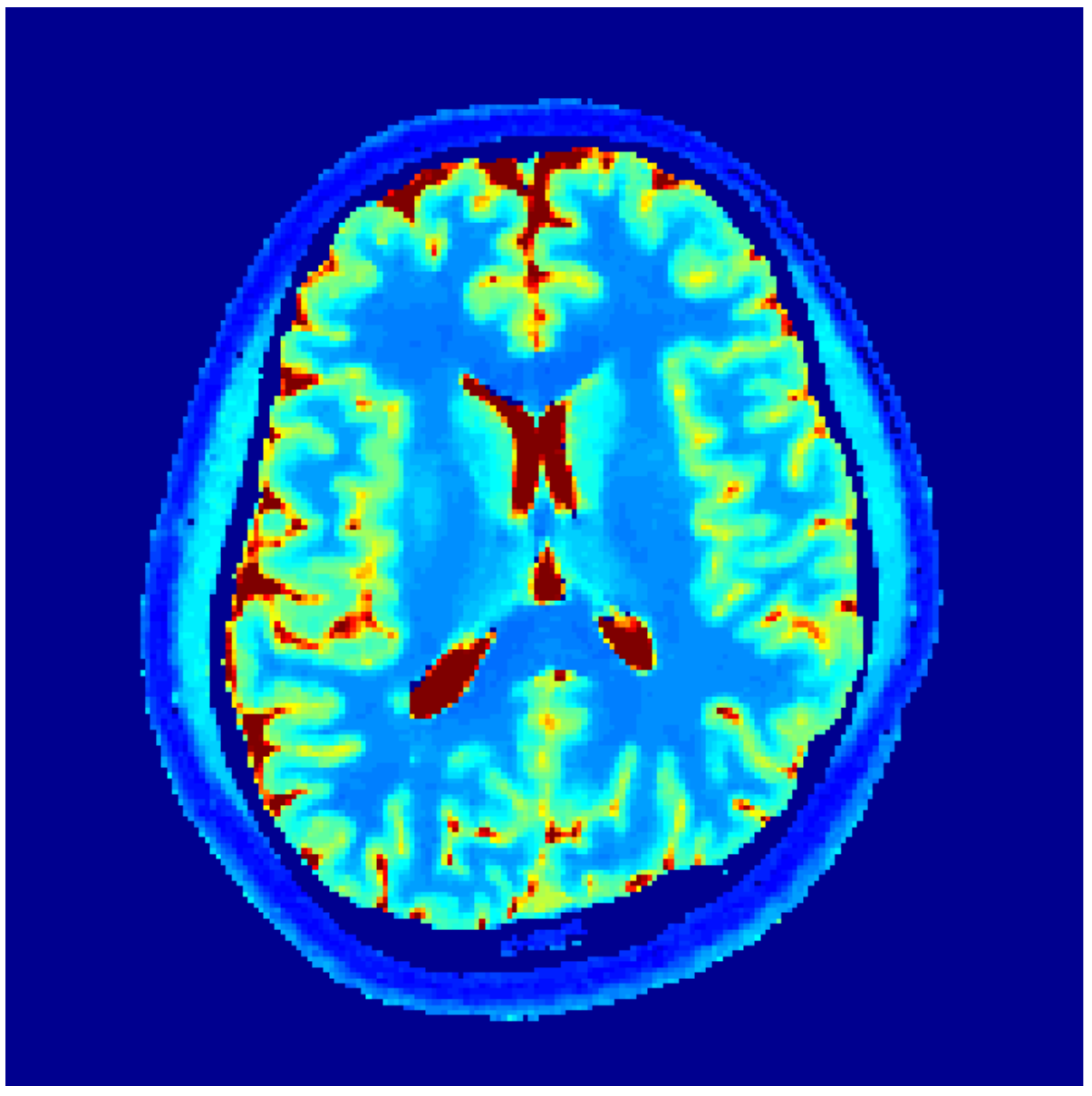}\hspace{-.05cm}
		\includegraphics[trim= 25 15 25 25, clip, width=.16\linewidth]{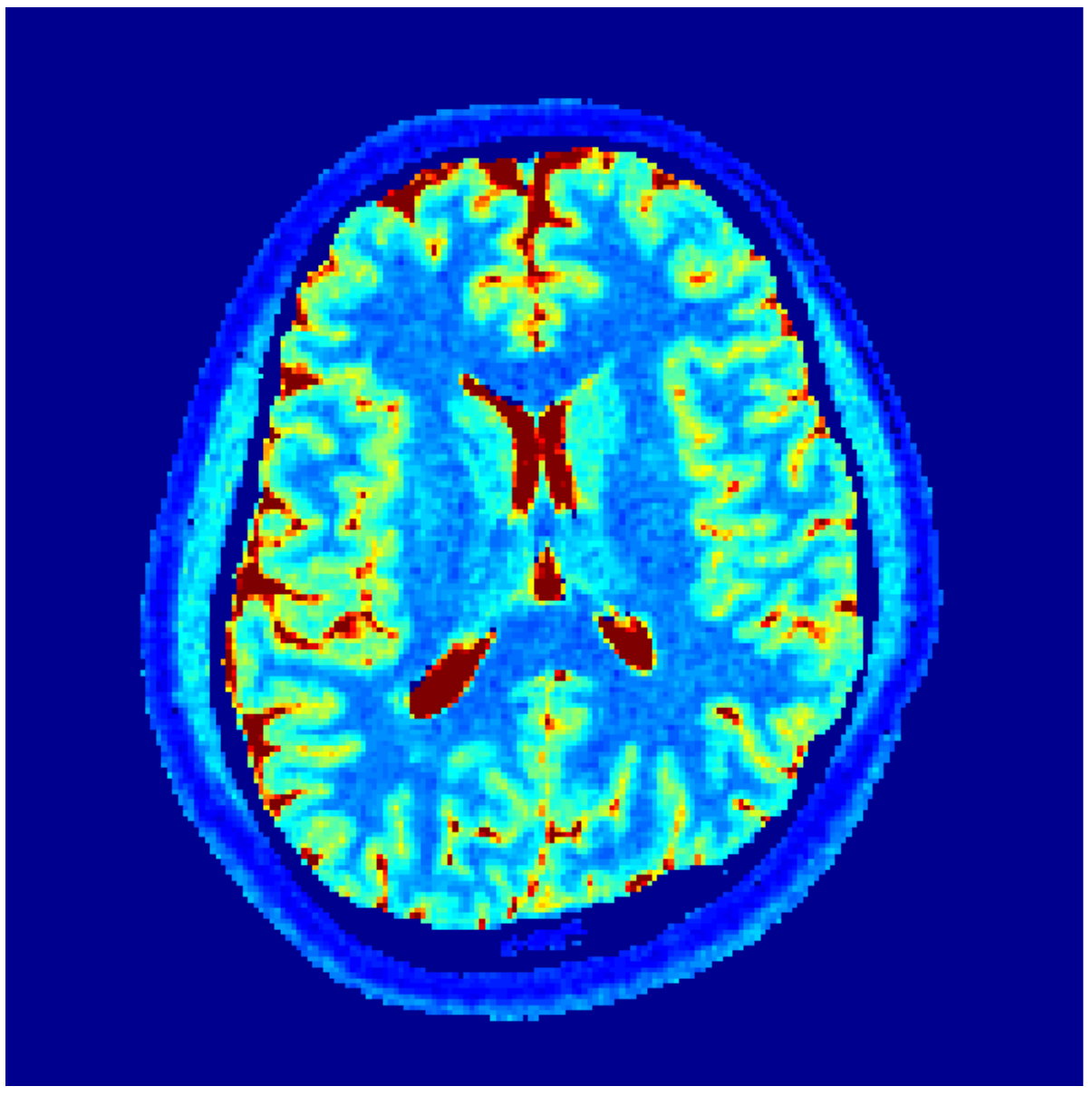}\hspace{.15cm}
		\includegraphics[trim= 25 15 25 25, clip, width=.16\linewidth]{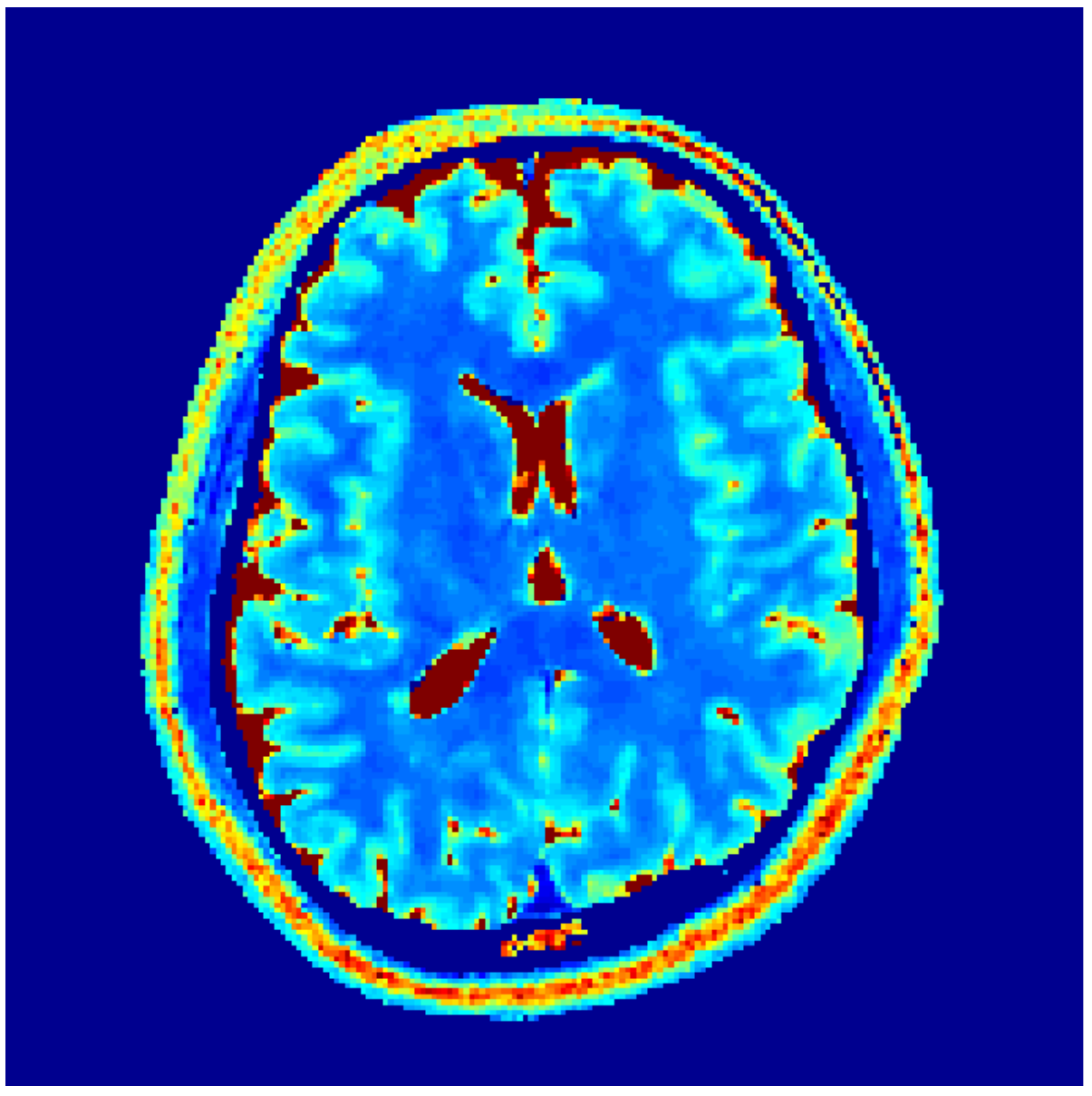}\hspace{-.05cm}	
		\includegraphics[trim= 25 15 25 25, clip, width=.16\linewidth]{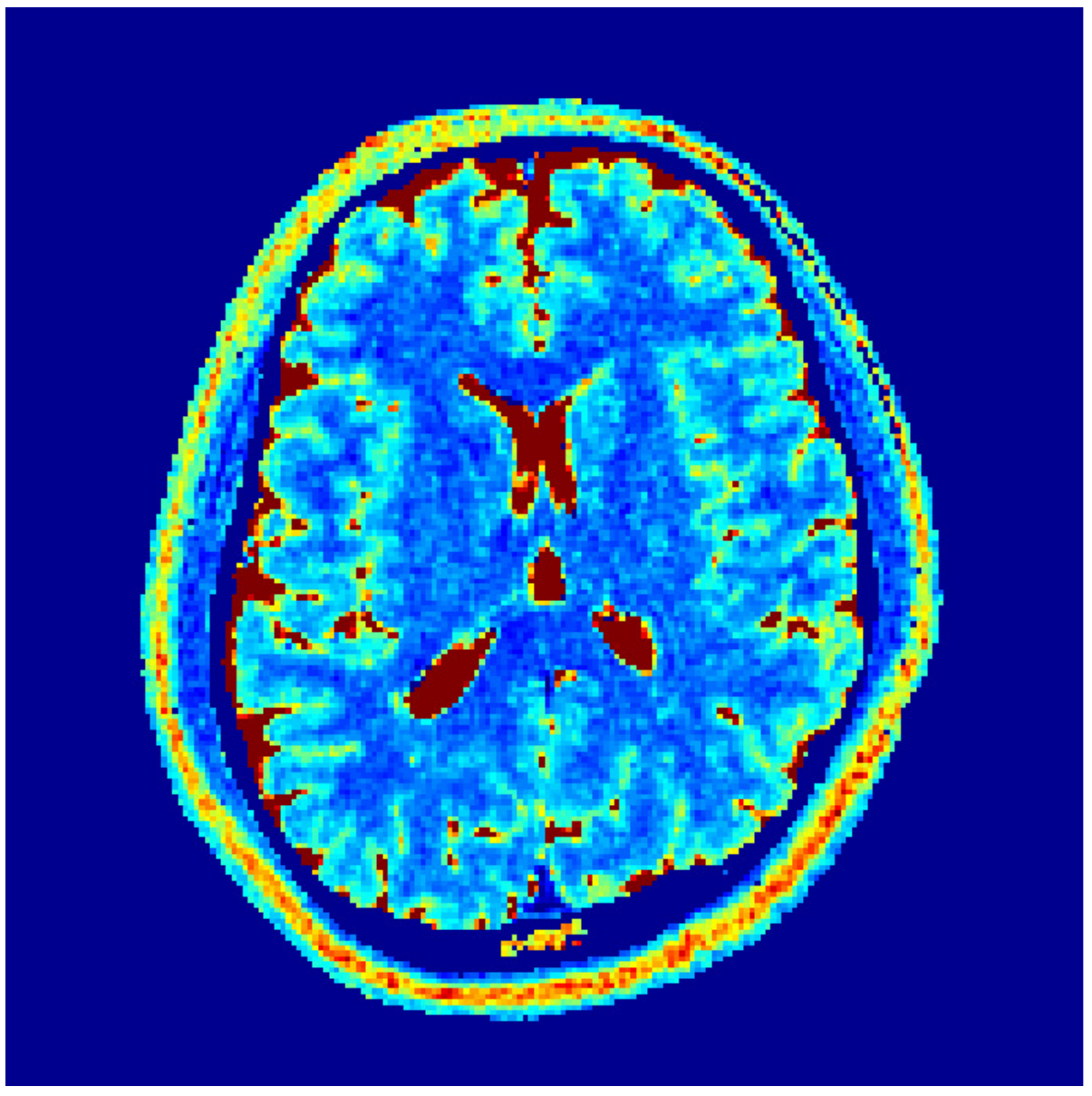}\hspace{.15cm}
		\includegraphics[trim= 25 15 25 25, clip, width=.16\linewidth]{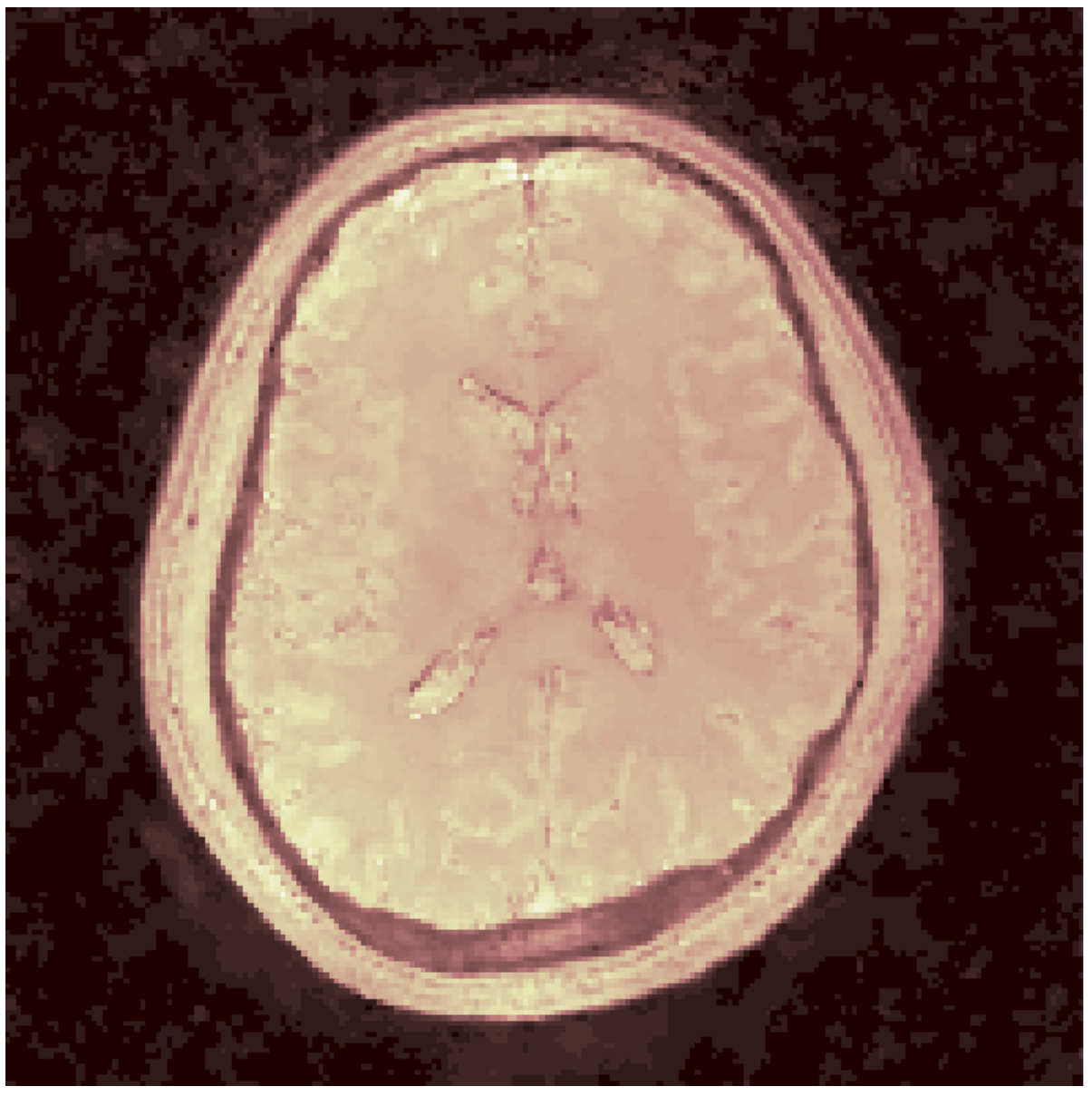}\hspace{-.05cm}	
		\includegraphics[trim= 25 15 25 25, clip, width=.16\linewidth]{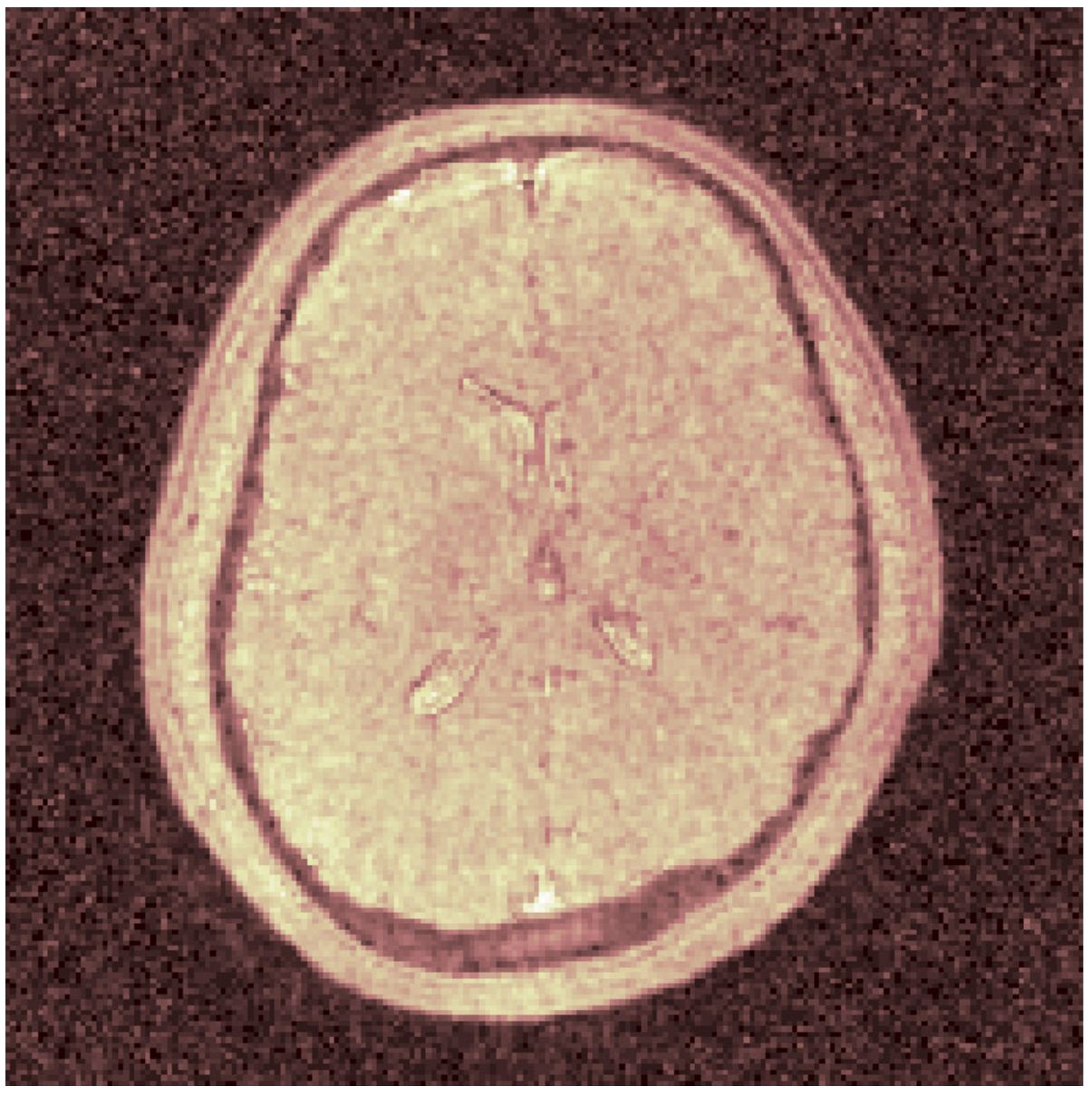}
		\\
		\vspace{.05cm}
		\includegraphics[trim= 22 0 22 60, clip, width=.16\linewidth]{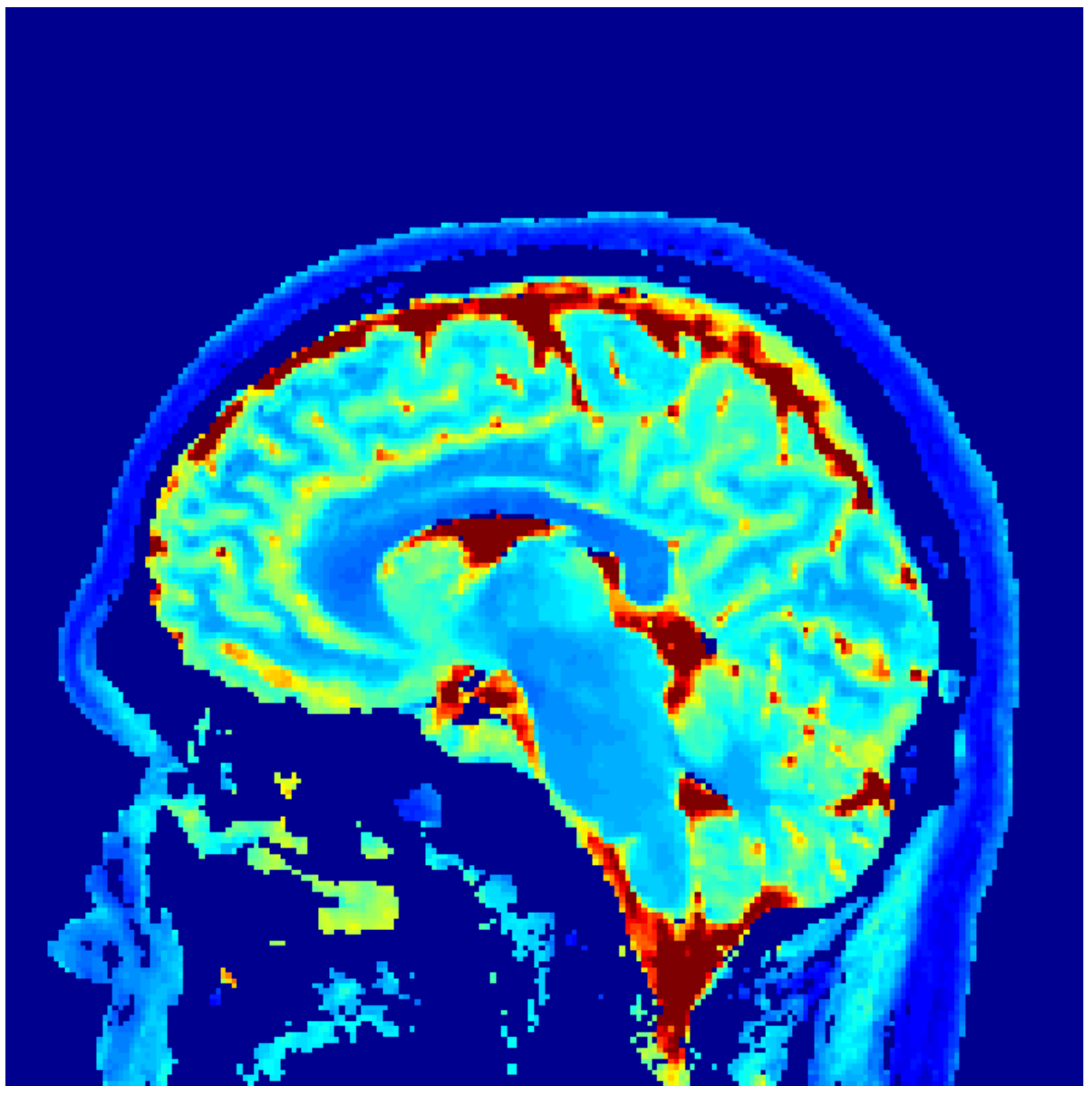}\hspace{-.05cm}
		\includegraphics[trim= 22 0 22 60, clip, width=.16\linewidth]{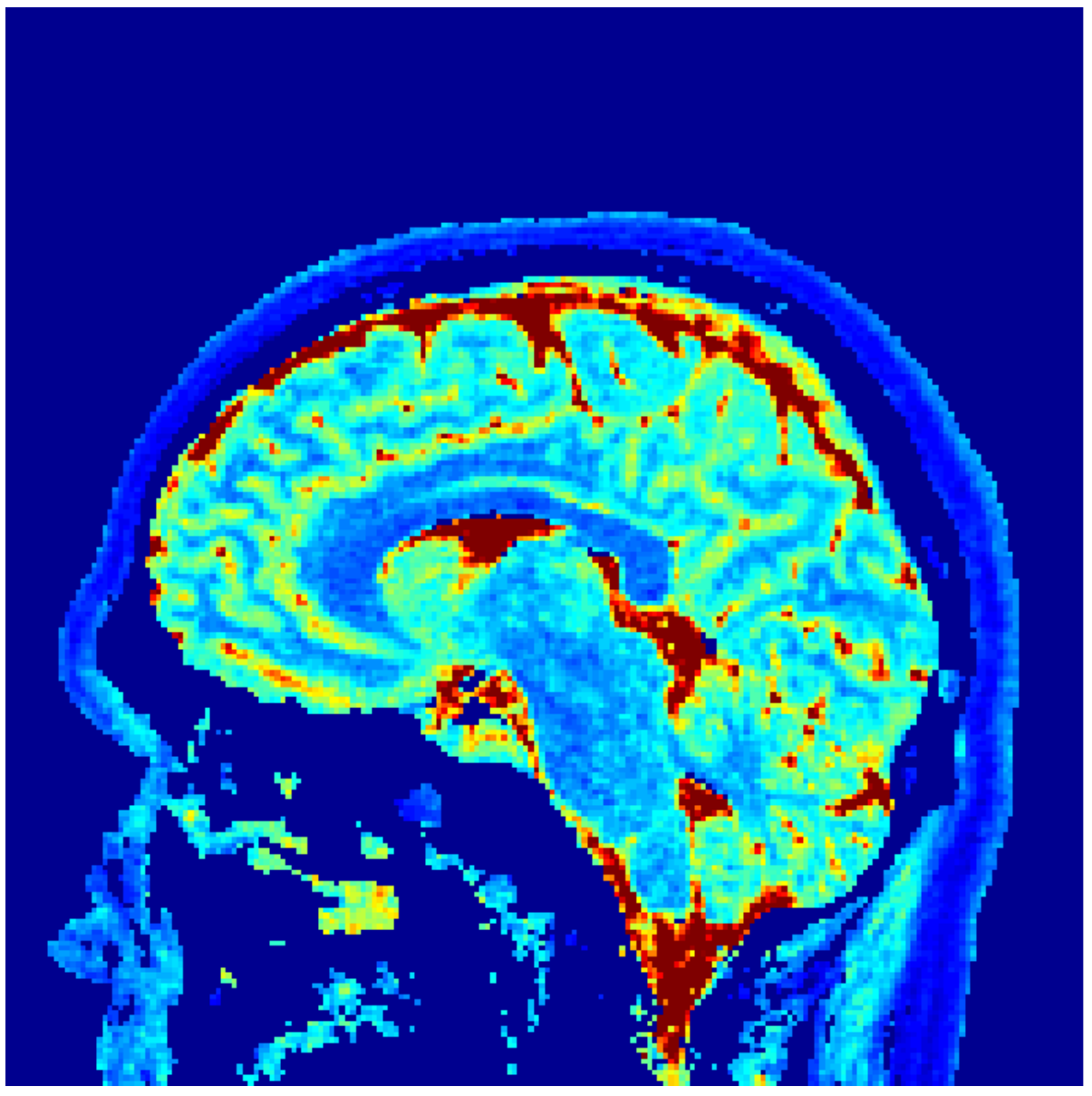}\hspace{.15cm}
		\includegraphics[trim= 22 0 22 60, clip, width=.16\linewidth]{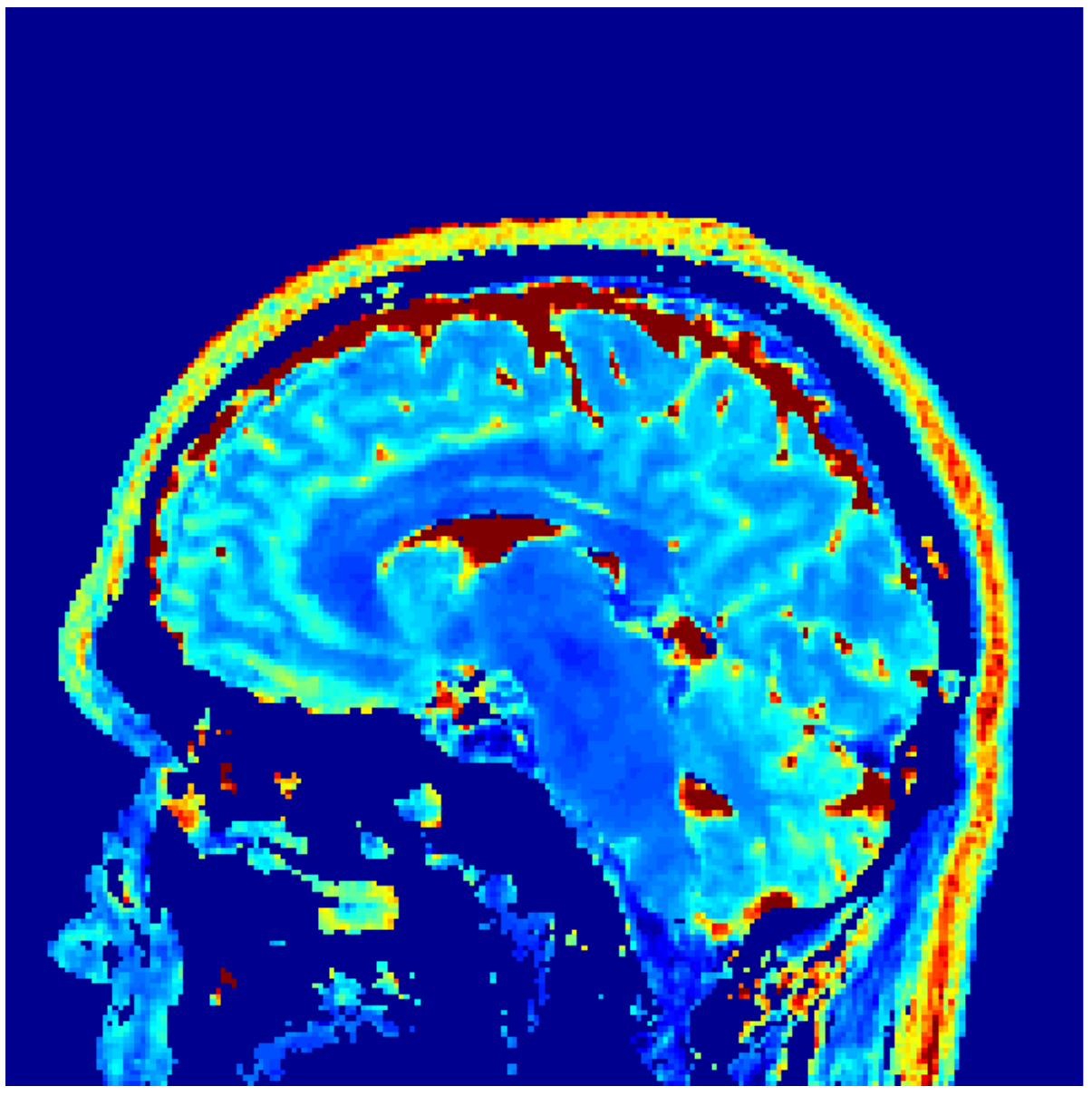}\hspace{-.05cm}
		\includegraphics[trim= 22 0 22 60, clip, width=.16\linewidth]{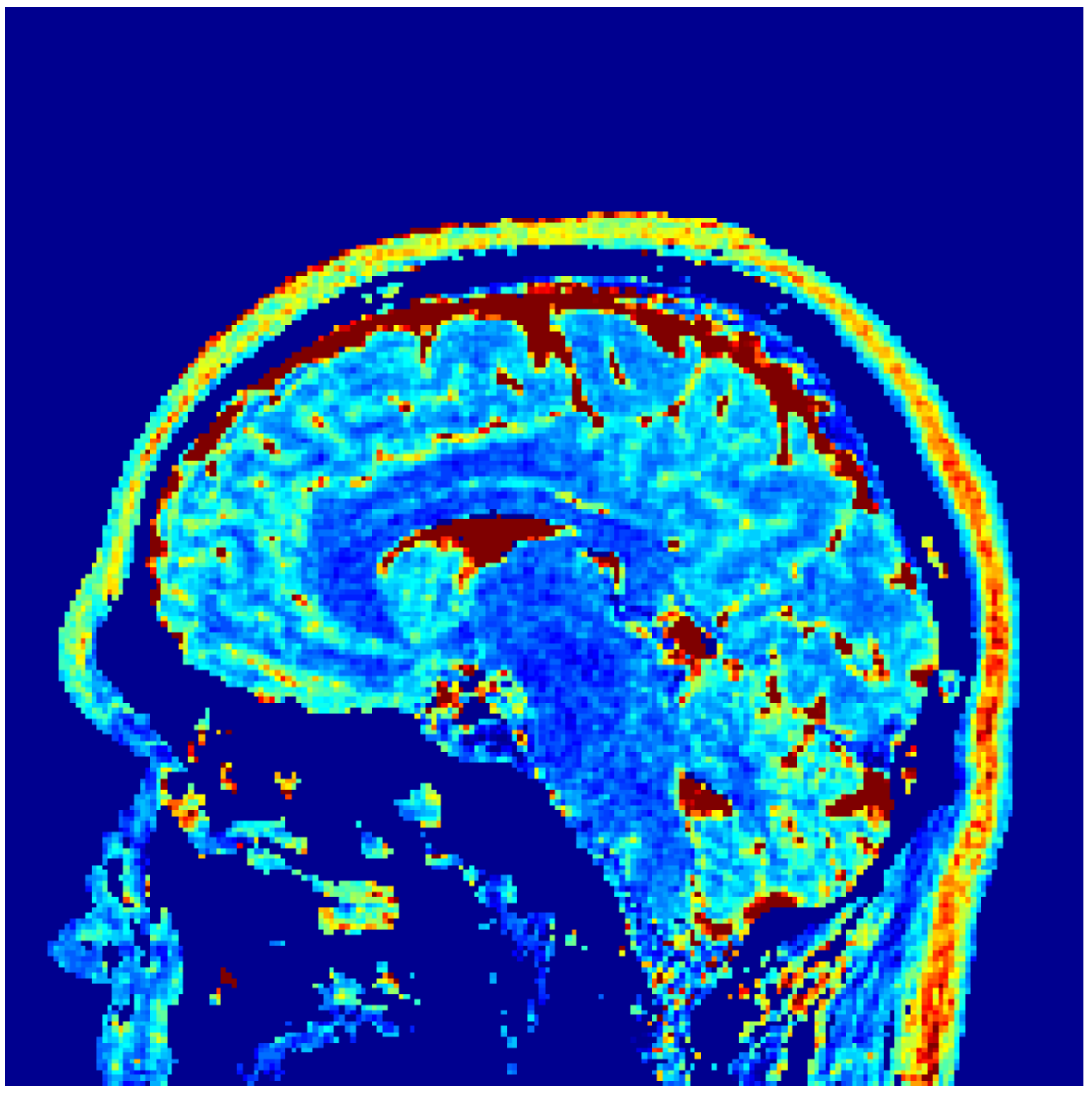}\hspace{.15cm}
		\includegraphics[trim= 22 0 22 60, clip, width=.16\linewidth]{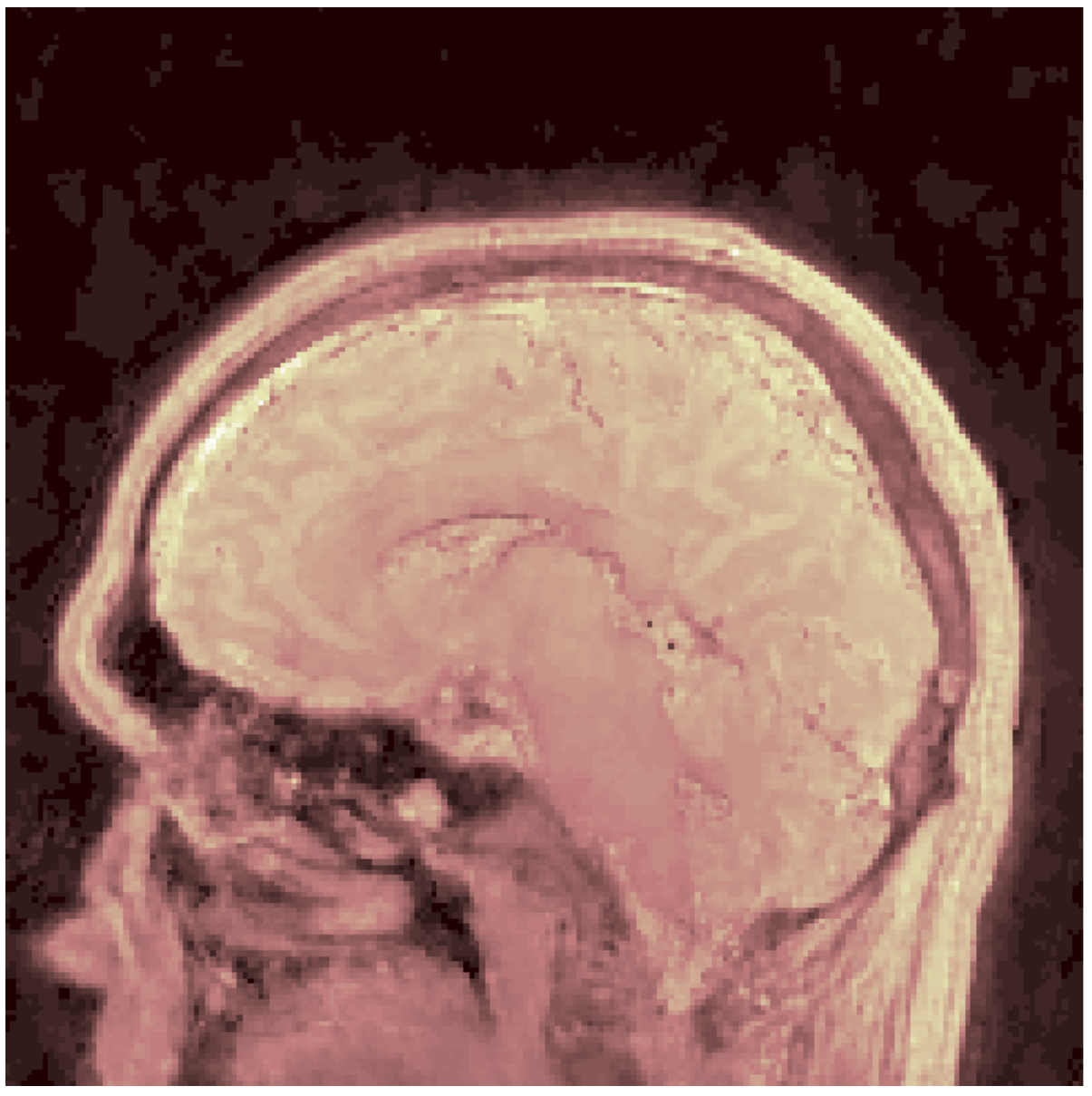}\hspace{-.05cm}
		\includegraphics[trim= 22 0 22 60, clip, width=.16\linewidth]{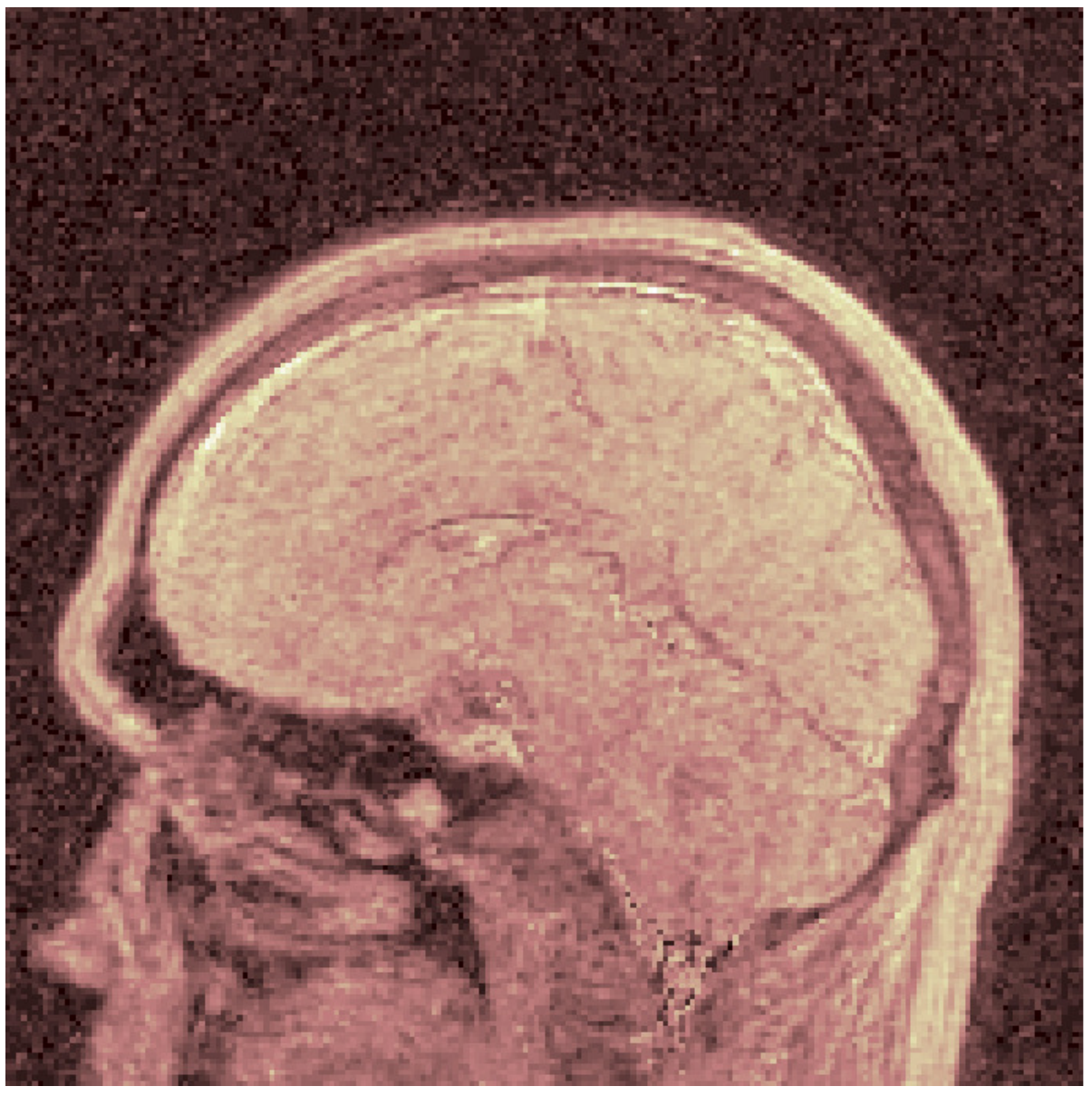}
		\\
		\vspace{.02cm}
		\includegraphics[trim= 25 0 25 60, clip, width=.16\linewidth]{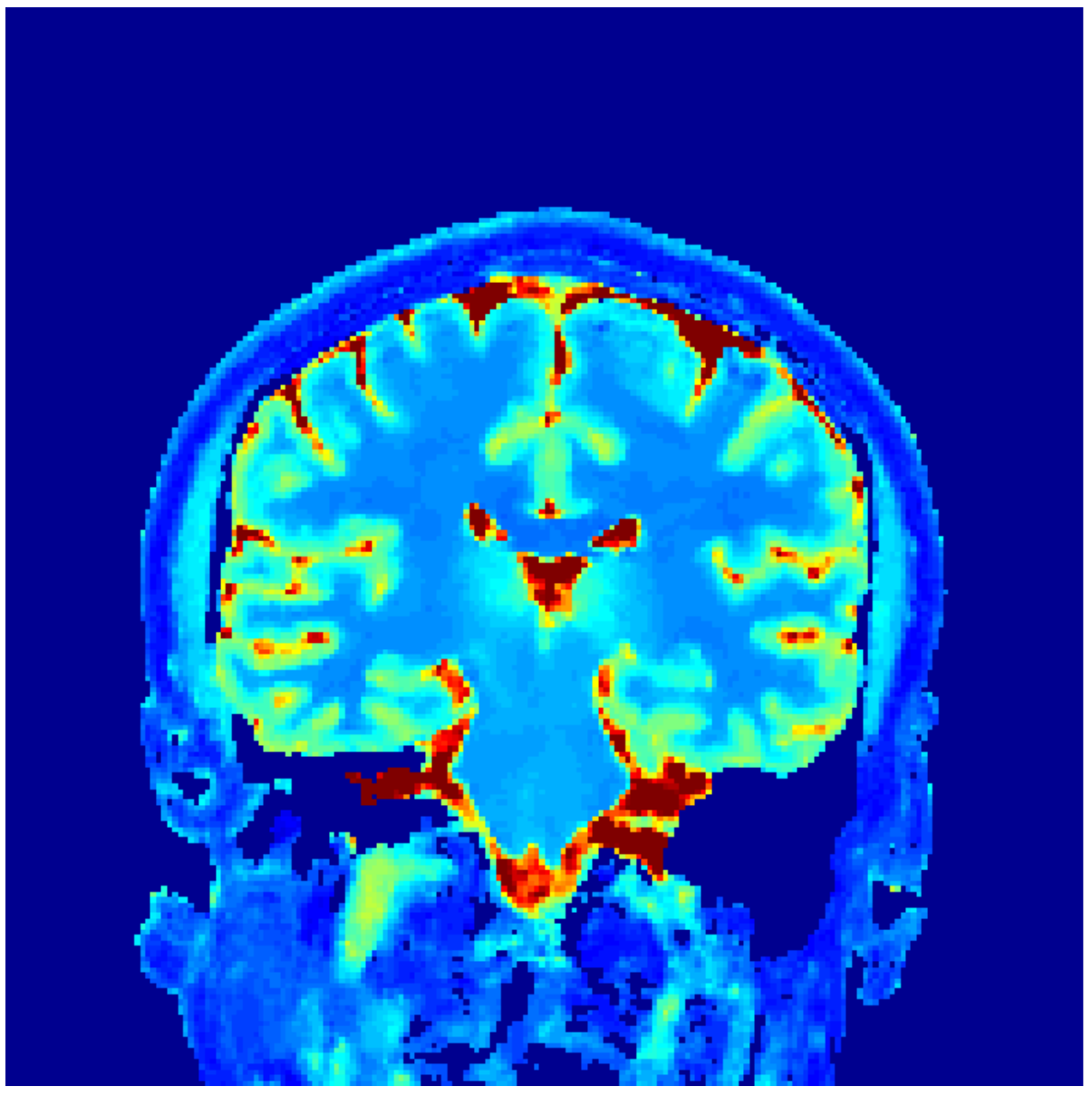}\hspace{-.05cm}	
		\includegraphics[trim= 25 0 25 60, clip, width=.16\linewidth]{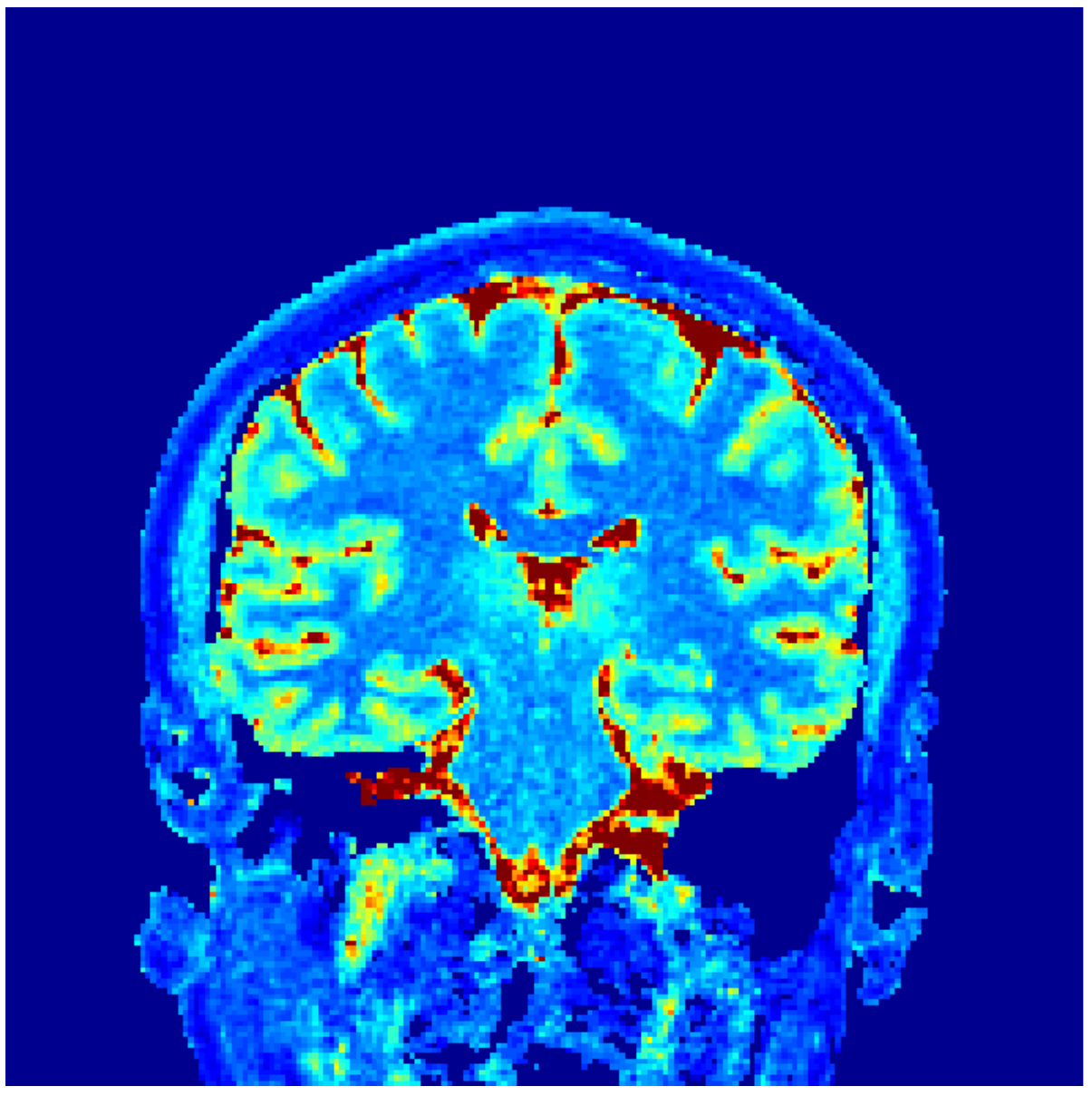}\hspace{.15cm}
		\includegraphics[trim= 25 0 25 60, clip, width=.16\linewidth]{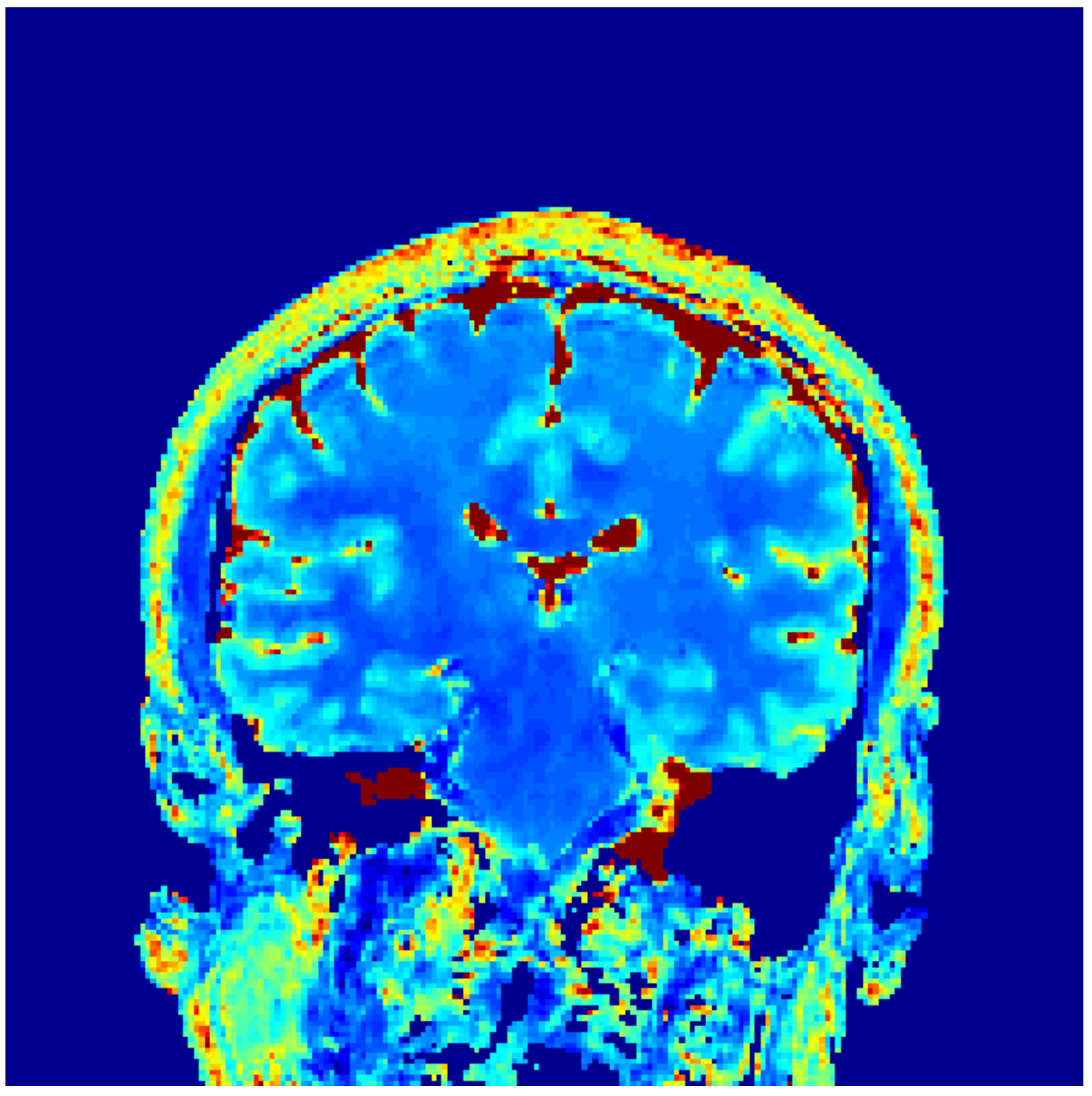}\hspace{-.05cm}
		\includegraphics[trim= 25 0 25 60, clip, width=.16\linewidth]{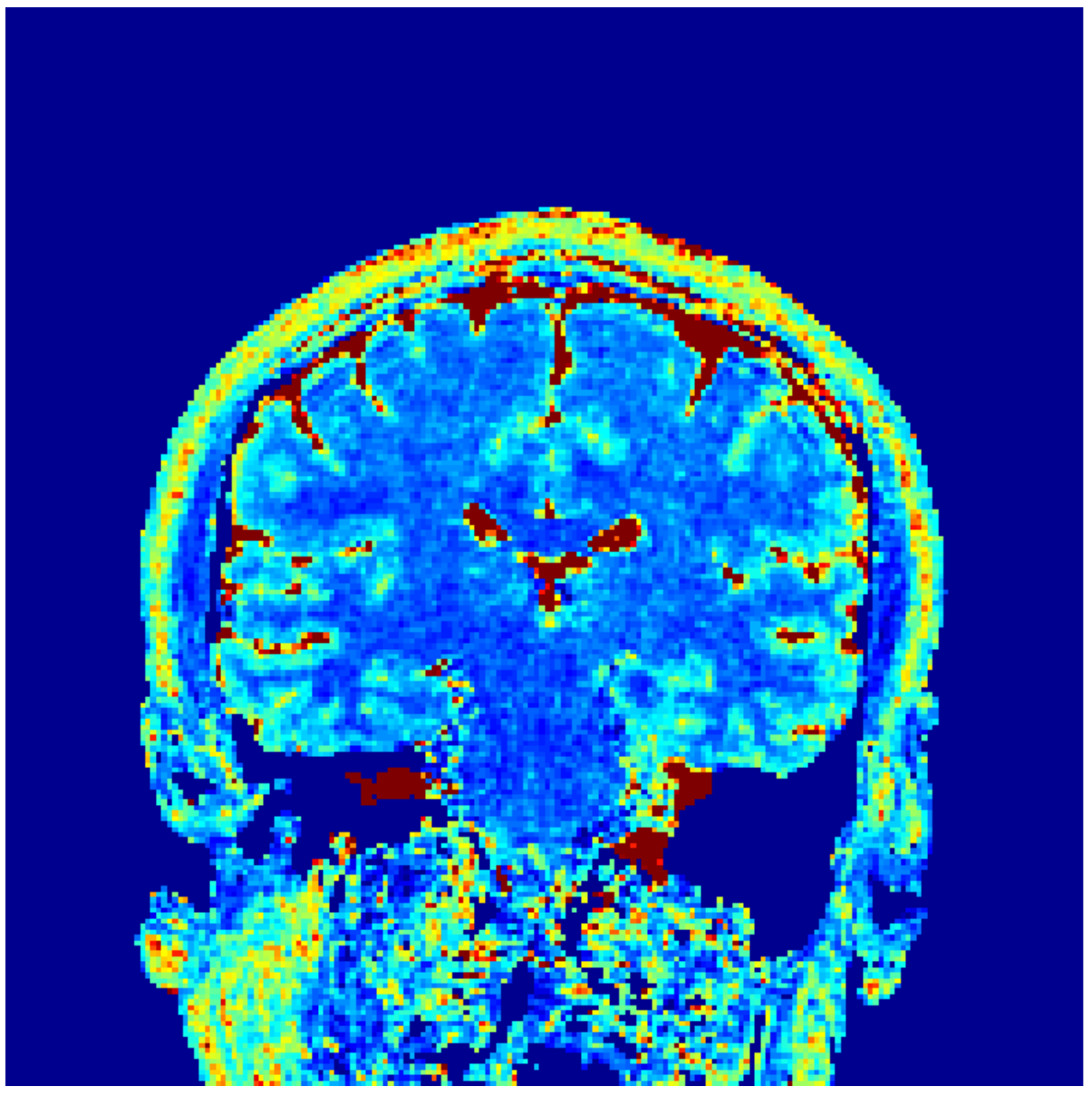}\hspace{.15cm}
		\includegraphics[trim= 25 0 25 60, clip, width=.16\linewidth]{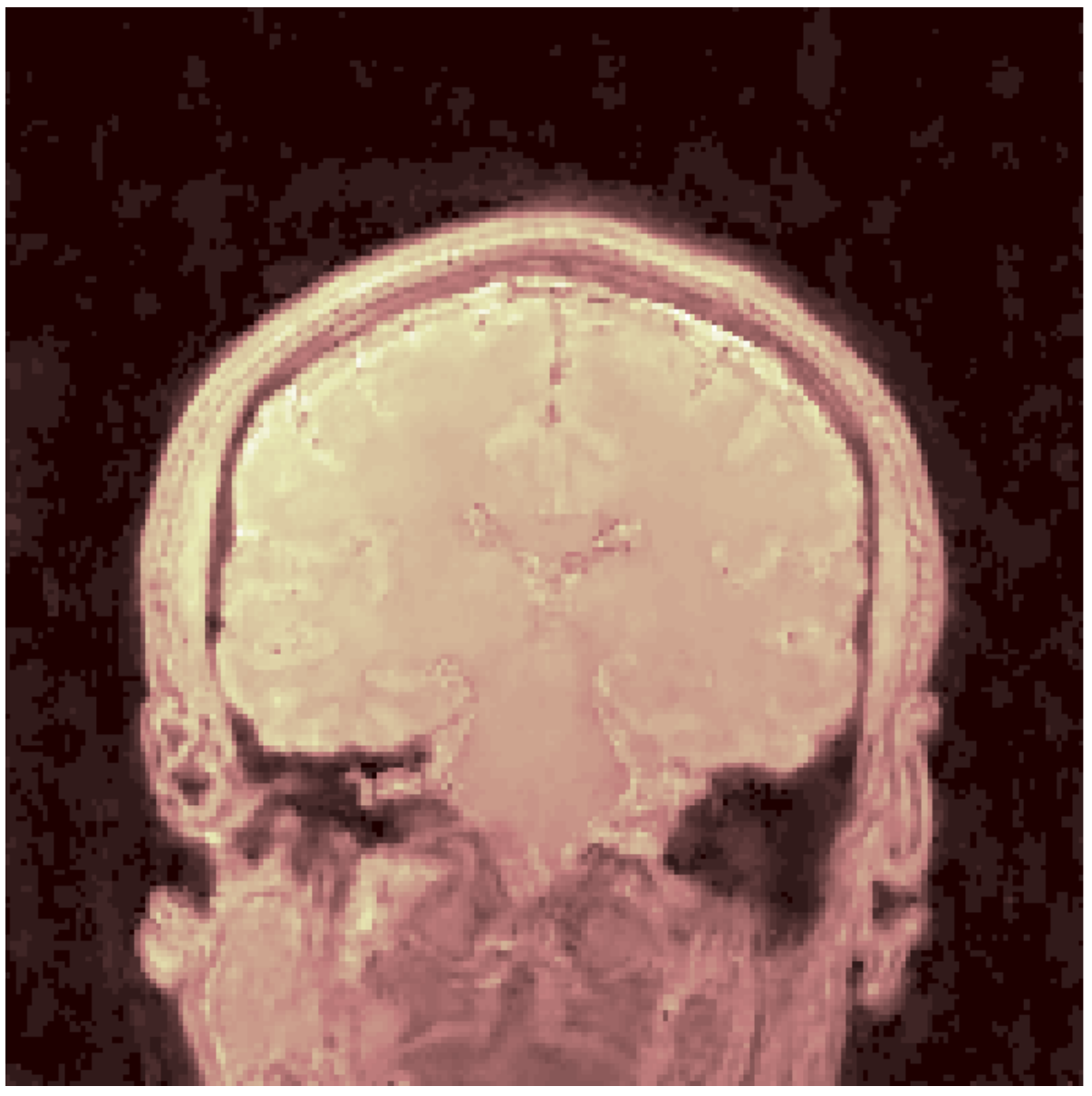}\hspace{-.05cm}	
		\includegraphics[trim= 25 0 25 60, clip, width=.16\linewidth]{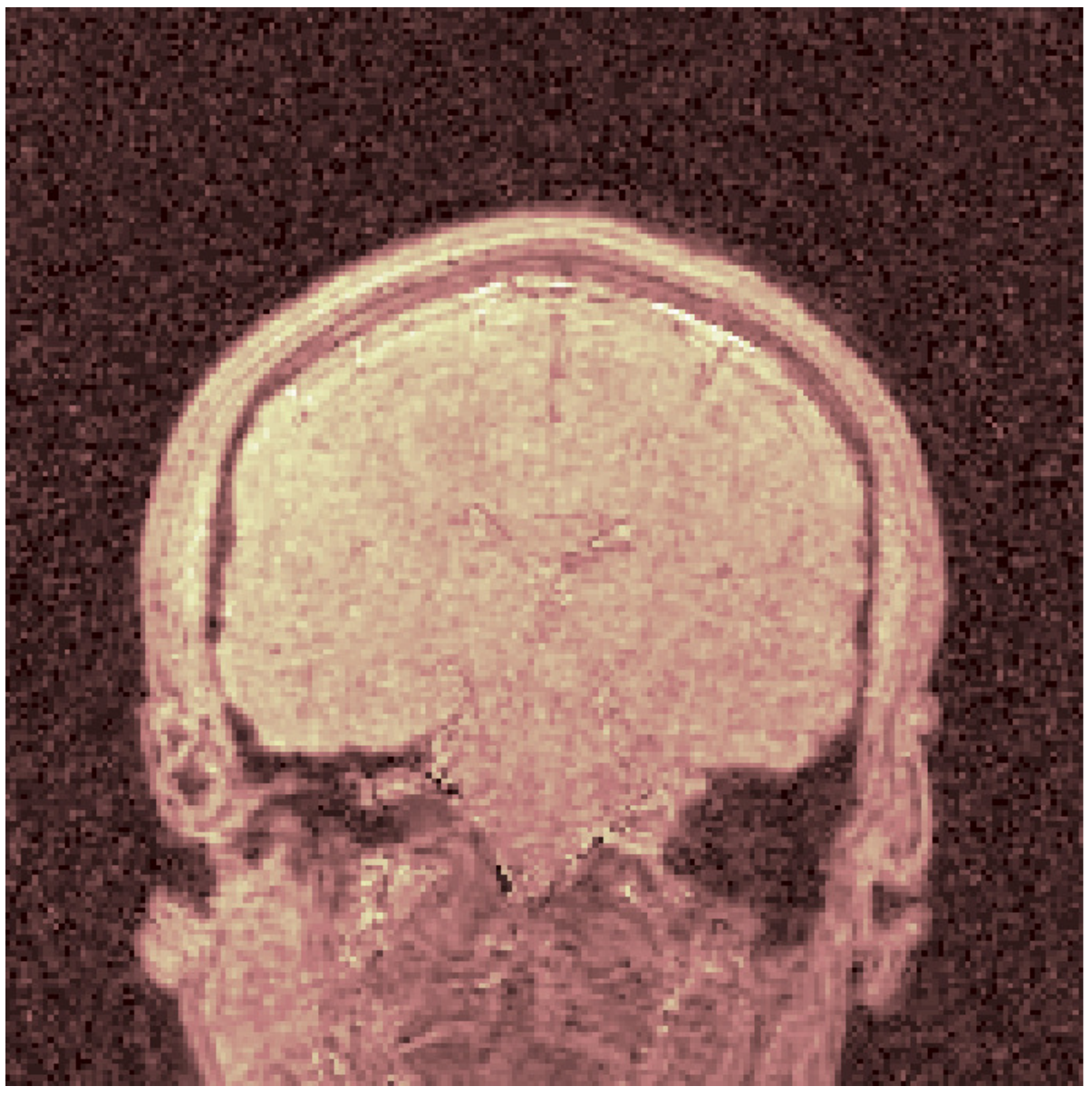}
			\\
\vspace{-.3cm}
		\includegraphics[trim= -10 50 -20 720, clip,width=.32\linewidth]{./figs/T1barvivo.jpg}
		\includegraphics[trim= -10 50 -20 700, clip,width=.32\linewidth]{./figs/T2barvivo.jpg}
		\includegraphics[trim= -10 50 -20 700, clip,width=.32\linewidth]{./figs/PDbarvivo.jpg}\vspace{1cm}
				\\
		T1(s) \hspace{4.5cm} T2 (s) \hspace{4.8cm}  PD (a.u.) \\		
		\caption{Reconstructed T1, T2 and PD maps using 3D acquisition (real-world scan) with spiral readouts. The (zoomed) 3D maps are computed using LRTV-MRFResnet (left sub-column) and the VS-DM (right sub-column) algorithms.\vspace{.2cm}}  \label{fig:3dvivo}
	\end{minipage}}
\end{figure*}


\subsection{In-vivo experiments}
\label{sec:vivoexpe}
Two sets of experiments were conducted here: 
first, we used the 2D and 3D acquisition sequences for scanning a healthy volunteer's brain (real-world acquisitions).
Figures~\ref{fig:2dvivo_spiral} and \ref{fig:2dvivo_radial} display the parametric maps reconstructed from 2D spiral and radial readouts. We computed the T1, T2 and proton density (PD) maps using baseline reconstruction algorithms ZF, VS, LR, FLOR, AIR-MRF and our proposed LRTV.
While baselines use DM either for quantitative inference or also during reconstruction (i.e. AIR-MRF), we further compare the DM-free LRTV's performance when cascaded to DM, KM and MRFResnet for quantitative inference.
For the 3D spiral acquisitions we compared LRTV and its closest competitor VS in Figure~\ref{fig:3dvivo}. Outcomes from other tested algorithm are displayed in the supplementary materials (Figure~S5). 
Since FLOR does not use dimensionality-reduction, our system ran out of memory during 3D reconstruction; hence results are not reported in this case.

Due to the lack of ground truth in the first experiment, 
we conducted a second experiment with retrospective validations against the ground-truth anatomical maps. Here the gold standard anatomical maps were acquired from a separate volunteer using MAGIC quantitative imaging protocol~\cite{magic2015}.  Figure~S1 
in supplementary materials shows these ground truth parameter maps. From these parametric maps we then constructed the corresponding TSMIs and noisy MRF measurements. We  used the same excitation sequence and the 2D spiral and radial k-space sampling patterns as in our real-world scans to construct a single-coil acquisition where the k-space measurements were corrupted by additive Gaussian noise with 35 dB SNR. Table~\ref{tab:expe_retro} compares the reconstruction accuracies of the baselines against our proposed algorithm for the T1, T2, and PD maps as well as the TSMI images. Accuracies were measured by the MAE and MAPE errors, reconstruction SNRs, and the Structural Similarity Index Metric (SSIM)~\cite{wang2004image}. Reconstructed maps and their differences with respect to the ground-truth can be found in Figures S2 and S3 
in the supplementary materials.

\subsubsection{Discussion}
The LRTV-DM and LRTV-MRFResnet perform on par, and both outperform all tested baselines for reconstructing T1, T2 and PD maps in all acquisition schemes. 
This can be observed both visually in Figures~\ref{fig:2dvivo_spiral}, \ref{fig:2dvivo_radial}, \ref{fig:3dvivo}, S2 and S3,
and quantitatively in Table~\ref{tab:expe_retro}  across all tested metrics.
Other baselines were unable to successfully remove the under-sampling artefacts in TSMIs, and these errors propagated to the parameter inference phase and resulted in inaccurate maps. Temporal-only priors incorporated within LR are shown insufficient to regularise the inverse problem and LR sometimes (e.g. 2D spiral acquisitions) can admit solutions with even stronger artefacts than the model-free ZF baseline. This issue was previously studied for other non-Cartesian MRF readouts that similar to our spiral/radial trajectories, miss to sample the corners of the k-space in all timeframes (see section~2.2.2 and figure~2~in~\cite{AIRMRF}). In the absence of reference for the k-space corners information, the LR iterations despite minimising the objective $\|Y-\Aa(VX)\|^2$ can converge to solutions with high-frequency artefacts, as visible in the computed maps. This highlights the need for adding an appropriate spatial-domain regularisation. FLOR reduces the LR's artefacts but this improvement is limited because the suggested nuclear norm penalty does not incorporate an explicit spatial regularisation. Further for reducing artefacts, FLOR can introduce an undesirable bias in the computed T1/T2 maps e.g. see error maps in Figures S2 and S3. 
The non model-based VS baseline incorporates spatial regularisation and results in spatially smoother maps than ZF and LR, but it is unable to output artefact-free images. Further and consistent with our \emph{in-vitro} experiment, we observe that VS overestimates the T2 values (e.g. in White and Grey matter regions) in tested 2D acquisitions i.e. the spatial regularisation trades off agains the quantification accuracy. The model-based AIR-MRF adds spatial regularisation through 2D/3D low-pass Gaussian filters however this trades off the sharpness of the computed maps and can increase the errors at the tissue boundaries (we searched Gaussian spreads that keep the blurs and high-frequency artefacts minimal). For our acquisition readouts, Gaussian filters performed better than disk filters of~\cite{AIRMRF} for avoiding strong Gibbs artefacts. 
On the other hand, the spatiotemporally regularised LRTV greatly improves the TSMI reconstructions i.e. 4 dB enhancement compared to the closest competitor baseline (Table~\ref{tab:expe_retro}). This enables computing accurate and aliased-free multi-parametric inference using DM or  the DM-free learning-based alternative MRFResnet as visible in Figures~\ref{fig:2dvivo_spiral}, \ref{fig:2dvivo_radial}, \ref{fig:3dvivo}, S2 and S3. 
MRResnet and DM score competitive quantitative inference results i.e. T1 and T2 MAPE less than 5\% and 9\%, respectively (Table~\ref{tab:expe_retro}). KM also  outputs comparably accurate T1 maps, however this shallow learning model despite having a model size larger than MRFResnet, is unable to learn accurate T2/PD quantification and it results in poor estimated maps, consistent with our observations in section~\ref{sec:deepvskm}.

\subsection{MRFResnet's consistency with DM}
\begin{figure}[t!]
	\centering
	\scalebox{1}{
	\begin{minipage}{\linewidth}
		\centering
		\includegraphics[width=.3\linewidth]{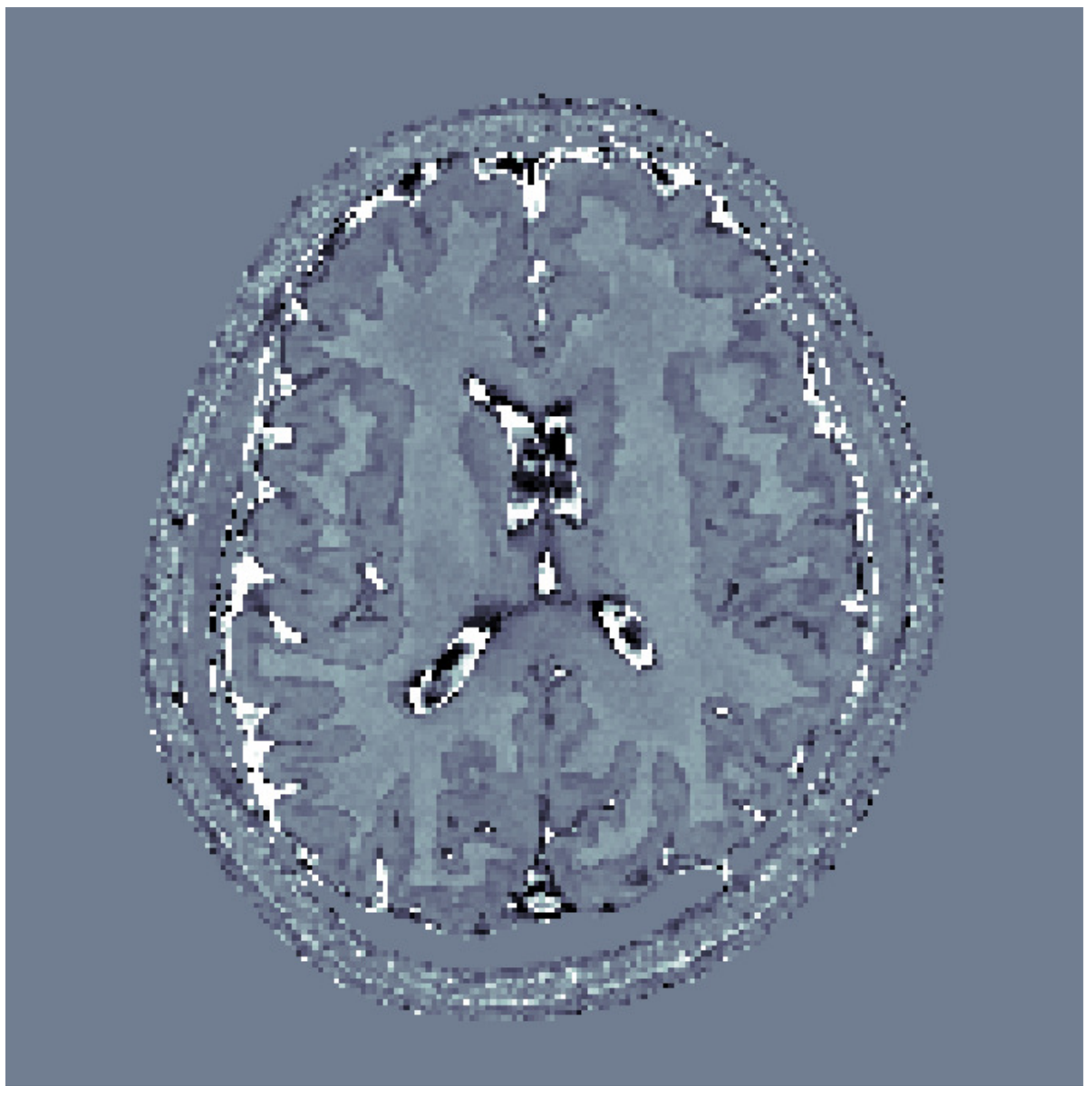}
		\includegraphics[width=.3\linewidth]{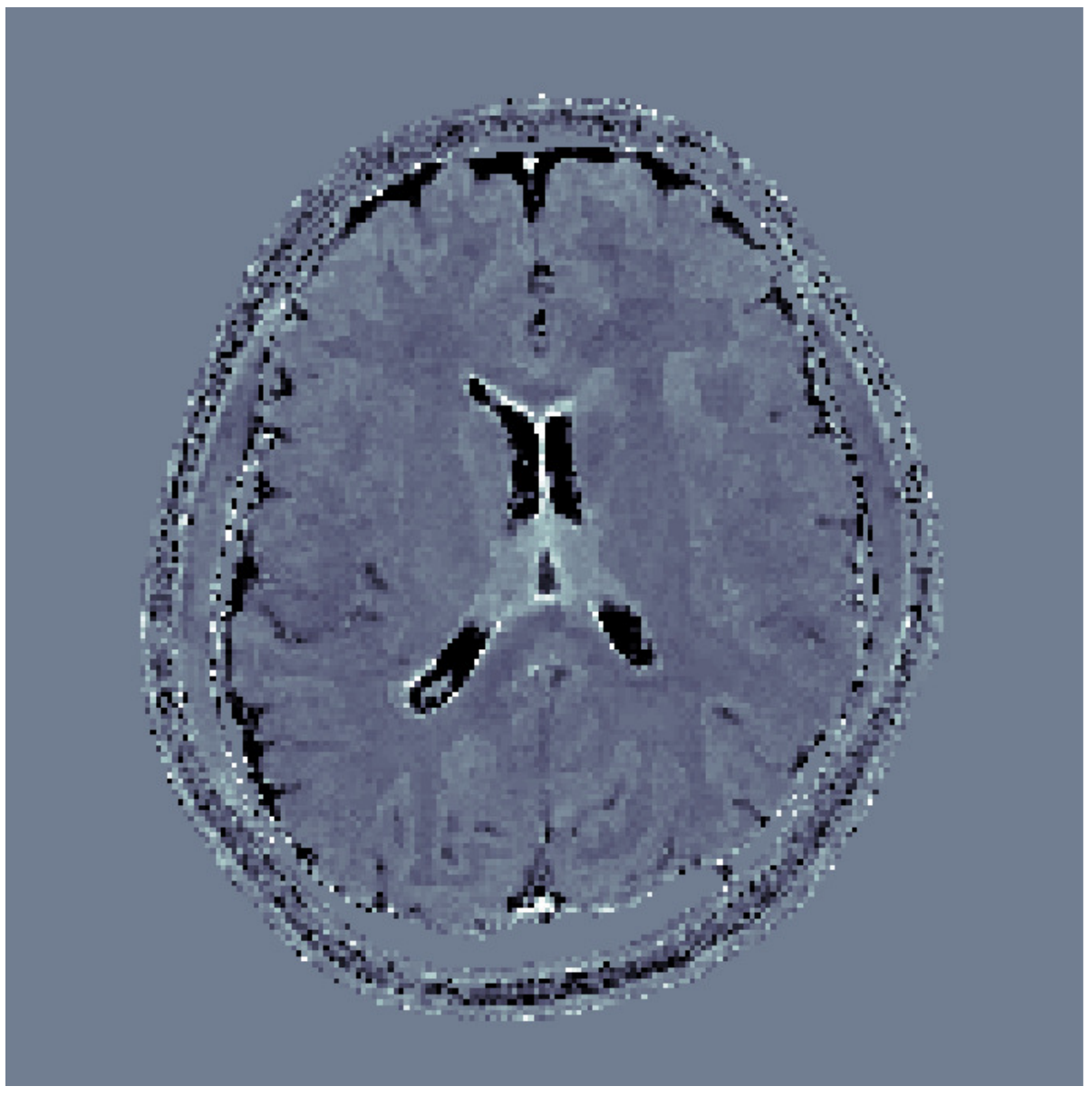}
		\includegraphics[width=.3\linewidth]{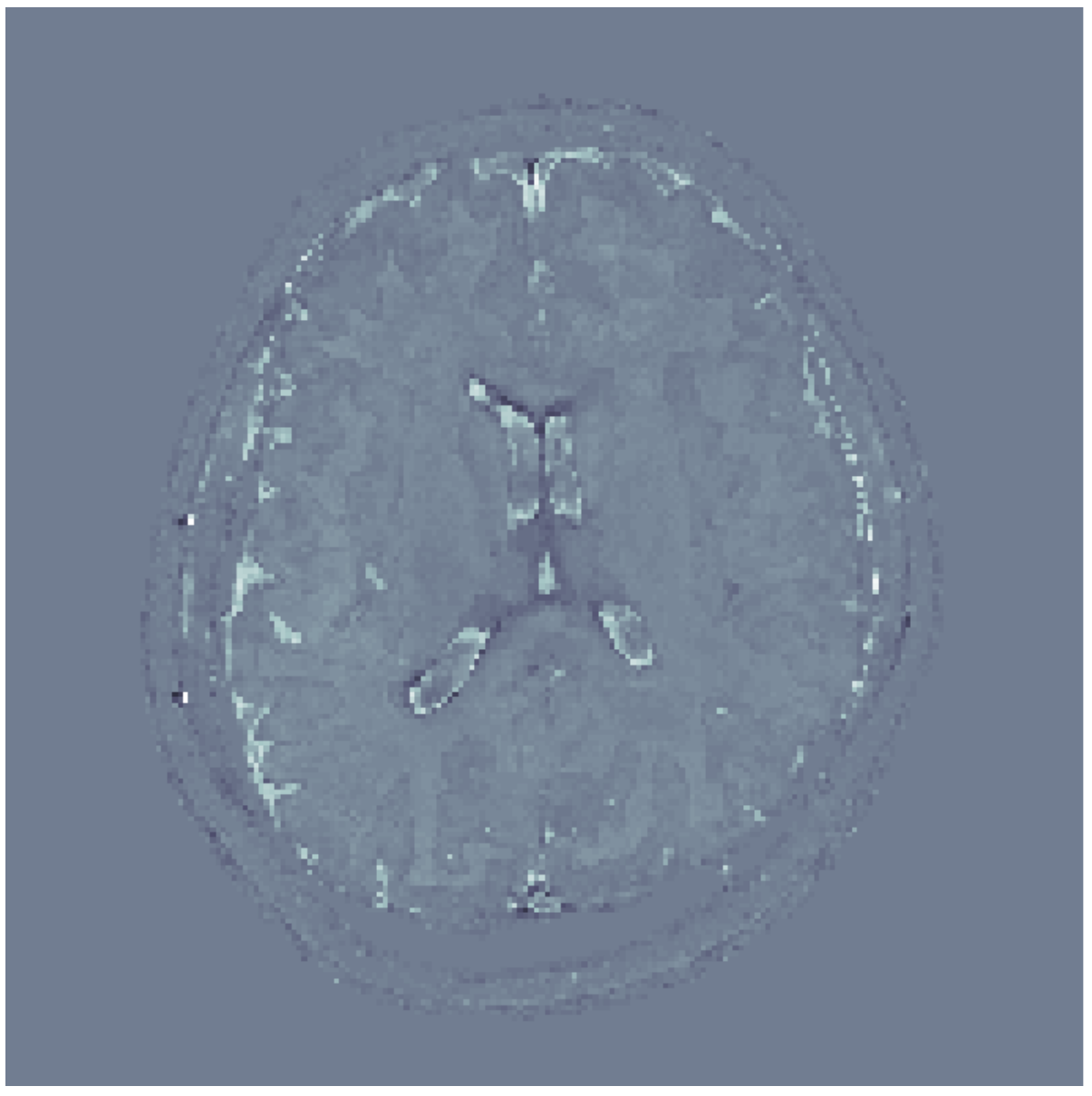}\vspace{-.3cm}
		\\
		\includegraphics[trim= -10 50 -20 720, clip,width=.32\linewidth]{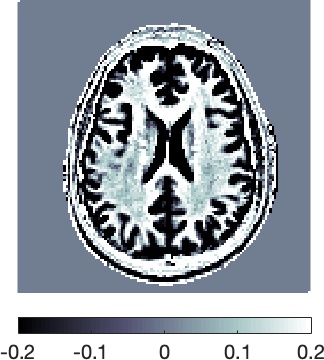}
		\includegraphics[trim= -10 50 -20 700, clip,width=.32\linewidth]{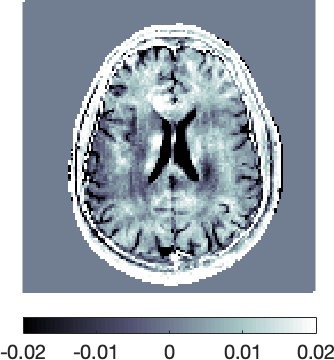}
		\includegraphics[trim= -10 50 -20 700, clip,width=.32\linewidth]{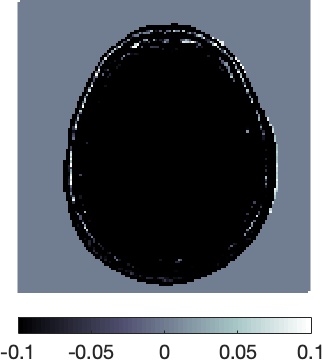}\vspace{.3cm}
				\\
		.\hspace{.5cm}diff T1(s) \hspace{1.2cm} diff T2 (s) \hspace{1.1cm}  diff PD (a.u.) \\
		\caption{Differences in the predicted T1, T2 and PD maps between MRFResnet and DM, after applying LRTV reconstruction.\vspace{.2cm}}  \label{fig:dmvsnet}
\end{minipage}}
\end{figure}

Further to section~\ref{sec:deepvskm} validations, 
we compare parametric maps computed by DM and MRFResnet for the \emph{in-vitro} and \emph{in-vivo} scans, where the LRTV algorithm was applied for TSMI reconstruction.
Results are summarised in Table~\ref{tab:dmvsnet} and 
for the \emph{in-vivo} 2D spiral scan is illustrated in Figure~\ref{fig:dmvsnet}. We observe very small differences in parametric maps (Table~\ref{tab:dmvsnet}) and particularly for the regions corresponding to white and grey matters predictions are highly consistent with each other (Figure~\ref{fig:dmvsnet}). 

\begin{table}[t!]
	\centering
	\scalebox{.93}{
		\begin{tabular}{c|cccc}
			\toprule[0.2em]
			
			NRMSE (\%) &  {T1} & T2 & PD \\		
			\midrule[0.1em]		
				\midrule[0.1em]		
			2D/3D phantom scans & 0.08 / 0.13  & 0.12 / 0.13  &0.78 / 1.43\\			
			2D/3D volunteer  scans & 3.25 / 1.28   & 7.15 / 2.68 &4.34 / 6.04\\
			\bottomrule[0.2em]
		\end{tabular}		}
		\caption{\footnotesize{The NRMSE between the T1, T2 and PD maps obtained from MRFResnet and DM, after LRTV reconstruction (real-world scans).}}\label{tab:dmvsnet}
	\end{table}

\begin{table*}[t!]
	\centering
	\fontsize{9}{9}\selectfont
	\scalebox{.8}{
		\begin{tabular}{c|cc|cc|cc|ccc||cc|cc|cc|ccc}
			\toprule[0.2em]
			\multicolumn{1}{c}{} &\multicolumn{9}{c}{2D spiral acquisition} & \multicolumn{9}{c}{2D radial acquisition} \\
			\midrule[0.1em]
			\multicolumn{1}{c}{} &\multicolumn{2}{c}{MAPE (\%)} & \multicolumn{2}{c}{MAE (ms)} & \multicolumn{2}{c}{SNR (dB)} & \multicolumn{3}{c}{SSIM}  &  \multicolumn{2}{c}{MAPE (\%)} & \multicolumn{2}{c}{MAE (ms)} & \multicolumn{2}{c}{SNR (dB)} &\multicolumn{3}{c}{SSIM}  \\
			\midrule[0.1em]
			{} &  {T1 } & { T2 } &{T1 } & { T2 } &  PD& TSMI & {T1 } & { T2 } &  PD &  
			{T1 } & { T2 } &{T1 } & { T2 } &  PD& TSMI & {T1 } & { T2 } &  PD \\
			\midrule[0.05em]
			\midrule[0.05em]
			ZF-DM  & 6.58   & 14.87 & 64.2 & 14.7 &  14.45  & 14.62  & 95.59&95.56 & 88.34&
			9.90 &  27.52 & 97.6 & 28.1 &  14.84 &   11.65  & 90.34& 87.43 & 89.01 \\  
			\midrule[0.1em]
			VS-DM & 9.19   & 23.74 & 99.2 & 23.8 & 17.02 & 19.16 & 94.40& 93.58& 90.08& 
			9.32 & 23.63  & 99.1 &23.8 &  16.52  & 18.28 & 93.40 & 93.09 & 89.87\\	
			\midrule[0.1em]
			LR-DM  & 9.96 & 19.39 & 98.6& 19.5 & 17.72 &  13.26 & 90.15& 93.13& 90.76& 
			7.49 & 13.46  &  82.2 & 14.1 & 19.97 & 15.83 & 92.96 & 96.23 & 95.90\\	
			\midrule[0.1em]
			AIRMRF & 10.64  & 15.38  & 126.1& 17.5 & 15.08   &  10.98   & 90.43&94.59 &94.99 & 
			 10.87& 16.28 &128.7 & 18.5&  14.93 & 10.72 & 90.12 & 93.96 & 94.72 \\
			 \midrule[0.1em]
			 FLOR & 7.69 & 17.97   &86.0 & 16.8 &  18.32   & 21.61   & 94.98& 96.80 & 97.75 & 						
			9.85 & 24.80 & 115.3 & 22.7 &  16.26 & 19.91 & 93.72 & 95.31 & 96.48\\
			\midrule[0.1em]
			\textbf{LRTV-DM}  & \textbf{4.06}  & \textbf{7.52} & \textbf{42.3} & \textbf{8.3} &  \textbf{26.59} &  \textbf{25.36}  & \textbf{97.21} & \textbf{98.61}& \textbf{98.93}& 	
			\textbf{4.90} &  \textbf{8.73} & \textbf{52.8} & \textbf{10.0} &  \textbf{25.15} & \textbf{23.79}  &\textbf{96.32} & \textbf{97.91} & \textbf{98.64} \\	
			*-KM  & 4.52   &16.81 &  46.5 & 15.6  & 15.55 &  --''-- & 96.34& 95.40 & 96.53 &
			  5.15 & 18.32 &55.0 & 17.5 & 15.21 &  --''-- & 97.79& 94.75 & 96.33\\	
			  \textbf{*-MRFResnet} & \textbf{4.11} & \textbf{7.94} & \textbf{42.9} & \textbf{8.6}  & \textbf{ 26.23}  &  --''--&  \textbf{97.27} & \textbf{98.53}& \textbf{98.87} &
			  \textbf{4.96} & \textbf{9.13} & \textbf{53.5} & \textbf{10.2} & \textbf{24.82} & --''-- & \textbf{96.42} & \textbf{97.83} & \textbf{98.59}\\
			\bottomrule[0.2em]
		\end{tabular}}
		\caption{The T1, T2, PD and TMSI reconstruction accuracies (measured by the MAE and MAPE errors, reconstruction SNR, and SSIM metrics) of the baselines and our proposed LRTV-MRFResnet algorithm, validated retrospectively against ground-truth anatomical maps.} \vspace{.3cm}\label{tab:expe_retro}
	\end{table*}

\subsection{Runtimes}
Computations were conducted  on an Intel Xeon E5-2667v4 processor (16 CPU cores), 32 GB RAM and a NVIDIA 2080Ti GPU. Where parallel computing was feasible, we adopted GPU implementation for speedup i.e. in forward/adjoint NUFFT operations\footnote{\url{https://www.opensourceimaging.org/project/gpunufft/}}, the TV shrinkage operator\footnote{\url{https://epfl-lts2.github.io/unlocbox-html/}}, VS, MRFResnet and DM. Table~\ref{tab:runtime} includes computation times of the tested methods for the 2D/3D \emph{in-vivo} experiments. 
LRTV benefits from momentum-acceleration and takes 7-11 iterations to converge, and its runtime faster than tested iterative schemes i.e.  about an order of magnitude faster than the DM-based iterative AIR-MRF with fast kd-tree searches. 
FLOR has the slowest runtime due to not using the subspace dimensionality-reduction, which makes it also memory-wise non-scalable for our 3D reconstruction experiment.  
We also observed that the LR method without spatial regularisation makes very slow progress towards its (inaccurate) solution and does not converge within our limit of 30 iterations. This indicates that exploiting additional (spatial) solution structure, despite introducing TV shrinkage computations, has an overall runtime advantage~(see e.g.\cite{Chandra-Jordan,GPIS}) by avoiding extra costly forward/adjoint iterations. The LRTV runs 2-3 times slower than its non-iterative competitor VS for achieving better predictions. DM-based inference methods are order(s) of magnitudes slower than MRFResnet, and therefore the great prediction consistency in both approaches suggests adopting neural inference in favour of runtime. 
\begin{table}[h!]
	\centering
	\scalebox{.8}{
		\begin{tabular}{cccccc|cc}
			\toprule[0.2em]
			\multicolumn{6}{c}{reconstruction times (s)} & \multicolumn{2}{c}{inference times (s)} \\
			\midrule[0.05em]
			&  ZF & VS & LRTV& AIRMRF&FLOR & DM & MRFResnet  \\		
			\midrule[0.1em]	
			\midrule[0.1em]	
			2D & 2.0 & 1.2e1  & 2.9e1 & 1.1e2  &6.2e4 & 8.5 & $< 0.5$ \\		
			3D & 1.5e2    & 9.1e2  & 2.2e3 & 4.7e4 &   \textemdash&1.6e3 & 54\\
			\bottomrule[0.2em]
		\end{tabular}		}
		\caption{\footnotesize{Tested runtimes for quantitative brain image computing (8-coil scans).\vspace{.2cm}}}\label{tab:runtime}
	\end{table}

\section{Conclusions}
\label{sec:conclusion}
We proposed a two-stage DM-free approach for multi-parametric QMRI image computing based on compressed sensing reconstruction and deep learning. The reconstruction is convex and incorporates efficient spatiotemporal regularisations within an accelerated iterative shrinkage algorithm to minimise undersampling artefacts in the computed TSMI. 
We proposed MRFResnet, a compact auto-encoder network with deep residual blocks, in order to embed Bloch manifold projections through multi-scale piecewise affine approximations, and to replace the non-scalable DM baseline for quantitative inference. 
We demonstrated the effectiveness of the proposed scheme through validations on a novel 2D/3D multi-parametric quantitative acquisition sequence.   
Future extensions  could address motion-artefacts and multi-compartment voxel quantification~\cite{MRFmotion_cruz, duarte_PVEMRF} that are currently un-modelled in our pipeline.  
Further accelerations could be studied through stochastic gradients~\cite{tang2019practicality} and/or learned proximity operations~\cite{restkatyusha} where the proposed scheme could complementarily be adopted for creating accurate labelled parametric maps for training.

\ifCLASSOPTIONcaptionsoff
  \newpage
\fi

\bibliographystyle{IEEEtran}
{\footnotesize{
\bibliography{mybiblio}
}}

\pagebreak
\newpage
\onecolumn
\begin{center}
	\textbf{\large Supplementary Materials: Compressive MRI quantification using convex spatiotemporal priors and deep auto-encoders }\\
	Mohammad Golbabaee, Guido Bounincontri, Carolin Pirkl, Marion Menzel, Bjoern  Menze, Mike Davies and Pedro G\`omez
\end{center}
\setcounter{equation}{0}
\setcounter{figure}{0}
\setcounter{table}{0}
\setcounter{section}{0}
\setcounter{page}{1}
\makeatletter
\renewcommand{\theequation}{S\arabic{equation}}
\renewcommand{\thefigure}{S\arabic{figure}}
\renewcommand{\thesection}{S\Roman{section}}
\renewcommand{\bibnumfmt}[1]{[S#1]}
\renewcommand{\citenumfont}[1]{S#1}
\section{Reconstructed parameter maps for the retrospective experiment}

\begin{figure*}[ht!]
	\centering
	\scalebox{1}{
	\begin{minipage}{\linewidth}
	\centering
		\includegraphics[width=.162\linewidth]{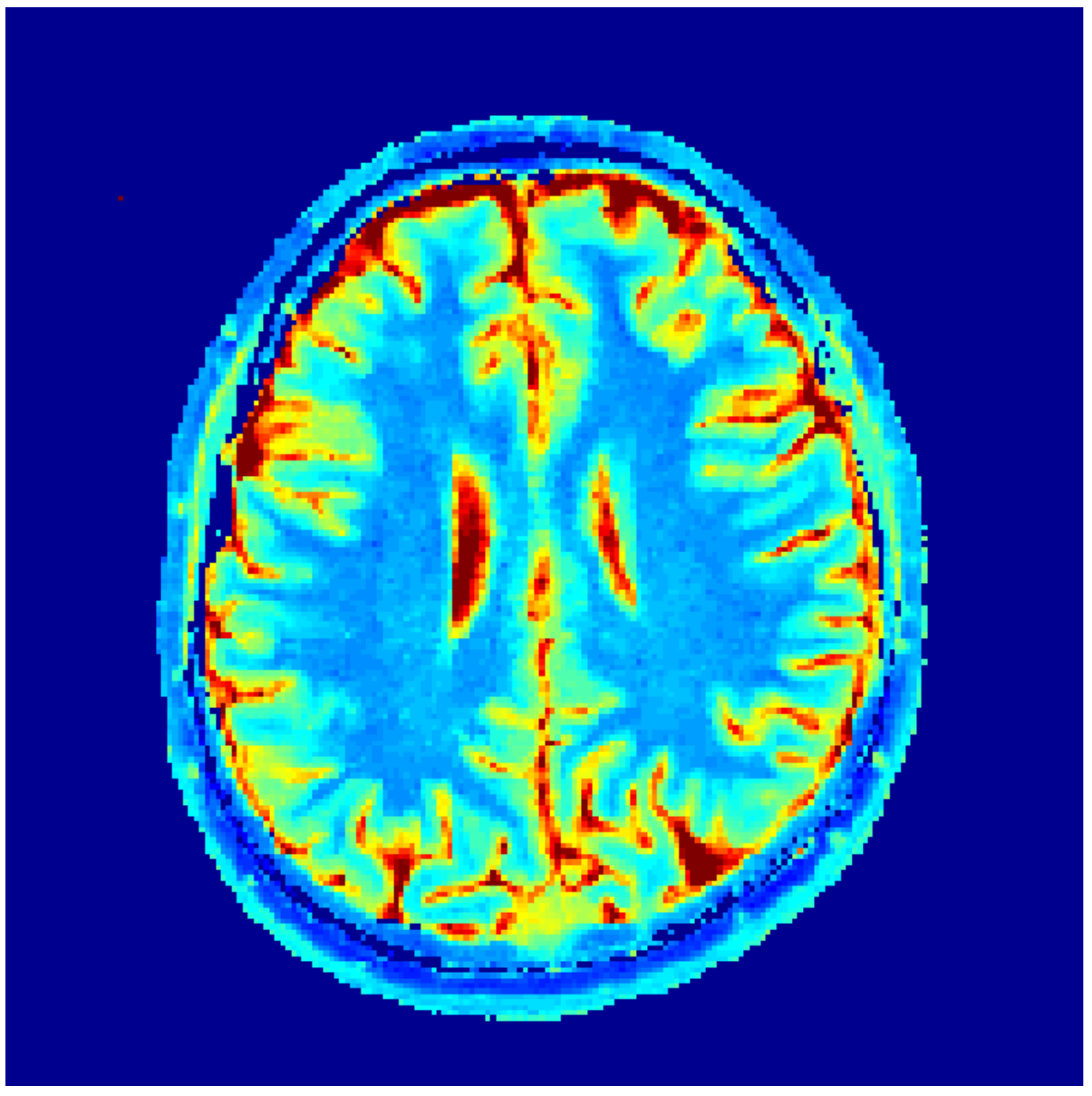}\hspace{-.1cm}
		\includegraphics[width=.162\linewidth]{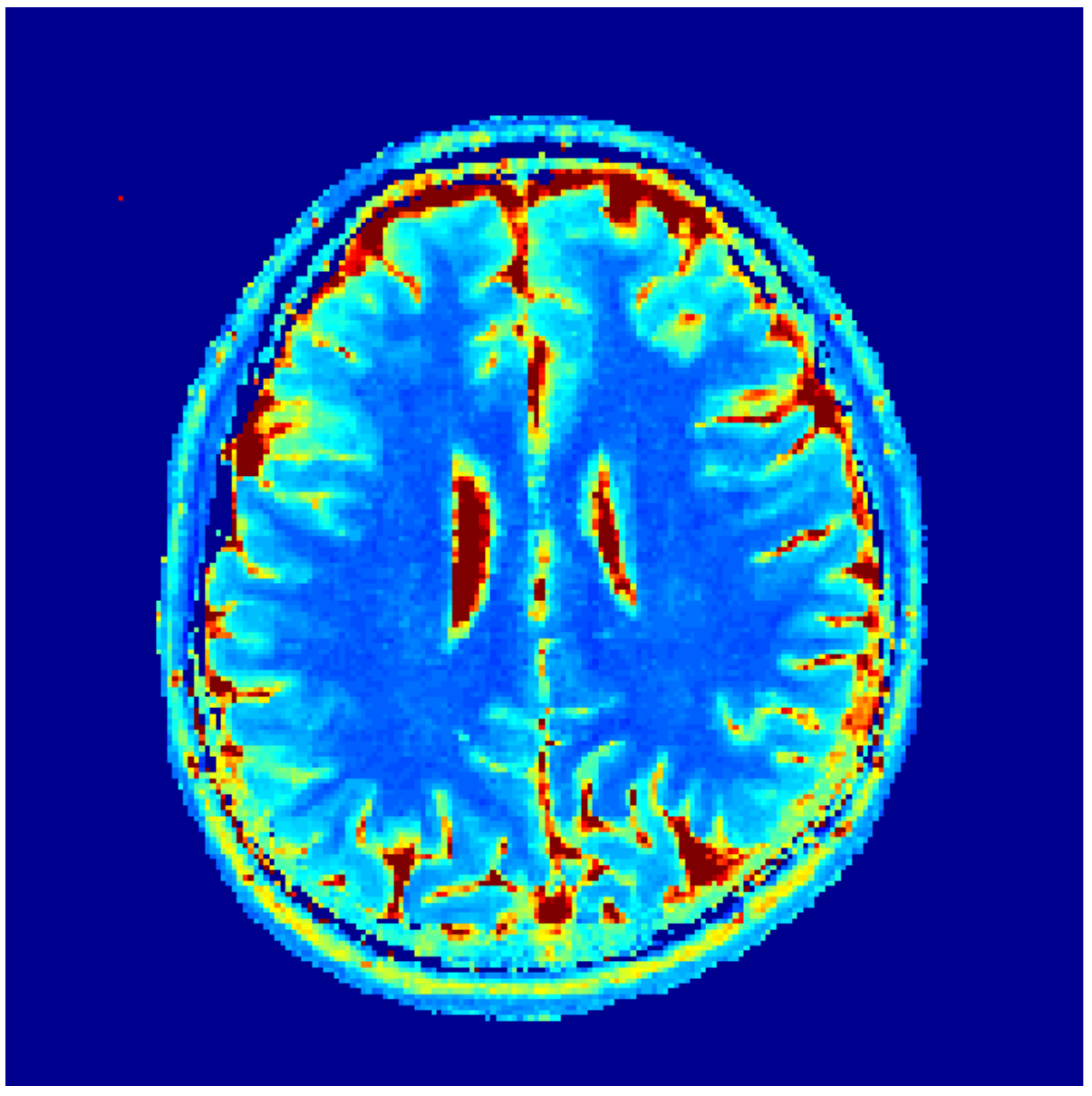}\hspace{-.1cm}
		\includegraphics[width=.162\linewidth]{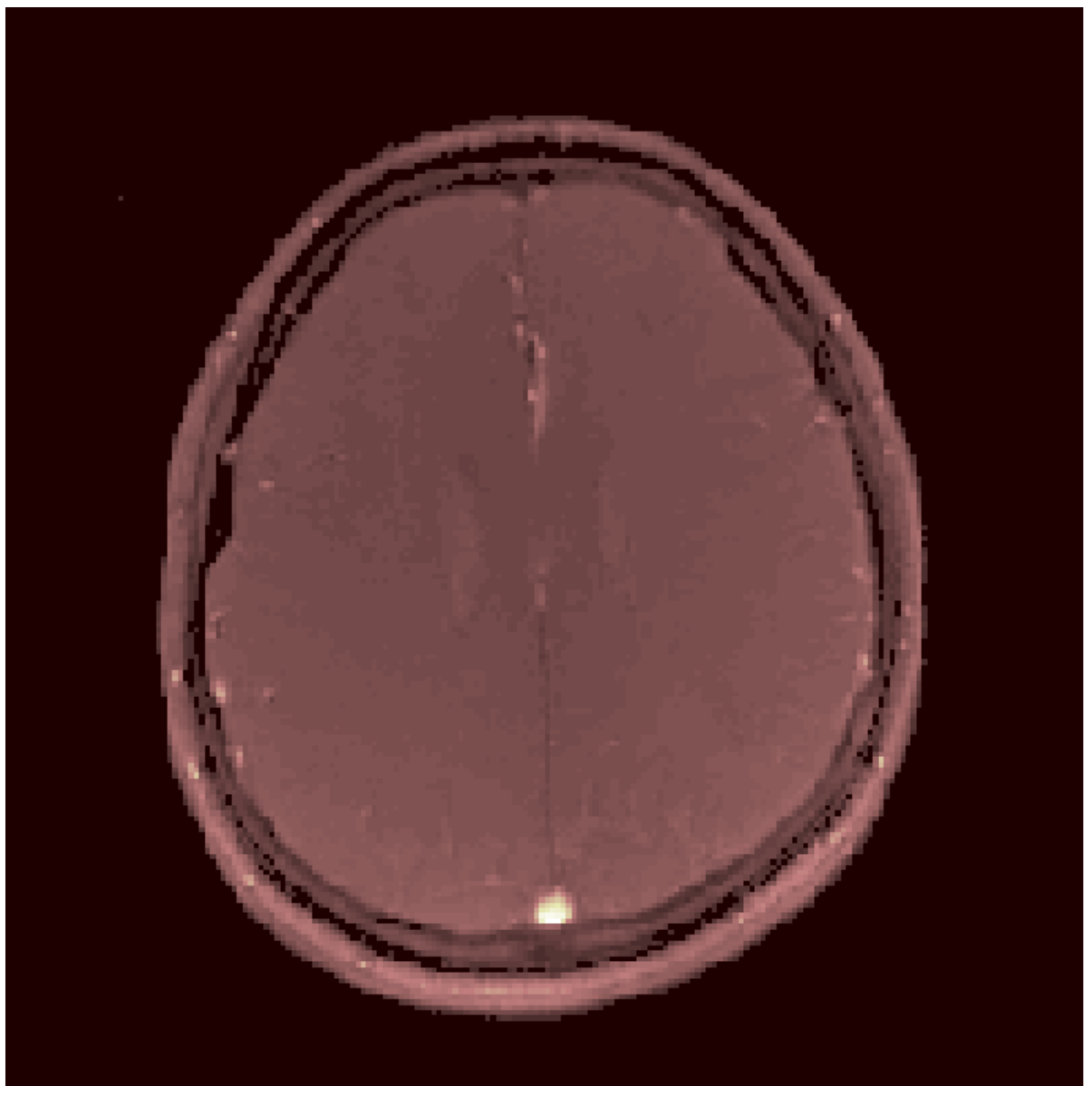}\vspace{-.3cm}
		\\
		\includegraphics[trim= -10 50 -20 720, clip,width=.17\linewidth]{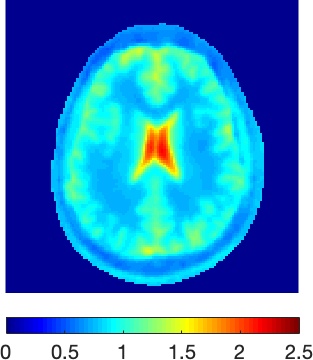}\hspace{-.2cm}
		\includegraphics[trim= -10 50 -20 700, clip,width=.17\linewidth]{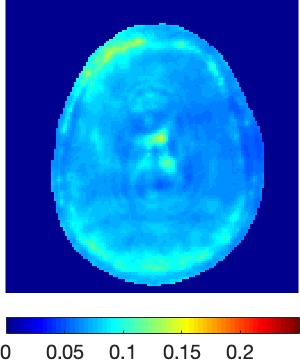}\hspace{-.2cm}
		\includegraphics[trim= -10 50 -20 700, clip,width=.17\linewidth]{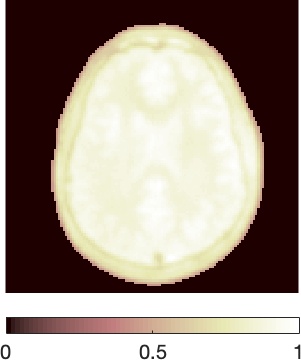}\vspace{.5cm}
				\\
		 .\hspace{2cm} T1(s)  \hspace{2cm} T2 (s) \hspace{2cm} PD (a.u.) \hspace{1.7cm} 
		\caption{The T1, T2 and PD maps acquired by the gold standard MAGIC quantitative acquisition protocol.  
		\label{fig:FISPGT}}
\end{minipage}}
\end{figure*}

\begin{figure*}[ht!]
	\centering
	\scalebox{.9}{
	\begin{minipage}{\linewidth}
	\begin{turn}{90} \quad\qquad ZF-DM \end{turn}
		\includegraphics[width=.162\linewidth]{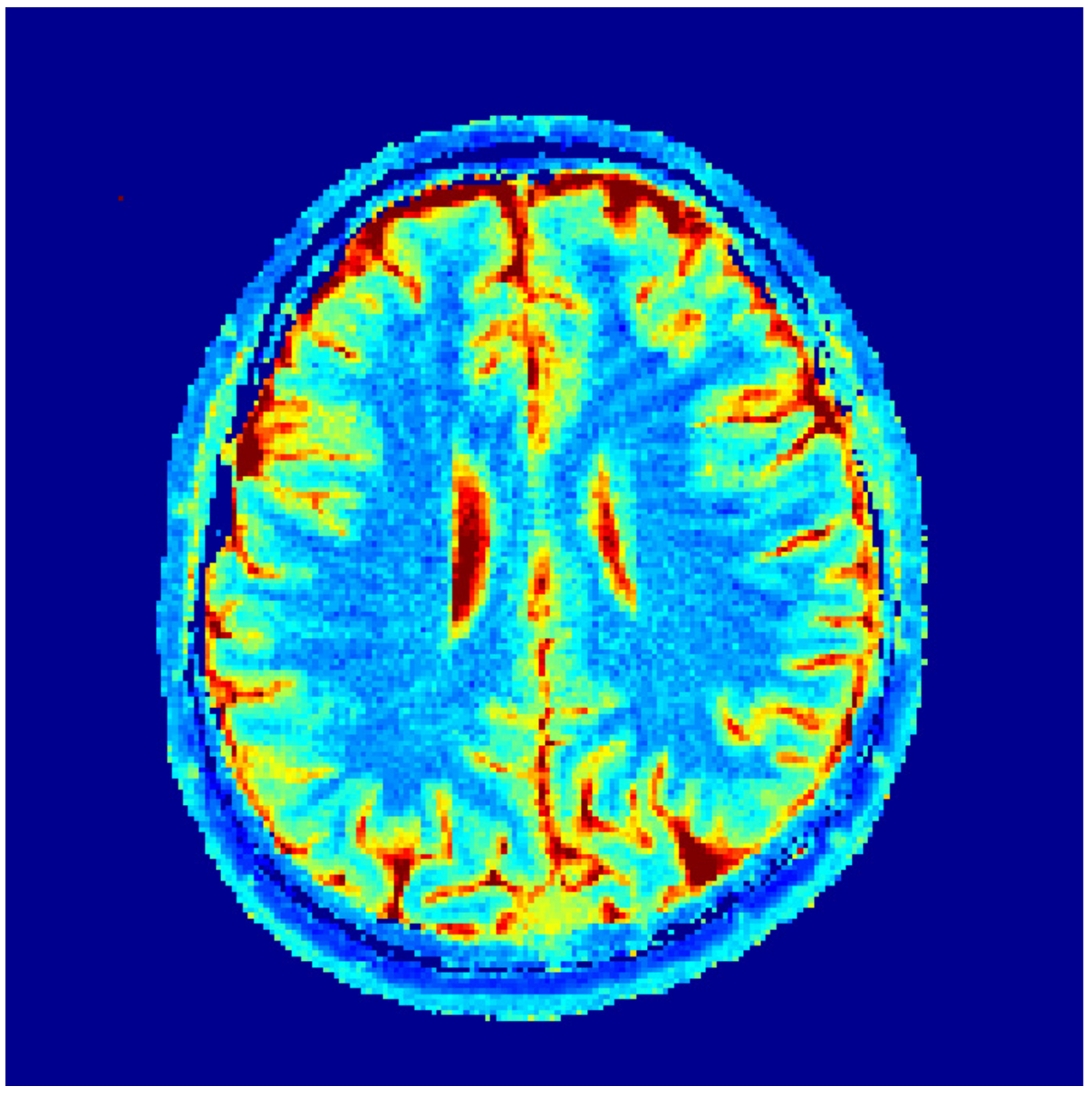}\hspace{-.1cm}
		\includegraphics[width=.162\linewidth]{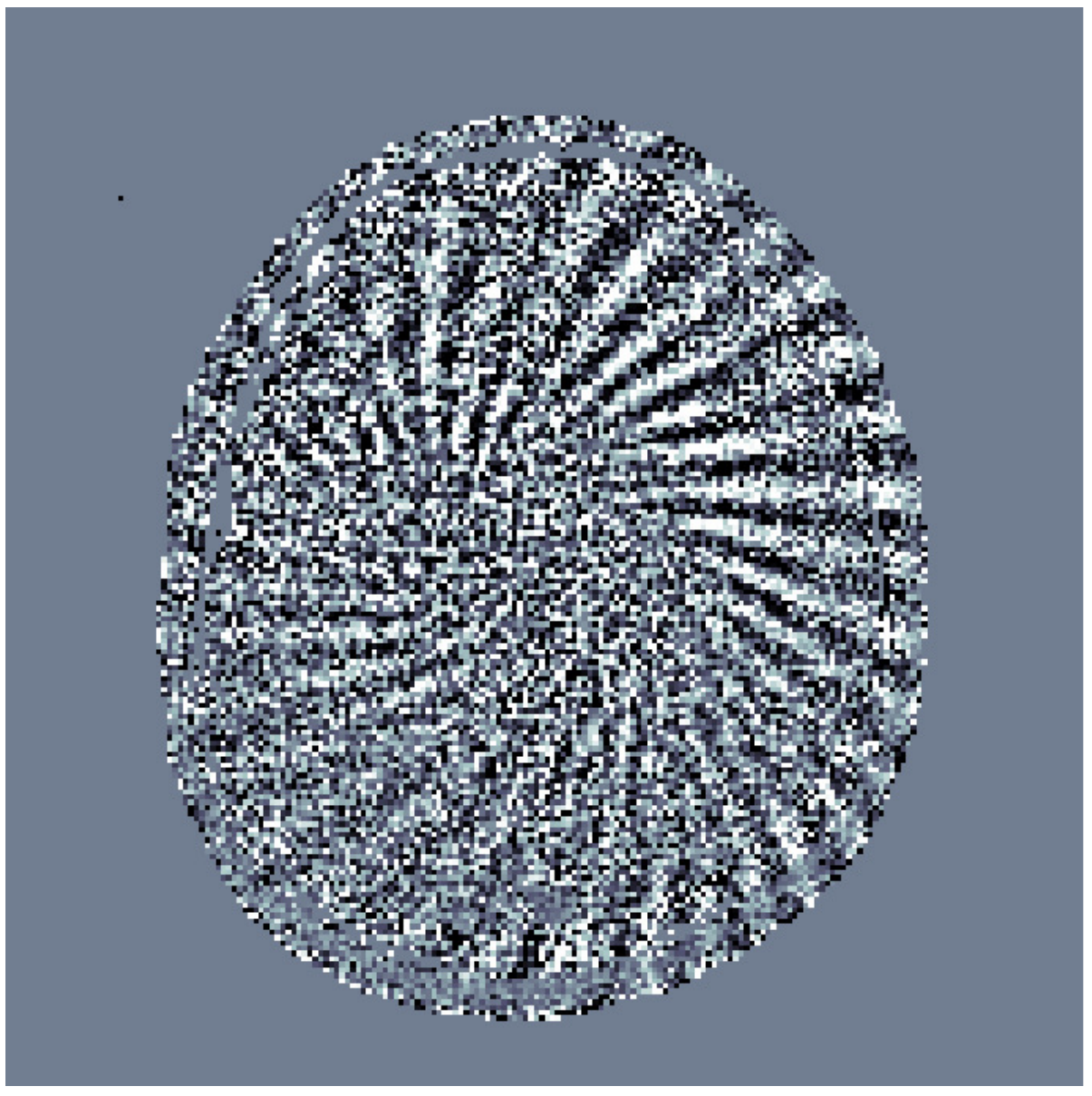}\hspace{.0cm}
		\includegraphics[width=.162\linewidth]{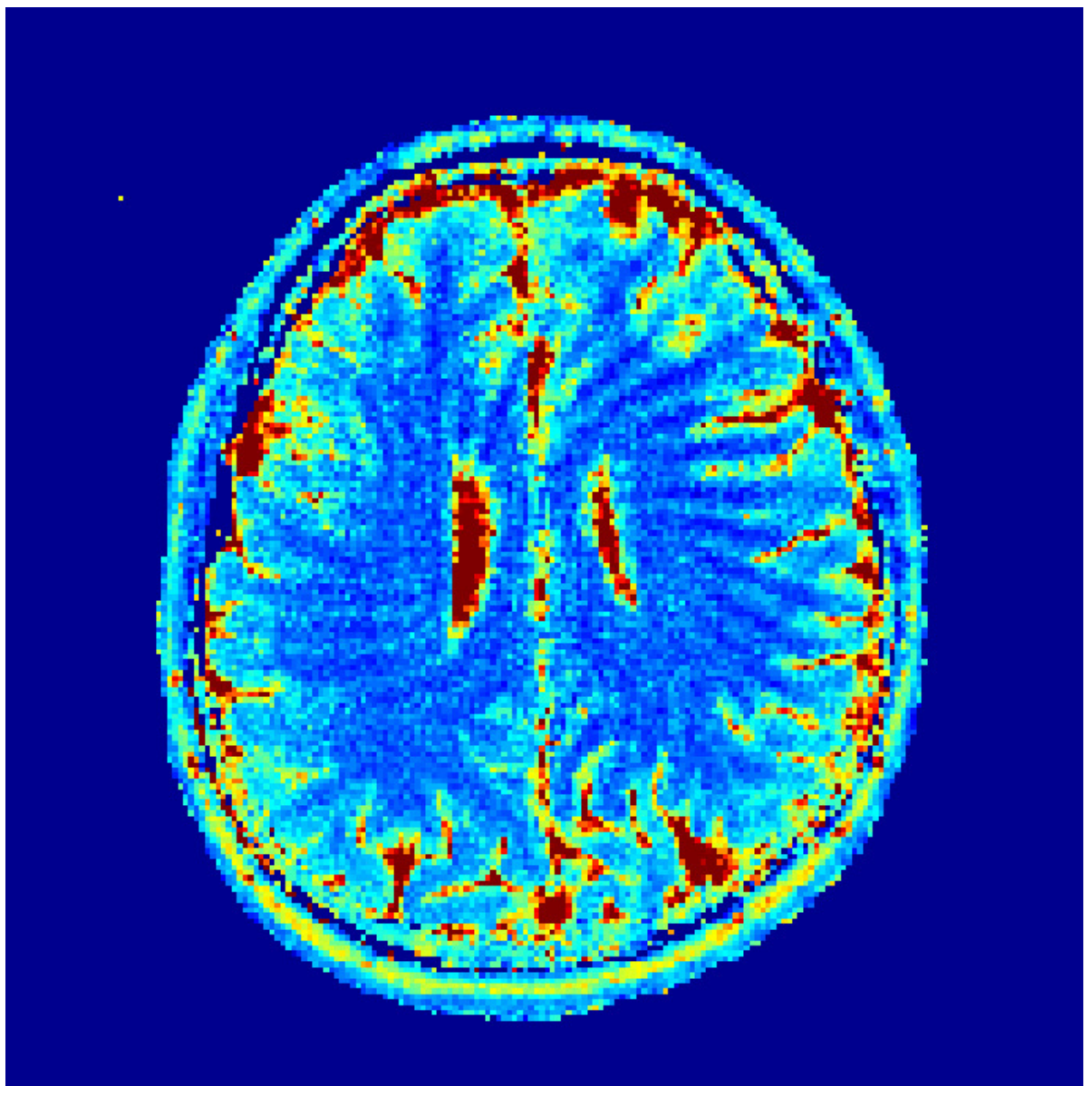}\hspace{-.1cm}
		\includegraphics[width=.162\linewidth]{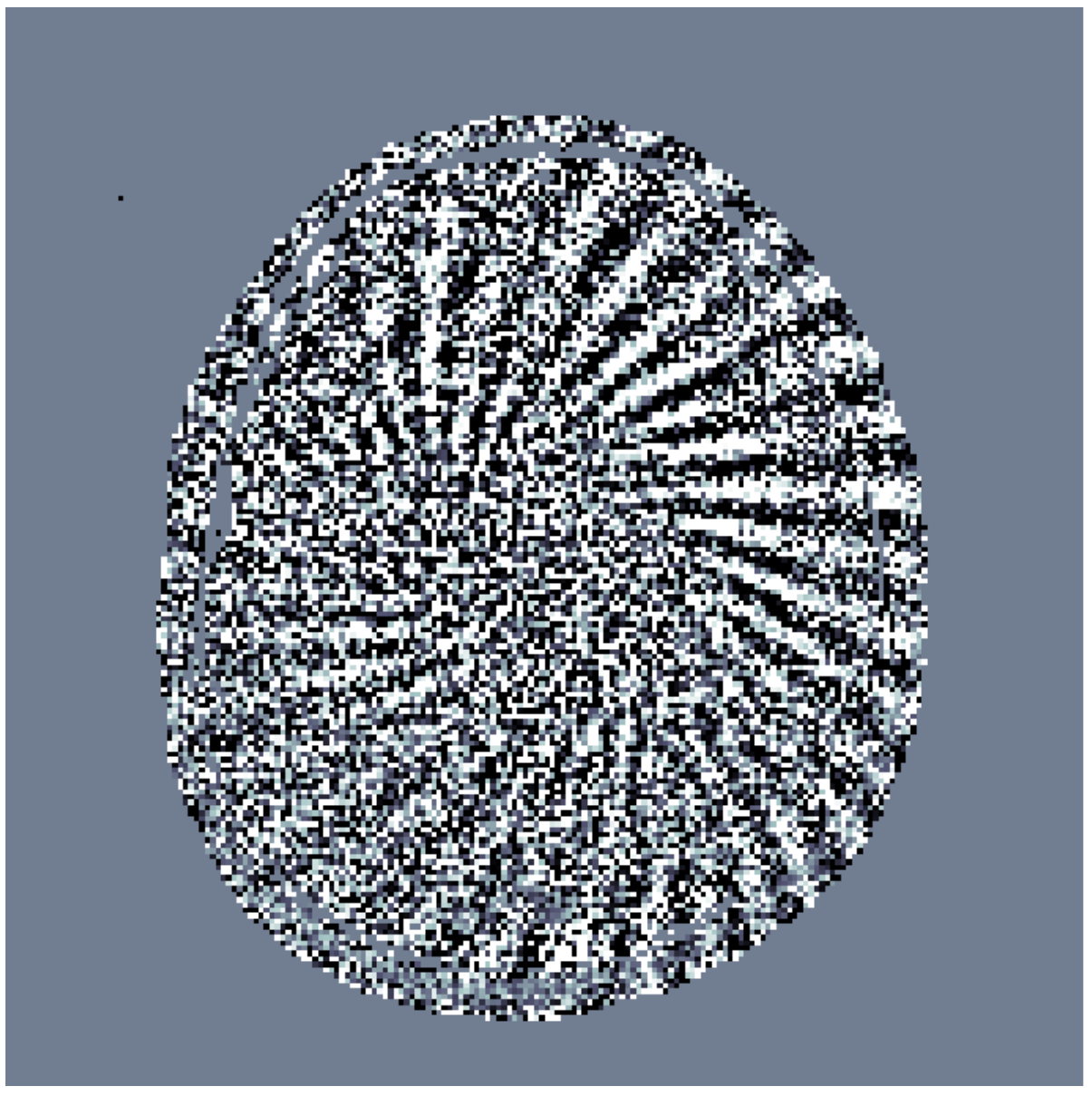}\hspace{.0cm}
		\includegraphics[width=.162\linewidth]{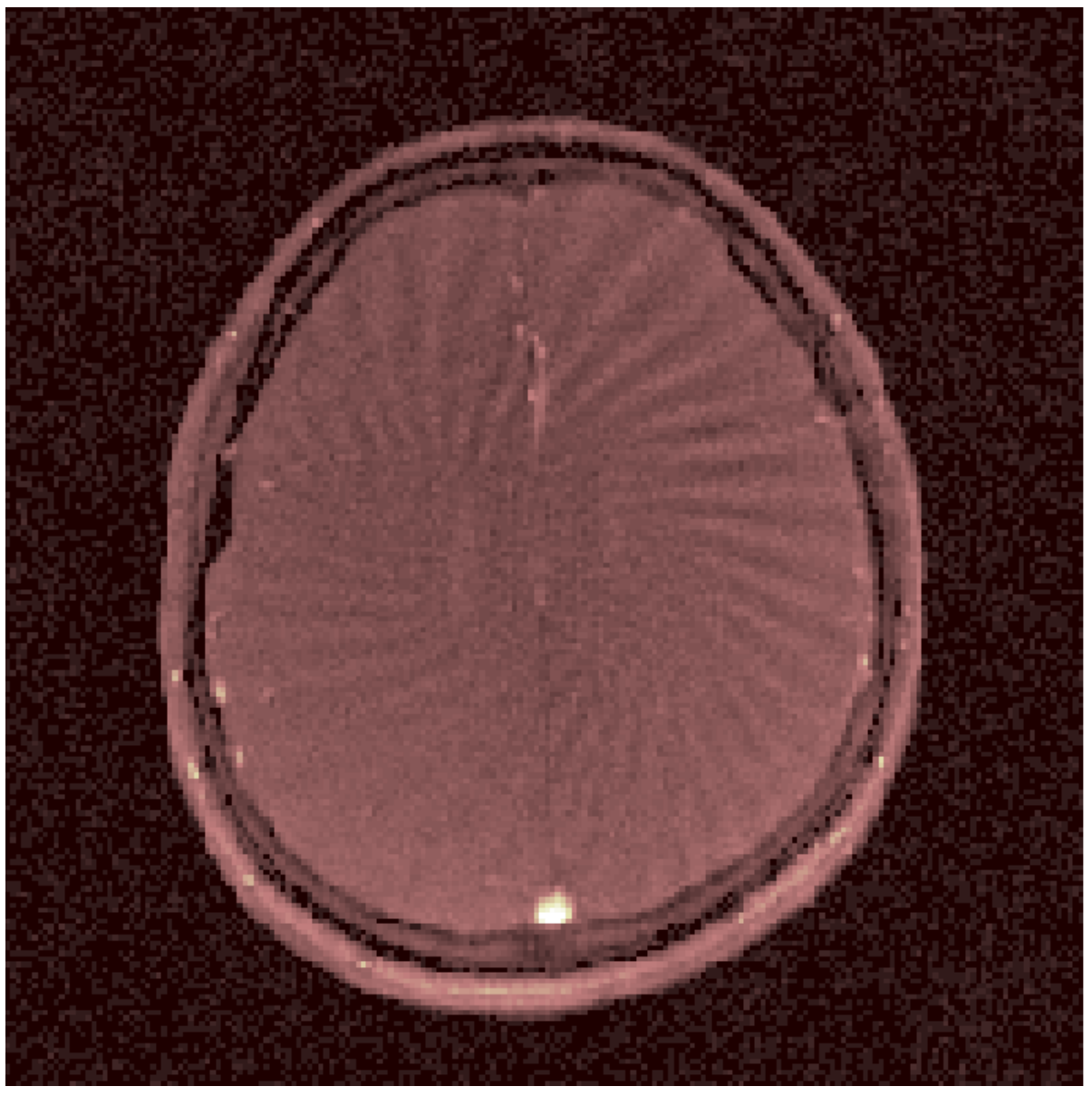}\hspace{-.1cm}
		\includegraphics[width=.162\linewidth]{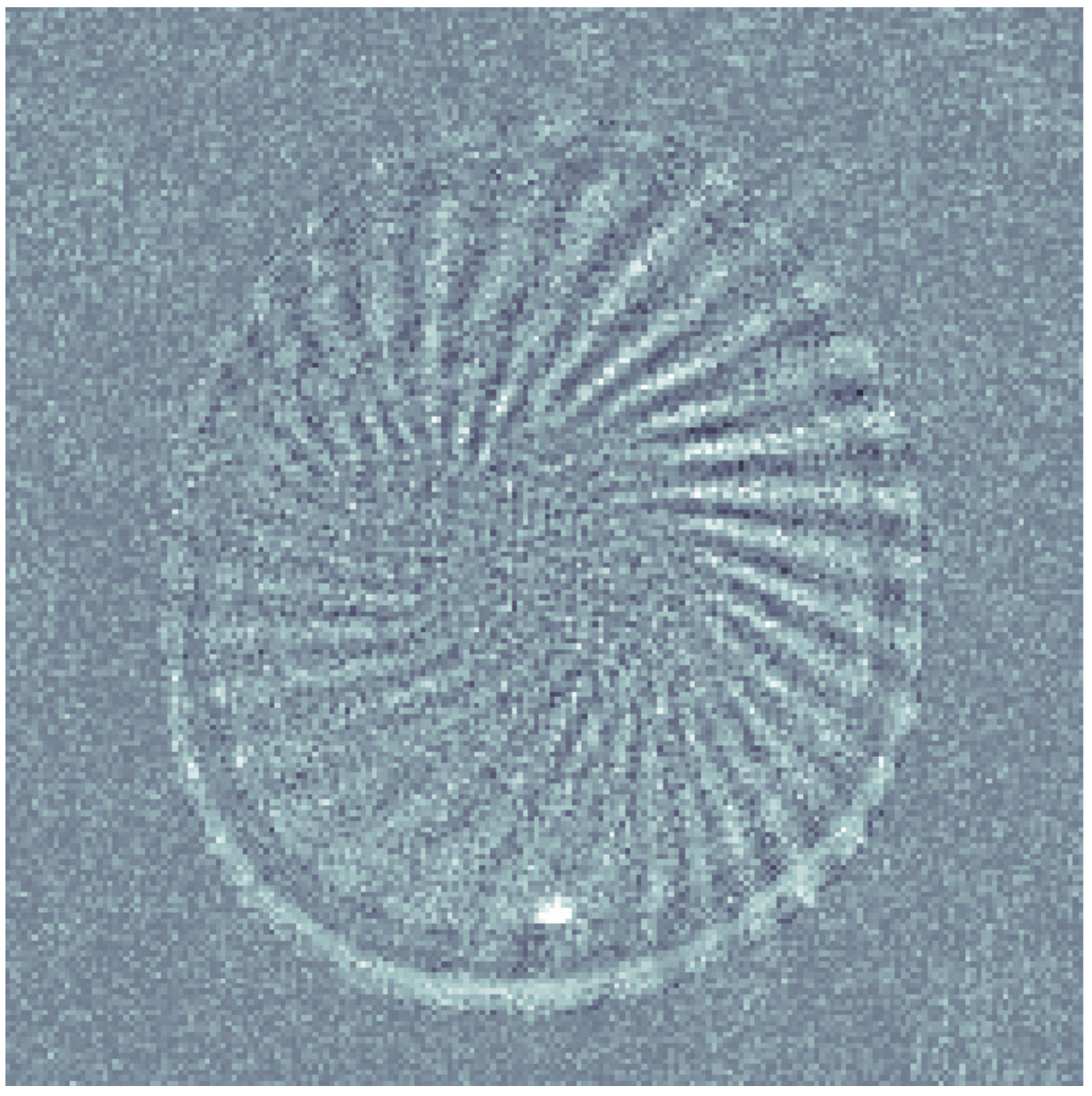}	
		\\
		\begin{turn}{90} \quad\qquad LR-DM \end{turn}
		\includegraphics[width=.162\linewidth]{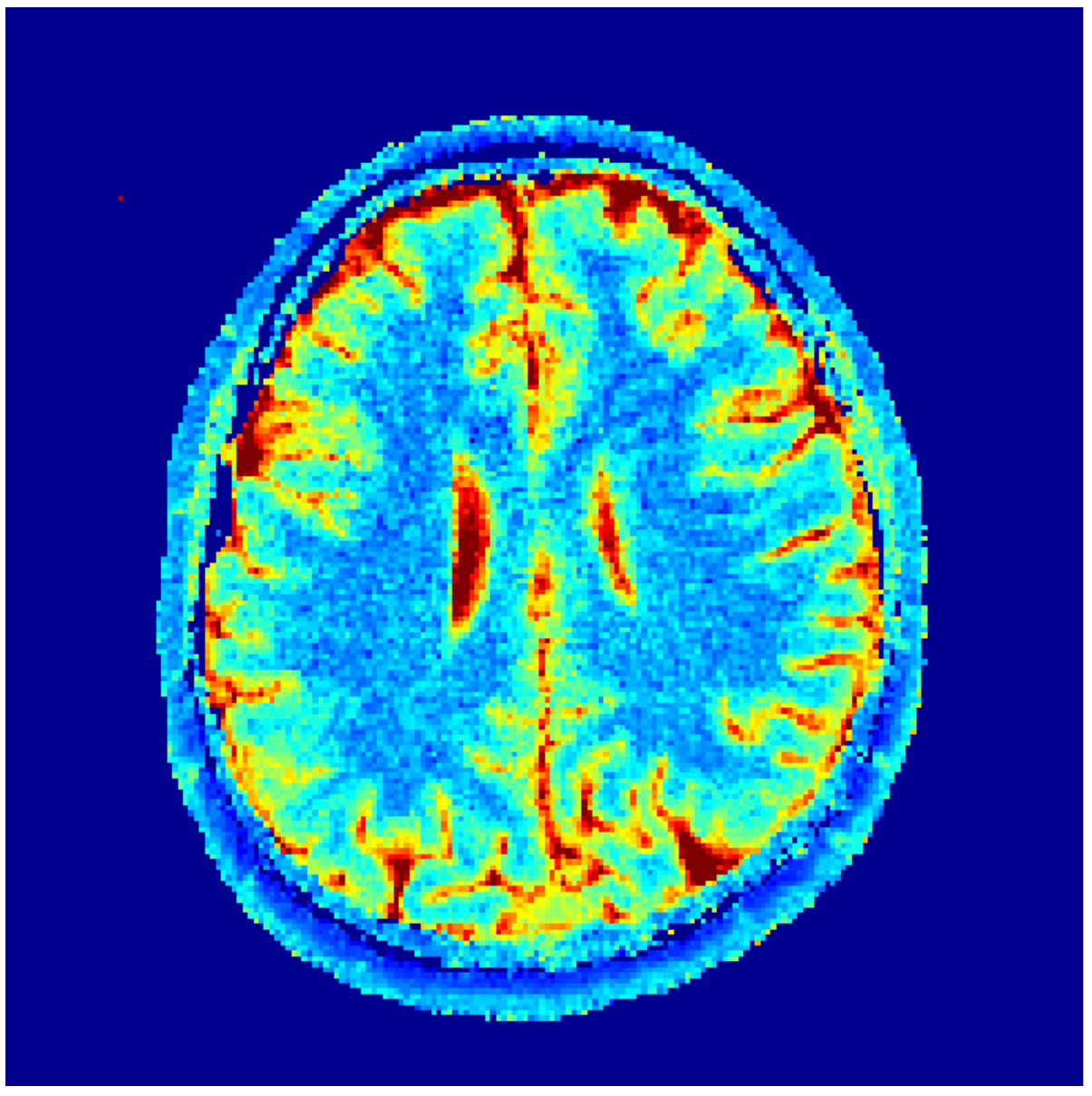}\hspace{-.1cm}
		\includegraphics[width=.162\linewidth]{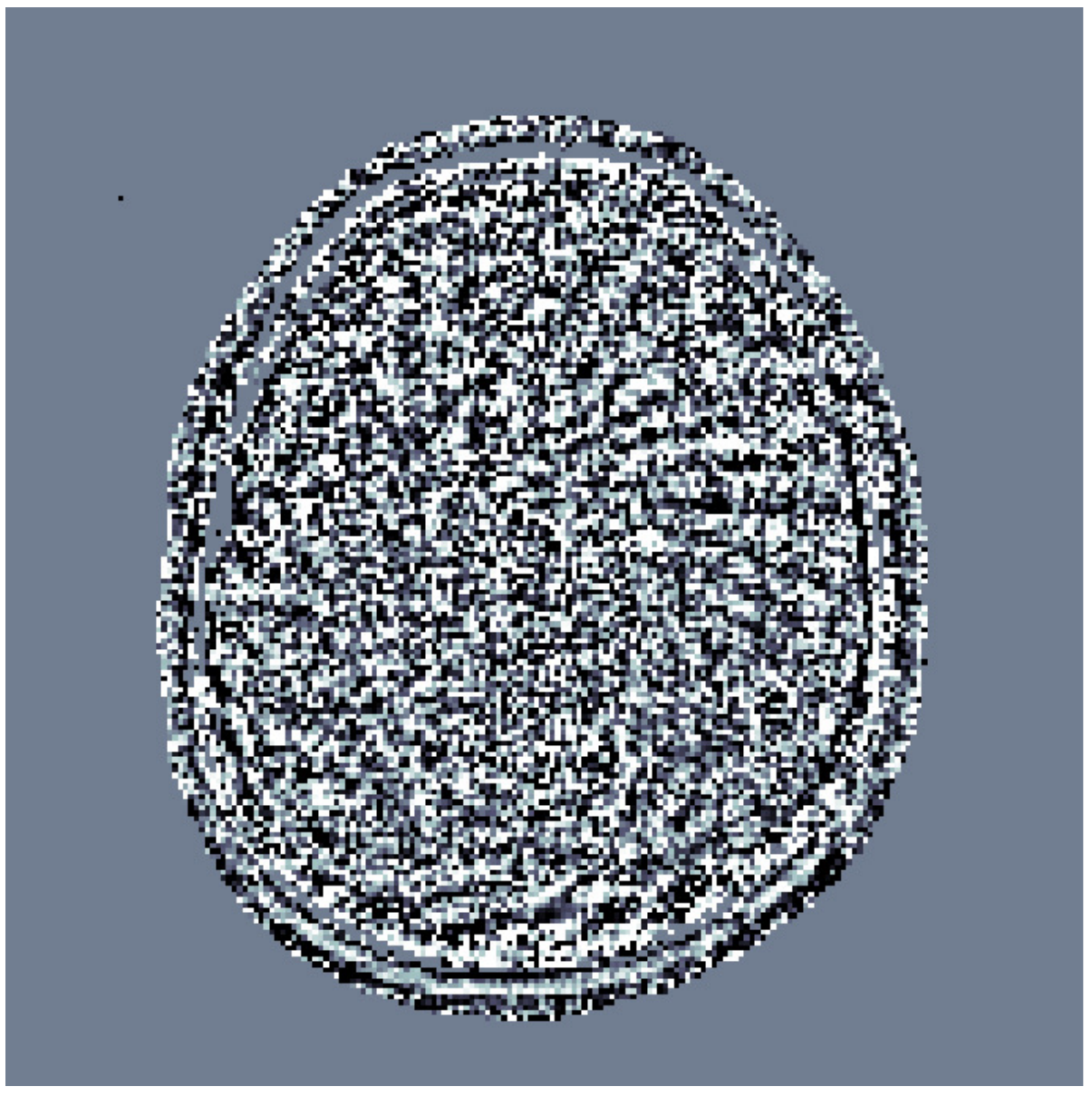}\hspace{.0cm}
		\includegraphics[width=.162\linewidth]{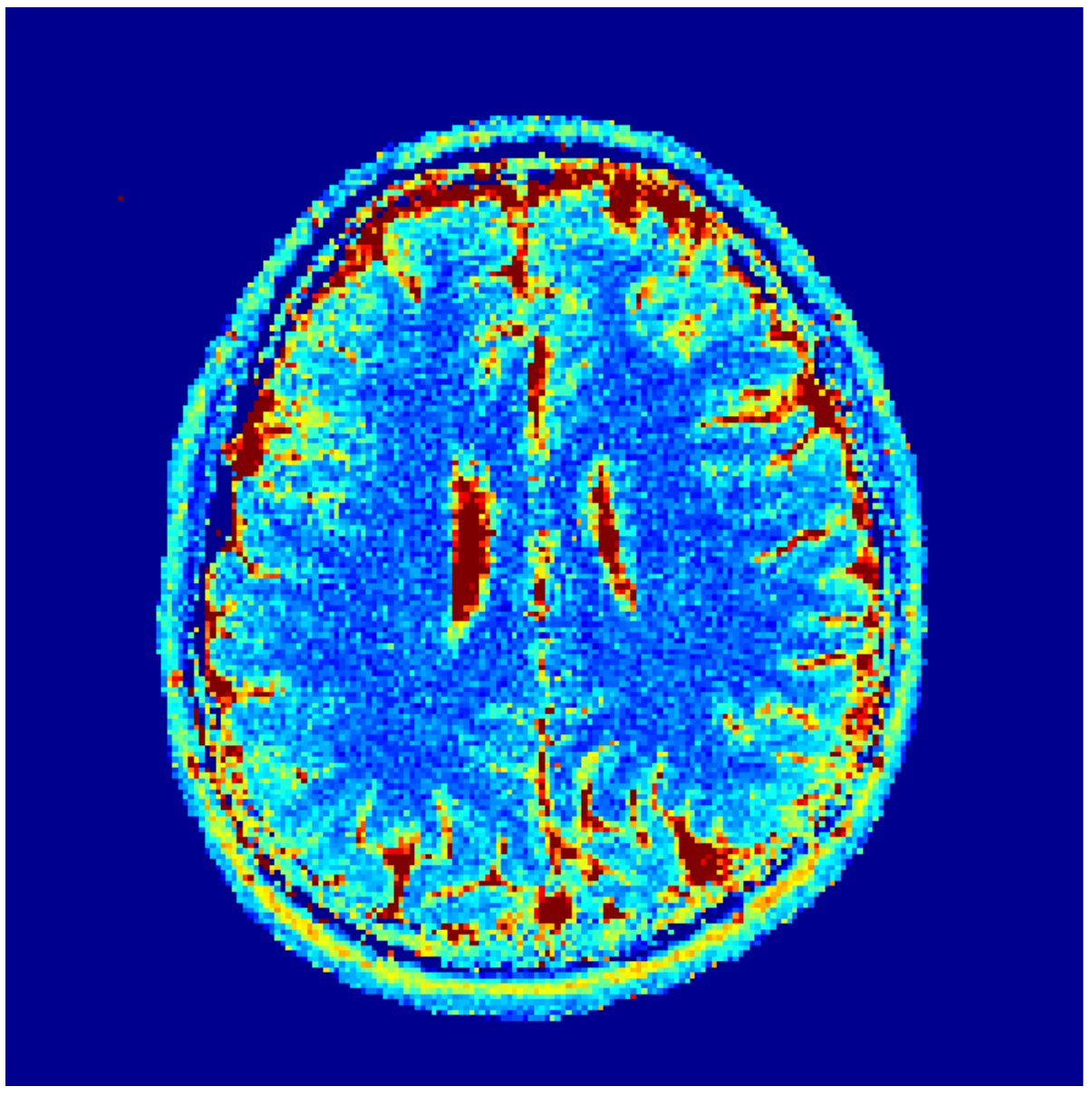}\hspace{-.1cm}
		\includegraphics[width=.162\linewidth]{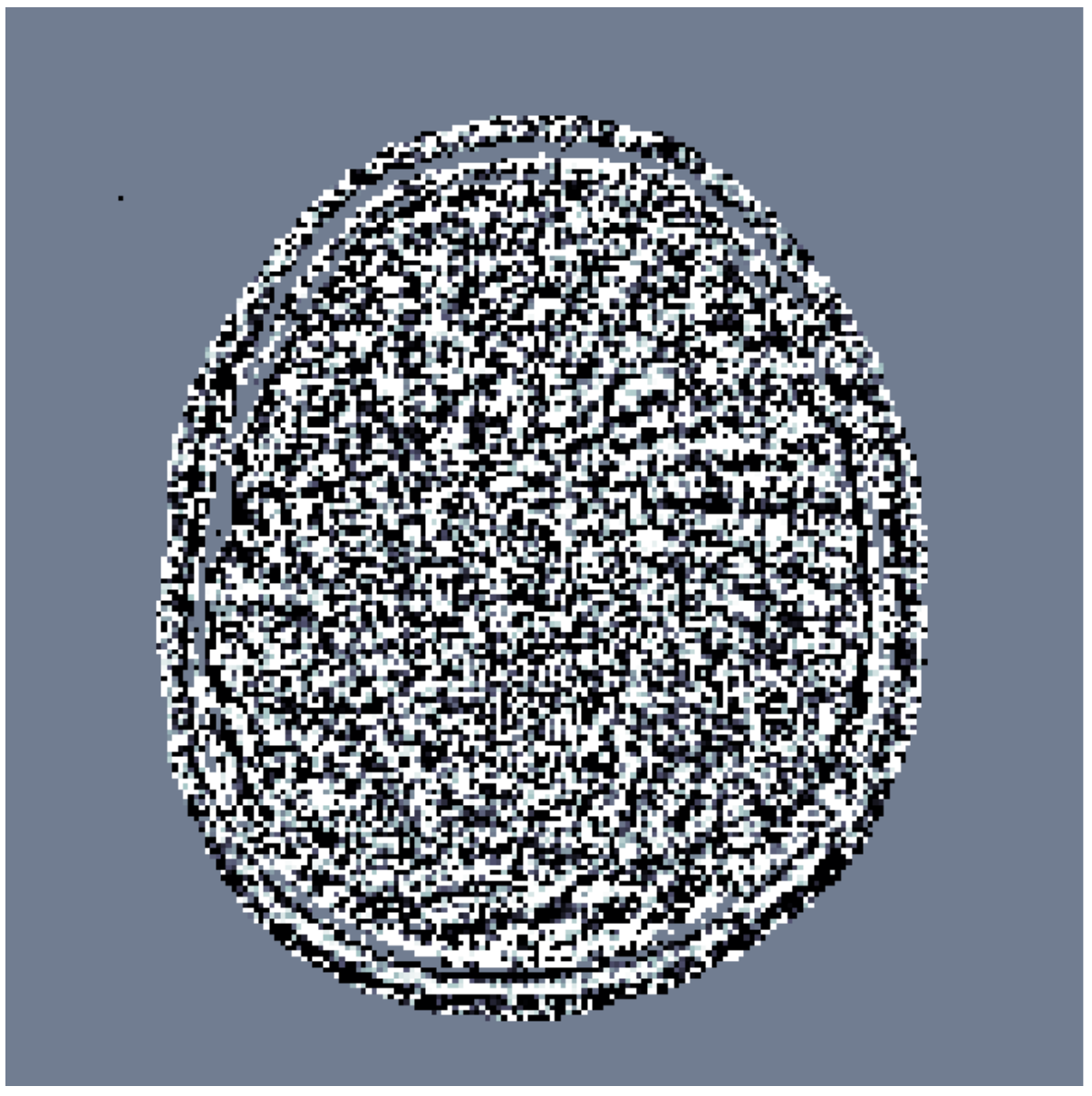}\hspace{.0cm}
		\includegraphics[width=.162\linewidth]{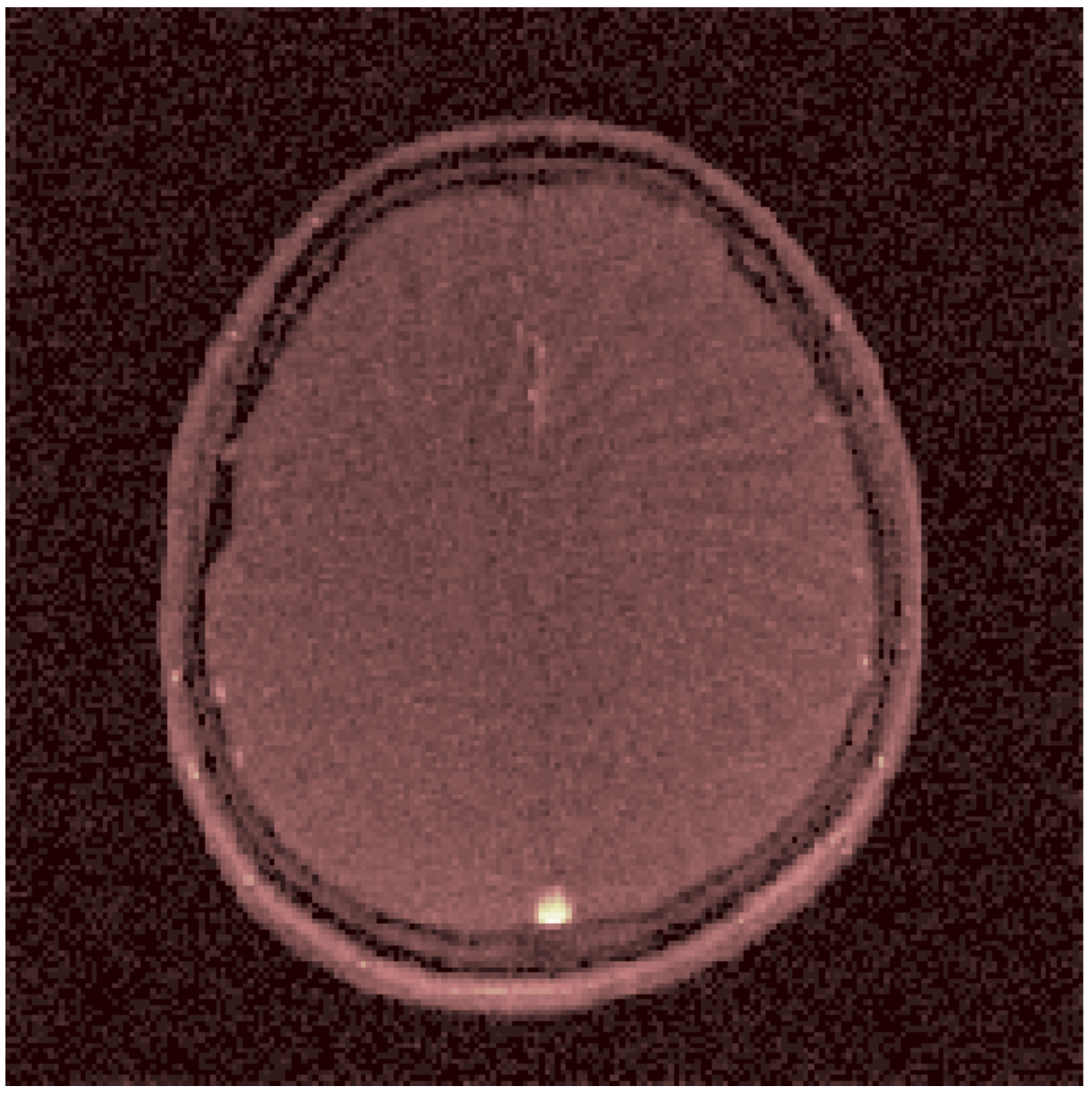}\hspace{-.1cm}
		\includegraphics[width=.162\linewidth]{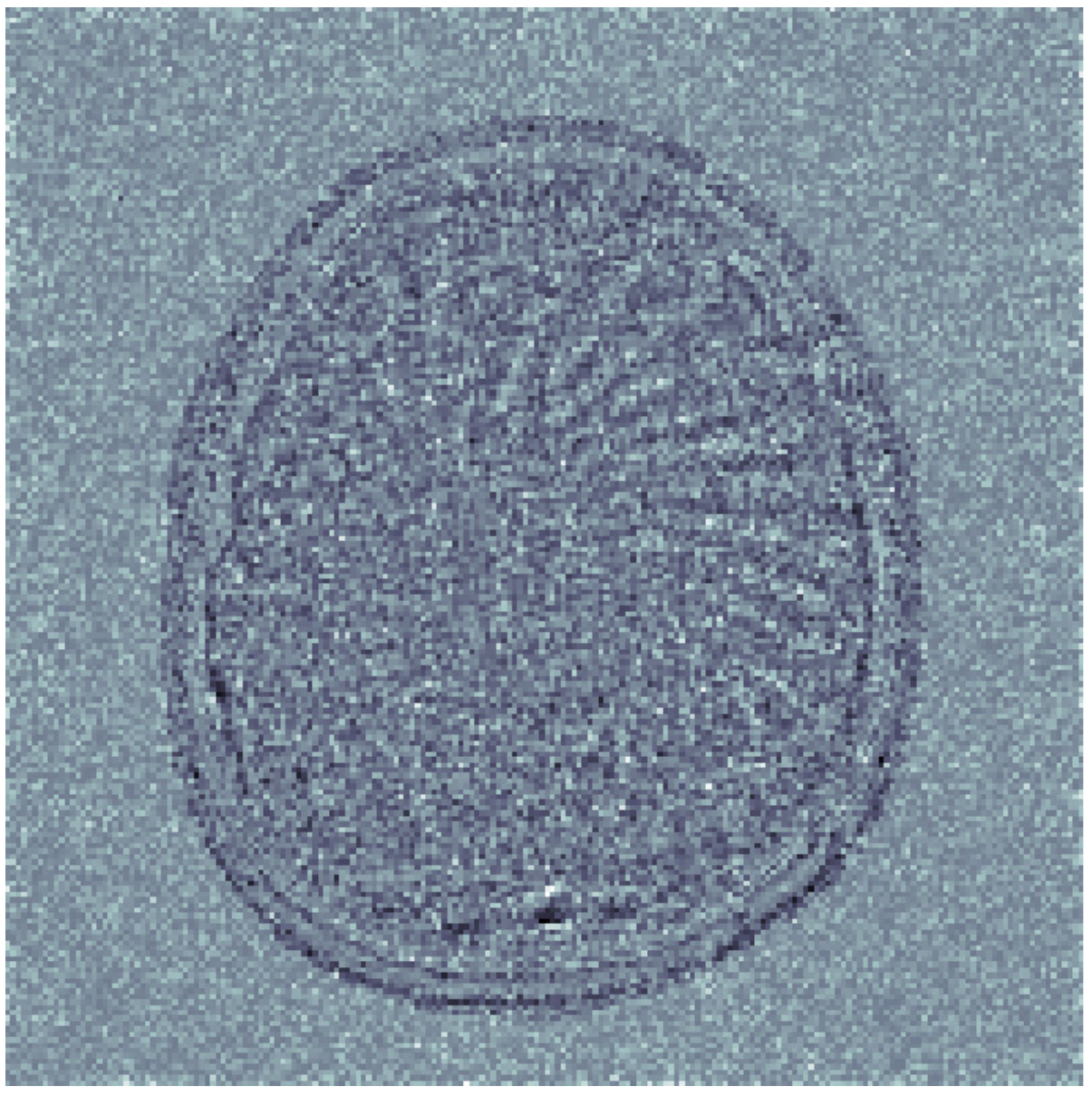}
		\\
		\begin{turn}{90} \quad\qquad VS-DM \end{turn}
		\includegraphics[width=.162\linewidth]{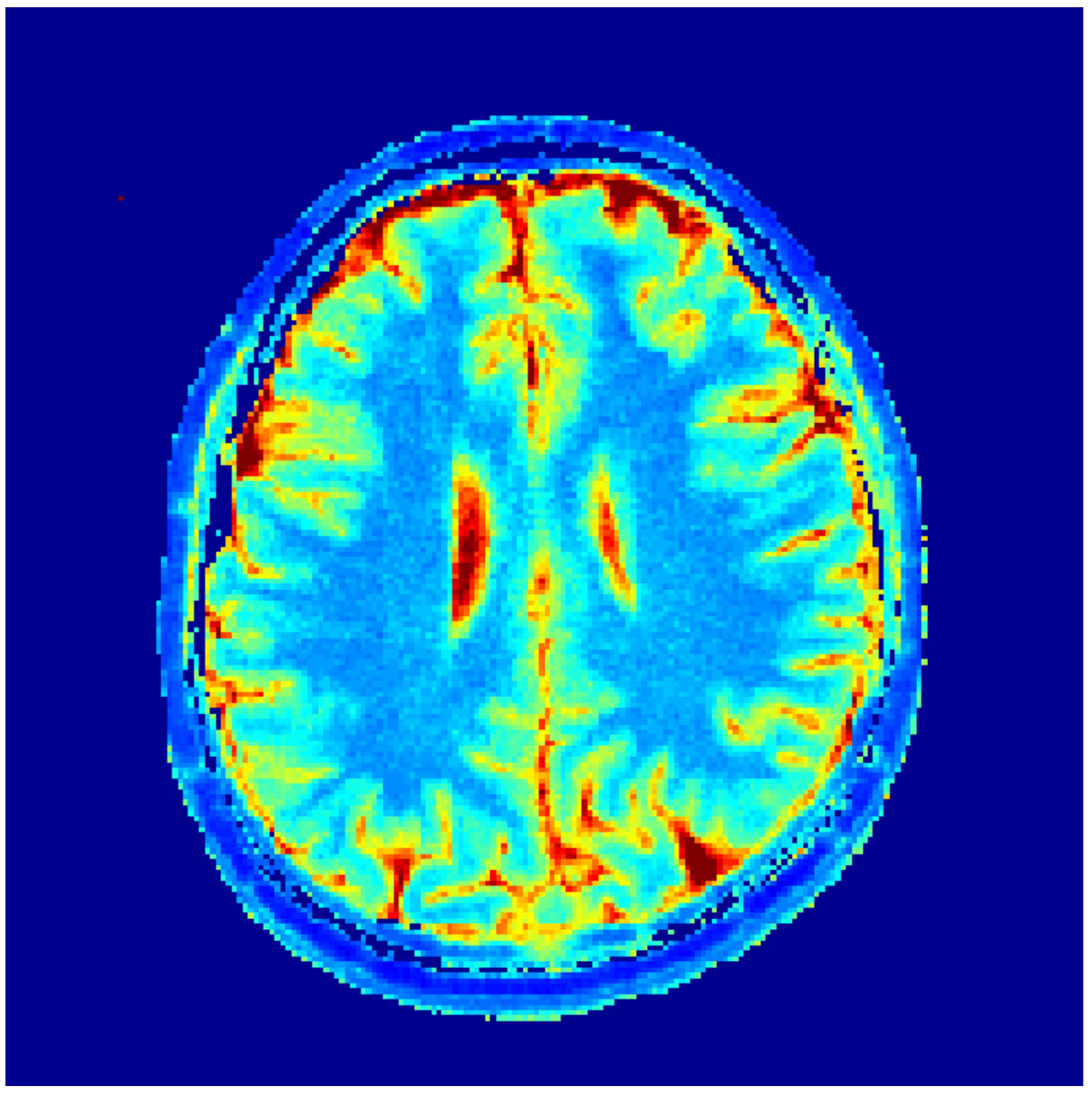}\hspace{-.1cm}
		\includegraphics[width=.162\linewidth]{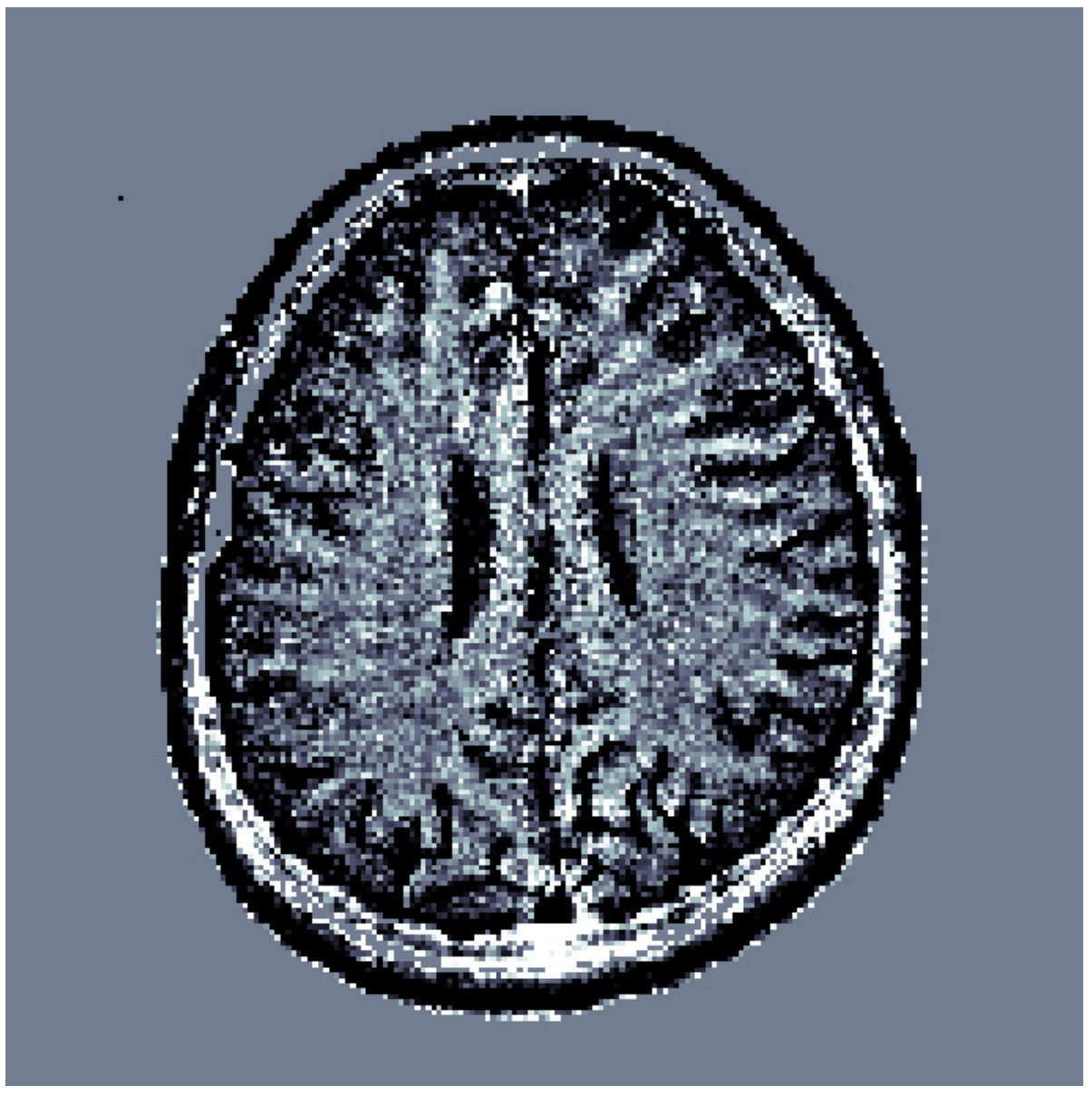}\hspace{.0cm}
		\includegraphics[width=.162\linewidth]{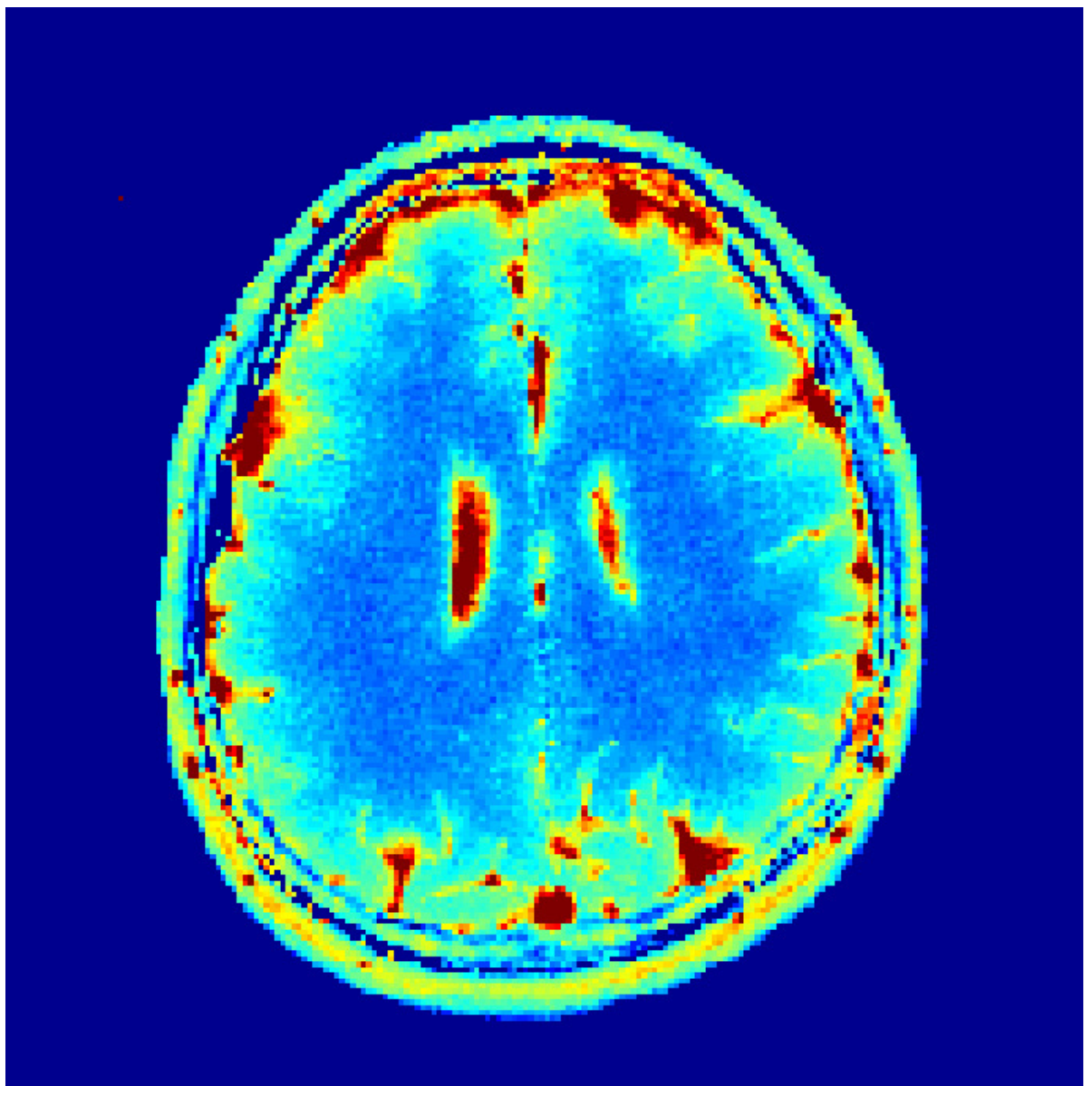}\hspace{-.1cm}
		\includegraphics[width=.162\linewidth]{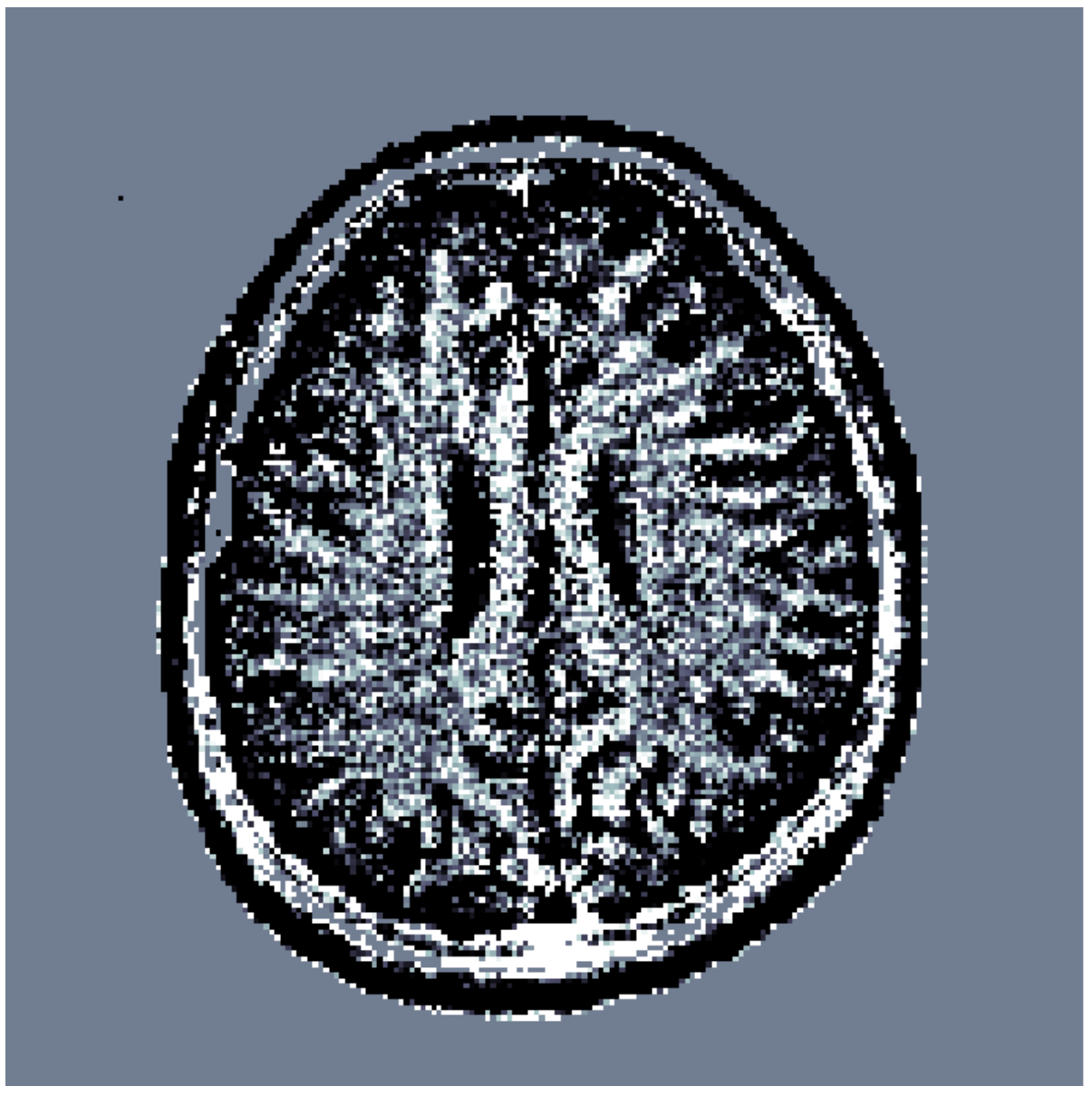}\hspace{.0cm}
		\includegraphics[width=.162\linewidth]{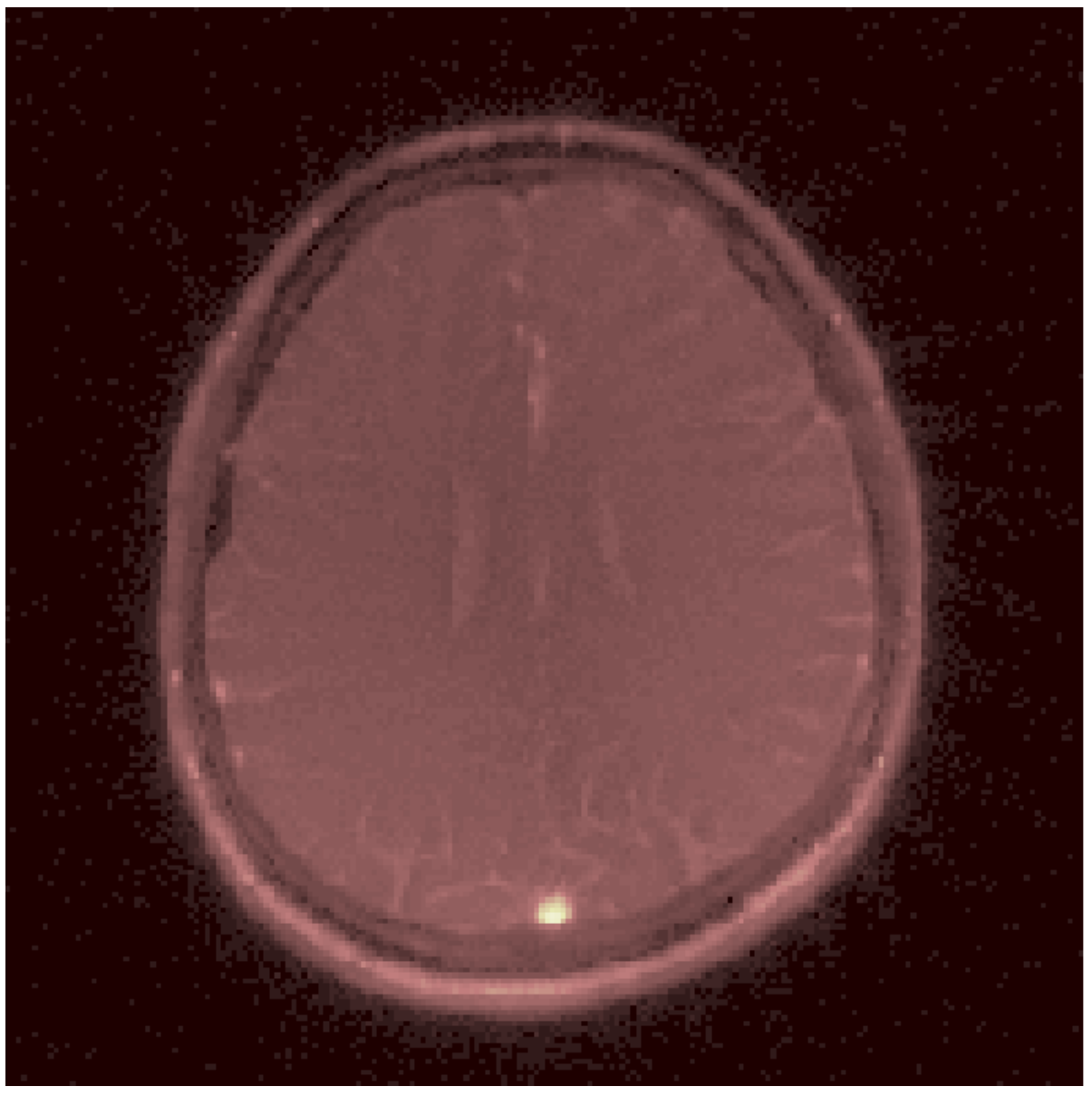}\hspace{-.1cm}
		\includegraphics[width=.162\linewidth]{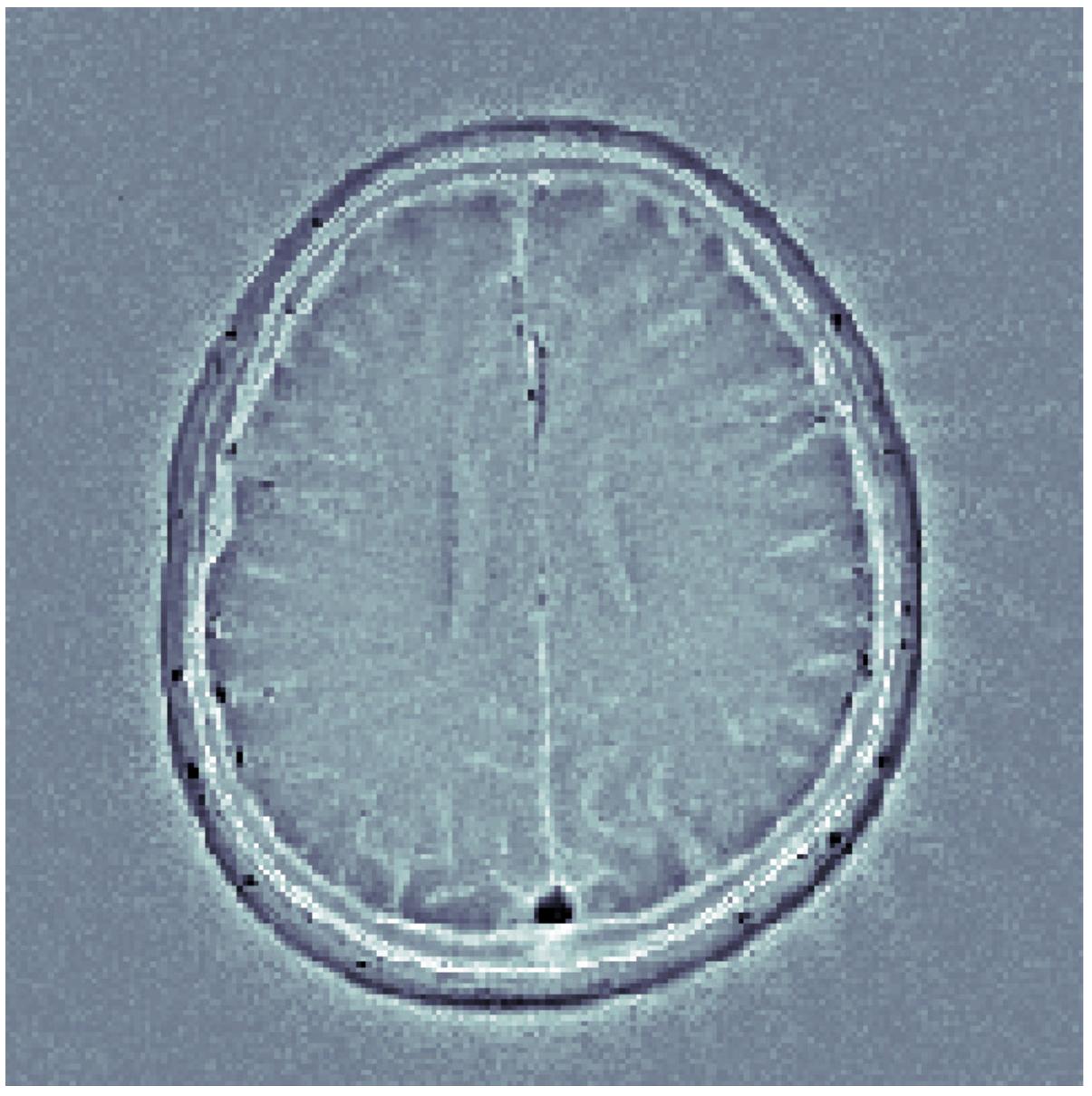}	\\
		\begin{turn}{90} \quad\qquad FLOR\end{turn}
		\includegraphics[width=.162\linewidth]{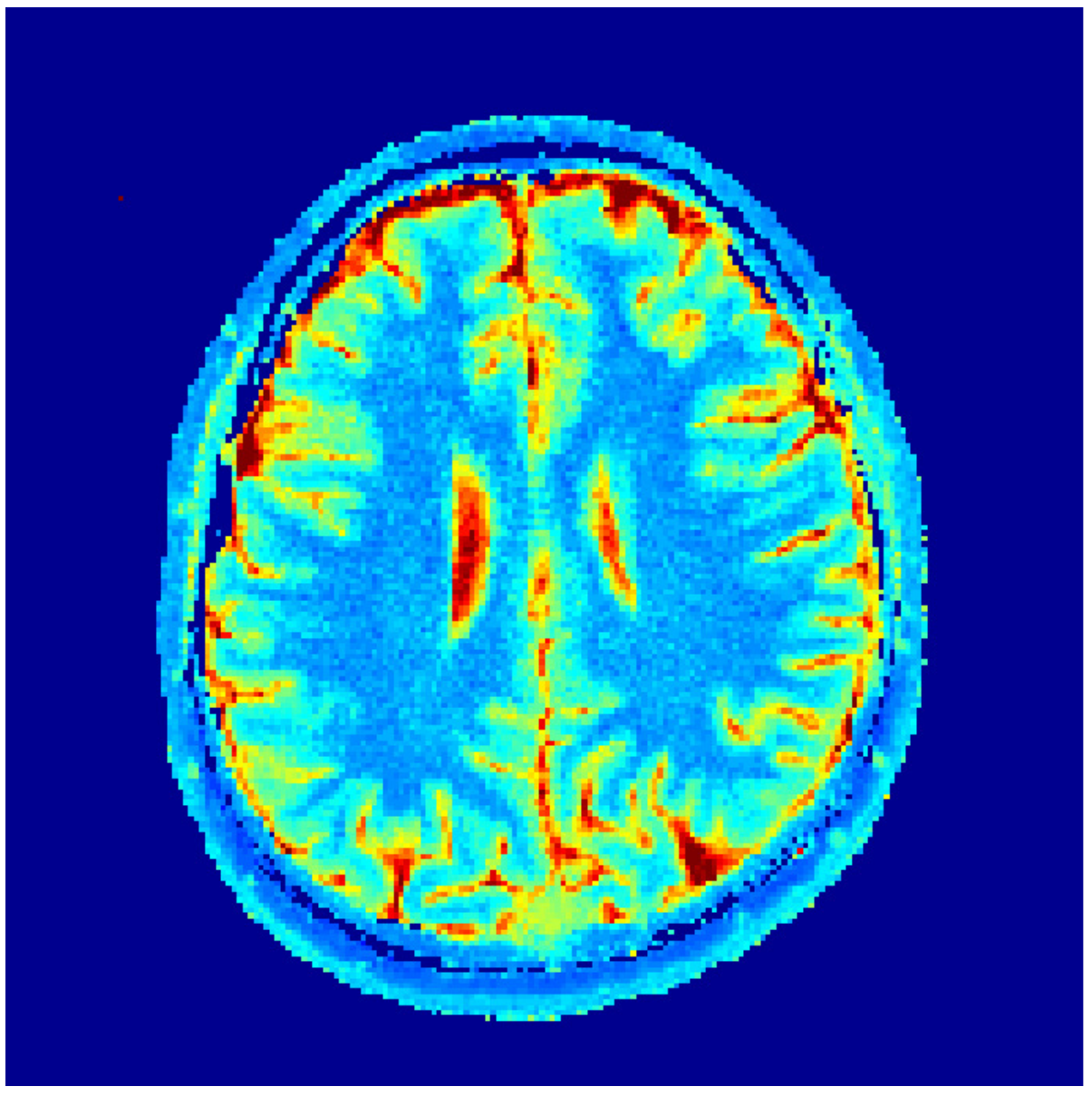}\hspace{-.1cm}
		\includegraphics[width=.162\linewidth]{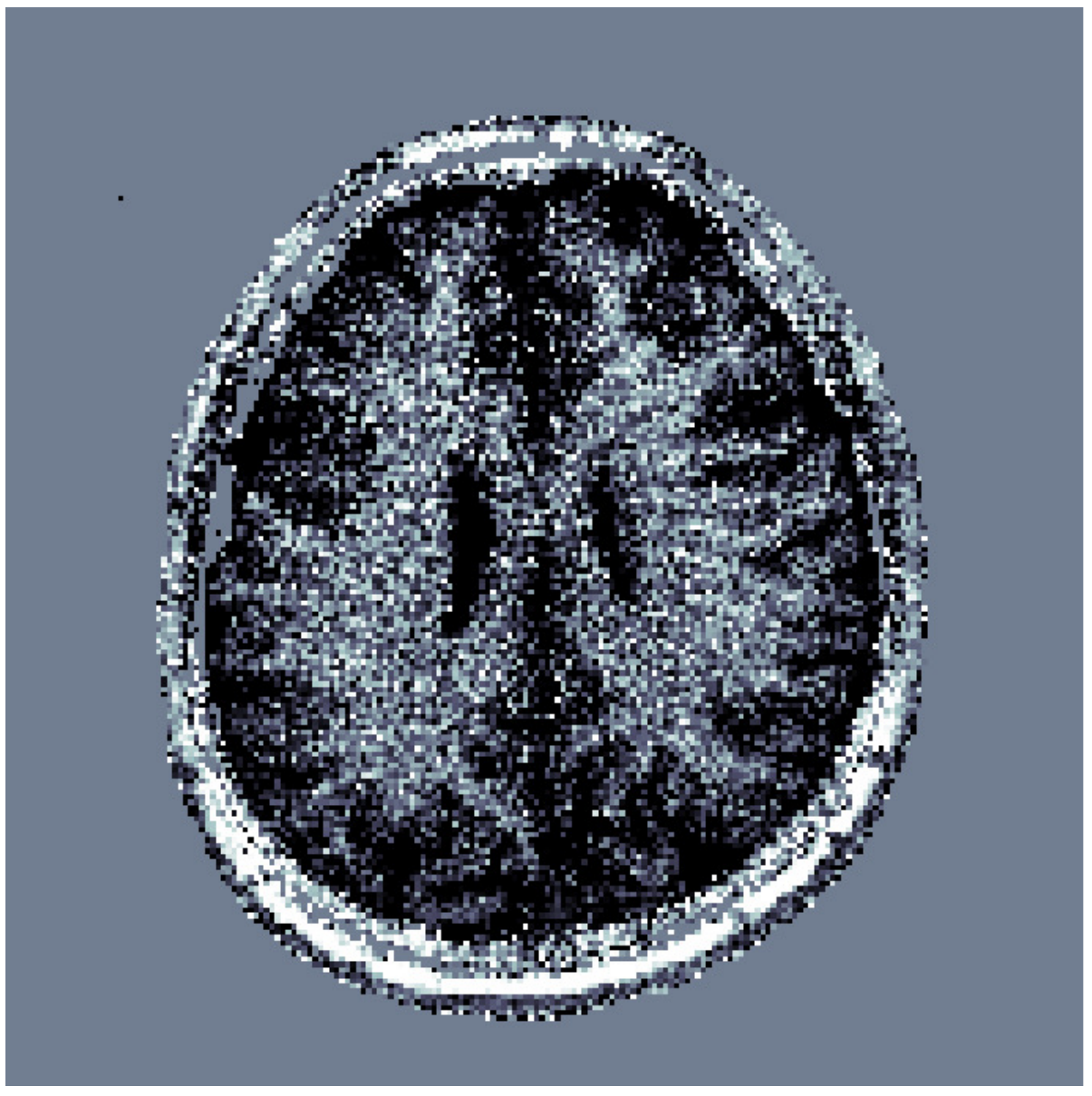}\hspace{.0cm}
		\includegraphics[width=.162\linewidth]{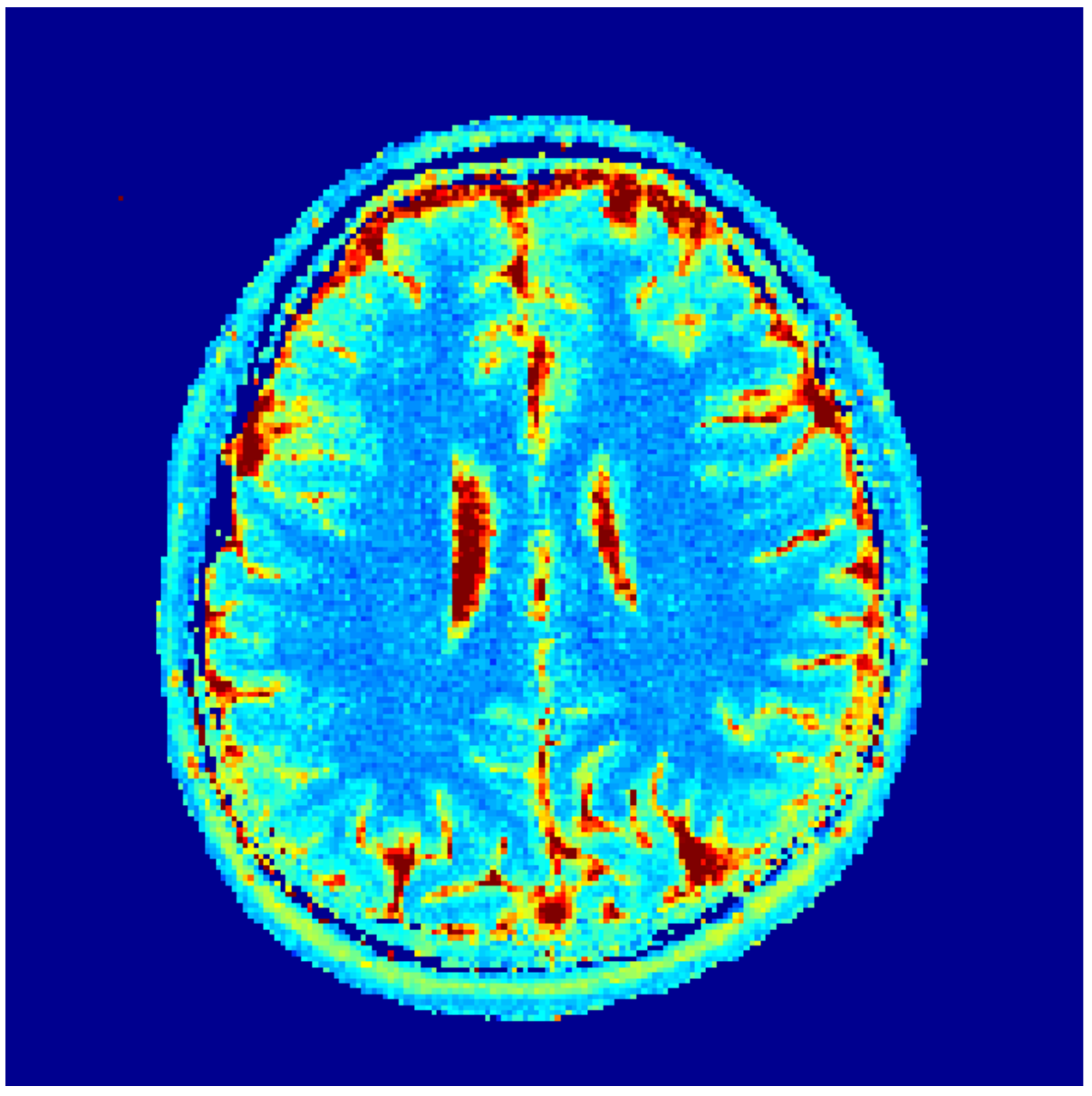}\hspace{-.1cm}
		\includegraphics[width=.162\linewidth]{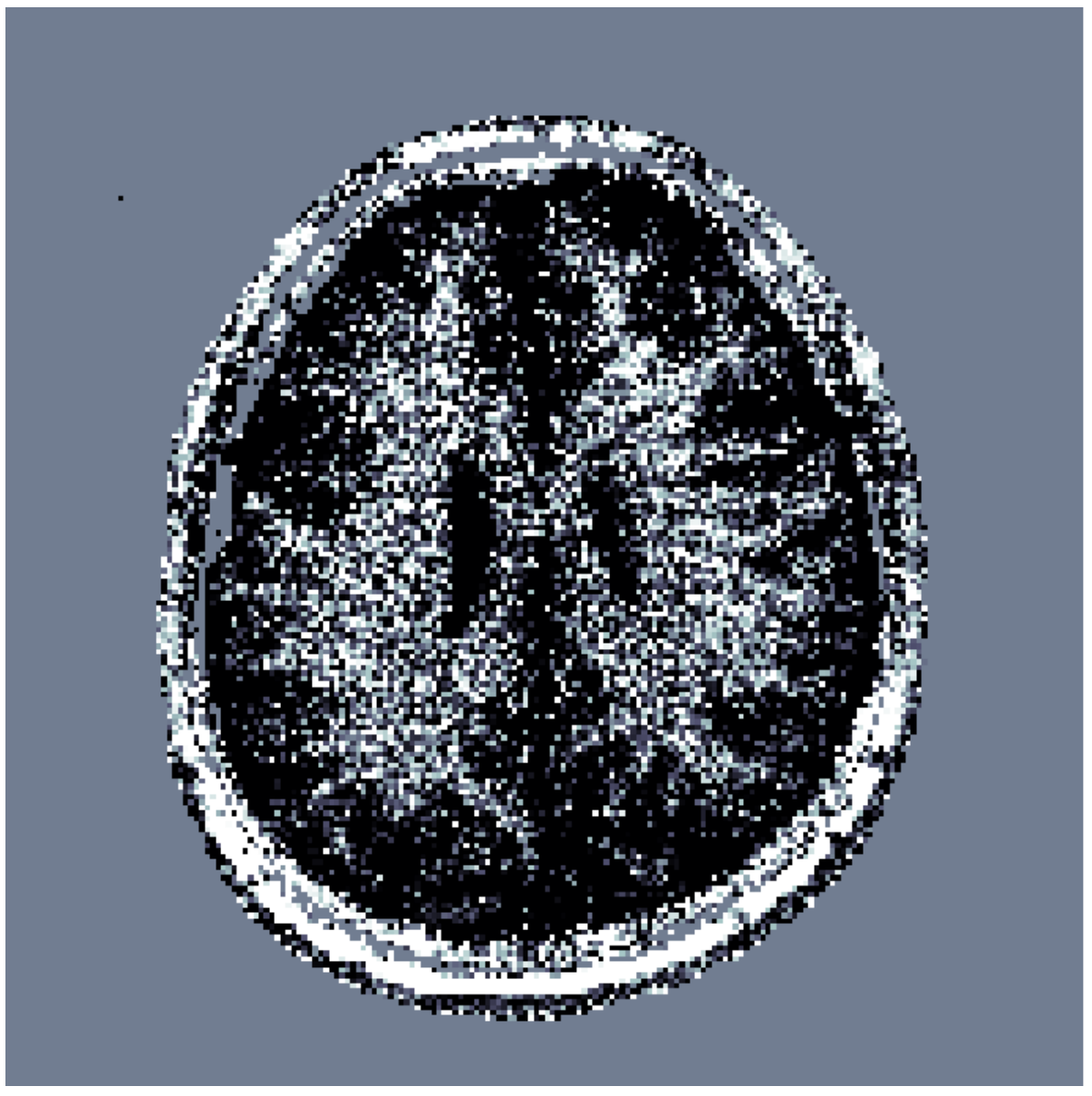}\hspace{.0cm}
		\includegraphics[width=.162\linewidth]{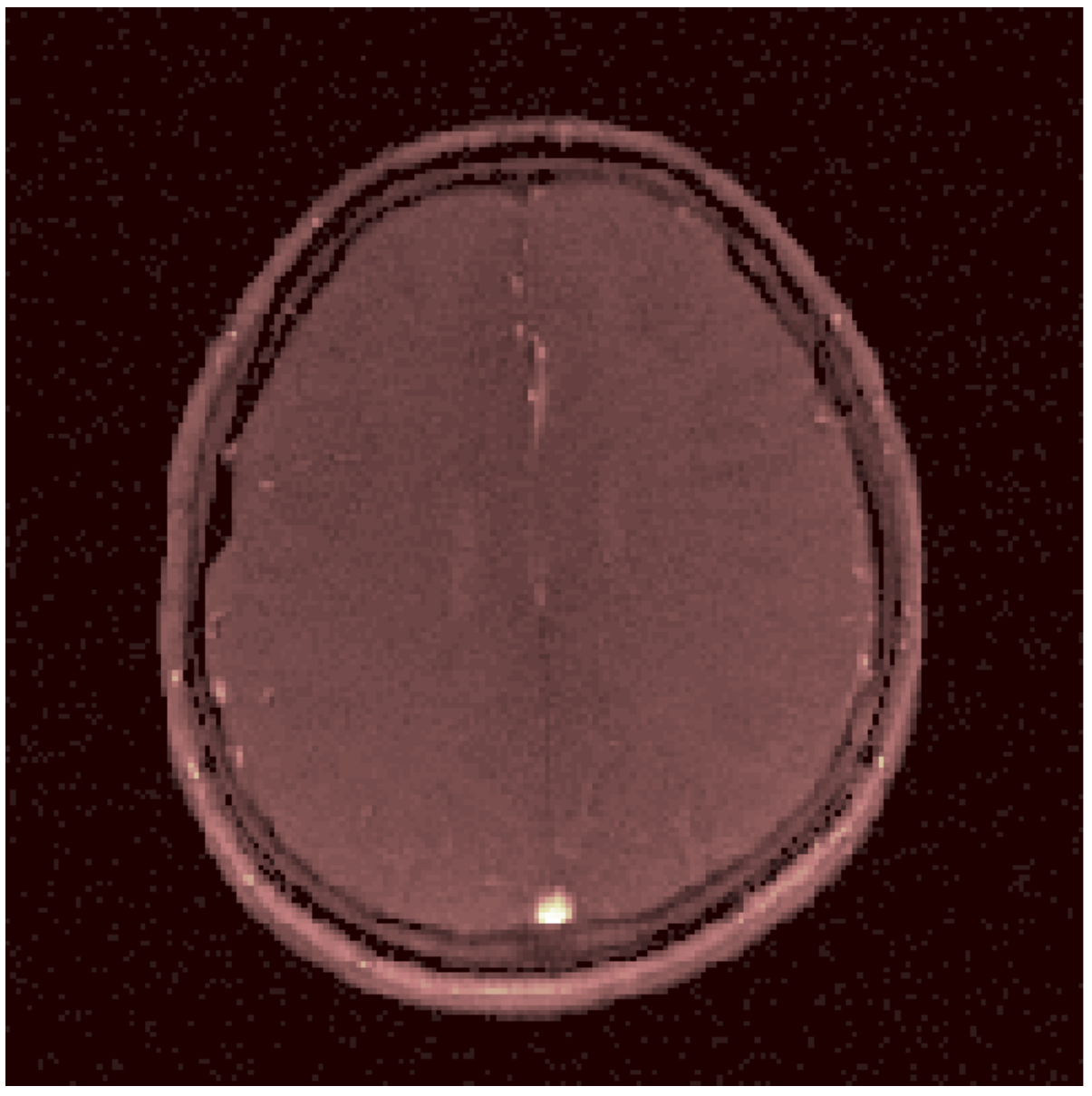}\hspace{-.1cm}
		\includegraphics[width=.162\linewidth]{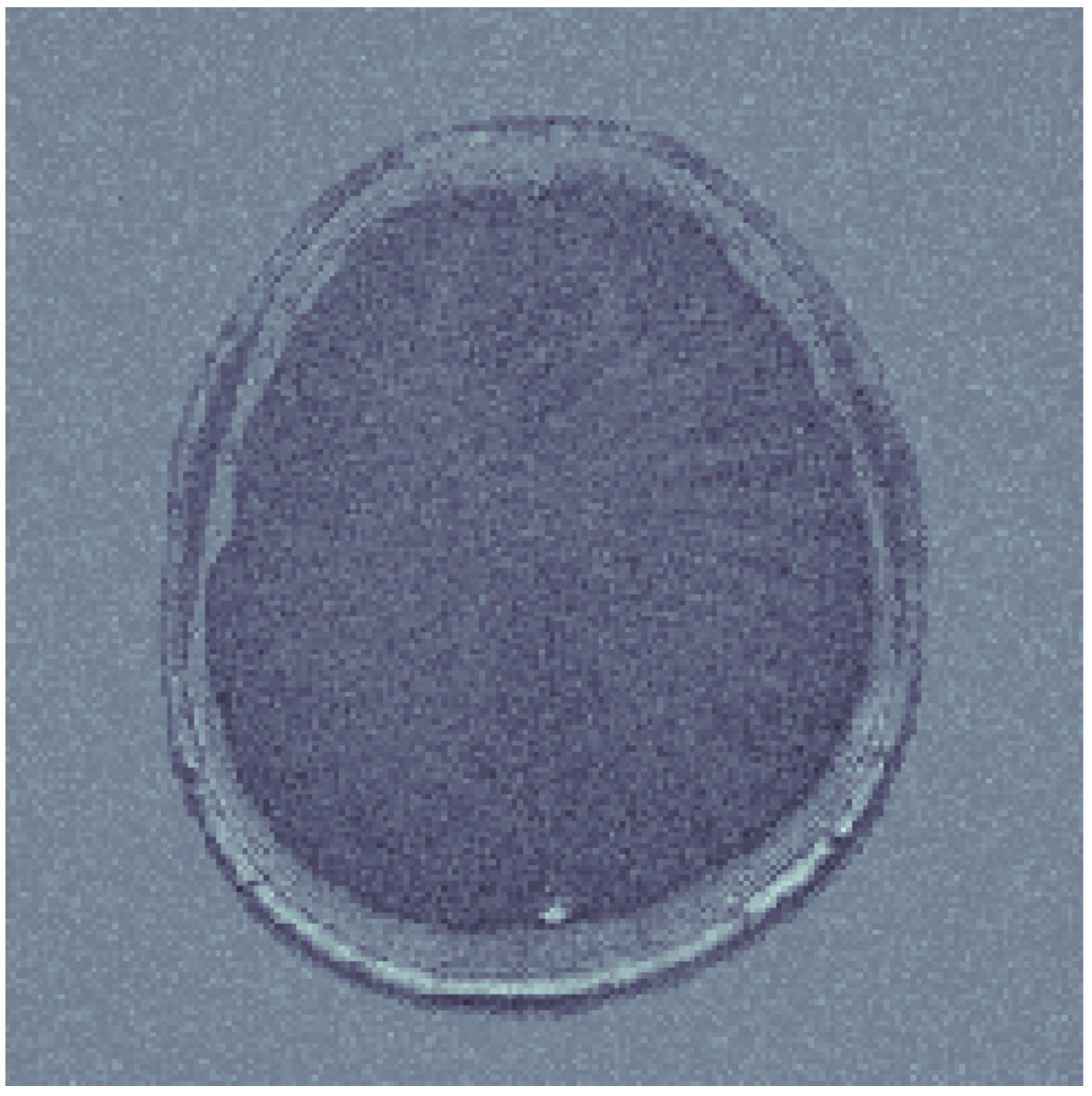}			\\
		\begin{turn}{90} \quad\quad AIR-MRF\end{turn}
		\includegraphics[width=.162\linewidth]{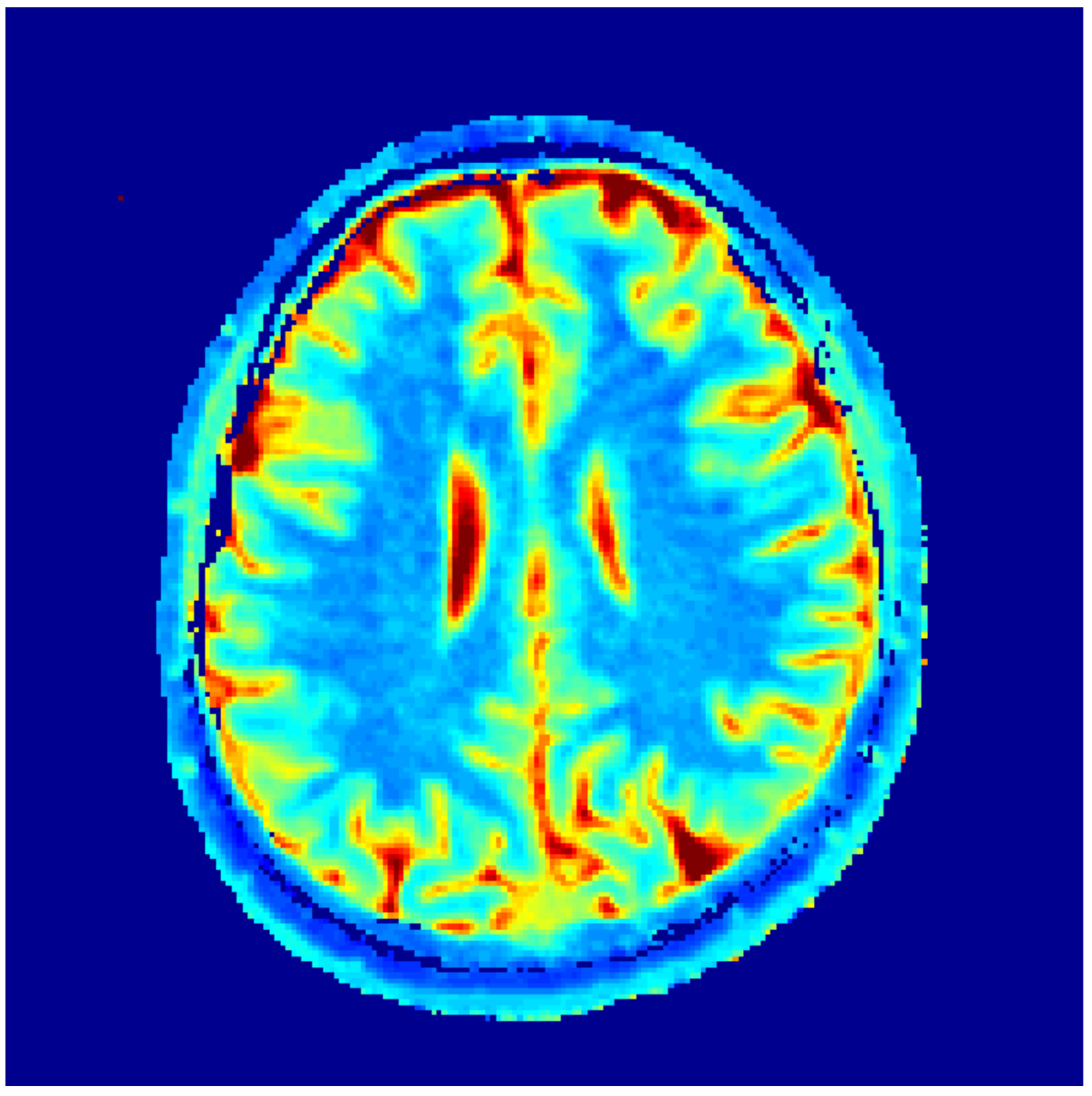}\hspace{-.1cm}
		\includegraphics[width=.162\linewidth]{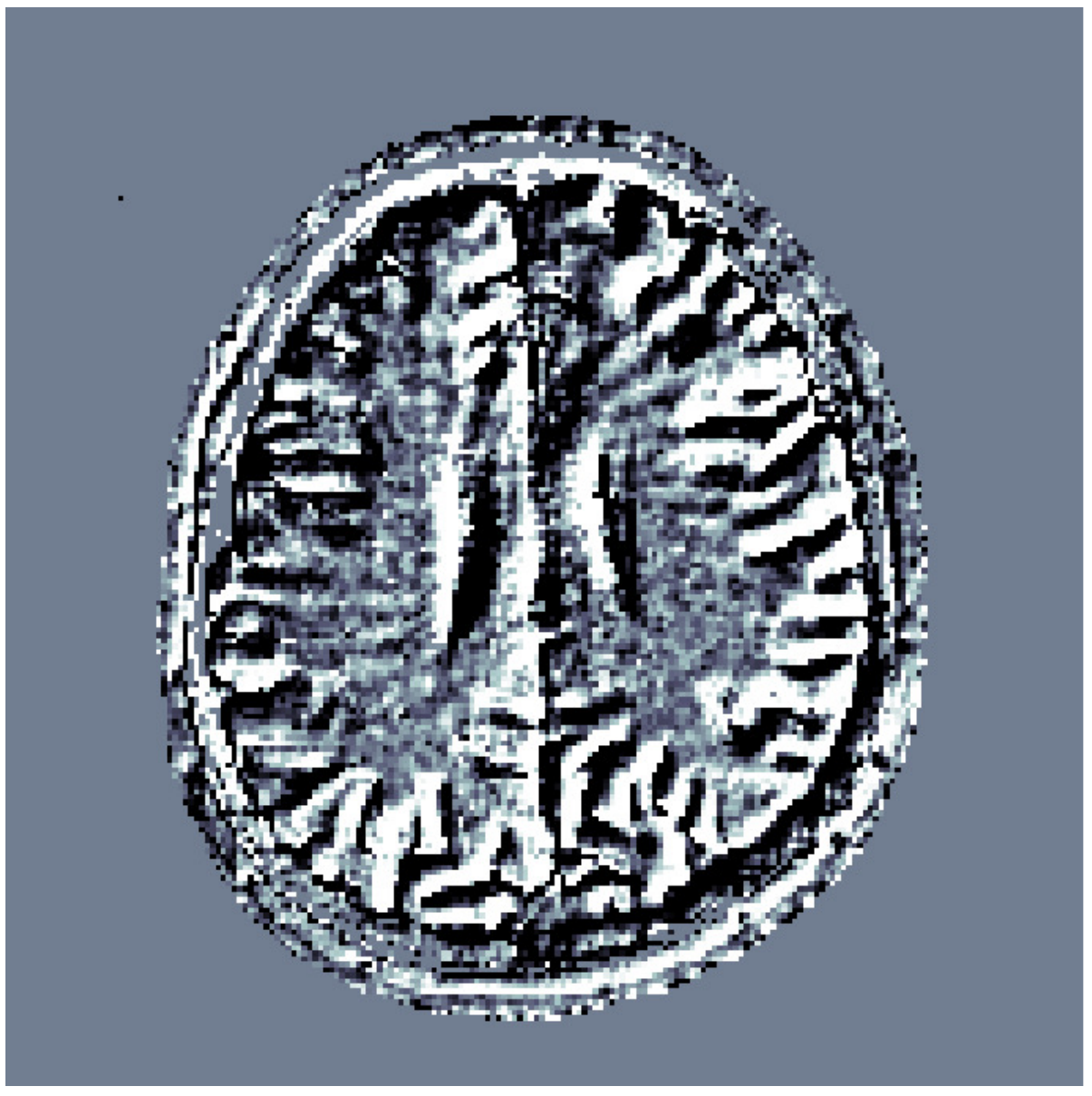}\hspace{.0cm}
		\includegraphics[width=.162\linewidth]{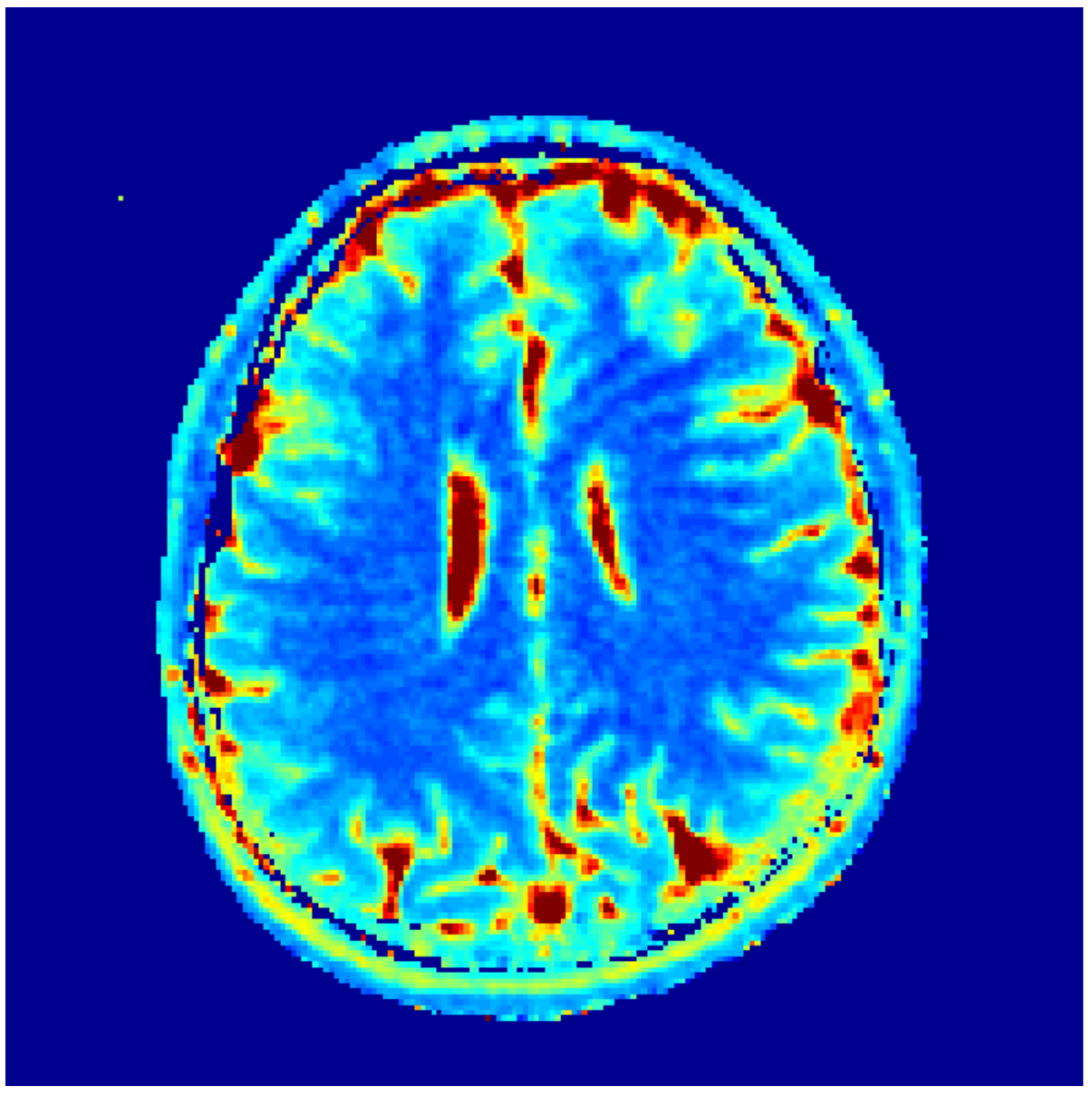}\hspace{-.1cm}
		\includegraphics[width=.162\linewidth]{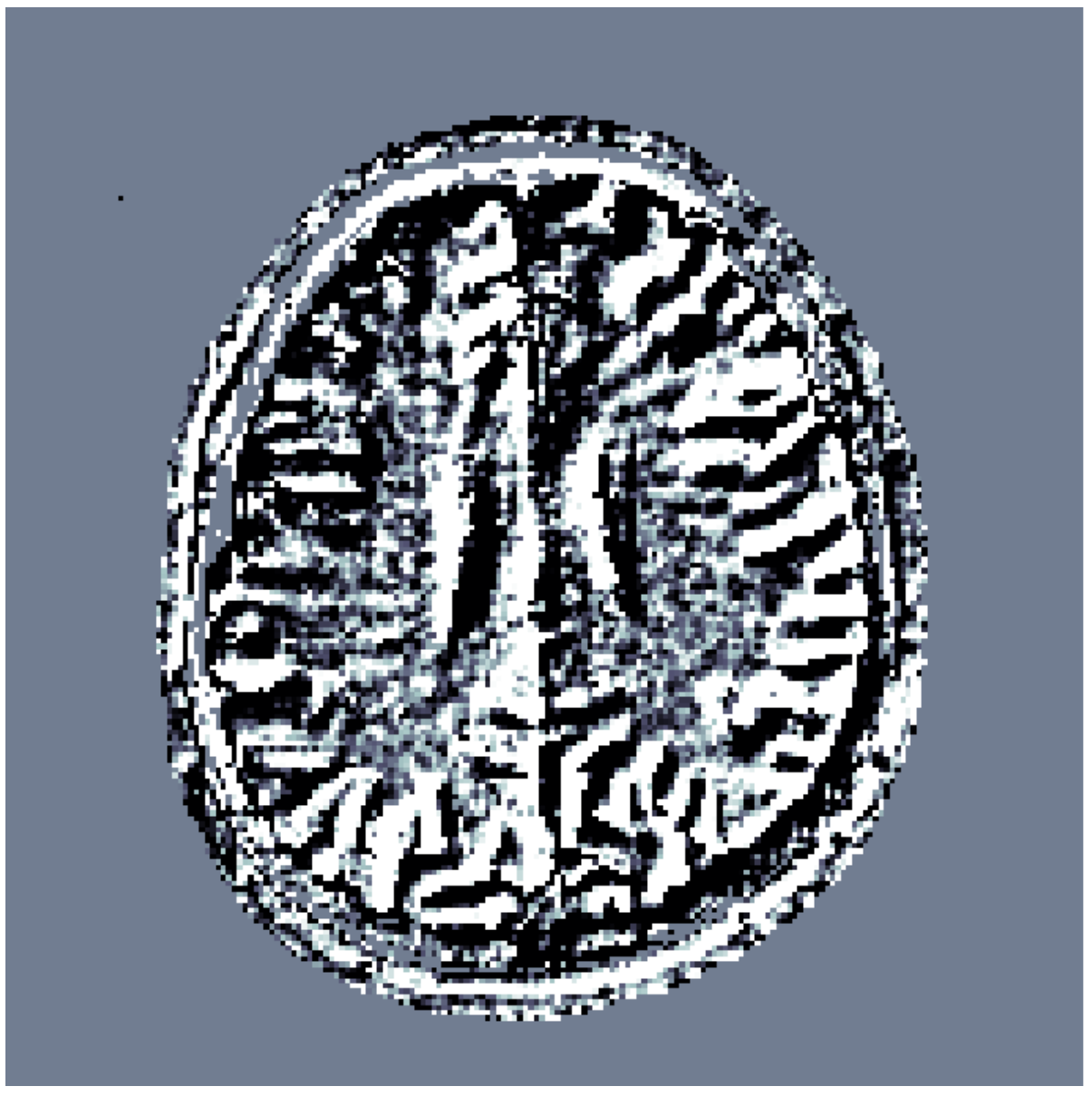}\hspace{.0cm}
		\includegraphics[width=.162\linewidth]{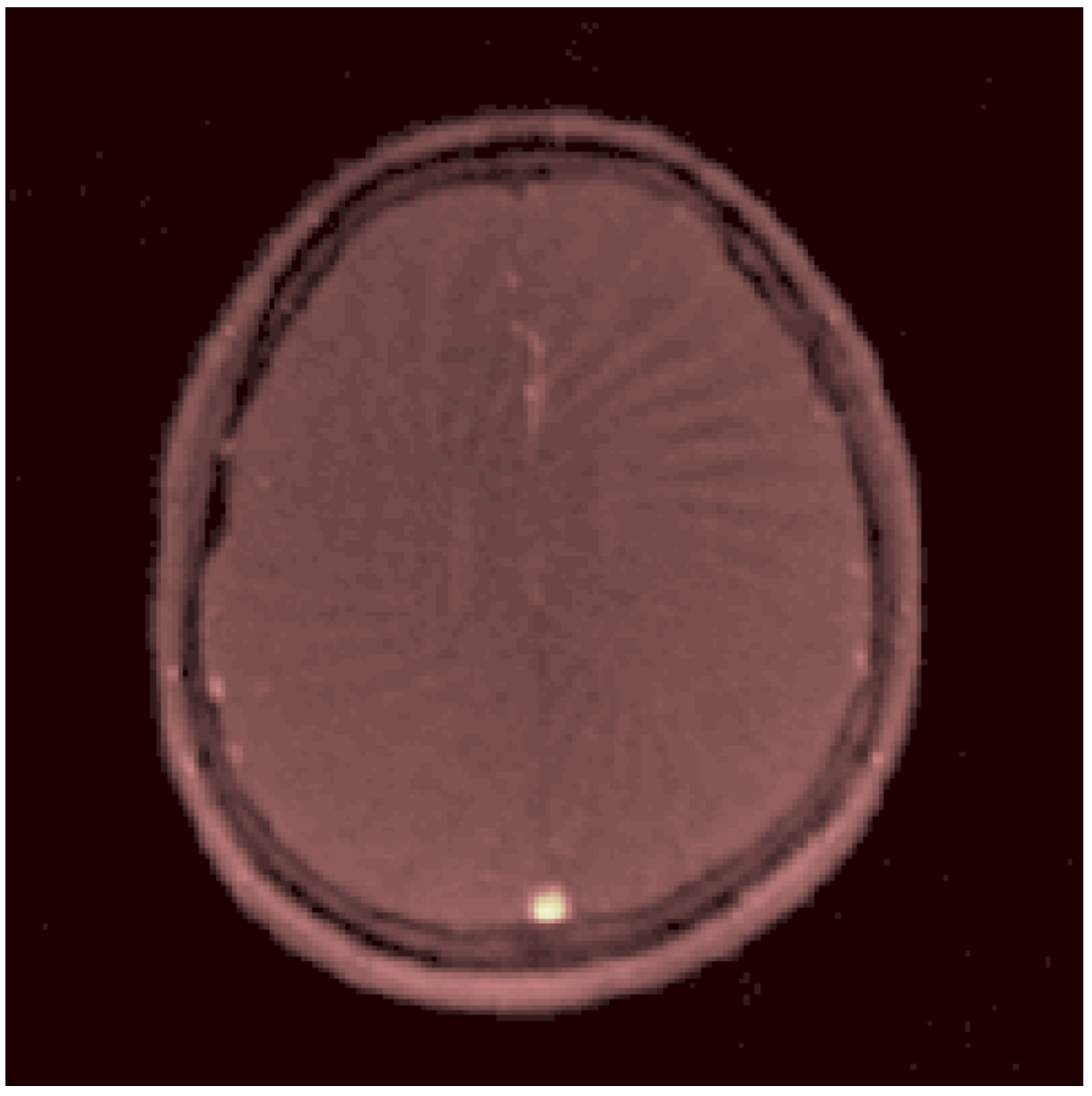}\hspace{-.1cm}
		\includegraphics[width=.162\linewidth]{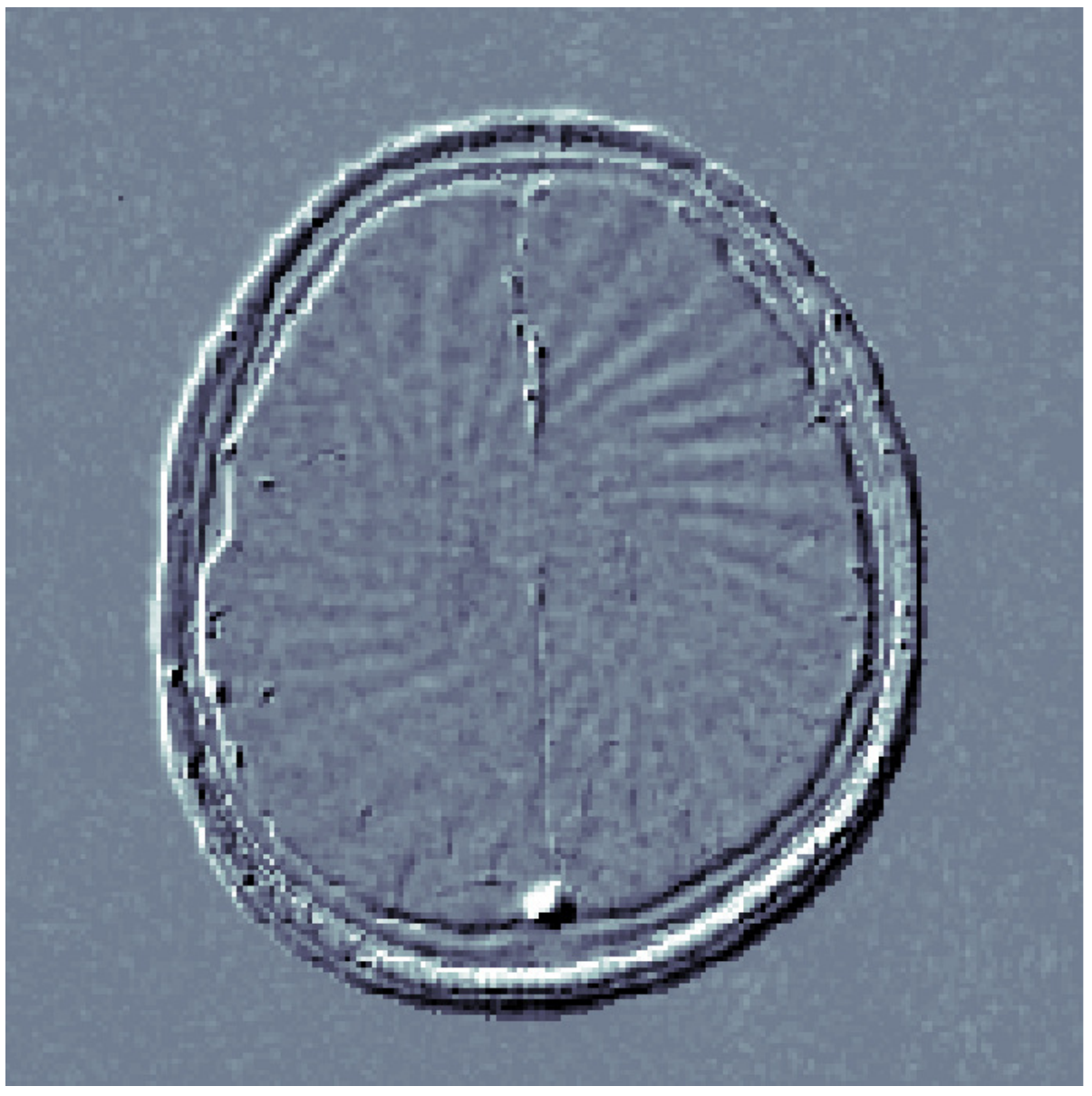}\vspace{.2cm}			
		\hrule
		\hrule
		\hrule
		\begin{turn}{90} \quad\quad LRTV-DM\end{turn}
		\includegraphics[width=.162\linewidth]{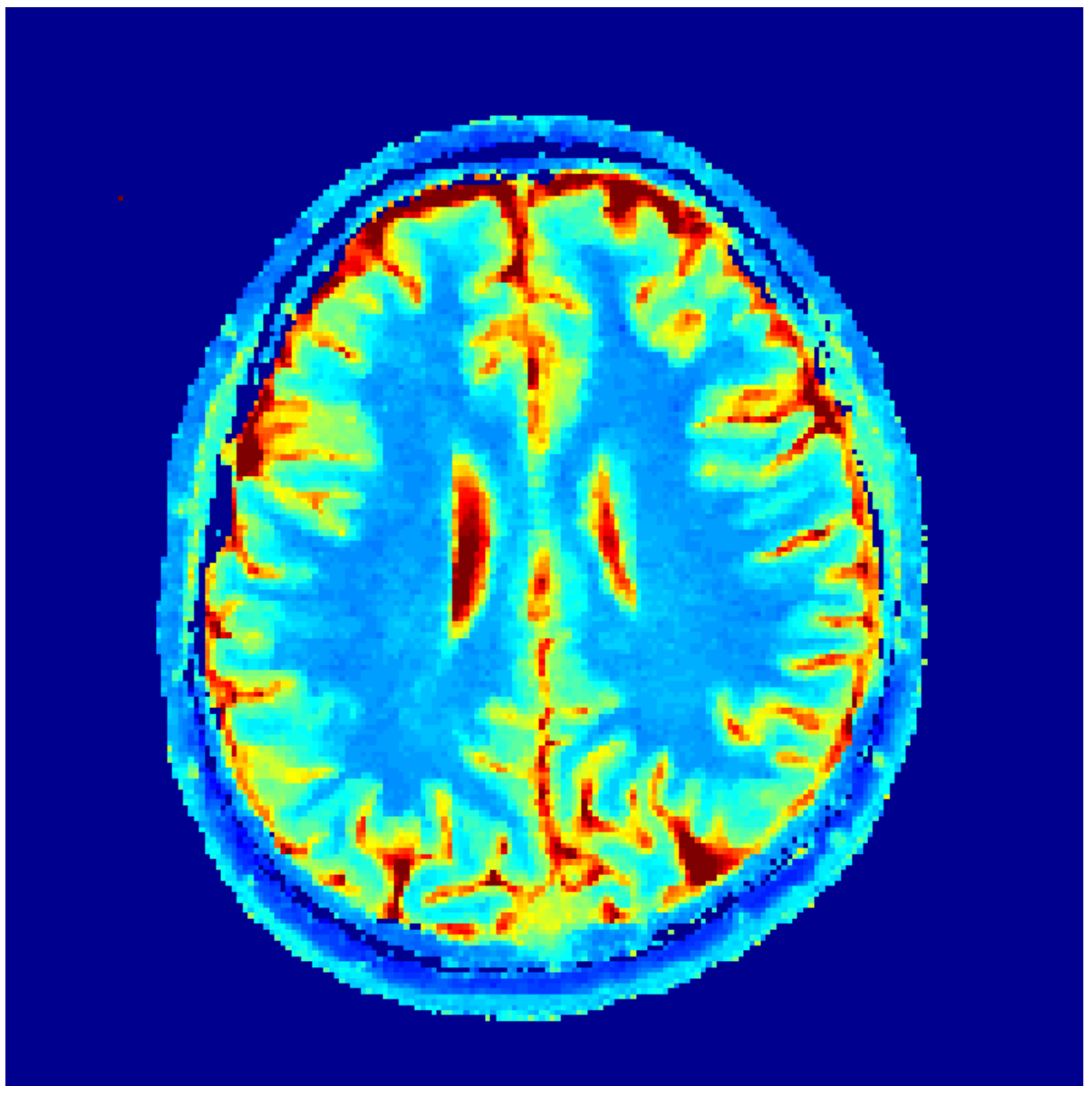}\hspace{-.1cm}
		\includegraphics[width=.162\linewidth]{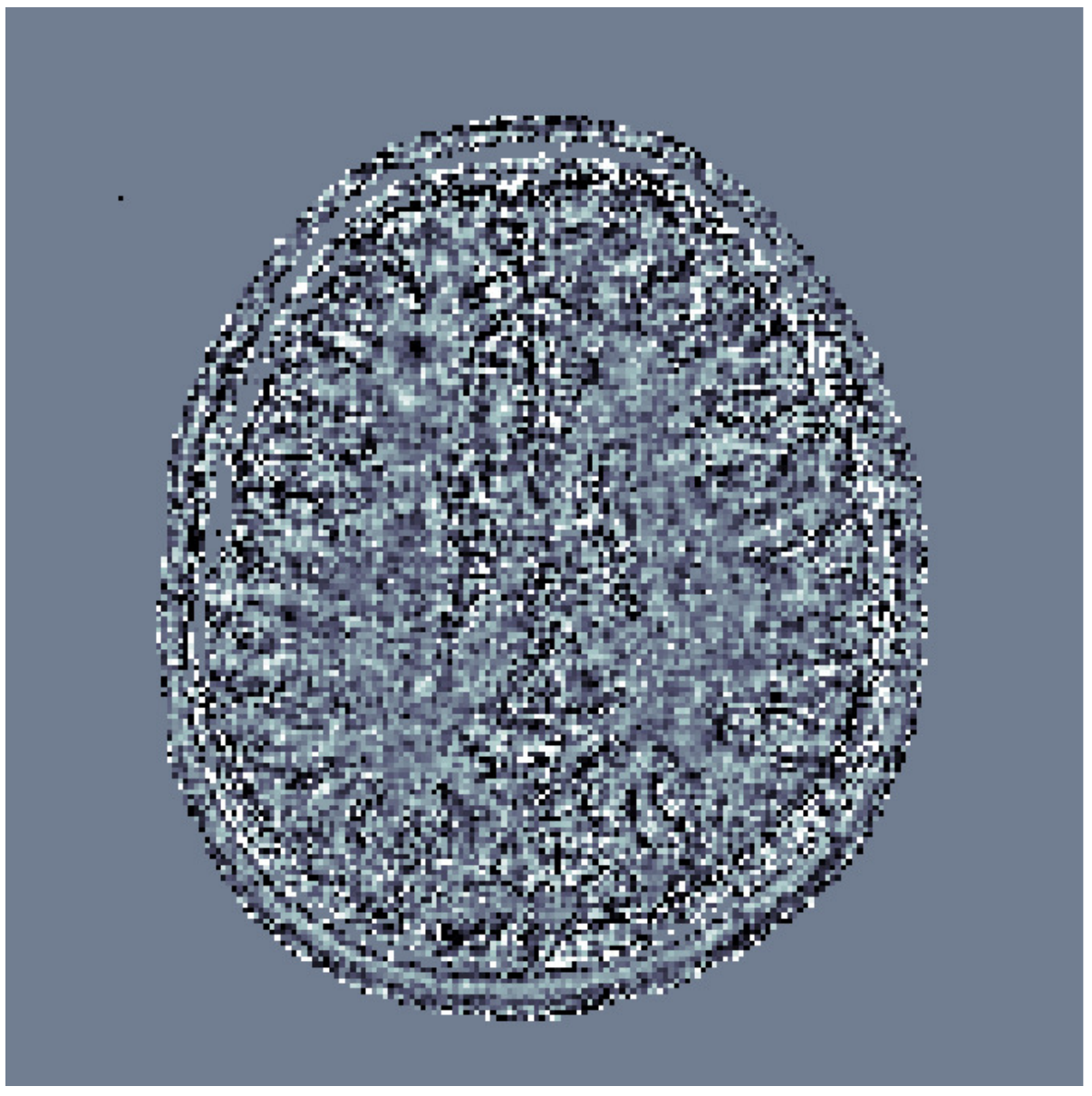}\hspace{.0cm}
		\includegraphics[width=.162\linewidth]{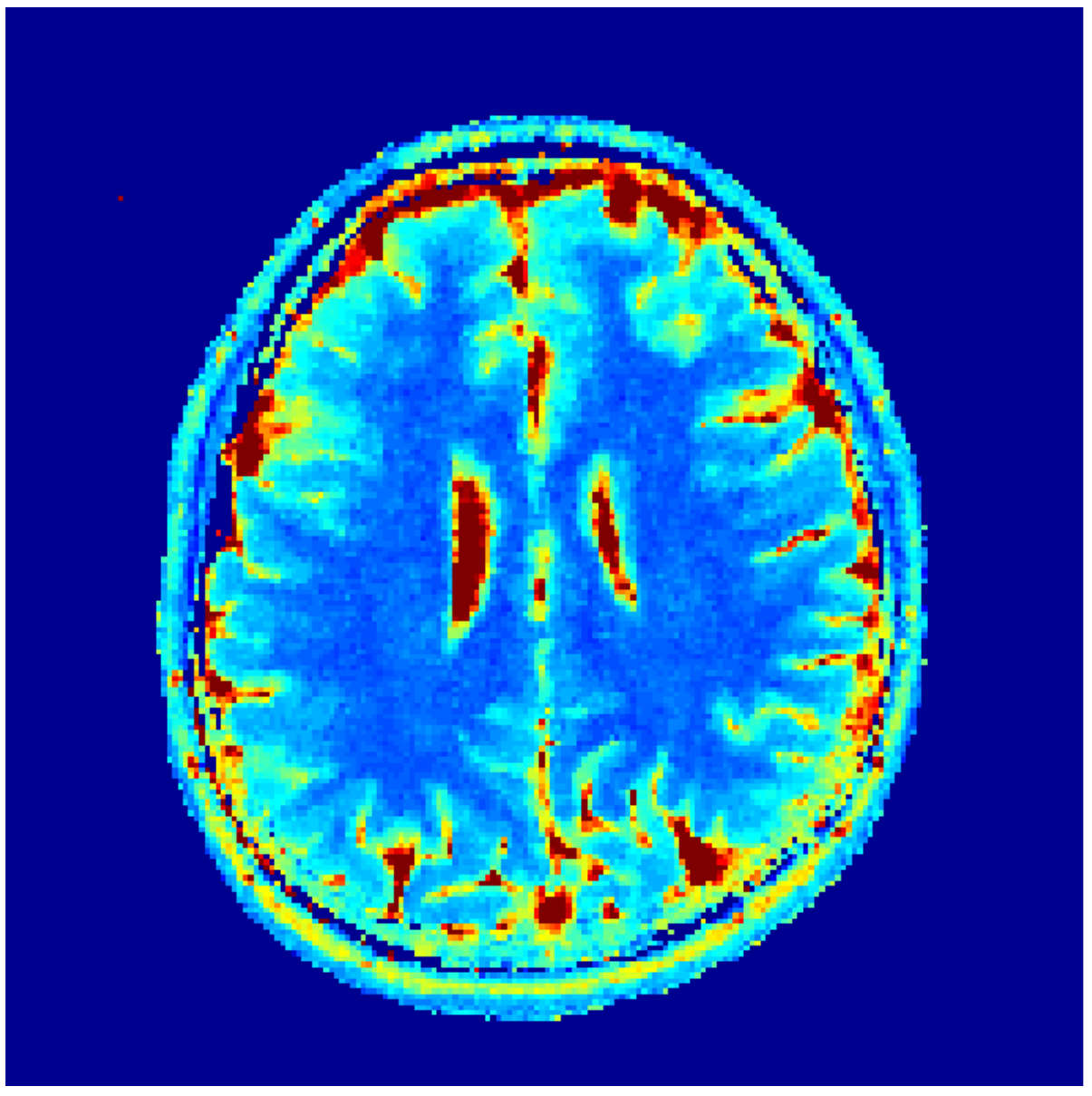}\hspace{-.1cm}
		\includegraphics[width=.162\linewidth]{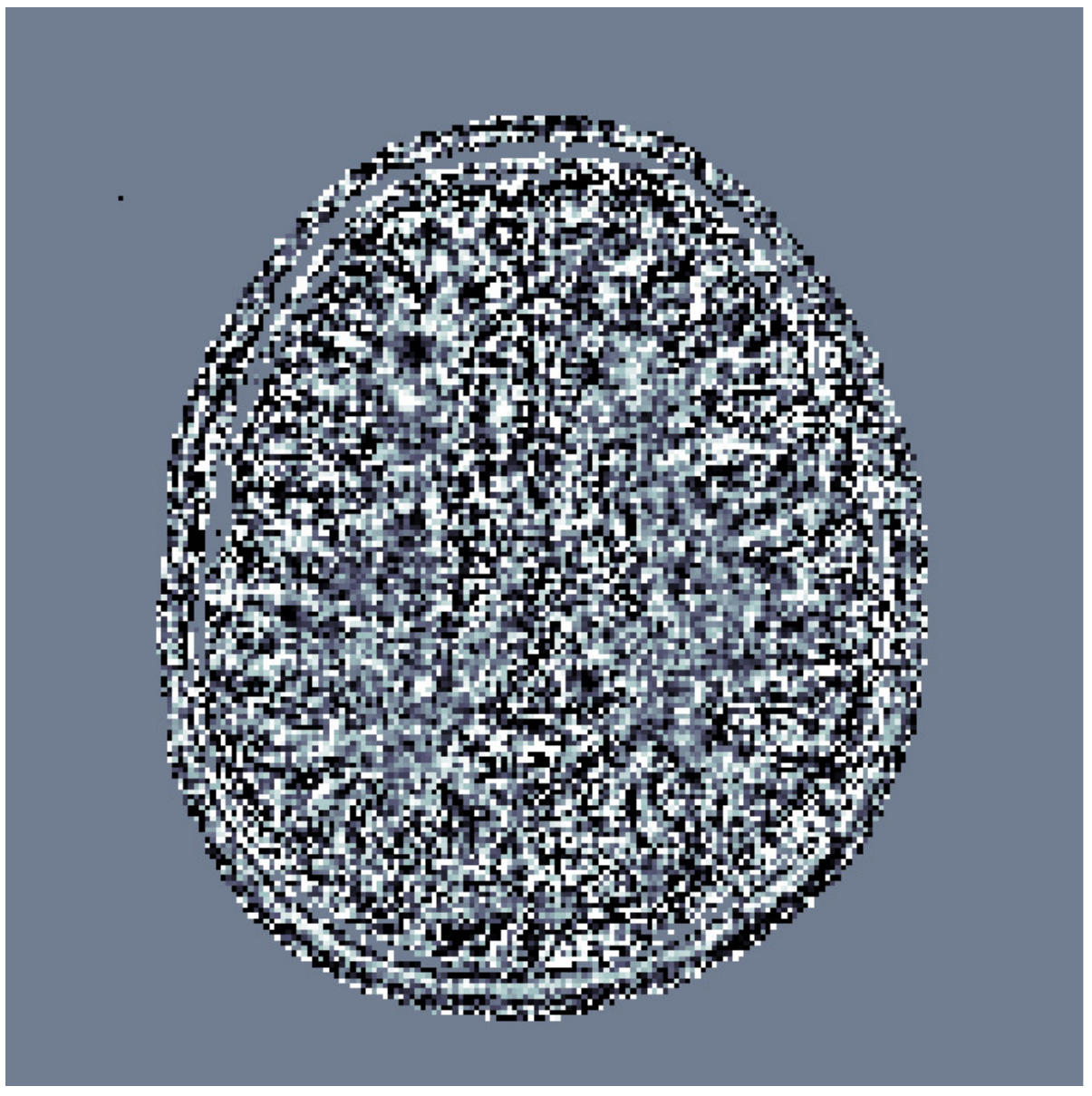}\hspace{.0cm}
		\includegraphics[width=.162\linewidth]{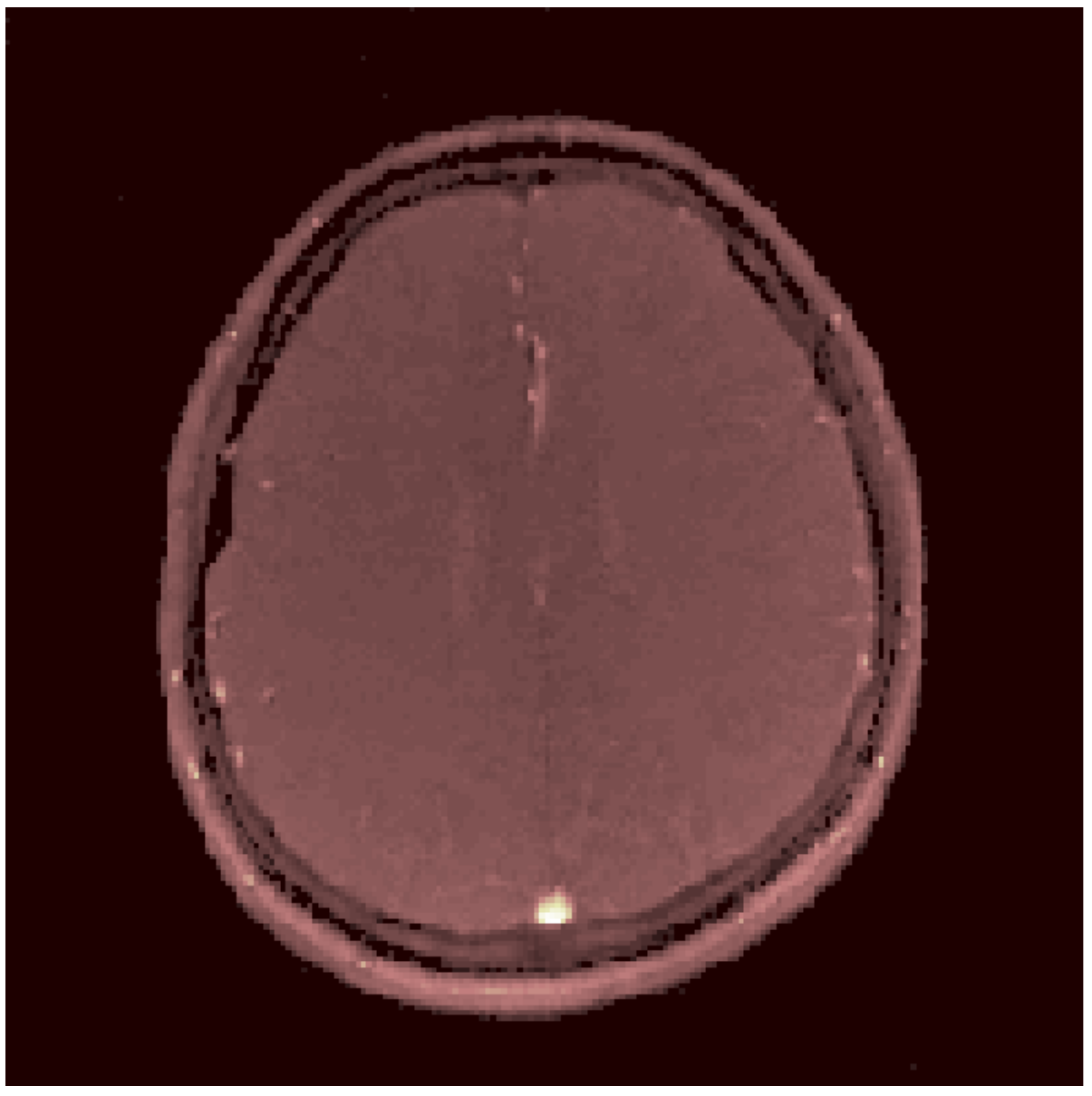}\hspace{-.1cm}
		\includegraphics[width=.162\linewidth]{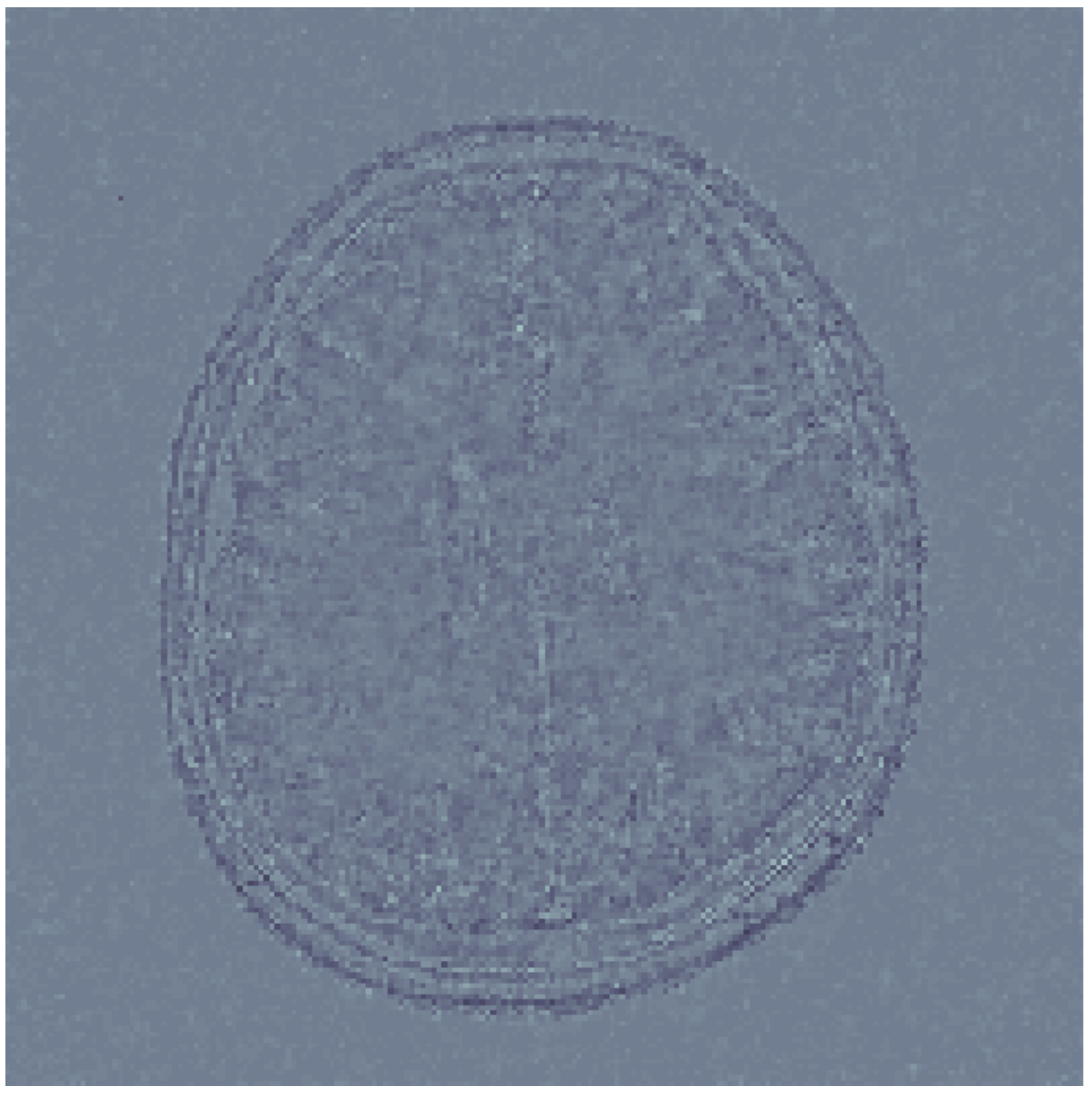}			\\
		\begin{turn}{90} \quad\quad LRTV-KM\end{turn}
		\includegraphics[width=.162\linewidth]{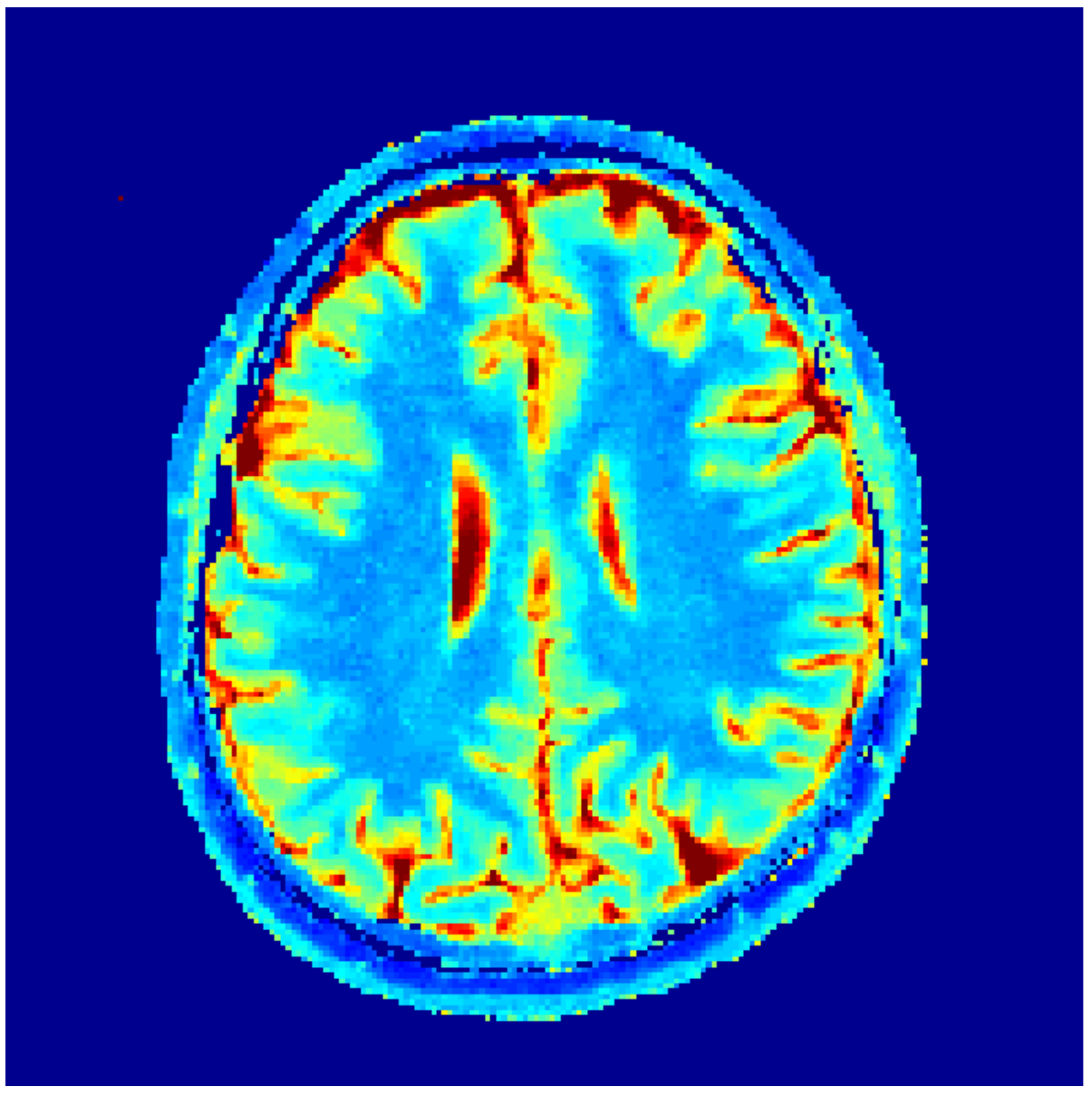}\hspace{-.1cm}
		\includegraphics[width=.162\linewidth]{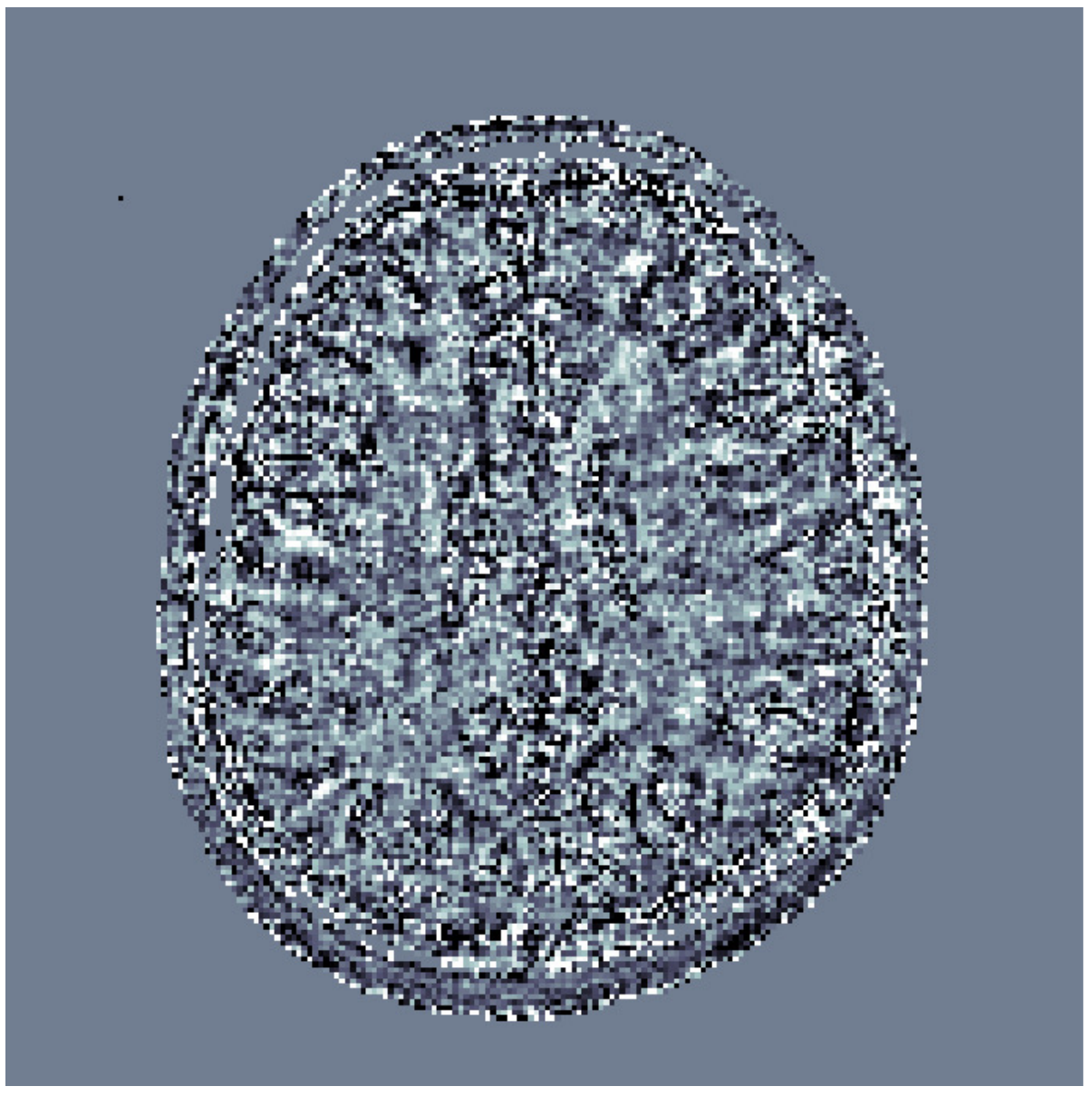}\hspace{.0cm}
		\includegraphics[width=.162\linewidth]{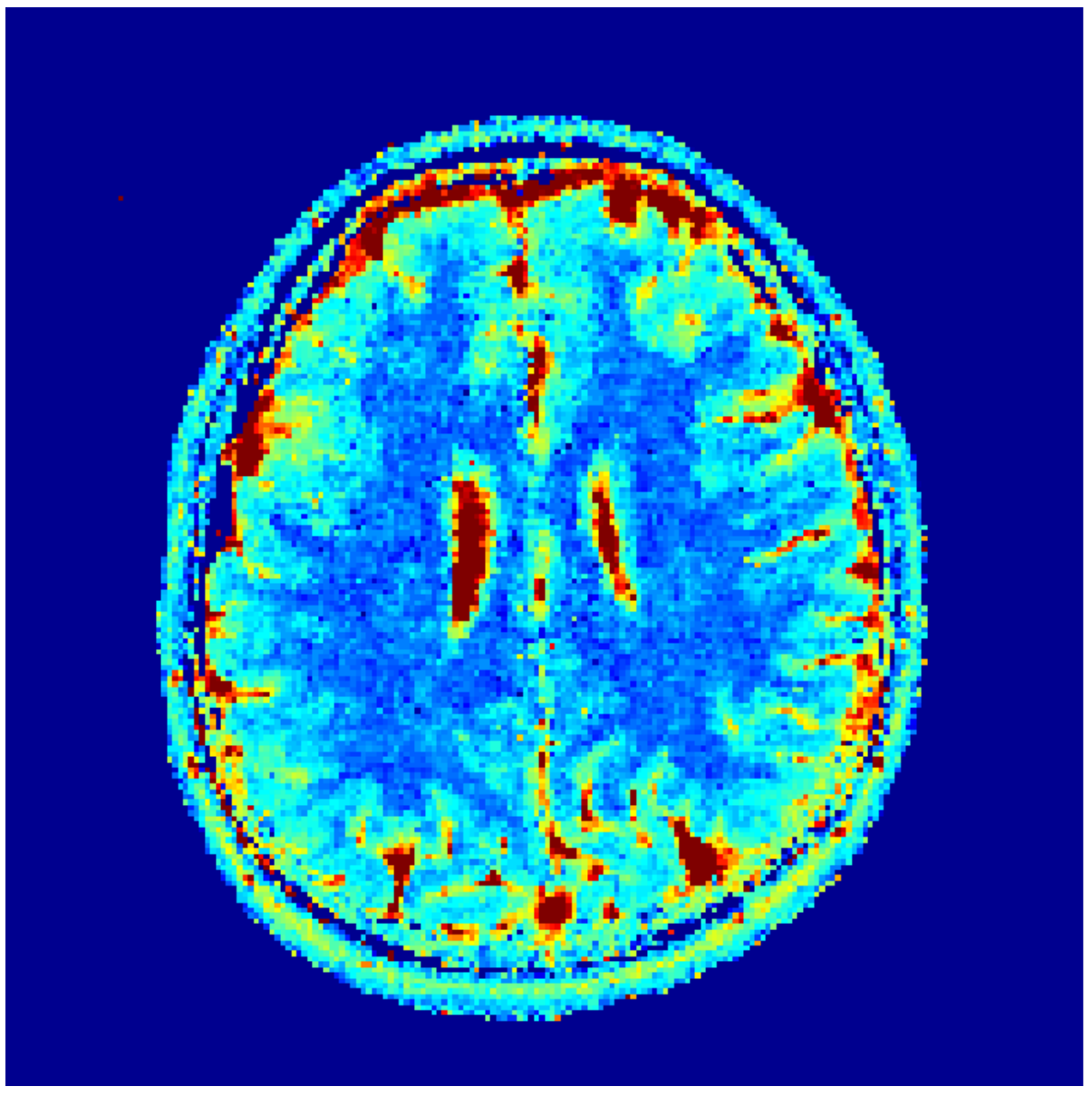}\hspace{-.1cm}
		\includegraphics[width=.162\linewidth]{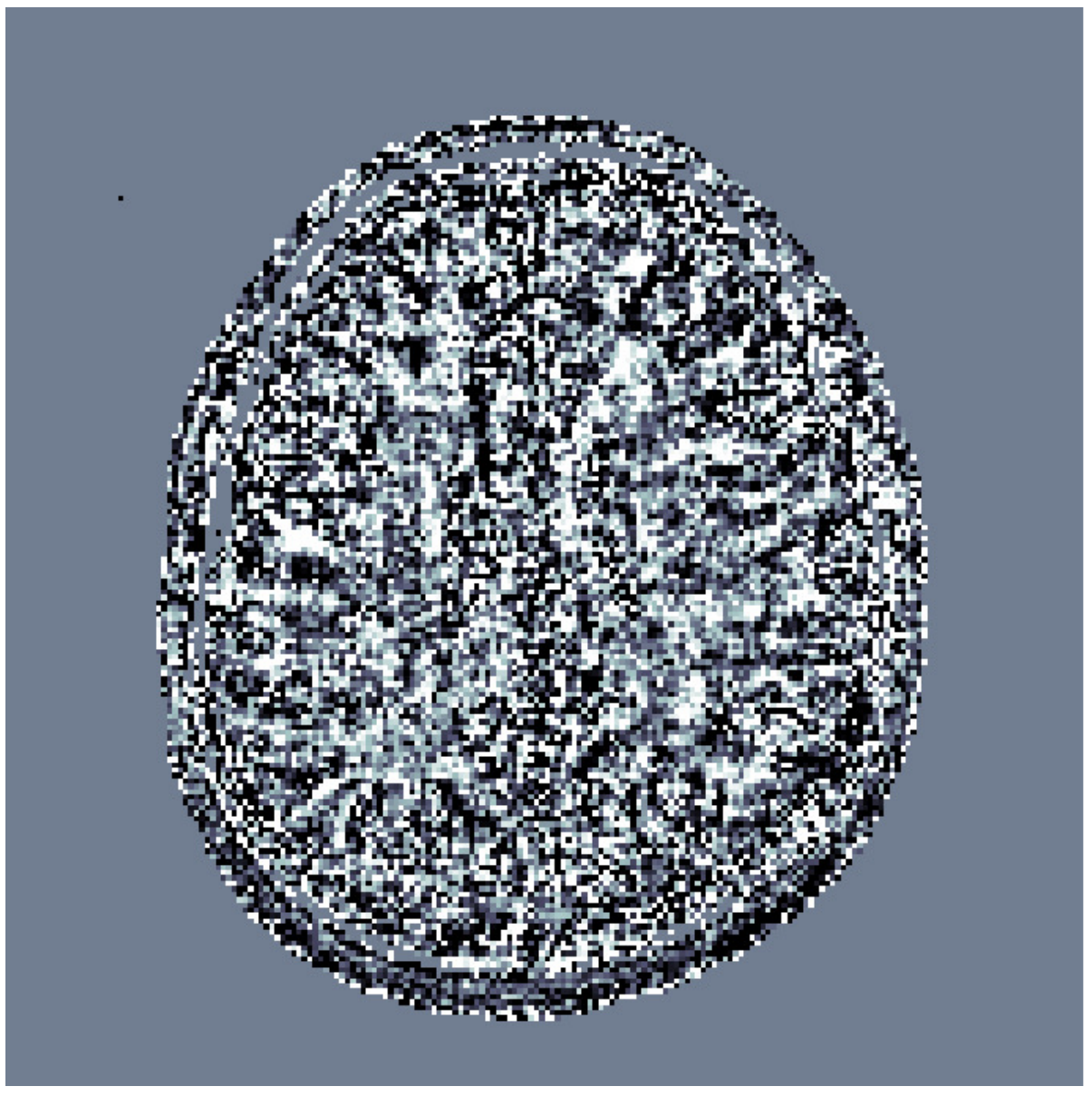}\hspace{.0cm}
		\includegraphics[width=.162\linewidth]{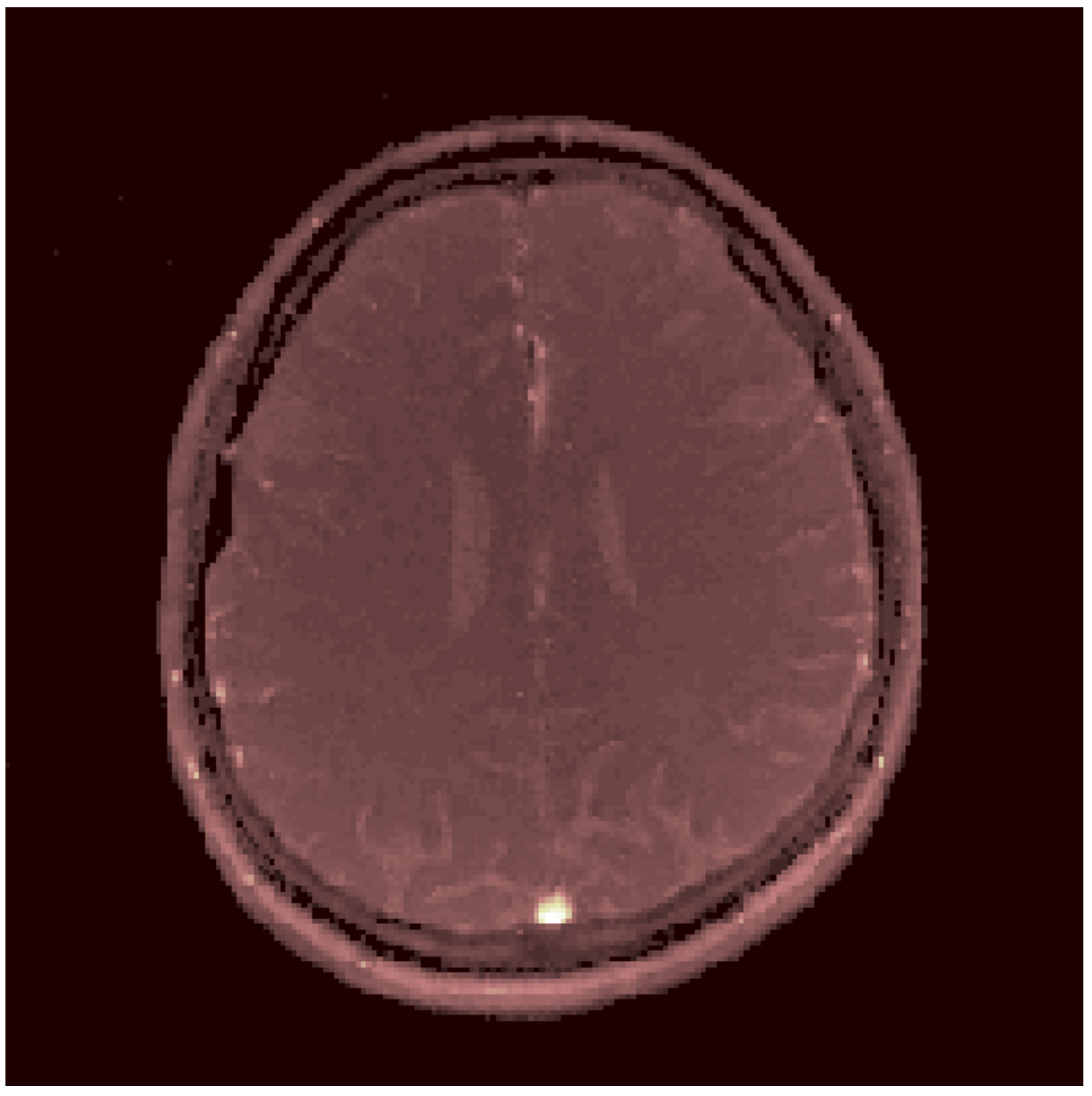}\hspace{-.1cm}
		\includegraphics[width=.162\linewidth]{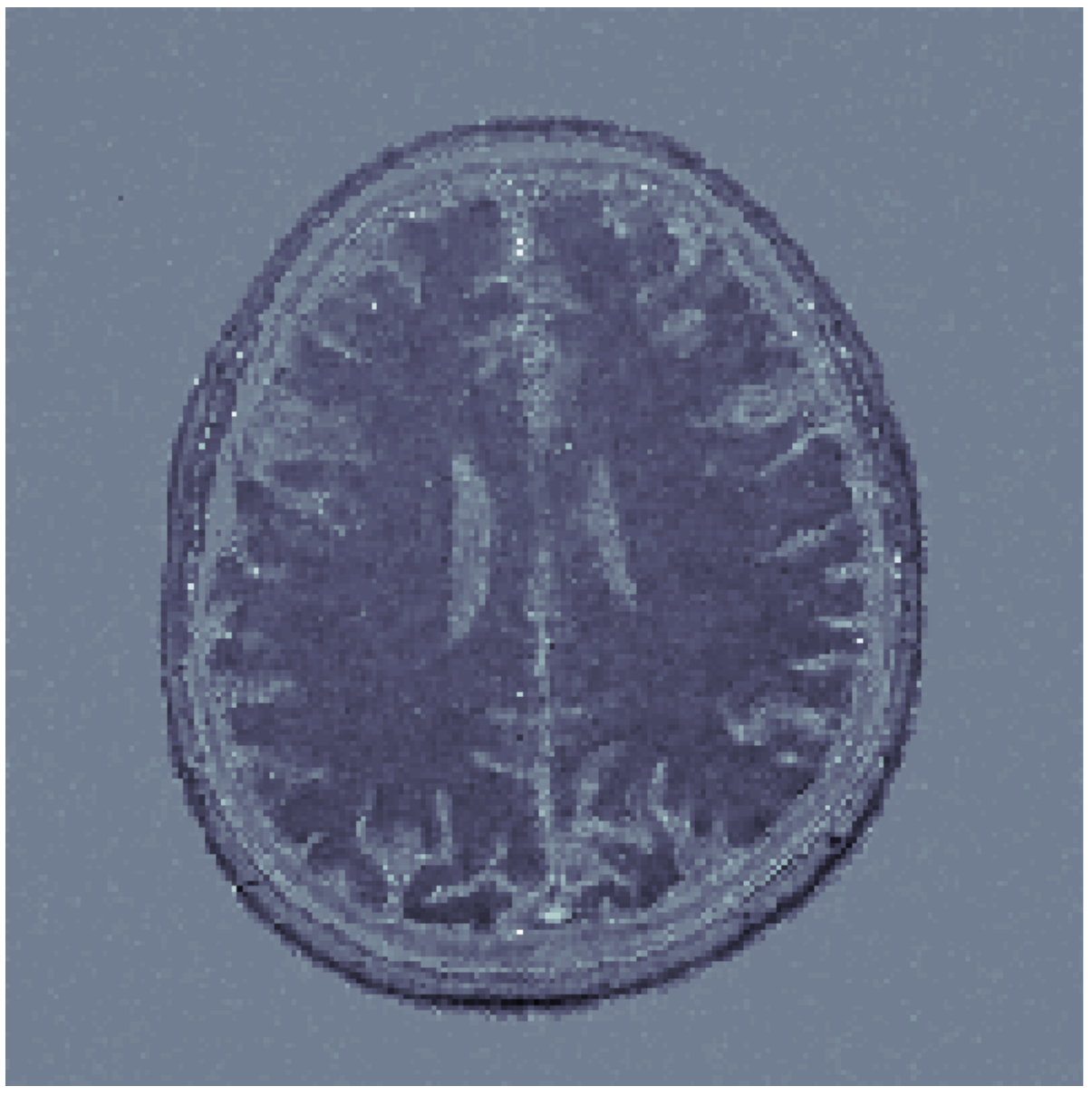}		\\
		\begin{turn}{90}LRTV-MRFResnet\end{turn}
		\includegraphics[width=.162\linewidth]{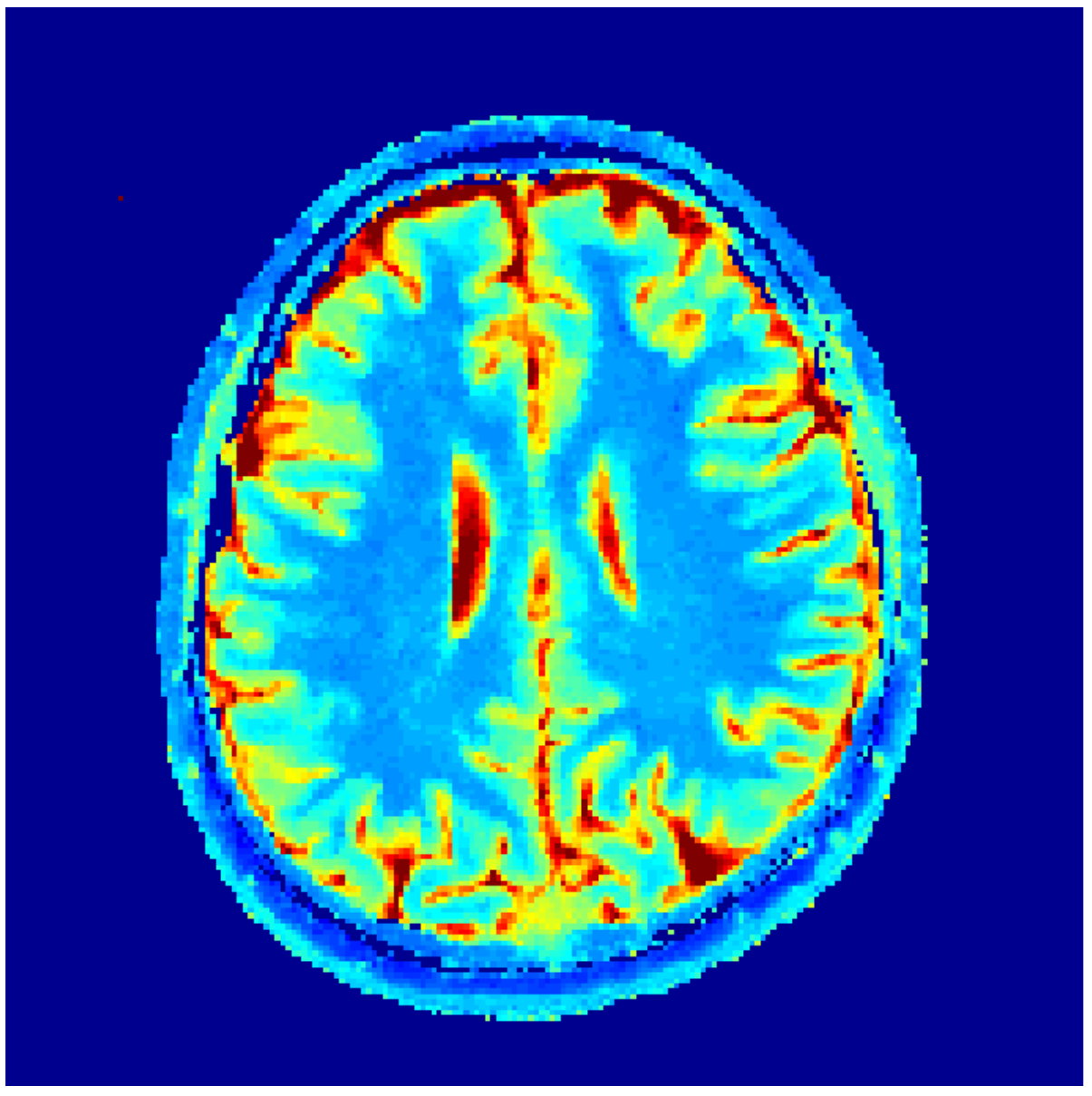}\hspace{-.1cm}
		\includegraphics[width=.162\linewidth]{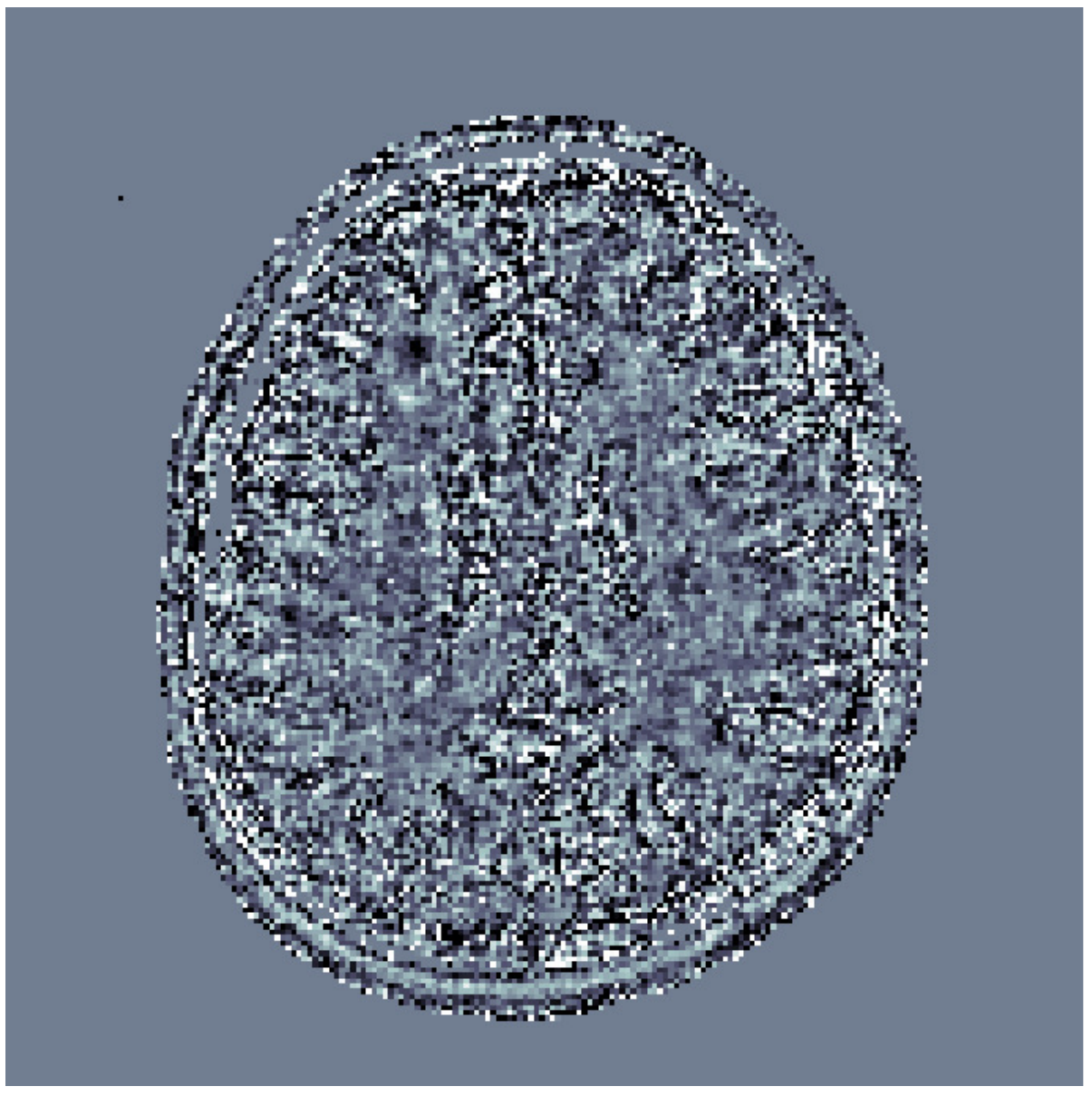}\hspace{.0cm}
		\includegraphics[width=.162\linewidth]{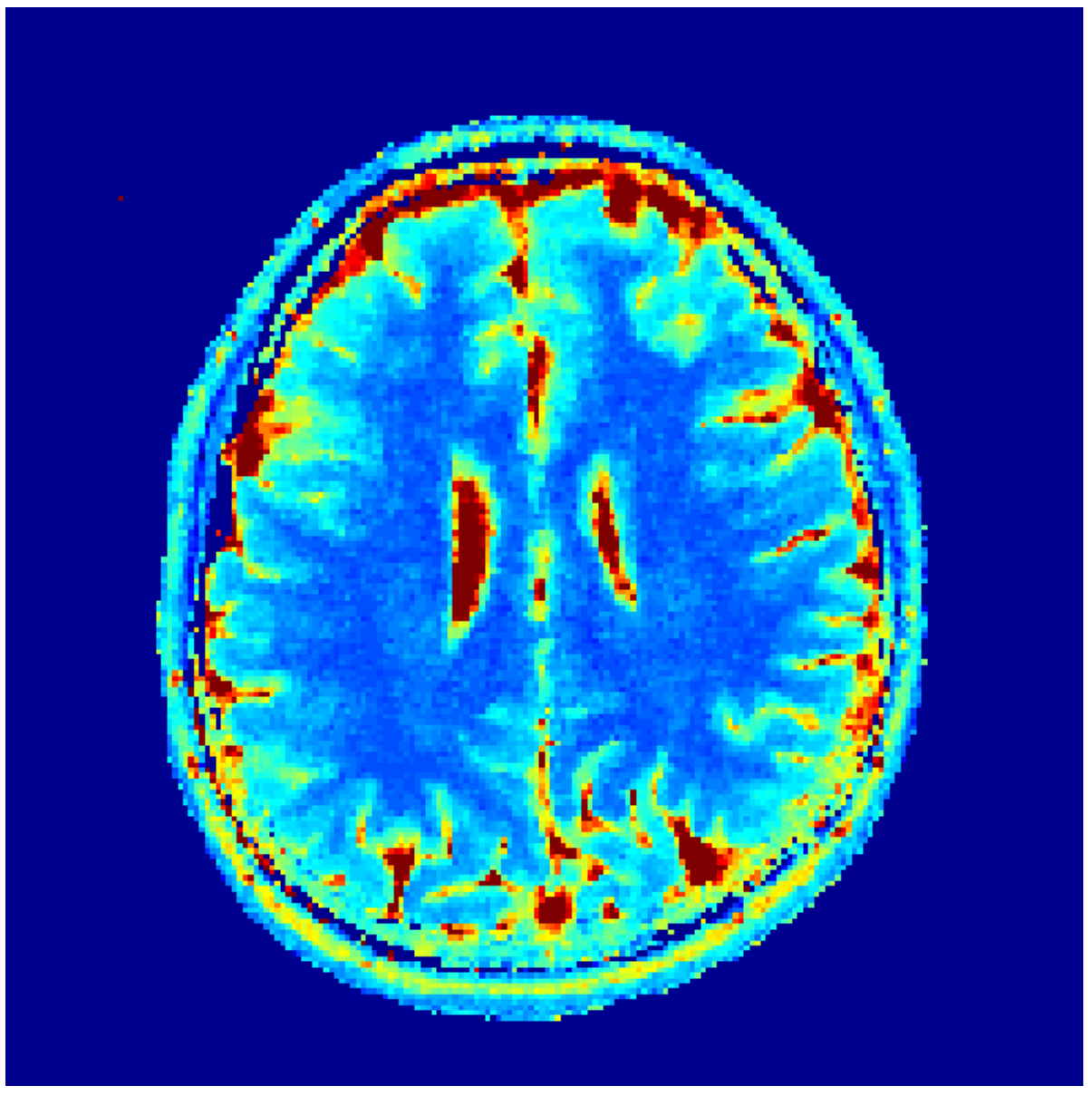}\hspace{-.1cm}
		\includegraphics[width=.162\linewidth]{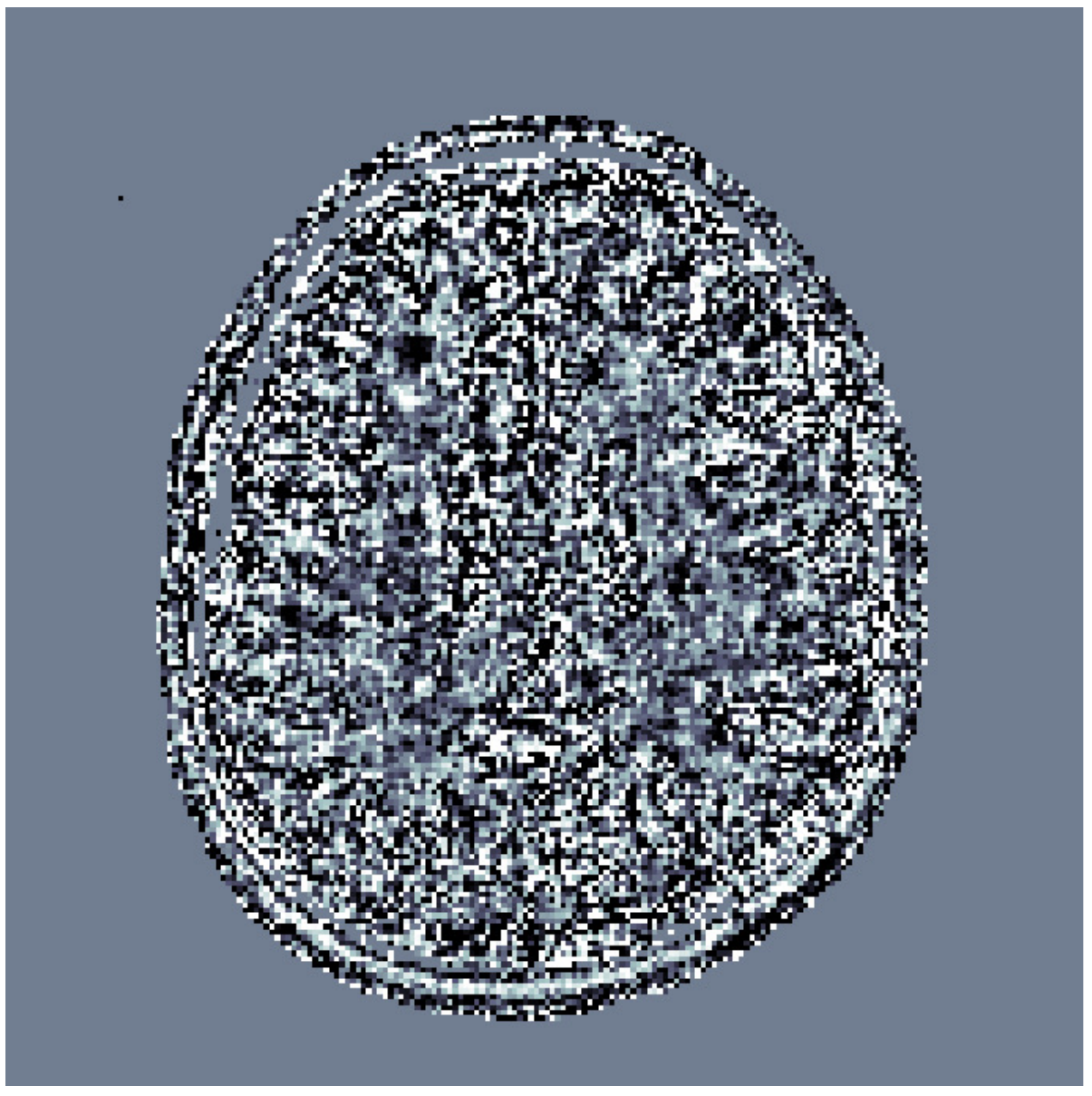}\hspace{.0cm}
		\includegraphics[width=.162\linewidth]{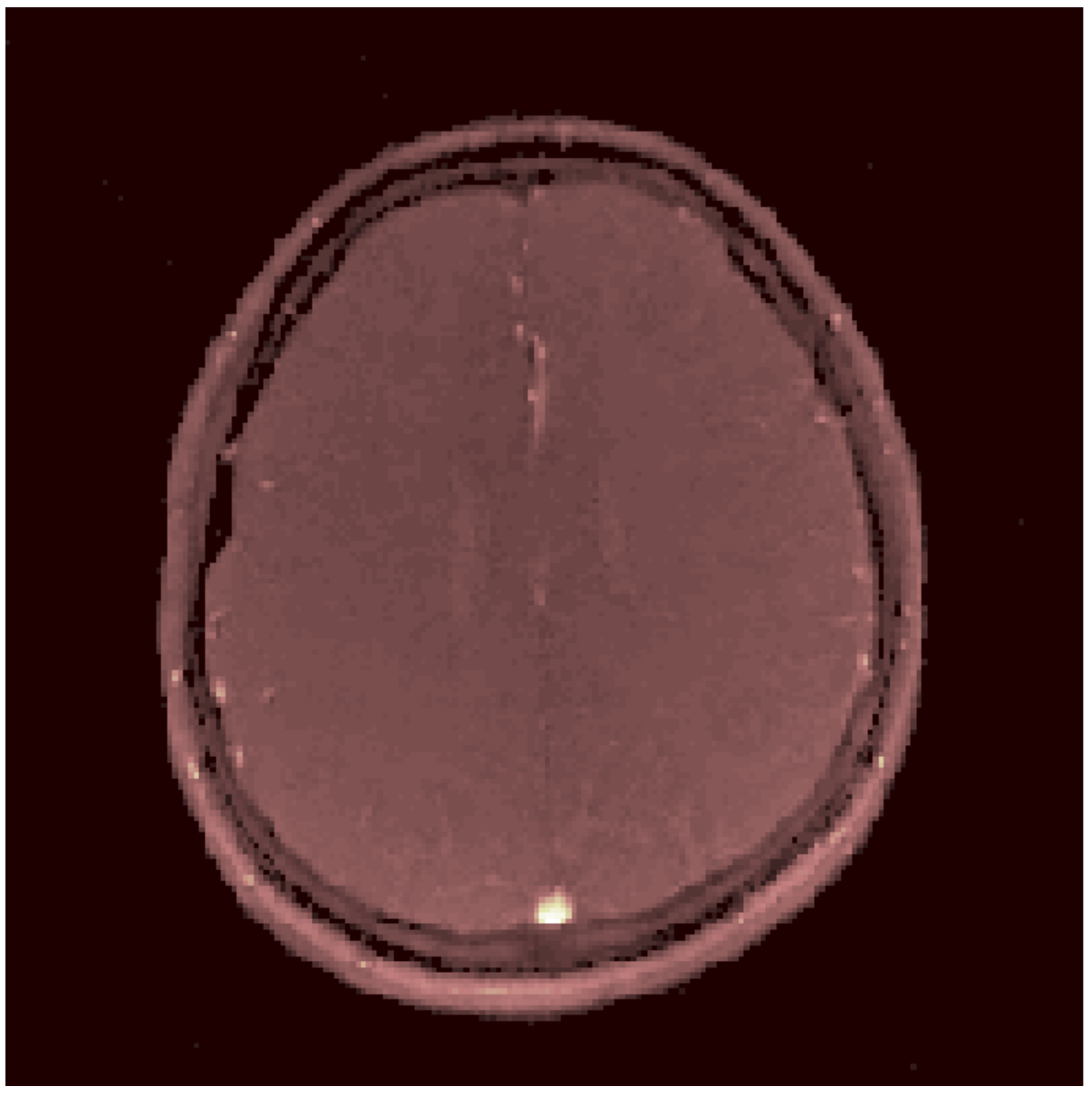}\hspace{-.1cm}
		\includegraphics[width=.162\linewidth]{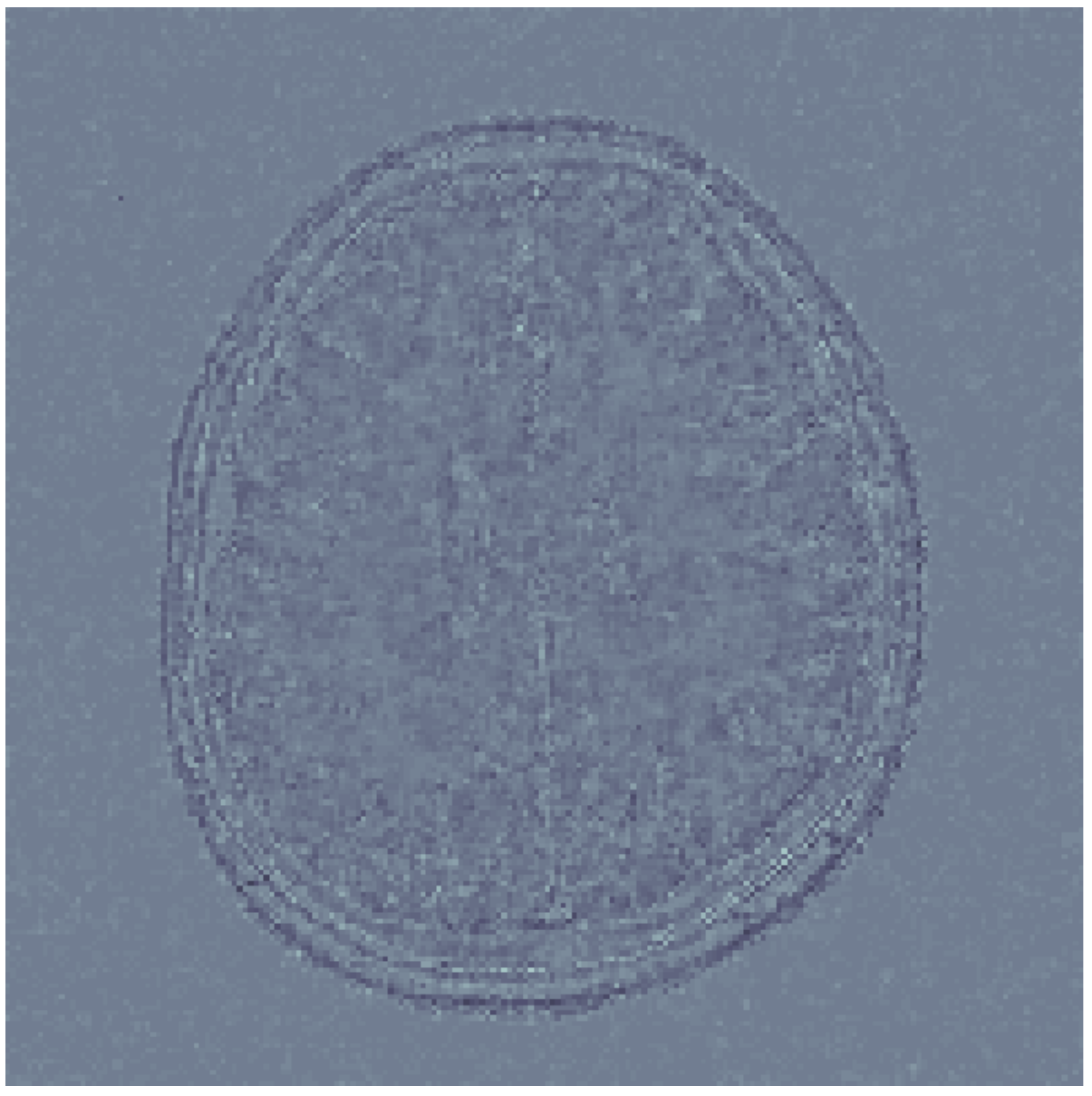}	\vspace{-.3cm}
		\\			
		\includegraphics[trim= -10 50 -20 720, clip,width=.17\linewidth]{./figs/retrospective/T1barvivo.jpg}\hspace{-.2cm}
		\includegraphics[trim= -10 50 -20 700, clip,width=.17\linewidth]{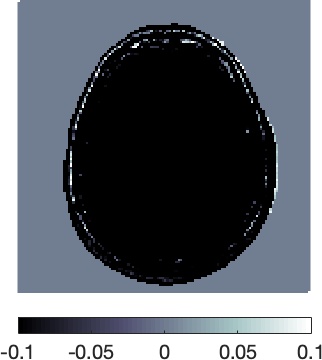}\hspace{-.2cm}
		\includegraphics[trim= -10 50 -20 700, clip,width=.17\linewidth]{./figs/retrospective/T2barvivo.jpg}\hspace{-.2cm}
		\includegraphics[trim= -10 50 -20 700, clip,width=.17\linewidth]{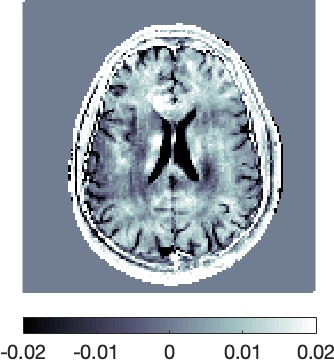}\hspace{-.2cm}
		\includegraphics[trim= -10 50 -20 700, clip,width=.17\linewidth]{./figs/retrospective/PDbarvivo.jpg}\hspace{-.2cm}
		\includegraphics[trim= -10 50 -20 700, clip,width=.17\linewidth]{./figs/retrospective/dPDbarvivo.jpg}\vspace{.5cm}
		\\
		 .\hspace{1cm} T1(sec) \hspace{1.7cm} T1 error  \hspace{1.7cm} T2 (sec) \hspace{1.7cm} T2 error \hspace{1.7cm} PD (a.u.) \hspace{1.7cm} PD error 
		\caption{The computed T1, T2, PD maps and their corresponding errors (with respect to MAGIC gold-standard) using 2D spiral k-space sampling, different reconstruction baselines and our proposed LRTV-MRFResnet algorithm. 
  \label{fig:spiral_recon_retro}}
\end{minipage}
}
\end{figure*}

The gold standard anatomical maps were acquired from a volunteer using the MAGIC quantisation protocol~\cite{magic2015}, using the 1.5T GE HDxT scanner with 8-channel receive-only head RF coil. Figure~\ref{fig:FISPGT} shows these ground truth parameter maps. From these parametric maps we constructed the corresponding TSMIs and MRF measurements. 
A single-coil acquisition with eight times less measurements were considered i.e. $S(\bar X)=\bar X$ identity sensitivity map. We  used the same excitation sequence and the 2D spiral and radial k-space sampling patterns as in our real-world scans (Section~VI). 
The k-space measurements were corrupted by additive i.i.d. Gaussian noise with 35 dB SNR. Figures~\ref{fig:spiral_recon_retro} and \ref{fig:spiral_recon_retro} show the reconstructed maps and their errors with respect to the ground-truth using various baselines and our proposed LRTV-MRFResnet algorithm.

\begin{figure*}[ht!]
	\centering
	\scalebox{.9}{
	\begin{minipage}{\linewidth}
	\begin{turn}{90} \quad\qquad ZF-DM \end{turn}
		\includegraphics[width=.162\linewidth]{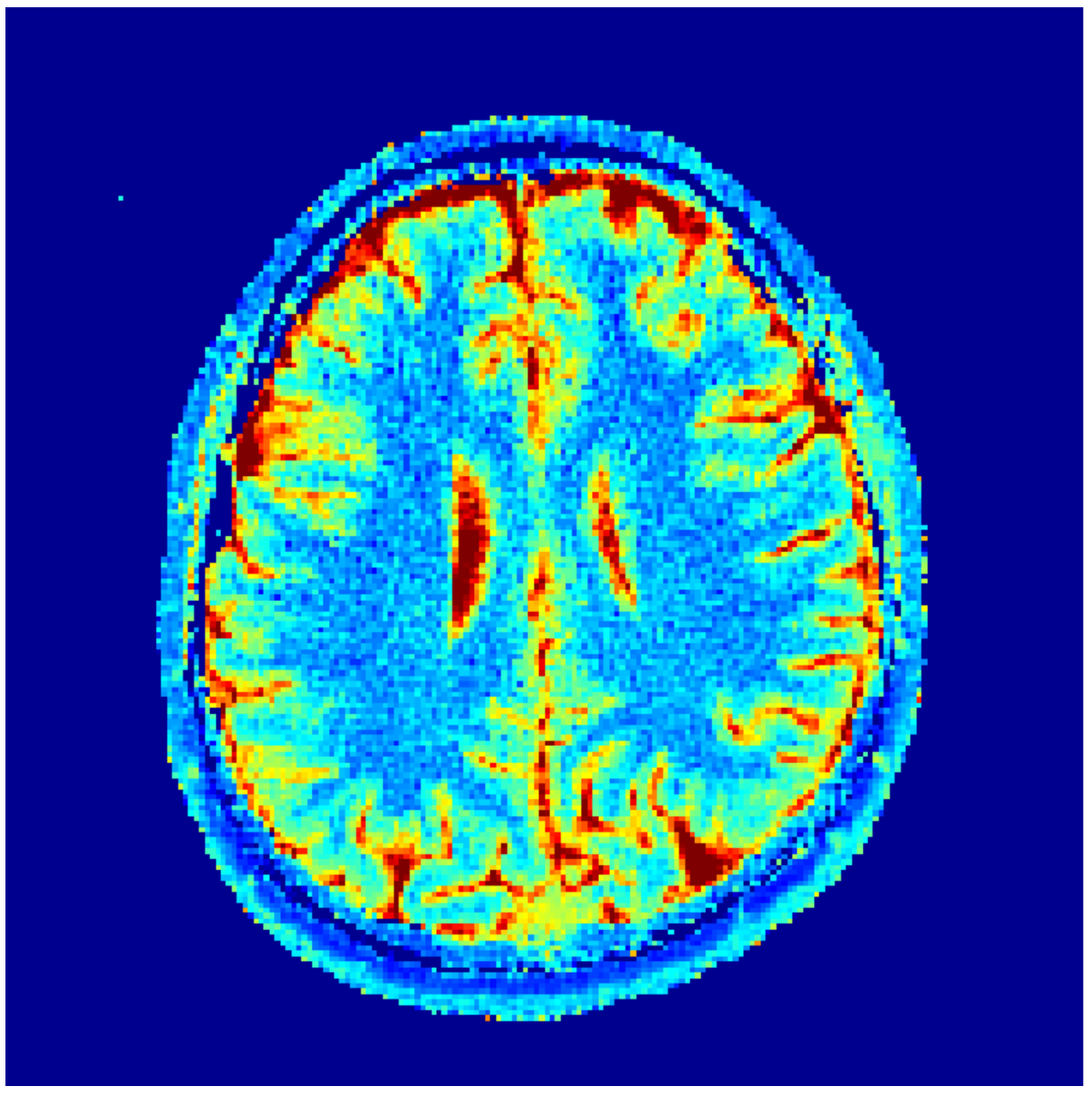}\hspace{-.1cm}
		\includegraphics[width=.162\linewidth]{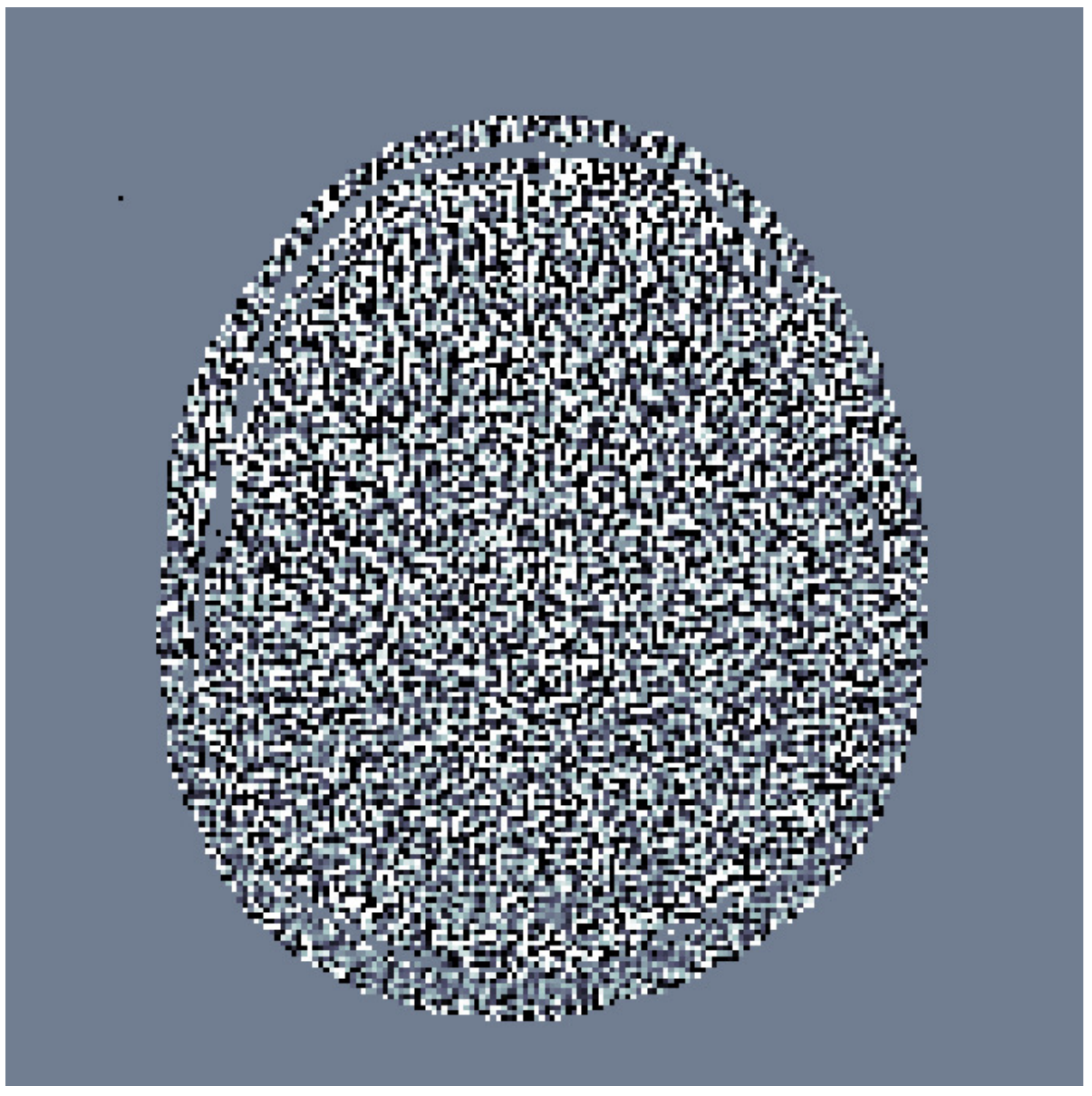}\hspace{.0cm}
		\includegraphics[width=.162\linewidth]{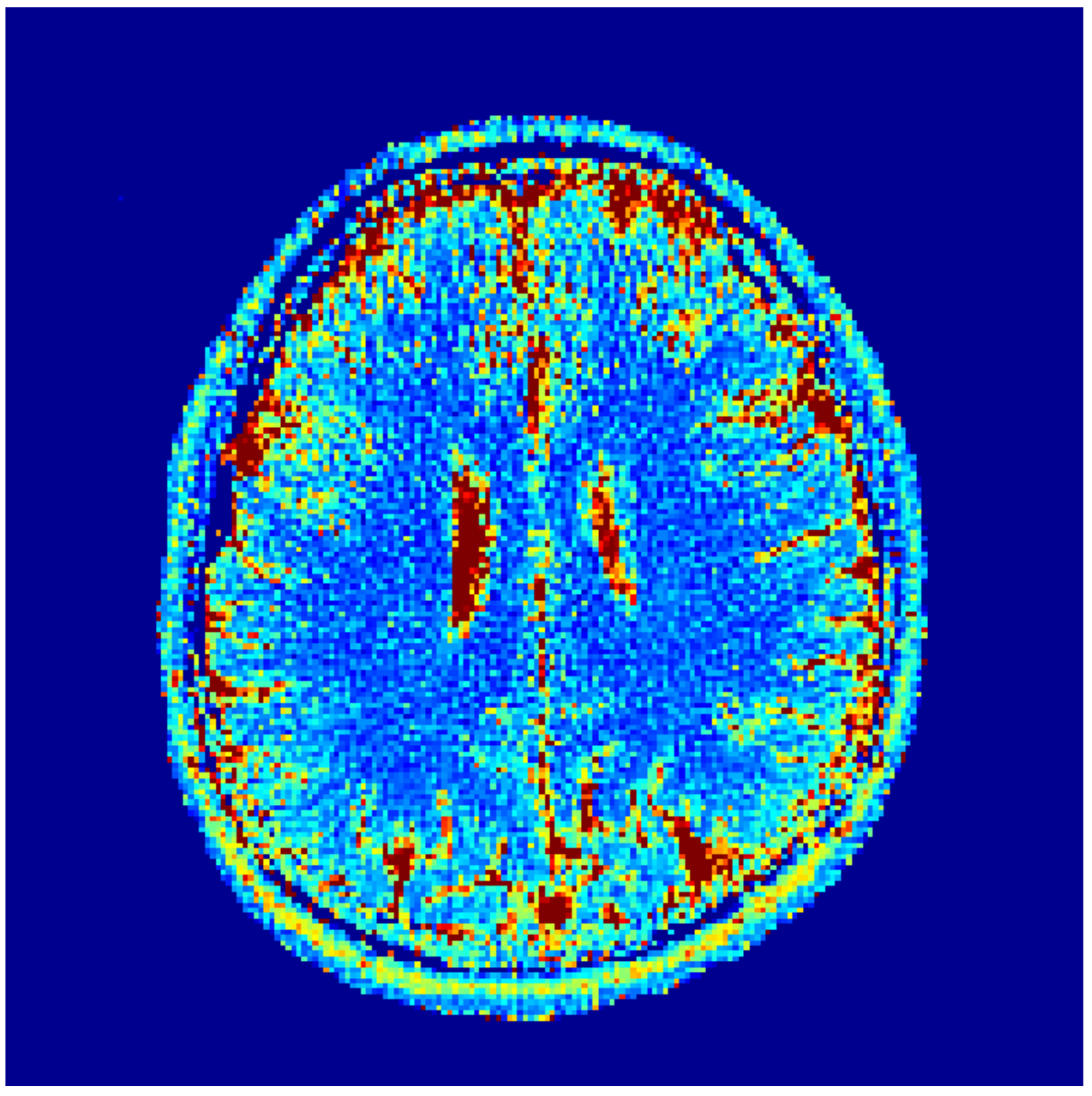}\hspace{-.1cm}
		\includegraphics[width=.162\linewidth]{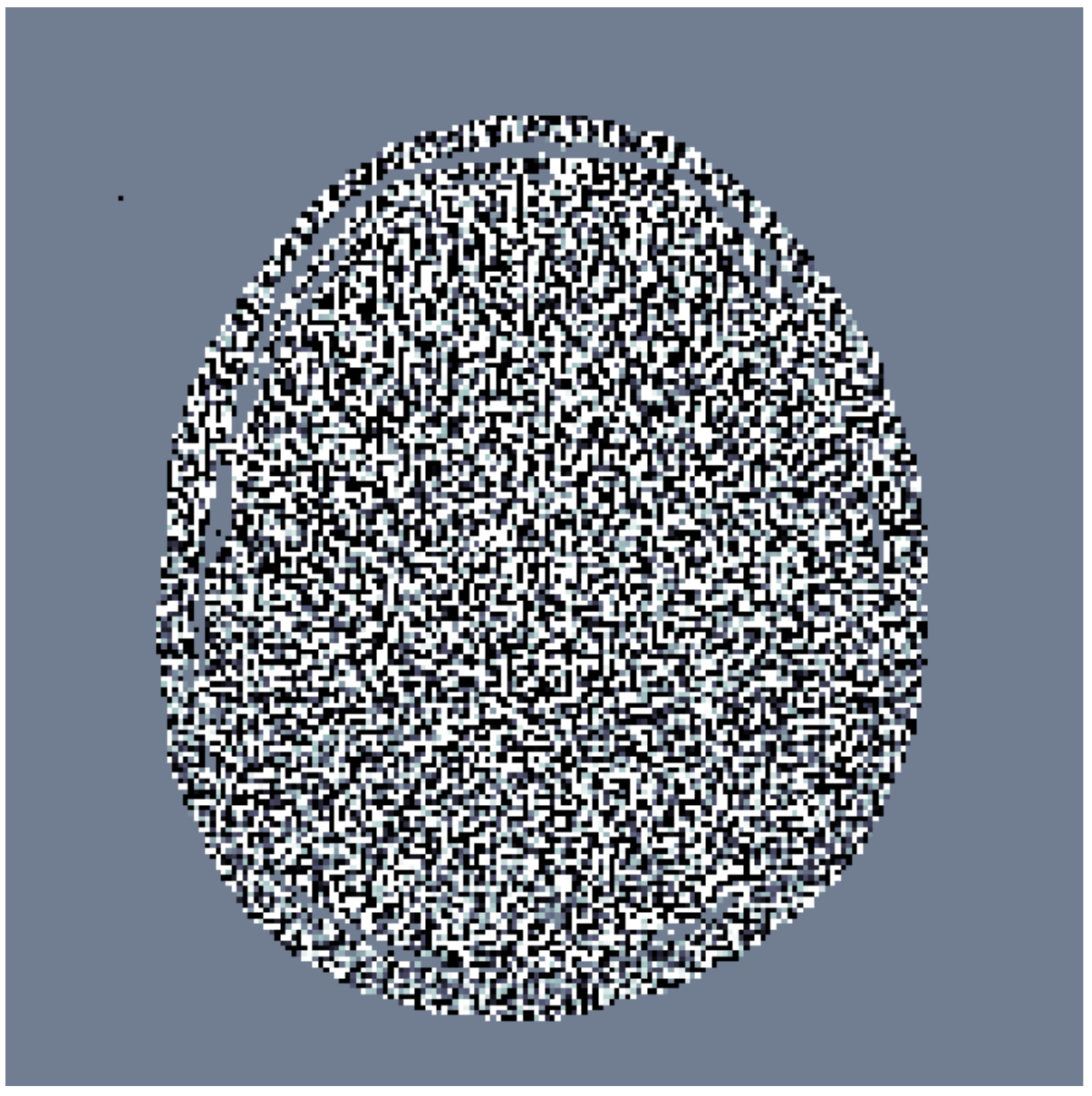}\hspace{.0cm}
		\includegraphics[width=.162\linewidth]{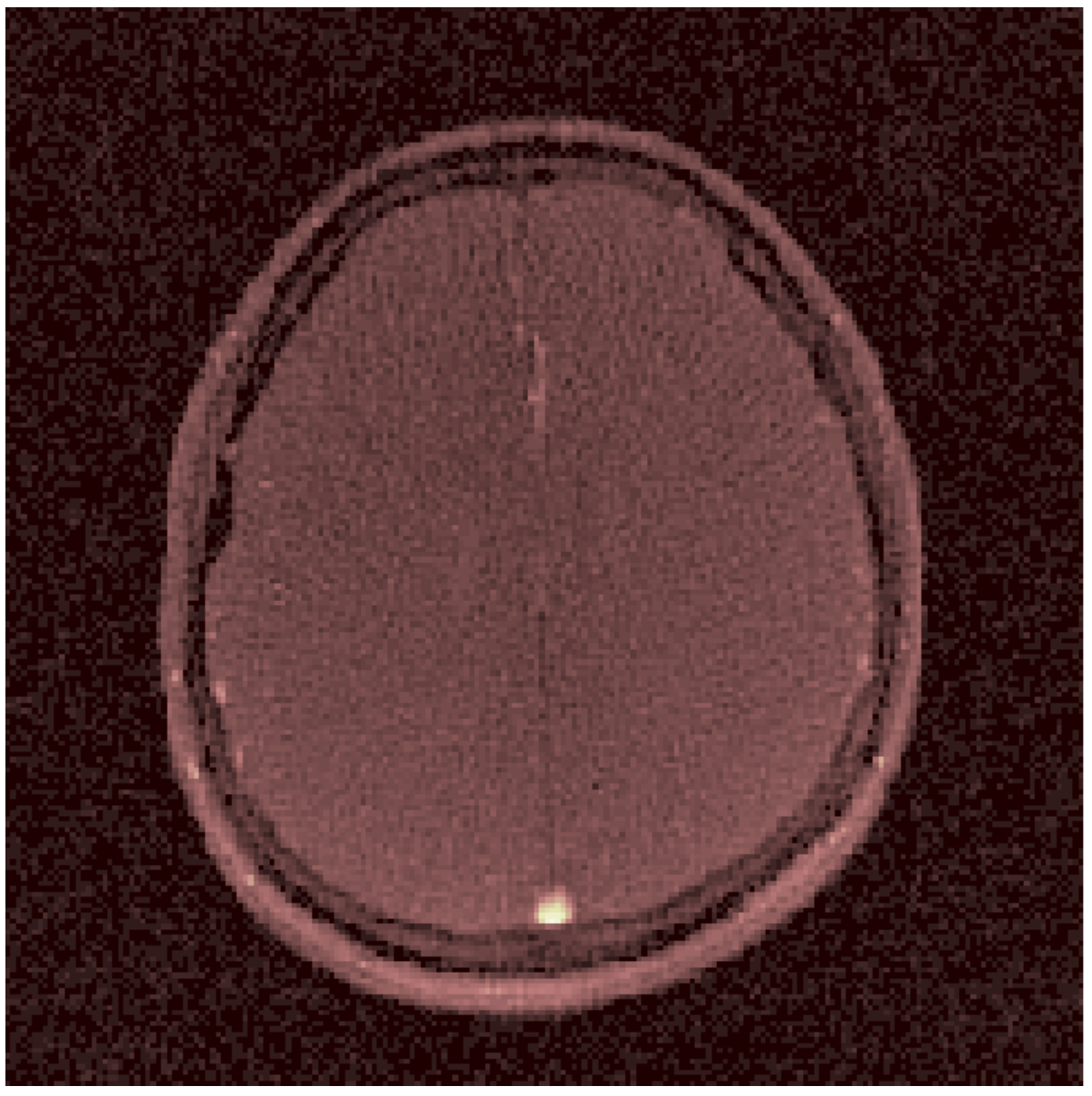}\hspace{-.1cm}
		\includegraphics[width=.162\linewidth]{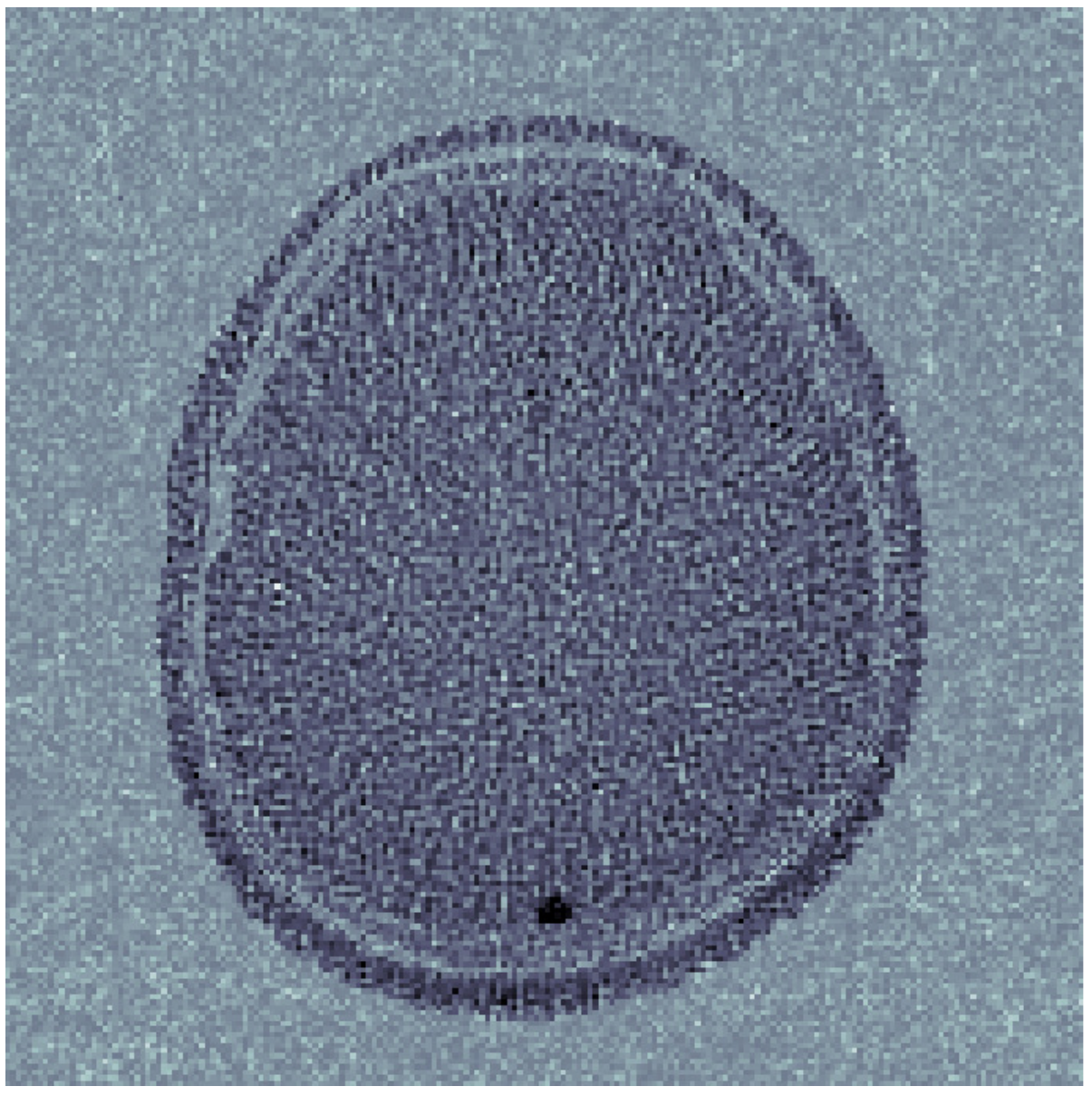}		
		\\
		\begin{turn}{90} \quad\qquad LR-DM \end{turn}
		\includegraphics[width=.162\linewidth]{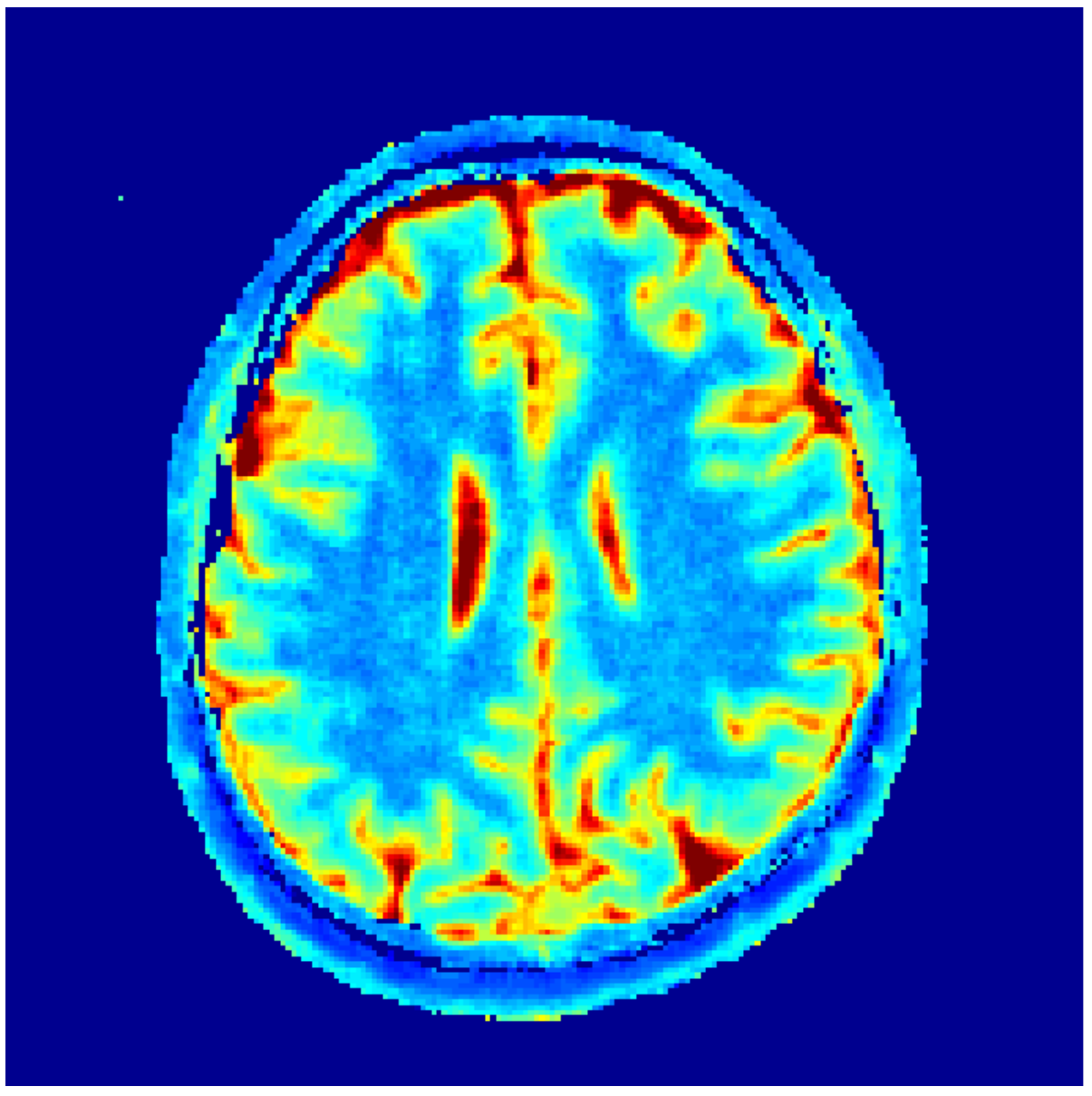}\hspace{-.1cm}
		\includegraphics[width=.162\linewidth]{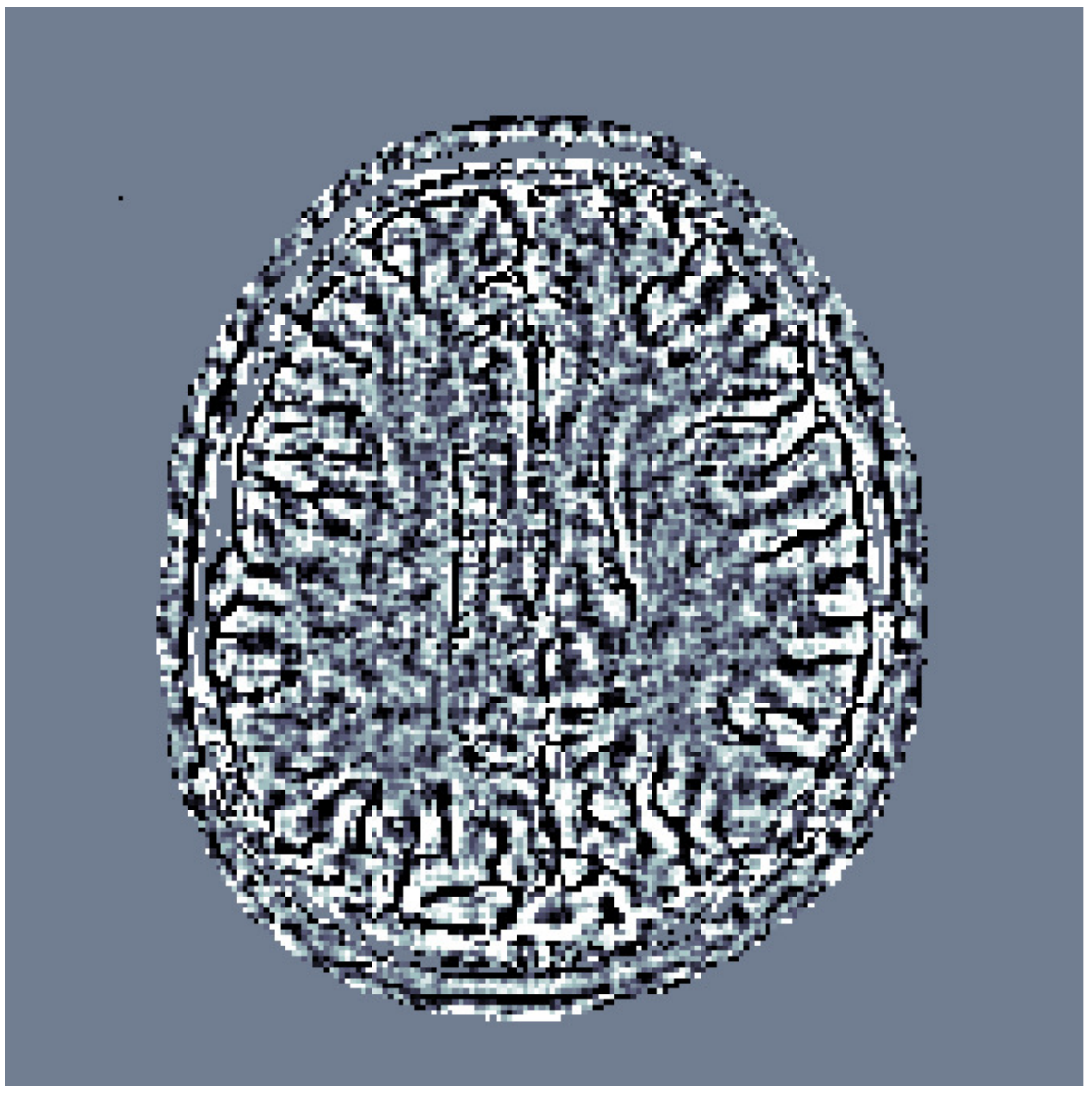}\hspace{.0cm}
		\includegraphics[width=.162\linewidth]{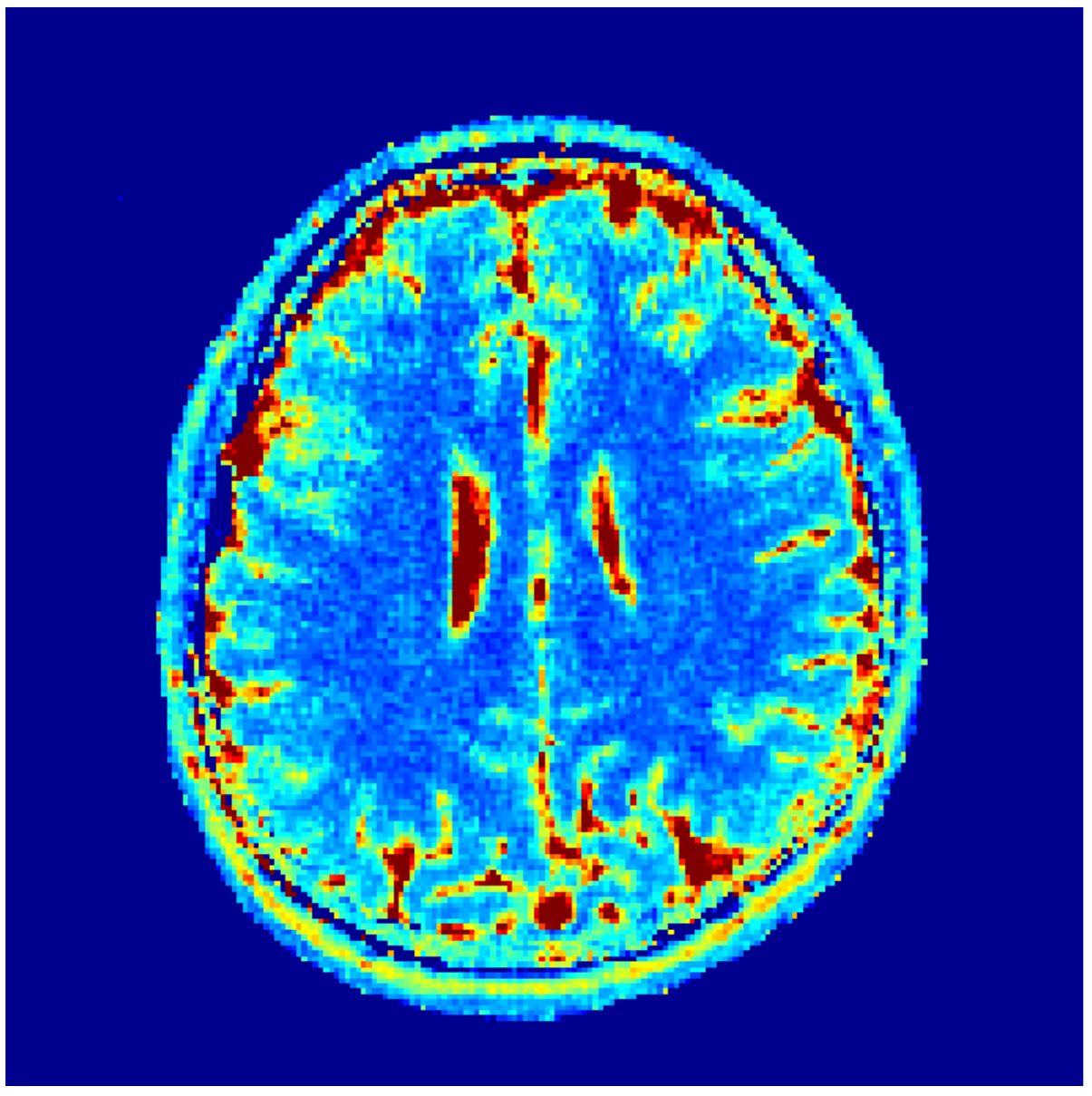}\hspace{-.1cm}
		\includegraphics[width=.162\linewidth]{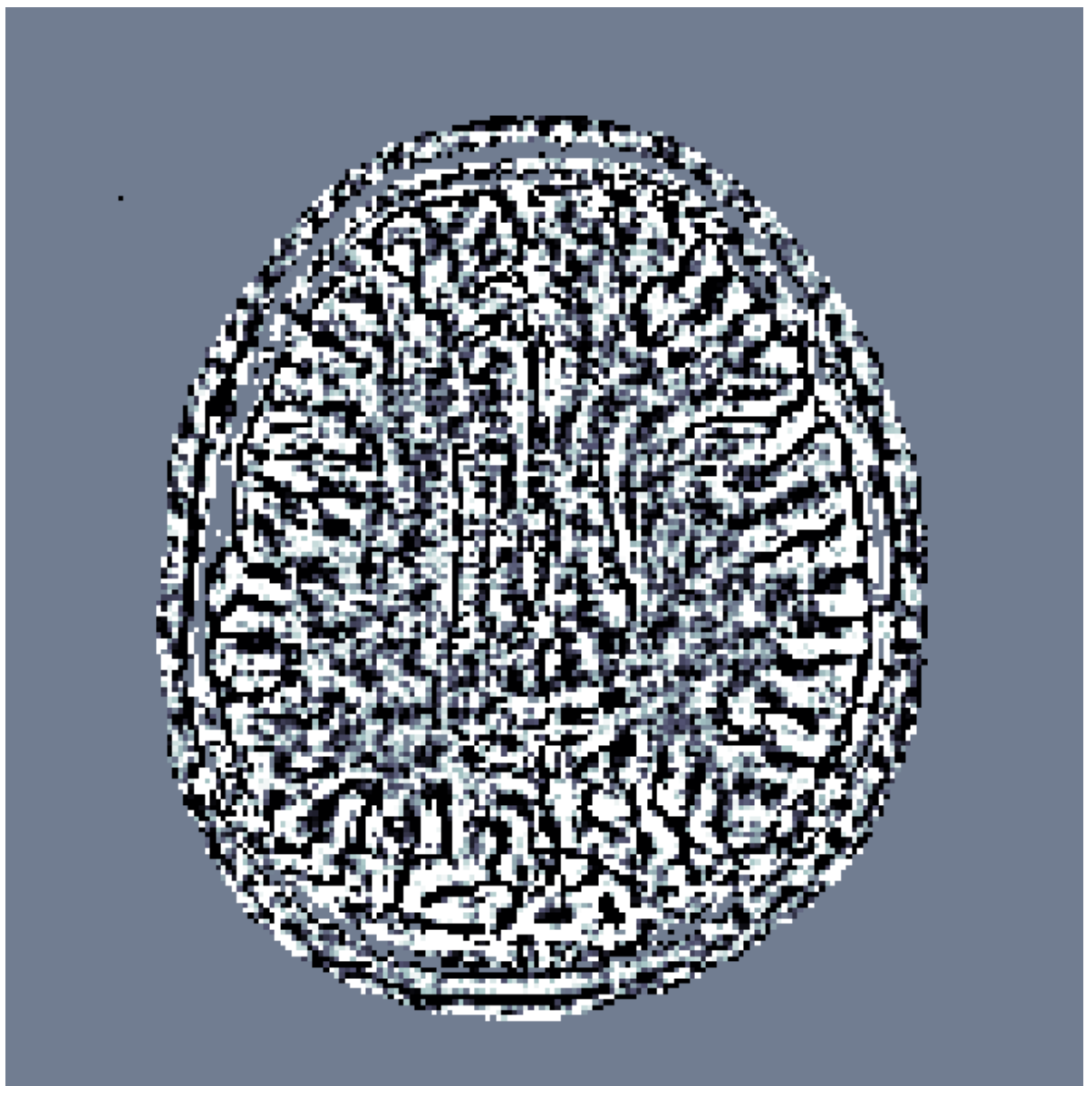}\hspace{.0cm}
		\includegraphics[width=.162\linewidth]{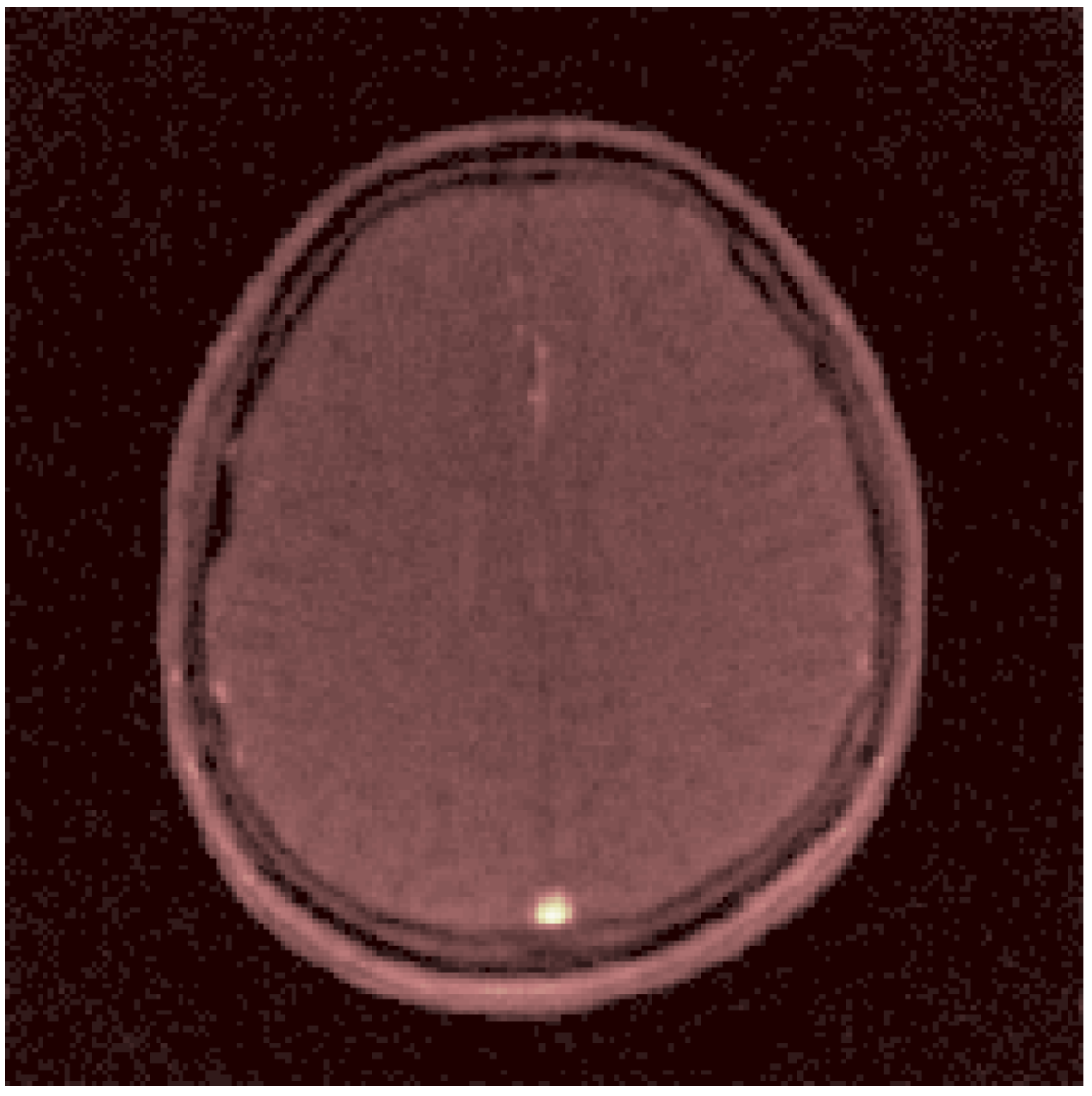}\hspace{-.1cm}
		\includegraphics[width=.162\linewidth]{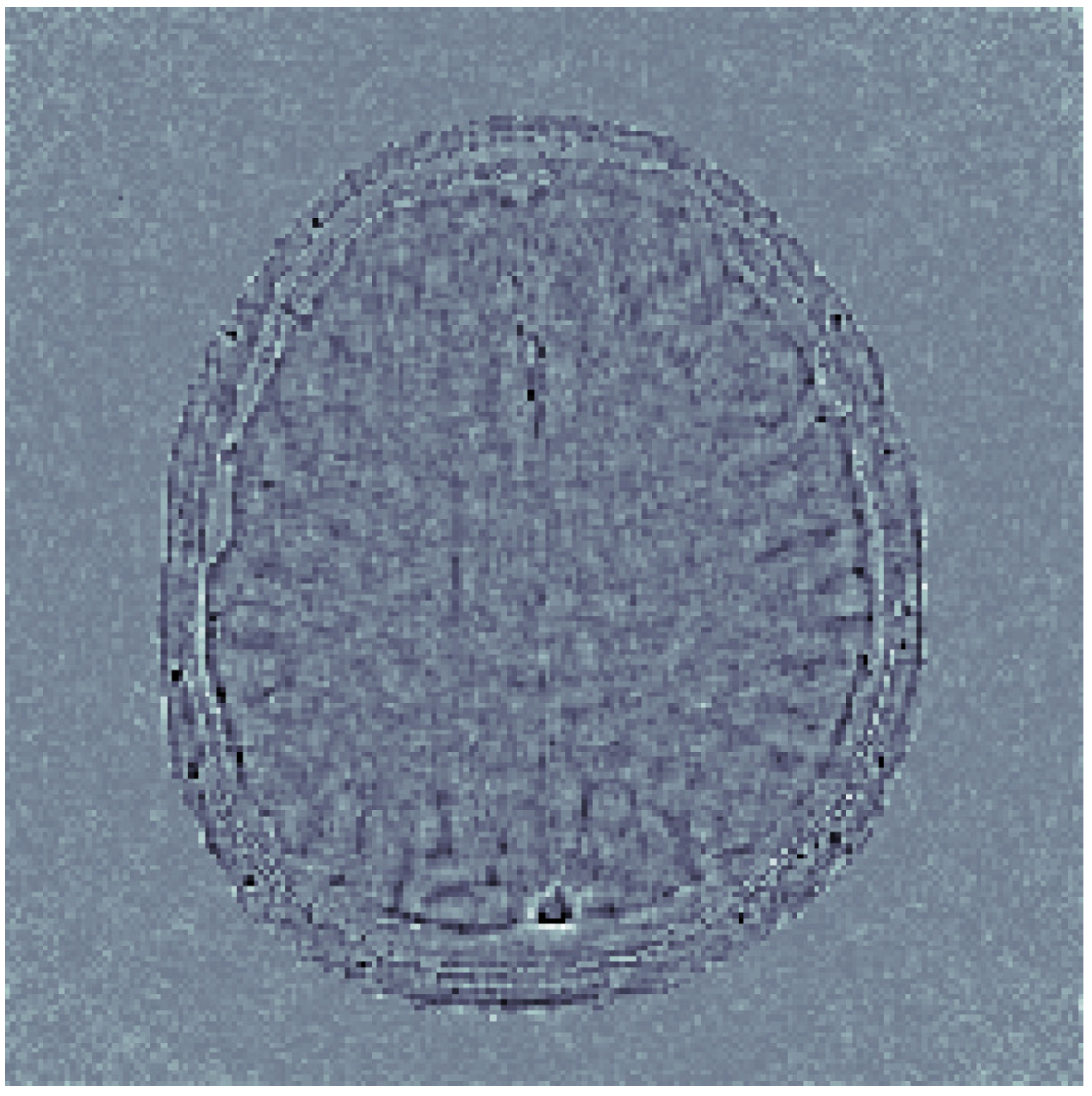}			
		\\
		\begin{turn}{90} \quad\qquad VS-DM \end{turn}
		\includegraphics[width=.162\linewidth]{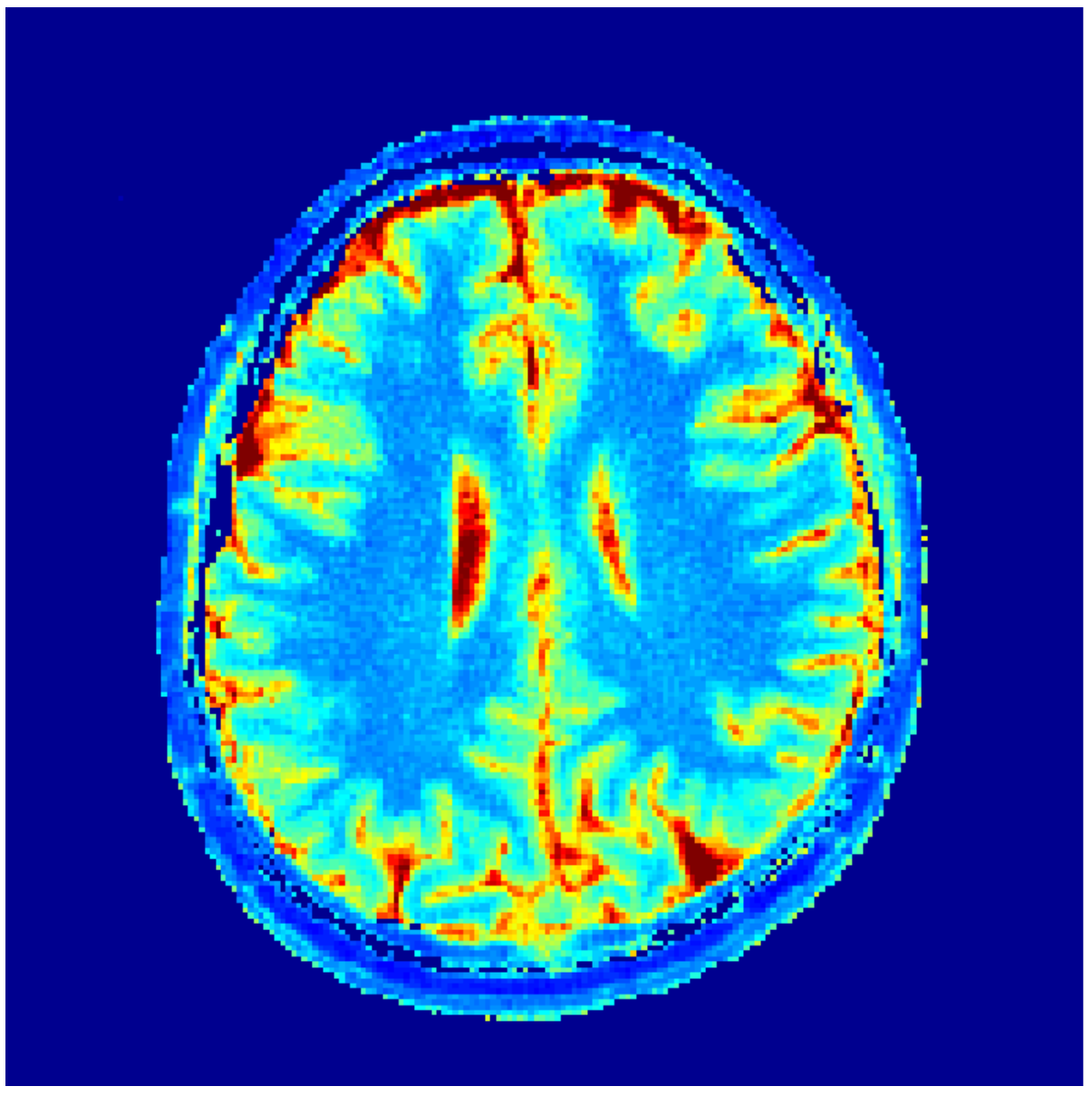}\hspace{-.1cm}
		\includegraphics[width=.162\linewidth]{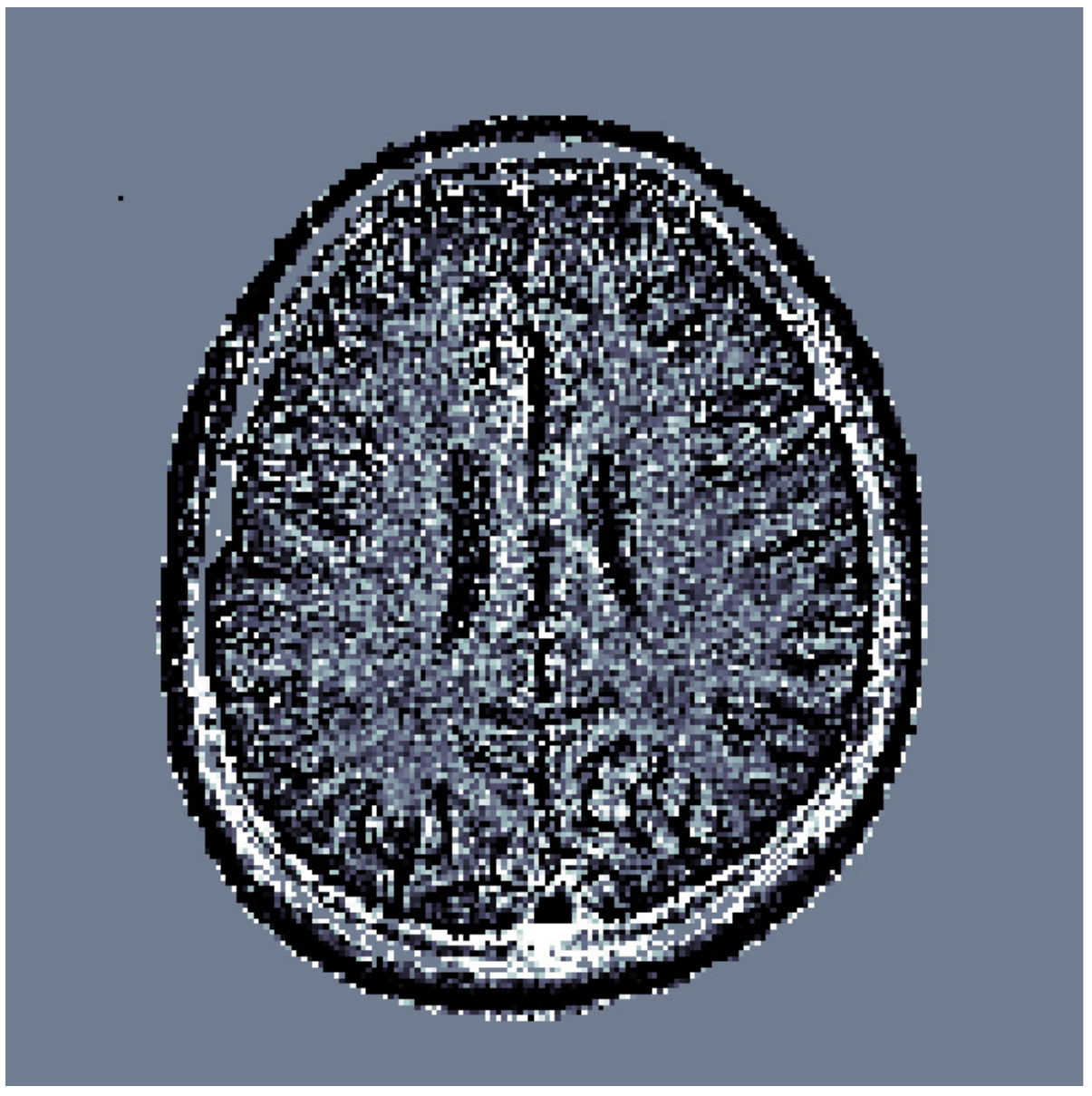}\hspace{.0cm}
		\includegraphics[width=.162\linewidth]{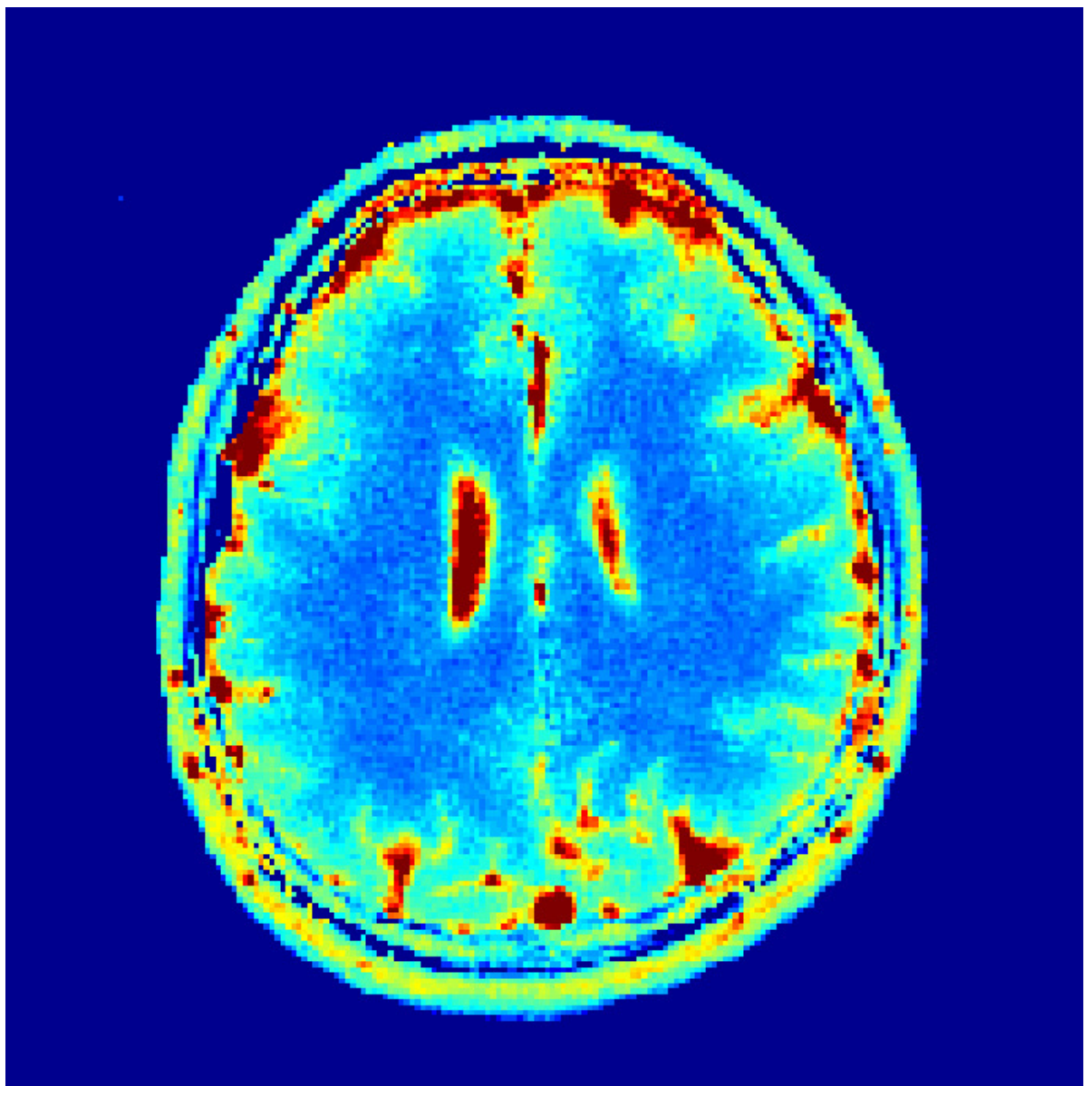}\hspace{-.1cm}
		\includegraphics[width=.162\linewidth]{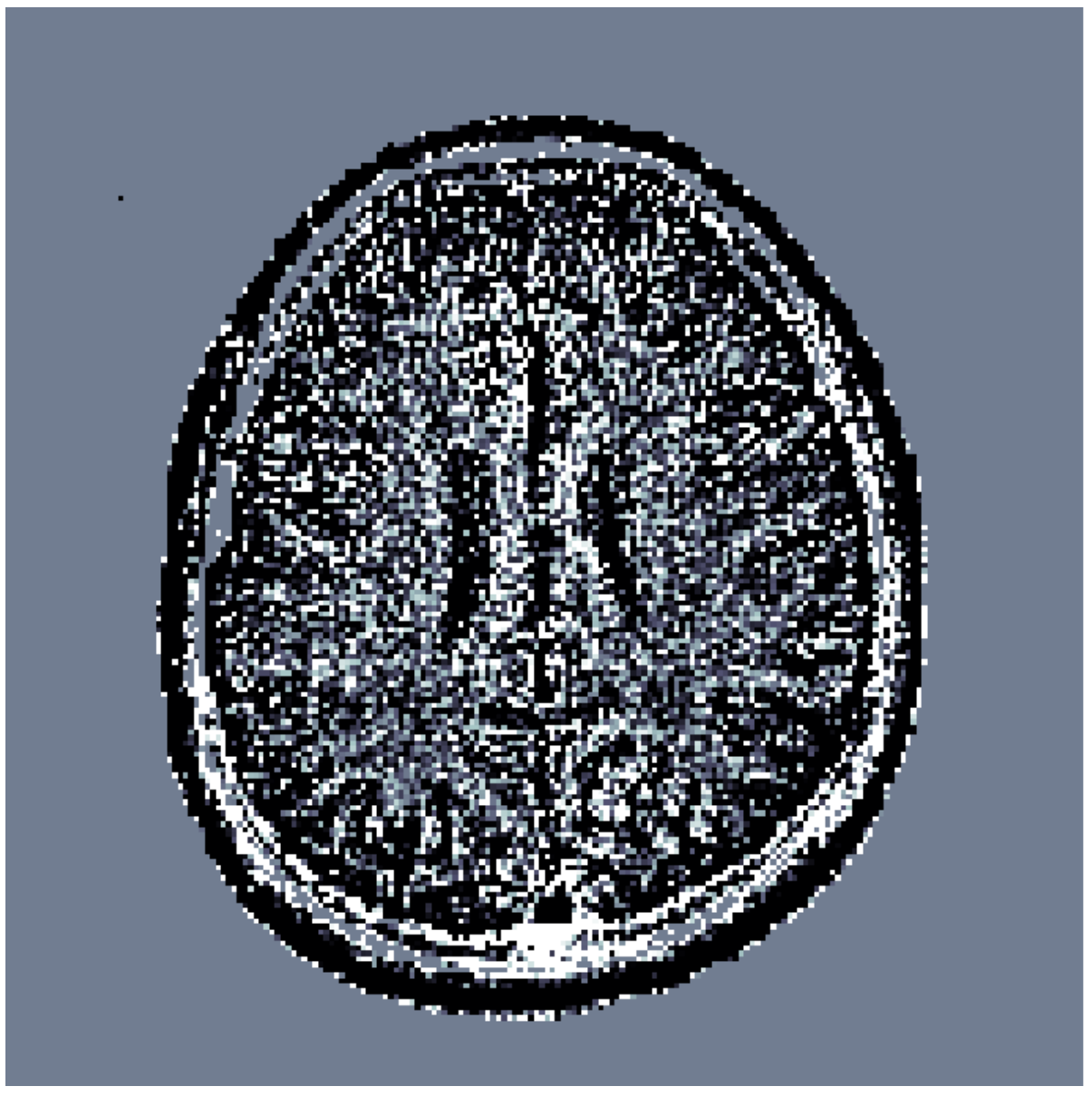}\hspace{.0cm}
		\includegraphics[width=.162\linewidth]{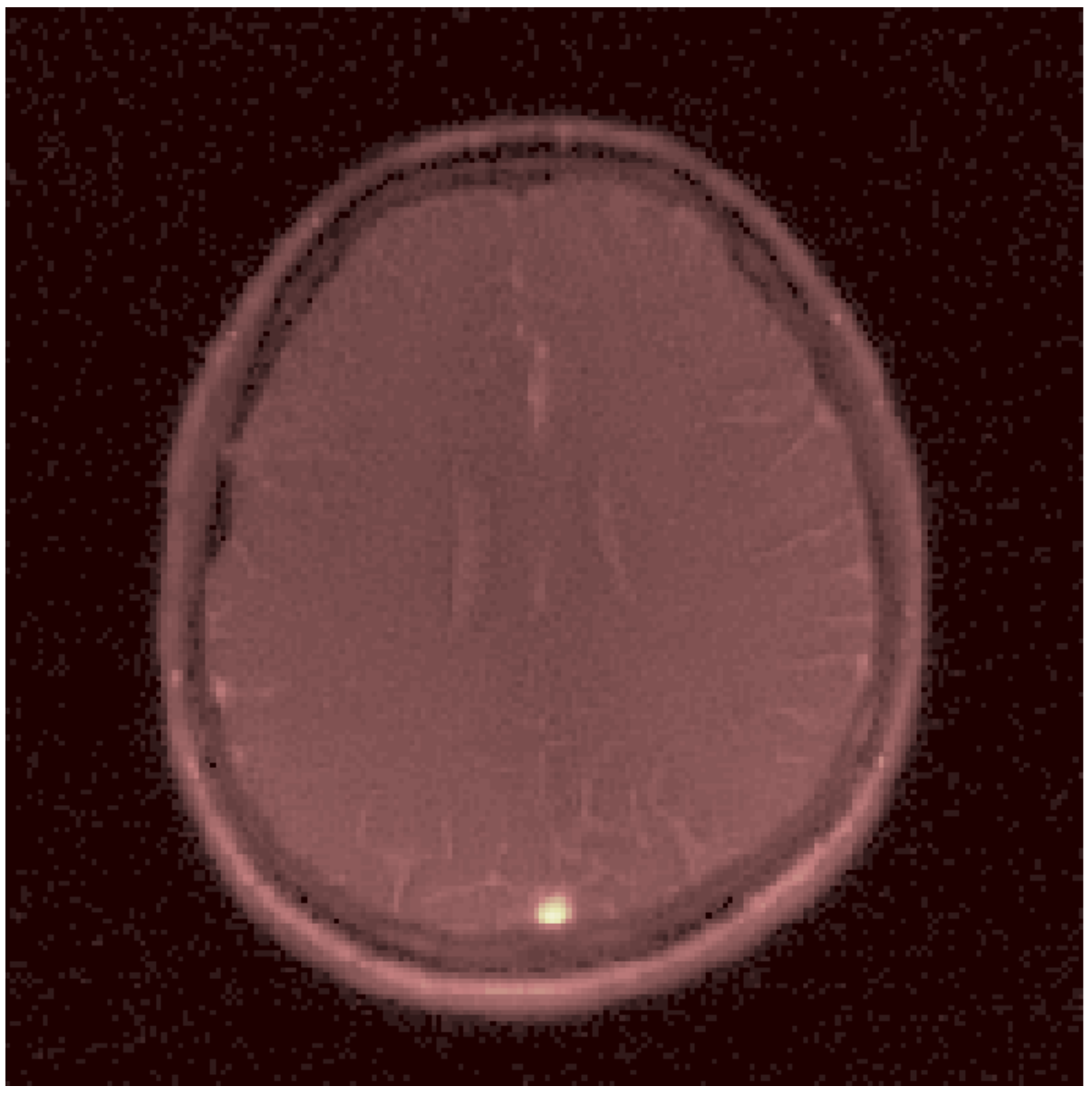}\hspace{-.1cm}
		\includegraphics[width=.162\linewidth]{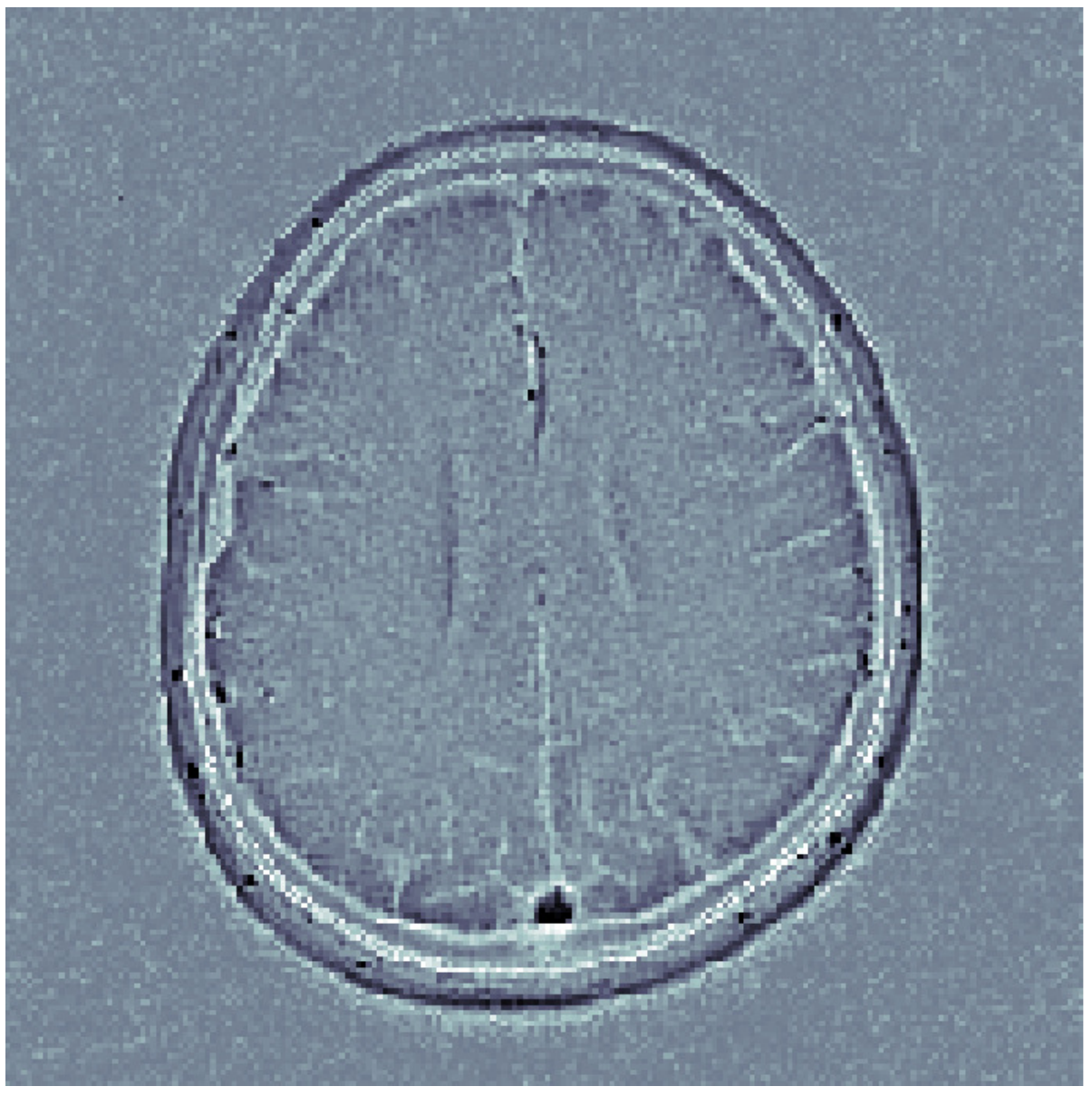}			
				\\
		\begin{turn}{90} \quad\qquad FLOR\end{turn}
		\includegraphics[width=.162\linewidth]{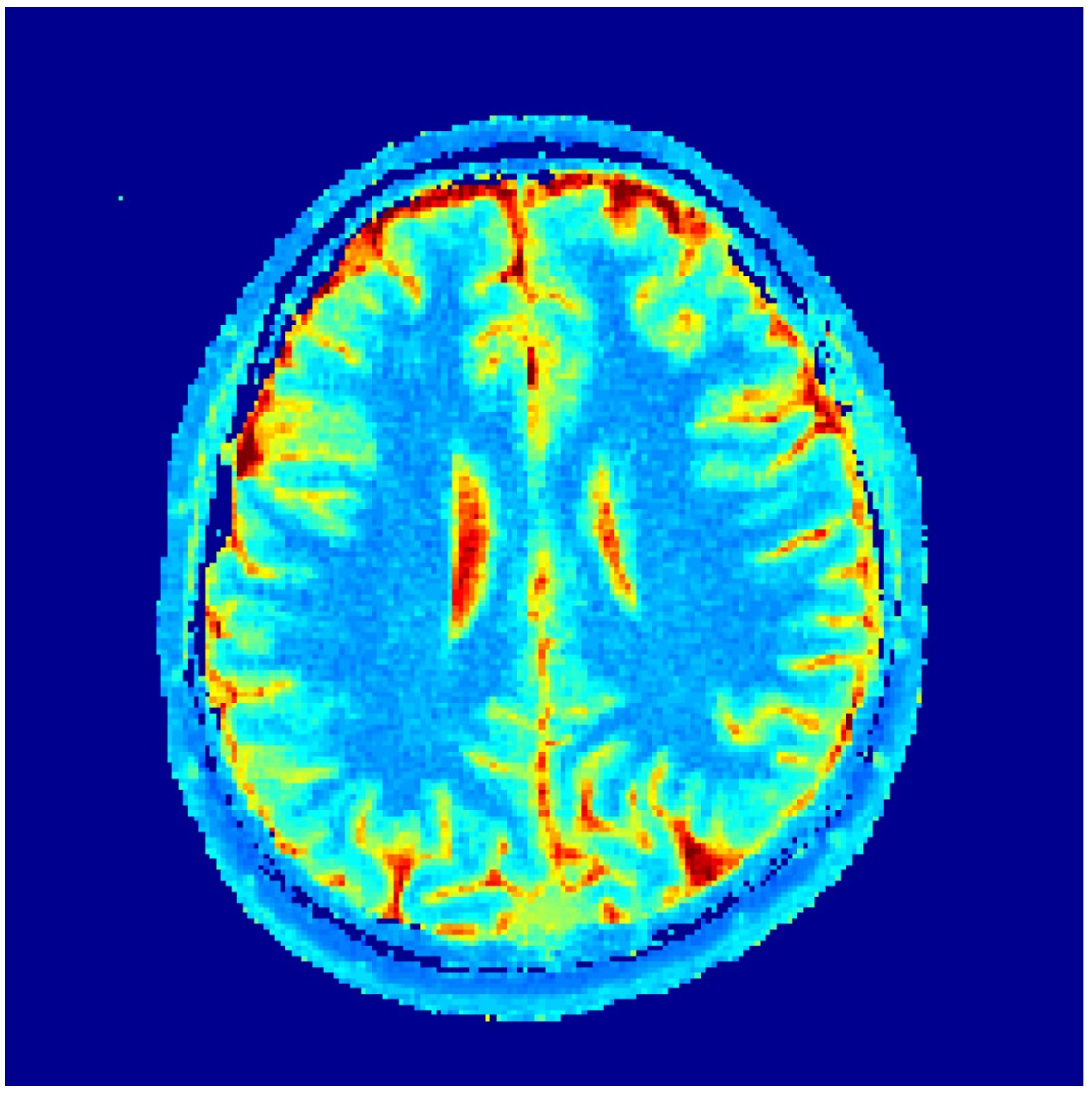}\hspace{-.1cm}
		\includegraphics[width=.162\linewidth]{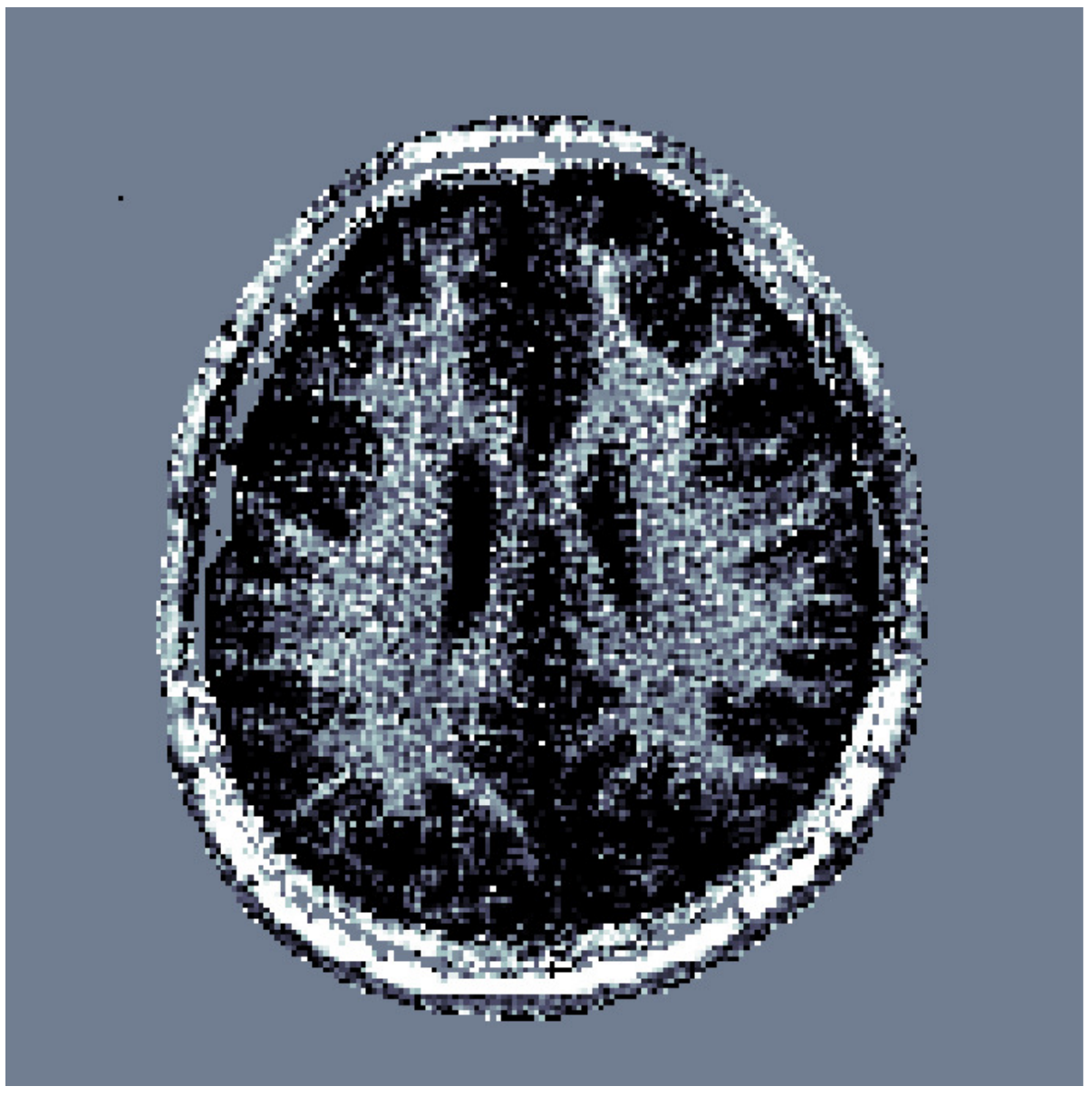}\hspace{.0cm}
		\includegraphics[width=.162\linewidth]{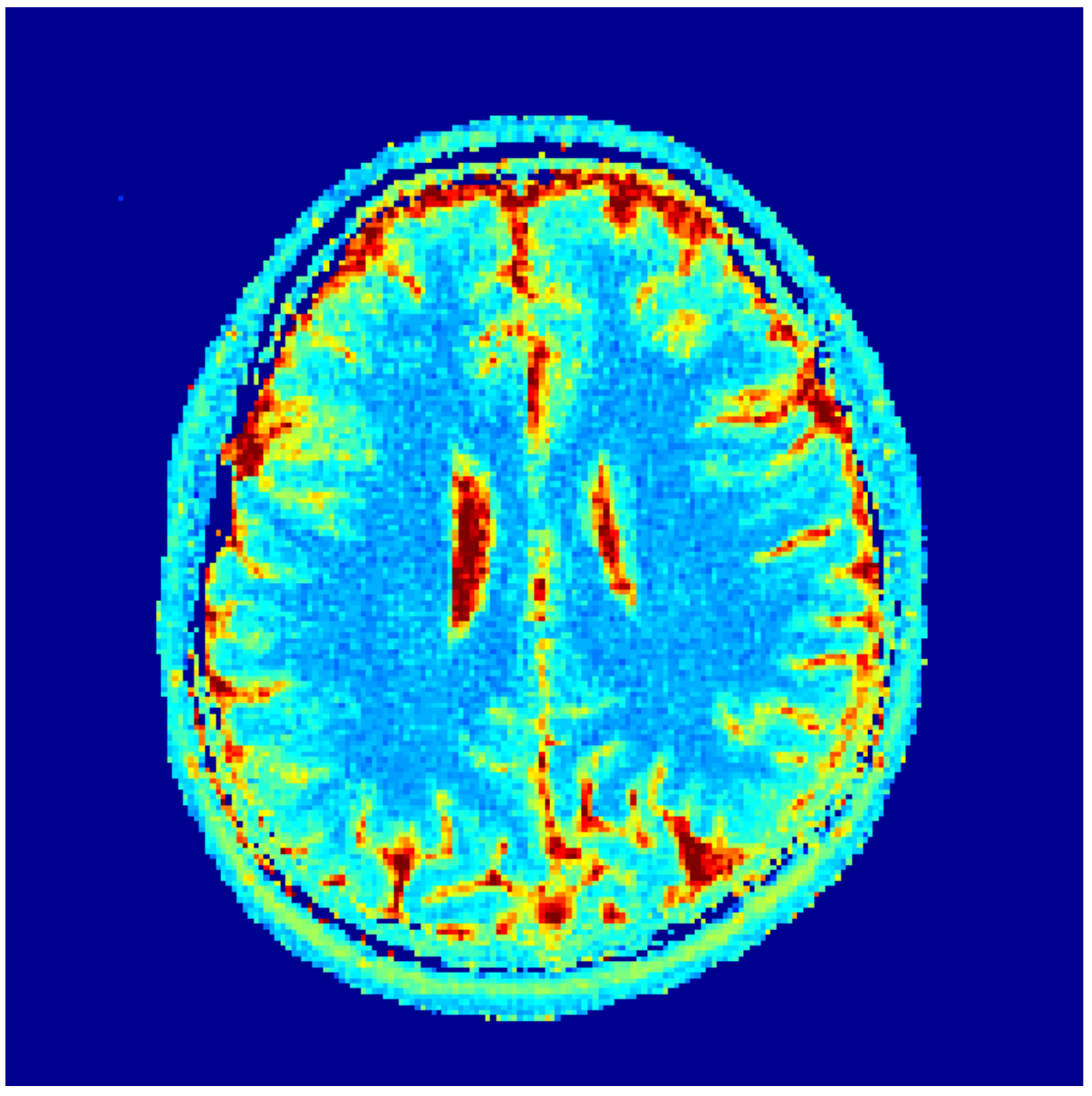}\hspace{-.1cm}
		\includegraphics[width=.162\linewidth]{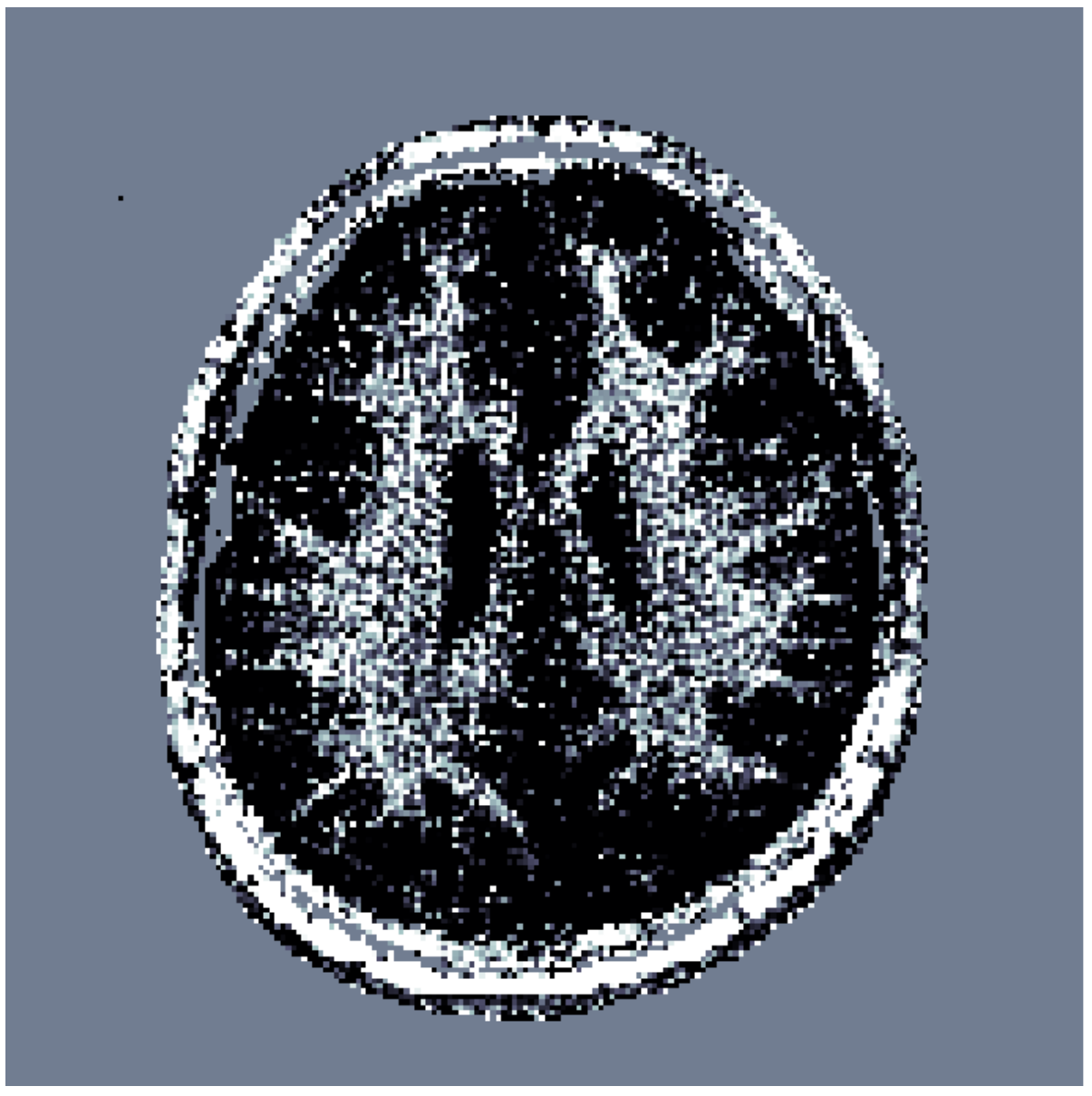}\hspace{.0cm}
		\includegraphics[width=.162\linewidth]{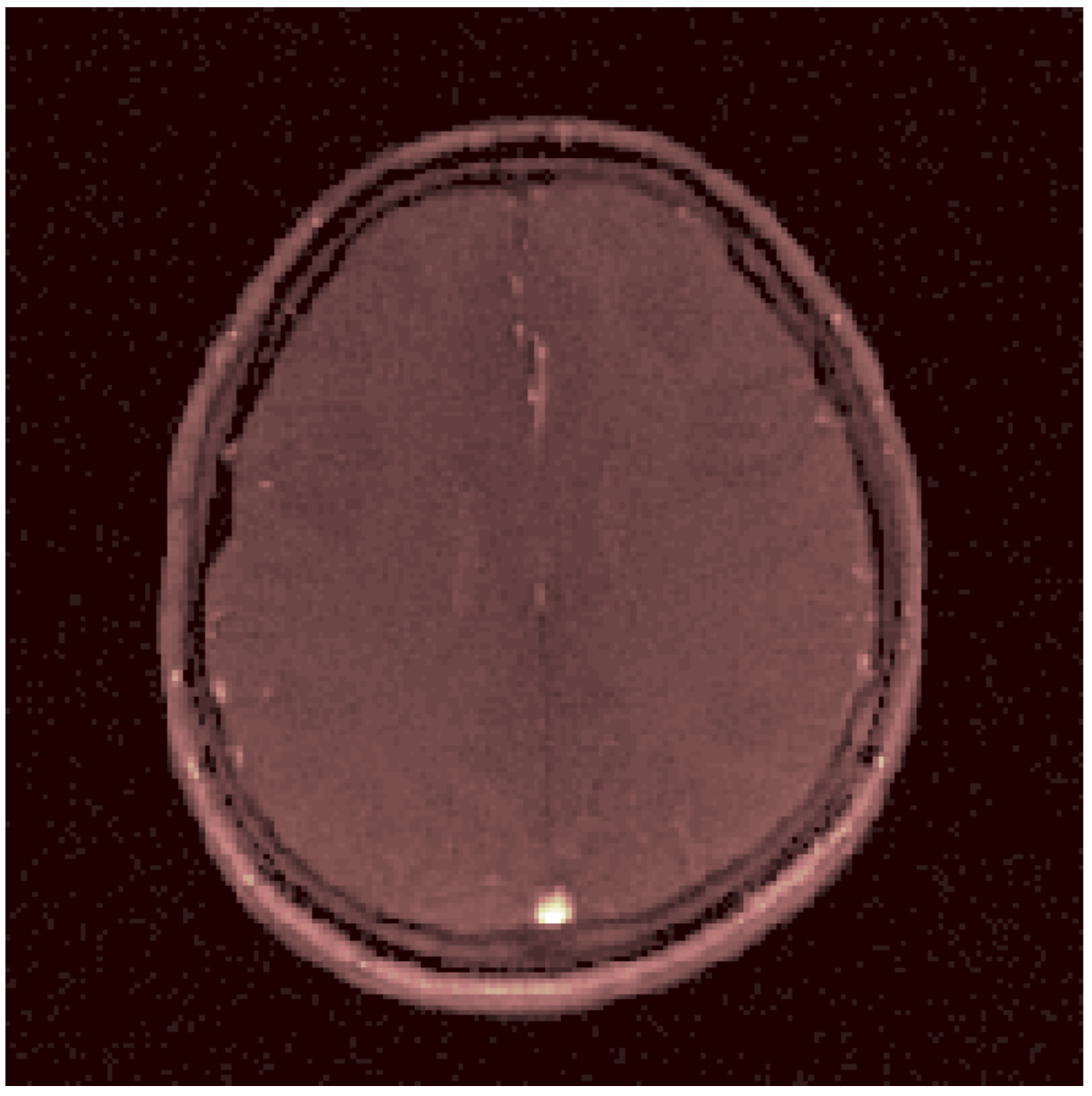}\hspace{-.1cm}
		\includegraphics[width=.162\linewidth]{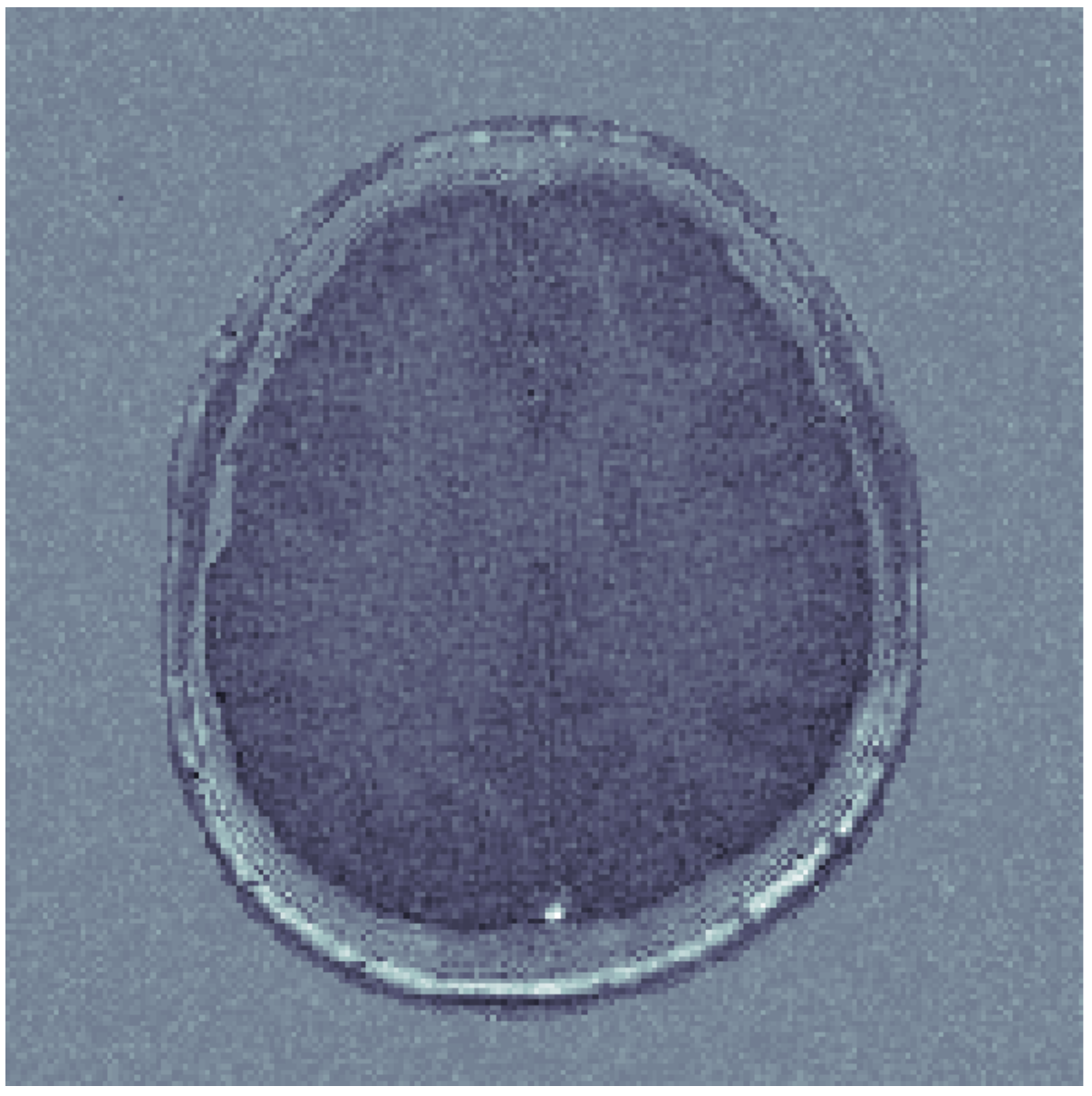}			
				\\
		\begin{turn}{90} \quad\quad AIR-MRF\end{turn}
		\includegraphics[width=.162\linewidth]{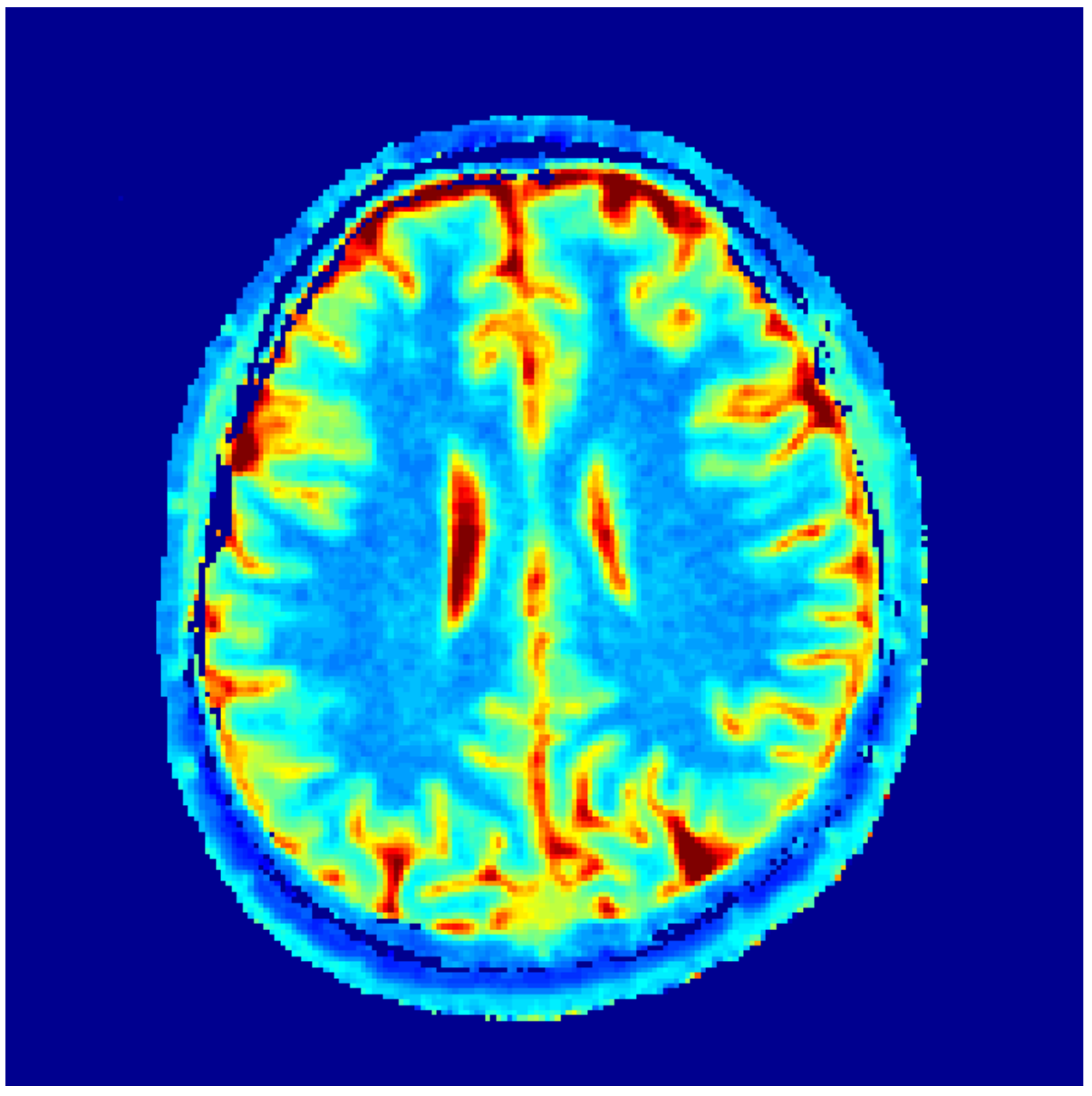}\hspace{-.1cm}
		\includegraphics[width=.162\linewidth]{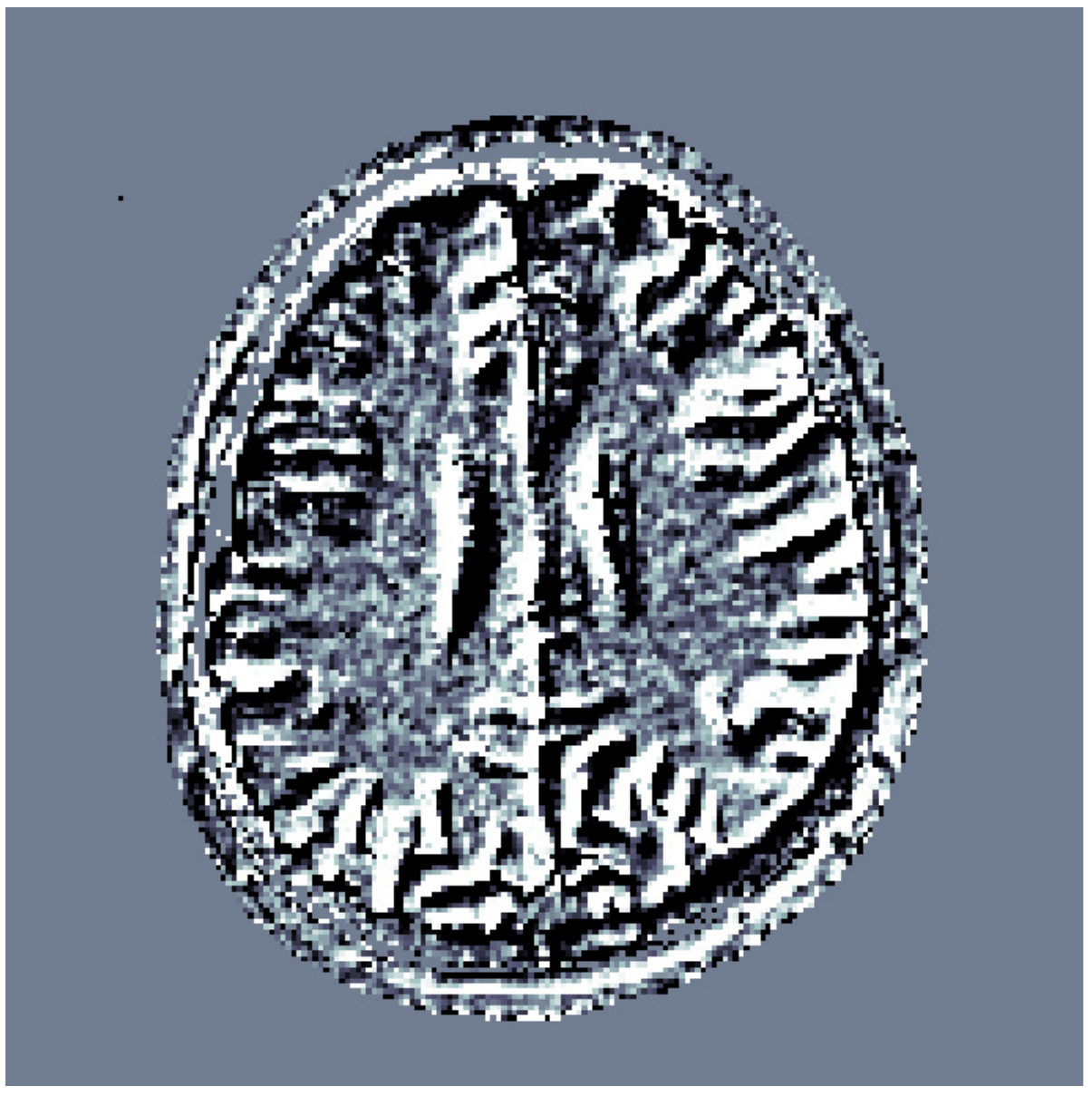}\hspace{.0cm}
		\includegraphics[width=.162\linewidth]{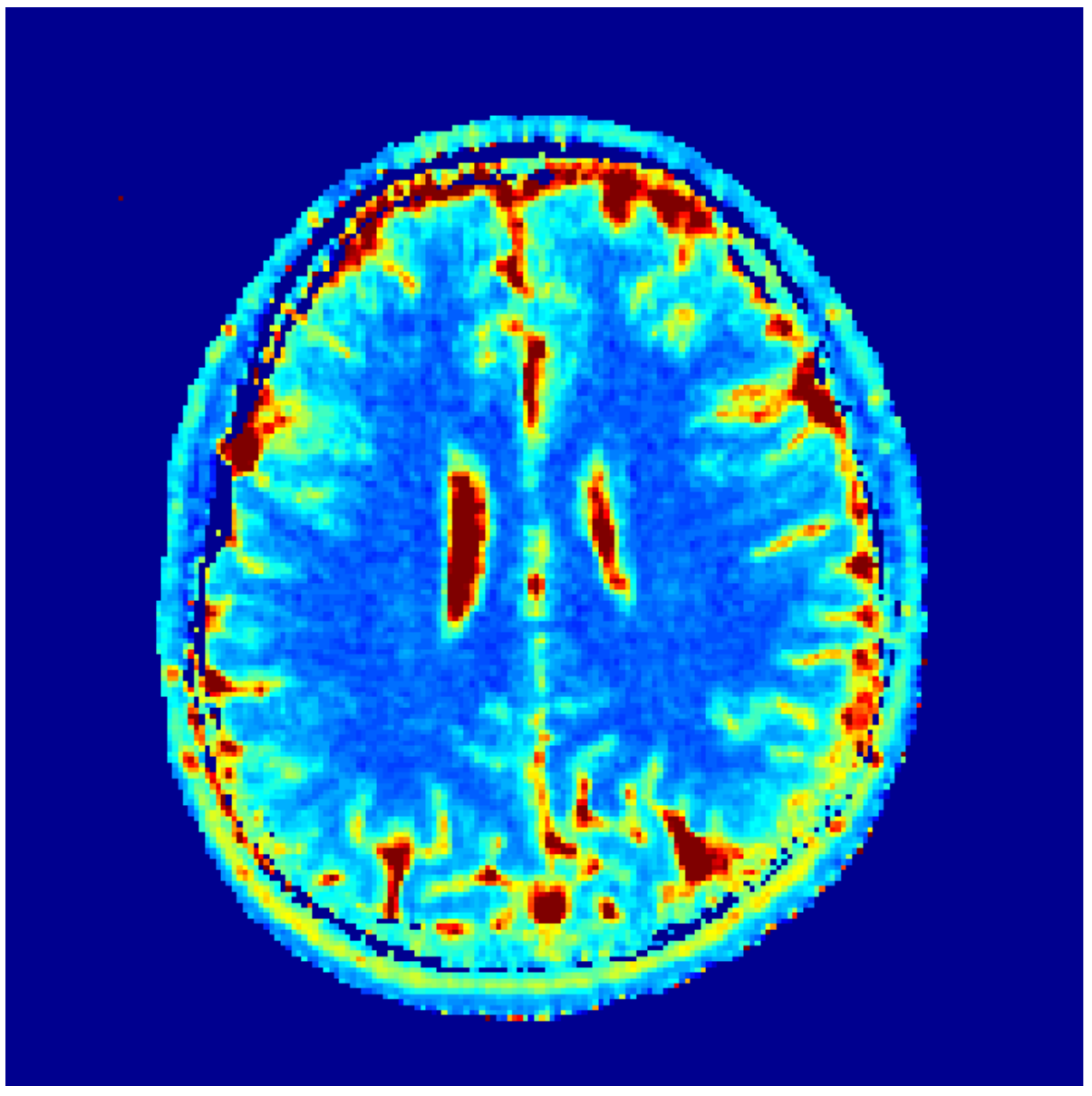}\hspace{-.1cm}
		\includegraphics[width=.162\linewidth]{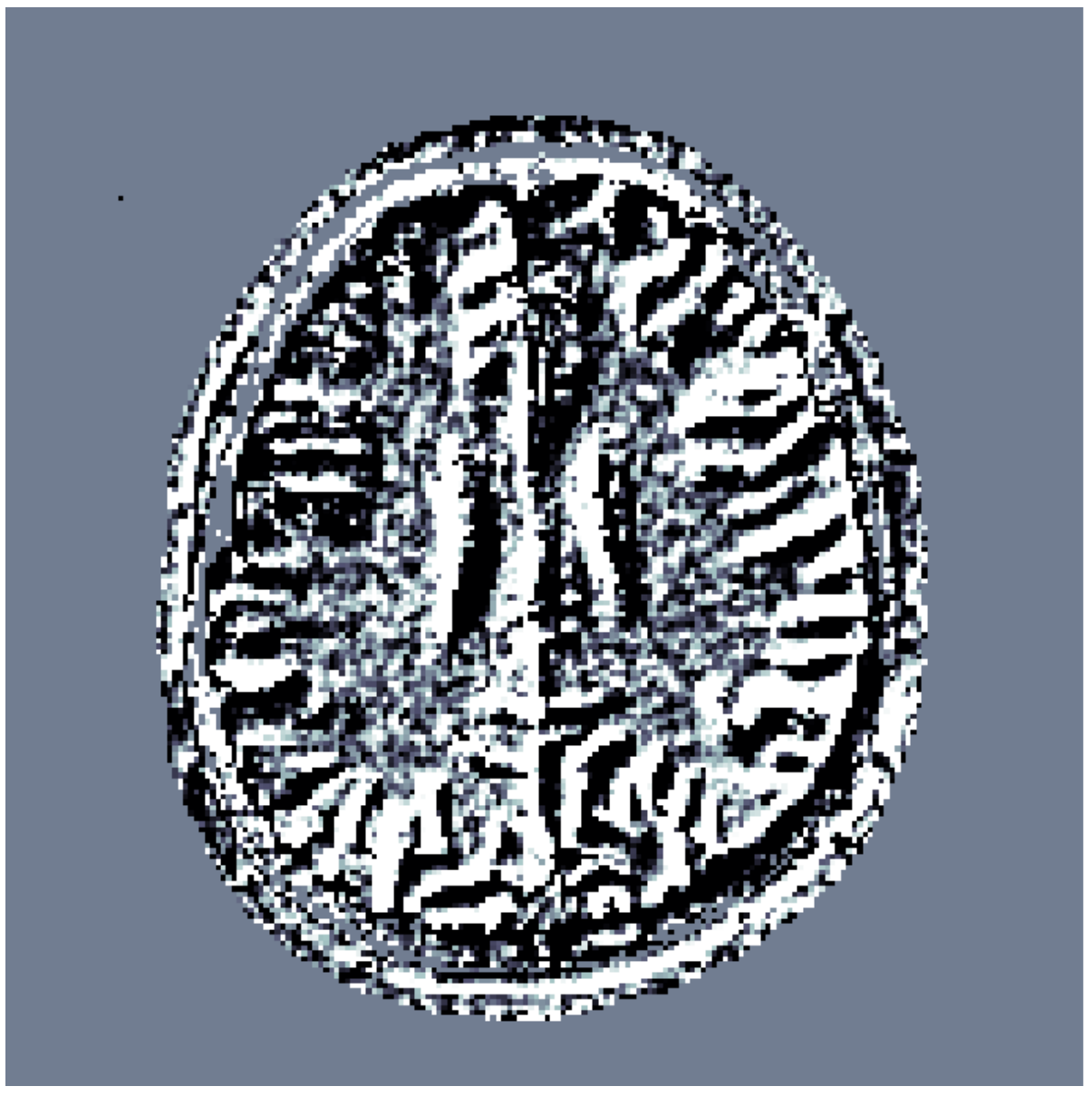}\hspace{.0cm}
		\includegraphics[width=.162\linewidth]{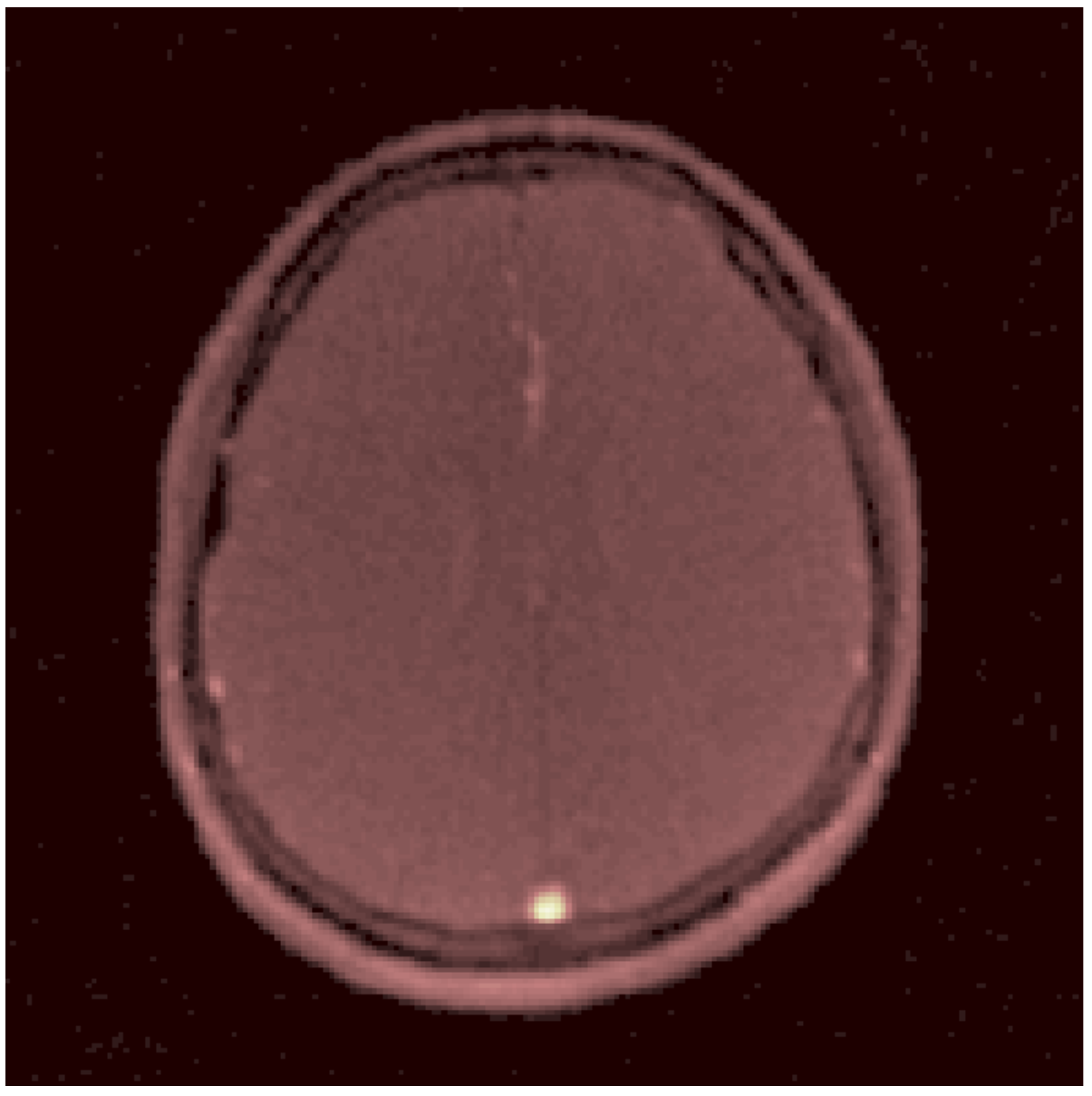}\hspace{-.1cm}
		\includegraphics[width=.162\linewidth]{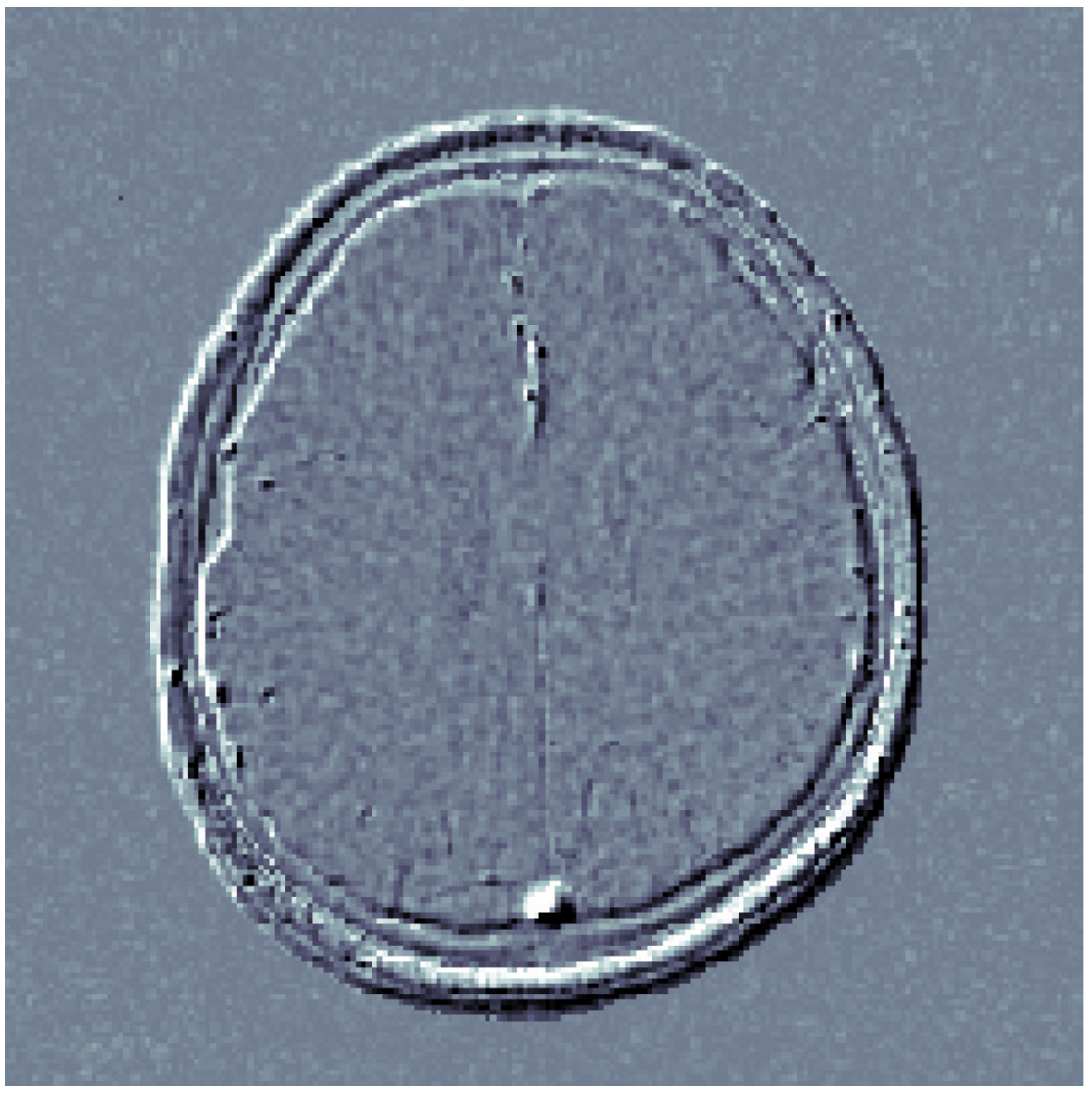}\vspace{.2cm}			
		\hrule
		\hrule
		\hrule
		\begin{turn}{90} \quad\quad LRTV-DM\end{turn}
		\includegraphics[width=.162\linewidth]{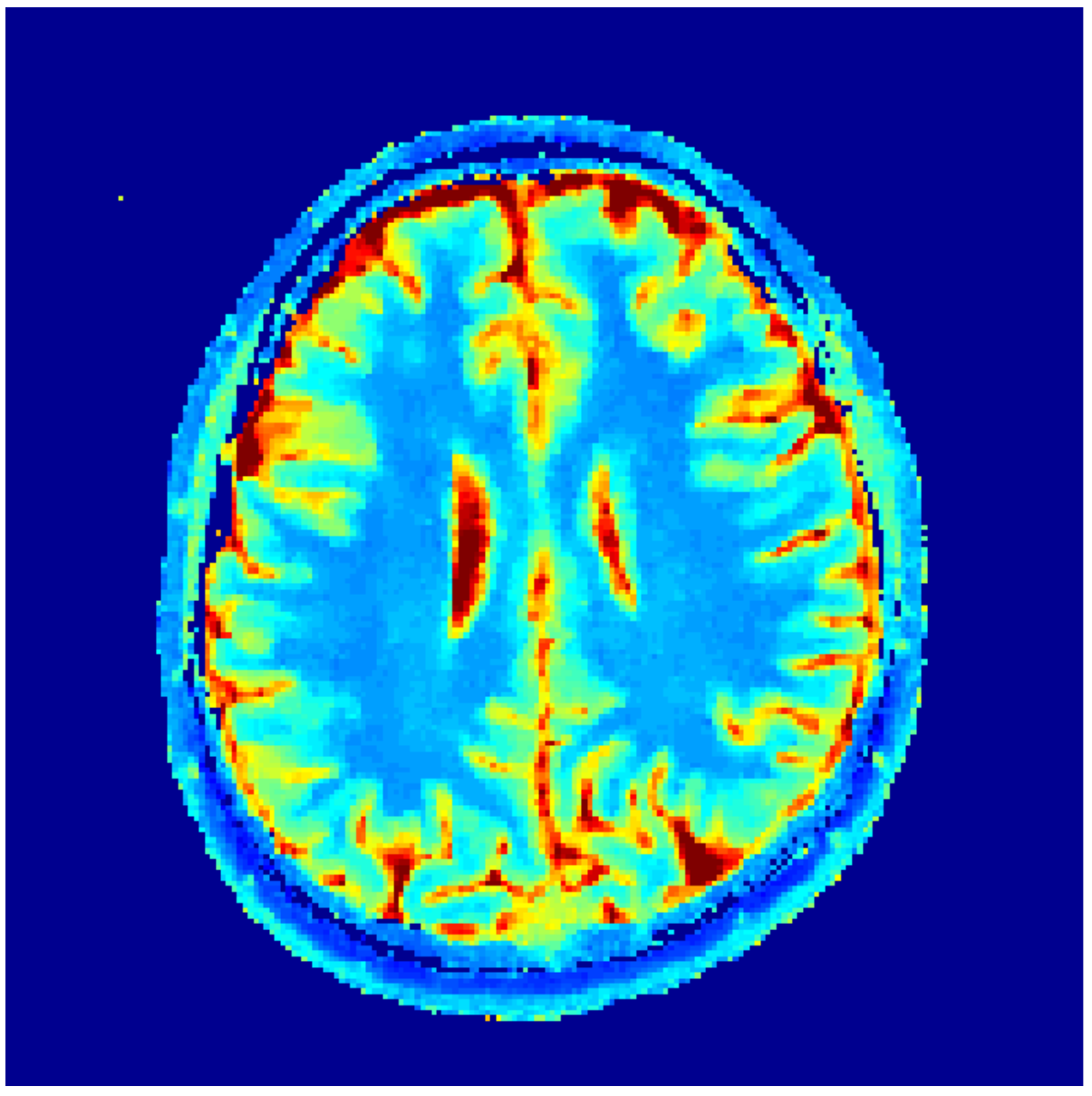}\hspace{-.1cm}
		\includegraphics[width=.162\linewidth]{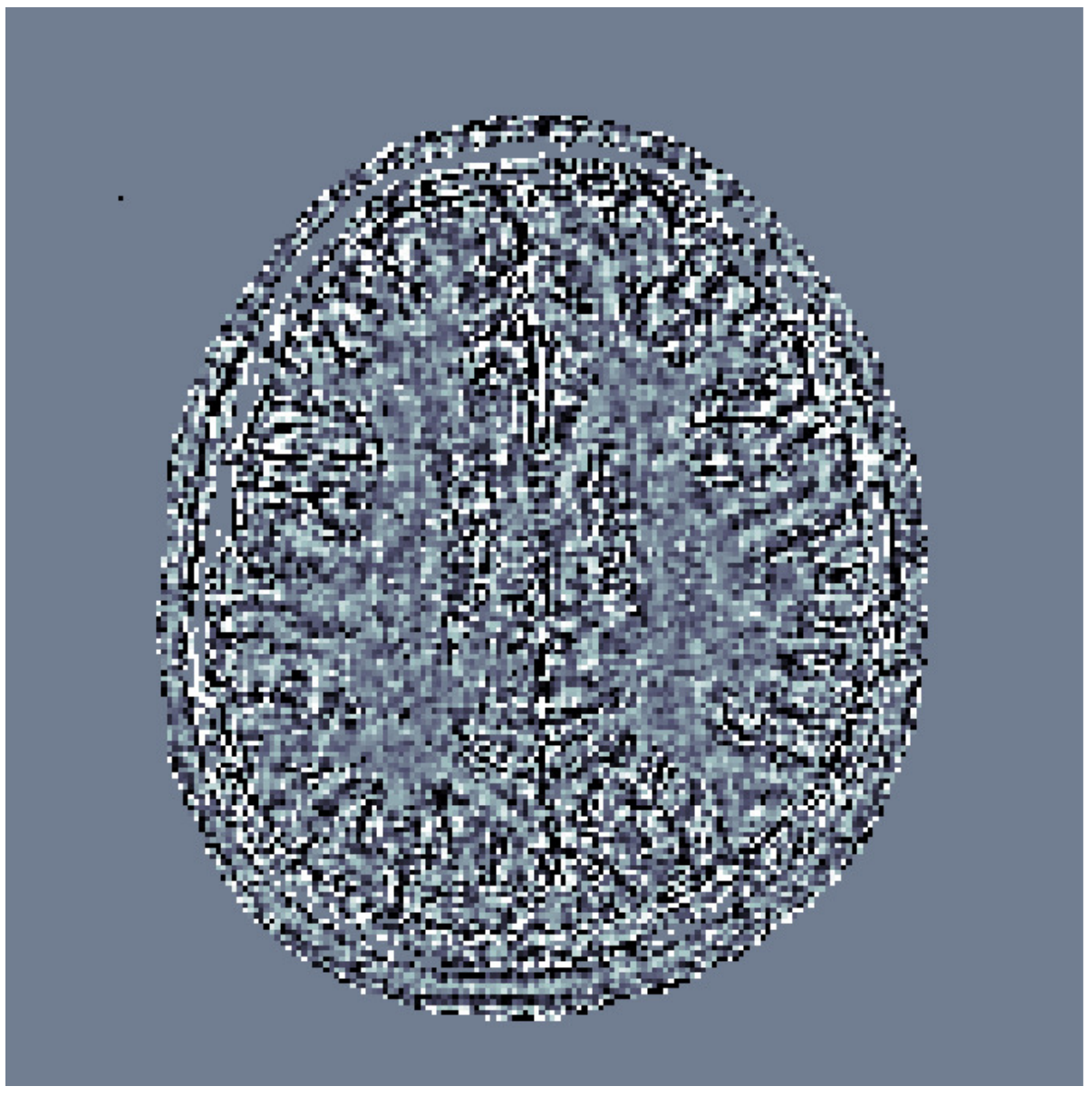}\hspace{.0cm}
		\includegraphics[width=.162\linewidth]{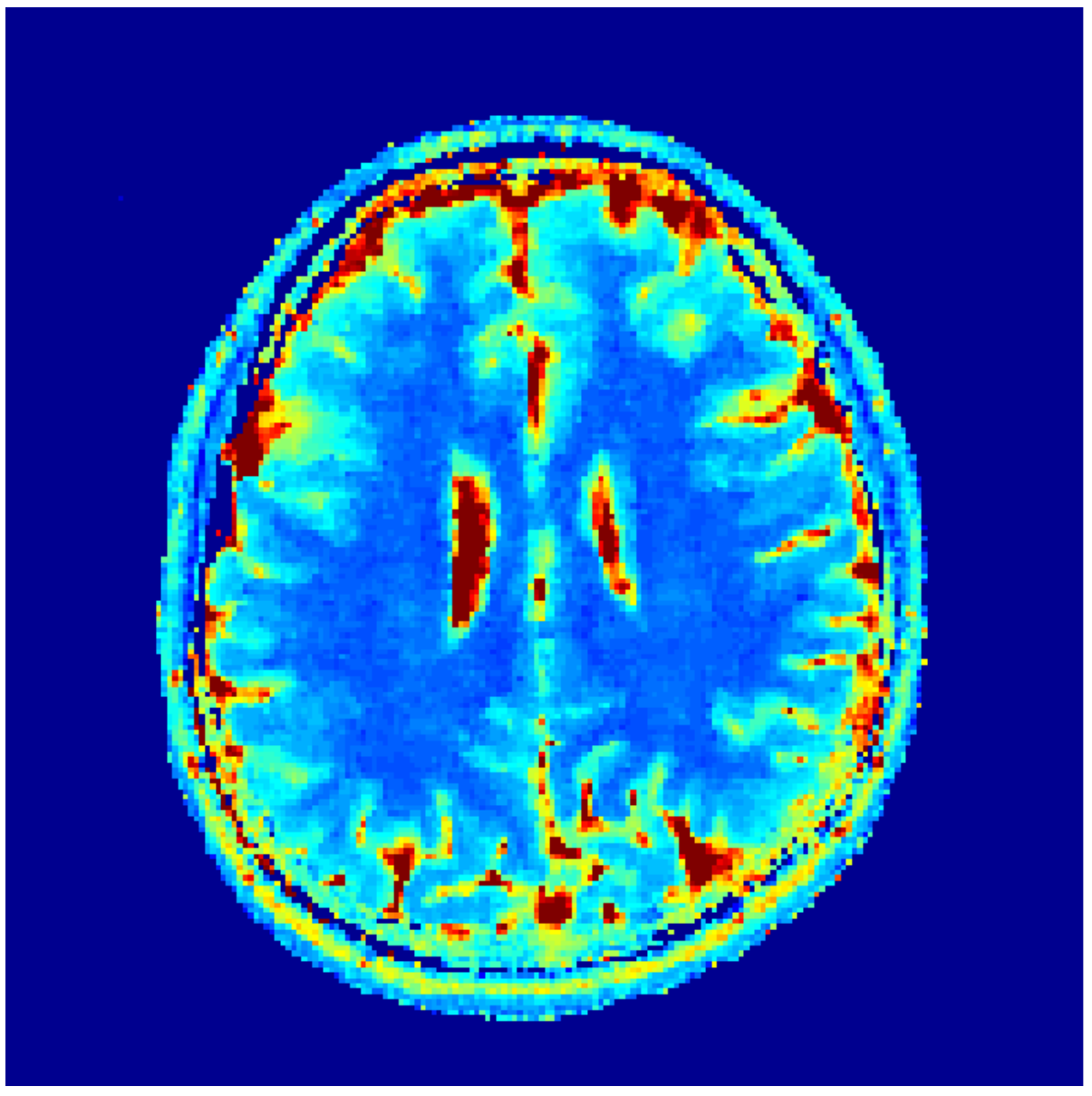}\hspace{-.1cm}
		\includegraphics[width=.162\linewidth]{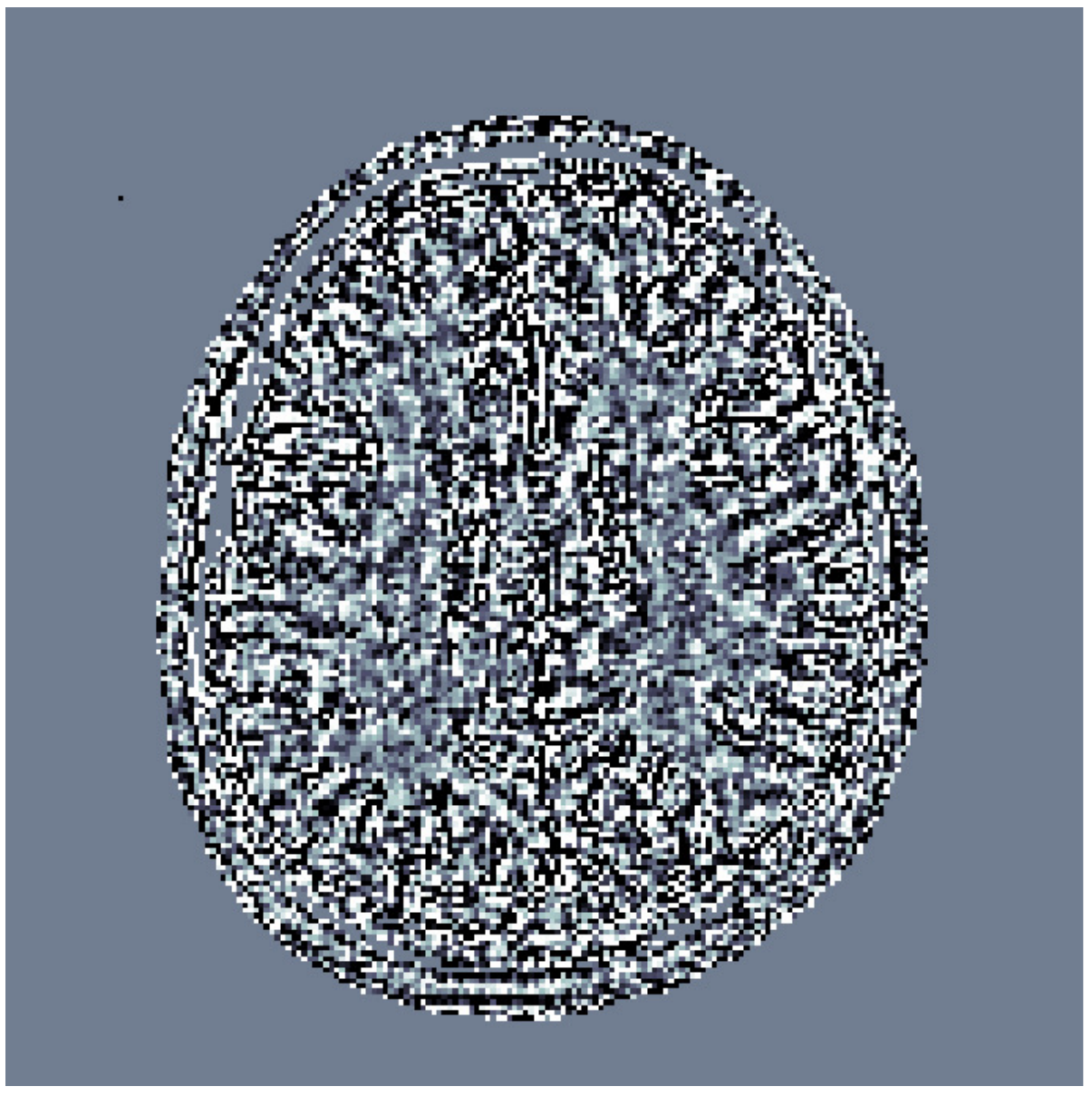}\hspace{.0cm}
		\includegraphics[width=.162\linewidth]{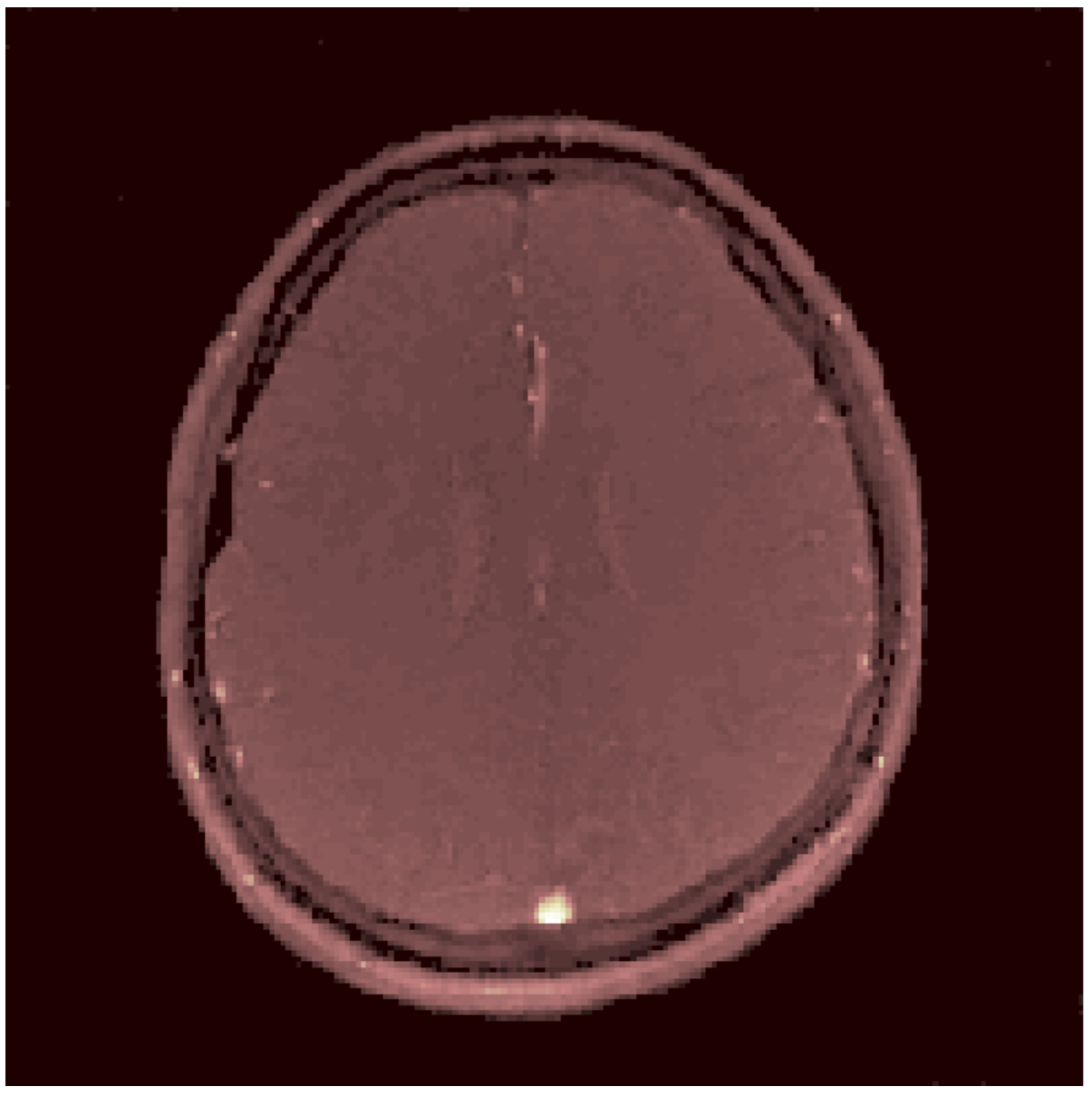}\hspace{-.1cm}
		\includegraphics[width=.162\linewidth]{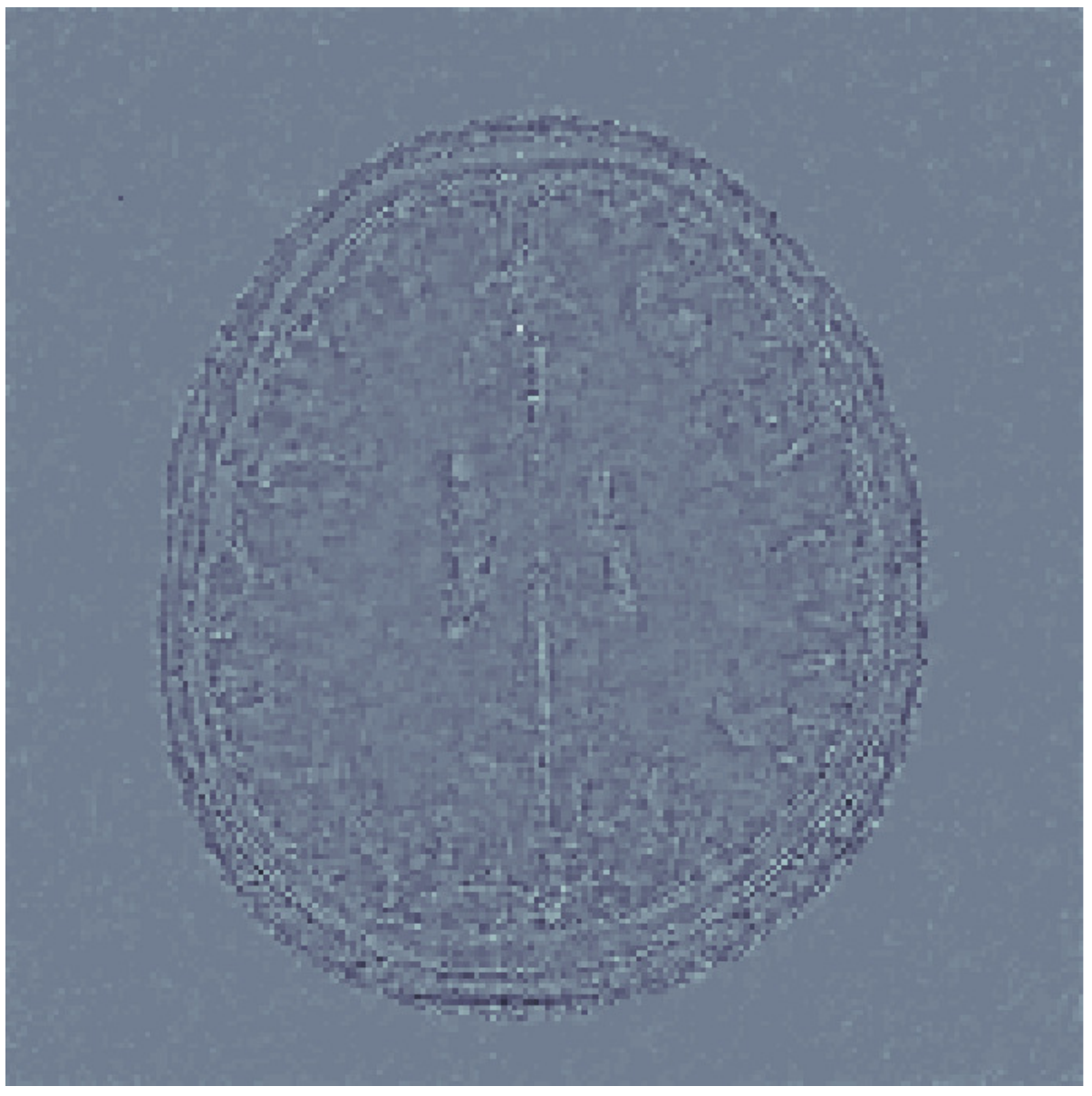}			
		\\
		\begin{turn}{90} \quad\quad LRTV-KM\end{turn}
		\includegraphics[width=.162\linewidth]{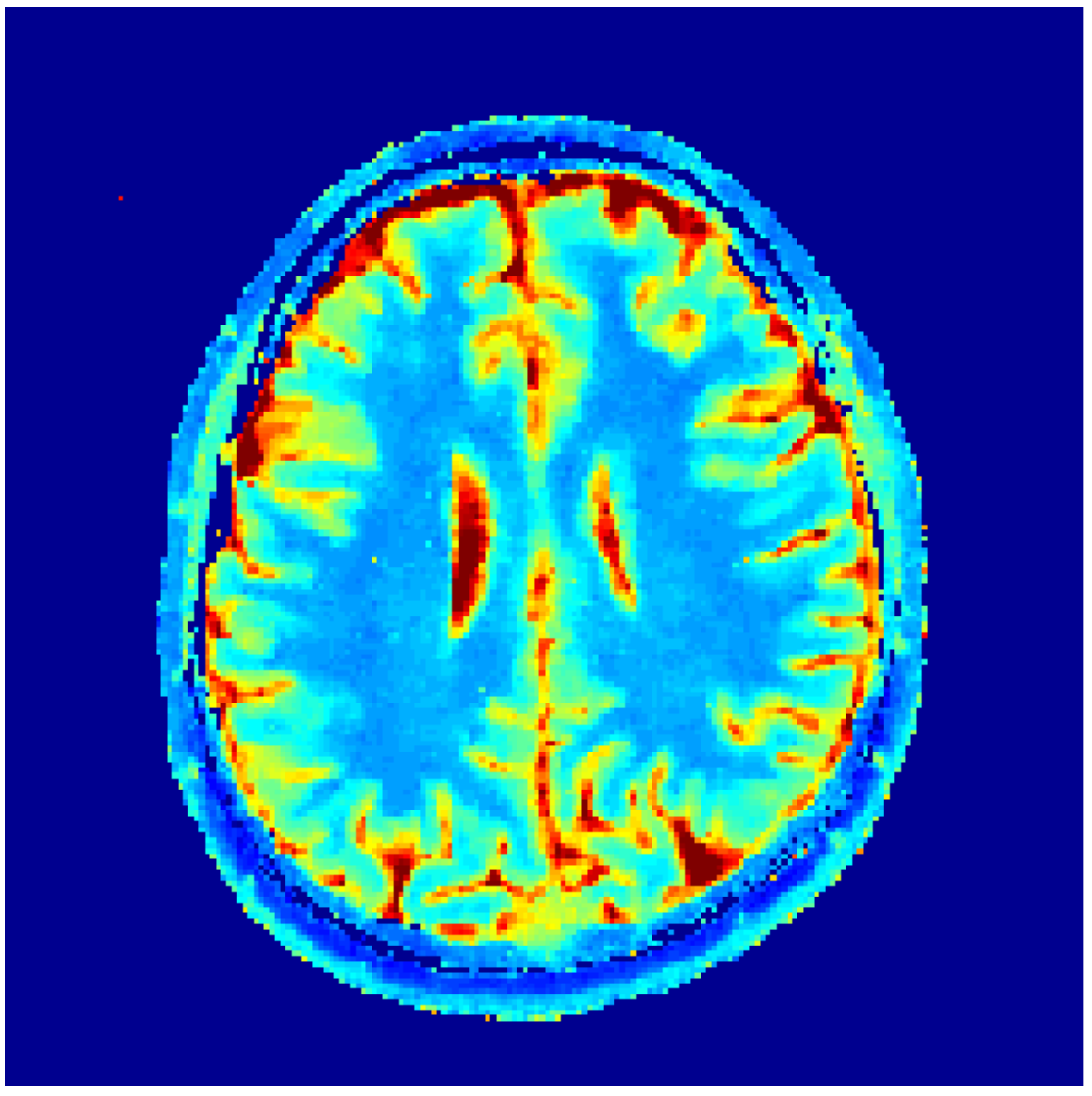}\hspace{-.1cm}
		\includegraphics[width=.162\linewidth]{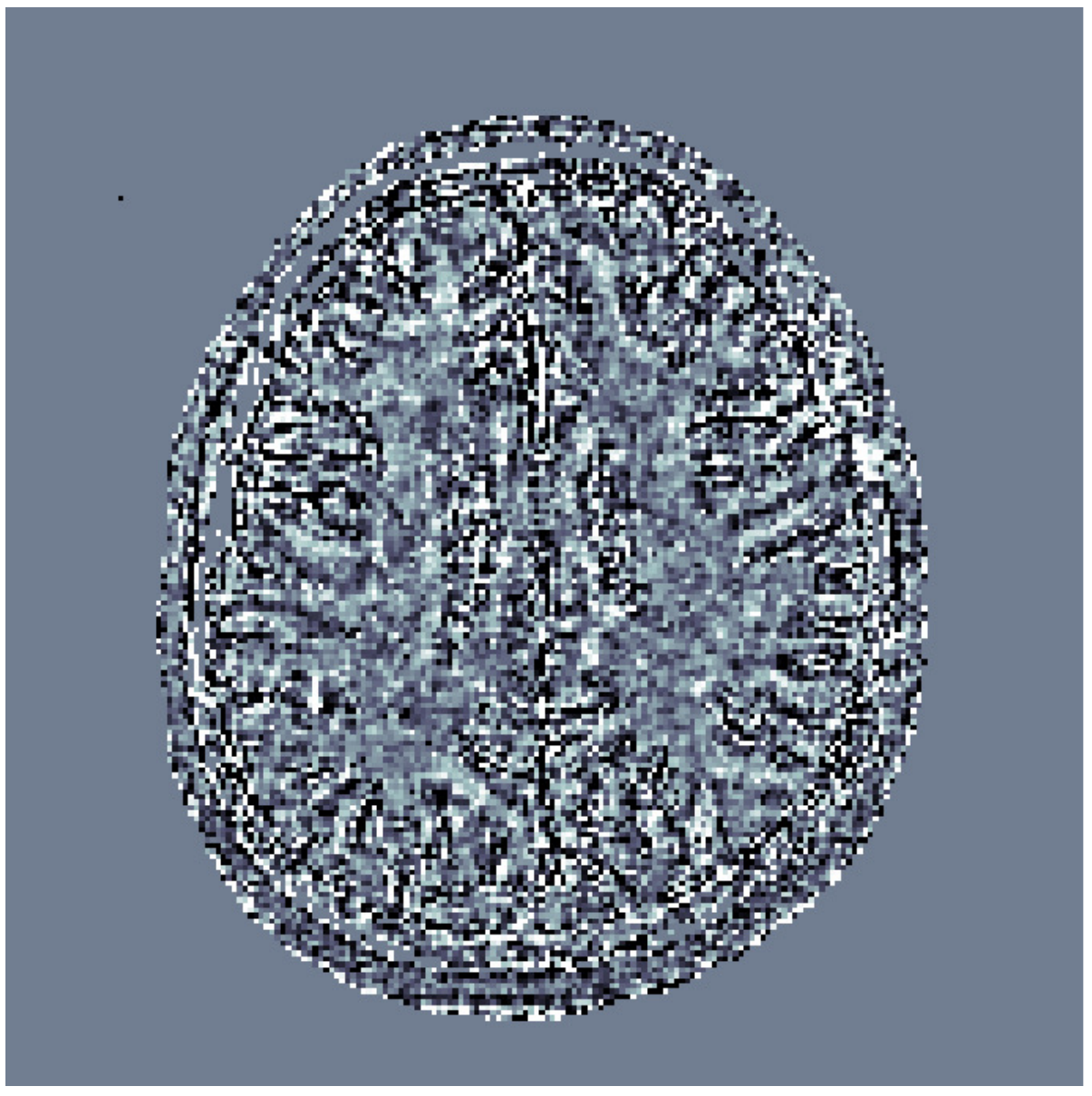}\hspace{.0cm}
		\includegraphics[width=.162\linewidth]{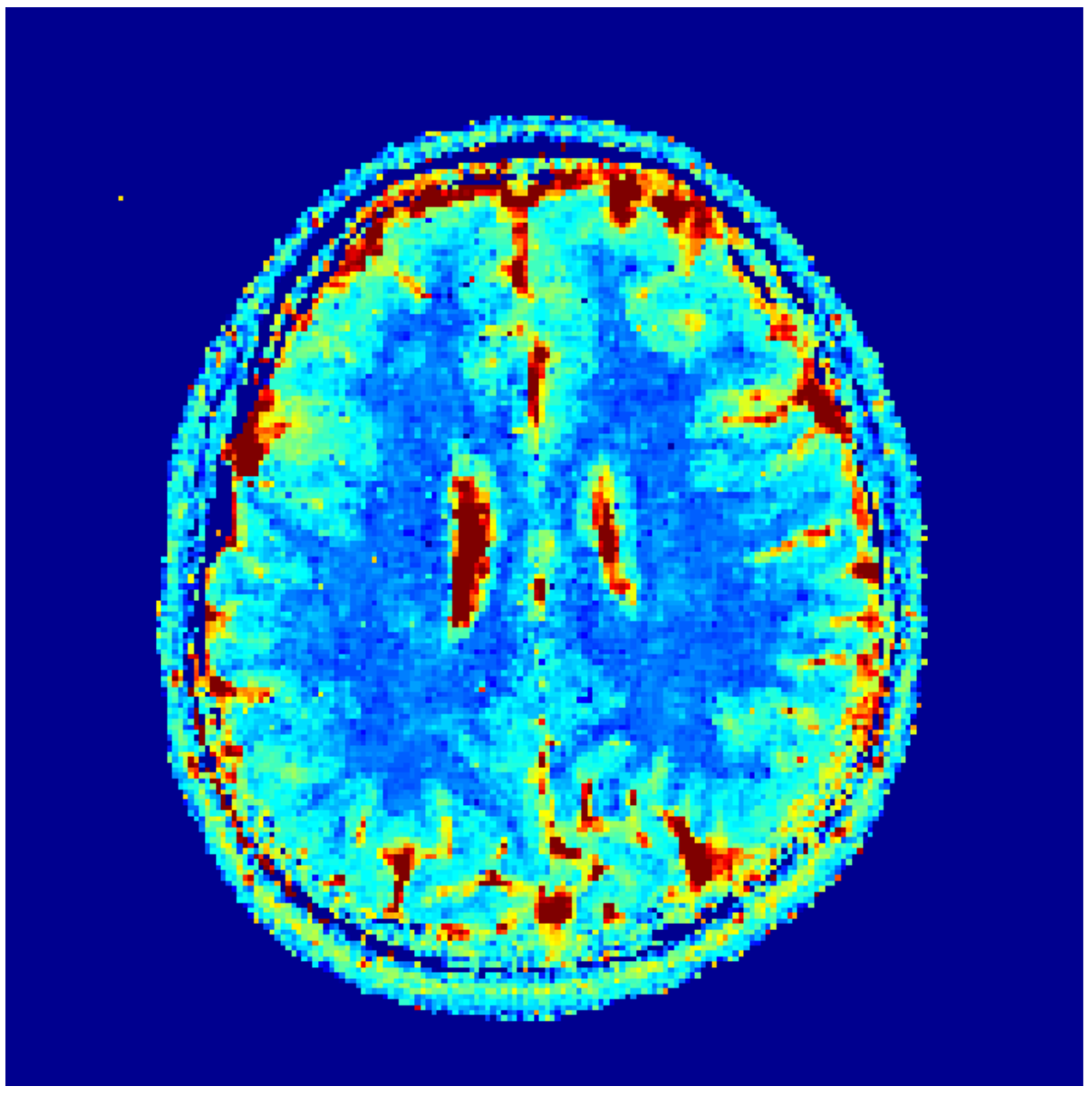}\hspace{-.1cm}
		\includegraphics[width=.162\linewidth]{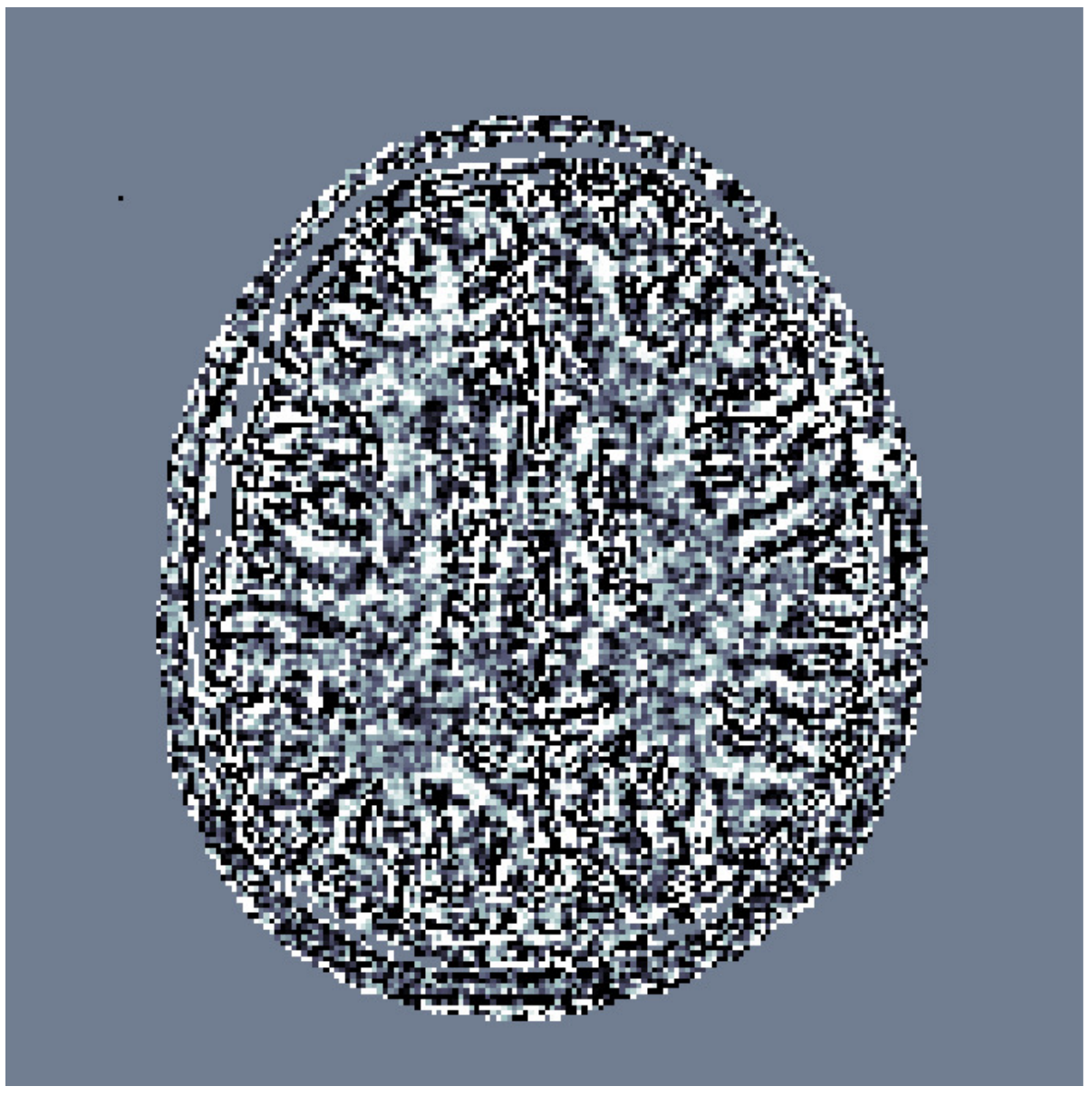}\hspace{.0cm}
		\includegraphics[width=.162\linewidth]{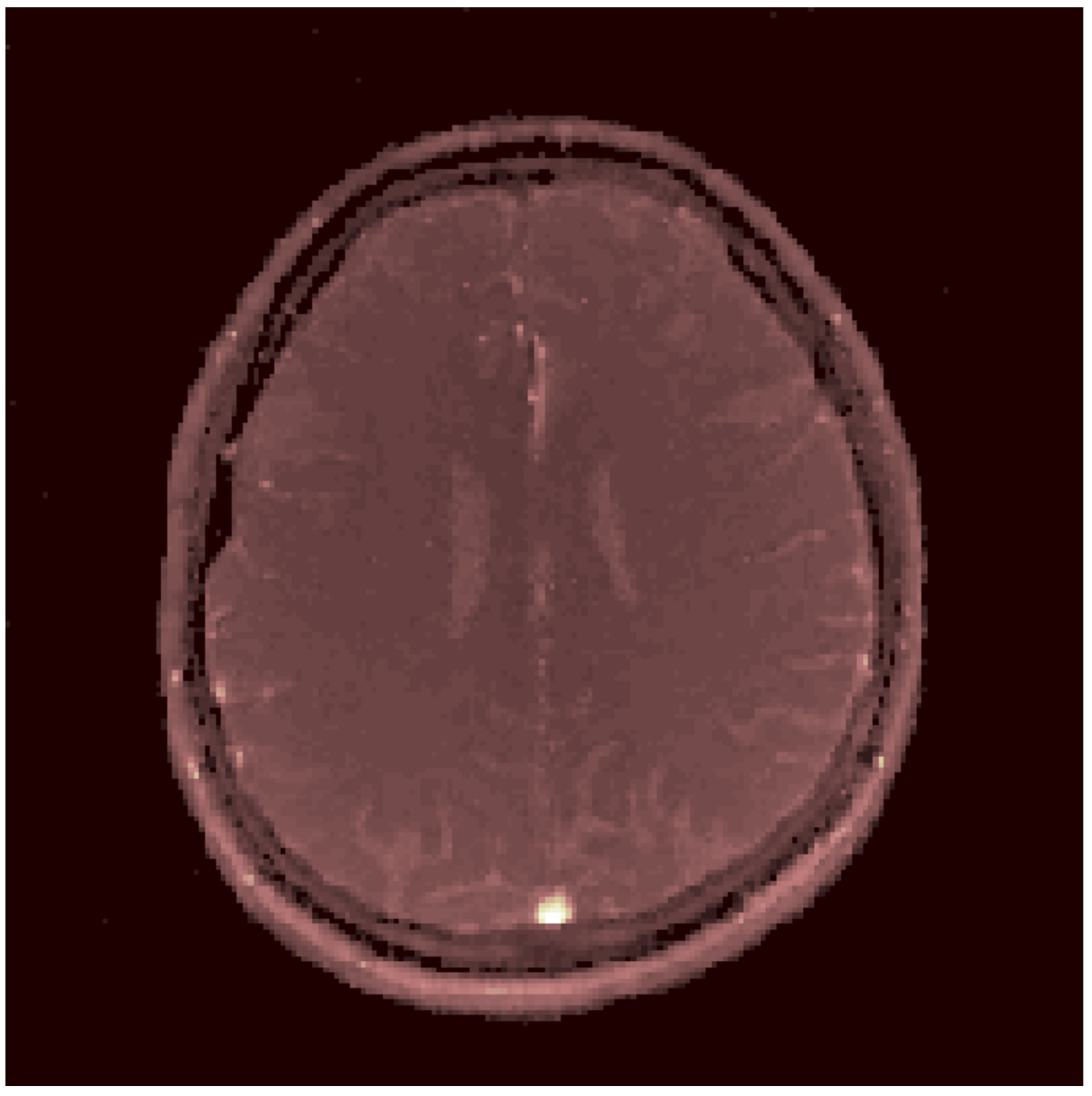}\hspace{-.1cm}
		\includegraphics[width=.162\linewidth]{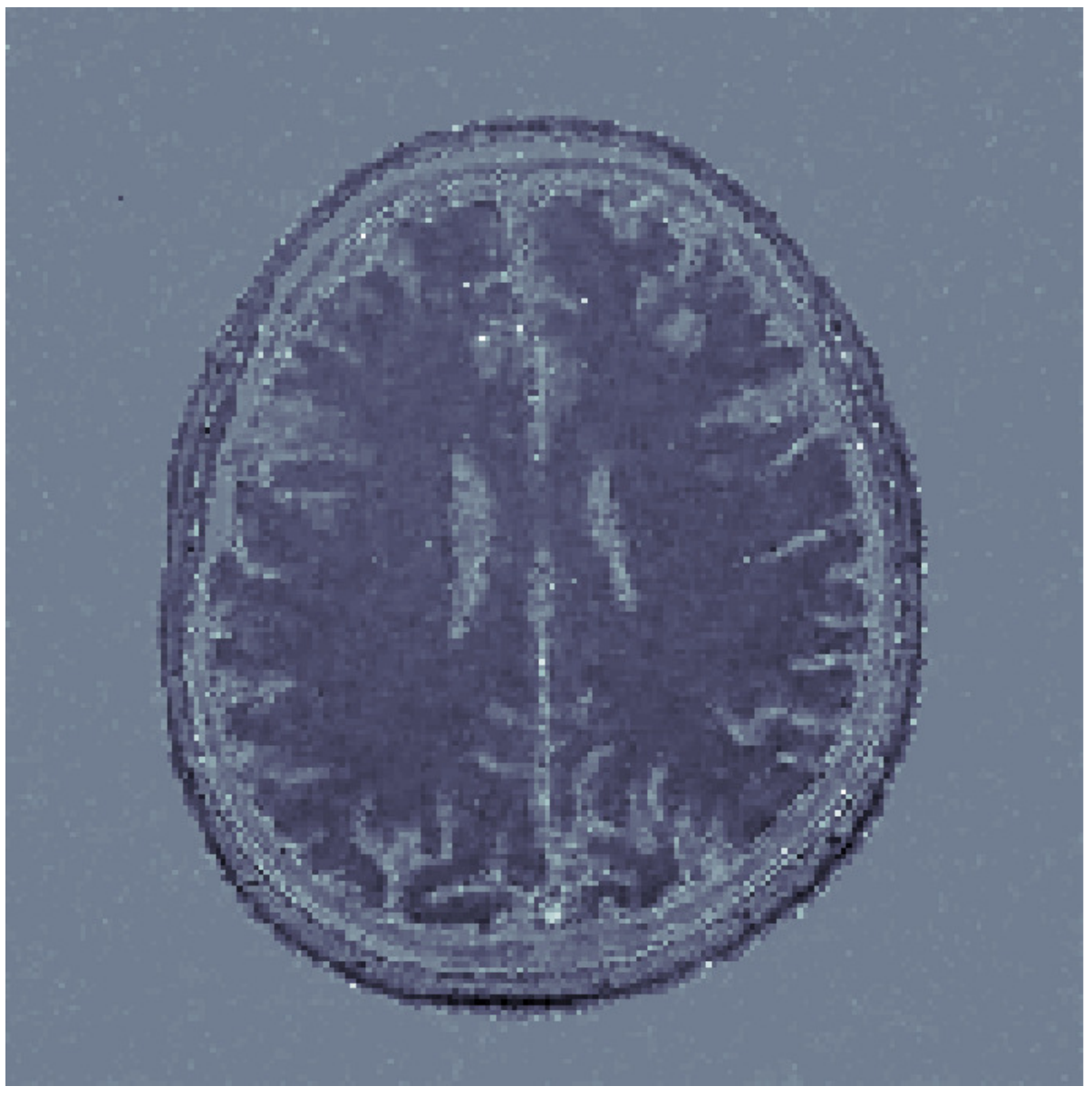}		
		\\
		\begin{turn}{90}LRTV-MRFResnet\end{turn}
		\includegraphics[width=.162\linewidth]{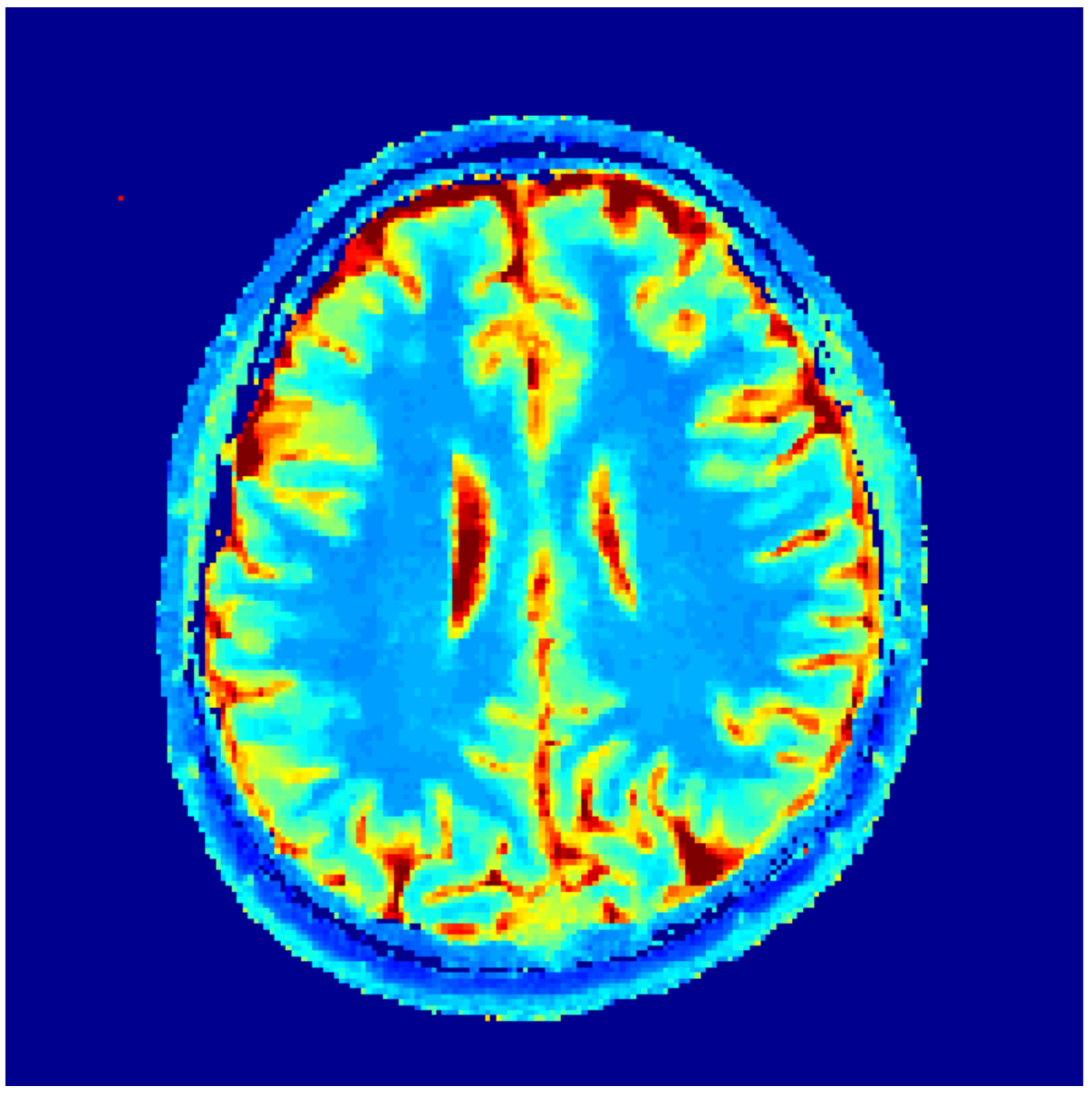}\hspace{-.1cm}
		\includegraphics[width=.162\linewidth]{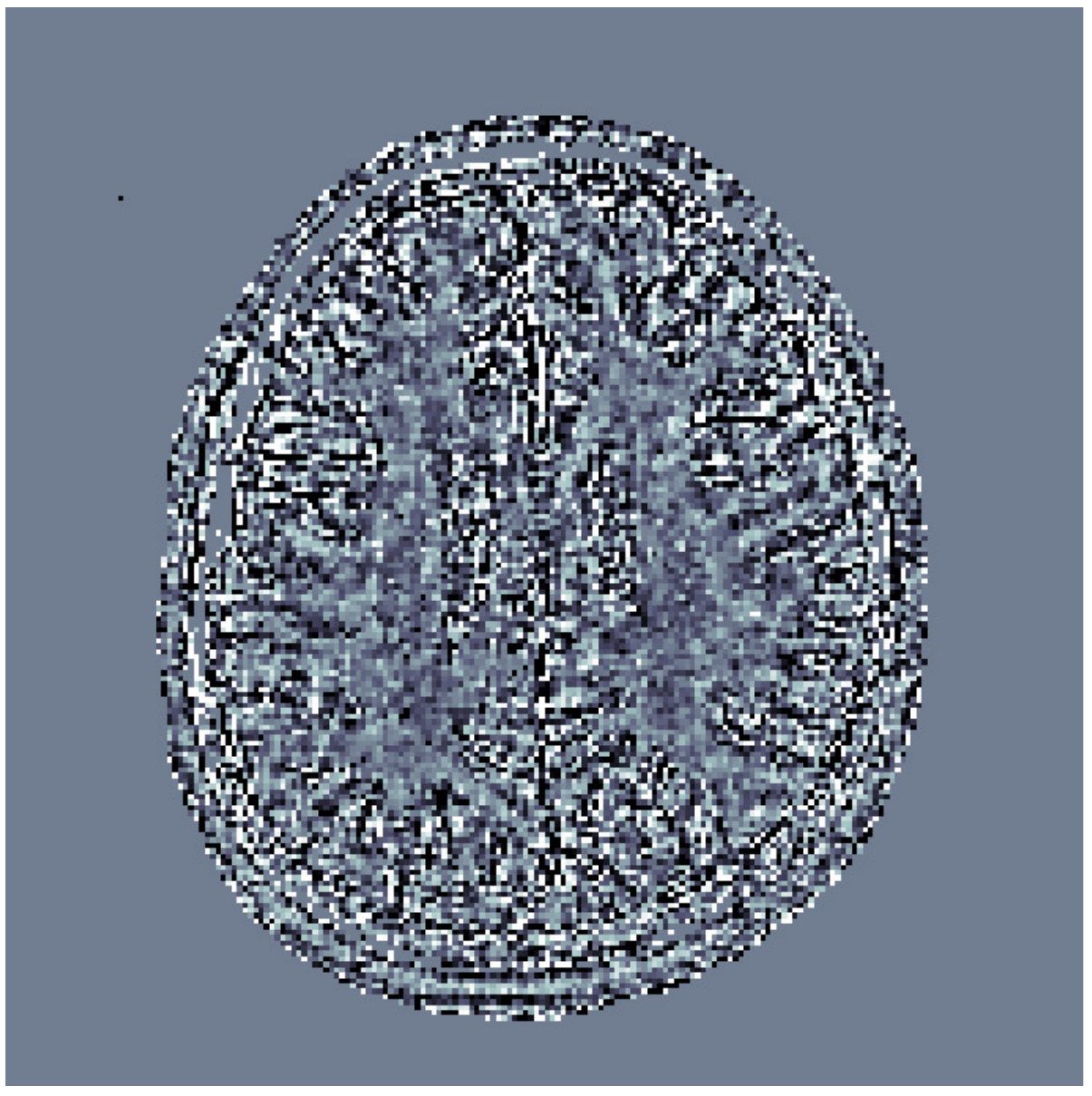}\hspace{.0cm}
		\includegraphics[width=.162\linewidth]{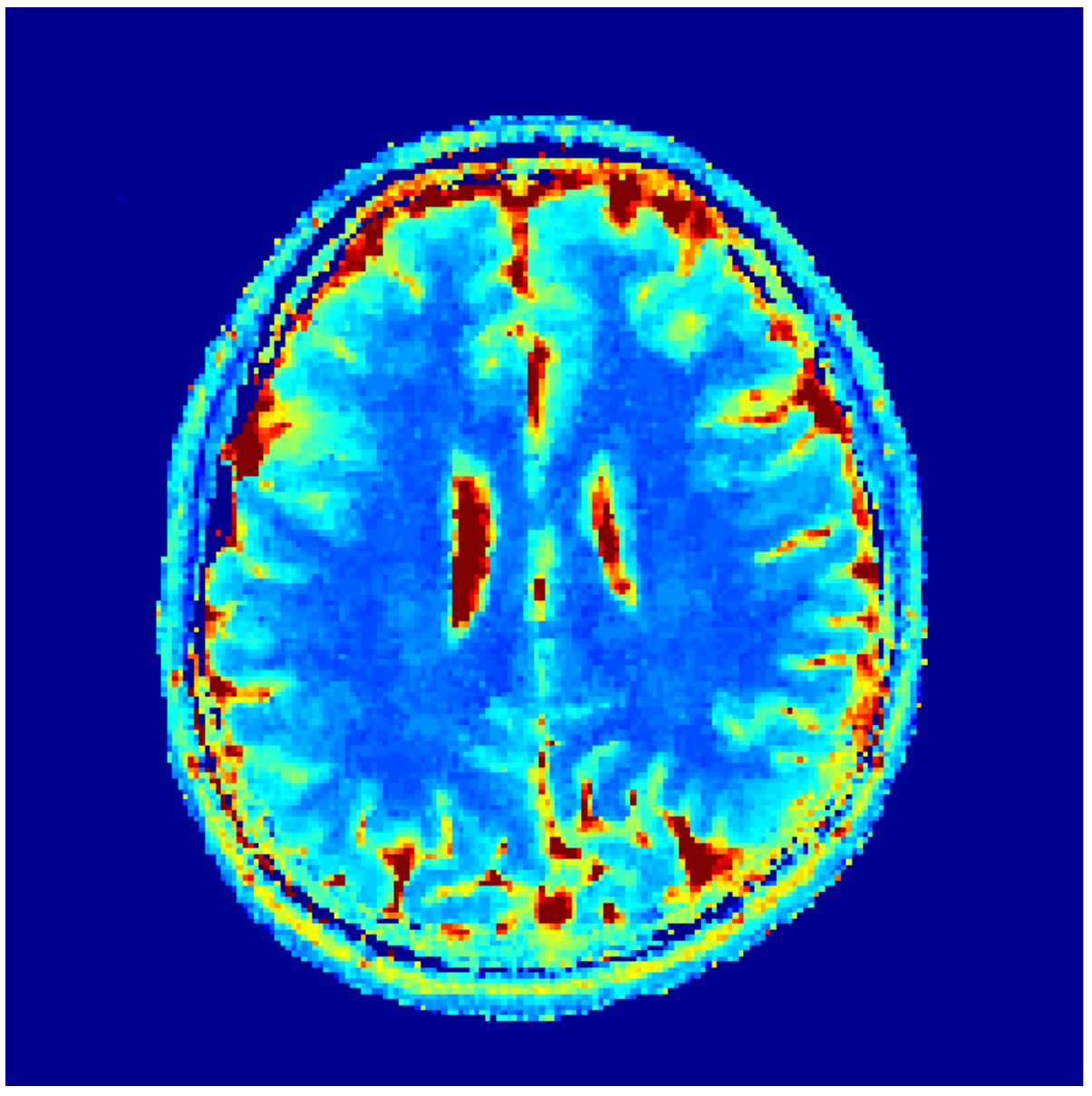}\hspace{-.1cm}
		\includegraphics[width=.162\linewidth]{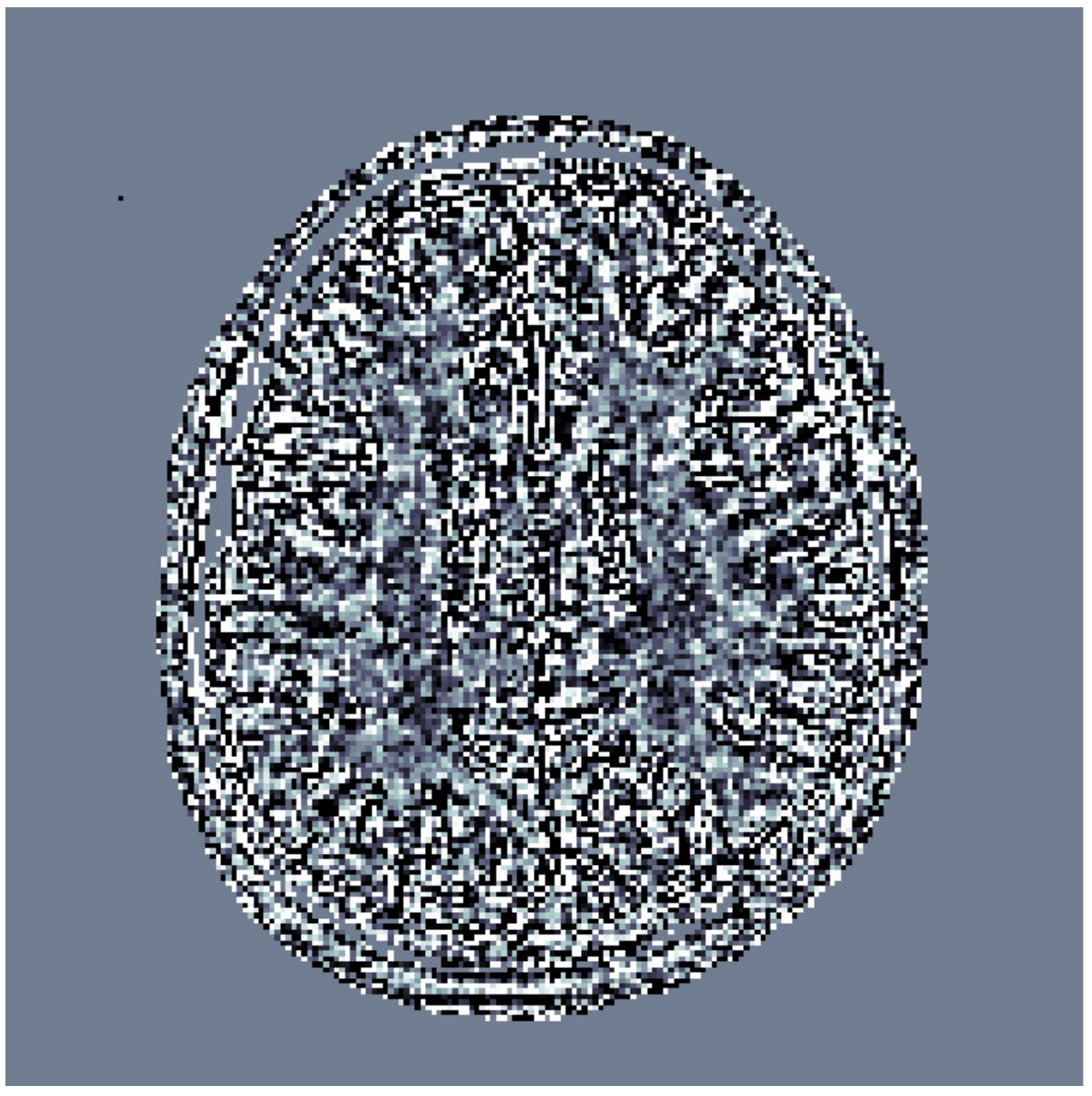}\hspace{.0cm}
		\includegraphics[width=.162\linewidth]{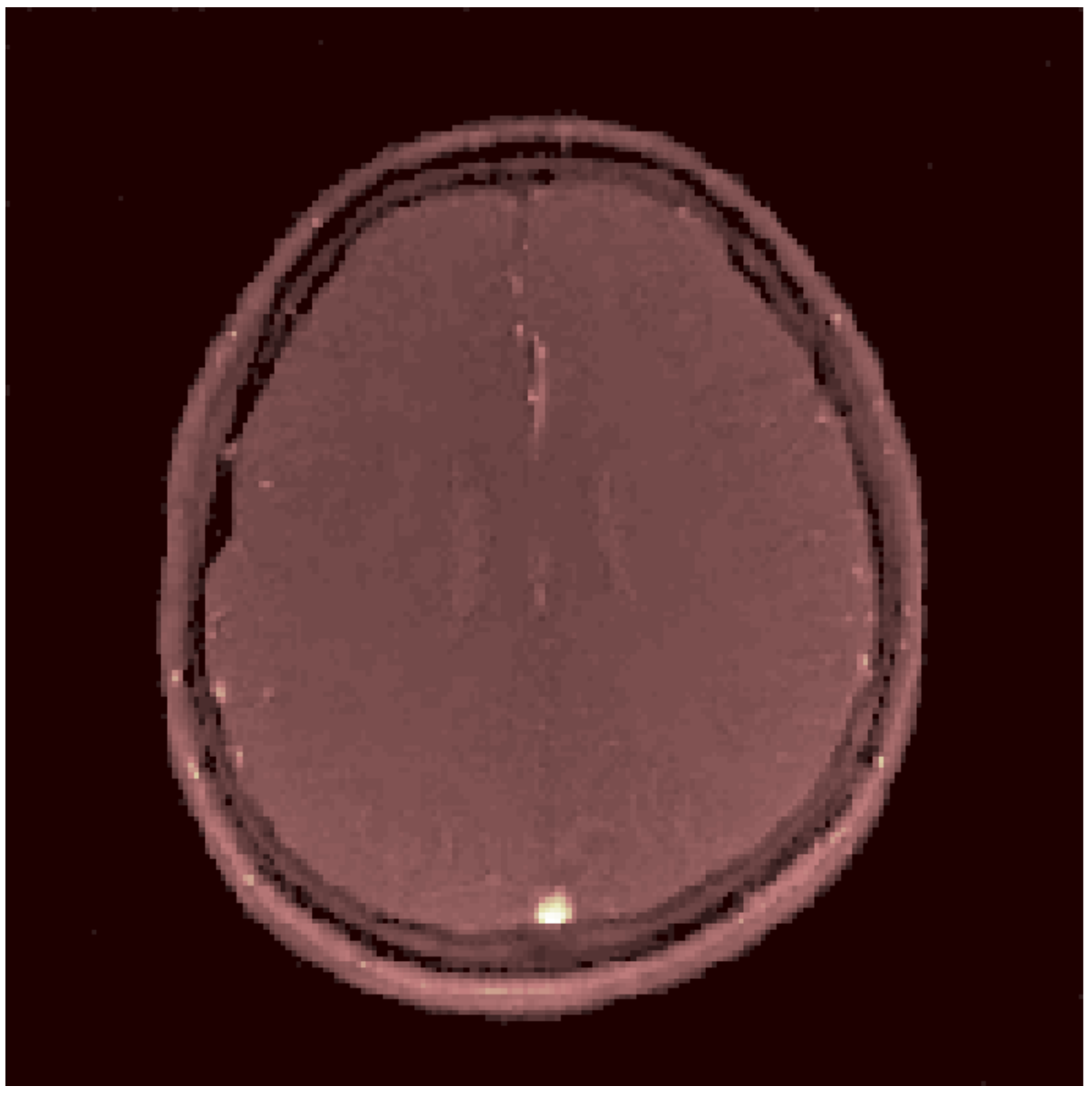}\hspace{-.1cm}
		\includegraphics[width=.162\linewidth]{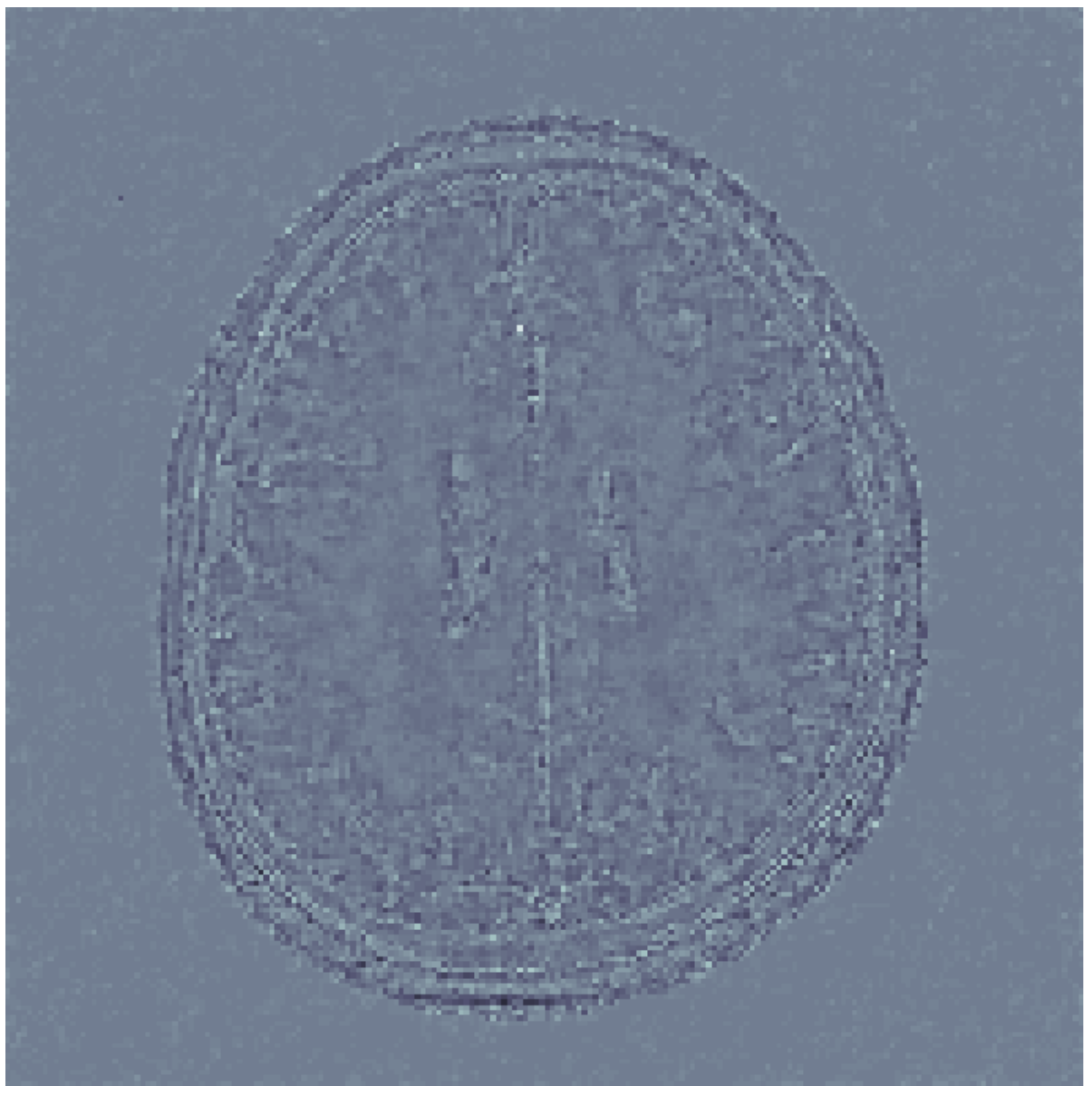}	\vspace{-.3cm}
		\\			
		\includegraphics[trim= -10 50 -20 720, clip,width=.17\linewidth]{./figs/retrospective/T1barvivo.jpg}\hspace{-.2cm}
		\includegraphics[trim= -10 50 -20 700, clip,width=.17\linewidth]{./figs/retrospective/dPDbarvivo.jpg}\hspace{-.2cm}
		\includegraphics[trim= -10 50 -20 700, clip,width=.17\linewidth]{./figs/retrospective/T2barvivo.jpg}\hspace{-.2cm}
		\includegraphics[trim= -10 50 -20 700, clip,width=.17\linewidth]{./figs/retrospective/dT2barvivo.jpg}\hspace{-.2cm}
		\includegraphics[trim= -10 50 -20 700, clip,width=.17\linewidth]{./figs/retrospective/PDbarvivo.jpg}\hspace{-.2cm}
		\includegraphics[trim= -10 50 -20 700, clip,width=.17\linewidth]{./figs/retrospective/dPDbarvivo.jpg}\vspace{.5cm}
		\\
		 .\hspace{1cm} T1(sec) \hspace{1.7cm} T1 error  \hspace{1.7cm} T2 (sec) \hspace{1.7cm} T2 error \hspace{1.7cm} PD (a.u.) \hspace{1.7cm} PD error 
		\caption{The computed T1, T2, PD maps and their corresponding errors (with respect to MAGIC gold-standard) using 2D radial k-space sampling, different reconstruction baselines and our proposed LRTV-MRFResnet algorithm. 
  \label{fig:radial_recon_retro}}
\end{minipage}}
\end{figure*}

Results are consistent with those obtained in previous experiments. KM despite great T1 accuracy outputs inaccurate T2/PD predictions i.e. overestimated T2 and underestimated PD maps. Due to the extremely low single coil k-space data for view sharing, VS introduces strong bias on the T1/T2 maps. 
Further, temporal priors used by LR and FLOR are insufficient to reject under-sampling artefacts, and FLOR can introduce bias in the estimated maps. AIR-MRF's spatial low-pass filtering trades off the sharpness of the computed maps meanwhile unsuccessful to fully remove aliasing artefacts (Figure~\ref{fig:spiral_recon_retro}). On the other hand, the spatiotemporally regularised LRTV significantly improves TSMI reconstructions (e.g. 4 dB improvement with respect to the closest competitor baseline, Table~IV) 
through successfully removing strong aliasing artefacts (see Figure~\ref{fig:spiral_recon_retro} and \ref{fig:radial_recon_retro}). This enables accurate parameter inference in the next stage using DM or the DM-free alternative MRFResnet. 

\section{\textit{in-vitro} Phantom reconstructed maps}
\begin{figure*}
	\centering
	\begin{minipage}{\textwidth}
		\centering
		\includegraphics[width=.16\linewidth]{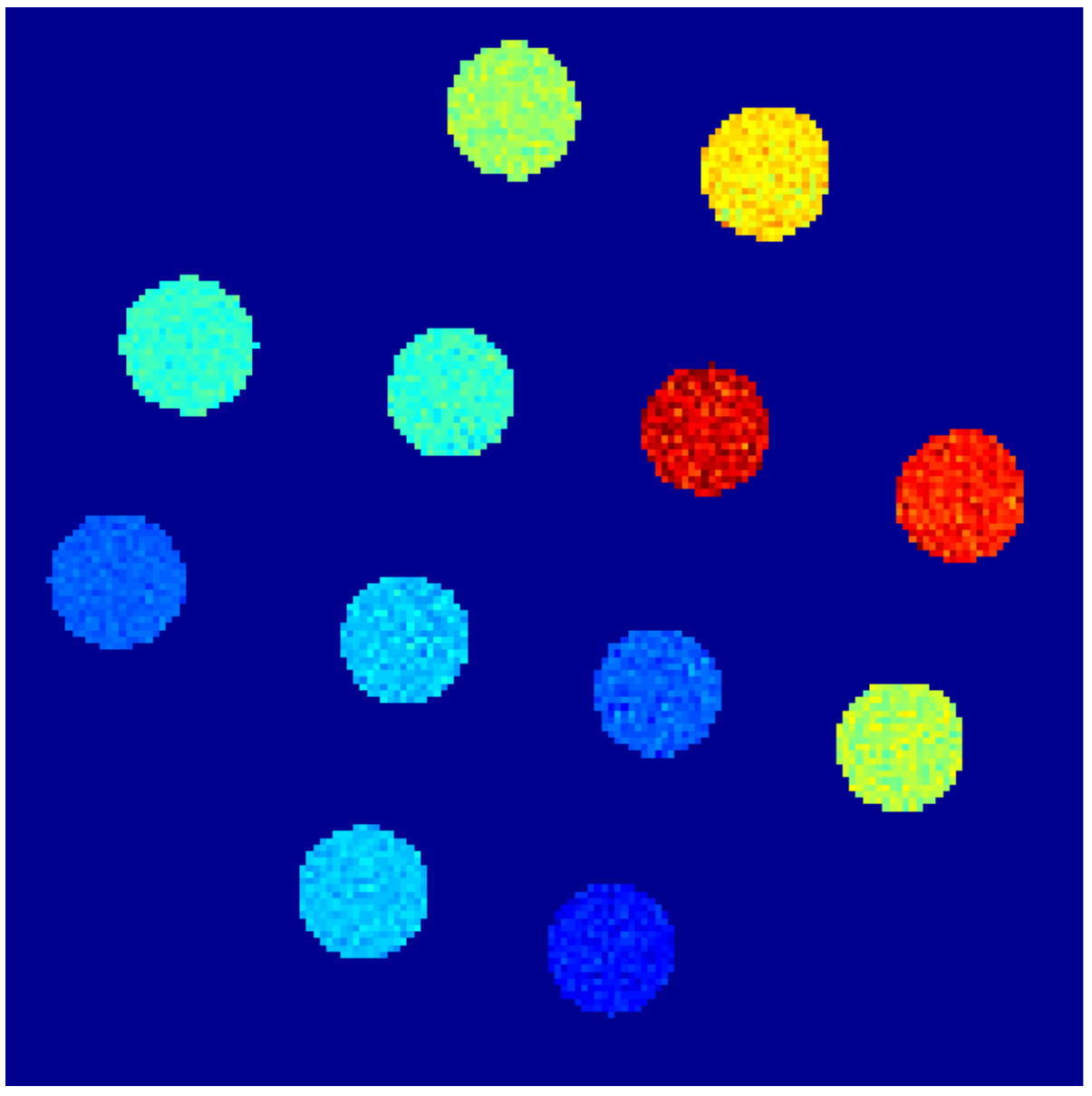} \hspace{-.25cm}	
		\includegraphics[width=.16\linewidth]{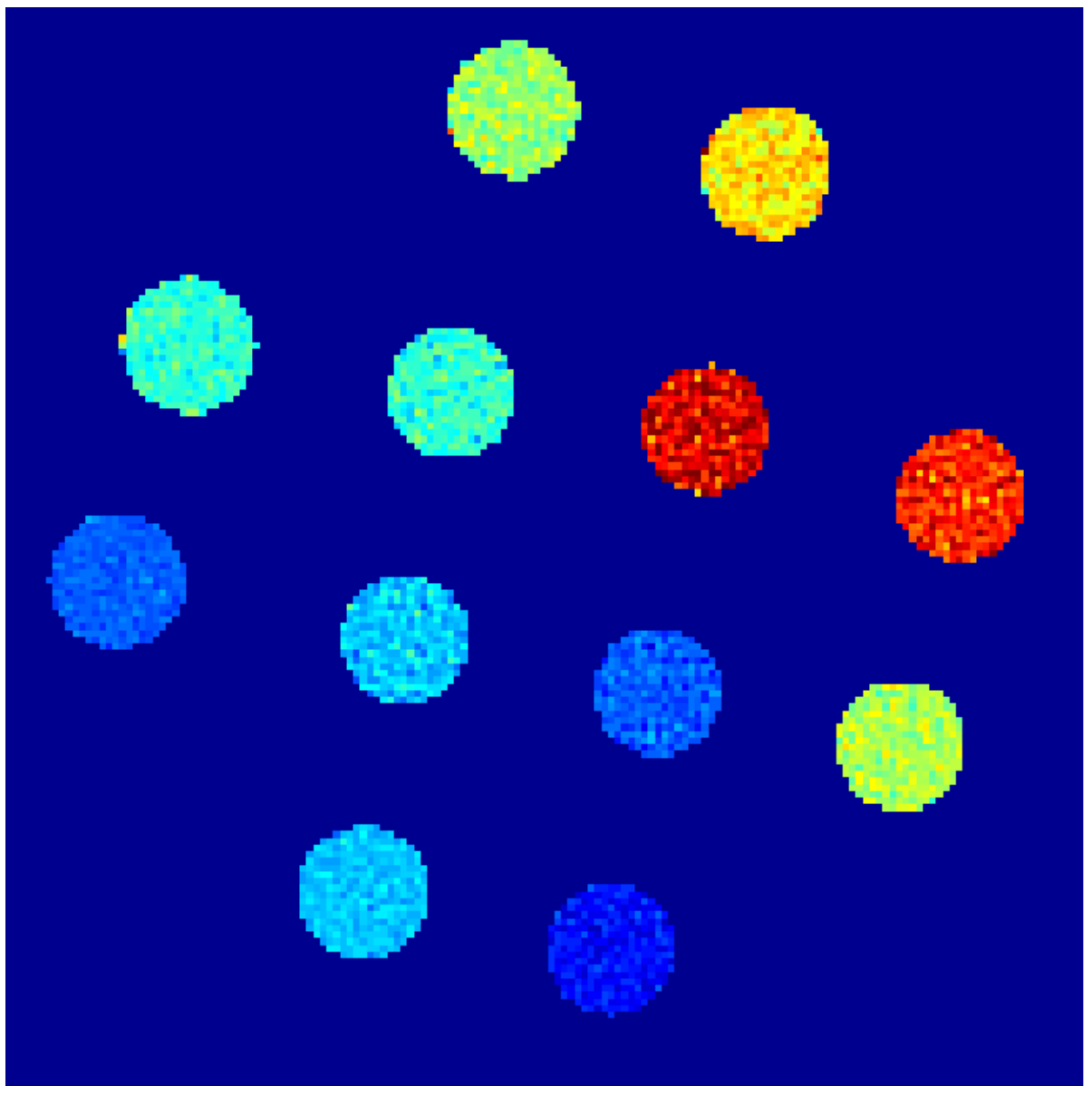} \hspace{-.25cm}		
		\includegraphics[width=.16\linewidth]{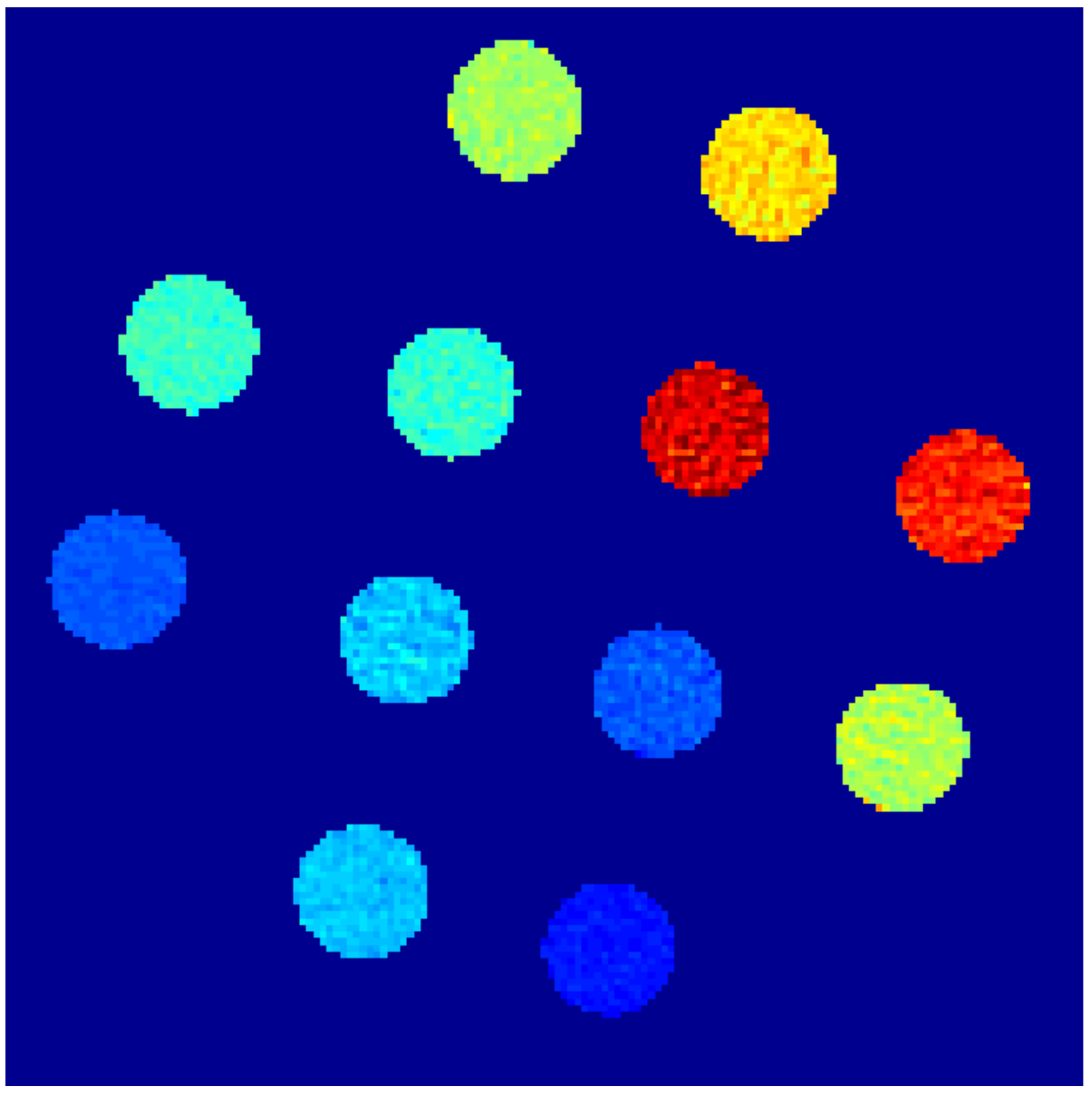} 		\hspace{.2cm}
		\includegraphics[width=.16\linewidth]{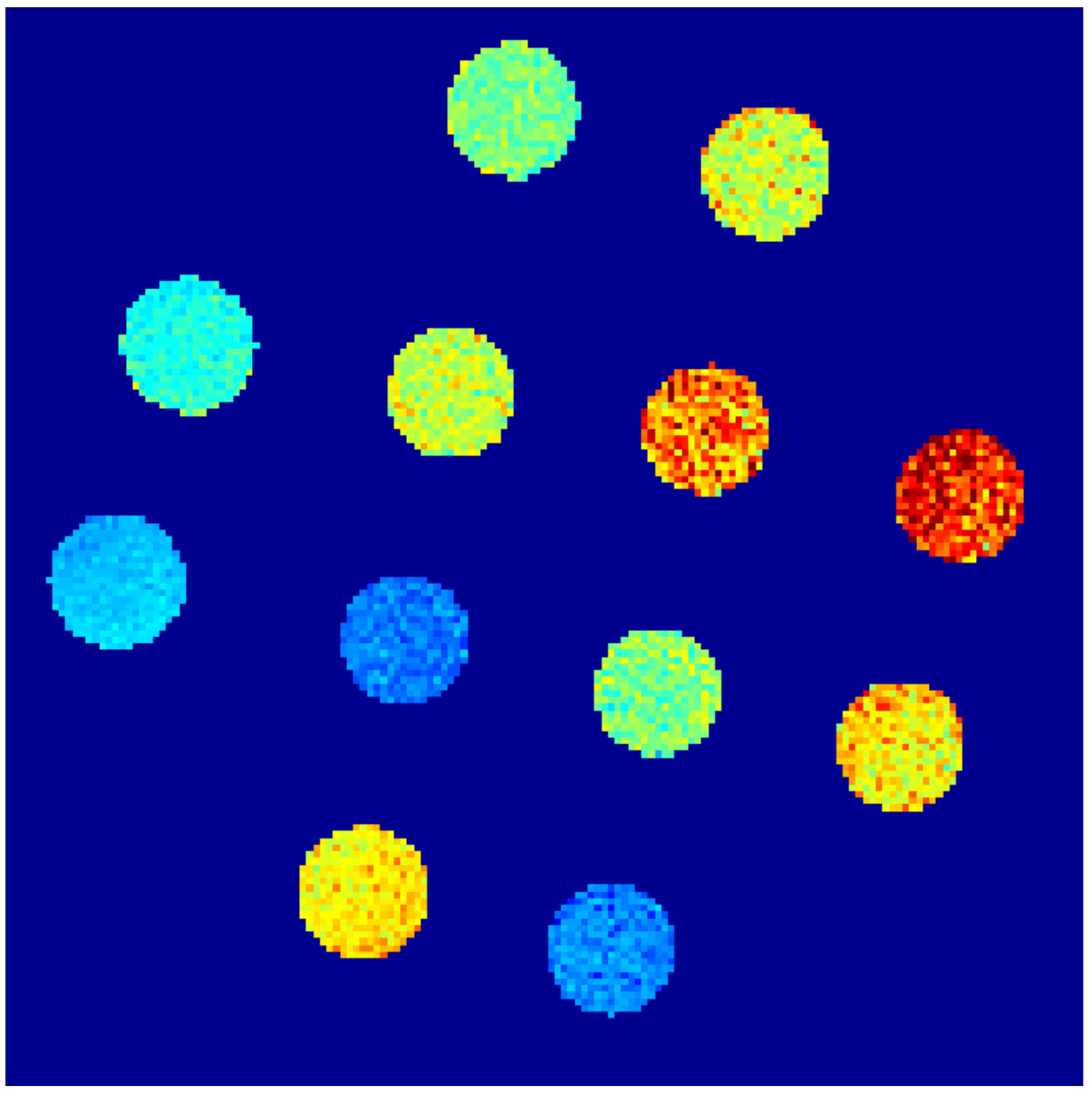} 		\hspace{-.25cm}
		\includegraphics[width=.16\linewidth]{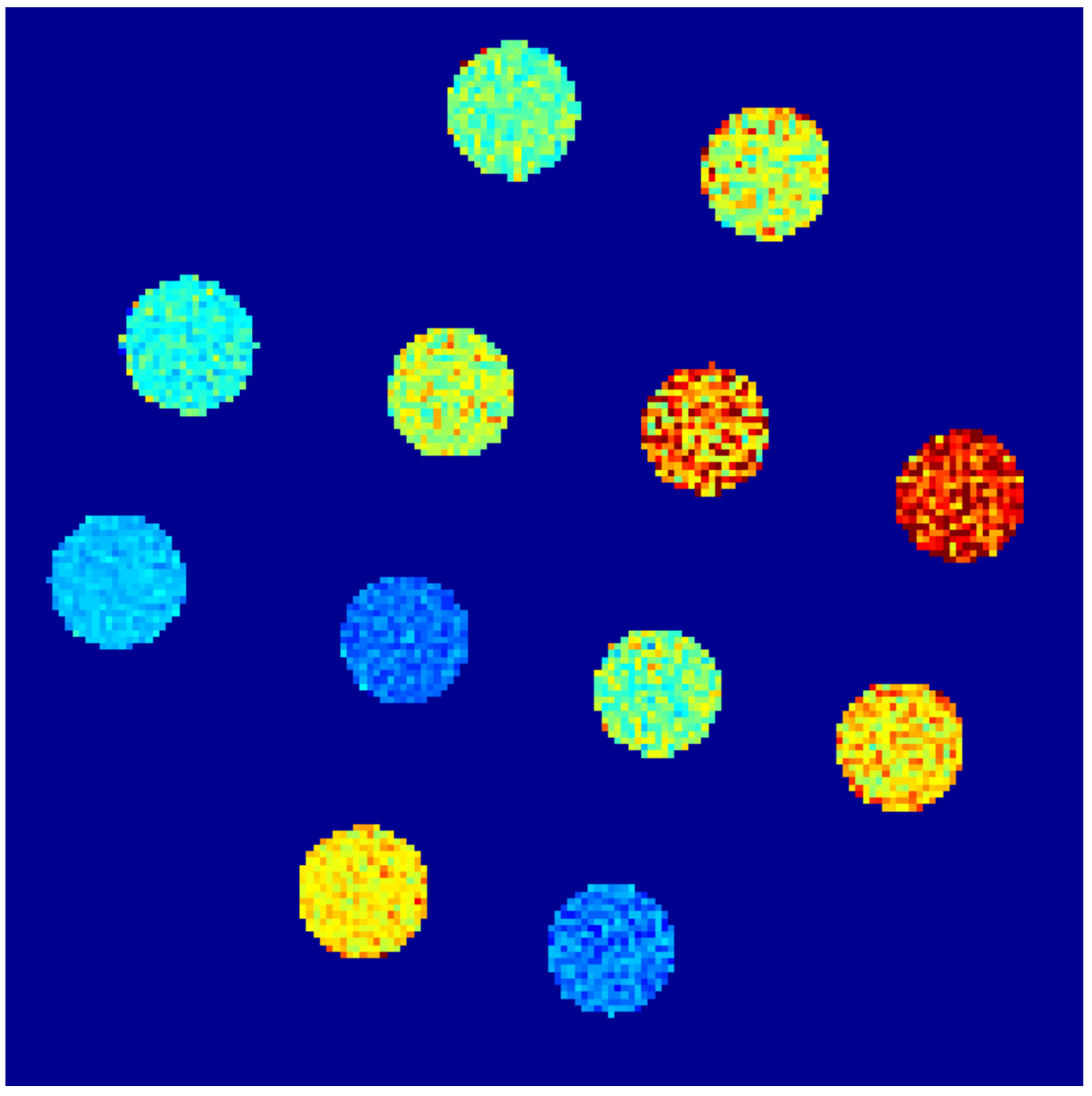} 		\hspace{-.25cm}
		\includegraphics[width=.16\linewidth]{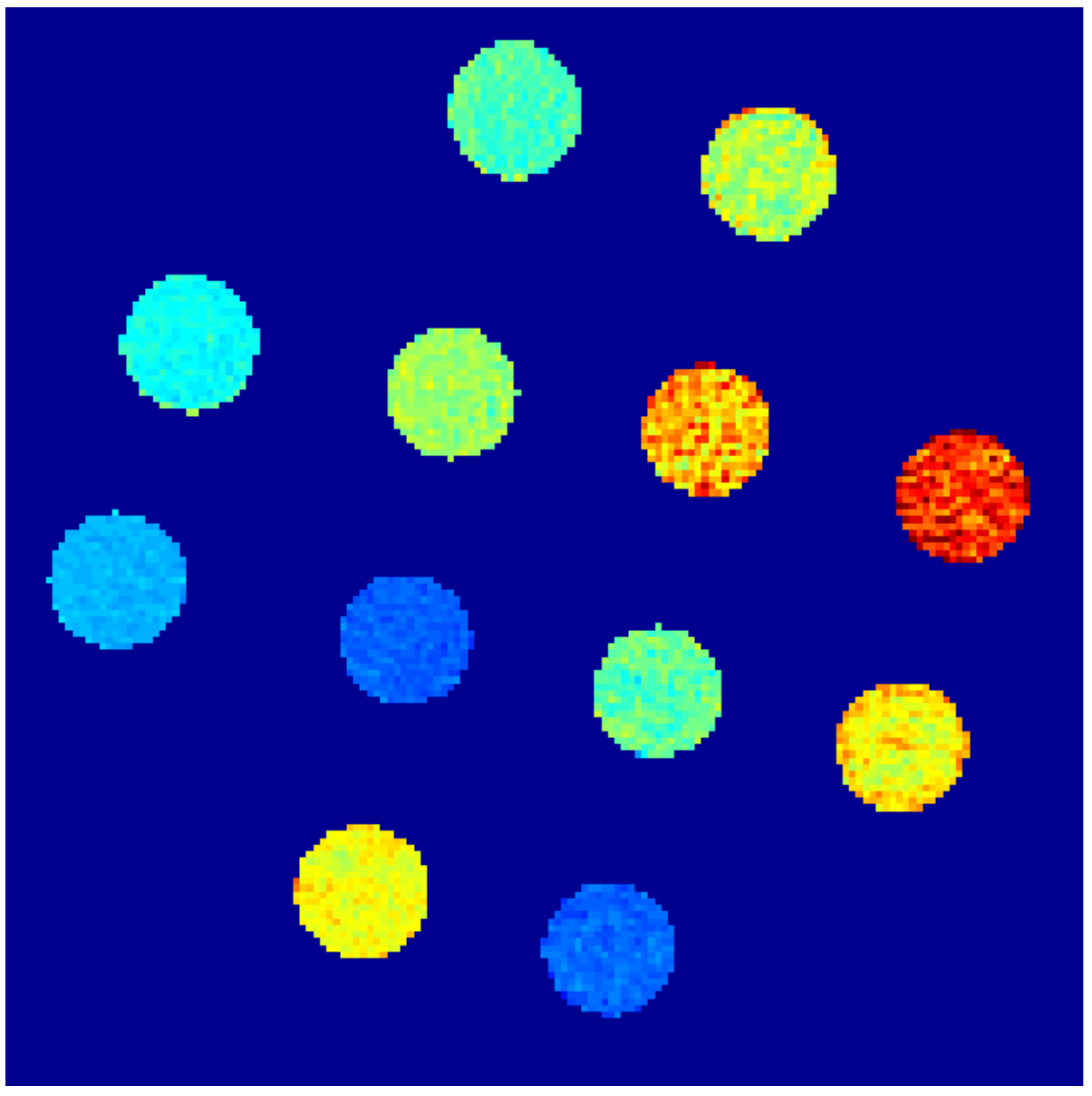} 	
		\\		
		\includegraphics[width=.16\linewidth]{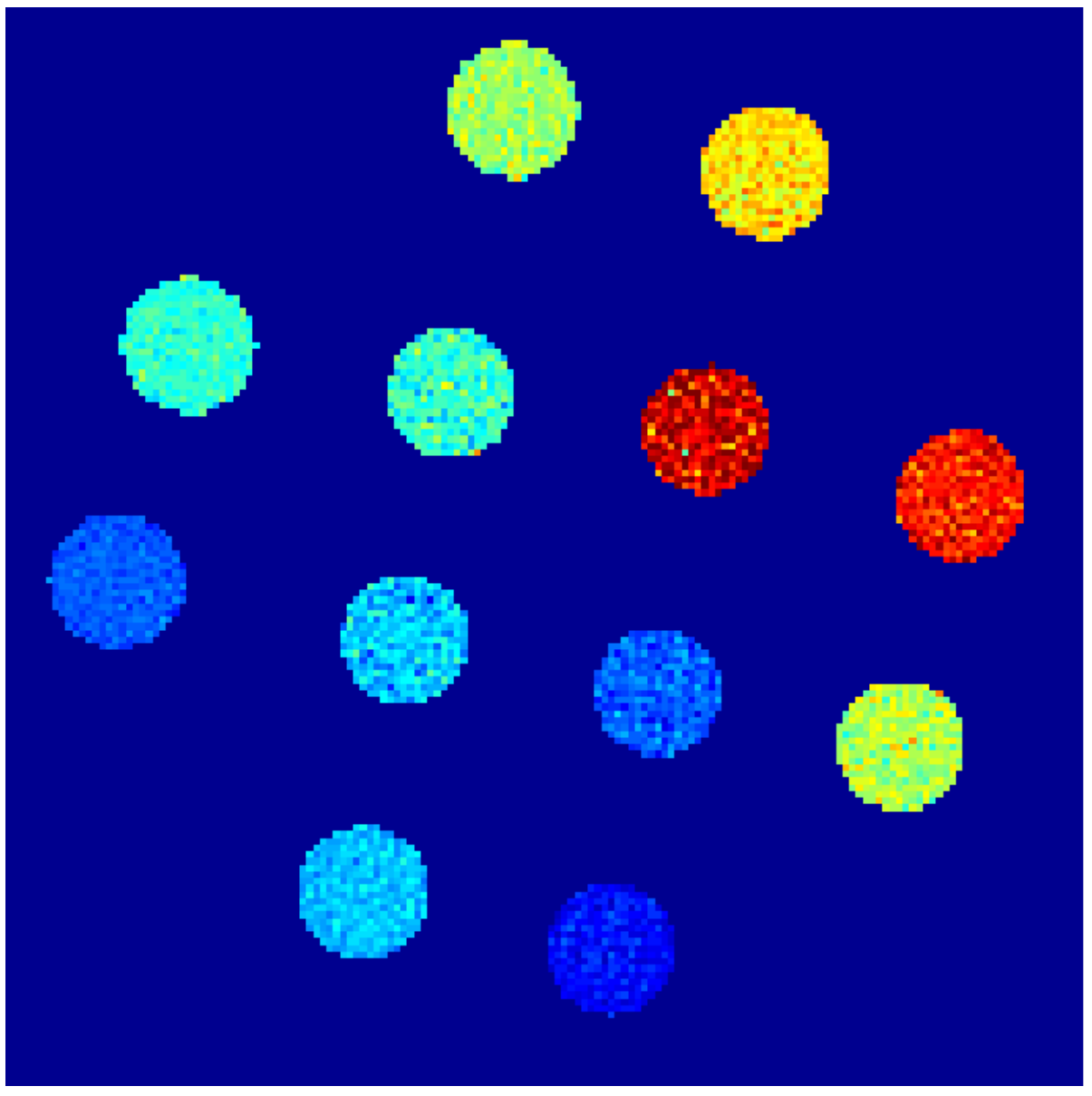} 	\hspace{-.25cm}	
		\includegraphics[width=.16\linewidth]{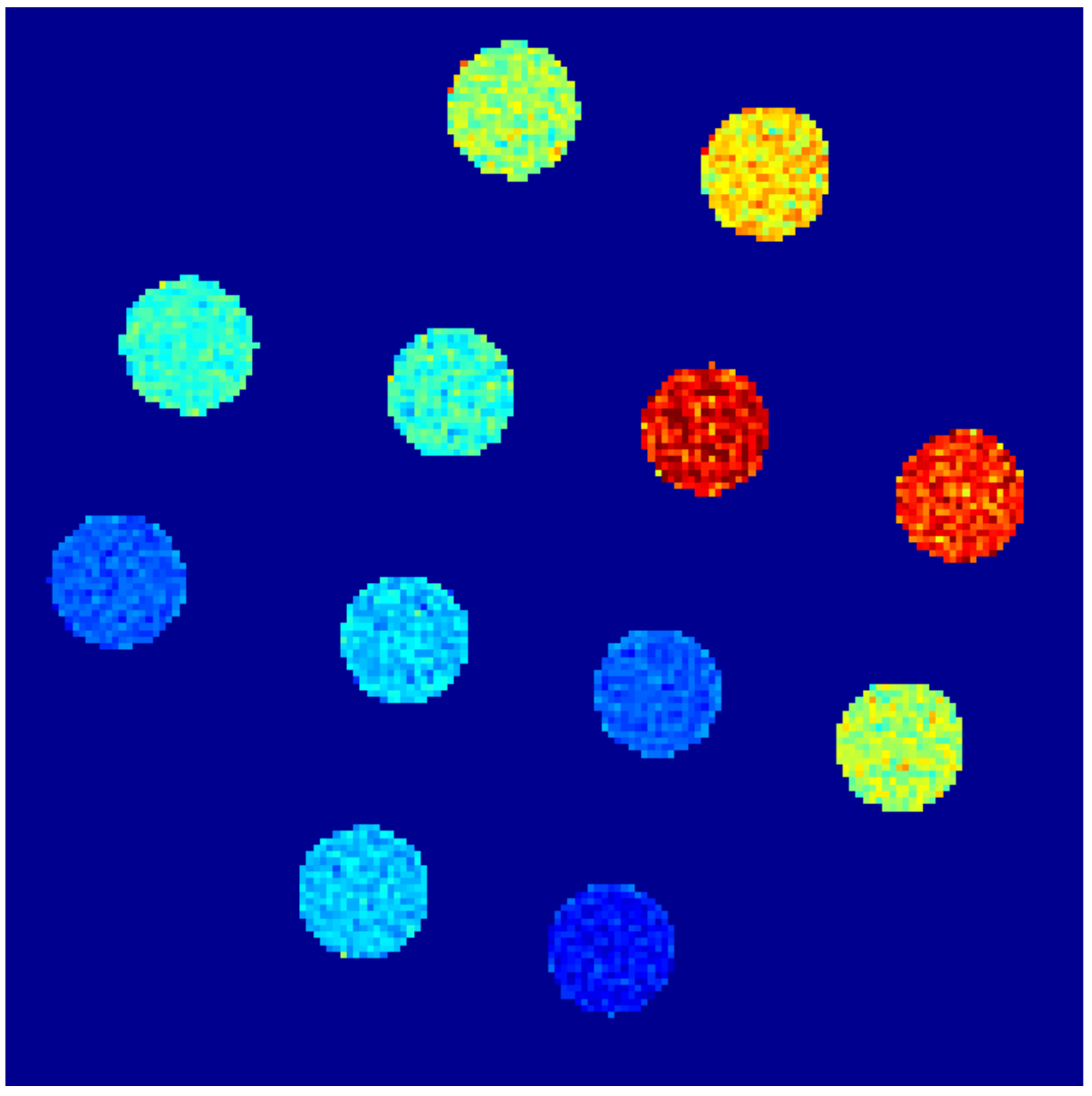} 		\hspace{-.25cm}	
		\includegraphics[width=.16\linewidth]{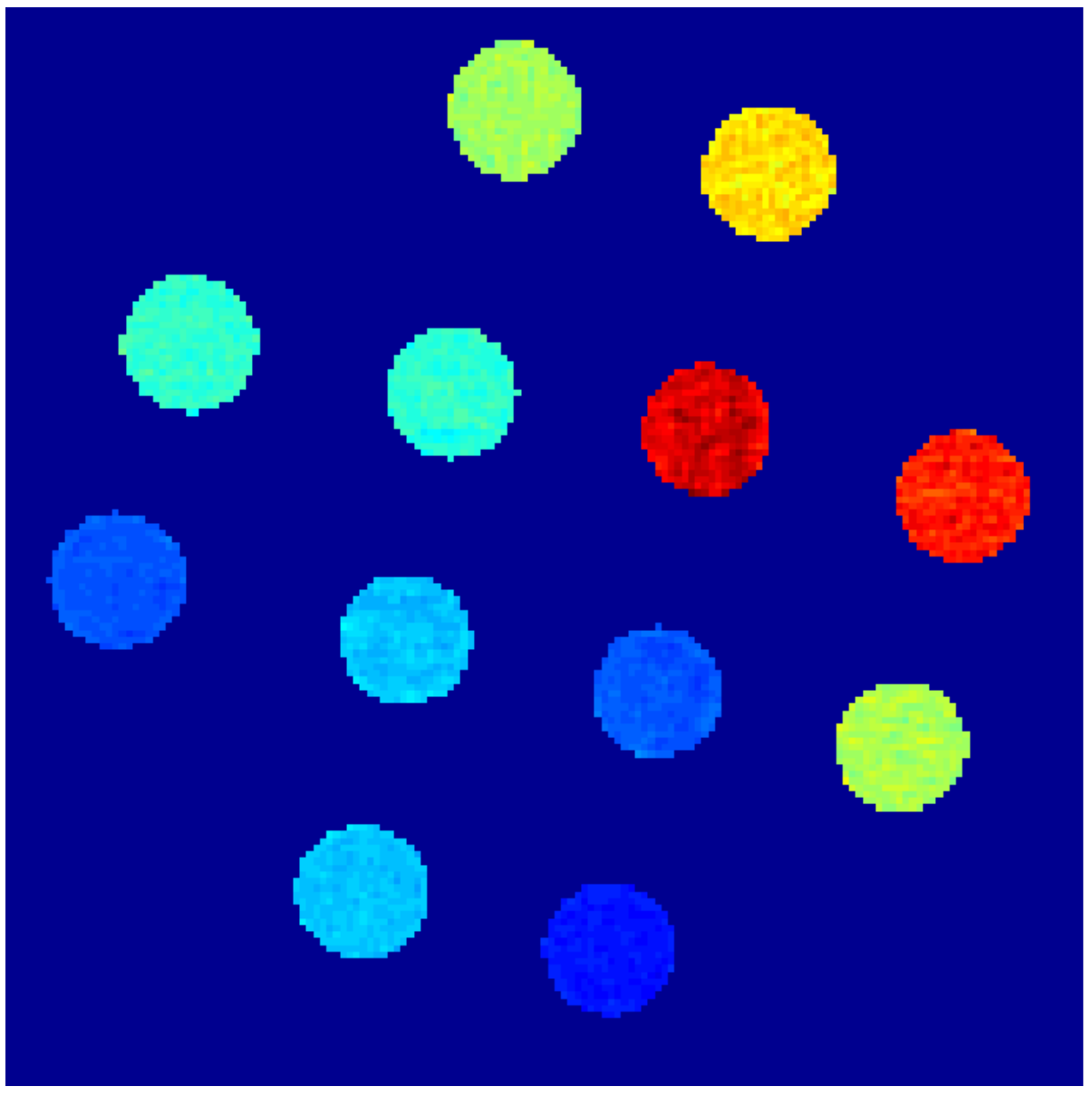} 		\hspace{.2cm}
		\includegraphics[width=.16\linewidth]{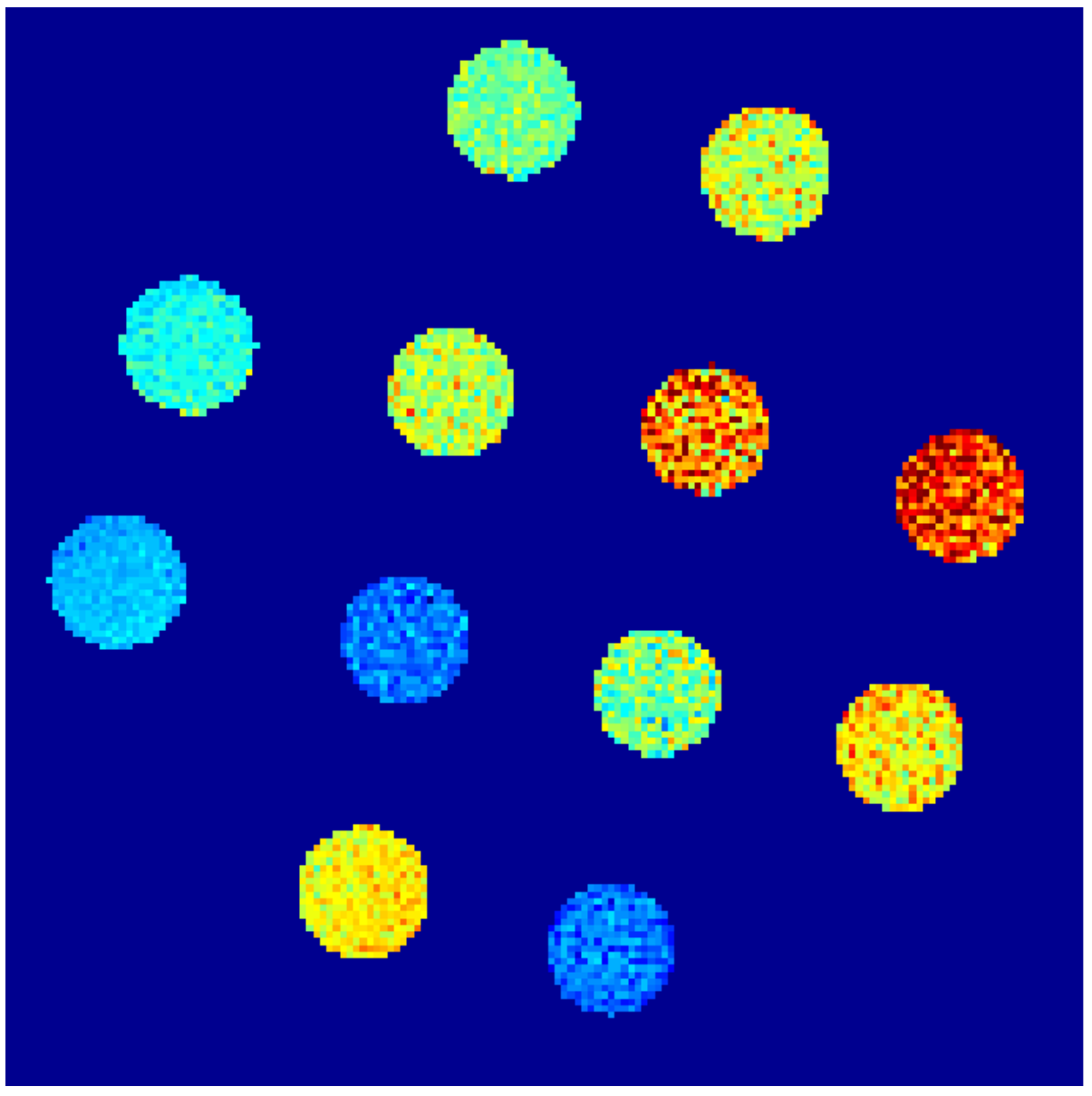} 		\hspace{-.25cm}	
		\includegraphics[width=.16\linewidth]{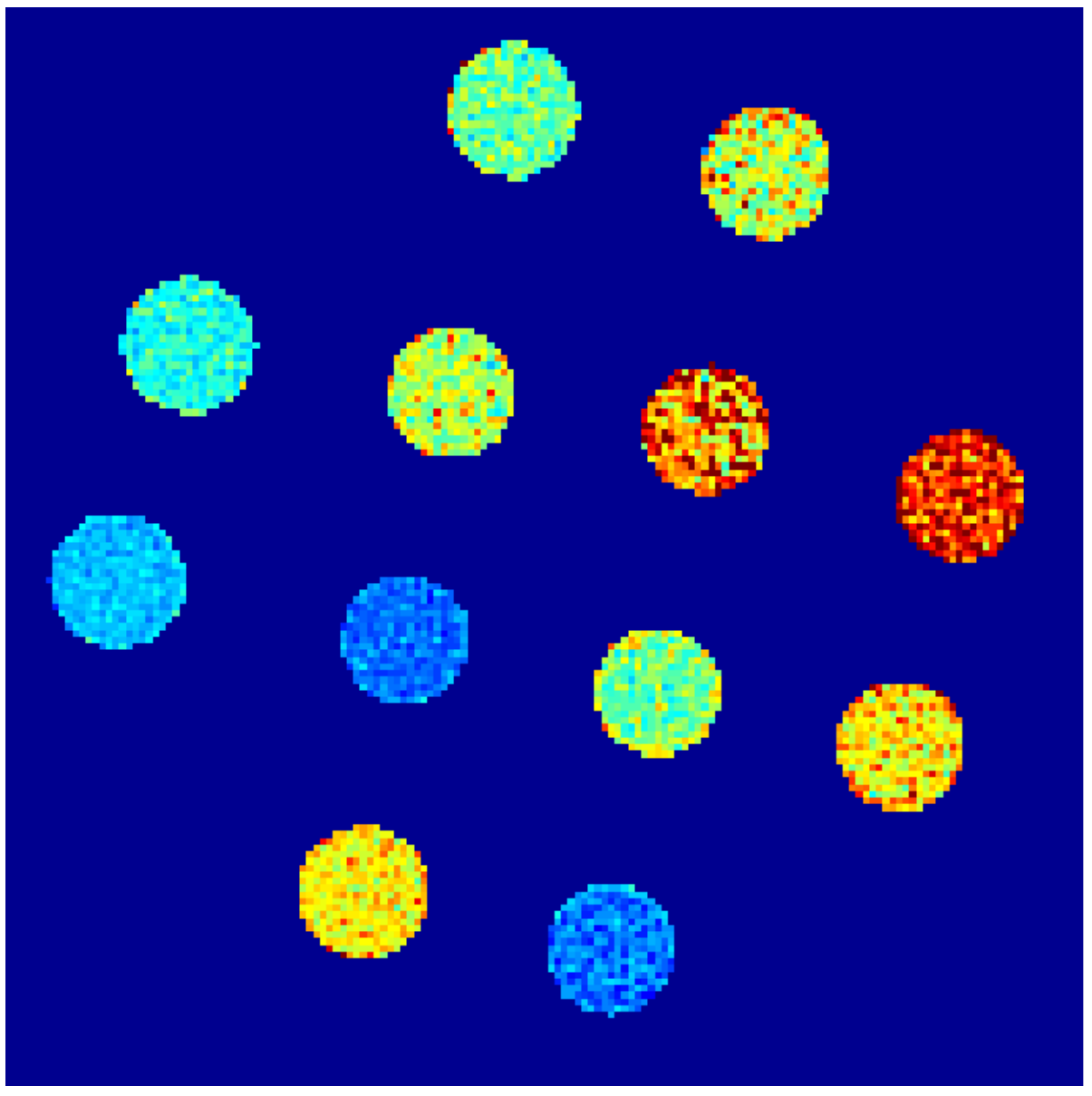} 		\hspace{-.25cm}	
		\includegraphics[width=.16\linewidth]{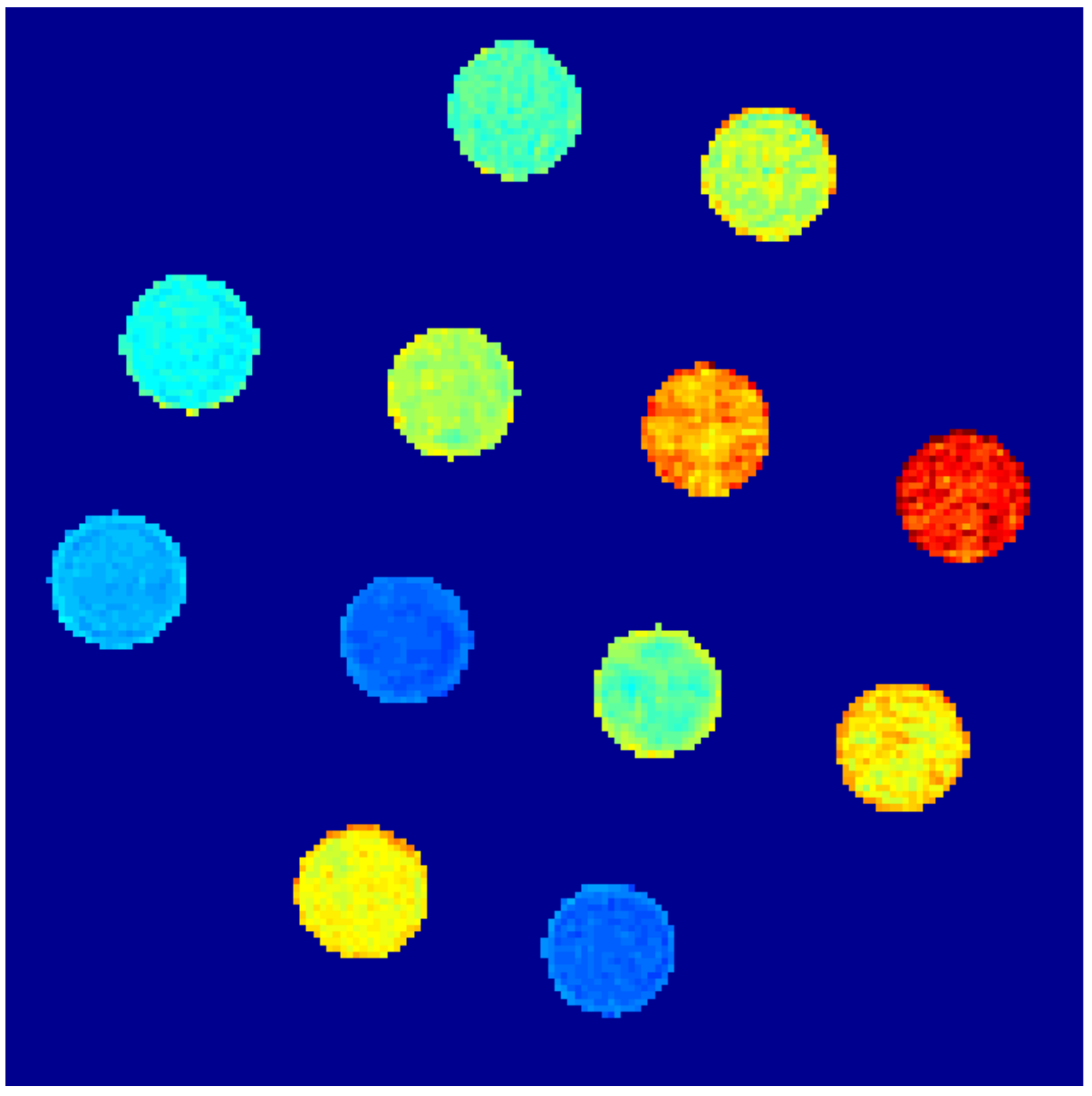} 	
		\\
		\includegraphics[width=.16\linewidth]{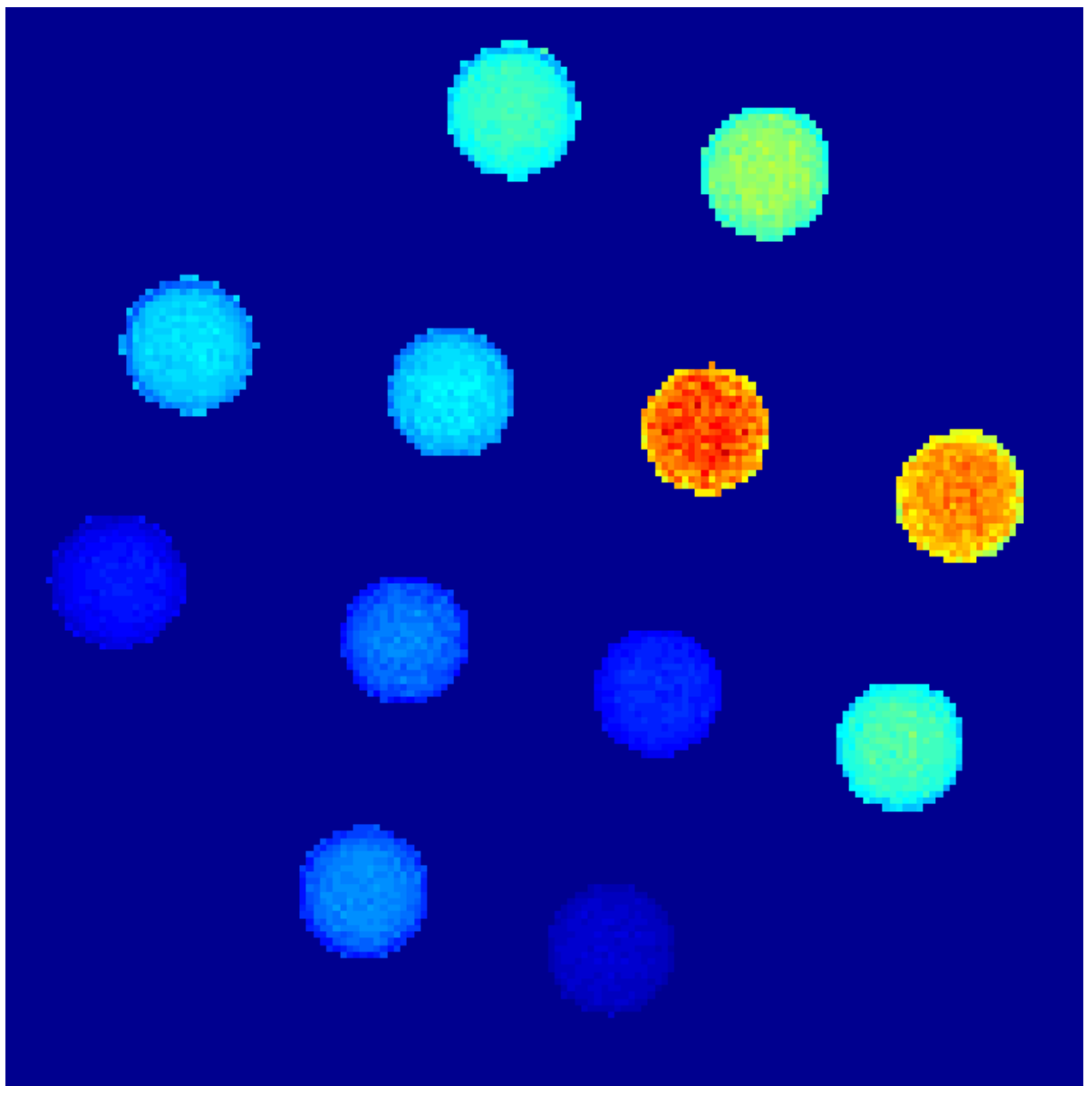} 	\hspace{-.25cm}	
		\includegraphics[width=.16\linewidth]{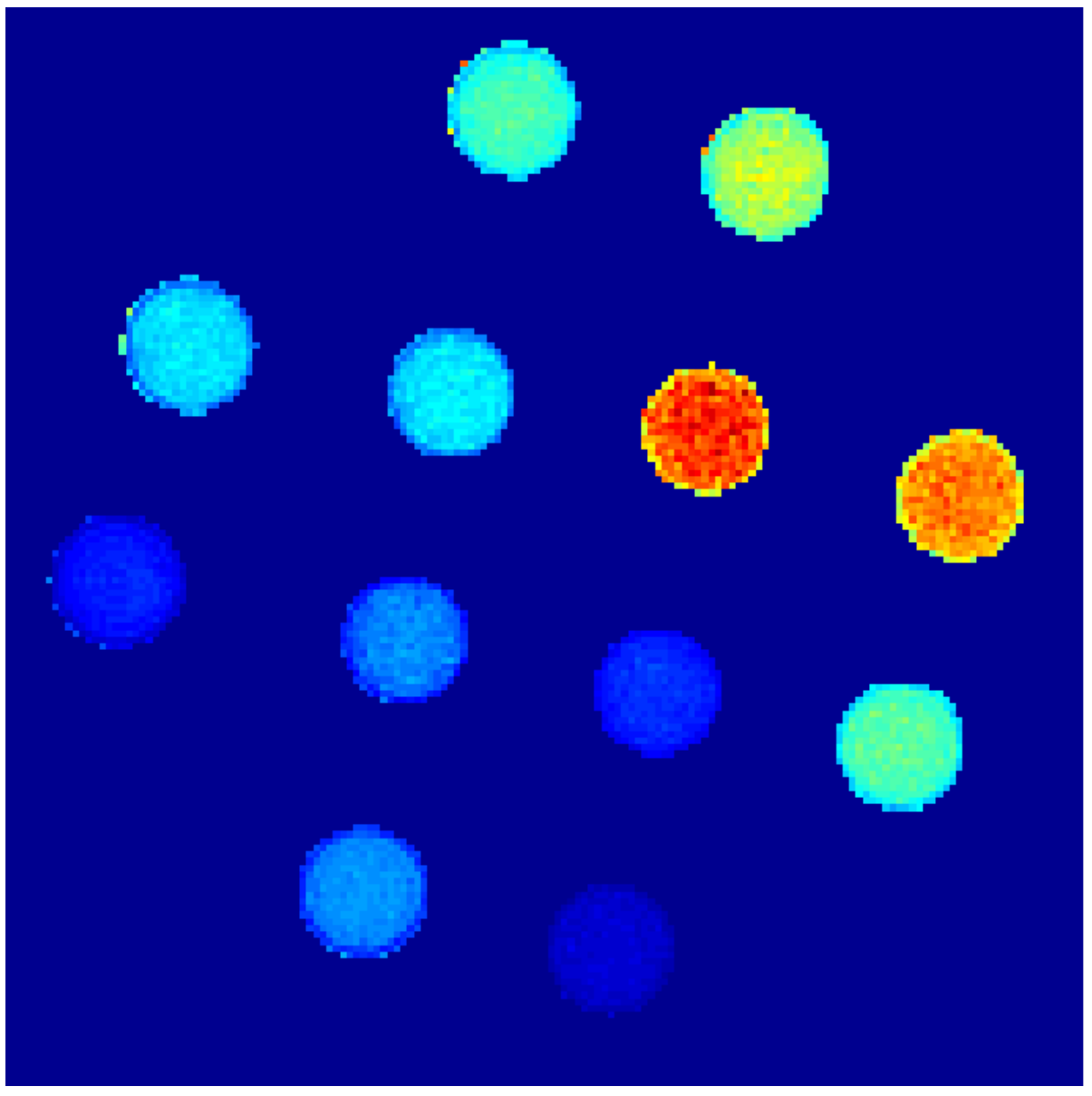} 		\hspace{-.25cm}	
		\includegraphics[width=.16\linewidth]{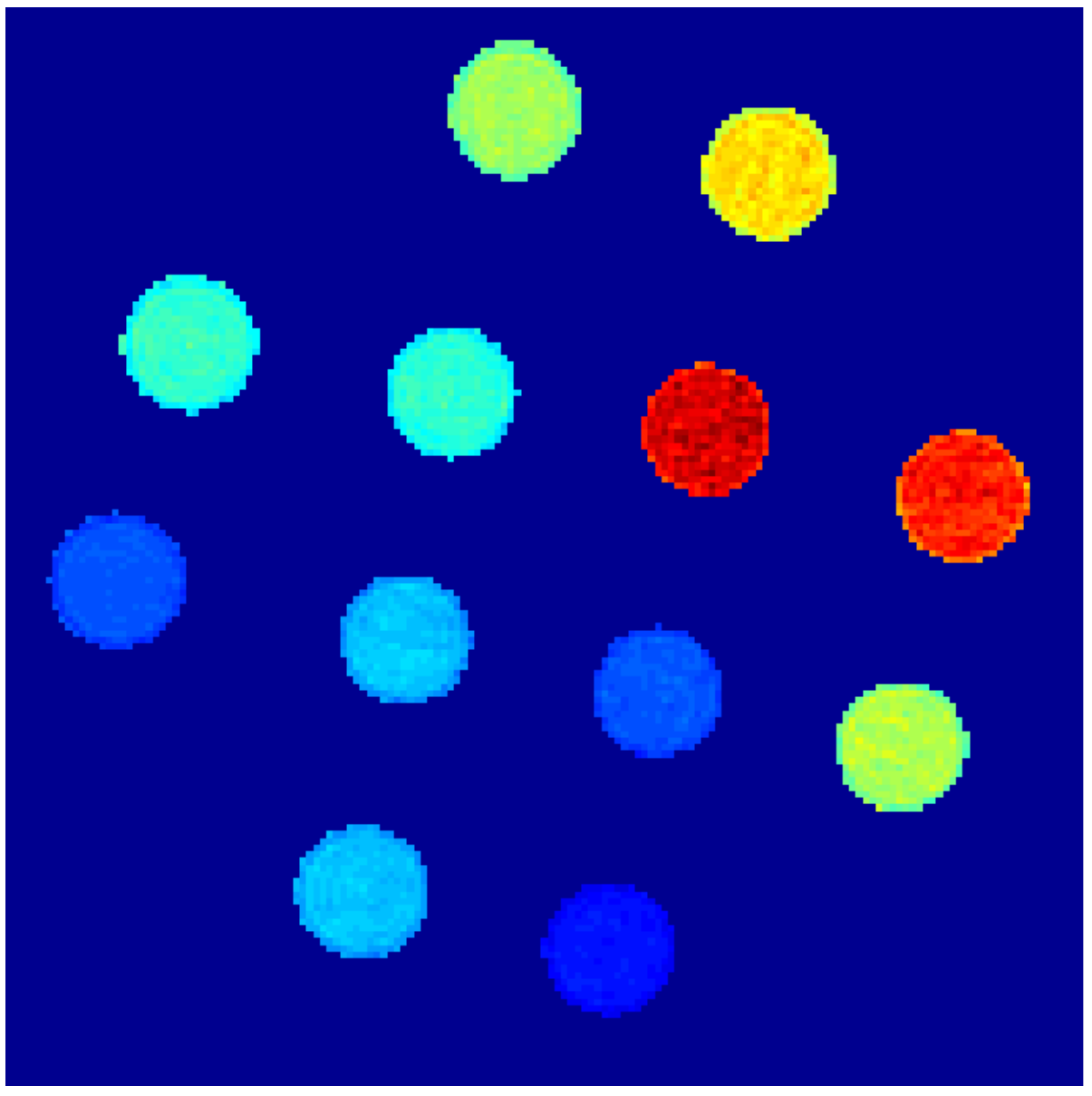} 		\hspace{.2cm}
		\includegraphics[width=.16\linewidth]{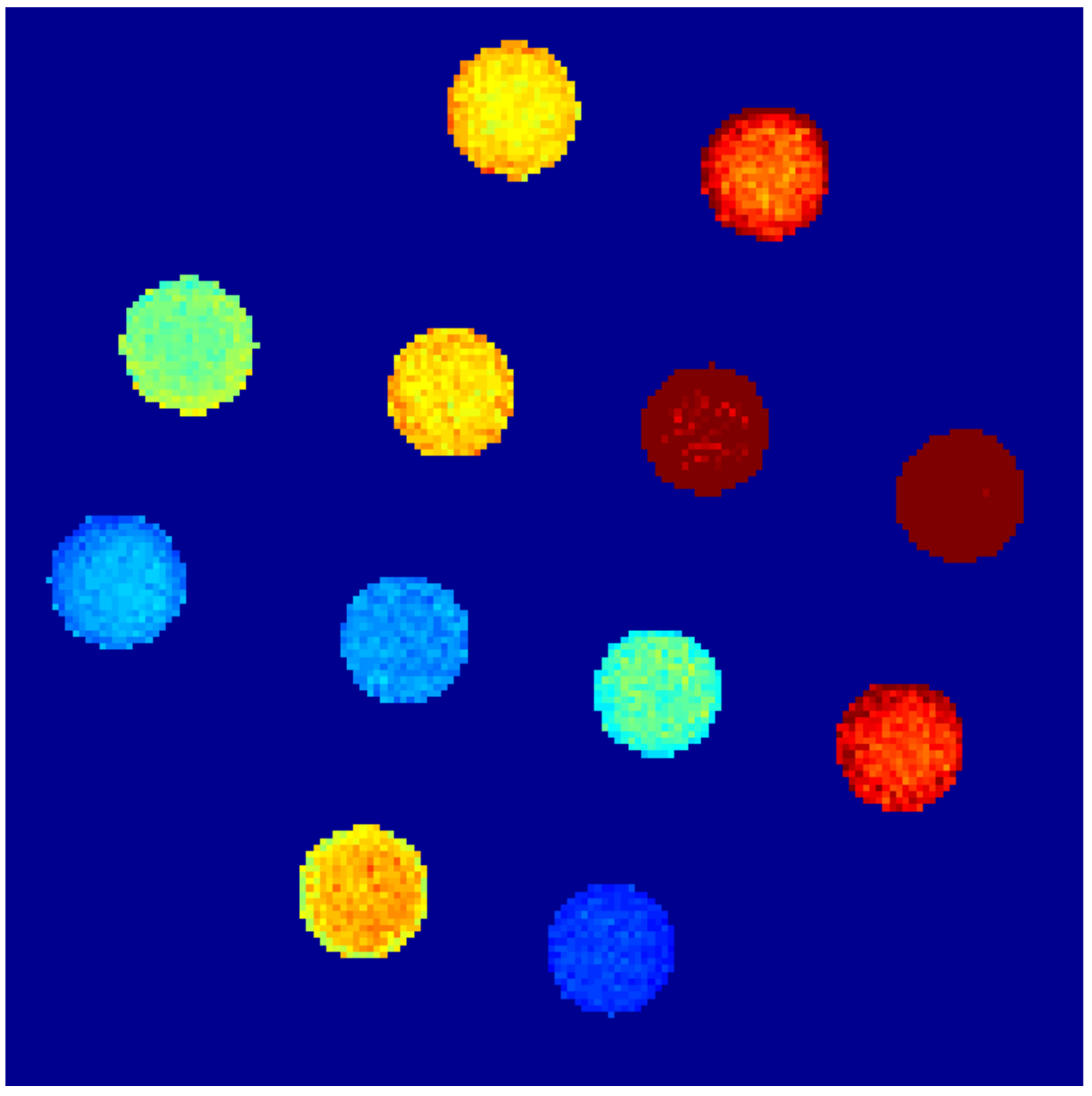} 		\hspace{-.25cm}	
		\includegraphics[width=.16\linewidth]{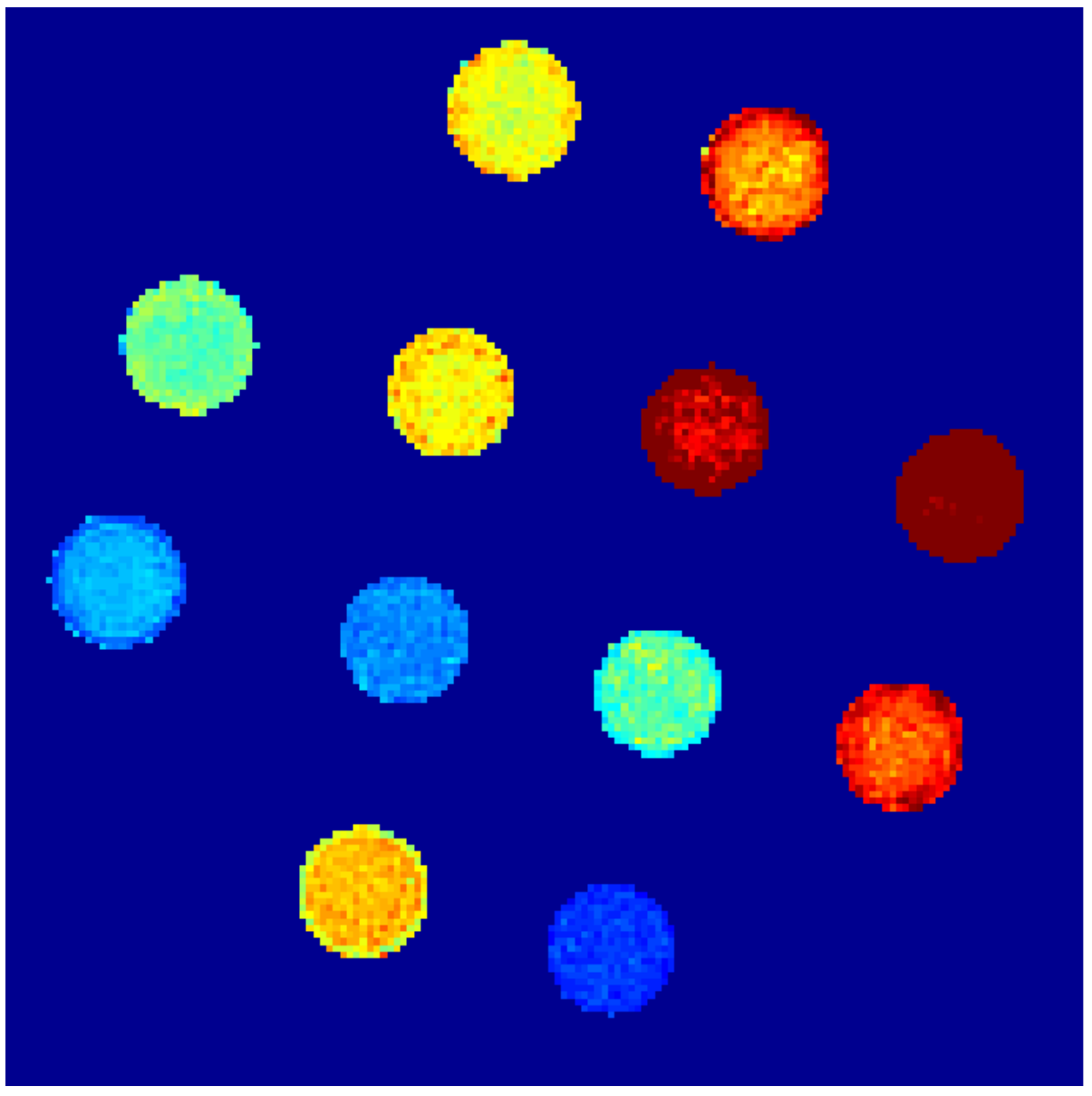} 		\hspace{-.25cm}	
		\includegraphics[width=.16\linewidth]{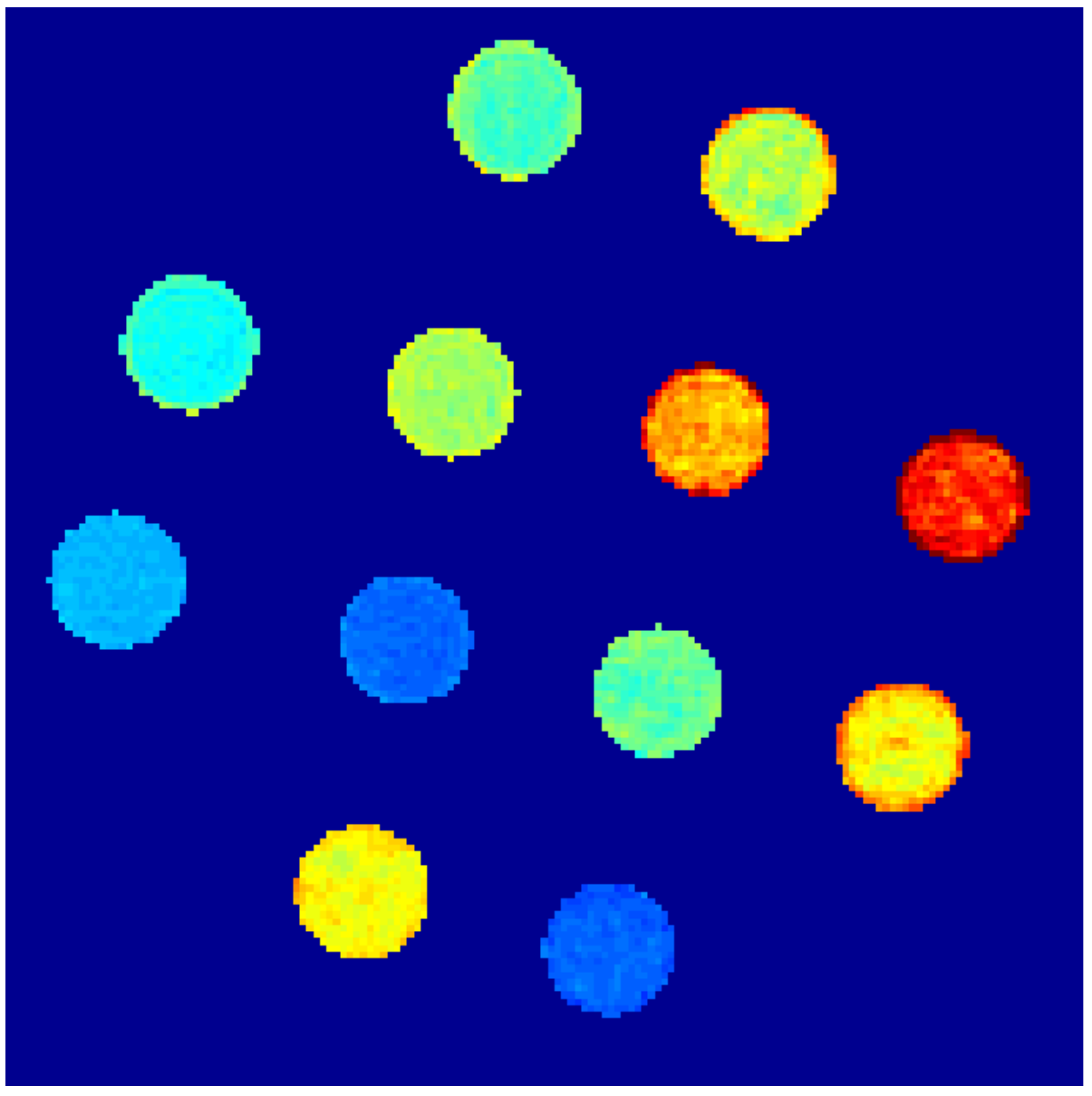} 
			\\
		\includegraphics[width=.16\linewidth]{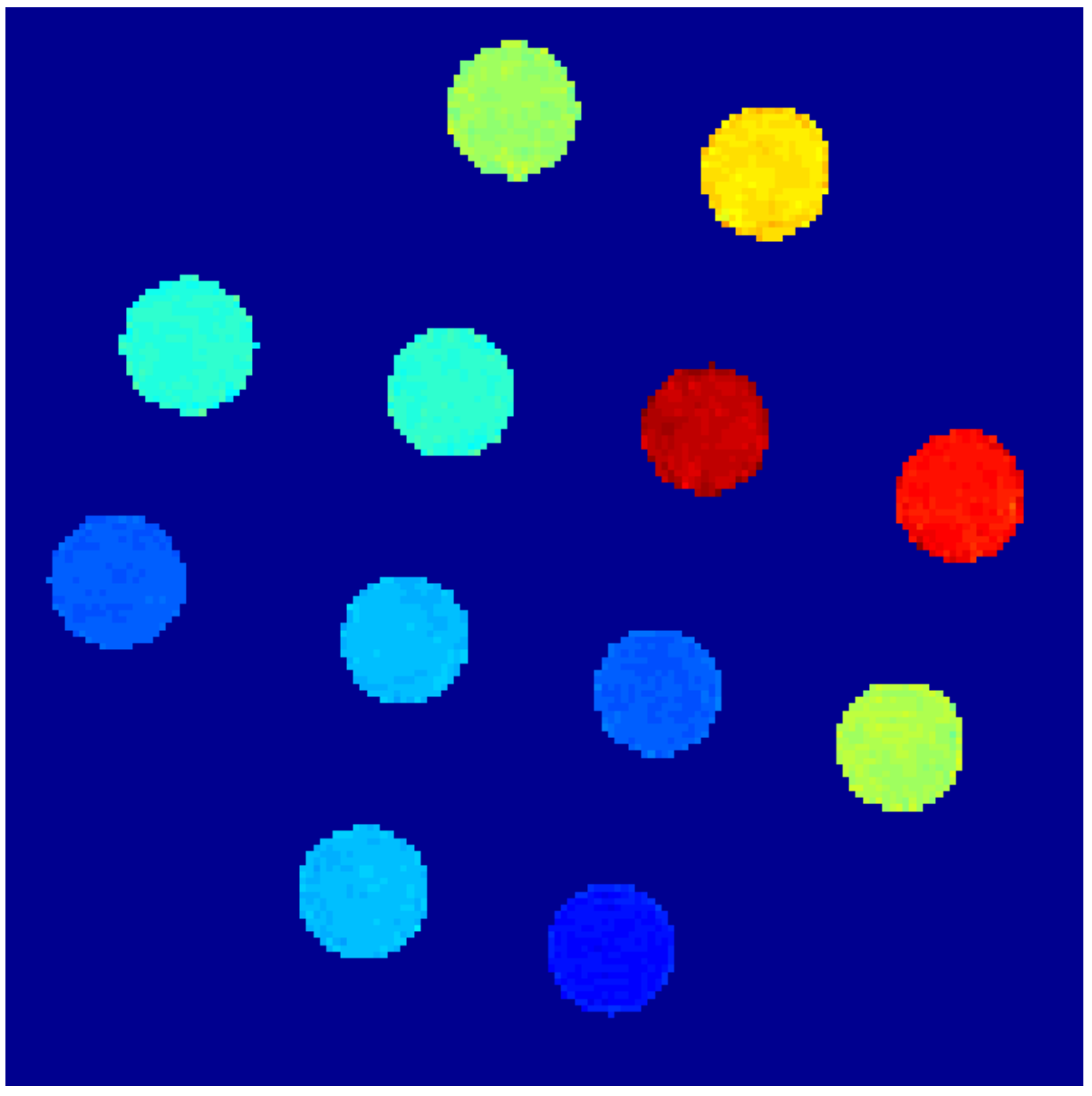} 	\hspace{-.25cm}	
		\includegraphics[width=.16\linewidth]{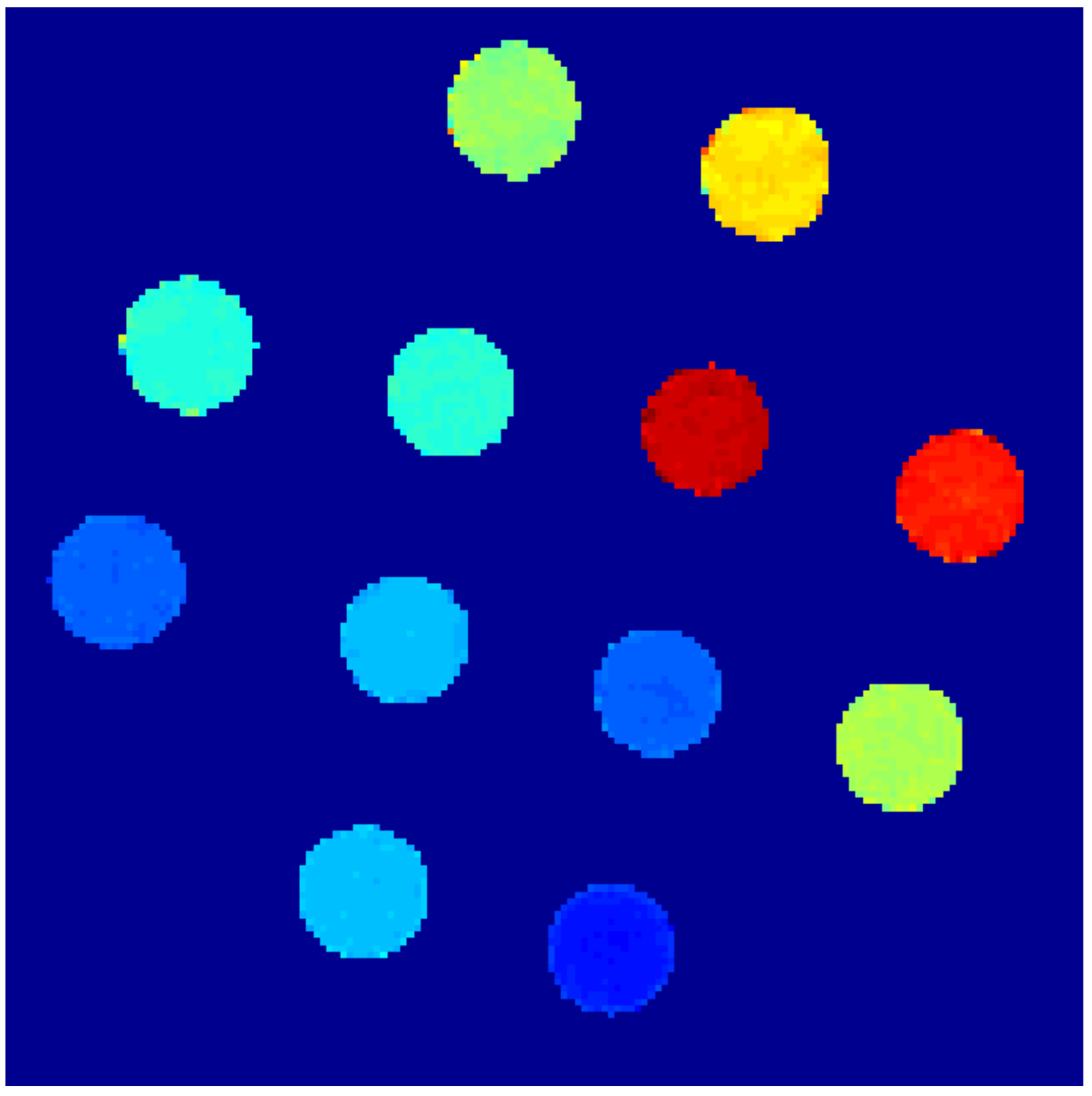} 		\hspace{-.25cm}	
		\includegraphics[width=.16\linewidth]{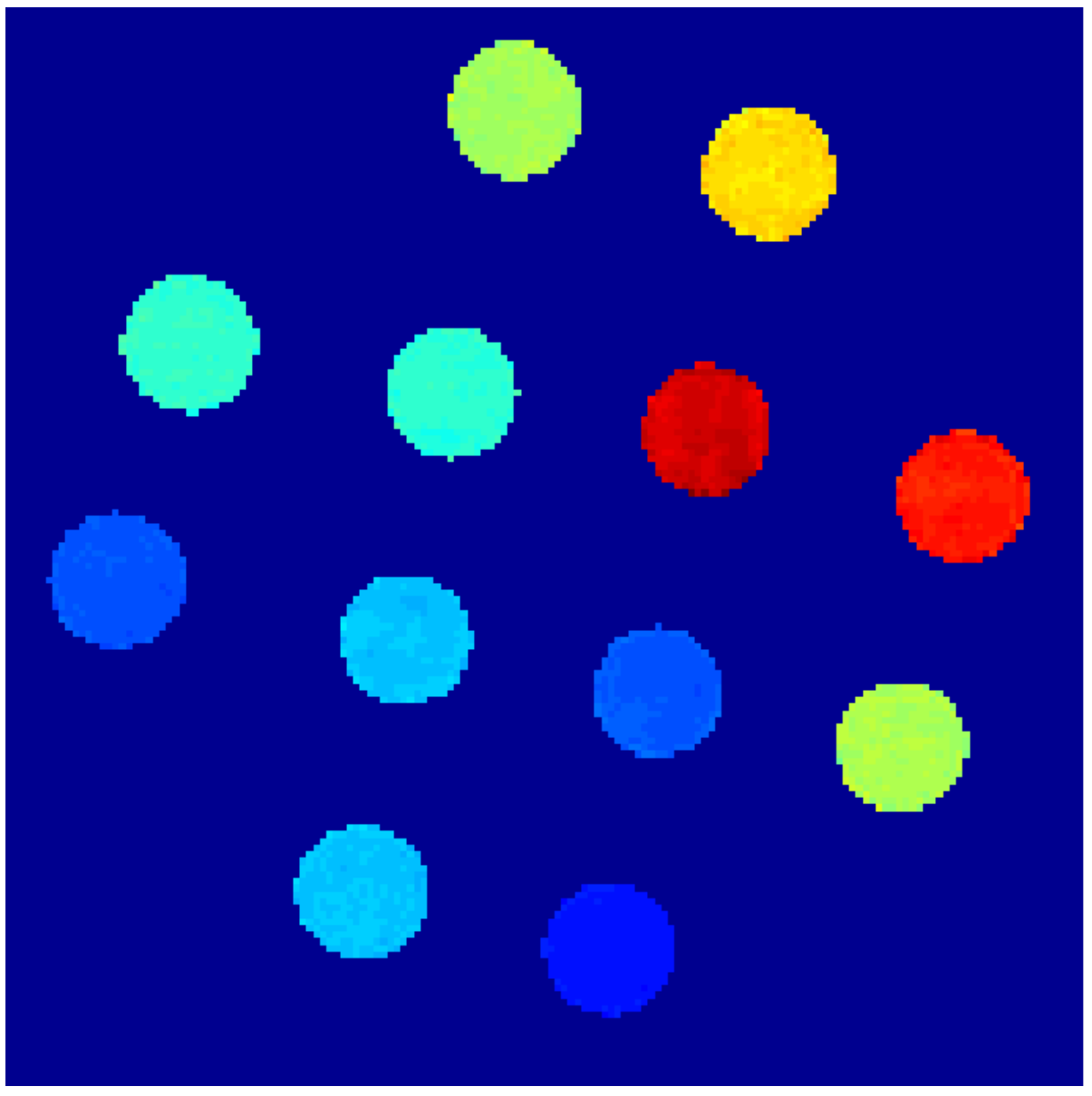} 		\hspace{.2cm}
		\includegraphics[width=.16\linewidth]{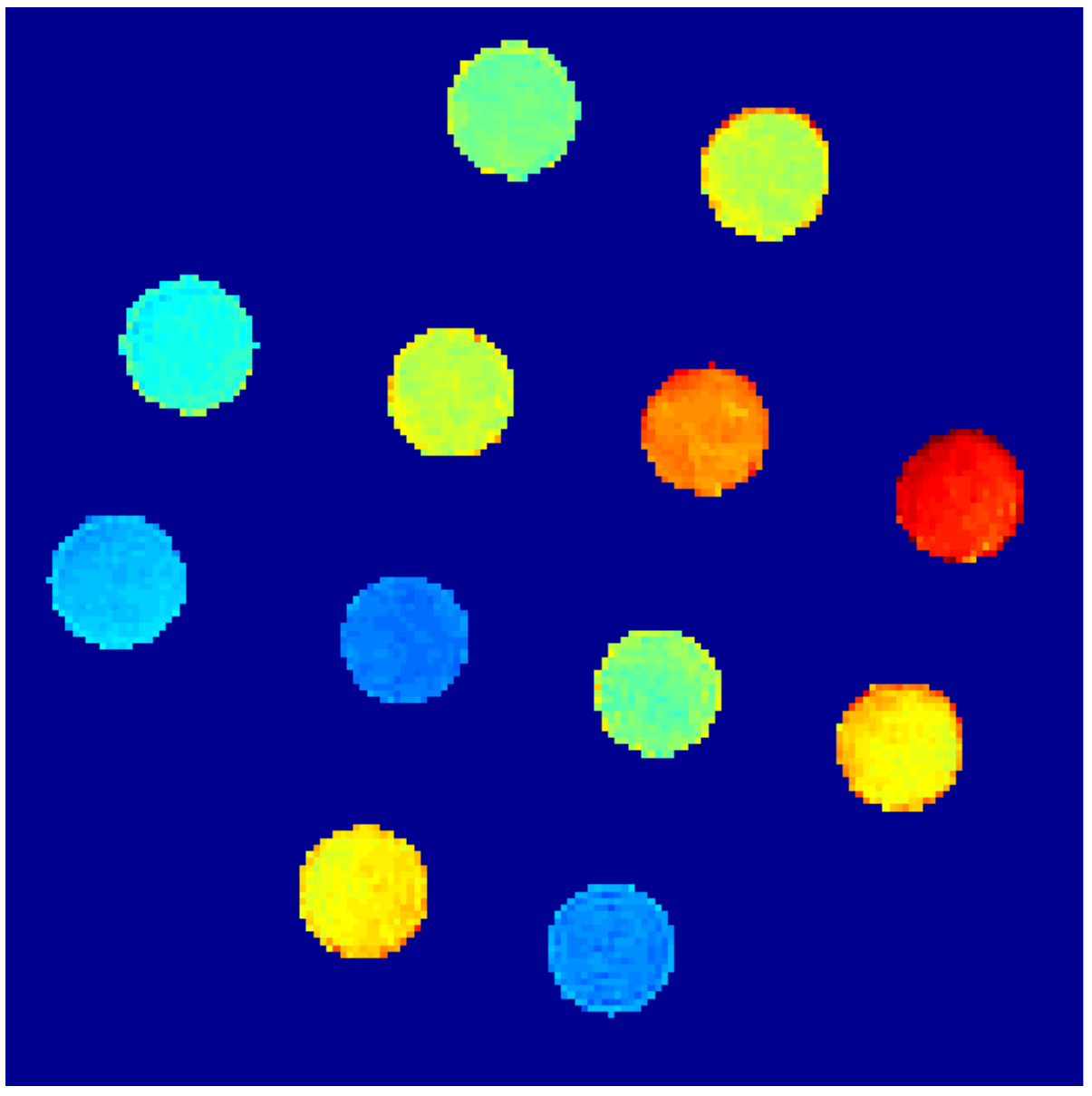} 		\hspace{-.25cm}	
		\includegraphics[width=.16\linewidth]{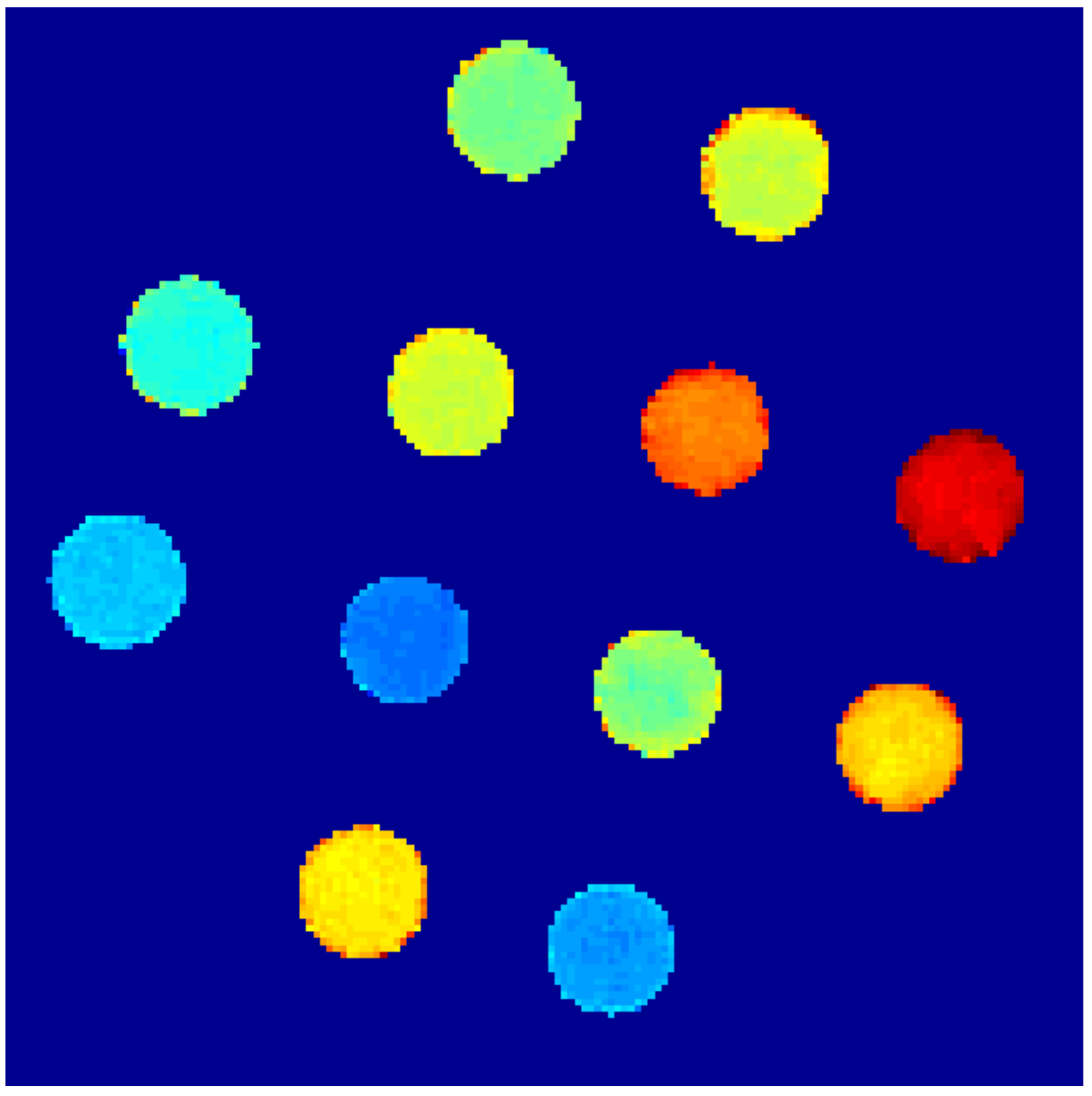} 		\hspace{-.25cm}	
		\includegraphics[width=.16\linewidth]{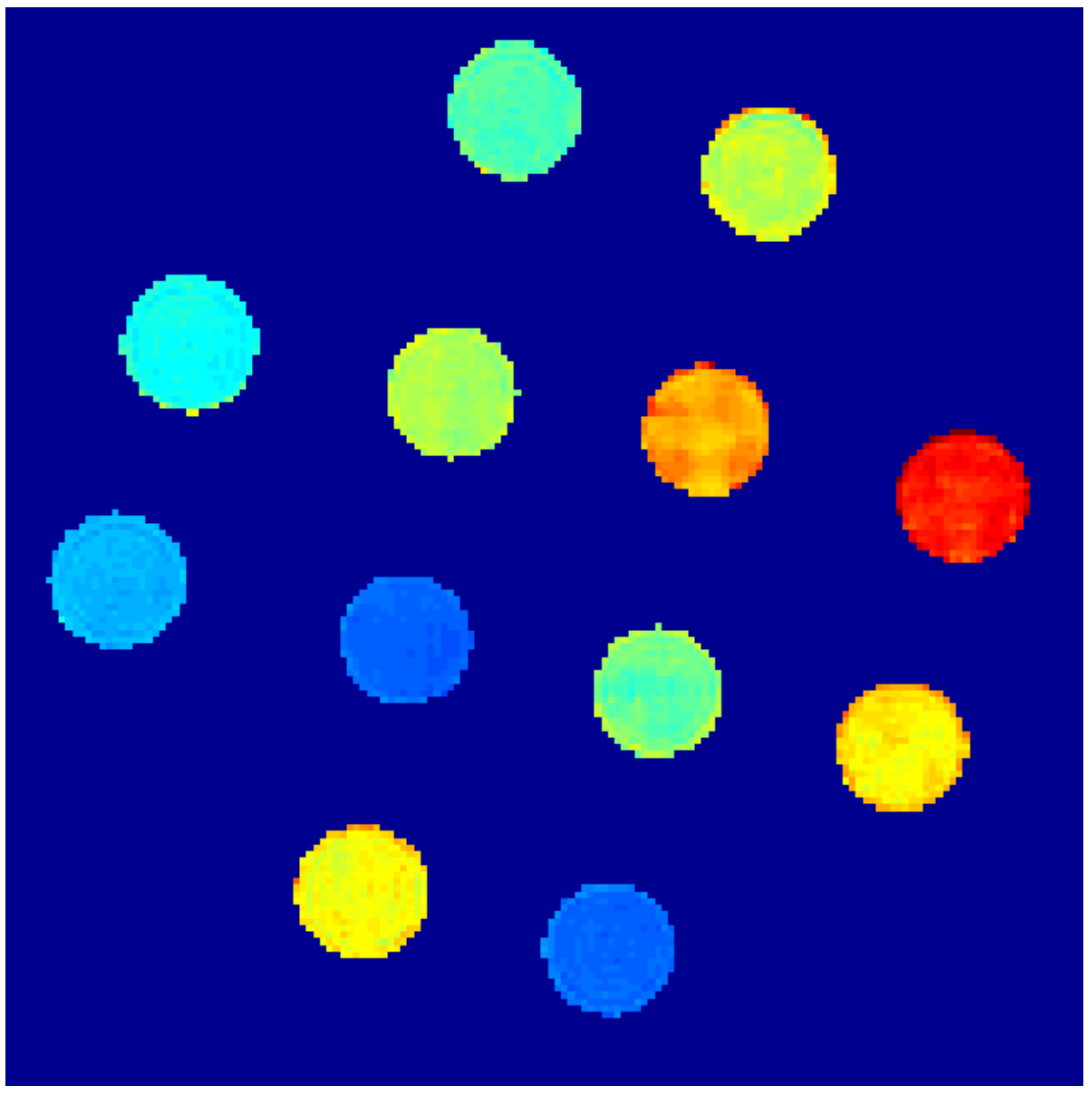} 
		\\
		\includegraphics[trim= 100 15 80 700, clip,width=.3\linewidth]{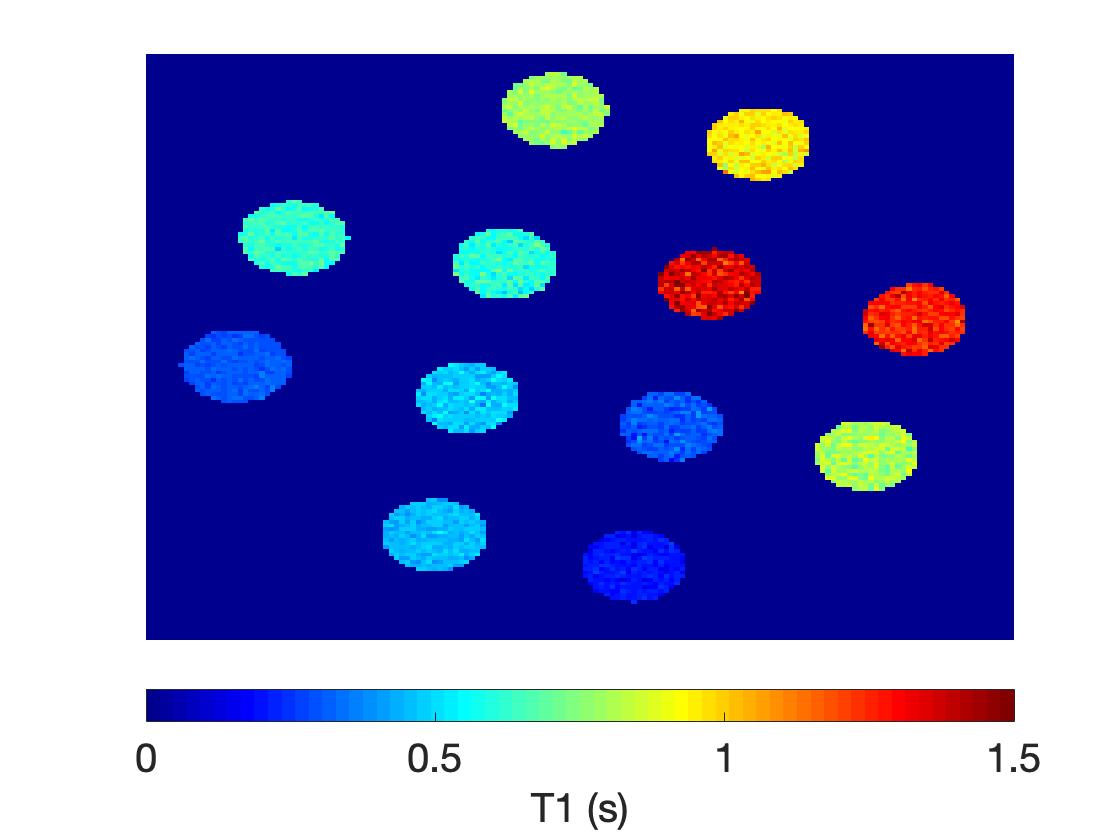}\qquad\qquad\qquad\qquad\qquad
		\includegraphics[trim= 100 15 80 700, clip,width=.3\linewidth]{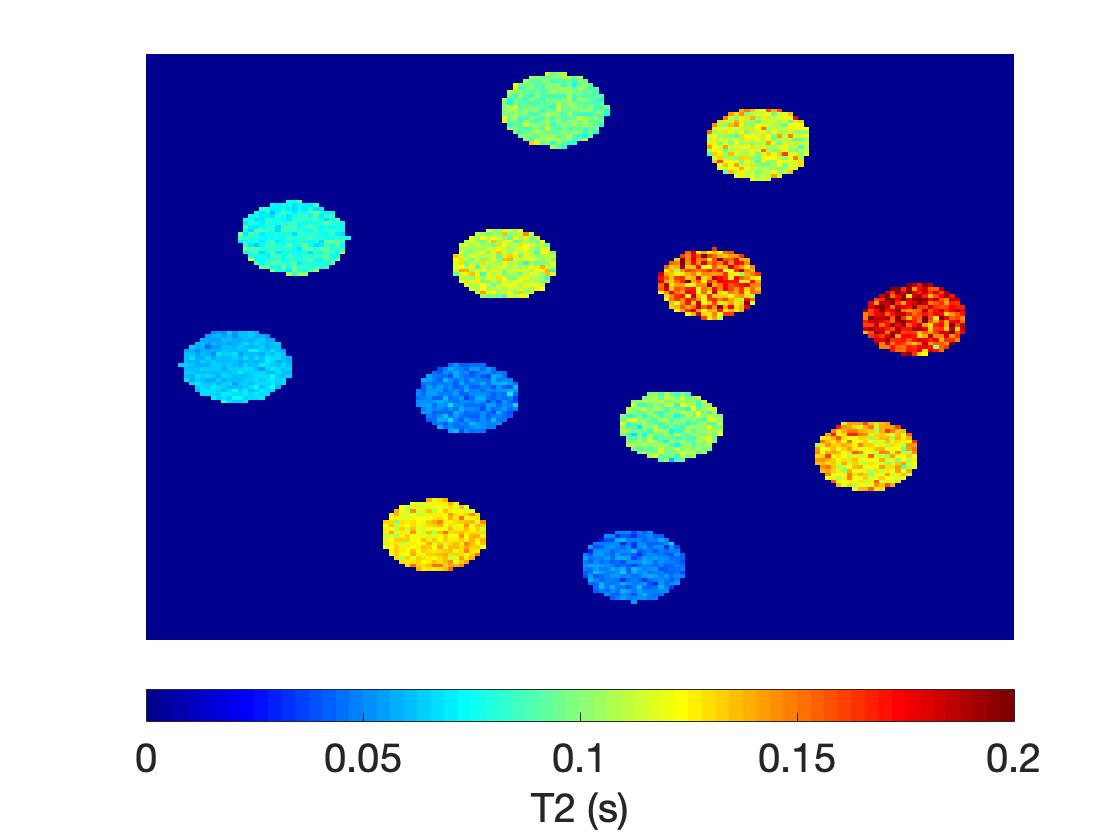}
		\caption{Reconstructed T1 (3 left columns) and T2 (3 right columns) maps of EUROSPIN TO5 phantom, imaged using the 2D spiral (1st sub-column), 2D radial (2nd sub-column) and 3D spiral (3rd sub-column) k-space acquisitions. Tested methods from top to the bottom row are ZF-DM, LR-DM, VS-DM and the proposed LRTV-MRFResnet algorithm. \label{fig:phanmaps}}
	\end{minipage}
\end{figure*}

In Figure~\ref{fig:phanmaps} we display the computed T1, T2 and PD maps for our \textit{in-vitro} phantom experiments in section~VI-D. 
Tested reconstruction methods are ZF, LR, VS and the proposed LRTV, all fed to the MRFResnet for quantitative inference. Methods ZF and LR result in noisy predictions. It can be observed that for the 2D acquisitions (spiral/radial) VS strongly compromises between outputting smoother images  and overestimated T2 values (bias). This issue is also present in \emph{in-vivo} and \emph{in-silico} experiments, where less k-space neighbourhood information are available to share (compared to the 3D acquisitions) and make the VS noncompetitive, and further the overall quantifications inconsistent across 2D/3D acquisitions. The proposed LRTV overcomes this issue through a model-based compressed sensing reconstruction.  

\section{Reconstructed maps for the 3D \textit{in-vivo} scans}

\begin{figure*}[h!]
	\centering
	\scalebox{.85}{
	\begin{minipage}{\textwidth}
		\centering
		\begin{turn}{90} \qquad\quad ZF-DM\end{turn}
		\includegraphics[trim= 25 15 25 25, clip, width=.15\linewidth]{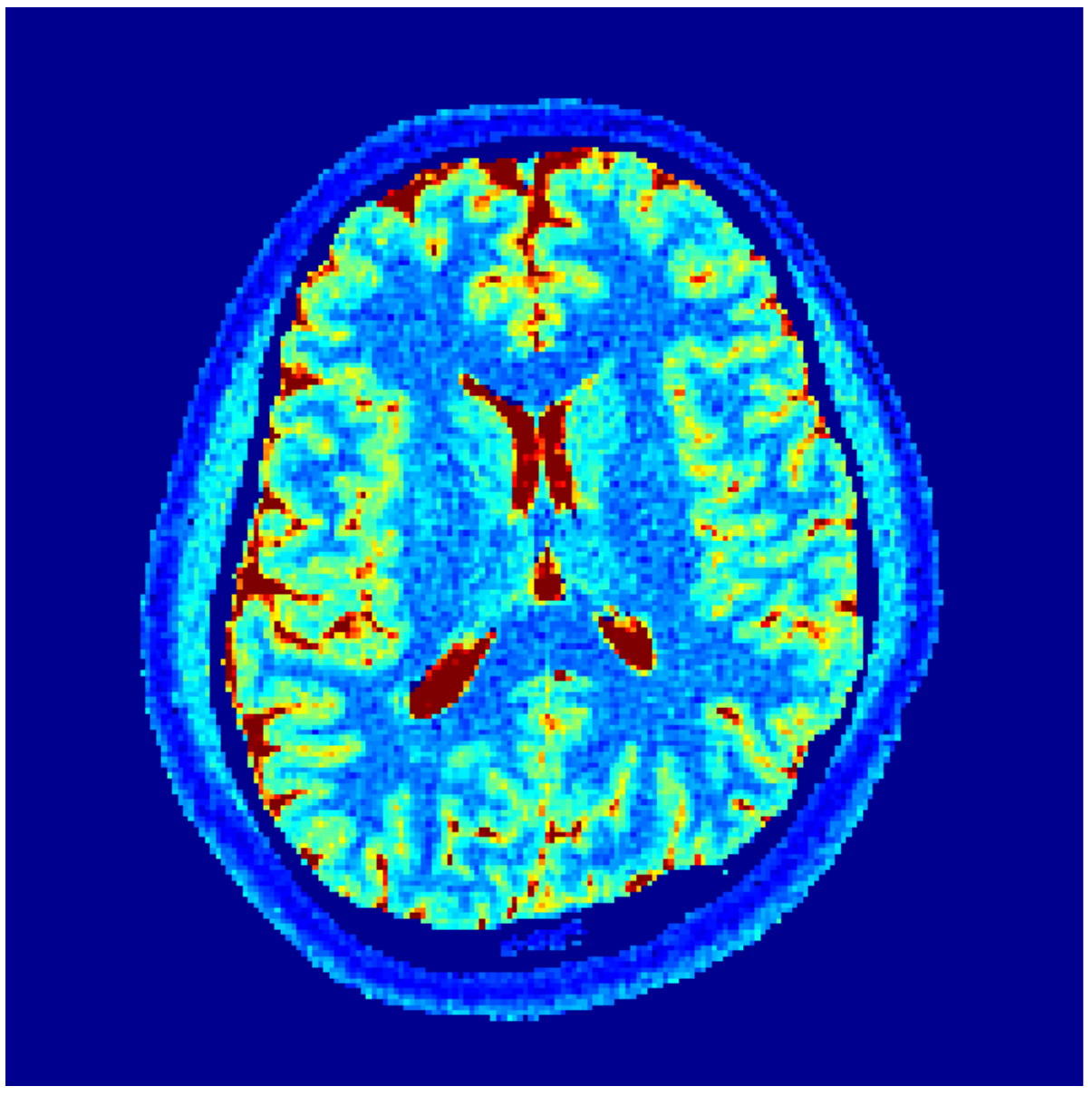}\hspace{-.05cm}
		\includegraphics[trim= 25 15 25 25, clip, width=.15\linewidth]{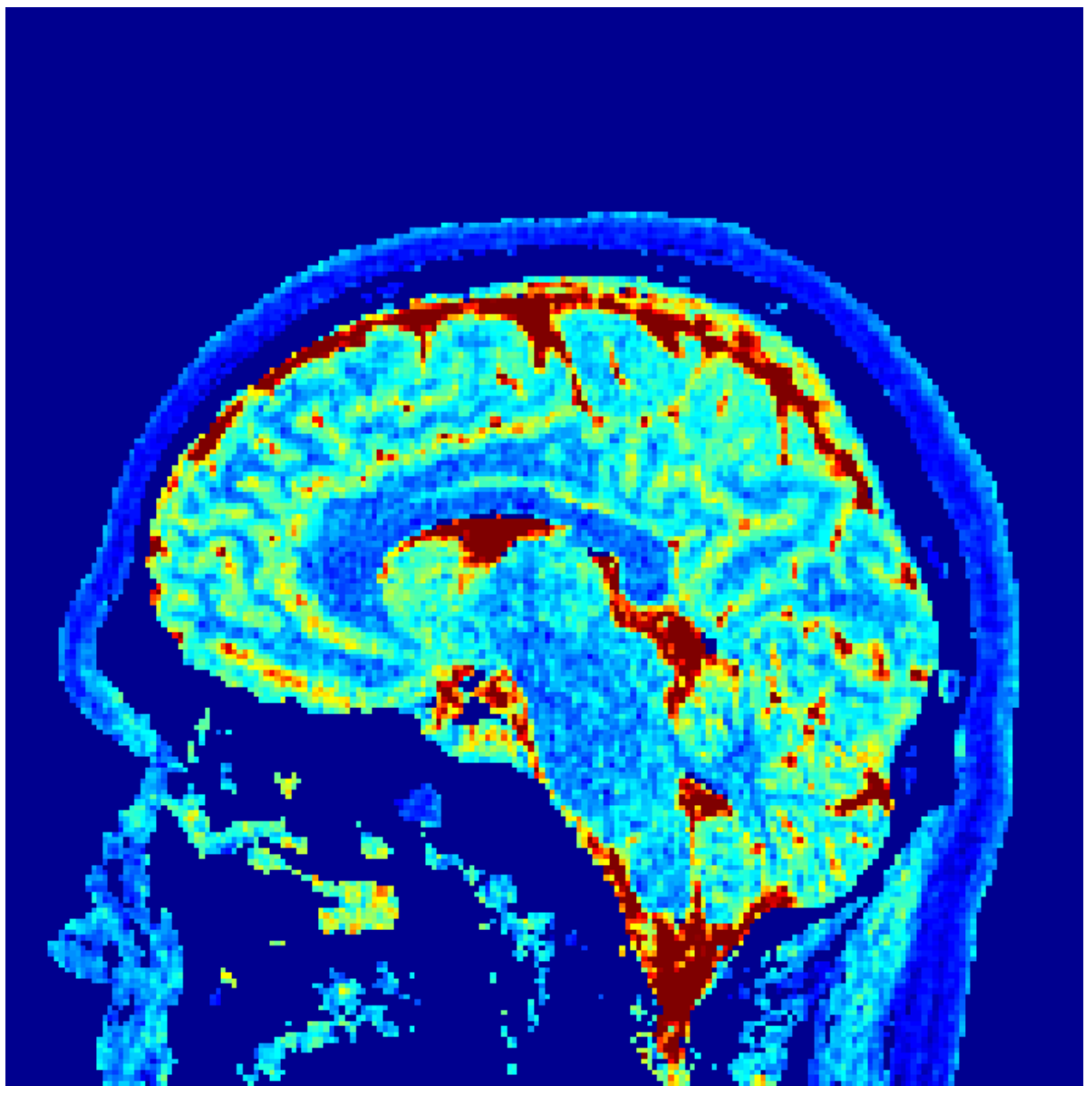}\hspace{-.05cm}
		\includegraphics[trim= 25 15 25 25, clip, width=.15\linewidth]{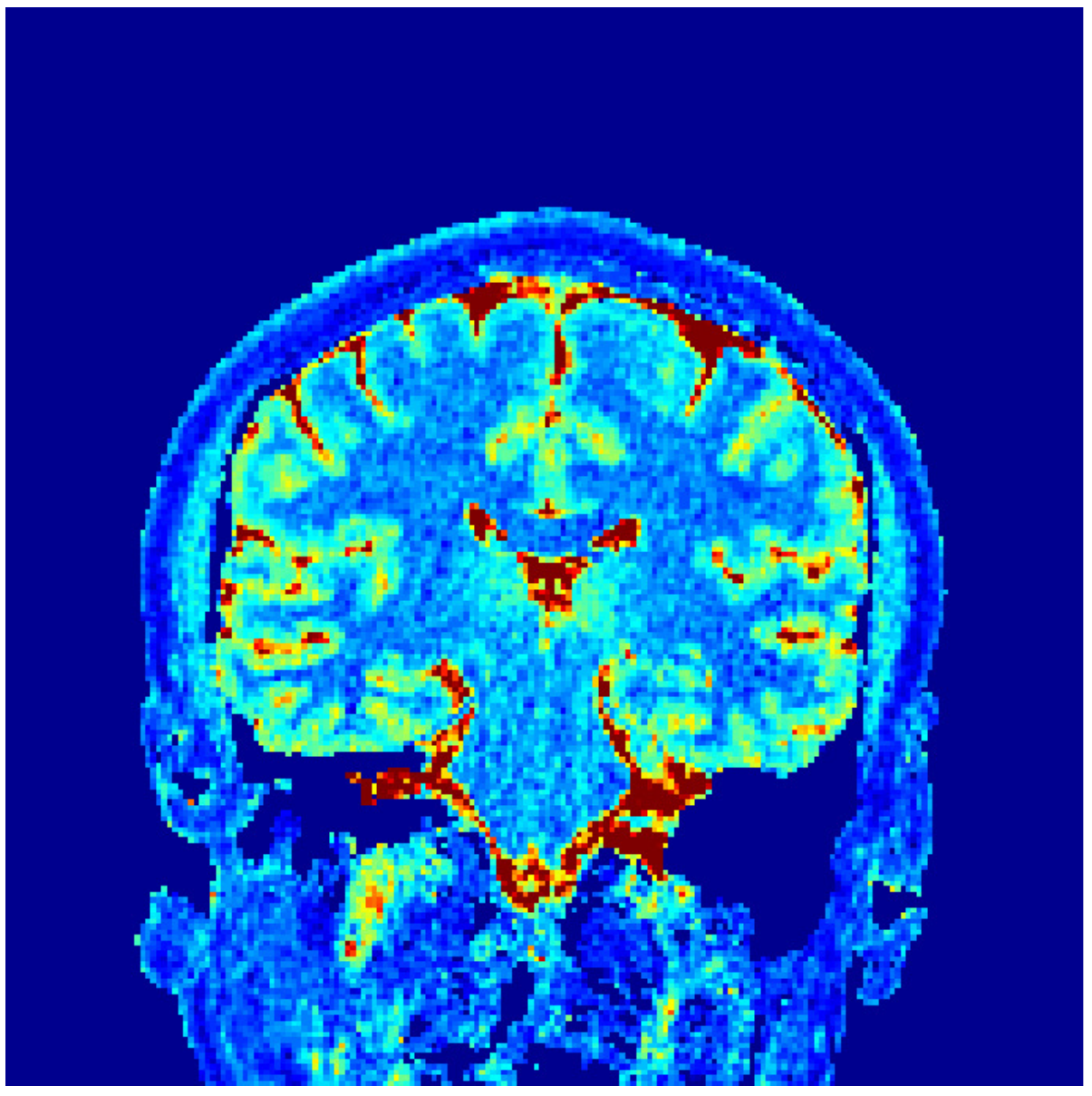}\hspace{.15cm}
		\includegraphics[trim= 25 15 25 25, clip, width=.15\linewidth]{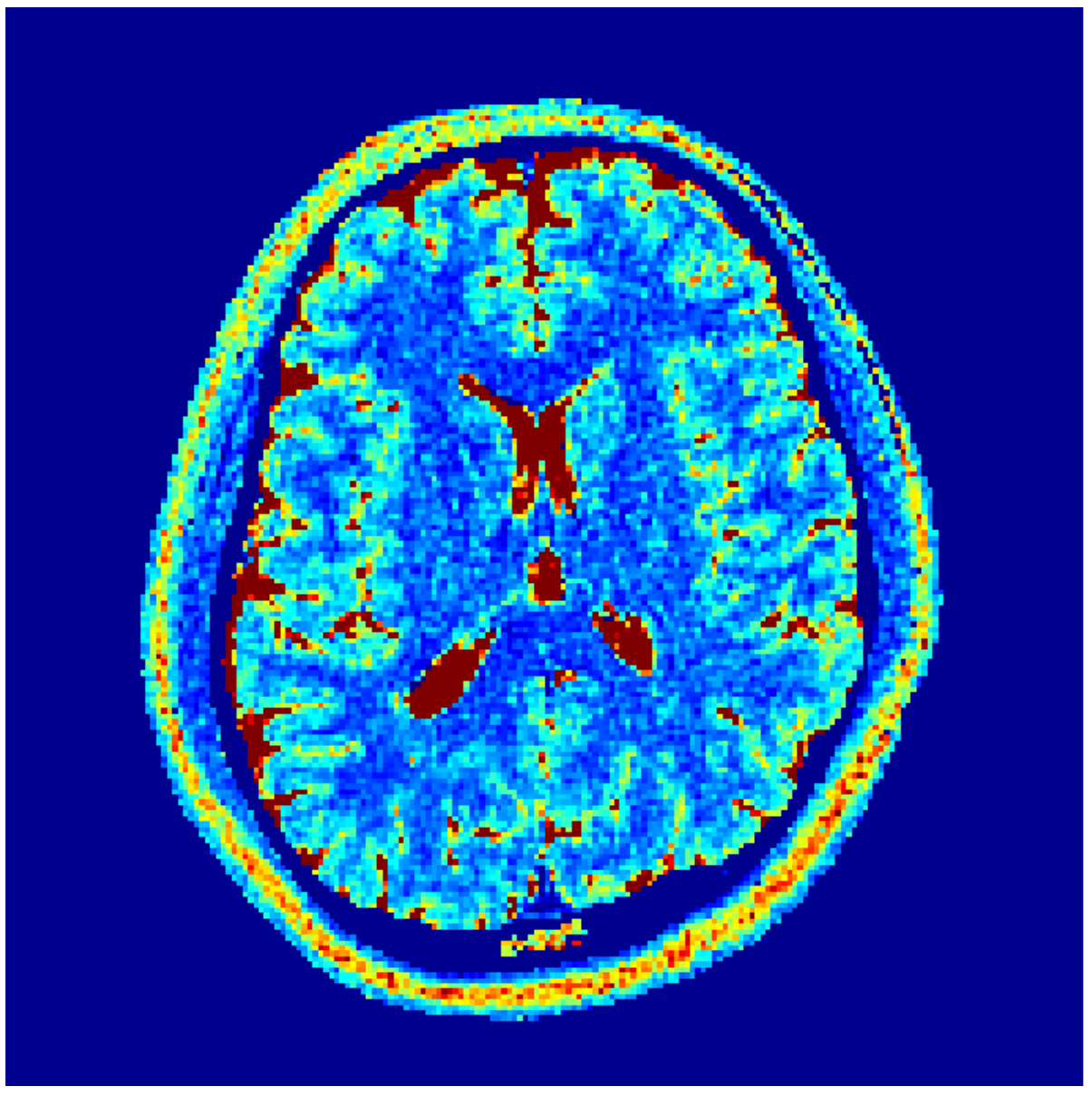}\hspace{-.05cm}
		\includegraphics[trim= 25 15 25 25, clip, width=.15\linewidth]{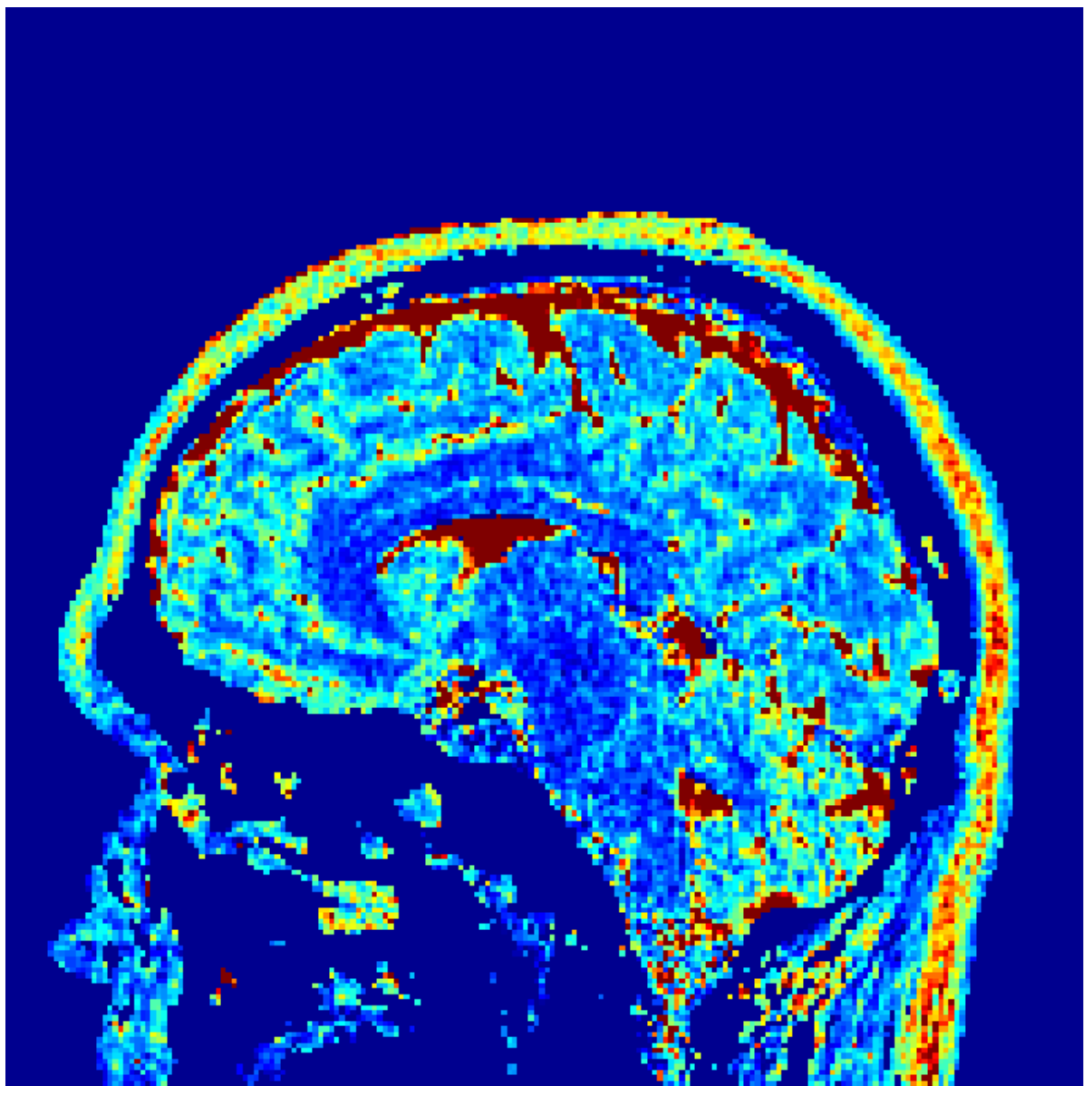}\hspace{-.05cm}
		\includegraphics[trim= 25 15 25 25, clip, width=.15\linewidth]{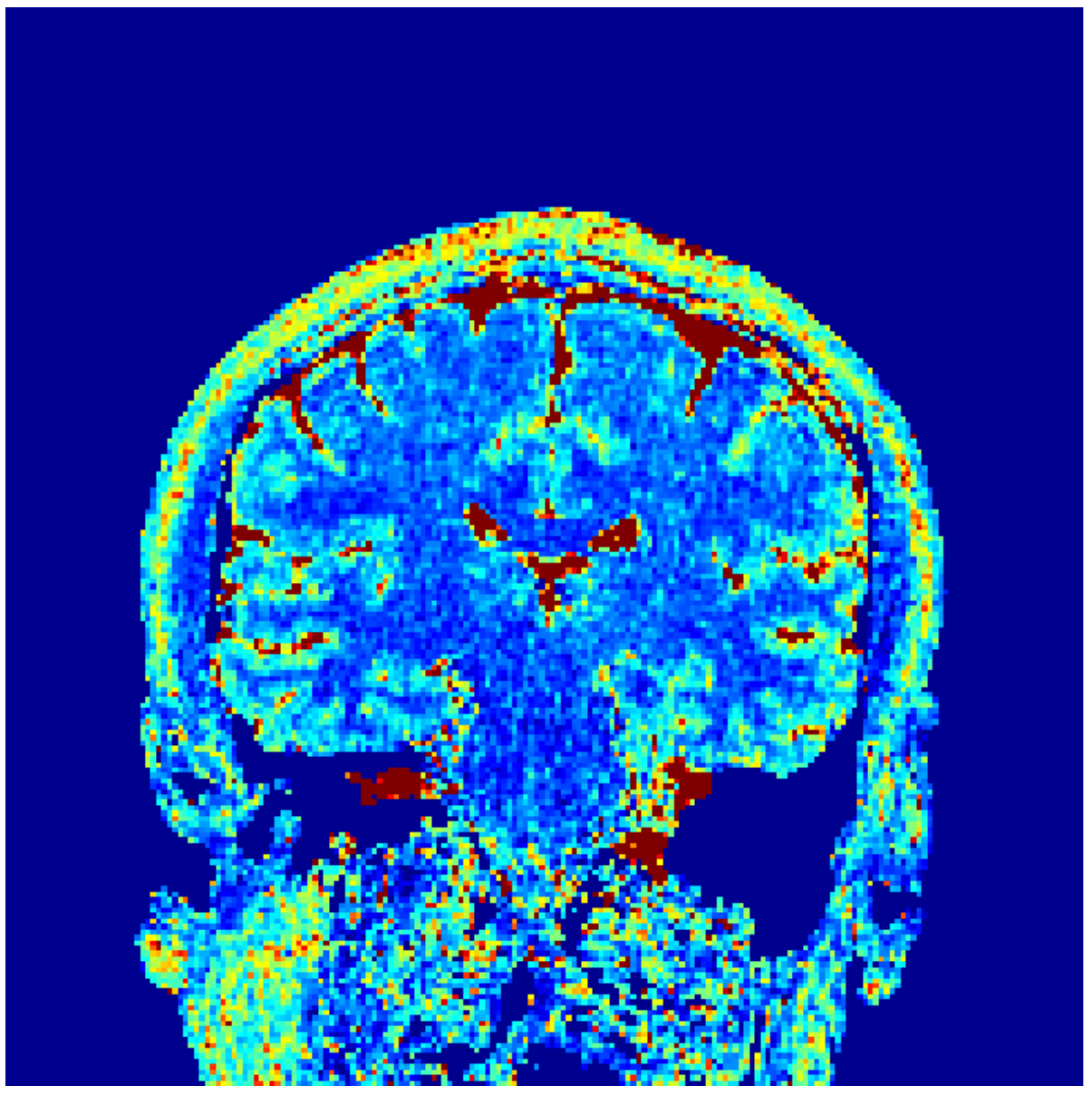}\hspace{-.05cm}
		\\
		\begin{turn}{90} \qquad\quad LR-DM\end{turn}
		\includegraphics[trim= 25 15 25 25, clip, width=.15\linewidth]{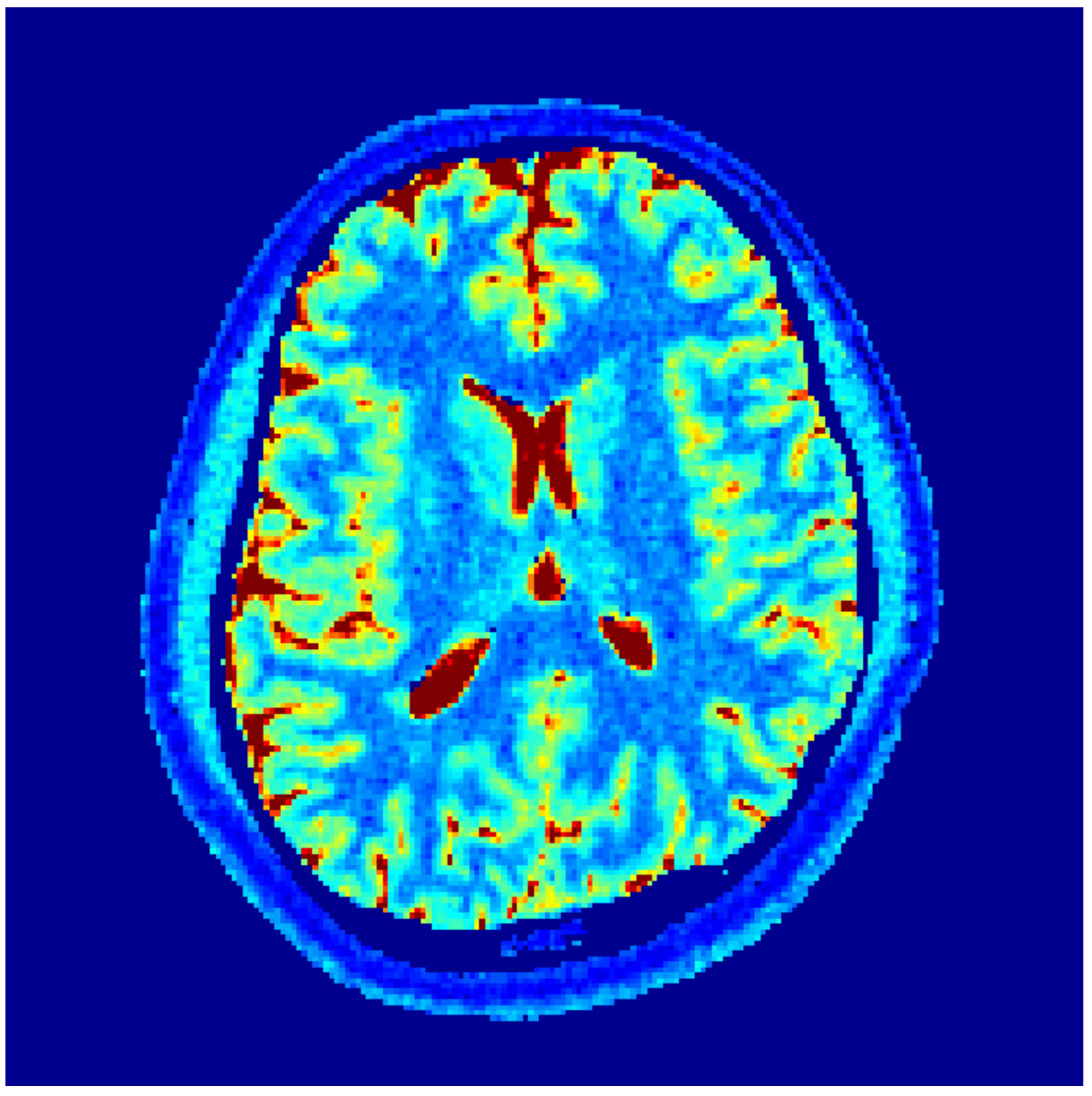}\hspace{-.05cm}
		\includegraphics[trim= 25 15 25 25, clip, width=.15\linewidth]{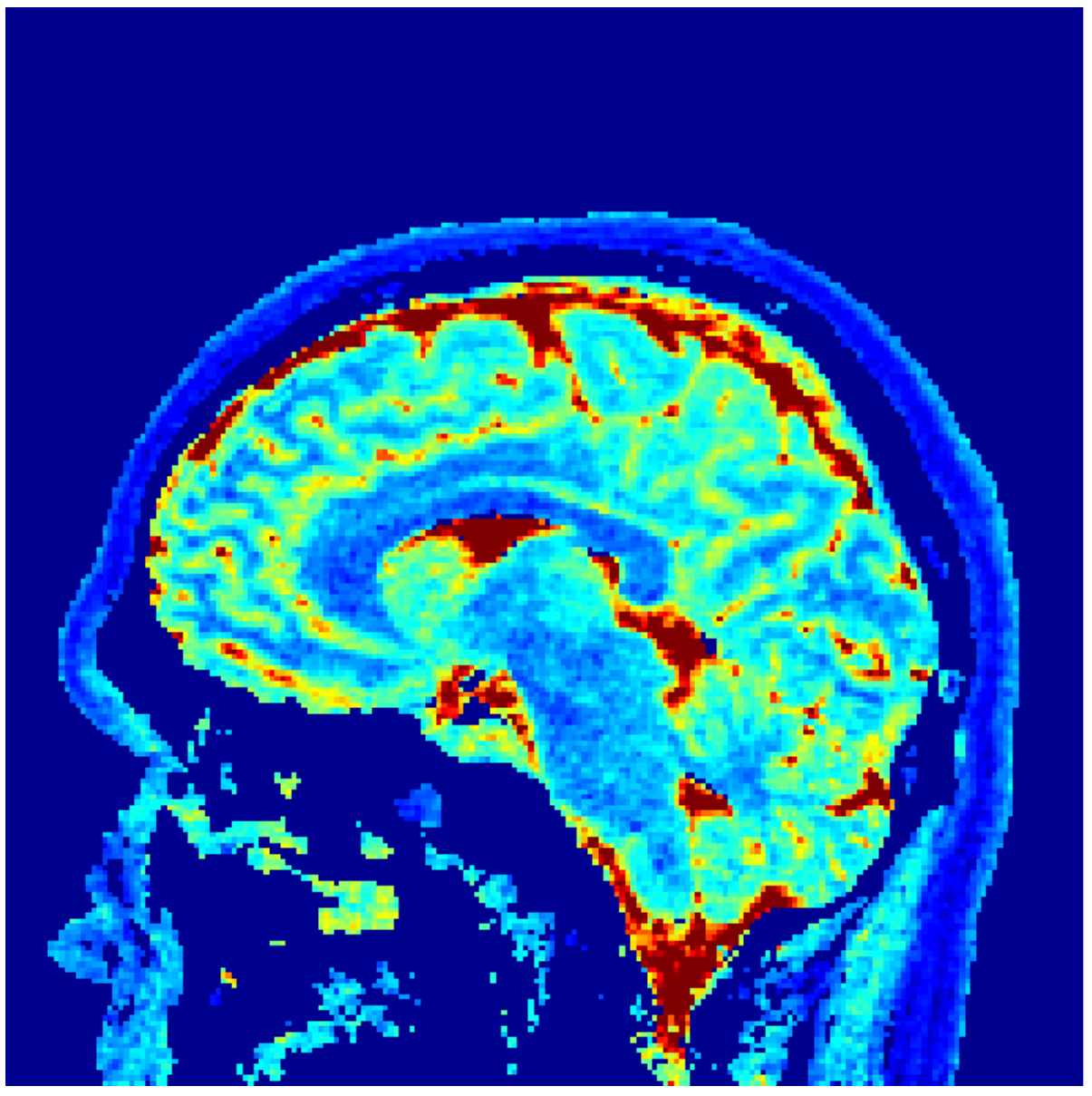}\hspace{-.05cm}
		\includegraphics[trim= 25 15 25 25, clip, width=.15\linewidth]{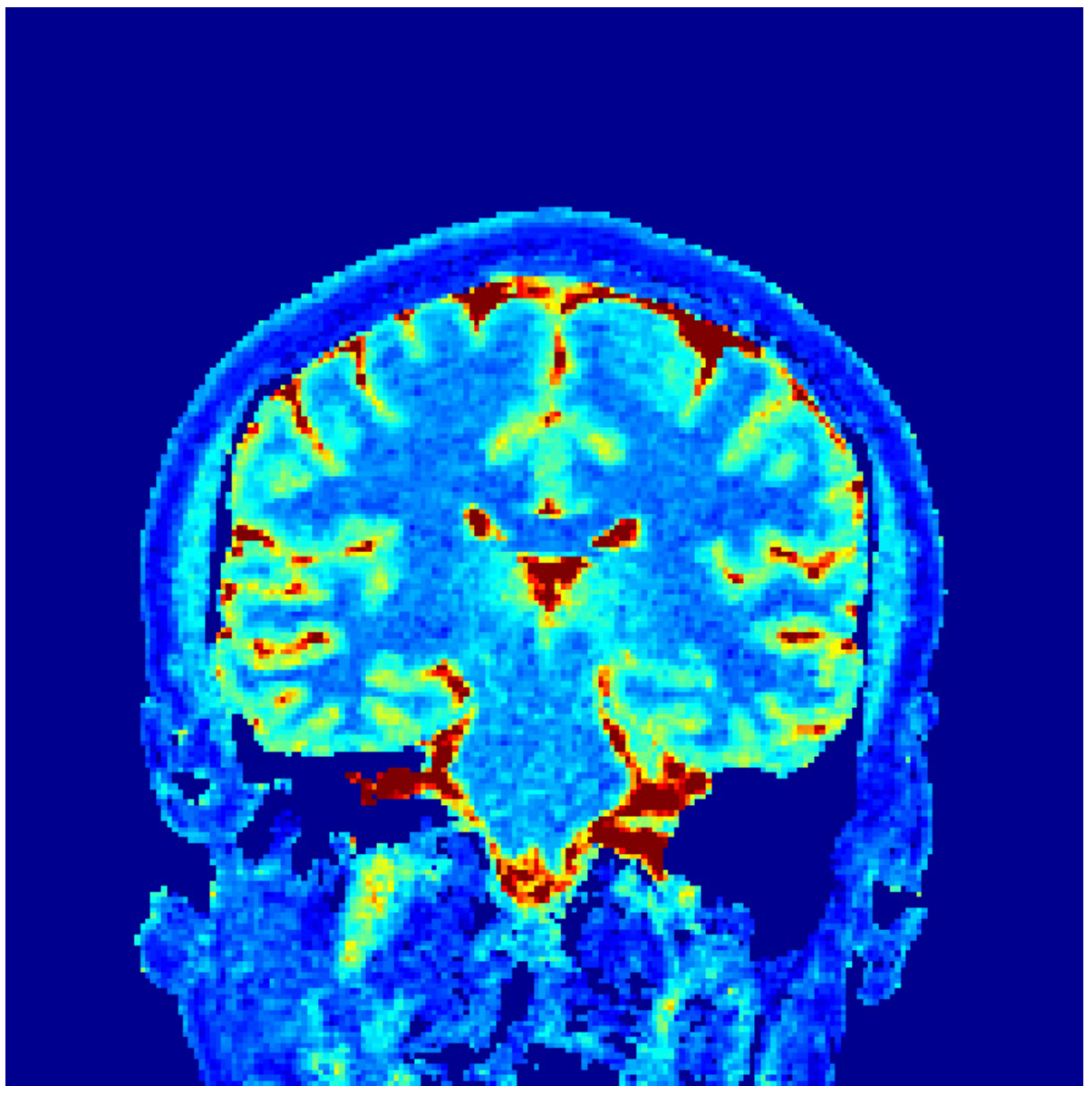}\hspace{.15cm}
		\includegraphics[trim= 25 15 25 25, clip, width=.15\linewidth]{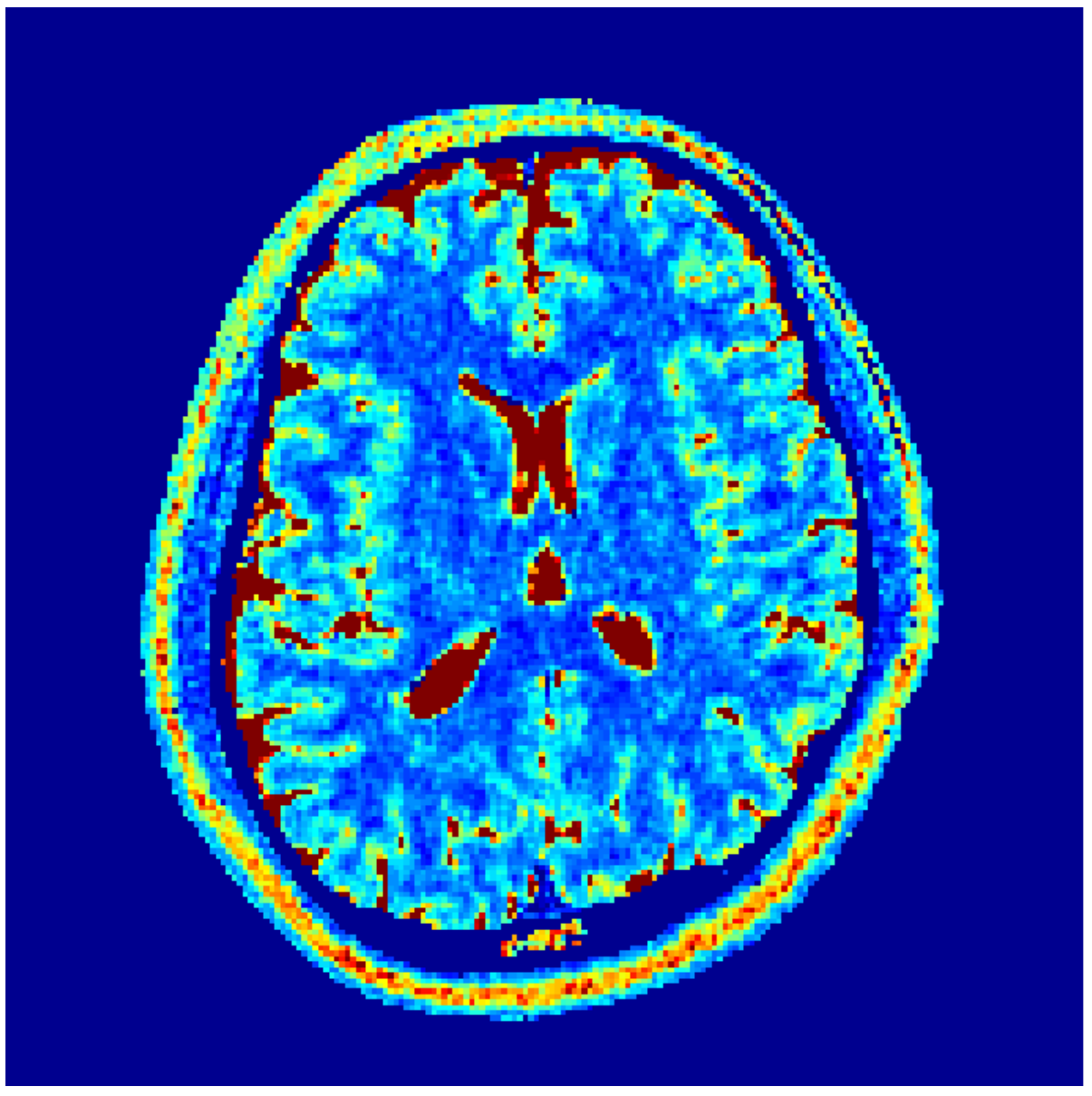}\hspace{-.05cm}
		\includegraphics[trim= 25 15 25 25, clip, width=.15\linewidth]{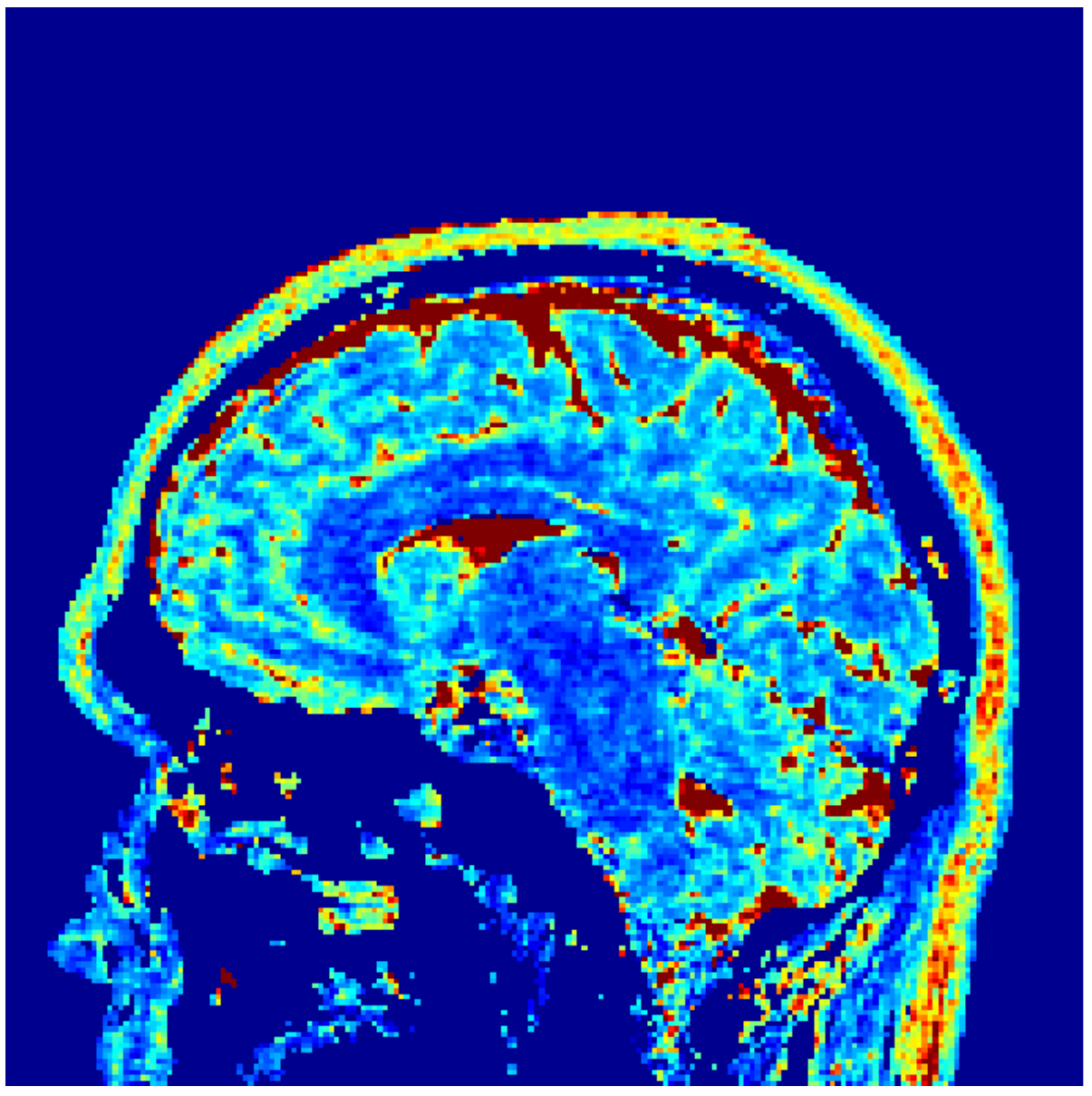}\hspace{-.05cm}
		\includegraphics[trim= 25 15 25 25, clip, width=.15\linewidth]{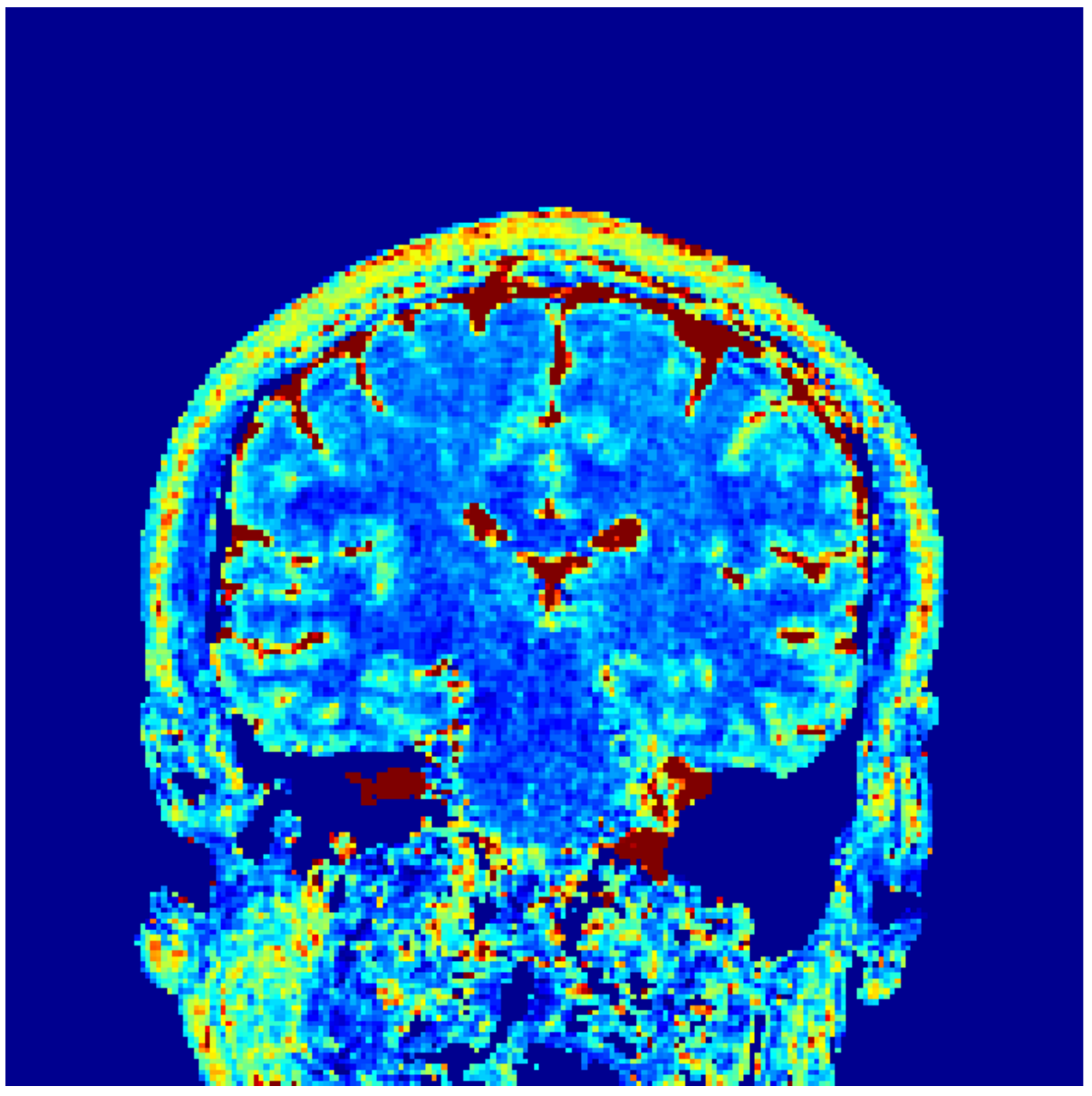}\hspace{-.05cm}
		\\
		\begin{turn}{90} \qquad\quad AIR-MRF\end{turn}
		\includegraphics[trim= 25 15 25 25, clip, width=.15\linewidth]{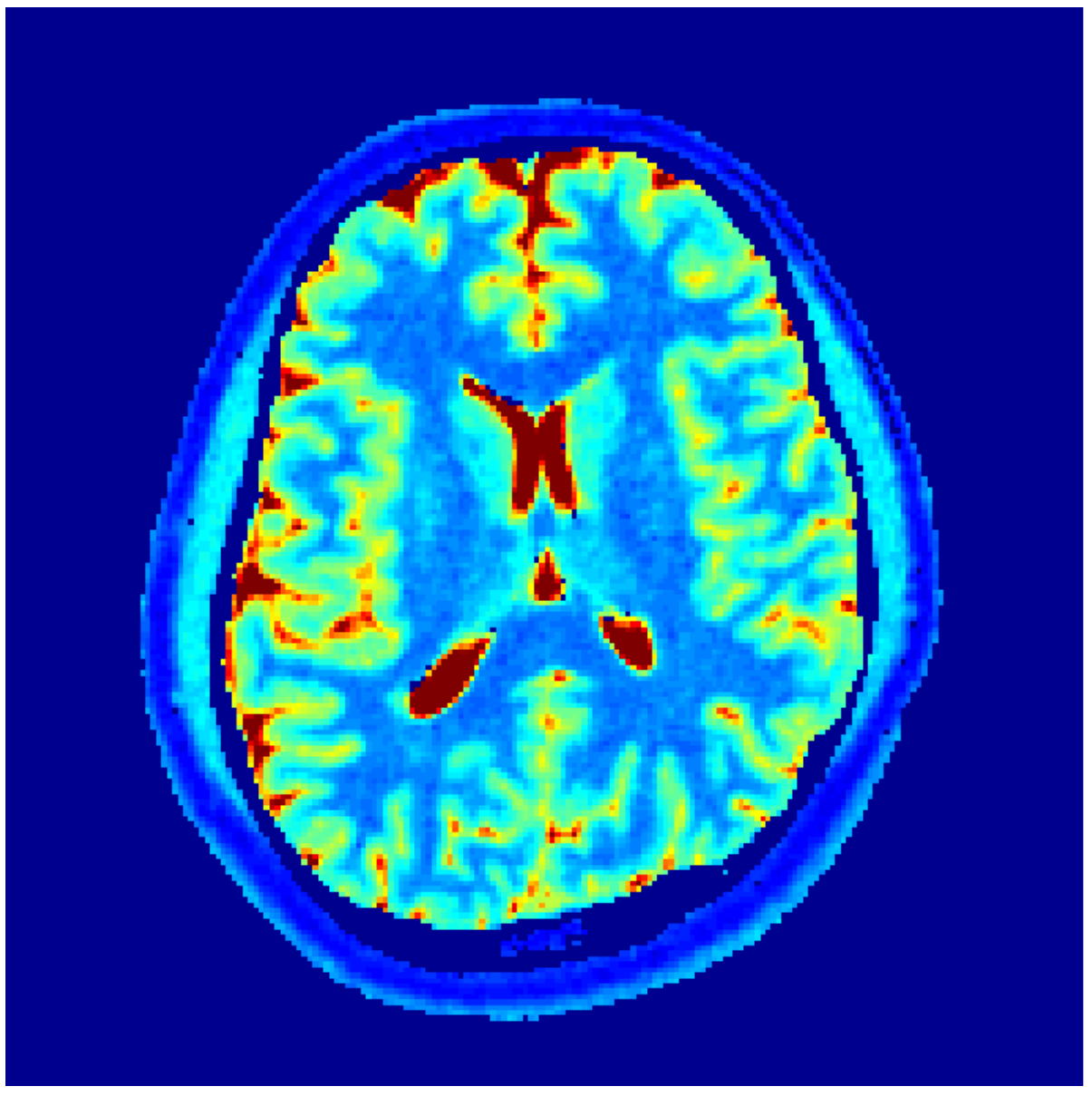}\hspace{-.05cm}
		\includegraphics[trim= 25 15 25 25, clip, width=.15\linewidth]{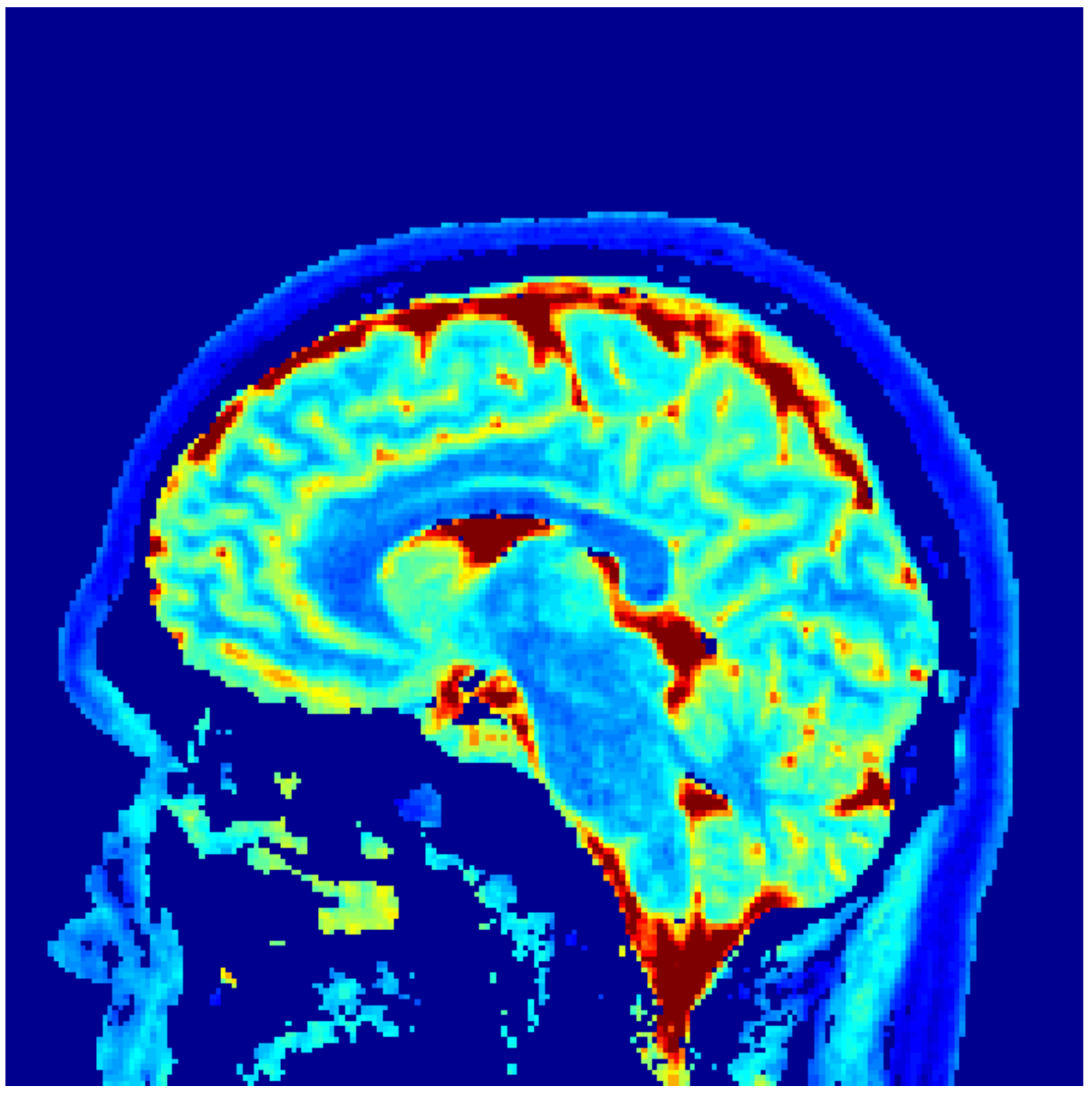}\hspace{-.05cm}
		\includegraphics[trim= 25 15 25 25, clip, width=.15\linewidth]{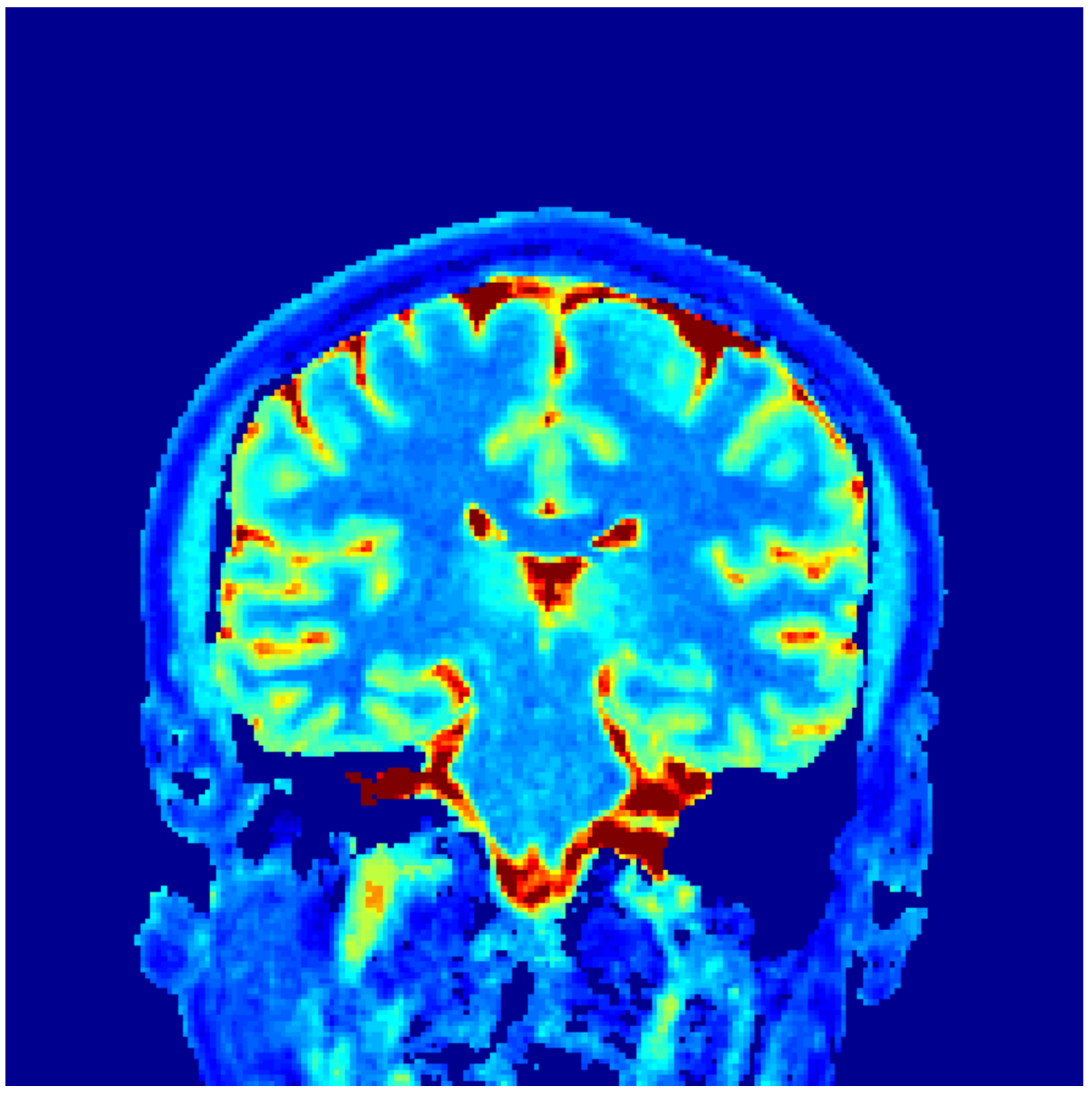}\hspace{.15cm}
		\includegraphics[trim= 25 15 25 25, clip, width=.15\linewidth]{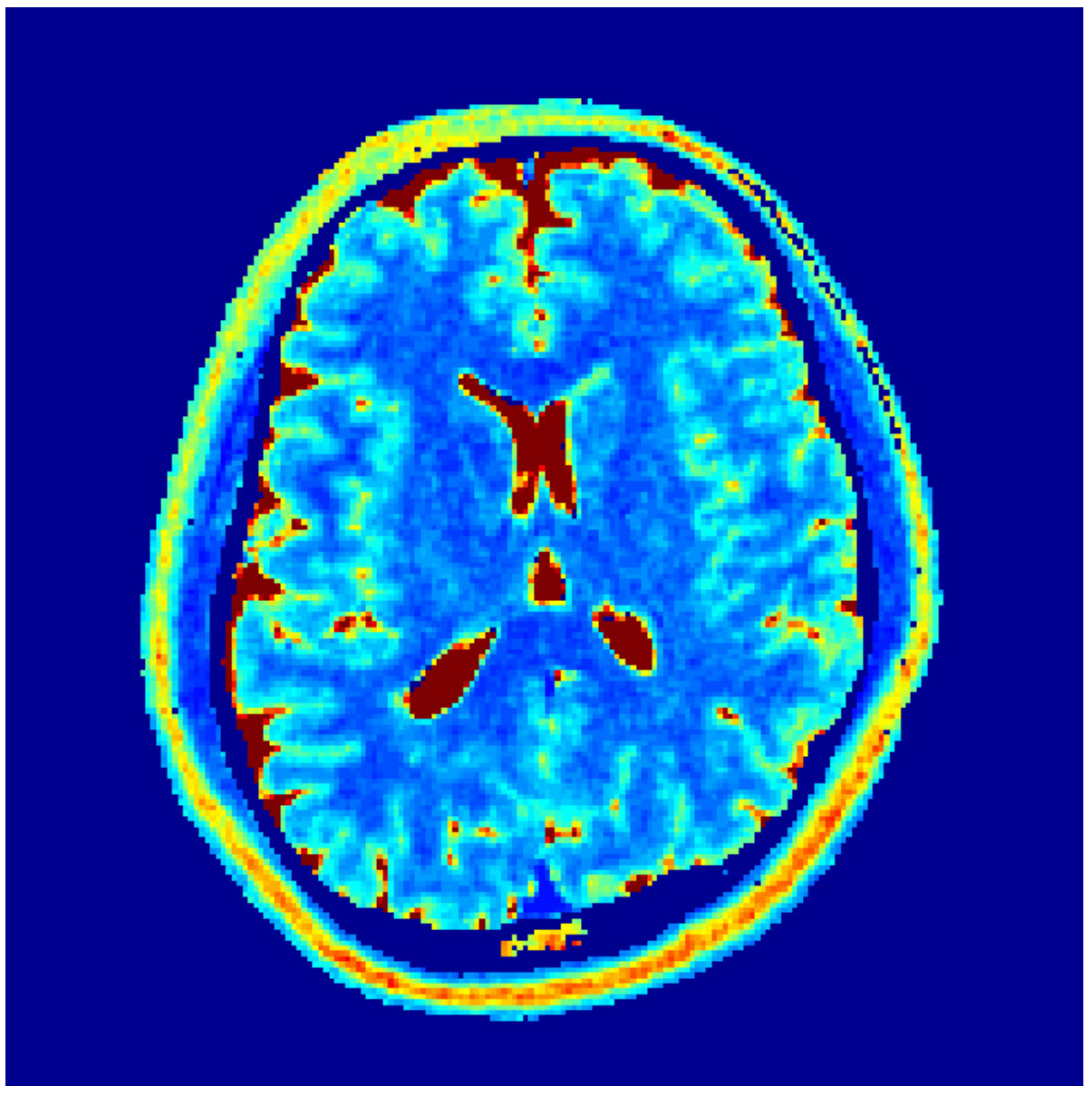}\hspace{-.05cm}
		\includegraphics[trim= 25 15 25 25, clip, width=.15\linewidth]{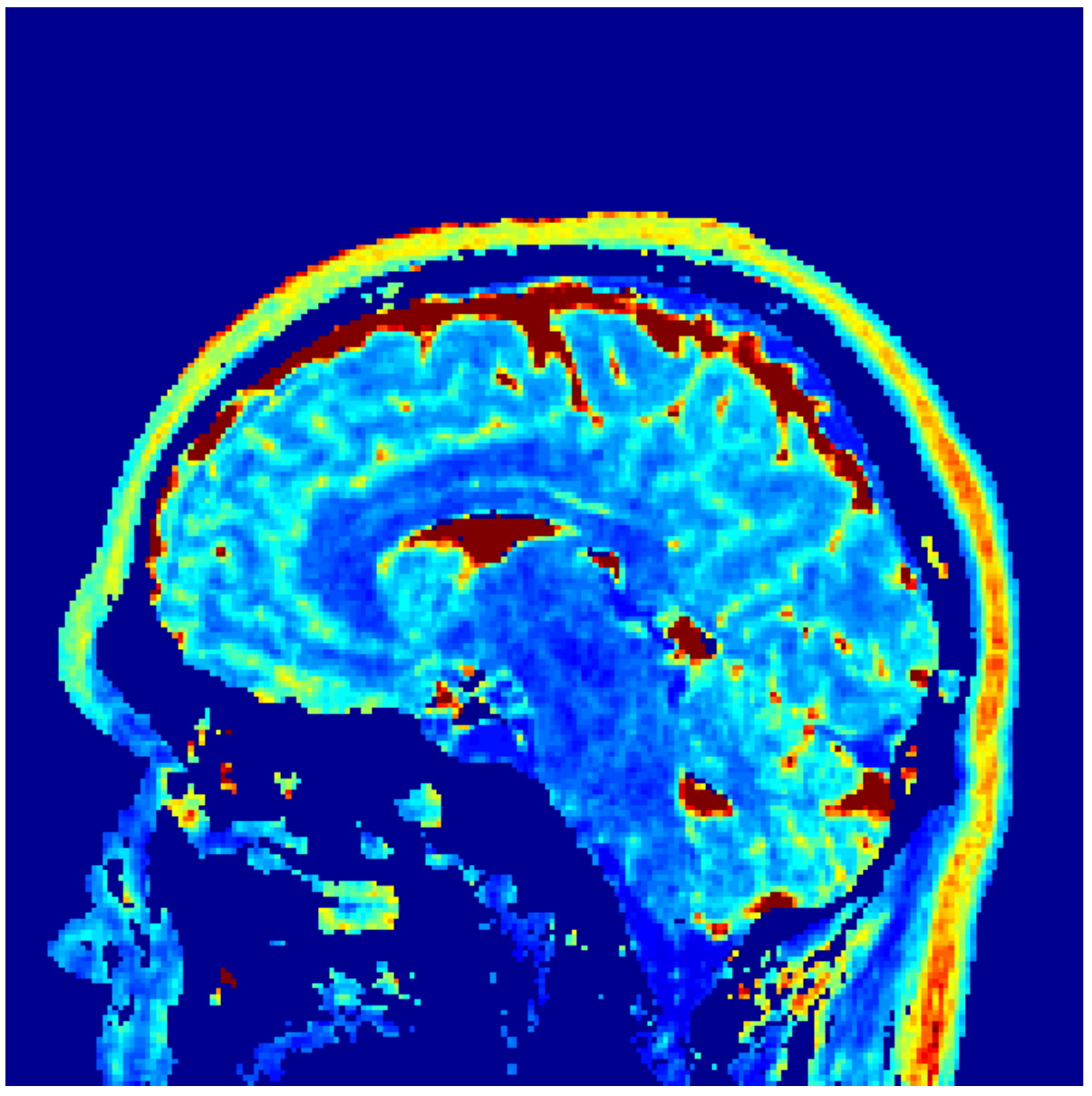}\hspace{-.05cm}
		\includegraphics[trim= 25 15 25 25, clip, width=.15\linewidth]{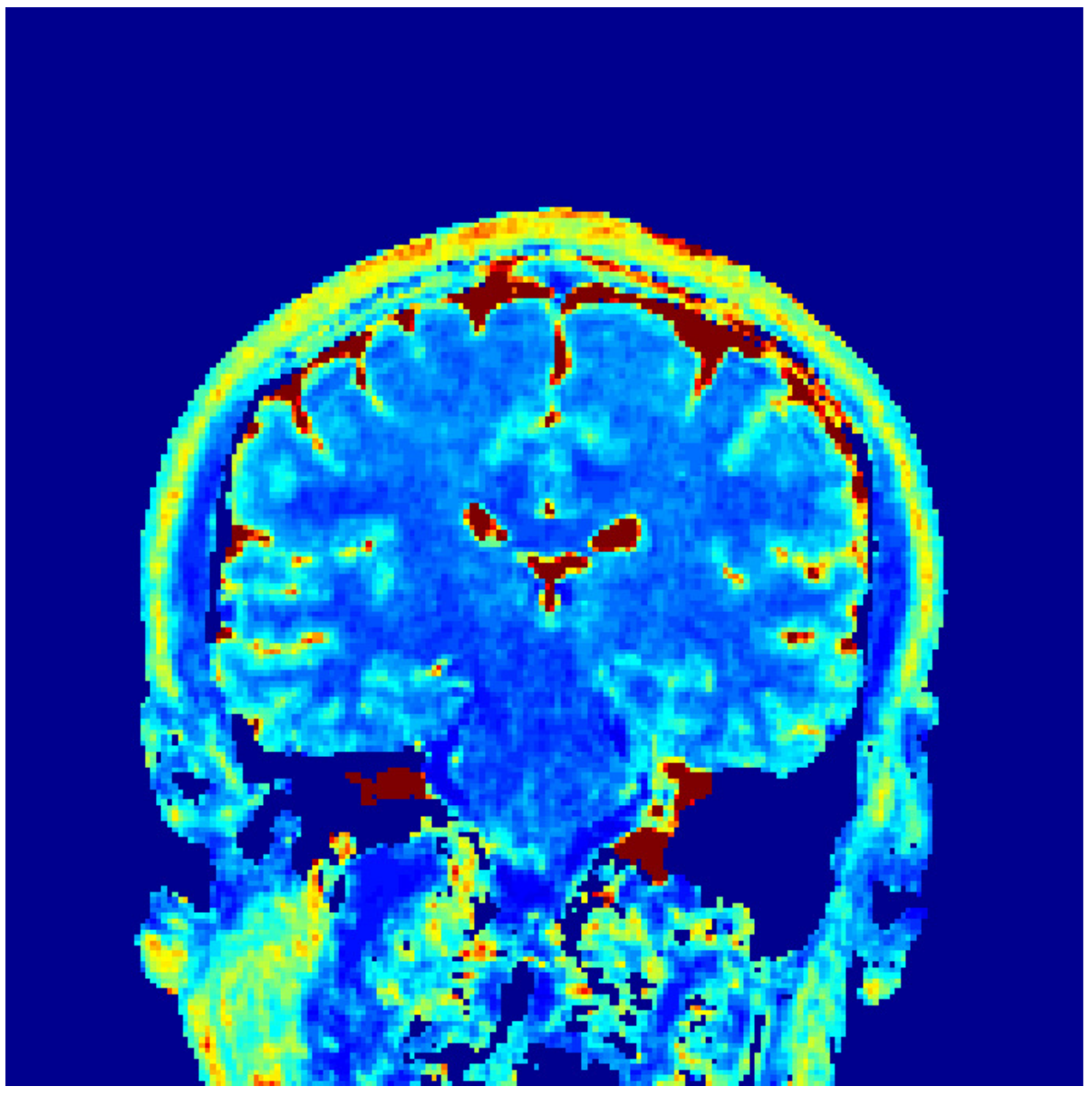}\vspace{-.3cm}
		\\
		\includegraphics[trim= -10 50 -20 720, clip,width=.25\linewidth]{./figs/T1barvivo.jpg}\hspace{4cm}
		\includegraphics[trim= -10 50 -20 720, clip,width=.25\linewidth]{./figs/T2barvivo.jpg}\vspace{.6cm}
		\\
		T1 (s) \hspace{8cm} T2 (s)\vspace{.3cm}
		 \\
		 \begin{turn}{90} \qquad\quad ZF-DM\end{turn}
		\includegraphics[trim= 25 15 25 25, clip, width=.15\linewidth]{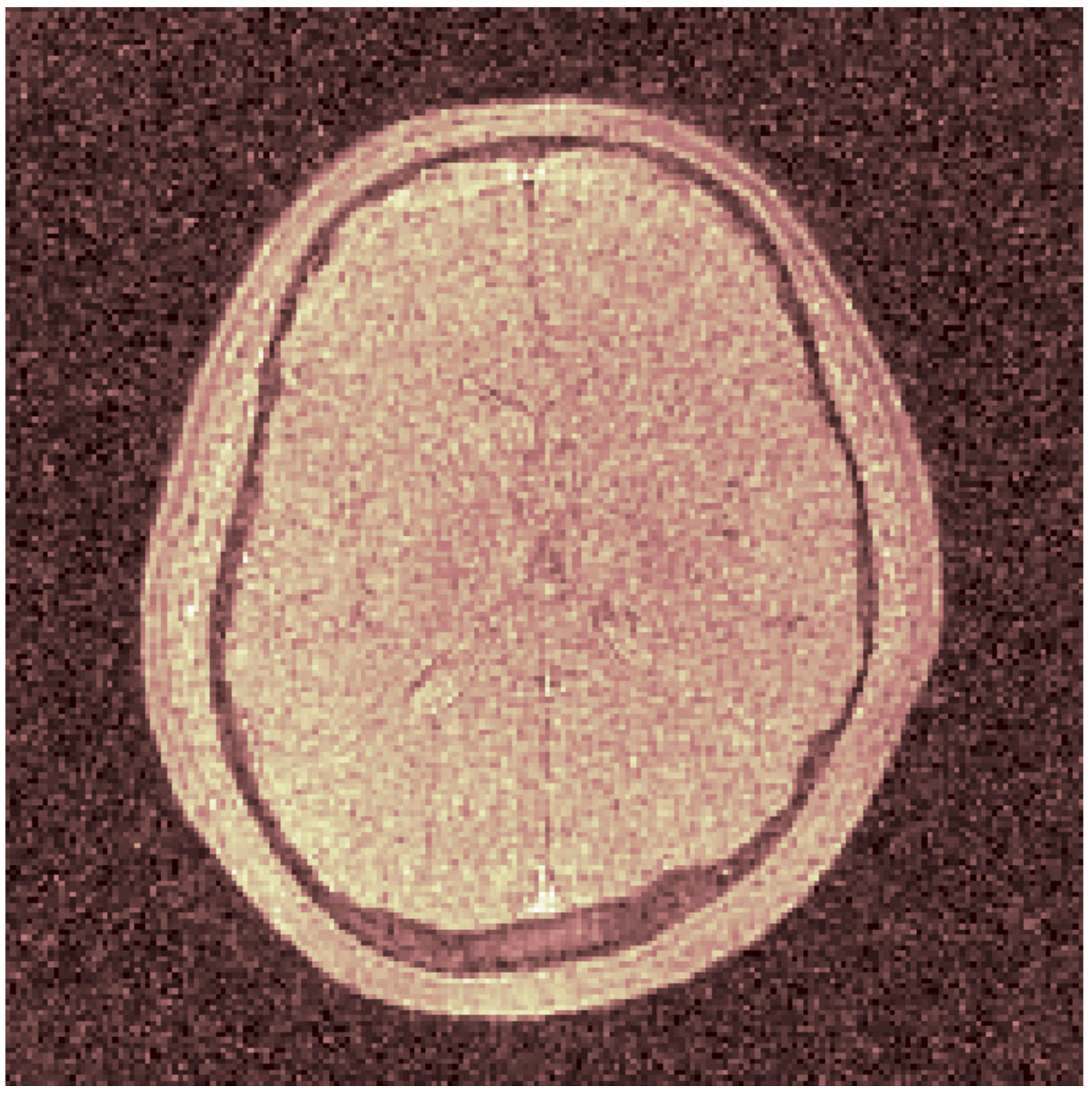}\hspace{-.05cm}
		\includegraphics[trim= 25 15 25 25, clip, width=.15\linewidth]{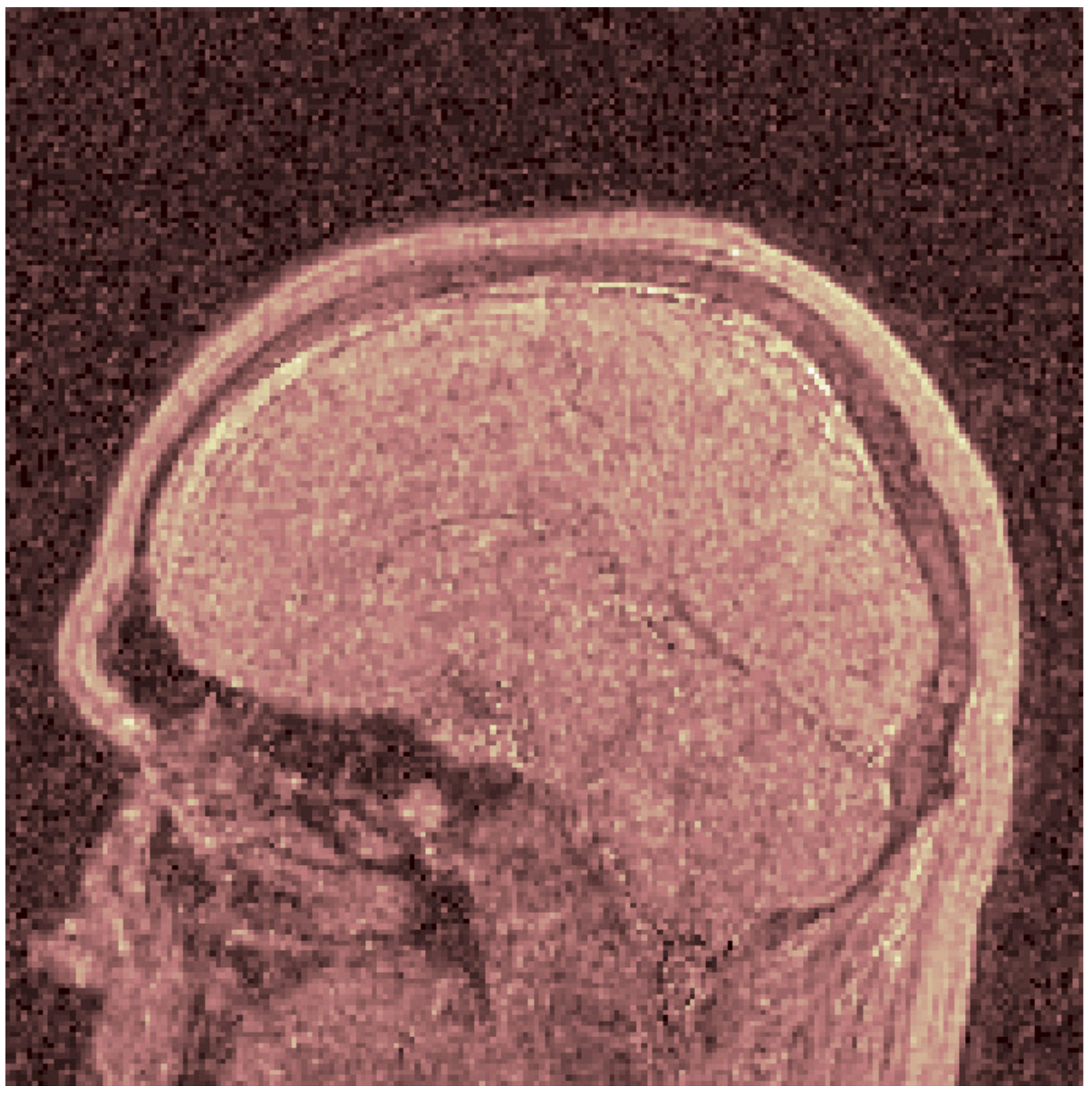}\hspace{-.05cm}
		\includegraphics[trim= 25 15 25 25, clip, width=.15\linewidth]{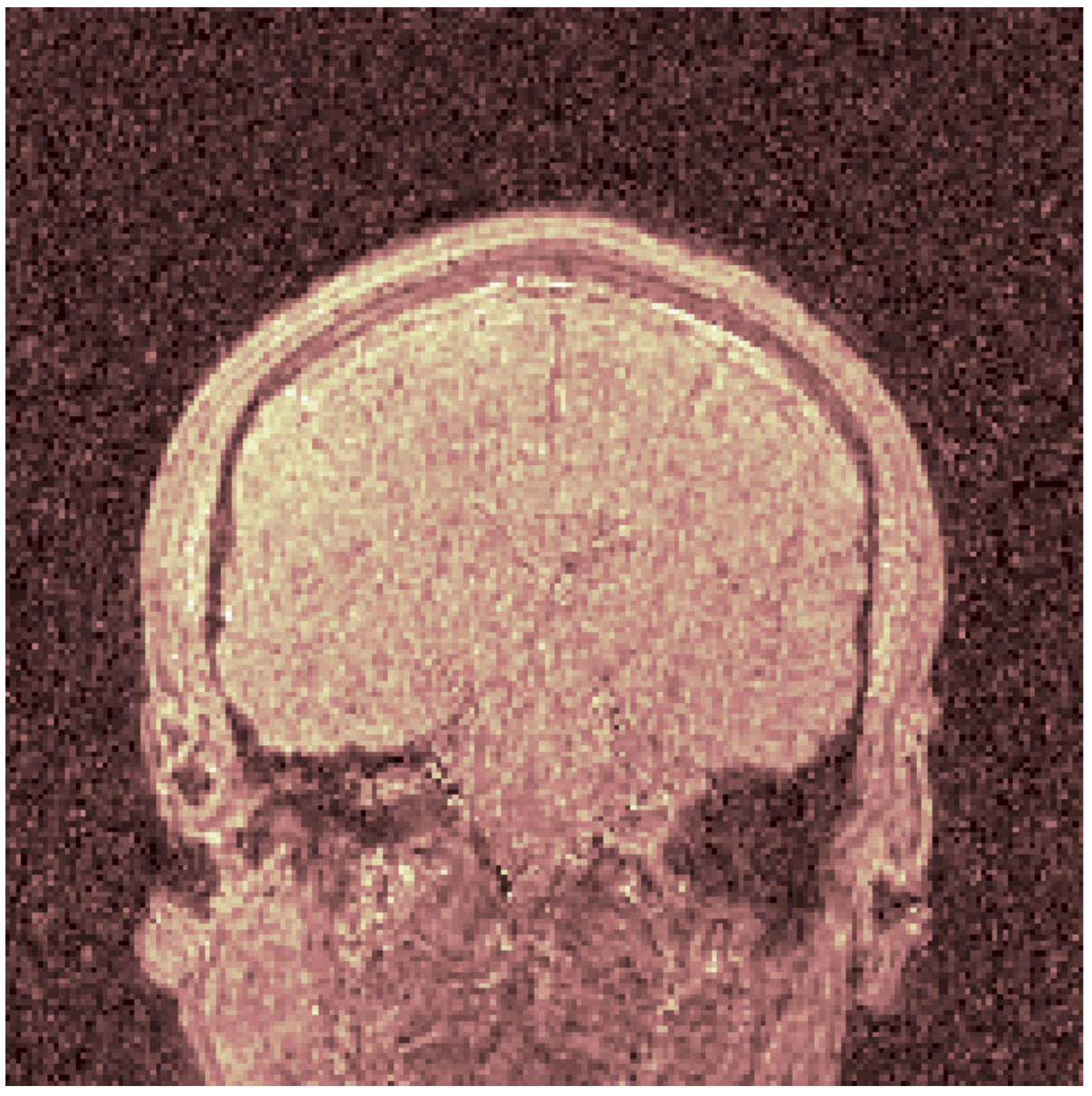}\hspace{-.05cm}
		\\
		\begin{turn}{90} \qquad\quad LR-DM\end{turn}
		\includegraphics[trim= 25 15 25 25, clip, width=.15\linewidth]{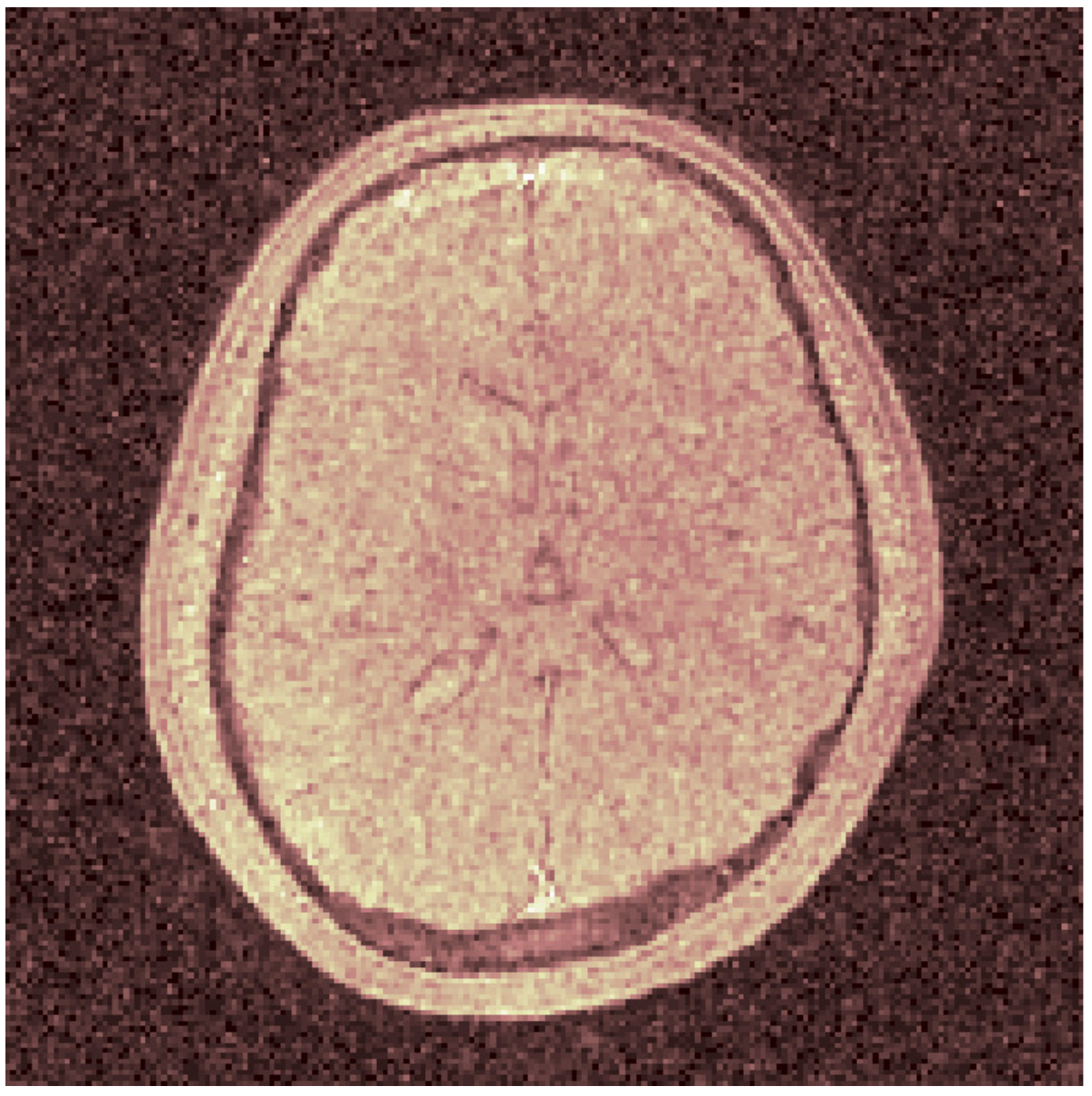}\hspace{-.05cm}
		\includegraphics[trim= 25 15 25 25, clip, width=.15\linewidth]{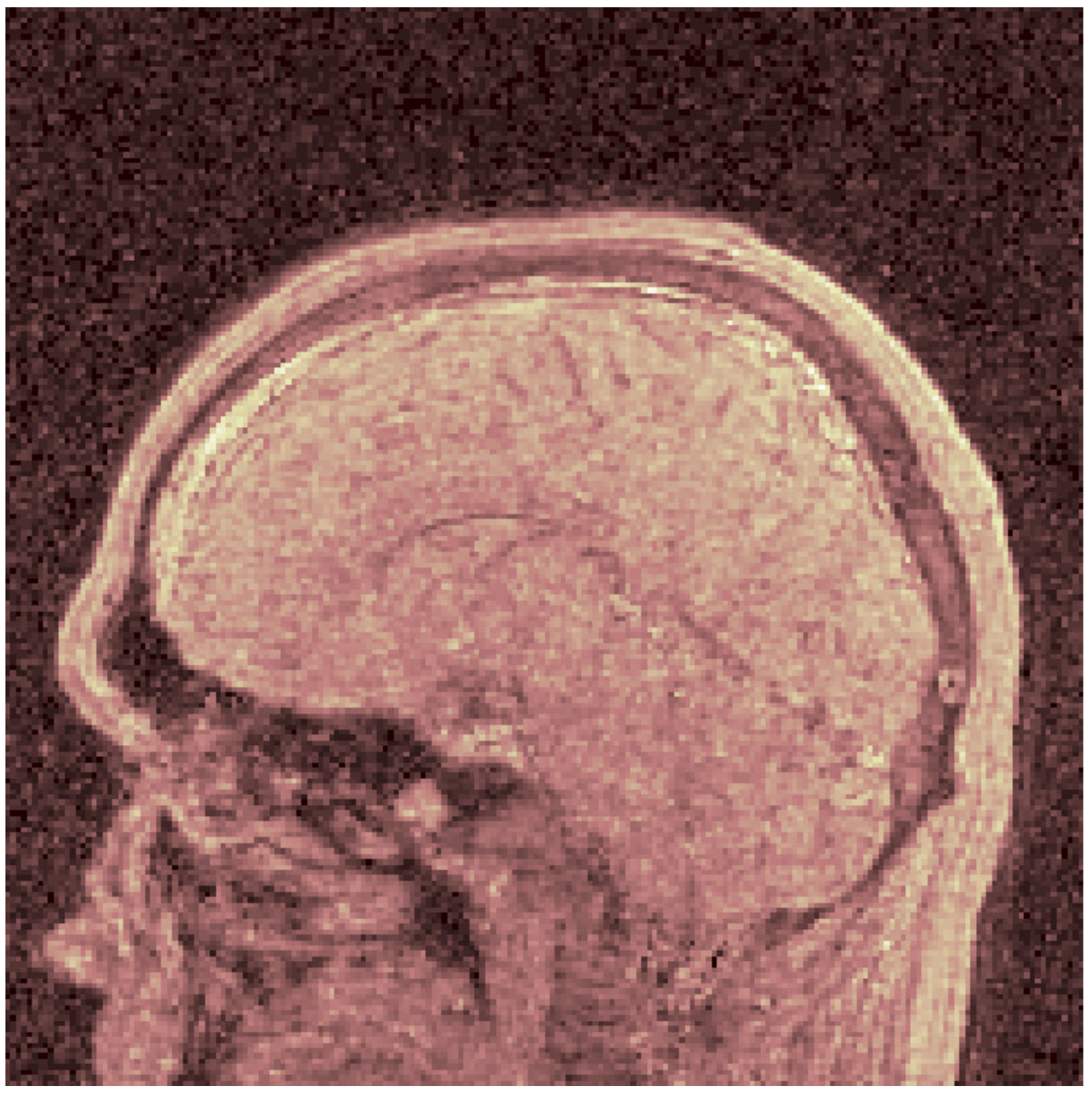}\hspace{-.05cm}
		\includegraphics[trim= 25 15 25 25, clip, width=.15\linewidth]{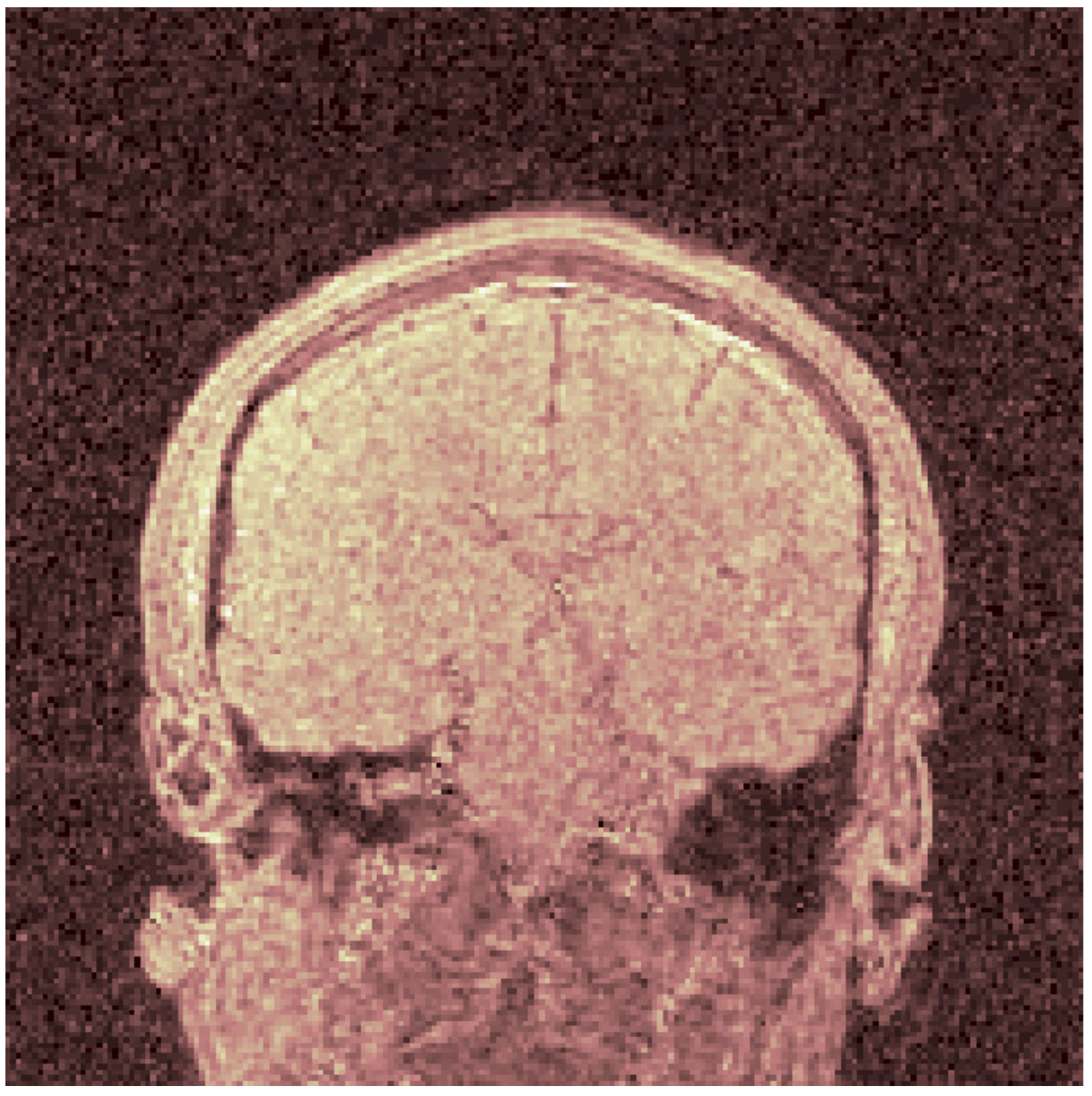}\hspace{-.05cm}
		\\
		\begin{turn}{90} \qquad\quad AIR-MRF\end{turn}
		\includegraphics[trim= 25 15 25 25, clip, width=.15\linewidth]{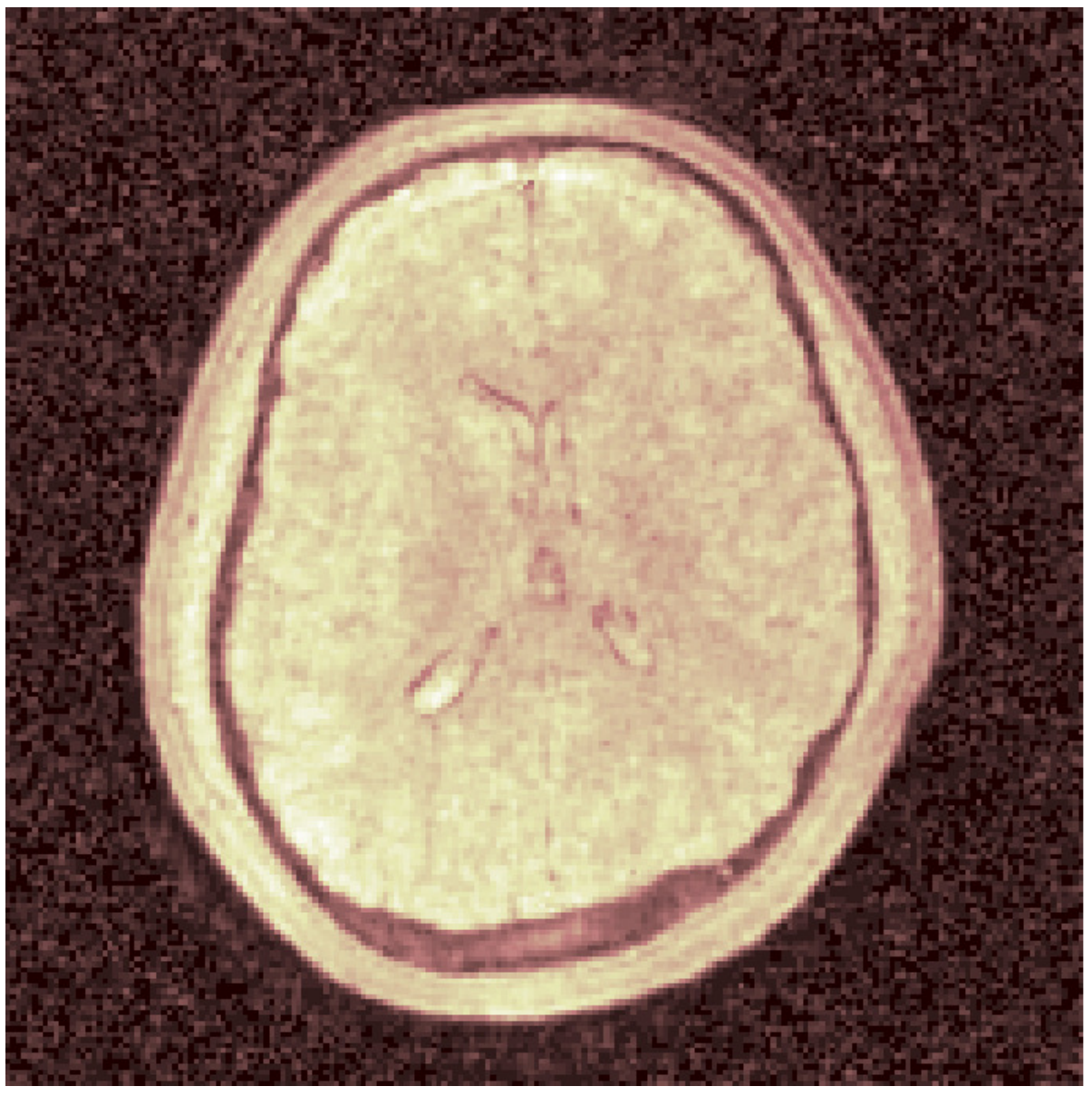}\hspace{-.05cm}
		\includegraphics[trim= 25 15 25 25, clip, width=.15\linewidth]{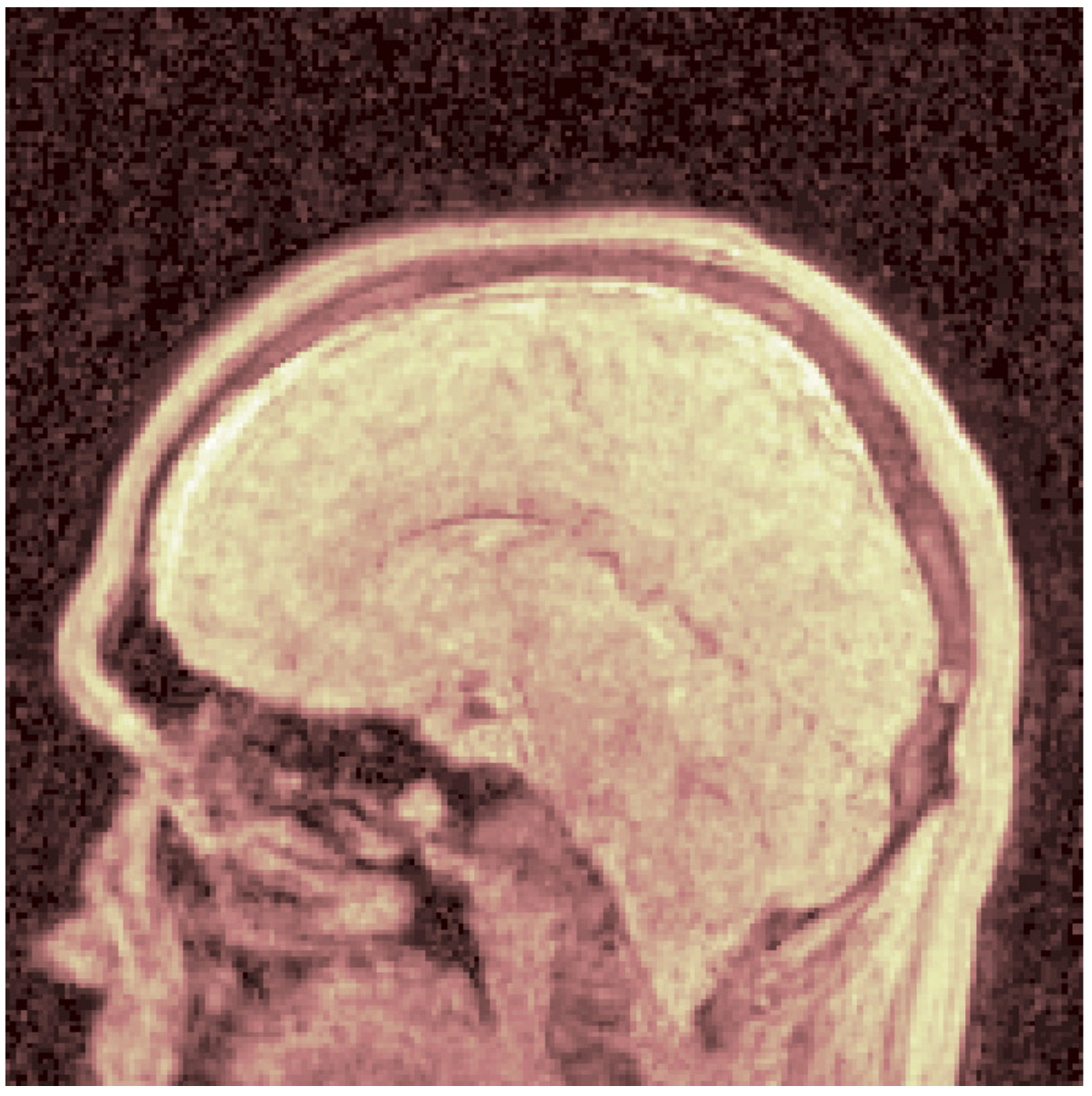}\hspace{-.05cm}
		\includegraphics[trim= 25 15 25 25, clip, width=.15\linewidth]{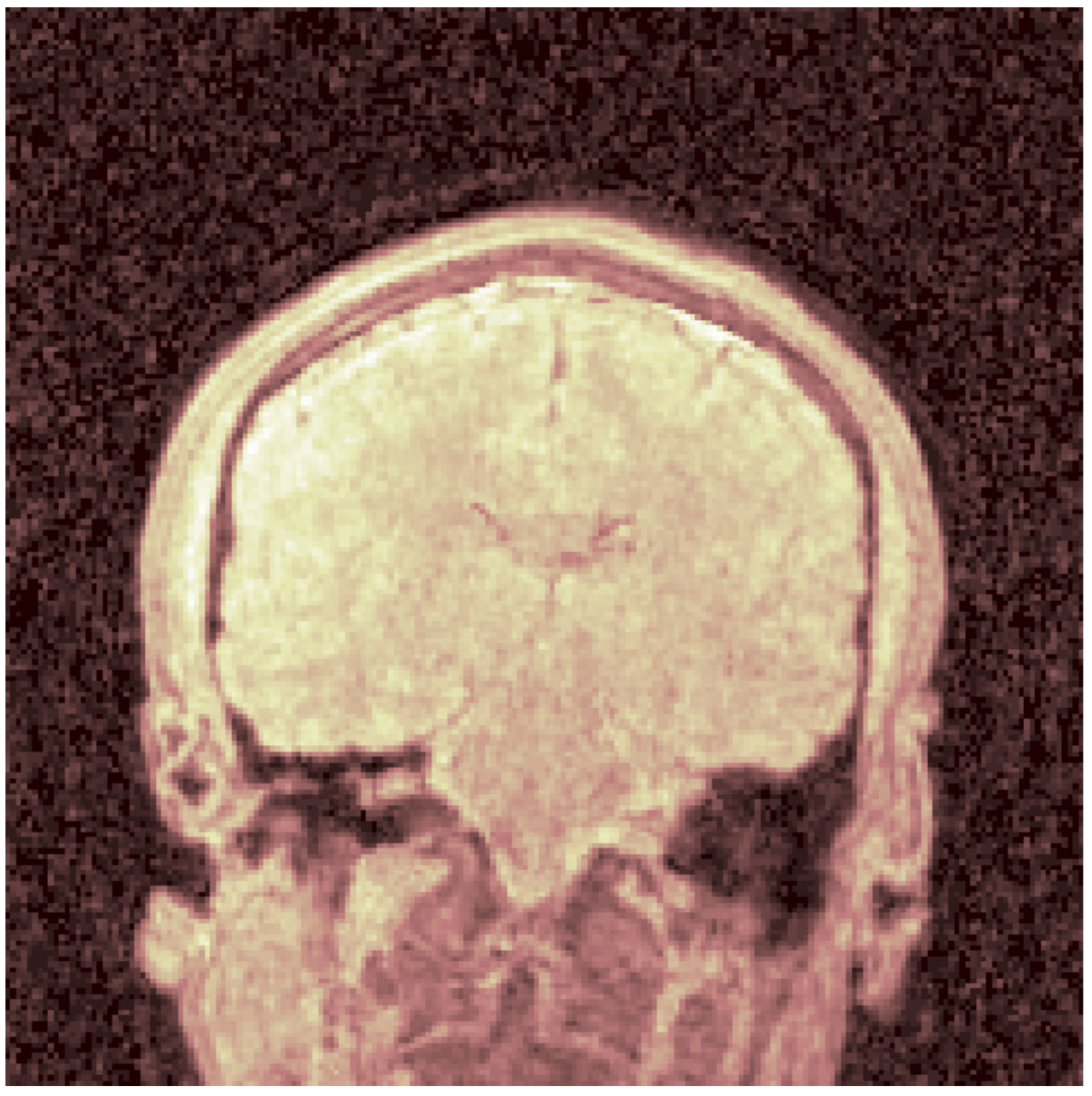}\vspace{-.3cm}
		\\
		\includegraphics[trim= -10 50 -20 720, clip,width=.25\linewidth]{./figs/PDbarvivo.jpg}\vspace{.6cm}
		\\
		PD (a.u.) 
		 \\
		\caption{Reconstructed T1, T2  and PD maps using a 3D scan with spiral readouts. The (zoomed) 3D maps are computed using ZF-DM, LR-DM and AIR-MRF baselines.}  \label{fig:3dvivo-bis}
	\end{minipage}}
\end{figure*}

To supplement our comparisons in section~VI-E (Figure~7) 
regarding the 3D quantitative brain imaging scans,  we display the parametric maps (Figure~\ref{fig:3dvivo-bis})  computed by the ZF-DM, LR-DM and AIR-MRF baselines.  
The ZF and LR predictions are suffering from strong undersampling artefacts due to lacking spatial-domain regularisation. Thanks to adding spatial regularisation, AIR-MRF to some extend reduces these artefacts, however low-pass filtering introduces an undesirable over-smoothing (blur) in the computed quantitative maps that are non-competitive with the LRTV results in Figure~8. 
\end{document}